\documentclass{article}

\IfFileExists{microtype.sty}{\usepackage{microtype}}{}
\usepackage{graphicx}
\usepackage{booktabs}
\usepackage{array}
\usepackage{xcolor}

\usepackage{hyperref}

\usepackage{preprint}
\hypersetup{pdfsubject={},pdftitle={Looping Deeper Without Growing the KV Cache: Recurrent Depth for Causal Encoder-Decoders}}

\usepackage{amsmath}
\usepackage{amssymb}
\usepackage{xspace}
\usepackage{xcolor}
\usepackage{url}

\usepackage{tikz}
\usetikzlibrary{positioning,fit,calc,decorations.pathreplacing,arrows.meta}

\makeatletter
\newcommand{\defval}[2]{\expandafter\def\csname pv@#1\endcsname{#2}}
\newcommand{\defpending}[3]{%
  \expandafter\def\csname pv@#1\endcsname{\pendingbox{#2}{#3}}%
  \expandafter\def\csname ps@#1\endcsname{\pendingshort}%
  \expandafter\def\csname pr@#1\endcsname{#2}}
\newcommand{\pendingbox}[2]{\textcolor{red}{\textbf{[pending: #2; run \pendingruns{#1}]}}}
\newcommand{\pendingshort}{\textcolor{red}{\textbf{TBD}}}
\newcommand{\pendingruns}[1]{%
  \def\pr@sep{}\@for\pr@run:=#1\do{\pr@sep\expandafter\path\expandafter{\pr@run}\def\pr@sep{, }}}
\newcommand{\val}[1]{\ifcsname pv@#1\endcsname\csname pv@#1\endcsname
  \else\textcolor{red}{\textbf{??#1??}}\PackageWarning{paper}{Undefined value #1}\fi}
\newcommand{\sval}[1]{\ifcsname ps@#1\endcsname\csname ps@#1\endcsname\else\val{#1}\fi}
\newcommand{\waitfor}[1]{%
  \begingroup
  \def\wf@list{}%
  \@for\wf@key:=#1\do{%
    \ifcsname pr@\wf@key\endcsname
      \edef\wf@runs{\csname pr@\wf@key\endcsname}%
      \@for\wf@run:=\wf@runs\do{%
        \ifcsname wfseen@\wf@run\endcsname\else
          \expandafter\def\csname wfseen@\wf@run\endcsname{}%
          \ifx\wf@list\@empty\edef\wf@list{\wf@run}\else\edef\wf@list{\wf@list,\wf@run}\fi
        \fi}%
    \fi}%
  \ifx\wf@list\@empty\else
    \textcolor{red}{\textbf{Pending (TBD):} \expandafter\pendingruns\expandafter{\wf@list}.}%
  \fi
  \endgroup}
\makeatother

\newcommand{\makecell}[1]{\begin{tabular}[c]{@{}c@{}}#1\end{tabular}}

\newcommand{\ced}{CED\xspace}
\newcommand{\ourmethod}{Looped CED\xspace}
\newcommand{\vfour}{DeepSeek-V4.1\xspace}

\newcommand{\Lr}[2]{L#1--#2}

\newcommand{\safegraphic}[2][]{\IfFileExists{#2}{\includegraphics[#1]{#2}}{%
  \fbox{\parbox[c][3cm][c]{0.92\columnwidth}{\centering\textcolor{red}{\textbf{[figure not generated yet:
  \path{#2}; run scripts/make\_figures.py]}}}}}\ignorespaces}

\makeatletter
\let\paper@makecaption\@makecaption
\long\def\@makecaption#1#2{\def\@tempa{figure}\ifx\@captype\@tempa\vskip -8pt\fi\paper@makecaption{#1}{#2}}
\newcommand{\captype}[1]{\def\@captype{#1}}
\makeatother

\defval{bpb/long/1024/ced-dec}{0.8295}
\defval{bpb/long/1024/ced-enc}{0.8124}
\defval{bpb/long/1024/ced-noloop}{0.8401}
\defval{bpb/long/1024/dense-l47}{0.8178}
\defval{bpb/long/1024/dense-noloop}{0.8406}
\defval{bpb/p140/768/p140-dec-fixed}{0.8937}
\defval{bpb/p140/768/p140-enc}{0.8844}
\defval{bpb/p140/768/p140-enc-fixed}{0.8844}
\defval{bpb/p140/768/p140-noloop}{0.9162}
\defval{bpb/x40/1024/ced-dec}{0.8648}
\defval{bpb/x40/1024/ced-enc}{0.8468}
\defval{bpb/x40/1024/ced-enc-fixed}{0.8487}
\defval{bpb/x40/1024/ced-noloop}{0.8716}
\defval{bpb/x40/1024/dense-l25}{0.8461}
\defval{bpb/x40/1024/dense-l47}{0.8514}
\defval{bpb/x40/1024/dense-noloop}{0.8747}
\defval{bpb/x40/1024/yocou-loop}{0.8506}
\defval{bpb/x40/1024/yocou-noloop}{0.8745}
\defval{bpb/x40/1536/ced-dec}{0.8141}
\defval{bpb/x40/1536/ced-enc}{0.7974}
\defval{bpb/x40/1536/ced-noloop}{0.8250}
\defval{bpb/x40/1536/dense-l25}{0.7971}
\defval{bpb/x40/1536/dense-noloop}{0.8267}
\defval{bpb/x40/768/ced-dec}{0.9109}
\defval{bpb/x40/768/ced-enc}{0.8924}
\defval{bpb/x40/768/ced-noloop}{0.9143}
\defval{bpb/x40/768/dense-l25}{0.8923}
\defval{bpb/x40/768/dense-l47}{0.8981}
\defval{bpb/x40/768/dense-noloop}{0.9171}
\defval{bpb/x40/768/yocou-loop}{0.8958}
\defval{bpb/x40/768/yocou-noloop}{0.9171}
\defval{bpb512/long/1024/ced-dec}{0.8297}
\defval{bpb512/long/1024/ced-enc}{0.8124}
\defval{bpb512/long/1024/ced-noloop}{0.8405}
\defval{bpb512/long/1024/dense-l47}{0.8178}
\defval{bpb512/long/1024/dense-noloop}{0.8406}
\defval{bpb512/x40/1024/ced-dec}{0.8648}
\defval{bpb512/x40/1024/ced-enc}{0.8465}
\defval{bpb512/x40/1024/ced-enc-fixed}{0.8484}
\defval{bpb512/x40/1024/ced-noloop}{0.8716}
\defval{bpb512/x40/1024/dense-l25}{0.8457}
\defval{bpb512/x40/1024/dense-l47}{0.8511}
\defval{bpb512/x40/1024/dense-noloop}{0.8746}
\defval{bpb512/x40/1024/yocou-loop}{0.8503}
\defval{bpb512/x40/1024/yocou-noloop}{0.8743}
\defval{bpb512/x40/1536/ced-dec}{0.8138}
\defval{bpb512/x40/1536/ced-enc}{0.7966}
\defval{bpb512/x40/1536/ced-noloop}{0.8247}
\defval{bpb512/x40/1536/dense-l25}{0.7963}
\defval{bpb512/x40/1536/dense-noloop}{0.8265}
\defval{bpb512/x40/768/ced-dec}{0.9108}
\defval{bpb512/x40/768/ced-enc}{0.8922}
\defval{bpb512/x40/768/ced-noloop}{0.9142}
\defval{bpb512/x40/768/dense-l25}{0.8919}
\defval{bpb512/x40/768/dense-l47}{0.8976}
\defval{bpb512/x40/768/dense-noloop}{0.9163}
\defval{bpb512/x40/768/yocou-loop}{0.8953}
\defval{bpb512/x40/768/yocou-noloop}{0.9172}
\defval{bpbtrain/long/1024/ced-dec}{0.8278}
\defval{bpbtrain/long/1024/ced-enc}{0.8106}
\defval{bpbtrain/long/1024/ced-noloop}{0.8386}
\defval{bpbtrain/long/1024/dense-l47}{0.8157}
\defval{bpbtrain/long/1024/dense-noloop}{0.8389}
\defval{bpbtrain/p140/768/p140-dec-fixed}{0.8905}
\defval{bpbtrain/p140/768/p140-enc}{0.8807}
\defval{bpbtrain/p140/768/p140-enc-fixed}{0.8815}
\defval{bpbtrain/p140/768/p140-noloop}{0.9132}
\defval{bpbtrain/x40/1024/ced-dec}{0.8648}
\defval{bpbtrain/x40/1024/ced-enc}{0.8465}
\defval{bpbtrain/x40/1024/ced-enc-fixed}{0.8484}
\defval{bpbtrain/x40/1024/ced-noloop}{0.8716}
\defval{bpbtrain/x40/1024/dense-l25}{0.8457}
\defval{bpbtrain/x40/1024/dense-l47}{0.8511}
\defval{bpbtrain/x40/1024/dense-noloop}{0.8746}
\defval{bpbtrain/x40/1024/yocou-loop}{0.8503}
\defval{bpbtrain/x40/1024/yocou-noloop}{0.8743}
\defval{bpbtrain/x40/1536/ced-dec}{0.8138}
\defval{bpbtrain/x40/1536/ced-enc}{0.7966}
\defval{bpbtrain/x40/1536/ced-noloop}{0.8247}
\defval{bpbtrain/x40/1536/dense-l25}{0.7963}
\defval{bpbtrain/x40/1536/dense-noloop}{0.8265}
\defval{bpbtrain/x40/768/ced-dec}{0.9108}
\defval{bpbtrain/x40/768/ced-enc}{0.8922}
\defval{bpbtrain/x40/768/ced-noloop}{0.9142}
\defval{bpbtrain/x40/768/dense-l25}{0.8919}
\defval{bpbtrain/x40/768/dense-l47}{0.8976}
\defval{bpbtrain/x40/768/dense-noloop}{0.9163}
\defval{bpbtrain/x40/768/yocou-loop}{0.8953}
\defval{bpbtrain/x40/768/yocou-noloop}{0.9172}
\defval{cedcommon/long/1024/ced-dec}{44.86}
\defval{cedcommon/long/1024/ced-enc}{45.75}
\defval{cedcommon/long/1024/ced-noloop}{44.23}
\defval{cedcommon/long/1024/dense-l47}{45.42}
\defval{cedcommon/long/1024/dense-noloop}{43.26}
\defval{cedcommon/p140/768/p140-dec-fixed}{41.92}
\defval{cedcommon/p140/768/p140-enc}{41.75}
\defval{cedcommon/p140/768/p140-enc-fixed}{42.11}
\defval{cedcommon/p140/768/p140-noloop}{40.00}
\defval{cedcommon/x40/1024/ced-dec}{42.43}
\defval{cedcommon/x40/1024/ced-enc}{43.47}
\defval{cedcommon/x40/1024/ced-enc-fixed}{43.76}
\defval{cedcommon/x40/1024/ced-noloop}{41.82}
\defval{cedcommon/x40/1024/dense-l25}{44.14}
\defval{cedcommon/x40/1024/dense-l47}{43.16}
\defval{cedcommon/x40/1024/dense-noloop}{41.99}
\defval{cedcommon/x40/1024/yocou-loop}{43.24}
\defval{cedcommon/x40/1024/yocou-noloop}{42.31}
\defval{cedcommon/x40/1536/ced-dec}{45.82}
\defval{cedcommon/x40/1536/ced-enc}{46.75}
\defval{cedcommon/x40/1536/ced-noloop}{44.34}
\defval{cedcommon/x40/1536/dense-l25}{46.09}
\defval{cedcommon/x40/1536/dense-noloop}{44.87}
\defval{cedcommon/x40/768/ced-dec}{40.65}
\defval{cedcommon/x40/768/ced-enc}{41.64}
\defval{cedcommon/x40/768/ced-noloop}{40.83}
\defval{cedcommon/x40/768/dense-l25}{41.91}
\defval{cedcommon/x40/768/dense-l47}{40.71}
\defval{cedcommon/x40/768/dense-noloop}{40.13}
\defval{cedcommon/x40/768/yocou-loop}{41.55}
\defval{cedcommon/x40/768/yocou-noloop}{39.72}
\defval{cfg/batch}{256}
\defval{cfg/blockparams/1024}{151.0}
\defval{cfg/blockparams/1536}{339.7}
\defval{cfg/blockparams/768}{84.9}
\defval{cfg/bootreps}{1000}
\defval{cfg/bwd}{4}
\defval{cfg/decay}{20}
\defval{cfg/decayinit}{0.2}
\defval{cfg/evalwin/long}{128}
\defval{cfg/evalwin/x40}{512}
\defval{cfg/ffnmult}{4}
\defval{cfg/fp/rounds}{32}
\defval{cfg/fp/windows/fw}{256}
\defval{cfg/fp/windows/pg19}{128}
\defval{cfg/fwtok}{5.8}
\defval{cfg/fwwin}{2829}
\defval{cfg/headdim}{128}
\defval{cfg/heads/1024}{8}
\defval{cfg/heads/1536}{12}
\defval{cfg/heads/768}{6}
\defval{cfg/kvfull/windows}{2829}
\defval{cfg/layers}{12}
\defval{cfg/long-over-x40}{4.9}
\defval{cfg/lr}{0.0025}
\defval{cfg/muonwd}{0.3}
\defval{cfg/poswin}{512}
\defval{cfg/ratio/x40/1024}{40}
\defval{cfg/ratio/x40/1536}{40}
\defval{cfg/ratio/x40/768}{40}
\defval{cfg/recall/windows}{1024}
\defval{cfg/rope}{50,000}
\defval{cfg/rounds}{8}
\defval{cfg/seq}{2048}
\defval{cfg/stateinit}{like}
\defval{cfg/steps/x40/1024}{11,444}
\defval{cfg/steps/x40/1536}{25,916}
\defval{cfg/steps/x40/768}{6,480}
\defval{cfg/tokens/long}{29.6}
\defval{cfg/tokens/x40/1024}{6.0}
\defval{cfg/tokens/x40/1536}{13.6}
\defval{cfg/tokens/x40/768}{3.4}
\defval{cfg/ve}{1, 3, 5, 7, 9, 11}
\defval{cfg/vocab}{151,936}
\defval{cfg/warmup}{5}
\defval{cfg/wd}{0.1}
\defval{cfg/window}{128}
\defval{core/long/1024/ced-dec}{21.56}
\defval{core/long/1024/ced-enc}{21.63}
\defval{core/long/1024/ced-noloop}{19.09}
\defval{core/long/1024/dense-l47}{22.60}
\defval{core/long/1024/dense-noloop}{18.50}
\defval{core/p140/768/p140-dec-fixed}{16.02}
\defval{core/p140/768/p140-enc}{17.30}
\defval{core/p140/768/p140-enc-fixed}{17.89}
\defval{core/p140/768/p140-noloop}{12.56}
\defval{core/x40/1024/ced-dec}{17.09}
\defval{core/x40/1024/ced-enc}{18.91}
\defval{core/x40/1024/ced-enc-fixed}{19.74}
\defval{core/x40/1024/ced-noloop}{17.06}
\defval{core/x40/1024/dense-l25}{20.15}
\defval{core/x40/1024/dense-l47}{18.81}
\defval{core/x40/1024/dense-noloop}{17.07}
\defval{core/x40/1024/yocou-loop}{18.99}
\defval{core/x40/1024/yocou-noloop}{16.12}
\defval{core/x40/1536/ced-dec}{20.64}
\defval{core/x40/1536/ced-enc}{24.04}
\defval{core/x40/1536/ced-noloop}{21.05}
\defval{core/x40/1536/dense-l25}{24.61}
\defval{core/x40/1536/dense-noloop}{21.59}
\defval{core/x40/768/ced-dec}{13.19}
\defval{core/x40/768/ced-enc}{15.18}
\defval{core/x40/768/ced-noloop}{13.83}
\defval{core/x40/768/dense-l25}{16.51}
\defval{core/x40/768/dense-l47}{14.43}
\defval{core/x40/768/dense-noloop}{13.47}
\defval{core/x40/768/yocou-loop}{15.69}
\defval{core/x40/768/yocou-noloop}{14.08}
\defval{corediff/long/1024/ced-dec-minus-ced-enc}{$-$0.1}
\defval{corediff/long/1024/ced-dec-minus-ced-enc/hi}{+1.2}
\defval{corediff/long/1024/ced-dec-minus-ced-enc/lo}{$-$1.3}
\defval{corediff/long/1024/ced-dec-minus-ced-enc/sig}{no}
\defval{corediff/long/1024/ced-enc-minus-ced-noloop}{+2.5}
\defval{corediff/long/1024/ced-enc-minus-ced-noloop/hi}{+3.6}
\defval{corediff/long/1024/ced-enc-minus-ced-noloop/lo}{+1.4}
\defval{corediff/long/1024/ced-enc-minus-ced-noloop/sig}{yes}
\defval{corediff/long/1024/ced-enc-minus-dense-l47}{$-$1.0}
\defval{corediff/long/1024/ced-enc-minus-dense-l47/hi}{+0.0}
\defval{corediff/long/1024/ced-enc-minus-dense-l47/lo}{$-$2.0}
\defval{corediff/long/1024/ced-enc-minus-dense-l47/sig}{no}
\defval{corediff/long/1024/ced-noloop-minus-dense-noloop}{+0.6}
\defval{corediff/long/1024/ced-noloop-minus-dense-noloop/hi}{+1.6}
\defval{corediff/long/1024/ced-noloop-minus-dense-noloop/lo}{$-$0.4}
\defval{corediff/long/1024/ced-noloop-minus-dense-noloop/sig}{no}
\defval{corediff/long/1024/dense-l47-minus-dense-noloop}{+4.1}
\defval{corediff/long/1024/dense-l47-minus-dense-noloop/hi}{+5.1}
\defval{corediff/long/1024/dense-l47-minus-dense-noloop/lo}{+3.1}
\defval{corediff/long/1024/dense-l47-minus-dense-noloop/sig}{yes}
\defval{corediff/p140/768/p140-enc-minus-p140-enc-fixed}{$-$0.6}
\defval{corediff/p140/768/p140-enc-minus-p140-enc-fixed/hi}{+0.4}
\defval{corediff/p140/768/p140-enc-minus-p140-enc-fixed/lo}{$-$1.5}
\defval{corediff/p140/768/p140-enc-minus-p140-enc-fixed/sig}{no}
\defval{corediff/p140/768/p140-enc-minus-p140-noloop}{+4.8}
\defval{corediff/p140/768/p140-enc-minus-p140-noloop/hi}{+5.8}
\defval{corediff/p140/768/p140-enc-minus-p140-noloop/lo}{+3.8}
\defval{corediff/p140/768/p140-enc-minus-p140-noloop/sig}{yes}
\defval{corediff/x40/1024/ced-dec-minus-ced-enc}{$-$1.8}
\defval{corediff/x40/1024/ced-dec-minus-ced-enc/hi}{$-$0.8}
\defval{corediff/x40/1024/ced-dec-minus-ced-enc/lo}{$-$2.9}
\defval{corediff/x40/1024/ced-dec-minus-ced-enc/sig}{yes}
\defval{corediff/x40/1024/ced-enc-minus-ced-enc-fixed}{$-$0.8}
\defval{corediff/x40/1024/ced-enc-minus-ced-enc-fixed/hi}{+0.3}
\defval{corediff/x40/1024/ced-enc-minus-ced-enc-fixed/lo}{$-$1.9}
\defval{corediff/x40/1024/ced-enc-minus-ced-enc-fixed/sig}{no}
\defval{corediff/x40/1024/ced-enc-minus-ced-noloop}{+1.8}
\defval{corediff/x40/1024/ced-enc-minus-ced-noloop/hi}{+2.9}
\defval{corediff/x40/1024/ced-enc-minus-ced-noloop/lo}{+0.8}
\defval{corediff/x40/1024/ced-enc-minus-ced-noloop/sig}{yes}
\defval{corediff/x40/1024/ced-enc-minus-dense-l25}{$-$1.2}
\defval{corediff/x40/1024/ced-enc-minus-dense-l25/hi}{$-$0.2}
\defval{corediff/x40/1024/ced-enc-minus-dense-l25/lo}{$-$2.3}
\defval{corediff/x40/1024/ced-enc-minus-dense-l25/sig}{yes}
\defval{corediff/x40/1024/ced-enc-minus-dense-l47}{+0.1}
\defval{corediff/x40/1024/ced-enc-minus-dense-l47/hi}{+1.2}
\defval{corediff/x40/1024/ced-enc-minus-dense-l47/lo}{$-$1.0}
\defval{corediff/x40/1024/ced-enc-minus-dense-l47/sig}{no}
\defval{corediff/x40/1024/ced-noloop-minus-dense-noloop}{$-$0.0}
\defval{corediff/x40/1024/ced-noloop-minus-dense-noloop/hi}{+1.0}
\defval{corediff/x40/1024/ced-noloop-minus-dense-noloop/lo}{$-$1.1}
\defval{corediff/x40/1024/ced-noloop-minus-dense-noloop/sig}{no}
\defval{corediff/x40/1024/dense-l25-minus-dense-noloop}{+3.1}
\defval{corediff/x40/1024/dense-l25-minus-dense-noloop/hi}{+4.2}
\defval{corediff/x40/1024/dense-l25-minus-dense-noloop/lo}{+1.9}
\defval{corediff/x40/1024/dense-l25-minus-dense-noloop/sig}{yes}
\defval{corediff/x40/1024/dense-l47-minus-dense-l25}{$-$1.3}
\defval{corediff/x40/1024/dense-l47-minus-dense-l25/hi}{$-$0.2}
\defval{corediff/x40/1024/dense-l47-minus-dense-l25/lo}{$-$2.4}
\defval{corediff/x40/1024/dense-l47-minus-dense-l25/sig}{yes}
\defval{corediff/x40/1024/dense-l47-minus-dense-noloop}{+1.7}
\defval{corediff/x40/1024/dense-l47-minus-dense-noloop/hi}{+2.8}
\defval{corediff/x40/1024/dense-l47-minus-dense-noloop/lo}{+0.6}
\defval{corediff/x40/1024/dense-l47-minus-dense-noloop/sig}{yes}
\defval{corediff/x40/1024/yocou-loop-minus-yocou-noloop}{+2.9}
\defval{corediff/x40/1024/yocou-loop-minus-yocou-noloop/hi}{+3.9}
\defval{corediff/x40/1024/yocou-loop-minus-yocou-noloop/lo}{+1.8}
\defval{corediff/x40/1024/yocou-loop-minus-yocou-noloop/sig}{yes}
\defval{corediff/x40/1536/ced-dec-minus-ced-enc}{$-$3.4}
\defval{corediff/x40/1536/ced-dec-minus-ced-enc/hi}{$-$2.4}
\defval{corediff/x40/1536/ced-dec-minus-ced-enc/lo}{$-$4.5}
\defval{corediff/x40/1536/ced-dec-minus-ced-enc/sig}{yes}
\defval{corediff/x40/1536/ced-enc-minus-ced-noloop}{+3.0}
\defval{corediff/x40/1536/ced-enc-minus-ced-noloop/hi}{+4.2}
\defval{corediff/x40/1536/ced-enc-minus-ced-noloop/lo}{+1.9}
\defval{corediff/x40/1536/ced-enc-minus-ced-noloop/sig}{yes}
\defval{corediff/x40/1536/ced-enc-minus-dense-l25}{$-$0.6}
\defval{corediff/x40/1536/ced-enc-minus-dense-l25/hi}{+0.4}
\defval{corediff/x40/1536/ced-enc-minus-dense-l25/lo}{$-$1.6}
\defval{corediff/x40/1536/ced-enc-minus-dense-l25/sig}{no}
\defval{corediff/x40/1536/ced-noloop-minus-dense-noloop}{$-$0.5}
\defval{corediff/x40/1536/ced-noloop-minus-dense-noloop/hi}{+0.5}
\defval{corediff/x40/1536/ced-noloop-minus-dense-noloop/lo}{$-$1.6}
\defval{corediff/x40/1536/ced-noloop-minus-dense-noloop/sig}{no}
\defval{corediff/x40/1536/dense-l25-minus-dense-noloop}{+3.0}
\defval{corediff/x40/1536/dense-l25-minus-dense-noloop/hi}{+4.2}
\defval{corediff/x40/1536/dense-l25-minus-dense-noloop/lo}{+1.9}
\defval{corediff/x40/1536/dense-l25-minus-dense-noloop/sig}{yes}
\defval{corediff/x40/768/ced-dec-minus-ced-enc}{$-$2.0}
\defval{corediff/x40/768/ced-dec-minus-ced-enc/hi}{$-$0.9}
\defval{corediff/x40/768/ced-dec-minus-ced-enc/lo}{$-$3.0}
\defval{corediff/x40/768/ced-dec-minus-ced-enc/sig}{yes}
\defval{corediff/x40/768/ced-enc-minus-ced-noloop}{+1.4}
\defval{corediff/x40/768/ced-enc-minus-ced-noloop/hi}{+2.4}
\defval{corediff/x40/768/ced-enc-minus-ced-noloop/lo}{+0.3}
\defval{corediff/x40/768/ced-enc-minus-ced-noloop/sig}{yes}
\defval{corediff/x40/768/ced-enc-minus-dense-l25}{$-$1.3}
\defval{corediff/x40/768/ced-enc-minus-dense-l25/hi}{$-$0.3}
\defval{corediff/x40/768/ced-enc-minus-dense-l25/lo}{$-$2.4}
\defval{corediff/x40/768/ced-enc-minus-dense-l25/sig}{yes}
\defval{corediff/x40/768/ced-enc-minus-dense-l47}{+0.8}
\defval{corediff/x40/768/ced-enc-minus-dense-l47/hi}{+1.8}
\defval{corediff/x40/768/ced-enc-minus-dense-l47/lo}{$-$0.5}
\defval{corediff/x40/768/ced-enc-minus-dense-l47/sig}{no}
\defval{corediff/x40/768/ced-noloop-minus-dense-noloop}{+0.4}
\defval{corediff/x40/768/ced-noloop-minus-dense-noloop/hi}{+1.4}
\defval{corediff/x40/768/ced-noloop-minus-dense-noloop/lo}{$-$0.7}
\defval{corediff/x40/768/ced-noloop-minus-dense-noloop/sig}{no}
\defval{corediff/x40/768/dense-l25-minus-dense-noloop}{+3.0}
\defval{corediff/x40/768/dense-l25-minus-dense-noloop/hi}{+4.3}
\defval{corediff/x40/768/dense-l25-minus-dense-noloop/lo}{+1.9}
\defval{corediff/x40/768/dense-l25-minus-dense-noloop/sig}{yes}
\defval{corediff/x40/768/dense-l47-minus-dense-l25}{$-$2.1}
\defval{corediff/x40/768/dense-l47-minus-dense-l25/hi}{$-$0.9}
\defval{corediff/x40/768/dense-l47-minus-dense-l25/lo}{$-$3.2}
\defval{corediff/x40/768/dense-l47-minus-dense-l25/sig}{yes}
\defval{corediff/x40/768/dense-l47-minus-dense-noloop}{+1.0}
\defval{corediff/x40/768/dense-l47-minus-dense-noloop/hi}{+2.1}
\defval{corediff/x40/768/dense-l47-minus-dense-noloop/lo}{$-$0.2}
\defval{corediff/x40/768/dense-l47-minus-dense-noloop/sig}{no}
\defval{corediff/x40/768/yocou-loop-minus-yocou-noloop}{+1.6}
\defval{corediff/x40/768/yocou-loop-minus-yocou-noloop/hi}{+2.7}
\defval{corediff/x40/768/yocou-loop-minus-yocou-noloop/lo}{+0.5}
\defval{corediff/x40/768/yocou-loop-minus-yocou-noloop/sig}{yes}
\defval{coreext/long/1024/ced-dec}{13.65}
\defval{coreext/long/1024/ced-enc}{13.54}
\defval{coreext/long/1024/ced-noloop}{12.32}
\defval{coreext/long/1024/dense-l47}{14.84}
\defval{coreext/long/1024/dense-noloop}{11.57}
\defval{coreext/p140/768/p140-dec-fixed}{11.00}
\defval{coreext/p140/768/p140-enc}{10.87}
\defval{coreext/p140/768/p140-enc-fixed}{11.82}
\defval{coreext/p140/768/p140-noloop}{8.26}
\defval{coreext/x40/1024/ced-dec}{10.90}
\defval{coreext/x40/1024/ced-enc}{13.15}
\defval{coreext/x40/1024/ced-enc-fixed}{13.47}
\defval{coreext/x40/1024/ced-noloop}{11.51}
\defval{coreext/x40/1024/dense-l25}{13.11}
\defval{coreext/x40/1024/dense-l47}{12.72}
\defval{coreext/x40/1024/dense-noloop}{11.47}
\defval{coreext/x40/1024/yocou-loop}{11.60}
\defval{coreext/x40/1024/yocou-noloop}{10.16}
\defval{coreext/x40/1536/ced-dec}{12.26}
\defval{coreext/x40/1536/ced-enc}{14.88}
\defval{coreext/x40/1536/ced-noloop}{13.13}
\defval{coreext/x40/1536/dense-l25}{14.54}
\defval{coreext/x40/1536/dense-noloop}{13.21}
\defval{coreext/x40/768/ced-dec}{8.90}
\defval{coreext/x40/768/ced-enc}{10.36}
\defval{coreext/x40/768/ced-noloop}{9.44}
\defval{coreext/x40/768/dense-l25}{11.43}
\defval{coreext/x40/768/dense-l47}{9.62}
\defval{coreext/x40/768/dense-noloop}{9.47}
\defval{coreext/x40/768/yocou-loop}{10.12}
\defval{coreext/x40/768/yocou-noloop}{9.29}
\defval{coreextdiff/long/1024/ced-dec-minus-ced-enc}{+0.1}
\defval{coreextdiff/long/1024/ced-dec-minus-ced-enc/hi}{+1.3}
\defval{coreextdiff/long/1024/ced-dec-minus-ced-enc/lo}{$-$1.2}
\defval{coreextdiff/long/1024/ced-dec-minus-ced-enc/sig}{no}
\defval{coreextdiff/long/1024/ced-enc-minus-ced-noloop}{+1.2}
\defval{coreextdiff/long/1024/ced-enc-minus-ced-noloop/hi}{+2.5}
\defval{coreextdiff/long/1024/ced-enc-minus-ced-noloop/lo}{+0.0}
\defval{coreextdiff/long/1024/ced-enc-minus-ced-noloop/sig}{yes}
\defval{coreextdiff/long/1024/ced-enc-minus-dense-l47}{$-$1.3}
\defval{coreextdiff/long/1024/ced-enc-minus-dense-l47/hi}{$-$0.1}
\defval{coreextdiff/long/1024/ced-enc-minus-dense-l47/lo}{$-$2.4}
\defval{coreextdiff/long/1024/ced-enc-minus-dense-l47/sig}{yes}
\defval{coreextdiff/long/1024/ced-noloop-minus-dense-noloop}{+0.8}
\defval{coreextdiff/long/1024/ced-noloop-minus-dense-noloop/hi}{+1.9}
\defval{coreextdiff/long/1024/ced-noloop-minus-dense-noloop/lo}{$-$0.3}
\defval{coreextdiff/long/1024/ced-noloop-minus-dense-noloop/sig}{no}
\defval{coreextdiff/long/1024/dense-l47-minus-dense-noloop}{+3.3}
\defval{coreextdiff/long/1024/dense-l47-minus-dense-noloop/hi}{+4.5}
\defval{coreextdiff/long/1024/dense-l47-minus-dense-noloop/lo}{+2.1}
\defval{coreextdiff/long/1024/dense-l47-minus-dense-noloop/sig}{yes}
\defval{coreextdiff/p140/768/p140-enc-minus-p140-enc-fixed}{$-$1.0}
\defval{coreextdiff/p140/768/p140-enc-minus-p140-enc-fixed/hi}{+0.2}
\defval{coreextdiff/p140/768/p140-enc-minus-p140-enc-fixed/lo}{$-$2.1}
\defval{coreextdiff/p140/768/p140-enc-minus-p140-enc-fixed/sig}{no}
\defval{coreextdiff/p140/768/p140-enc-minus-p140-noloop}{+2.6}
\defval{coreextdiff/p140/768/p140-enc-minus-p140-noloop/hi}{+4.0}
\defval{coreextdiff/p140/768/p140-enc-minus-p140-noloop/lo}{+1.2}
\defval{coreextdiff/p140/768/p140-enc-minus-p140-noloop/sig}{yes}
\defval{coreextdiff/x40/1024/ced-dec-minus-ced-enc}{$-$2.2}
\defval{coreextdiff/x40/1024/ced-dec-minus-ced-enc/hi}{$-$0.8}
\defval{coreextdiff/x40/1024/ced-dec-minus-ced-enc/lo}{$-$3.5}
\defval{coreextdiff/x40/1024/ced-dec-minus-ced-enc/sig}{yes}
\defval{coreextdiff/x40/1024/ced-enc-minus-ced-enc-fixed}{$-$0.3}
\defval{coreextdiff/x40/1024/ced-enc-minus-ced-enc-fixed/hi}{+0.9}
\defval{coreextdiff/x40/1024/ced-enc-minus-ced-enc-fixed/lo}{$-$1.6}
\defval{coreextdiff/x40/1024/ced-enc-minus-ced-enc-fixed/sig}{no}
\defval{coreextdiff/x40/1024/ced-enc-minus-ced-noloop}{+1.6}
\defval{coreextdiff/x40/1024/ced-enc-minus-ced-noloop/hi}{+2.9}
\defval{coreextdiff/x40/1024/ced-enc-minus-ced-noloop/lo}{+0.3}
\defval{coreextdiff/x40/1024/ced-enc-minus-ced-noloop/sig}{yes}
\defval{coreextdiff/x40/1024/ced-enc-minus-dense-l25}{+0.0}
\defval{coreextdiff/x40/1024/ced-enc-minus-dense-l25/hi}{+1.2}
\defval{coreextdiff/x40/1024/ced-enc-minus-dense-l25/lo}{$-$1.3}
\defval{coreextdiff/x40/1024/ced-enc-minus-dense-l25/sig}{no}
\defval{coreextdiff/x40/1024/ced-enc-minus-dense-l47}{+0.4}
\defval{coreextdiff/x40/1024/ced-enc-minus-dense-l47/hi}{+1.7}
\defval{coreextdiff/x40/1024/ced-enc-minus-dense-l47/lo}{$-$0.8}
\defval{coreextdiff/x40/1024/ced-enc-minus-dense-l47/sig}{no}
\defval{coreextdiff/x40/1024/ced-noloop-minus-dense-noloop}{+0.0}
\defval{coreextdiff/x40/1024/ced-noloop-minus-dense-noloop/hi}{+1.2}
\defval{coreextdiff/x40/1024/ced-noloop-minus-dense-noloop/lo}{$-$1.0}
\defval{coreextdiff/x40/1024/ced-noloop-minus-dense-noloop/sig}{no}
\defval{coreextdiff/x40/1024/dense-l25-minus-dense-noloop}{+1.7}
\defval{coreextdiff/x40/1024/dense-l25-minus-dense-noloop/hi}{+2.9}
\defval{coreextdiff/x40/1024/dense-l25-minus-dense-noloop/lo}{+0.4}
\defval{coreextdiff/x40/1024/dense-l25-minus-dense-noloop/sig}{yes}
\defval{coreextdiff/x40/1024/dense-l47-minus-dense-l25}{$-$0.4}
\defval{coreextdiff/x40/1024/dense-l47-minus-dense-l25/hi}{+0.8}
\defval{coreextdiff/x40/1024/dense-l47-minus-dense-l25/lo}{$-$1.6}
\defval{coreextdiff/x40/1024/dense-l47-minus-dense-l25/sig}{no}
\defval{coreextdiff/x40/1024/dense-l47-minus-dense-noloop}{+1.2}
\defval{coreextdiff/x40/1024/dense-l47-minus-dense-noloop/hi}{+2.4}
\defval{coreextdiff/x40/1024/dense-l47-minus-dense-noloop/lo}{+0.1}
\defval{coreextdiff/x40/1024/dense-l47-minus-dense-noloop/sig}{yes}
\defval{coreextdiff/x40/1024/yocou-loop-minus-yocou-noloop}{+1.5}
\defval{coreextdiff/x40/1024/yocou-loop-minus-yocou-noloop/hi}{+2.6}
\defval{coreextdiff/x40/1024/yocou-loop-minus-yocou-noloop/lo}{+0.3}
\defval{coreextdiff/x40/1024/yocou-loop-minus-yocou-noloop/sig}{yes}
\defval{coreextdiff/x40/1536/ced-dec-minus-ced-enc}{$-$2.6}
\defval{coreextdiff/x40/1536/ced-dec-minus-ced-enc/hi}{$-$1.5}
\defval{coreextdiff/x40/1536/ced-dec-minus-ced-enc/lo}{$-$3.9}
\defval{coreextdiff/x40/1536/ced-dec-minus-ced-enc/sig}{yes}
\defval{coreextdiff/x40/1536/ced-enc-minus-ced-noloop}{+1.8}
\defval{coreextdiff/x40/1536/ced-enc-minus-ced-noloop/hi}{+3.0}
\defval{coreextdiff/x40/1536/ced-enc-minus-ced-noloop/lo}{+0.5}
\defval{coreextdiff/x40/1536/ced-enc-minus-ced-noloop/sig}{yes}
\defval{coreextdiff/x40/1536/ced-enc-minus-dense-l25}{+0.3}
\defval{coreextdiff/x40/1536/ced-enc-minus-dense-l25/hi}{+1.6}
\defval{coreextdiff/x40/1536/ced-enc-minus-dense-l25/lo}{$-$1.0}
\defval{coreextdiff/x40/1536/ced-enc-minus-dense-l25/sig}{no}
\defval{coreextdiff/x40/1536/ced-noloop-minus-dense-noloop}{$-$0.1}
\defval{coreextdiff/x40/1536/ced-noloop-minus-dense-noloop/hi}{+1.1}
\defval{coreextdiff/x40/1536/ced-noloop-minus-dense-noloop/lo}{$-$1.3}
\defval{coreextdiff/x40/1536/ced-noloop-minus-dense-noloop/sig}{no}
\defval{coreextdiff/x40/1536/dense-l25-minus-dense-noloop}{+1.3}
\defval{coreextdiff/x40/1536/dense-l25-minus-dense-noloop/hi}{+2.6}
\defval{coreextdiff/x40/1536/dense-l25-minus-dense-noloop/lo}{+0.1}
\defval{coreextdiff/x40/1536/dense-l25-minus-dense-noloop/sig}{yes}
\defval{coreextdiff/x40/768/ced-dec-minus-ced-enc}{$-$1.5}
\defval{coreextdiff/x40/768/ced-dec-minus-ced-enc/hi}{$-$0.4}
\defval{coreextdiff/x40/768/ced-dec-minus-ced-enc/lo}{$-$2.6}
\defval{coreextdiff/x40/768/ced-dec-minus-ced-enc/sig}{yes}
\defval{coreextdiff/x40/768/ced-enc-minus-ced-noloop}{+0.9}
\defval{coreextdiff/x40/768/ced-enc-minus-ced-noloop/hi}{+2.2}
\defval{coreextdiff/x40/768/ced-enc-minus-ced-noloop/lo}{$-$0.3}
\defval{coreextdiff/x40/768/ced-enc-minus-ced-noloop/sig}{no}
\defval{coreextdiff/x40/768/ced-enc-minus-dense-l25}{$-$1.1}
\defval{coreextdiff/x40/768/ced-enc-minus-dense-l25/hi}{+0.1}
\defval{coreextdiff/x40/768/ced-enc-minus-dense-l25/lo}{$-$2.1}
\defval{coreextdiff/x40/768/ced-enc-minus-dense-l25/sig}{no}
\defval{coreextdiff/x40/768/ced-enc-minus-dense-l47}{+0.8}
\defval{coreextdiff/x40/768/ced-enc-minus-dense-l47/hi}{+1.9}
\defval{coreextdiff/x40/768/ced-enc-minus-dense-l47/lo}{$-$0.4}
\defval{coreextdiff/x40/768/ced-enc-minus-dense-l47/sig}{no}
\defval{coreextdiff/x40/768/ced-noloop-minus-dense-noloop}{$-$0.0}
\defval{coreextdiff/x40/768/ced-noloop-minus-dense-noloop/hi}{+1.1}
\defval{coreextdiff/x40/768/ced-noloop-minus-dense-noloop/lo}{$-$1.1}
\defval{coreextdiff/x40/768/ced-noloop-minus-dense-noloop/sig}{no}
\defval{coreextdiff/x40/768/dense-l25-minus-dense-noloop}{+2.0}
\defval{coreextdiff/x40/768/dense-l25-minus-dense-noloop/hi}{+3.2}
\defval{coreextdiff/x40/768/dense-l25-minus-dense-noloop/lo}{+0.6}
\defval{coreextdiff/x40/768/dense-l25-minus-dense-noloop/sig}{yes}
\defval{coreextdiff/x40/768/dense-l47-minus-dense-l25}{$-$1.8}
\defval{coreextdiff/x40/768/dense-l47-minus-dense-l25/hi}{$-$0.7}
\defval{coreextdiff/x40/768/dense-l47-minus-dense-l25/lo}{$-$2.9}
\defval{coreextdiff/x40/768/dense-l47-minus-dense-l25/sig}{yes}
\defval{coreextdiff/x40/768/dense-l47-minus-dense-noloop}{+0.1}
\defval{coreextdiff/x40/768/dense-l47-minus-dense-noloop/hi}{+1.3}
\defval{coreextdiff/x40/768/dense-l47-minus-dense-noloop/lo}{$-$1.0}
\defval{coreextdiff/x40/768/dense-l47-minus-dense-noloop/sig}{no}
\defval{coreextdiff/x40/768/yocou-loop-minus-yocou-noloop}{+0.8}
\defval{coreextdiff/x40/768/yocou-loop-minus-yocou-noloop/hi}{+2.0}
\defval{coreextdiff/x40/768/yocou-loop-minus-yocou-noloop/lo}{$-$0.3}
\defval{coreextdiff/x40/768/yocou-loop-minus-yocou-noloop/sig}{no}
\defval{coreextgain/lit/parcae}{+1.04}
\defval{coreextgain/long/1024/ced-enc}{+1.22}
\defval{coregain/lit/parcae}{+2.54}
\defval{coregain/long/1024/ced-enc}{+2.54}
\defval{cost/batch}{128}
\defval{cost/c-dec-us}{17.30}
\defval{cost/c-pre-us}{5.40}
\defval{cost/coresize}{18}
\defval{cost/ctx}{65,536}
\defval{cost/dec-over-pre}{3.2}
\defval{cost/dec/core}{21--38}
\defval{cost/dec/decode}{3.85}
\defval{cost/dec/decodepasses}{166}
\defval{cost/dec/exact-span}{18,543}
\defval{cost/dec/kv}{890}
\defval{cost/dec/kvx}{1.0}
\defval{cost/dec/prefill}{1.01}
\defval{cost/dec/prefill-exact}{2.03}
\defval{cost/dec/prefillpasses}{20}
\defval{cost/dec/tailpasses}{146}
\defval{cost/enc-p4/decode}{3.99}
\defval{cost/enc-p4/decodepasses}{166}
\defval{cost/enc-p4/kv}{890}
\defval{cost/enc-p4/kvx}{1.0}
\defval{cost/enc-p4/prefill}{3.73}
\defval{cost/enc-p4/prefillpasses}{74}
\defval{cost/enc-round/decode}{3.99}
\defval{cost/enc-round/decodepasses}{166}
\defval{cost/enc-round/kv}{4628}
\defval{cost/enc-round/kvx}{5.2}
\defval{cost/enc-round/prefill}{7.38}
\defval{cost/enc-round/prefillpasses}{146}
\defval{cost/enc/core}{2--19}
\defval{cost/enc/decode}{3.99}
\defval{cost/enc/decodepasses}{166}
\defval{cost/enc/kv}{890}
\defval{cost/enc/kvx}{1.0}
\defval{cost/enc/prefill}{7.38}
\defval{cost/enc/prefill-exact}{7.25}
\defval{cost/enc/prefillpasses}{146}
\defval{cost/encsources}{3}
\defval{cost/ep}{64}
\defval{cost/experts}{384}
\defval{cost/hit-frac}{5}
\defval{cost/kvratio/enc-round-over-enc}{5.2}
\defval{cost/noloop/decode}{1.00}
\defval{cost/noloop/decodepasses}{40}
\defval{cost/noloop/kv}{890}
\defval{cost/noloop/kvx}{1.0}
\defval{cost/noloop/prefill}{1.00}
\defval{cost/noloop/prefillpasses}{20}
\defval{cost/prompt}{65,536}
\defval{cost/ratio-at-0}{1.04}
\defval{cost/ratio-at-1}{7.29}
\defval{cost/rounds}{8}
\defval{cost/shared}{1}
\defval{cost/topk}{6}
\defval{cost/window}{128}
\defval{data/beyond/128}{88}
\defval{data/beyond/2048}{23}
\defval{data/beyond/4096}{13}
\defval{data/beyond/762}{47}
\defval{data/beyond/8192}{6}
\defval{data/doc-mean}{1,032}
\defval{data/doc-median}{617}
\defval{diff/long/1024/dense-l47-minus-ced-enc}{0.0054}
\defval{diff/x40/1024/ced-enc-minus-dense-l25}{+0.0008}
\defval{diff/x40/1024/ced-noloop-minus-dense-noloop}{$-$0.0031}
\defval{diff/x40/1024/dense-l47-minus-dense-l25}{+0.0053}
\defval{diff/x40/1024/enc-gain-minus-dec-gain}{0.0180}
\defval{diff/x40/1024/yocou-loop-minus-ced-enc}{+0.0037}
\defval{diff/x40/1024/yocou-noloop-minus-ced-noloop}{+0.0029}
\defval{diff/x40/1536/ced-enc-minus-dense-l25}{+0.0003}
\defval{diff/x40/1536/ced-noloop-minus-dense-noloop}{$-$0.0017}
\defval{diff/x40/768/ced-enc-minus-dense-l25}{+0.0001}
\defval{diff/x40/768/ced-noloop-minus-dense-noloop}{$-$0.0028}
\defval{diff/x40/768/dense-l47-minus-dense-l25}{+0.0057}
\defval{diff128/long/1024/dense-l47-minus-ced-enc}{0.0051}
\defval{fp/long/1024/ced-enc/fw/bpb/r16}{0.8071}
\defval{fp/long/1024/ced-enc/fw/bpb/r32}{0.8071}
\defval{fp/long/1024/ced-enc/fw/bpb/r4}{0.8156}
\defval{fp/long/1024/ced-enc/fw/bpb/r8}{0.8073}
\defval{fp/long/1024/ced-enc/fw/bpbspread/late}{0.0003}
\defval{fp/long/1024/ced-enc/fw/contract}{0.61}
\defval{fp/long/1024/ced-enc/fw/kvstep/r16}{0.51}
\defval{fp/long/1024/ced-enc/fw/kvstep/r32}{0.4}
\defval{fp/long/1024/ced-enc/fw/kvstep/r4}{23}
\defval{fp/long/1024/ced-enc/fw/kvstep/r8}{3.5}
\defval{fp/long/1024/ced-enc/fw/rho/r16}{0.088}
\defval{fp/long/1024/ced-enc/fw/rho/r32}{0.0071}
\defval{fp/long/1024/ced-enc/fw/rho/r4}{13}
\defval{fp/long/1024/ced-enc/fw/rho/r8}{1.8}
\defval{fp/long/1024/ced-enc/fw/rhomin/late}{0.0071}
\defval{fp/long/1024/ced-enc/pg19/bpb/r16}{1.2785}
\defval{fp/long/1024/ced-enc/pg19/bpb/r32}{1.2786}
\defval{fp/long/1024/ced-enc/pg19/bpb/r4}{1.2958}
\defval{fp/long/1024/ced-enc/pg19/bpb/r8}{1.2790}
\defval{fp/long/1024/ced-enc/pg19/bpbspread/late}{0.0005}
\defval{fp/long/1024/ced-enc/pg19/contract}{0.63}
\defval{fp/long/1024/ced-enc/pg19/kvstep/r16}{0.68}
\defval{fp/long/1024/ced-enc/pg19/kvstep/r32}{0.49}
\defval{fp/long/1024/ced-enc/pg19/kvstep/r4}{24}
\defval{fp/long/1024/ced-enc/pg19/kvstep/r8}{4.4}
\defval{fp/long/1024/ced-enc/pg19/rho/r16}{0.16}
\defval{fp/long/1024/ced-enc/pg19/rho/r32}{0.028}
\defval{fp/long/1024/ced-enc/pg19/rho/r4}{14}
\defval{fp/long/1024/ced-enc/pg19/rho/r8}{2.2}
\defval{fp/long/1024/ced-enc/pg19/rhomin/late}{0.028}
\defval{fp/p140/768/p140-enc-fixed/fw/bpb/r16}{0.8770}
\defval{fp/p140/768/p140-enc-fixed/fw/bpb/r32}{0.8775}
\defval{fp/p140/768/p140-enc-fixed/fw/bpb/r4}{1.0699}
\defval{fp/p140/768/p140-enc-fixed/fw/bpb/r8}{0.8779}
\defval{fp/p140/768/p140-enc-fixed/fw/bpbbad/max}{0.13}
\defval{fp/p140/768/p140-enc-fixed/fw/bpbbad/min}{0.07}
\defval{fp/p140/768/p140-enc-fixed/fw/bpbspread/late}{0.1292}
\defval{fp/p140/768/p140-enc-fixed/fw/contract}{1.00}
\defval{fp/p140/768/p140-enc-fixed/fw/kvstep/r16}{27}
\defval{fp/p140/768/p140-enc-fixed/fw/kvstep/r32}{27}
\defval{fp/p140/768/p140-enc-fixed/fw/kvstep/r4}{48}
\defval{fp/p140/768/p140-enc-fixed/fw/kvstep/r8}{28}
\defval{fp/p140/768/p140-enc-fixed/fw/period}{4}
\defval{fp/p140/768/p140-enc-fixed/fw/rho/r16}{67}
\defval{fp/p140/768/p140-enc-fixed/fw/rho/r32}{67}
\defval{fp/p140/768/p140-enc-fixed/fw/rho/r4}{68}
\defval{fp/p140/768/p140-enc-fixed/fw/rho/r8}{68}
\defval{fp/p140/768/p140-enc-fixed/fw/rhomin/late}{38}
\defval{fp/p140/768/p140-enc-fixed/pg19/bpb/r16}{1.3911}
\defval{fp/p140/768/p140-enc-fixed/pg19/bpb/r32}{1.3913}
\defval{fp/p140/768/p140-enc-fixed/pg19/bpb/r4}{1.6389}
\defval{fp/p140/768/p140-enc-fixed/pg19/bpb/r8}{1.3929}
\defval{fp/p140/768/p140-enc-fixed/pg19/bpbbad/max}{0.20}
\defval{fp/p140/768/p140-enc-fixed/pg19/bpbbad/min}{0.12}
\defval{fp/p140/768/p140-enc-fixed/pg19/bpbspread/late}{0.1991}
\defval{fp/p140/768/p140-enc-fixed/pg19/contract}{1.00}
\defval{fp/p140/768/p140-enc-fixed/pg19/kvstep/r16}{28}
\defval{fp/p140/768/p140-enc-fixed/pg19/kvstep/r32}{28}
\defval{fp/p140/768/p140-enc-fixed/pg19/kvstep/r4}{51}
\defval{fp/p140/768/p140-enc-fixed/pg19/kvstep/r8}{29}
\defval{fp/p140/768/p140-enc-fixed/pg19/period}{4}
\defval{fp/p140/768/p140-enc-fixed/pg19/rho/r16}{70}
\defval{fp/p140/768/p140-enc-fixed/pg19/rho/r32}{70}
\defval{fp/p140/768/p140-enc-fixed/pg19/rho/r4}{70}
\defval{fp/p140/768/p140-enc-fixed/pg19/rho/r8}{70}
\defval{fp/p140/768/p140-enc-fixed/pg19/rhomin/late}{39}
\defval{fp/p140/768/p140-enc/fw/bpb/r16}{0.8776}
\defval{fp/p140/768/p140-enc/fw/bpb/r32}{0.8776}
\defval{fp/p140/768/p140-enc/fw/bpb/r4}{0.8884}
\defval{fp/p140/768/p140-enc/fw/bpb/r8}{0.8779}
\defval{fp/p140/768/p140-enc/fw/bpbspread/late}{0.0003}
\defval{fp/p140/768/p140-enc/fw/contract}{0.62}
\defval{fp/p140/768/p140-enc/fw/kvstep/r16}{0.43}
\defval{fp/p140/768/p140-enc/fw/kvstep/r32}{0.36}
\defval{fp/p140/768/p140-enc/fw/kvstep/r4}{19}
\defval{fp/p140/768/p140-enc/fw/kvstep/r8}{3}
\defval{fp/p140/768/p140-enc/fw/rho/r16}{0.12}
\defval{fp/p140/768/p140-enc/fw/rho/r32}{0.0075}
\defval{fp/p140/768/p140-enc/fw/rho/r4}{19}
\defval{fp/p140/768/p140-enc/fw/rho/r8}{2.7}
\defval{fp/p140/768/p140-enc/fw/rhomin/late}{0.0075}
\defval{fp/p140/768/p140-enc/pg19/bpb/r16}{1.3892}
\defval{fp/p140/768/p140-enc/pg19/bpb/r32}{1.3892}
\defval{fp/p140/768/p140-enc/pg19/bpb/r4}{1.4080}
\defval{fp/p140/768/p140-enc/pg19/bpb/r8}{1.3909}
\defval{fp/p140/768/p140-enc/pg19/bpbspread/late}{0.0017}
\defval{fp/p140/768/p140-enc/pg19/contract}{0.64}
\defval{fp/p140/768/p140-enc/pg19/kvstep/r16}{2.2}
\defval{fp/p140/768/p140-enc/pg19/kvstep/r32}{2.3}
\defval{fp/p140/768/p140-enc/pg19/kvstep/r4}{19}
\defval{fp/p140/768/p140-enc/pg19/kvstep/r8}{4.1}
\defval{fp/p140/768/p140-enc/pg19/rho/r16}{0.45}
\defval{fp/p140/768/p140-enc/pg19/rho/r32}{0.34}
\defval{fp/p140/768/p140-enc/pg19/rho/r4}{20}
\defval{fp/p140/768/p140-enc/pg19/rho/r8}{3.3}
\defval{fp/p140/768/p140-enc/pg19/rhomin/late}{0.34}
\defval{fp/x40/1024/ced-dec/fw/bpb/r16}{0.8590}
\defval{fp/x40/1024/ced-dec/fw/bpb/r32}{0.8590}
\defval{fp/x40/1024/ced-dec/fw/bpb/r4}{0.8609}
\defval{fp/x40/1024/ced-dec/fw/bpb/r8}{0.8590}
\defval{fp/x40/1024/ced-dec/fw/bpbspread/late}{0.0000}
\defval{fp/x40/1024/ced-dec/fw/contract}{0.58}
\defval{fp/x40/1024/ced-dec/fw/rho/r16}{0.033}
\defval{fp/x40/1024/ced-dec/fw/rho/r32}{0.001}
\defval{fp/x40/1024/ced-dec/fw/rho/r4}{5.5}
\defval{fp/x40/1024/ced-dec/fw/rho/r8}{0.62}
\defval{fp/x40/1024/ced-dec/fw/rhomin/late}{0.001}
\defval{fp/x40/1024/ced-dec/pg19/bpb/r16}{1.3607}
\defval{fp/x40/1024/ced-dec/pg19/bpb/r32}{1.3608}
\defval{fp/x40/1024/ced-dec/pg19/bpb/r4}{1.3626}
\defval{fp/x40/1024/ced-dec/pg19/bpb/r8}{1.3607}
\defval{fp/x40/1024/ced-dec/pg19/bpbspread/late}{0.0001}
\defval{fp/x40/1024/ced-dec/pg19/contract}{0.59}
\defval{fp/x40/1024/ced-dec/pg19/rho/r16}{0.038}
\defval{fp/x40/1024/ced-dec/pg19/rho/r32}{0.00084}
\defval{fp/x40/1024/ced-dec/pg19/rho/r4}{5.5}
\defval{fp/x40/1024/ced-dec/pg19/rho/r8}{0.68}
\defval{fp/x40/1024/ced-dec/pg19/rhomin/late}{0.00084}
\defval{fp/x40/1024/ced-enc-fixed/fw/bpb/r16}{0.8421}
\defval{fp/x40/1024/ced-enc-fixed/fw/bpb/r32}{0.8430}
\defval{fp/x40/1024/ced-enc-fixed/fw/bpb/r4}{1.1125}
\defval{fp/x40/1024/ced-enc-fixed/fw/bpb/r8}{0.8428}
\defval{fp/x40/1024/ced-enc-fixed/fw/bpbbad/max}{0.08}
\defval{fp/x40/1024/ced-enc-fixed/fw/bpbbad/min}{0.06}
\defval{fp/x40/1024/ced-enc-fixed/fw/bpbspread/late}{0.0821}
\defval{fp/x40/1024/ced-enc-fixed/fw/contract}{0.98}
\defval{fp/x40/1024/ced-enc-fixed/fw/kvstep/r16}{29}
\defval{fp/x40/1024/ced-enc-fixed/fw/kvstep/r32}{29}
\defval{fp/x40/1024/ced-enc-fixed/fw/kvstep/r4}{61}
\defval{fp/x40/1024/ced-enc-fixed/fw/kvstep/r8}{31}
\defval{fp/x40/1024/ced-enc-fixed/fw/period}{4}
\defval{fp/x40/1024/ced-enc-fixed/fw/rho/r16}{61}
\defval{fp/x40/1024/ced-enc-fixed/fw/rho/r32}{61}
\defval{fp/x40/1024/ced-enc-fixed/fw/rho/r4}{66}
\defval{fp/x40/1024/ced-enc-fixed/fw/rho/r8}{62}
\defval{fp/x40/1024/ced-enc-fixed/fw/rhomin/late}{30}
\defval{fp/x40/1024/ced-enc-fixed/pg19/bpb/r16}{1.3347}
\defval{fp/x40/1024/ced-enc-fixed/pg19/bpb/r32}{1.3390}
\defval{fp/x40/1024/ced-enc-fixed/pg19/bpb/r4}{1.6541}
\defval{fp/x40/1024/ced-enc-fixed/pg19/bpb/r8}{1.3367}
\defval{fp/x40/1024/ced-enc-fixed/pg19/bpbbad/max}{0.13}
\defval{fp/x40/1024/ced-enc-fixed/pg19/bpbbad/min}{0.10}
\defval{fp/x40/1024/ced-enc-fixed/pg19/bpbspread/late}{0.1332}
\defval{fp/x40/1024/ced-enc-fixed/pg19/contract}{0.97}
\defval{fp/x40/1024/ced-enc-fixed/pg19/kvstep/r16}{29}
\defval{fp/x40/1024/ced-enc-fixed/pg19/kvstep/r32}{29}
\defval{fp/x40/1024/ced-enc-fixed/pg19/kvstep/r4}{63}
\defval{fp/x40/1024/ced-enc-fixed/pg19/kvstep/r8}{32}
\defval{fp/x40/1024/ced-enc-fixed/pg19/period}{4}
\defval{fp/x40/1024/ced-enc-fixed/pg19/rho/r16}{62}
\defval{fp/x40/1024/ced-enc-fixed/pg19/rho/r32}{62}
\defval{fp/x40/1024/ced-enc-fixed/pg19/rho/r4}{70}
\defval{fp/x40/1024/ced-enc-fixed/pg19/rho/r8}{62}
\defval{fp/x40/1024/ced-enc-fixed/pg19/rhomin/late}{32}
\defval{fp/x40/1024/ced-enc/fw/bpb/r16}{0.8403}
\defval{fp/x40/1024/ced-enc/fw/bpb/r32}{0.8404}
\defval{fp/x40/1024/ced-enc/fw/bpb/r4}{0.8478}
\defval{fp/x40/1024/ced-enc/fw/bpb/r8}{0.8406}
\defval{fp/x40/1024/ced-enc/fw/bpbspread/late}{0.0003}
\defval{fp/x40/1024/ced-enc/fw/contract}{0.61}
\defval{fp/x40/1024/ced-enc/fw/kvstep/r16}{0.55}
\defval{fp/x40/1024/ced-enc/fw/kvstep/r32}{0.45}
\defval{fp/x40/1024/ced-enc/fw/kvstep/r4}{20}
\defval{fp/x40/1024/ced-enc/fw/kvstep/r8}{3}
\defval{fp/x40/1024/ced-enc/fw/rho/r16}{0.1}
\defval{fp/x40/1024/ced-enc/fw/rho/r32}{0.014}
\defval{fp/x40/1024/ced-enc/fw/rho/r4}{13}
\defval{fp/x40/1024/ced-enc/fw/rho/r8}{1.7}
\defval{fp/x40/1024/ced-enc/fw/rhomin/late}{0.014}
\defval{fp/x40/1024/ced-enc/pg19/bpb/r16}{1.3303}
\defval{fp/x40/1024/ced-enc/pg19/bpb/r32}{1.3303}
\defval{fp/x40/1024/ced-enc/pg19/bpb/r4}{1.3454}
\defval{fp/x40/1024/ced-enc/pg19/bpb/r8}{1.3313}
\defval{fp/x40/1024/ced-enc/pg19/bpbspread/late}{0.0010}
\defval{fp/x40/1024/ced-enc/pg19/contract}{0.63}
\defval{fp/x40/1024/ced-enc/pg19/kvstep/r16}{0.68}
\defval{fp/x40/1024/ced-enc/pg19/kvstep/r32}{0.53}
\defval{fp/x40/1024/ced-enc/pg19/kvstep/r4}{21}
\defval{fp/x40/1024/ced-enc/pg19/kvstep/r8}{3.7}
\defval{fp/x40/1024/ced-enc/pg19/rho/r16}{0.18}
\defval{fp/x40/1024/ced-enc/pg19/rho/r32}{0.033}
\defval{fp/x40/1024/ced-enc/pg19/rho/r4}{14}
\defval{fp/x40/1024/ced-enc/pg19/rho/r8}{2.2}
\defval{fp/x40/1024/ced-enc/pg19/rhomin/late}{0.033}
\defval{frac/long/1024/dec-over-enc}{38}
\defval{frac/long/matched-early/dec-over-enc/max}{37}
\defval{frac/long/matched-early/dec-over-enc/min}{12}
\defval{frac/long/matched-early/from}{3.1}
\defval{frac/long/matched-early/to}{7.9}
\defval{frac/long/matched/dec-over-enc/max}{43}
\defval{frac/long/matched/dec-over-enc/median}{37}
\defval{frac/long/matched/dec-over-enc/min}{31}
\defval{frac/long/matched/from}{10.5}
\defval{frac/long/matched/to}{24.1}
\defval{frac/x40/1024/ced-over-dense}{86}
\defval{frac/x40/1024/dec-over-enc}{27}
\defval{frac/x40/1024/yocou-over-enc}{97}
\defval{frac/x40/1536/ced-over-dense}{93}
\defval{frac/x40/1536/dec-over-enc}{40}
\defval{frac/x40/768/ced-over-dense}{89}
\defval{frac/x40/768/dec-over-enc}{16}
\defval{gain/long/1024/ced-dec}{0.0107}
\defval{gain/long/1024/ced-enc}{0.0277}
\defval{gain/x40/1024/ced-dec}{0.0068}
\defval{gain/x40/1024/ced-enc}{0.0248}
\defval{gain/x40/1024/ced-enc-fixed}{0.0229}
\defval{gain/x40/1024/dense-l25}{0.0287}
\defval{gain/x40/1024/dense-l47}{0.0233}
\defval{gain/x40/1024/yocou-loop}{0.0239}
\defval{gain/x40/1536/ced-dec}{0.0109}
\defval{gain/x40/1536/ced-enc}{0.0276}
\defval{gain/x40/1536/dense-l25}{0.0296}
\defval{gain/x40/768/ced-dec}{0.0034}
\defval{gain/x40/768/ced-enc}{0.0219}
\defval{gain/x40/768/dense-l25}{0.0248}
\defval{gain/x40/768/dense-l47}{0.0191}
\defval{gain/x40/768/yocou-loop}{0.0214}
\defval{gain128/long/1024/ced-dec}{0.0108}
\defval{gain128/long/1024/ced-enc}{0.0280}
\defval{gain128/long/enc-range/from}{5.2}
\defval{gain128/long/enc-range/max}{0.0274}
\defval{gain128/long/enc-range/min}{0.0240}
\defval{gap/long/decay-start}{23.7}
\defval{gap/long/dense-minus-enc/from}{5.2}
\defval{gap/long/dense-minus-enc/max}{0.0052}
\defval{gap/long/dense-minus-enc/min}{0.0023}
\defval{gap6b/ced-enc-minus-dense-l25/from}{2.1}
\defval{gap6b/ced-enc-minus-dense-l25/maxabs}{0.0015}
\defval{kvfull/long/1024/ced-enc/delta}{$-$0.00018}
\defval{kvfull/long/1024/ced-enc/delta/hi}{$-$0.00014}
\defval{kvfull/long/1024/ced-enc/delta/lo}{$-$0.00021}
\defval{kvfull/max}{0.0002}
\defval{kvfull/maxci}{0.0002}
\defval{kvfull/x40/1024/ced-enc/delta}{$-$0.00015}
\defval{kvfull/x40/1024/ced-enc/delta/hi}{$-$0.00012}
\defval{kvfull/x40/1024/ced-enc/delta/lo}{$-$0.00018}
\defval{kvfull/x40/1536/ced-enc/delta}{0.00000}
\defval{kvfull/x40/1536/ced-enc/delta/hi}{+0.00004}
\defval{kvfull/x40/1536/ced-enc/delta/lo}{$-$0.00003}
\defval{kvfull/x40/768/ced-enc/delta}{$-$0.00018}
\defval{kvfull/x40/768/ced-enc/delta/hi}{$-$0.00015}
\defval{kvfull/x40/768/ced-enc/delta/lo}{$-$0.00021}
\defval{kvlast/long/1024/ced-dec}{8}
\defval{kvlast/long/1024/ced-enc}{8}
\defval{kvlast/long/1024/ced-noloop}{8}
\defval{kvlast/long/1024/dense-l47}{48}
\defval{kvlast/long/1024/dense-noloop}{48}
\defval{kvlast/p140/768/p140-dec-fixed}{3}
\defval{kvlast/p140/768/p140-enc}{3}
\defval{kvlast/p140/768/p140-enc-fixed}{3}
\defval{kvlast/p140/768/p140-noloop}{3}
\defval{kvlast/x40/1024/ced-dec}{8}
\defval{kvlast/x40/1024/ced-enc}{8}
\defval{kvlast/x40/1024/ced-enc-fixed}{8}
\defval{kvlast/x40/1024/ced-noloop}{8}
\defval{kvlast/x40/1024/dense-l25}{48}
\defval{kvlast/x40/1024/dense-l47}{48}
\defval{kvlast/x40/1024/dense-noloop}{48}
\defval{kvlast/x40/1024/yocou-loop}{4}
\defval{kvlast/x40/1024/yocou-noloop}{4}
\defval{kvlast/x40/1536/ced-dec}{12}
\defval{kvlast/x40/1536/ced-enc}{12}
\defval{kvlast/x40/1536/ced-noloop}{12}
\defval{kvlast/x40/1536/dense-l25}{72}
\defval{kvlast/x40/1536/dense-noloop}{72}
\defval{kvlast/x40/768/ced-dec}{6}
\defval{kvlast/x40/768/ced-enc}{6}
\defval{kvlast/x40/768/ced-noloop}{6}
\defval{kvlast/x40/768/dense-l25}{36}
\defval{kvlast/x40/768/dense-l47}{36}
\defval{kvlast/x40/768/dense-noloop}{36}
\defval{kvlast/x40/768/yocou-loop}{3}
\defval{kvlast/x40/768/yocou-noloop}{3}
\defval{kvratio/x40/1024/dense-over-ced}{6}
\defval{kvratio/x40/1024/denseround-over-ced}{20}
\defval{kvratio/x40/1536/dense-over-ced}{6}
\defval{kvratio/x40/768/dense-over-ced}{6}
\defval{kvround/long/1024/ced-dec}{8}
\defval{kvround/long/1024/ced-enc}{36}
\defval{kvround/long/1024/ced-noloop}{8}
\defval{kvround/long/1024/dense-l47}{160}
\defval{kvround/long/1024/dense-noloop}{48}
\defval{kvround/p140/768/p140-dec-fixed}{3}
\defval{kvround/p140/768/p140-enc}{24}
\defval{kvround/p140/768/p140-enc-fixed}{24}
\defval{kvround/p140/768/p140-noloop}{3}
\defval{kvround/x40/1024/ced-dec}{8}
\defval{kvround/x40/1024/ced-enc}{36}
\defval{kvround/x40/1024/ced-enc-fixed}{36}
\defval{kvround/x40/1024/ced-noloop}{8}
\defval{kvround/x40/1024/dense-l25}{160}
\defval{kvround/x40/1024/dense-l47}{160}
\defval{kvround/x40/1024/dense-noloop}{48}
\defval{kvround/x40/1024/yocou-loop}{4}
\defval{kvround/x40/1024/yocou-noloop}{4}
\defval{kvround/x40/1536/ced-dec}{12}
\defval{kvround/x40/1536/ced-enc}{54}
\defval{kvround/x40/1536/ced-noloop}{12}
\defval{kvround/x40/1536/dense-l25}{240}
\defval{kvround/x40/1536/dense-noloop}{72}
\defval{kvround/x40/768/ced-dec}{6}
\defval{kvround/x40/768/ced-enc}{27}
\defval{kvround/x40/768/ced-noloop}{6}
\defval{kvround/x40/768/dense-l25}{120}
\defval{kvround/x40/768/dense-l47}{120}
\defval{kvround/x40/768/dense-noloop}{36}
\defval{kvround/x40/768/yocou-loop}{3}
\defval{kvround/x40/768/yocou-noloop}{3}
\defval{kvseg/long/1024/ced-dec/finepdfs-long/1-128/delta}{+0.00004}
\defval{kvseg/long/1024/ced-dec/finepdfs-long/1-128/hi}{+0.00032}
\defval{kvseg/long/1024/ced-dec/finepdfs-long/1-128/last}{1.3791}
\defval{kvseg/long/1024/ced-dec/finepdfs-long/1-128/lo}{$-$0.00022}
\defval{kvseg/long/1024/ced-dec/finepdfs-long/1-128/round}{1.3791}
\defval{kvseg/long/1024/ced-dec/finepdfs-long/128-256/delta}{+0.00009}
\defval{kvseg/long/1024/ced-dec/finepdfs-long/128-256/hi}{+0.00026}
\defval{kvseg/long/1024/ced-dec/finepdfs-long/128-256/last}{1.1446}
\defval{kvseg/long/1024/ced-dec/finepdfs-long/128-256/lo}{$-$0.00009}
\defval{kvseg/long/1024/ced-dec/finepdfs-long/128-256/round}{1.1445}
\defval{kvseg/long/1024/ced-dec/finepdfs-long/1920-2048/delta}{$-$0.00016}
\defval{kvseg/long/1024/ced-dec/finepdfs-long/1920-2048/hi}{+0.00005}
\defval{kvseg/long/1024/ced-dec/finepdfs-long/1920-2048/last}{1.0129}
\defval{kvseg/long/1024/ced-dec/finepdfs-long/1920-2048/lo}{$-$0.00037}
\defval{kvseg/long/1024/ced-dec/finepdfs-long/1920-2048/round}{1.0131}
\defval{kvseg/long/1024/ced-dec/finepdfs-long/384-512/delta}{+0.00011}
\defval{kvseg/long/1024/ced-dec/finepdfs-long/384-512/hi}{+0.00035}
\defval{kvseg/long/1024/ced-dec/finepdfs-long/384-512/last}{1.0811}
\defval{kvseg/long/1024/ced-dec/finepdfs-long/384-512/lo}{$-$0.00011}
\defval{kvseg/long/1024/ced-dec/finepdfs-long/384-512/round}{1.0810}
\defval{kvseg/long/1024/ced-dec/finepdfs-long/896-1024/delta}{+0.00015}
\defval{kvseg/long/1024/ced-dec/finepdfs-long/896-1024/hi}{+0.00036}
\defval{kvseg/long/1024/ced-dec/finepdfs-long/896-1024/last}{1.0647}
\defval{kvseg/long/1024/ced-dec/finepdfs-long/896-1024/lo}{$-$0.00004}
\defval{kvseg/long/1024/ced-dec/finepdfs-long/896-1024/round}{1.0645}
\defval{kvseg/long/1024/ced-dec/finepdfs-long/all/delta}{+0.00005}
\defval{kvseg/long/1024/ced-dec/finepdfs-long/all/hi}{+0.00015}
\defval{kvseg/long/1024/ced-dec/finepdfs-long/all/last}{1.1371}
\defval{kvseg/long/1024/ced-dec/finepdfs-long/all/lo}{$-$0.00005}
\defval{kvseg/long/1024/ced-dec/finepdfs-long/all/round}{1.1371}
\defval{kvseg/long/1024/ced-dec/fineweb-edu-100bt/1-128/delta}{+0.00002}
\defval{kvseg/long/1024/ced-dec/fineweb-edu-100bt/1-128/hi}{+0.00019}
\defval{kvseg/long/1024/ced-dec/fineweb-edu-100bt/1-128/last}{0.9426}
\defval{kvseg/long/1024/ced-dec/fineweb-edu-100bt/1-128/lo}{$-$0.00016}
\defval{kvseg/long/1024/ced-dec/fineweb-edu-100bt/1-128/round}{0.9426}
\defval{kvseg/long/1024/ced-dec/fineweb-edu-100bt/128-256/delta}{$-$0.00003}
\defval{kvseg/long/1024/ced-dec/fineweb-edu-100bt/128-256/hi}{+0.00013}
\defval{kvseg/long/1024/ced-dec/fineweb-edu-100bt/128-256/last}{0.8408}
\defval{kvseg/long/1024/ced-dec/fineweb-edu-100bt/128-256/lo}{$-$0.00018}
\defval{kvseg/long/1024/ced-dec/fineweb-edu-100bt/128-256/round}{0.8409}
\defval{kvseg/long/1024/ced-dec/fineweb-edu-100bt/1920-2048/delta}{$-$0.00005}
\defval{kvseg/long/1024/ced-dec/fineweb-edu-100bt/1920-2048/hi}{+0.00012}
\defval{kvseg/long/1024/ced-dec/fineweb-edu-100bt/1920-2048/last}{0.8196}
\defval{kvseg/long/1024/ced-dec/fineweb-edu-100bt/1920-2048/lo}{$-$0.00021}
\defval{kvseg/long/1024/ced-dec/fineweb-edu-100bt/1920-2048/round}{0.8196}
\defval{kvseg/long/1024/ced-dec/fineweb-edu-100bt/384-512/delta}{+0.00002}
\defval{kvseg/long/1024/ced-dec/fineweb-edu-100bt/384-512/hi}{+0.00017}
\defval{kvseg/long/1024/ced-dec/fineweb-edu-100bt/384-512/last}{0.8221}
\defval{kvseg/long/1024/ced-dec/fineweb-edu-100bt/384-512/lo}{$-$0.00012}
\defval{kvseg/long/1024/ced-dec/fineweb-edu-100bt/384-512/round}{0.8221}
\defval{kvseg/long/1024/ced-dec/fineweb-edu-100bt/896-1024/delta}{$-$0.00015}
\defval{kvseg/long/1024/ced-dec/fineweb-edu-100bt/896-1024/hi}{$-$0.00002}
\defval{kvseg/long/1024/ced-dec/fineweb-edu-100bt/896-1024/last}{0.8188}
\defval{kvseg/long/1024/ced-dec/fineweb-edu-100bt/896-1024/lo}{$-$0.00029}
\defval{kvseg/long/1024/ced-dec/fineweb-edu-100bt/896-1024/round}{0.8190}
\defval{kvseg/long/1024/ced-dec/fineweb-edu-100bt/all/delta}{$-$0.00004}
\defval{kvseg/long/1024/ced-dec/fineweb-edu-100bt/all/hi}{+0.00004}
\defval{kvseg/long/1024/ced-dec/fineweb-edu-100bt/all/last}{0.8486}
\defval{kvseg/long/1024/ced-dec/fineweb-edu-100bt/all/lo}{$-$0.00011}
\defval{kvseg/long/1024/ced-dec/fineweb-edu-100bt/all/round}{0.8486}
\defval{kvseg/long/1024/ced-dec/govreport-test/1-128/delta}{+0.00011}
\defval{kvseg/long/1024/ced-dec/govreport-test/1-128/hi}{+0.00027}
\defval{kvseg/long/1024/ced-dec/govreport-test/1-128/last}{0.8453}
\defval{kvseg/long/1024/ced-dec/govreport-test/1-128/lo}{$-$0.00002}
\defval{kvseg/long/1024/ced-dec/govreport-test/1-128/round}{0.8452}
\defval{kvseg/long/1024/ced-dec/govreport-test/128-256/delta}{+0.00005}
\defval{kvseg/long/1024/ced-dec/govreport-test/128-256/hi}{+0.00017}
\defval{kvseg/long/1024/ced-dec/govreport-test/128-256/last}{0.7184}
\defval{kvseg/long/1024/ced-dec/govreport-test/128-256/lo}{$-$0.00007}
\defval{kvseg/long/1024/ced-dec/govreport-test/128-256/round}{0.7184}
\defval{kvseg/long/1024/ced-dec/govreport-test/1920-2048/delta}{$-$0.00003}
\defval{kvseg/long/1024/ced-dec/govreport-test/1920-2048/hi}{+0.00008}
\defval{kvseg/long/1024/ced-dec/govreport-test/1920-2048/last}{0.6535}
\defval{kvseg/long/1024/ced-dec/govreport-test/1920-2048/lo}{$-$0.00014}
\defval{kvseg/long/1024/ced-dec/govreport-test/1920-2048/round}{0.6536}
\defval{kvseg/long/1024/ced-dec/govreport-test/384-512/delta}{$-$0.00010}
\defval{kvseg/long/1024/ced-dec/govreport-test/384-512/hi}{+0.00003}
\defval{kvseg/long/1024/ced-dec/govreport-test/384-512/last}{0.6816}
\defval{kvseg/long/1024/ced-dec/govreport-test/384-512/lo}{$-$0.00024}
\defval{kvseg/long/1024/ced-dec/govreport-test/384-512/round}{0.6817}
\defval{kvseg/long/1024/ced-dec/govreport-test/896-1024/delta}{$-$0.00001}
\defval{kvseg/long/1024/ced-dec/govreport-test/896-1024/hi}{+0.00011}
\defval{kvseg/long/1024/ced-dec/govreport-test/896-1024/last}{0.6629}
\defval{kvseg/long/1024/ced-dec/govreport-test/896-1024/lo}{$-$0.00012}
\defval{kvseg/long/1024/ced-dec/govreport-test/896-1024/round}{0.6629}
\defval{kvseg/long/1024/ced-dec/govreport-test/all/delta}{+0.00000}
\defval{kvseg/long/1024/ced-dec/govreport-test/all/hi}{+0.00006}
\defval{kvseg/long/1024/ced-dec/govreport-test/all/last}{0.7123}
\defval{kvseg/long/1024/ced-dec/govreport-test/all/lo}{$-$0.00005}
\defval{kvseg/long/1024/ced-dec/govreport-test/all/round}{0.7123}
\defval{kvseg/long/1024/ced-dec/longdocs/1-128/delta}{+0.00004}
\defval{kvseg/long/1024/ced-dec/longdocs/1-128/hi}{+0.00016}
\defval{kvseg/long/1024/ced-dec/longdocs/1-128/last}{1.1947}
\defval{kvseg/long/1024/ced-dec/longdocs/1-128/lo}{$-$0.00008}
\defval{kvseg/long/1024/ced-dec/longdocs/1-128/round}{1.1947}
\defval{kvseg/long/1024/ced-dec/longdocs/128-256/delta}{+0.00006}
\defval{kvseg/long/1024/ced-dec/longdocs/128-256/hi}{+0.00015}
\defval{kvseg/long/1024/ced-dec/longdocs/128-256/last}{1.0230}
\defval{kvseg/long/1024/ced-dec/longdocs/128-256/lo}{$-$0.00004}
\defval{kvseg/long/1024/ced-dec/longdocs/128-256/round}{1.0229}
\defval{kvseg/long/1024/ced-dec/longdocs/1920-2048/delta}{$-$0.00012}
\defval{kvseg/long/1024/ced-dec/longdocs/1920-2048/hi}{$-$0.00002}
\defval{kvseg/long/1024/ced-dec/longdocs/1920-2048/last}{0.9354}
\defval{kvseg/long/1024/ced-dec/longdocs/1920-2048/lo}{$-$0.00022}
\defval{kvseg/long/1024/ced-dec/longdocs/1920-2048/round}{0.9356}
\defval{kvseg/long/1024/ced-dec/longdocs/384-512/delta}{$-$0.00002}
\defval{kvseg/long/1024/ced-dec/longdocs/384-512/hi}{+0.00009}
\defval{kvseg/long/1024/ced-dec/longdocs/384-512/last}{0.9766}
\defval{kvseg/long/1024/ced-dec/longdocs/384-512/lo}{$-$0.00014}
\defval{kvseg/long/1024/ced-dec/longdocs/384-512/round}{0.9766}
\defval{kvseg/long/1024/ced-dec/longdocs/896-1024/delta}{$-$0.00000}
\defval{kvseg/long/1024/ced-dec/longdocs/896-1024/hi}{+0.00009}
\defval{kvseg/long/1024/ced-dec/longdocs/896-1024/last}{0.9630}
\defval{kvseg/long/1024/ced-dec/longdocs/896-1024/lo}{$-$0.00010}
\defval{kvseg/long/1024/ced-dec/longdocs/896-1024/round}{0.9630}
\defval{kvseg/long/1024/ced-dec/longdocs/all/delta}{$-$0.00001}
\defval{kvseg/long/1024/ced-dec/longdocs/all/hi}{+0.00004}
\defval{kvseg/long/1024/ced-dec/longdocs/all/last}{1.0185}
\defval{kvseg/long/1024/ced-dec/longdocs/all/lo}{$-$0.00006}
\defval{kvseg/long/1024/ced-dec/longdocs/all/round}{1.0186}
\defval{kvseg/long/1024/ced-dec/pg19-test/1-128/delta}{$-$0.00007}
\defval{kvseg/long/1024/ced-dec/pg19-test/1-128/hi}{+0.00015}
\defval{kvseg/long/1024/ced-dec/pg19-test/1-128/last}{1.4685}
\defval{kvseg/long/1024/ced-dec/pg19-test/1-128/lo}{$-$0.00029}
\defval{kvseg/long/1024/ced-dec/pg19-test/1-128/round}{1.4686}
\defval{kvseg/long/1024/ced-dec/pg19-test/128-256/delta}{+0.00004}
\defval{kvseg/long/1024/ced-dec/pg19-test/128-256/hi}{+0.00025}
\defval{kvseg/long/1024/ced-dec/pg19-test/128-256/last}{1.2958}
\defval{kvseg/long/1024/ced-dec/pg19-test/128-256/lo}{$-$0.00018}
\defval{kvseg/long/1024/ced-dec/pg19-test/128-256/round}{1.2958}
\defval{kvseg/long/1024/ced-dec/pg19-test/1920-2048/delta}{$-$0.00019}
\defval{kvseg/long/1024/ced-dec/pg19-test/1920-2048/hi}{+0.00003}
\defval{kvseg/long/1024/ced-dec/pg19-test/1920-2048/last}{1.2252}
\defval{kvseg/long/1024/ced-dec/pg19-test/1920-2048/lo}{$-$0.00043}
\defval{kvseg/long/1024/ced-dec/pg19-test/1920-2048/round}{1.2254}
\defval{kvseg/long/1024/ced-dec/pg19-test/384-512/delta}{$-$0.00004}
\defval{kvseg/long/1024/ced-dec/pg19-test/384-512/hi}{+0.00018}
\defval{kvseg/long/1024/ced-dec/pg19-test/384-512/last}{1.2553}
\defval{kvseg/long/1024/ced-dec/pg19-test/384-512/lo}{$-$0.00028}
\defval{kvseg/long/1024/ced-dec/pg19-test/384-512/round}{1.2553}
\defval{kvseg/long/1024/ced-dec/pg19-test/896-1024/delta}{$-$0.00015}
\defval{kvseg/long/1024/ced-dec/pg19-test/896-1024/hi}{+0.00006}
\defval{kvseg/long/1024/ced-dec/pg19-test/896-1024/last}{1.2525}
\defval{kvseg/long/1024/ced-dec/pg19-test/896-1024/lo}{$-$0.00038}
\defval{kvseg/long/1024/ced-dec/pg19-test/896-1024/round}{1.2526}
\defval{kvseg/long/1024/ced-dec/pg19-test/all/delta}{$-$0.00008}
\defval{kvseg/long/1024/ced-dec/pg19-test/all/hi}{+0.00001}
\defval{kvseg/long/1024/ced-dec/pg19-test/all/last}{1.2991}
\defval{kvseg/long/1024/ced-dec/pg19-test/all/lo}{$-$0.00018}
\defval{kvseg/long/1024/ced-dec/pg19-test/all/round}{1.2991}
\defval{kvseg/long/1024/ced-dec/pg19-val/1-128/delta}{$-$0.00025}
\defval{kvseg/long/1024/ced-dec/pg19-val/1-128/hi}{$-$0.00003}
\defval{kvseg/long/1024/ced-dec/pg19-val/1-128/last}{1.4177}
\defval{kvseg/long/1024/ced-dec/pg19-val/1-128/lo}{$-$0.00048}
\defval{kvseg/long/1024/ced-dec/pg19-val/1-128/round}{1.4179}
\defval{kvseg/long/1024/ced-dec/pg19-val/128-256/delta}{+0.00013}
\defval{kvseg/long/1024/ced-dec/pg19-val/128-256/hi}{+0.00034}
\defval{kvseg/long/1024/ced-dec/pg19-val/128-256/last}{1.2563}
\defval{kvseg/long/1024/ced-dec/pg19-val/128-256/lo}{$-$0.00007}
\defval{kvseg/long/1024/ced-dec/pg19-val/128-256/round}{1.2561}
\defval{kvseg/long/1024/ced-dec/pg19-val/1920-2048/delta}{$-$0.00006}
\defval{kvseg/long/1024/ced-dec/pg19-val/1920-2048/hi}{+0.00019}
\defval{kvseg/long/1024/ced-dec/pg19-val/1920-2048/last}{1.1868}
\defval{kvseg/long/1024/ced-dec/pg19-val/1920-2048/lo}{$-$0.00028}
\defval{kvseg/long/1024/ced-dec/pg19-val/1920-2048/round}{1.1868}
\defval{kvseg/long/1024/ced-dec/pg19-val/384-512/delta}{+0.00011}
\defval{kvseg/long/1024/ced-dec/pg19-val/384-512/hi}{+0.00036}
\defval{kvseg/long/1024/ced-dec/pg19-val/384-512/last}{1.2286}
\defval{kvseg/long/1024/ced-dec/pg19-val/384-512/lo}{$-$0.00011}
\defval{kvseg/long/1024/ced-dec/pg19-val/384-512/round}{1.2285}
\defval{kvseg/long/1024/ced-dec/pg19-val/896-1024/delta}{$-$0.00017}
\defval{kvseg/long/1024/ced-dec/pg19-val/896-1024/hi}{+0.00001}
\defval{kvseg/long/1024/ced-dec/pg19-val/896-1024/last}{1.2025}
\defval{kvseg/long/1024/ced-dec/pg19-val/896-1024/lo}{$-$0.00037}
\defval{kvseg/long/1024/ced-dec/pg19-val/896-1024/round}{1.2027}
\defval{kvseg/long/1024/ced-dec/pg19-val/all/delta}{$-$0.00005}
\defval{kvseg/long/1024/ced-dec/pg19-val/all/hi}{+0.00005}
\defval{kvseg/long/1024/ced-dec/pg19-val/all/last}{1.2582}
\defval{kvseg/long/1024/ced-dec/pg19-val/all/lo}{$-$0.00013}
\defval{kvseg/long/1024/ced-dec/pg19-val/all/round}{1.2583}
\defval{kvseg/long/1024/ced-enc/finepdfs-long/1-128/delta}{$-$0.00064}
\defval{kvseg/long/1024/ced-enc/finepdfs-long/1-128/hi}{$-$0.00020}
\defval{kvseg/long/1024/ced-enc/finepdfs-long/1-128/last}{1.3636}
\defval{kvseg/long/1024/ced-enc/finepdfs-long/1-128/lo}{$-$0.00111}
\defval{kvseg/long/1024/ced-enc/finepdfs-long/1-128/round}{1.3642}
\defval{kvseg/long/1024/ced-enc/finepdfs-long/128-256/delta}{$-$0.00020}
\defval{kvseg/long/1024/ced-enc/finepdfs-long/128-256/hi}{+0.00016}
\defval{kvseg/long/1024/ced-enc/finepdfs-long/128-256/last}{1.1145}
\defval{kvseg/long/1024/ced-enc/finepdfs-long/128-256/lo}{$-$0.00056}
\defval{kvseg/long/1024/ced-enc/finepdfs-long/128-256/round}{1.1147}
\defval{kvseg/long/1024/ced-enc/finepdfs-long/1920-2048/delta}{$-$0.00043}
\defval{kvseg/long/1024/ced-enc/finepdfs-long/1920-2048/hi}{$-$0.00001}
\defval{kvseg/long/1024/ced-enc/finepdfs-long/1920-2048/last}{0.9777}
\defval{kvseg/long/1024/ced-enc/finepdfs-long/1920-2048/lo}{$-$0.00083}
\defval{kvseg/long/1024/ced-enc/finepdfs-long/1920-2048/round}{0.9781}
\defval{kvseg/long/1024/ced-enc/finepdfs-long/384-512/delta}{$-$0.00059}
\defval{kvseg/long/1024/ced-enc/finepdfs-long/384-512/hi}{$-$0.00019}
\defval{kvseg/long/1024/ced-enc/finepdfs-long/384-512/last}{1.0510}
\defval{kvseg/long/1024/ced-enc/finepdfs-long/384-512/lo}{$-$0.00099}
\defval{kvseg/long/1024/ced-enc/finepdfs-long/384-512/round}{1.0516}
\defval{kvseg/long/1024/ced-enc/finepdfs-long/896-1024/delta}{$-$0.00084}
\defval{kvseg/long/1024/ced-enc/finepdfs-long/896-1024/hi}{$-$0.00048}
\defval{kvseg/long/1024/ced-enc/finepdfs-long/896-1024/last}{1.0332}
\defval{kvseg/long/1024/ced-enc/finepdfs-long/896-1024/lo}{$-$0.00124}
\defval{kvseg/long/1024/ced-enc/finepdfs-long/896-1024/round}{1.0341}
\defval{kvseg/long/1024/ced-enc/finepdfs-long/all/delta}{$-$0.00054}
\defval{kvseg/long/1024/ced-enc/finepdfs-long/all/hi}{$-$0.00035}
\defval{kvseg/long/1024/ced-enc/finepdfs-long/all/last}{1.1087}
\defval{kvseg/long/1024/ced-enc/finepdfs-long/all/lo}{$-$0.00074}
\defval{kvseg/long/1024/ced-enc/finepdfs-long/all/round}{1.1092}
\defval{kvseg/long/1024/ced-enc/fineweb-edu-100bt/1-128/delta}{$-$0.00018}
\defval{kvseg/long/1024/ced-enc/fineweb-edu-100bt/1-128/hi}{+0.00010}
\defval{kvseg/long/1024/ced-enc/fineweb-edu-100bt/1-128/last}{0.9291}
\defval{kvseg/long/1024/ced-enc/fineweb-edu-100bt/1-128/lo}{$-$0.00044}
\defval{kvseg/long/1024/ced-enc/fineweb-edu-100bt/1-128/round}{0.9292}
\defval{kvseg/long/1024/ced-enc/fineweb-edu-100bt/128-256/delta}{$-$0.00003}
\defval{kvseg/long/1024/ced-enc/fineweb-edu-100bt/128-256/hi}{+0.00022}
\defval{kvseg/long/1024/ced-enc/fineweb-edu-100bt/128-256/last}{0.8226}
\defval{kvseg/long/1024/ced-enc/fineweb-edu-100bt/128-256/lo}{$-$0.00030}
\defval{kvseg/long/1024/ced-enc/fineweb-edu-100bt/128-256/round}{0.8226}
\defval{kvseg/long/1024/ced-enc/fineweb-edu-100bt/1920-2048/delta}{$-$0.00034}
\defval{kvseg/long/1024/ced-enc/fineweb-edu-100bt/1920-2048/hi}{$-$0.00004}
\defval{kvseg/long/1024/ced-enc/fineweb-edu-100bt/1920-2048/last}{0.8005}
\defval{kvseg/long/1024/ced-enc/fineweb-edu-100bt/1920-2048/lo}{$-$0.00067}
\defval{kvseg/long/1024/ced-enc/fineweb-edu-100bt/1920-2048/round}{0.8009}
\defval{kvseg/long/1024/ced-enc/fineweb-edu-100bt/384-512/delta}{$-$0.00012}
\defval{kvseg/long/1024/ced-enc/fineweb-edu-100bt/384-512/hi}{+0.00016}
\defval{kvseg/long/1024/ced-enc/fineweb-edu-100bt/384-512/last}{0.8035}
\defval{kvseg/long/1024/ced-enc/fineweb-edu-100bt/384-512/lo}{$-$0.00039}
\defval{kvseg/long/1024/ced-enc/fineweb-edu-100bt/384-512/round}{0.8036}
\defval{kvseg/long/1024/ced-enc/fineweb-edu-100bt/896-1024/delta}{$-$0.00005}
\defval{kvseg/long/1024/ced-enc/fineweb-edu-100bt/896-1024/hi}{+0.00023}
\defval{kvseg/long/1024/ced-enc/fineweb-edu-100bt/896-1024/last}{0.8024}
\defval{kvseg/long/1024/ced-enc/fineweb-edu-100bt/896-1024/lo}{$-$0.00029}
\defval{kvseg/long/1024/ced-enc/fineweb-edu-100bt/896-1024/round}{0.8024}
\defval{kvseg/long/1024/ced-enc/fineweb-edu-100bt/all/delta}{$-$0.00014}
\defval{kvseg/long/1024/ced-enc/fineweb-edu-100bt/all/hi}{$-$0.00003}
\defval{kvseg/long/1024/ced-enc/fineweb-edu-100bt/all/last}{0.8314}
\defval{kvseg/long/1024/ced-enc/fineweb-edu-100bt/all/lo}{$-$0.00026}
\defval{kvseg/long/1024/ced-enc/fineweb-edu-100bt/all/round}{0.8315}
\defval{kvseg/long/1024/ced-enc/govreport-test/1-128/delta}{$-$0.00008}
\defval{kvseg/long/1024/ced-enc/govreport-test/1-128/hi}{+0.00018}
\defval{kvseg/long/1024/ced-enc/govreport-test/1-128/last}{0.8314}
\defval{kvseg/long/1024/ced-enc/govreport-test/1-128/lo}{$-$0.00032}
\defval{kvseg/long/1024/ced-enc/govreport-test/1-128/round}{0.8314}
\defval{kvseg/long/1024/ced-enc/govreport-test/128-256/delta}{$-$0.00016}
\defval{kvseg/long/1024/ced-enc/govreport-test/128-256/hi}{+0.00008}
\defval{kvseg/long/1024/ced-enc/govreport-test/128-256/last}{0.7021}
\defval{kvseg/long/1024/ced-enc/govreport-test/128-256/lo}{$-$0.00039}
\defval{kvseg/long/1024/ced-enc/govreport-test/128-256/round}{0.7023}
\defval{kvseg/long/1024/ced-enc/govreport-test/1920-2048/delta}{$-$0.00028}
\defval{kvseg/long/1024/ced-enc/govreport-test/1920-2048/hi}{$-$0.00008}
\defval{kvseg/long/1024/ced-enc/govreport-test/1920-2048/last}{0.6329}
\defval{kvseg/long/1024/ced-enc/govreport-test/1920-2048/lo}{$-$0.00049}
\defval{kvseg/long/1024/ced-enc/govreport-test/1920-2048/round}{0.6332}
\defval{kvseg/long/1024/ced-enc/govreport-test/384-512/delta}{$-$0.00023}
\defval{kvseg/long/1024/ced-enc/govreport-test/384-512/hi}{+0.00000}
\defval{kvseg/long/1024/ced-enc/govreport-test/384-512/last}{0.6641}
\defval{kvseg/long/1024/ced-enc/govreport-test/384-512/lo}{$-$0.00044}
\defval{kvseg/long/1024/ced-enc/govreport-test/384-512/round}{0.6643}
\defval{kvseg/long/1024/ced-enc/govreport-test/896-1024/delta}{$-$0.00026}
\defval{kvseg/long/1024/ced-enc/govreport-test/896-1024/hi}{+0.00002}
\defval{kvseg/long/1024/ced-enc/govreport-test/896-1024/last}{0.6440}
\defval{kvseg/long/1024/ced-enc/govreport-test/896-1024/lo}{$-$0.00051}
\defval{kvseg/long/1024/ced-enc/govreport-test/896-1024/round}{0.6442}
\defval{kvseg/long/1024/ced-enc/govreport-test/all/delta}{$-$0.00020}
\defval{kvseg/long/1024/ced-enc/govreport-test/all/hi}{$-$0.00010}
\defval{kvseg/long/1024/ced-enc/govreport-test/all/last}{0.6949}
\defval{kvseg/long/1024/ced-enc/govreport-test/all/lo}{$-$0.00030}
\defval{kvseg/long/1024/ced-enc/govreport-test/all/round}{0.6951}
\defval{kvseg/long/1024/ced-enc/longdocs/1-128/delta}{$-$0.00063}
\defval{kvseg/long/1024/ced-enc/longdocs/1-128/hi}{$-$0.00041}
\defval{kvseg/long/1024/ced-enc/longdocs/1-128/last}{1.1815}
\defval{kvseg/long/1024/ced-enc/longdocs/1-128/lo}{$-$0.00085}
\defval{kvseg/long/1024/ced-enc/longdocs/1-128/round}{1.1821}
\defval{kvseg/long/1024/ced-enc/longdocs/128-256/delta}{$-$0.00022}
\defval{kvseg/long/1024/ced-enc/longdocs/128-256/hi}{$-$0.00004}
\defval{kvseg/long/1024/ced-enc/longdocs/128-256/last}{0.9995}
\defval{kvseg/long/1024/ced-enc/longdocs/128-256/lo}{$-$0.00041}
\defval{kvseg/long/1024/ced-enc/longdocs/128-256/round}{0.9997}
\defval{kvseg/long/1024/ced-enc/longdocs/1920-2048/delta}{$-$0.00039}
\defval{kvseg/long/1024/ced-enc/longdocs/1920-2048/hi}{$-$0.00018}
\defval{kvseg/long/1024/ced-enc/longdocs/1920-2048/last}{0.9089}
\defval{kvseg/long/1024/ced-enc/longdocs/1920-2048/lo}{$-$0.00059}
\defval{kvseg/long/1024/ced-enc/longdocs/1920-2048/round}{0.9092}
\defval{kvseg/long/1024/ced-enc/longdocs/384-512/delta}{$-$0.00032}
\defval{kvseg/long/1024/ced-enc/longdocs/384-512/hi}{$-$0.00010}
\defval{kvseg/long/1024/ced-enc/longdocs/384-512/last}{0.9522}
\defval{kvseg/long/1024/ced-enc/longdocs/384-512/lo}{$-$0.00054}
\defval{kvseg/long/1024/ced-enc/longdocs/384-512/round}{0.9526}
\defval{kvseg/long/1024/ced-enc/longdocs/896-1024/delta}{$-$0.00065}
\defval{kvseg/long/1024/ced-enc/longdocs/896-1024/hi}{$-$0.00043}
\defval{kvseg/long/1024/ced-enc/longdocs/896-1024/last}{0.9370}
\defval{kvseg/long/1024/ced-enc/longdocs/896-1024/lo}{$-$0.00087}
\defval{kvseg/long/1024/ced-enc/longdocs/896-1024/round}{0.9377}
\defval{kvseg/long/1024/ced-enc/longdocs/all/delta}{$-$0.00044}
\defval{kvseg/long/1024/ced-enc/longdocs/all/hi}{$-$0.00034}
\defval{kvseg/long/1024/ced-enc/longdocs/all/last}{0.9958}
\defval{kvseg/long/1024/ced-enc/longdocs/all/lo}{$-$0.00053}
\defval{kvseg/long/1024/ced-enc/longdocs/all/round}{0.9963}
\defval{kvseg/long/1024/ced-enc/pg19-test/1-128/delta}{$-$0.00134}
\defval{kvseg/long/1024/ced-enc/pg19-test/1-128/hi}{$-$0.00083}
\defval{kvseg/long/1024/ced-enc/pg19-test/1-128/last}{1.4585}
\defval{kvseg/long/1024/ced-enc/pg19-test/1-128/lo}{$-$0.00182}
\defval{kvseg/long/1024/ced-enc/pg19-test/1-128/round}{1.4598}
\defval{kvseg/long/1024/ced-enc/pg19-test/128-256/delta}{$-$0.00031}
\defval{kvseg/long/1024/ced-enc/pg19-test/128-256/hi}{+0.00005}
\defval{kvseg/long/1024/ced-enc/pg19-test/128-256/last}{1.2695}
\defval{kvseg/long/1024/ced-enc/pg19-test/128-256/lo}{$-$0.00070}
\defval{kvseg/long/1024/ced-enc/pg19-test/128-256/round}{1.2698}
\defval{kvseg/long/1024/ced-enc/pg19-test/1920-2048/delta}{$-$0.00049}
\defval{kvseg/long/1024/ced-enc/pg19-test/1920-2048/hi}{$-$0.00008}
\defval{kvseg/long/1024/ced-enc/pg19-test/1920-2048/last}{1.1993}
\defval{kvseg/long/1024/ced-enc/pg19-test/1920-2048/lo}{$-$0.00093}
\defval{kvseg/long/1024/ced-enc/pg19-test/1920-2048/round}{1.1997}
\defval{kvseg/long/1024/ced-enc/pg19-test/384-512/delta}{$-$0.00016}
\defval{kvseg/long/1024/ced-enc/pg19-test/384-512/hi}{+0.00032}
\defval{kvseg/long/1024/ced-enc/pg19-test/384-512/last}{1.2276}
\defval{kvseg/long/1024/ced-enc/pg19-test/384-512/lo}{$-$0.00068}
\defval{kvseg/long/1024/ced-enc/pg19-test/384-512/round}{1.2278}
\defval{kvseg/long/1024/ced-enc/pg19-test/896-1024/delta}{$-$0.00096}
\defval{kvseg/long/1024/ced-enc/pg19-test/896-1024/hi}{$-$0.00045}
\defval{kvseg/long/1024/ced-enc/pg19-test/896-1024/last}{1.2228}
\defval{kvseg/long/1024/ced-enc/pg19-test/896-1024/lo}{$-$0.00147}
\defval{kvseg/long/1024/ced-enc/pg19-test/896-1024/round}{1.2238}
\defval{kvseg/long/1024/ced-enc/pg19-test/all/delta}{$-$0.00065}
\defval{kvseg/long/1024/ced-enc/pg19-test/all/hi}{$-$0.00044}
\defval{kvseg/long/1024/ced-enc/pg19-test/all/last}{1.2751}
\defval{kvseg/long/1024/ced-enc/pg19-test/all/lo}{$-$0.00087}
\defval{kvseg/long/1024/ced-enc/pg19-test/all/round}{1.2757}
\defval{kvseg/long/1024/ced-enc/pg19-val/1-128/delta}{$-$0.00070}
\defval{kvseg/long/1024/ced-enc/pg19-val/1-128/hi}{$-$0.00026}
\defval{kvseg/long/1024/ced-enc/pg19-val/1-128/last}{1.4095}
\defval{kvseg/long/1024/ced-enc/pg19-val/1-128/lo}{$-$0.00116}
\defval{kvseg/long/1024/ced-enc/pg19-val/1-128/round}{1.4102}
\defval{kvseg/long/1024/ced-enc/pg19-val/128-256/delta}{$-$0.00054}
\defval{kvseg/long/1024/ced-enc/pg19-val/128-256/hi}{$-$0.00012}
\defval{kvseg/long/1024/ced-enc/pg19-val/128-256/last}{1.2299}
\defval{kvseg/long/1024/ced-enc/pg19-val/128-256/lo}{$-$0.00099}
\defval{kvseg/long/1024/ced-enc/pg19-val/128-256/round}{1.2305}
\defval{kvseg/long/1024/ced-enc/pg19-val/1920-2048/delta}{$-$0.00003}
\defval{kvseg/long/1024/ced-enc/pg19-val/1920-2048/hi}{+0.00042}
\defval{kvseg/long/1024/ced-enc/pg19-val/1920-2048/last}{1.1592}
\defval{kvseg/long/1024/ced-enc/pg19-val/1920-2048/lo}{$-$0.00042}
\defval{kvseg/long/1024/ced-enc/pg19-val/1920-2048/round}{1.1592}
\defval{kvseg/long/1024/ced-enc/pg19-val/384-512/delta}{$-$0.00038}
\defval{kvseg/long/1024/ced-enc/pg19-val/384-512/hi}{+0.00002}
\defval{kvseg/long/1024/ced-enc/pg19-val/384-512/last}{1.2001}
\defval{kvseg/long/1024/ced-enc/pg19-val/384-512/lo}{$-$0.00081}
\defval{kvseg/long/1024/ced-enc/pg19-val/384-512/round}{1.2005}
\defval{kvseg/long/1024/ced-enc/pg19-val/896-1024/delta}{$-$0.00027}
\defval{kvseg/long/1024/ced-enc/pg19-val/896-1024/hi}{+0.00018}
\defval{kvseg/long/1024/ced-enc/pg19-val/896-1024/last}{1.1765}
\defval{kvseg/long/1024/ced-enc/pg19-val/896-1024/lo}{$-$0.00068}
\defval{kvseg/long/1024/ced-enc/pg19-val/896-1024/round}{1.1768}
\defval{kvseg/long/1024/ced-enc/pg19-val/all/delta}{$-$0.00038}
\defval{kvseg/long/1024/ced-enc/pg19-val/all/hi}{$-$0.00019}
\defval{kvseg/long/1024/ced-enc/pg19-val/all/last}{1.2349}
\defval{kvseg/long/1024/ced-enc/pg19-val/all/lo}{$-$0.00057}
\defval{kvseg/long/1024/ced-enc/pg19-val/all/round}{1.2353}
\defval{kvseg/long/1024/dense-l47/finepdfs-long/1-128/delta}{$-$0.00030}
\defval{kvseg/long/1024/dense-l47/finepdfs-long/1-128/hi}{+0.00010}
\defval{kvseg/long/1024/dense-l47/finepdfs-long/1-128/last}{1.3749}
\defval{kvseg/long/1024/dense-l47/finepdfs-long/1-128/lo}{$-$0.00074}
\defval{kvseg/long/1024/dense-l47/finepdfs-long/1-128/round}{1.3752}
\defval{kvseg/long/1024/dense-l47/finepdfs-long/128-256/delta}{+0.00020}
\defval{kvseg/long/1024/dense-l47/finepdfs-long/128-256/hi}{+0.00054}
\defval{kvseg/long/1024/dense-l47/finepdfs-long/128-256/last}{1.1314}
\defval{kvseg/long/1024/dense-l47/finepdfs-long/128-256/lo}{$-$0.00013}
\defval{kvseg/long/1024/dense-l47/finepdfs-long/128-256/round}{1.1312}
\defval{kvseg/long/1024/dense-l47/finepdfs-long/1920-2048/delta}{$-$0.00037}
\defval{kvseg/long/1024/dense-l47/finepdfs-long/1920-2048/hi}{+0.00002}
\defval{kvseg/long/1024/dense-l47/finepdfs-long/1920-2048/last}{0.9911}
\defval{kvseg/long/1024/dense-l47/finepdfs-long/1920-2048/lo}{$-$0.00076}
\defval{kvseg/long/1024/dense-l47/finepdfs-long/1920-2048/round}{0.9915}
\defval{kvseg/long/1024/dense-l47/finepdfs-long/384-512/delta}{$-$0.00050}
\defval{kvseg/long/1024/dense-l47/finepdfs-long/384-512/hi}{$-$0.00015}
\defval{kvseg/long/1024/dense-l47/finepdfs-long/384-512/last}{1.0658}
\defval{kvseg/long/1024/dense-l47/finepdfs-long/384-512/lo}{$-$0.00083}
\defval{kvseg/long/1024/dense-l47/finepdfs-long/384-512/round}{1.0663}
\defval{kvseg/long/1024/dense-l47/finepdfs-long/896-1024/delta}{$-$0.00011}
\defval{kvseg/long/1024/dense-l47/finepdfs-long/896-1024/hi}{+0.00023}
\defval{kvseg/long/1024/dense-l47/finepdfs-long/896-1024/last}{1.0455}
\defval{kvseg/long/1024/dense-l47/finepdfs-long/896-1024/lo}{$-$0.00046}
\defval{kvseg/long/1024/dense-l47/finepdfs-long/896-1024/round}{1.0456}
\defval{kvseg/long/1024/dense-l47/finepdfs-long/all/delta}{$-$0.00022}
\defval{kvseg/long/1024/dense-l47/finepdfs-long/all/hi}{$-$0.00005}
\defval{kvseg/long/1024/dense-l47/finepdfs-long/all/last}{1.1224}
\defval{kvseg/long/1024/dense-l47/finepdfs-long/all/lo}{$-$0.00038}
\defval{kvseg/long/1024/dense-l47/finepdfs-long/all/round}{1.1226}
\defval{kvseg/long/1024/dense-l47/fineweb-edu-100bt/1-128/delta}{+0.00011}
\defval{kvseg/long/1024/dense-l47/fineweb-edu-100bt/1-128/hi}{+0.00039}
\defval{kvseg/long/1024/dense-l47/fineweb-edu-100bt/1-128/last}{0.9357}
\defval{kvseg/long/1024/dense-l47/fineweb-edu-100bt/1-128/lo}{$-$0.00017}
\defval{kvseg/long/1024/dense-l47/fineweb-edu-100bt/1-128/round}{0.9356}
\defval{kvseg/long/1024/dense-l47/fineweb-edu-100bt/128-256/delta}{$-$0.00011}
\defval{kvseg/long/1024/dense-l47/fineweb-edu-100bt/128-256/hi}{+0.00016}
\defval{kvseg/long/1024/dense-l47/fineweb-edu-100bt/128-256/last}{0.8296}
\defval{kvseg/long/1024/dense-l47/fineweb-edu-100bt/128-256/lo}{$-$0.00035}
\defval{kvseg/long/1024/dense-l47/fineweb-edu-100bt/128-256/round}{0.8297}
\defval{kvseg/long/1024/dense-l47/fineweb-edu-100bt/1920-2048/delta}{+0.00000}
\defval{kvseg/long/1024/dense-l47/fineweb-edu-100bt/1920-2048/hi}{+0.00026}
\defval{kvseg/long/1024/dense-l47/fineweb-edu-100bt/1920-2048/last}{0.8061}
\defval{kvseg/long/1024/dense-l47/fineweb-edu-100bt/1920-2048/lo}{$-$0.00027}
\defval{kvseg/long/1024/dense-l47/fineweb-edu-100bt/1920-2048/round}{0.8061}
\defval{kvseg/long/1024/dense-l47/fineweb-edu-100bt/384-512/delta}{$-$0.00015}
\defval{kvseg/long/1024/dense-l47/fineweb-edu-100bt/384-512/hi}{+0.00009}
\defval{kvseg/long/1024/dense-l47/fineweb-edu-100bt/384-512/last}{0.8095}
\defval{kvseg/long/1024/dense-l47/fineweb-edu-100bt/384-512/lo}{$-$0.00040}
\defval{kvseg/long/1024/dense-l47/fineweb-edu-100bt/384-512/round}{0.8096}
\defval{kvseg/long/1024/dense-l47/fineweb-edu-100bt/896-1024/delta}{$-$0.00004}
\defval{kvseg/long/1024/dense-l47/fineweb-edu-100bt/896-1024/hi}{+0.00022}
\defval{kvseg/long/1024/dense-l47/fineweb-edu-100bt/896-1024/last}{0.8079}
\defval{kvseg/long/1024/dense-l47/fineweb-edu-100bt/896-1024/lo}{$-$0.00031}
\defval{kvseg/long/1024/dense-l47/fineweb-edu-100bt/896-1024/round}{0.8080}
\defval{kvseg/long/1024/dense-l47/fineweb-edu-100bt/all/delta}{$-$0.00004}
\defval{kvseg/long/1024/dense-l47/fineweb-edu-100bt/all/hi}{+0.00009}
\defval{kvseg/long/1024/dense-l47/fineweb-edu-100bt/all/last}{0.8376}
\defval{kvseg/long/1024/dense-l47/fineweb-edu-100bt/all/lo}{$-$0.00016}
\defval{kvseg/long/1024/dense-l47/fineweb-edu-100bt/all/round}{0.8376}
\defval{kvseg/long/1024/dense-l47/govreport-test/1-128/delta}{+0.00012}
\defval{kvseg/long/1024/dense-l47/govreport-test/1-128/hi}{+0.00033}
\defval{kvseg/long/1024/dense-l47/govreport-test/1-128/last}{0.8368}
\defval{kvseg/long/1024/dense-l47/govreport-test/1-128/lo}{$-$0.00010}
\defval{kvseg/long/1024/dense-l47/govreport-test/1-128/round}{0.8367}
\defval{kvseg/long/1024/dense-l47/govreport-test/128-256/delta}{+0.00011}
\defval{kvseg/long/1024/dense-l47/govreport-test/128-256/hi}{+0.00031}
\defval{kvseg/long/1024/dense-l47/govreport-test/128-256/last}{0.7075}
\defval{kvseg/long/1024/dense-l47/govreport-test/128-256/lo}{$-$0.00007}
\defval{kvseg/long/1024/dense-l47/govreport-test/128-256/round}{0.7073}
\defval{kvseg/long/1024/dense-l47/govreport-test/1920-2048/delta}{$-$0.00007}
\defval{kvseg/long/1024/dense-l47/govreport-test/1920-2048/hi}{+0.00014}
\defval{kvseg/long/1024/dense-l47/govreport-test/1920-2048/last}{0.6367}
\defval{kvseg/long/1024/dense-l47/govreport-test/1920-2048/lo}{$-$0.00027}
\defval{kvseg/long/1024/dense-l47/govreport-test/1920-2048/round}{0.6367}
\defval{kvseg/long/1024/dense-l47/govreport-test/384-512/delta}{+0.00012}
\defval{kvseg/long/1024/dense-l47/govreport-test/384-512/hi}{+0.00033}
\defval{kvseg/long/1024/dense-l47/govreport-test/384-512/last}{0.6670}
\defval{kvseg/long/1024/dense-l47/govreport-test/384-512/lo}{$-$0.00011}
\defval{kvseg/long/1024/dense-l47/govreport-test/384-512/round}{0.6668}
\defval{kvseg/long/1024/dense-l47/govreport-test/896-1024/delta}{$-$0.00001}
\defval{kvseg/long/1024/dense-l47/govreport-test/896-1024/hi}{+0.00021}
\defval{kvseg/long/1024/dense-l47/govreport-test/896-1024/last}{0.6473}
\defval{kvseg/long/1024/dense-l47/govreport-test/896-1024/lo}{$-$0.00025}
\defval{kvseg/long/1024/dense-l47/govreport-test/896-1024/round}{0.6473}
\defval{kvseg/long/1024/dense-l47/govreport-test/all/delta}{+0.00005}
\defval{kvseg/long/1024/dense-l47/govreport-test/all/hi}{+0.00014}
\defval{kvseg/long/1024/dense-l47/govreport-test/all/last}{0.6990}
\defval{kvseg/long/1024/dense-l47/govreport-test/all/lo}{$-$0.00004}
\defval{kvseg/long/1024/dense-l47/govreport-test/all/round}{0.6990}
\defval{kvseg/long/1024/dense-l47/longdocs/1-128/delta}{$-$0.00018}
\defval{kvseg/long/1024/dense-l47/longdocs/1-128/hi}{$-$0.00000}
\defval{kvseg/long/1024/dense-l47/longdocs/1-128/last}{1.1910}
\defval{kvseg/long/1024/dense-l47/longdocs/1-128/lo}{$-$0.00038}
\defval{kvseg/long/1024/dense-l47/longdocs/1-128/round}{1.1911}
\defval{kvseg/long/1024/dense-l47/longdocs/128-256/delta}{+0.00010}
\defval{kvseg/long/1024/dense-l47/longdocs/128-256/hi}{+0.00026}
\defval{kvseg/long/1024/dense-l47/longdocs/128-256/last}{1.0116}
\defval{kvseg/long/1024/dense-l47/longdocs/128-256/lo}{$-$0.00006}
\defval{kvseg/long/1024/dense-l47/longdocs/128-256/round}{1.0115}
\defval{kvseg/long/1024/dense-l47/longdocs/1920-2048/delta}{$-$0.00015}
\defval{kvseg/long/1024/dense-l47/longdocs/1920-2048/hi}{+0.00004}
\defval{kvseg/long/1024/dense-l47/longdocs/1920-2048/last}{0.9164}
\defval{kvseg/long/1024/dense-l47/longdocs/1920-2048/lo}{$-$0.00033}
\defval{kvseg/long/1024/dense-l47/longdocs/1920-2048/round}{0.9165}
\defval{kvseg/long/1024/dense-l47/longdocs/384-512/delta}{$-$0.00021}
\defval{kvseg/long/1024/dense-l47/longdocs/384-512/hi}{$-$0.00005}
\defval{kvseg/long/1024/dense-l47/longdocs/384-512/last}{0.9620}
\defval{kvseg/long/1024/dense-l47/longdocs/384-512/lo}{$-$0.00039}
\defval{kvseg/long/1024/dense-l47/longdocs/384-512/round}{0.9622}
\defval{kvseg/long/1024/dense-l47/longdocs/896-1024/delta}{$-$0.00013}
\defval{kvseg/long/1024/dense-l47/longdocs/896-1024/hi}{+0.00005}
\defval{kvseg/long/1024/dense-l47/longdocs/896-1024/last}{0.9450}
\defval{kvseg/long/1024/dense-l47/longdocs/896-1024/lo}{$-$0.00032}
\defval{kvseg/long/1024/dense-l47/longdocs/896-1024/round}{0.9452}
\defval{kvseg/long/1024/dense-l47/longdocs/all/delta}{$-$0.00011}
\defval{kvseg/long/1024/dense-l47/longdocs/all/hi}{$-$0.00004}
\defval{kvseg/long/1024/dense-l47/longdocs/all/last}{1.0052}
\defval{kvseg/long/1024/dense-l47/longdocs/all/lo}{$-$0.00019}
\defval{kvseg/long/1024/dense-l47/longdocs/all/round}{1.0053}
\defval{kvseg/long/1024/dense-l47/pg19-test/1-128/delta}{$-$0.00046}
\defval{kvseg/long/1024/dense-l47/pg19-test/1-128/hi}{$-$0.00008}
\defval{kvseg/long/1024/dense-l47/pg19-test/1-128/last}{1.4714}
\defval{kvseg/long/1024/dense-l47/pg19-test/1-128/lo}{$-$0.00082}
\defval{kvseg/long/1024/dense-l47/pg19-test/1-128/round}{1.4719}
\defval{kvseg/long/1024/dense-l47/pg19-test/128-256/delta}{$-$0.00001}
\defval{kvseg/long/1024/dense-l47/pg19-test/128-256/hi}{+0.00034}
\defval{kvseg/long/1024/dense-l47/pg19-test/128-256/last}{1.2858}
\defval{kvseg/long/1024/dense-l47/pg19-test/128-256/lo}{$-$0.00034}
\defval{kvseg/long/1024/dense-l47/pg19-test/128-256/round}{1.2858}
\defval{kvseg/long/1024/dense-l47/pg19-test/1920-2048/delta}{$-$0.00004}
\defval{kvseg/long/1024/dense-l47/pg19-test/1920-2048/hi}{+0.00032}
\defval{kvseg/long/1024/dense-l47/pg19-test/1920-2048/last}{1.2059}
\defval{kvseg/long/1024/dense-l47/pg19-test/1920-2048/lo}{$-$0.00041}
\defval{kvseg/long/1024/dense-l47/pg19-test/1920-2048/round}{1.2059}
\defval{kvseg/long/1024/dense-l47/pg19-test/384-512/delta}{$-$0.00037}
\defval{kvseg/long/1024/dense-l47/pg19-test/384-512/hi}{+0.00000}
\defval{kvseg/long/1024/dense-l47/pg19-test/384-512/last}{1.2414}
\defval{kvseg/long/1024/dense-l47/pg19-test/384-512/lo}{$-$0.00073}
\defval{kvseg/long/1024/dense-l47/pg19-test/384-512/round}{1.2418}
\defval{kvseg/long/1024/dense-l47/pg19-test/896-1024/delta}{$-$0.00029}
\defval{kvseg/long/1024/dense-l47/pg19-test/896-1024/hi}{+0.00013}
\defval{kvseg/long/1024/dense-l47/pg19-test/896-1024/last}{1.2327}
\defval{kvseg/long/1024/dense-l47/pg19-test/896-1024/lo}{$-$0.00070}
\defval{kvseg/long/1024/dense-l47/pg19-test/896-1024/round}{1.2330}
\defval{kvseg/long/1024/dense-l47/pg19-test/all/delta}{$-$0.00023}
\defval{kvseg/long/1024/dense-l47/pg19-test/all/hi}{$-$0.00006}
\defval{kvseg/long/1024/dense-l47/pg19-test/all/last}{1.2870}
\defval{kvseg/long/1024/dense-l47/pg19-test/all/lo}{$-$0.00041}
\defval{kvseg/long/1024/dense-l47/pg19-test/all/round}{1.2872}
\defval{kvseg/long/1024/dense-l47/pg19-val/1-128/delta}{$-$0.00017}
\defval{kvseg/long/1024/dense-l47/pg19-val/1-128/hi}{+0.00022}
\defval{kvseg/long/1024/dense-l47/pg19-val/1-128/last}{1.4211}
\defval{kvseg/long/1024/dense-l47/pg19-val/1-128/lo}{$-$0.00058}
\defval{kvseg/long/1024/dense-l47/pg19-val/1-128/round}{1.4213}
\defval{kvseg/long/1024/dense-l47/pg19-val/128-256/delta}{$-$0.00029}
\defval{kvseg/long/1024/dense-l47/pg19-val/128-256/hi}{+0.00005}
\defval{kvseg/long/1024/dense-l47/pg19-val/128-256/last}{1.2472}
\defval{kvseg/long/1024/dense-l47/pg19-val/128-256/lo}{$-$0.00063}
\defval{kvseg/long/1024/dense-l47/pg19-val/128-256/round}{1.2475}
\defval{kvseg/long/1024/dense-l47/pg19-val/1920-2048/delta}{$-$0.00010}
\defval{kvseg/long/1024/dense-l47/pg19-val/1920-2048/hi}{+0.00028}
\defval{kvseg/long/1024/dense-l47/pg19-val/1920-2048/last}{1.1685}
\defval{kvseg/long/1024/dense-l47/pg19-val/1920-2048/lo}{$-$0.00045}
\defval{kvseg/long/1024/dense-l47/pg19-val/1920-2048/round}{1.1686}
\defval{kvseg/long/1024/dense-l47/pg19-val/384-512/delta}{$-$0.00026}
\defval{kvseg/long/1024/dense-l47/pg19-val/384-512/hi}{+0.00017}
\defval{kvseg/long/1024/dense-l47/pg19-val/384-512/last}{1.2136}
\defval{kvseg/long/1024/dense-l47/pg19-val/384-512/lo}{$-$0.00066}
\defval{kvseg/long/1024/dense-l47/pg19-val/384-512/round}{1.2139}
\defval{kvseg/long/1024/dense-l47/pg19-val/896-1024/delta}{$-$0.00033}
\defval{kvseg/long/1024/dense-l47/pg19-val/896-1024/hi}{+0.00004}
\defval{kvseg/long/1024/dense-l47/pg19-val/896-1024/last}{1.1866}
\defval{kvseg/long/1024/dense-l47/pg19-val/896-1024/lo}{$-$0.00074}
\defval{kvseg/long/1024/dense-l47/pg19-val/896-1024/round}{1.1869}
\defval{kvseg/long/1024/dense-l47/pg19-val/all/delta}{$-$0.00023}
\defval{kvseg/long/1024/dense-l47/pg19-val/all/hi}{$-$0.00004}
\defval{kvseg/long/1024/dense-l47/pg19-val/all/last}{1.2473}
\defval{kvseg/long/1024/dense-l47/pg19-val/all/lo}{$-$0.00040}
\defval{kvseg/long/1024/dense-l47/pg19-val/all/round}{1.2475}
\defval{kvseg/p140/768/p140-dec-fixed/finepdfs-long/1-128/delta}{+0.08350}
\defval{kvseg/p140/768/p140-dec-fixed/finepdfs-long/1-128/hi}{+0.08753}
\defval{kvseg/p140/768/p140-dec-fixed/finepdfs-long/1-128/last}{1.5278}
\defval{kvseg/p140/768/p140-dec-fixed/finepdfs-long/1-128/lo}{+0.07988}
\defval{kvseg/p140/768/p140-dec-fixed/finepdfs-long/1-128/round}{1.4443}
\defval{kvseg/p140/768/p140-dec-fixed/finepdfs-long/128-256/delta}{+0.09169}
\defval{kvseg/p140/768/p140-dec-fixed/finepdfs-long/128-256/hi}{+0.09656}
\defval{kvseg/p140/768/p140-dec-fixed/finepdfs-long/128-256/last}{1.3164}
\defval{kvseg/p140/768/p140-dec-fixed/finepdfs-long/128-256/lo}{+0.08701}
\defval{kvseg/p140/768/p140-dec-fixed/finepdfs-long/128-256/round}{1.2247}
\defval{kvseg/p140/768/p140-dec-fixed/finepdfs-long/1920-2048/delta}{+0.09922}
\defval{kvseg/p140/768/p140-dec-fixed/finepdfs-long/1920-2048/hi}{+0.10443}
\defval{kvseg/p140/768/p140-dec-fixed/finepdfs-long/1920-2048/last}{1.2063}
\defval{kvseg/p140/768/p140-dec-fixed/finepdfs-long/1920-2048/lo}{+0.09389}
\defval{kvseg/p140/768/p140-dec-fixed/finepdfs-long/1920-2048/round}{1.1071}
\defval{kvseg/p140/768/p140-dec-fixed/finepdfs-long/384-512/delta}{+0.09498}
\defval{kvseg/p140/768/p140-dec-fixed/finepdfs-long/384-512/hi}{+0.10006}
\defval{kvseg/p140/768/p140-dec-fixed/finepdfs-long/384-512/last}{1.2669}
\defval{kvseg/p140/768/p140-dec-fixed/finepdfs-long/384-512/lo}{+0.09052}
\defval{kvseg/p140/768/p140-dec-fixed/finepdfs-long/384-512/round}{1.1719}
\defval{kvseg/p140/768/p140-dec-fixed/finepdfs-long/896-1024/delta}{+0.09459}
\defval{kvseg/p140/768/p140-dec-fixed/finepdfs-long/896-1024/hi}{+0.10016}
\defval{kvseg/p140/768/p140-dec-fixed/finepdfs-long/896-1024/last}{1.2528}
\defval{kvseg/p140/768/p140-dec-fixed/finepdfs-long/896-1024/lo}{+0.08811}
\defval{kvseg/p140/768/p140-dec-fixed/finepdfs-long/896-1024/round}{1.1582}
\defval{kvseg/p140/768/p140-dec-fixed/finepdfs-long/all/delta}{+0.09277}
\defval{kvseg/p140/768/p140-dec-fixed/finepdfs-long/all/hi}{+0.09645}
\defval{kvseg/p140/768/p140-dec-fixed/finepdfs-long/all/last}{1.3146}
\defval{kvseg/p140/768/p140-dec-fixed/finepdfs-long/all/lo}{+0.08894}
\defval{kvseg/p140/768/p140-dec-fixed/finepdfs-long/all/round}{1.2218}
\defval{kvseg/p140/768/p140-dec-fixed/fineweb-edu-100bt/1-128/delta}{+0.05115}
\defval{kvseg/p140/768/p140-dec-fixed/fineweb-edu-100bt/1-128/hi}{+0.05324}
\defval{kvseg/p140/768/p140-dec-fixed/fineweb-edu-100bt/1-128/last}{1.0533}
\defval{kvseg/p140/768/p140-dec-fixed/fineweb-edu-100bt/1-128/lo}{+0.04931}
\defval{kvseg/p140/768/p140-dec-fixed/fineweb-edu-100bt/1-128/round}{1.0022}
\defval{kvseg/p140/768/p140-dec-fixed/fineweb-edu-100bt/128-256/delta}{+0.06018}
\defval{kvseg/p140/768/p140-dec-fixed/fineweb-edu-100bt/128-256/hi}{+0.06227}
\defval{kvseg/p140/768/p140-dec-fixed/fineweb-edu-100bt/128-256/last}{0.9647}
\defval{kvseg/p140/768/p140-dec-fixed/fineweb-edu-100bt/128-256/lo}{+0.05797}
\defval{kvseg/p140/768/p140-dec-fixed/fineweb-edu-100bt/128-256/round}{0.9045}
\defval{kvseg/p140/768/p140-dec-fixed/fineweb-edu-100bt/1920-2048/delta}{+0.05985}
\defval{kvseg/p140/768/p140-dec-fixed/fineweb-edu-100bt/1920-2048/hi}{+0.06213}
\defval{kvseg/p140/768/p140-dec-fixed/fineweb-edu-100bt/1920-2048/last}{0.9441}
\defval{kvseg/p140/768/p140-dec-fixed/fineweb-edu-100bt/1920-2048/lo}{+0.05752}
\defval{kvseg/p140/768/p140-dec-fixed/fineweb-edu-100bt/1920-2048/round}{0.8842}
\defval{kvseg/p140/768/p140-dec-fixed/fineweb-edu-100bt/384-512/delta}{+0.05874}
\defval{kvseg/p140/768/p140-dec-fixed/fineweb-edu-100bt/384-512/hi}{+0.06121}
\defval{kvseg/p140/768/p140-dec-fixed/fineweb-edu-100bt/384-512/last}{0.9444}
\defval{kvseg/p140/768/p140-dec-fixed/fineweb-edu-100bt/384-512/lo}{+0.05629}
\defval{kvseg/p140/768/p140-dec-fixed/fineweb-edu-100bt/384-512/round}{0.8856}
\defval{kvseg/p140/768/p140-dec-fixed/fineweb-edu-100bt/896-1024/delta}{+0.05809}
\defval{kvseg/p140/768/p140-dec-fixed/fineweb-edu-100bt/896-1024/hi}{+0.06043}
\defval{kvseg/p140/768/p140-dec-fixed/fineweb-edu-100bt/896-1024/last}{0.9425}
\defval{kvseg/p140/768/p140-dec-fixed/fineweb-edu-100bt/896-1024/lo}{+0.05582}
\defval{kvseg/p140/768/p140-dec-fixed/fineweb-edu-100bt/896-1024/round}{0.8844}
\defval{kvseg/p140/768/p140-dec-fixed/fineweb-edu-100bt/all/delta}{+0.05762}
\defval{kvseg/p140/768/p140-dec-fixed/fineweb-edu-100bt/all/hi}{+0.05871}
\defval{kvseg/p140/768/p140-dec-fixed/fineweb-edu-100bt/all/last}{0.9696}
\defval{kvseg/p140/768/p140-dec-fixed/fineweb-edu-100bt/all/lo}{+0.05658}
\defval{kvseg/p140/768/p140-dec-fixed/fineweb-edu-100bt/all/round}{0.9120}
\defval{kvseg/p140/768/p140-dec-fixed/govreport-test/1-128/delta}{+0.05086}
\defval{kvseg/p140/768/p140-dec-fixed/govreport-test/1-128/hi}{+0.05259}
\defval{kvseg/p140/768/p140-dec-fixed/govreport-test/1-128/last}{0.9443}
\defval{kvseg/p140/768/p140-dec-fixed/govreport-test/1-128/lo}{+0.04918}
\defval{kvseg/p140/768/p140-dec-fixed/govreport-test/1-128/round}{0.8934}
\defval{kvseg/p140/768/p140-dec-fixed/govreport-test/128-256/delta}{+0.06465}
\defval{kvseg/p140/768/p140-dec-fixed/govreport-test/128-256/hi}{+0.06675}
\defval{kvseg/p140/768/p140-dec-fixed/govreport-test/128-256/last}{0.8368}
\defval{kvseg/p140/768/p140-dec-fixed/govreport-test/128-256/lo}{+0.06261}
\defval{kvseg/p140/768/p140-dec-fixed/govreport-test/128-256/round}{0.7721}
\defval{kvseg/p140/768/p140-dec-fixed/govreport-test/1920-2048/delta}{+0.06978}
\defval{kvseg/p140/768/p140-dec-fixed/govreport-test/1920-2048/hi}{+0.07192}
\defval{kvseg/p140/768/p140-dec-fixed/govreport-test/1920-2048/last}{0.7823}
\defval{kvseg/p140/768/p140-dec-fixed/govreport-test/1920-2048/lo}{+0.06754}
\defval{kvseg/p140/768/p140-dec-fixed/govreport-test/1920-2048/round}{0.7125}
\defval{kvseg/p140/768/p140-dec-fixed/govreport-test/384-512/delta}{+0.06782}
\defval{kvseg/p140/768/p140-dec-fixed/govreport-test/384-512/hi}{+0.06994}
\defval{kvseg/p140/768/p140-dec-fixed/govreport-test/384-512/last}{0.8046}
\defval{kvseg/p140/768/p140-dec-fixed/govreport-test/384-512/lo}{+0.06560}
\defval{kvseg/p140/768/p140-dec-fixed/govreport-test/384-512/round}{0.7368}
\defval{kvseg/p140/768/p140-dec-fixed/govreport-test/896-1024/delta}{+0.06944}
\defval{kvseg/p140/768/p140-dec-fixed/govreport-test/896-1024/hi}{+0.07187}
\defval{kvseg/p140/768/p140-dec-fixed/govreport-test/896-1024/last}{0.7894}
\defval{kvseg/p140/768/p140-dec-fixed/govreport-test/896-1024/lo}{+0.06724}
\defval{kvseg/p140/768/p140-dec-fixed/govreport-test/896-1024/round}{0.7200}
\defval{kvseg/p140/768/p140-dec-fixed/govreport-test/all/delta}{+0.06451}
\defval{kvseg/p140/768/p140-dec-fixed/govreport-test/all/hi}{+0.06561}
\defval{kvseg/p140/768/p140-dec-fixed/govreport-test/all/last}{0.8315}
\defval{kvseg/p140/768/p140-dec-fixed/govreport-test/all/lo}{+0.06346}
\defval{kvseg/p140/768/p140-dec-fixed/govreport-test/all/round}{0.7669}
\defval{kvseg/p140/768/p140-dec-fixed/longdocs/1-128/delta}{+0.06771}
\defval{kvseg/p140/768/p140-dec-fixed/longdocs/1-128/hi}{+0.06936}
\defval{kvseg/p140/768/p140-dec-fixed/longdocs/1-128/last}{1.3260}
\defval{kvseg/p140/768/p140-dec-fixed/longdocs/1-128/lo}{+0.06617}
\defval{kvseg/p140/768/p140-dec-fixed/longdocs/1-128/round}{1.2583}
\defval{kvseg/p140/768/p140-dec-fixed/longdocs/128-256/delta}{+0.07972}
\defval{kvseg/p140/768/p140-dec-fixed/longdocs/128-256/hi}{+0.08153}
\defval{kvseg/p140/768/p140-dec-fixed/longdocs/128-256/last}{1.1784}
\defval{kvseg/p140/768/p140-dec-fixed/longdocs/128-256/lo}{+0.07780}
\defval{kvseg/p140/768/p140-dec-fixed/longdocs/128-256/round}{1.0987}
\defval{kvseg/p140/768/p140-dec-fixed/longdocs/1920-2048/delta}{+0.08447}
\defval{kvseg/p140/768/p140-dec-fixed/longdocs/1920-2048/hi}{+0.08648}
\defval{kvseg/p140/768/p140-dec-fixed/longdocs/1920-2048/last}{1.1056}
\defval{kvseg/p140/768/p140-dec-fixed/longdocs/1920-2048/lo}{+0.08235}
\defval{kvseg/p140/768/p140-dec-fixed/longdocs/1920-2048/round}{1.0211}
\defval{kvseg/p140/768/p140-dec-fixed/longdocs/384-512/delta}{+0.08149}
\defval{kvseg/p140/768/p140-dec-fixed/longdocs/384-512/hi}{+0.08343}
\defval{kvseg/p140/768/p140-dec-fixed/longdocs/384-512/last}{1.1404}
\defval{kvseg/p140/768/p140-dec-fixed/longdocs/384-512/lo}{+0.07959}
\defval{kvseg/p140/768/p140-dec-fixed/longdocs/384-512/round}{1.0589}
\defval{kvseg/p140/768/p140-dec-fixed/longdocs/896-1024/delta}{+0.08292}
\defval{kvseg/p140/768/p140-dec-fixed/longdocs/896-1024/hi}{+0.08500}
\defval{kvseg/p140/768/p140-dec-fixed/longdocs/896-1024/last}{1.1296}
\defval{kvseg/p140/768/p140-dec-fixed/longdocs/896-1024/lo}{+0.08072}
\defval{kvseg/p140/768/p140-dec-fixed/longdocs/896-1024/round}{1.0467}
\defval{kvseg/p140/768/p140-dec-fixed/longdocs/all/delta}{+0.07926}
\defval{kvseg/p140/768/p140-dec-fixed/longdocs/all/hi}{+0.08045}
\defval{kvseg/p140/768/p140-dec-fixed/longdocs/all/last}{1.1760}
\defval{kvseg/p140/768/p140-dec-fixed/longdocs/all/lo}{+0.07797}
\defval{kvseg/p140/768/p140-dec-fixed/longdocs/all/round}{1.0967}
\defval{kvseg/p140/768/p140-dec-fixed/pg19-test/1-128/delta}{+0.07410}
\defval{kvseg/p140/768/p140-dec-fixed/pg19-test/1-128/hi}{+0.07751}
\defval{kvseg/p140/768/p140-dec-fixed/pg19-test/1-128/last}{1.6248}
\defval{kvseg/p140/768/p140-dec-fixed/pg19-test/1-128/lo}{+0.07098}
\defval{kvseg/p140/768/p140-dec-fixed/pg19-test/1-128/round}{1.5507}
\defval{kvseg/p140/768/p140-dec-fixed/pg19-test/128-256/delta}{+0.08739}
\defval{kvseg/p140/768/p140-dec-fixed/pg19-test/128-256/hi}{+0.09088}
\defval{kvseg/p140/768/p140-dec-fixed/pg19-test/128-256/last}{1.4828}
\defval{kvseg/p140/768/p140-dec-fixed/pg19-test/128-256/lo}{+0.08409}
\defval{kvseg/p140/768/p140-dec-fixed/pg19-test/128-256/round}{1.3954}
\defval{kvseg/p140/768/p140-dec-fixed/pg19-test/1920-2048/delta}{+0.08925}
\defval{kvseg/p140/768/p140-dec-fixed/pg19-test/1920-2048/hi}{+0.09293}
\defval{kvseg/p140/768/p140-dec-fixed/pg19-test/1920-2048/last}{1.4264}
\defval{kvseg/p140/768/p140-dec-fixed/pg19-test/1920-2048/lo}{+0.08570}
\defval{kvseg/p140/768/p140-dec-fixed/pg19-test/1920-2048/round}{1.3372}
\defval{kvseg/p140/768/p140-dec-fixed/pg19-test/384-512/delta}{+0.08612}
\defval{kvseg/p140/768/p140-dec-fixed/pg19-test/384-512/hi}{+0.08956}
\defval{kvseg/p140/768/p140-dec-fixed/pg19-test/384-512/last}{1.4503}
\defval{kvseg/p140/768/p140-dec-fixed/pg19-test/384-512/lo}{+0.08276}
\defval{kvseg/p140/768/p140-dec-fixed/pg19-test/384-512/round}{1.3642}
\defval{kvseg/p140/768/p140-dec-fixed/pg19-test/896-1024/delta}{+0.08905}
\defval{kvseg/p140/768/p140-dec-fixed/pg19-test/896-1024/hi}{+0.09283}
\defval{kvseg/p140/768/p140-dec-fixed/pg19-test/896-1024/last}{1.4501}
\defval{kvseg/p140/768/p140-dec-fixed/pg19-test/896-1024/lo}{+0.08533}
\defval{kvseg/p140/768/p140-dec-fixed/pg19-test/896-1024/round}{1.3610}
\defval{kvseg/p140/768/p140-dec-fixed/pg19-test/all/delta}{+0.08521}
\defval{kvseg/p140/768/p140-dec-fixed/pg19-test/all/hi}{+0.08755}
\defval{kvseg/p140/768/p140-dec-fixed/pg19-test/all/last}{1.4865}
\defval{kvseg/p140/768/p140-dec-fixed/pg19-test/all/lo}{+0.08300}
\defval{kvseg/p140/768/p140-dec-fixed/pg19-test/all/round}{1.4013}
\defval{kvseg/p140/768/p140-dec-fixed/pg19-val/1-128/delta}{+0.06855}
\defval{kvseg/p140/768/p140-dec-fixed/pg19-val/1-128/hi}{+0.07170}
\defval{kvseg/p140/768/p140-dec-fixed/pg19-val/1-128/last}{1.5642}
\defval{kvseg/p140/768/p140-dec-fixed/pg19-val/1-128/lo}{+0.06558}
\defval{kvseg/p140/768/p140-dec-fixed/pg19-val/1-128/round}{1.4956}
\defval{kvseg/p140/768/p140-dec-fixed/pg19-val/128-256/delta}{+0.08331}
\defval{kvseg/p140/768/p140-dec-fixed/pg19-val/128-256/hi}{+0.08692}
\defval{kvseg/p140/768/p140-dec-fixed/pg19-val/128-256/last}{1.4350}
\defval{kvseg/p140/768/p140-dec-fixed/pg19-val/128-256/lo}{+0.08006}
\defval{kvseg/p140/768/p140-dec-fixed/pg19-val/128-256/round}{1.3517}
\defval{kvseg/p140/768/p140-dec-fixed/pg19-val/1920-2048/delta}{+0.08131}
\defval{kvseg/p140/768/p140-dec-fixed/pg19-val/1920-2048/hi}{+0.08459}
\defval{kvseg/p140/768/p140-dec-fixed/pg19-val/1920-2048/last}{1.3764}
\defval{kvseg/p140/768/p140-dec-fixed/pg19-val/1920-2048/lo}{+0.07800}
\defval{kvseg/p140/768/p140-dec-fixed/pg19-val/1920-2048/round}{1.2951}
\defval{kvseg/p140/768/p140-dec-fixed/pg19-val/384-512/delta}{+0.08547}
\defval{kvseg/p140/768/p140-dec-fixed/pg19-val/384-512/hi}{+0.08884}
\defval{kvseg/p140/768/p140-dec-fixed/pg19-val/384-512/last}{1.4147}
\defval{kvseg/p140/768/p140-dec-fixed/pg19-val/384-512/lo}{+0.08249}
\defval{kvseg/p140/768/p140-dec-fixed/pg19-val/384-512/round}{1.3292}
\defval{kvseg/p140/768/p140-dec-fixed/pg19-val/896-1024/delta}{+0.08489}
\defval{kvseg/p140/768/p140-dec-fixed/pg19-val/896-1024/hi}{+0.08783}
\defval{kvseg/p140/768/p140-dec-fixed/pg19-val/896-1024/last}{1.3952}
\defval{kvseg/p140/768/p140-dec-fixed/pg19-val/896-1024/lo}{+0.08162}
\defval{kvseg/p140/768/p140-dec-fixed/pg19-val/896-1024/round}{1.3104}
\defval{kvseg/p140/768/p140-dec-fixed/pg19-val/all/delta}{+0.08071}
\defval{kvseg/p140/768/p140-dec-fixed/pg19-val/all/hi}{+0.08262}
\defval{kvseg/p140/768/p140-dec-fixed/pg19-val/all/last}{1.4370}
\defval{kvseg/p140/768/p140-dec-fixed/pg19-val/all/lo}{+0.07889}
\defval{kvseg/p140/768/p140-dec-fixed/pg19-val/all/round}{1.3563}
\defval{kvseg/p140/768/p140-enc-fixed/finepdfs-long/1-128/delta}{+0.06600}
\defval{kvseg/p140/768/p140-enc-fixed/finepdfs-long/1-128/hi}{+0.07010}
\defval{kvseg/p140/768/p140-enc-fixed/finepdfs-long/1-128/last}{1.5009}
\defval{kvseg/p140/768/p140-enc-fixed/finepdfs-long/1-128/lo}{+0.06217}
\defval{kvseg/p140/768/p140-enc-fixed/finepdfs-long/1-128/round}{1.4349}
\defval{kvseg/p140/768/p140-enc-fixed/finepdfs-long/128-256/delta}{+0.06002}
\defval{kvseg/p140/768/p140-enc-fixed/finepdfs-long/128-256/hi}{+0.06387}
\defval{kvseg/p140/768/p140-enc-fixed/finepdfs-long/128-256/last}{1.2673}
\defval{kvseg/p140/768/p140-enc-fixed/finepdfs-long/128-256/lo}{+0.05655}
\defval{kvseg/p140/768/p140-enc-fixed/finepdfs-long/128-256/round}{1.2073}
\defval{kvseg/p140/768/p140-enc-fixed/finepdfs-long/1920-2048/delta}{+0.07663}
\defval{kvseg/p140/768/p140-enc-fixed/finepdfs-long/1920-2048/hi}{+0.08203}
\defval{kvseg/p140/768/p140-enc-fixed/finepdfs-long/1920-2048/last}{1.1498}
\defval{kvseg/p140/768/p140-enc-fixed/finepdfs-long/1920-2048/lo}{+0.07136}
\defval{kvseg/p140/768/p140-enc-fixed/finepdfs-long/1920-2048/round}{1.0732}
\defval{kvseg/p140/768/p140-enc-fixed/finepdfs-long/384-512/delta}{+0.06135}
\defval{kvseg/p140/768/p140-enc-fixed/finepdfs-long/384-512/hi}{+0.06494}
\defval{kvseg/p140/768/p140-enc-fixed/finepdfs-long/384-512/last}{1.2094}
\defval{kvseg/p140/768/p140-enc-fixed/finepdfs-long/384-512/lo}{+0.05791}
\defval{kvseg/p140/768/p140-enc-fixed/finepdfs-long/384-512/round}{1.1481}
\defval{kvseg/p140/768/p140-enc-fixed/finepdfs-long/896-1024/delta}{+0.06341}
\defval{kvseg/p140/768/p140-enc-fixed/finepdfs-long/896-1024/hi}{+0.06802}
\defval{kvseg/p140/768/p140-enc-fixed/finepdfs-long/896-1024/last}{1.1894}
\defval{kvseg/p140/768/p140-enc-fixed/finepdfs-long/896-1024/lo}{+0.05860}
\defval{kvseg/p140/768/p140-enc-fixed/finepdfs-long/896-1024/round}{1.1260}
\defval{kvseg/p140/768/p140-enc-fixed/finepdfs-long/all/delta}{+0.06546}
\defval{kvseg/p140/768/p140-enc-fixed/finepdfs-long/all/hi}{+0.06840}
\defval{kvseg/p140/768/p140-enc-fixed/finepdfs-long/all/last}{1.2640}
\defval{kvseg/p140/768/p140-enc-fixed/finepdfs-long/all/lo}{+0.06269}
\defval{kvseg/p140/768/p140-enc-fixed/finepdfs-long/all/round}{1.1985}
\defval{kvseg/p140/768/p140-enc-fixed/fineweb-edu-100bt/1-128/delta}{+0.03688}
\defval{kvseg/p140/768/p140-enc-fixed/fineweb-edu-100bt/1-128/hi}{+0.03862}
\defval{kvseg/p140/768/p140-enc-fixed/fineweb-edu-100bt/1-128/last}{1.0328}
\defval{kvseg/p140/768/p140-enc-fixed/fineweb-edu-100bt/1-128/lo}{+0.03521}
\defval{kvseg/p140/768/p140-enc-fixed/fineweb-edu-100bt/1-128/round}{0.9960}
\defval{kvseg/p140/768/p140-enc-fixed/fineweb-edu-100bt/128-256/delta}{+0.03610}
\defval{kvseg/p140/768/p140-enc-fixed/fineweb-edu-100bt/128-256/hi}{+0.03781}
\defval{kvseg/p140/768/p140-enc-fixed/fineweb-edu-100bt/128-256/last}{0.9322}
\defval{kvseg/p140/768/p140-enc-fixed/fineweb-edu-100bt/128-256/lo}{+0.03431}
\defval{kvseg/p140/768/p140-enc-fixed/fineweb-edu-100bt/128-256/round}{0.8961}
\defval{kvseg/p140/768/p140-enc-fixed/fineweb-edu-100bt/1920-2048/delta}{+0.03802}
\defval{kvseg/p140/768/p140-enc-fixed/fineweb-edu-100bt/1920-2048/hi}{+0.03999}
\defval{kvseg/p140/768/p140-enc-fixed/fineweb-edu-100bt/1920-2048/last}{0.9114}
\defval{kvseg/p140/768/p140-enc-fixed/fineweb-edu-100bt/1920-2048/lo}{+0.03620}
\defval{kvseg/p140/768/p140-enc-fixed/fineweb-edu-100bt/1920-2048/round}{0.8734}
\defval{kvseg/p140/768/p140-enc-fixed/fineweb-edu-100bt/384-512/delta}{+0.03743}
\defval{kvseg/p140/768/p140-enc-fixed/fineweb-edu-100bt/384-512/hi}{+0.03928}
\defval{kvseg/p140/768/p140-enc-fixed/fineweb-edu-100bt/384-512/last}{0.9135}
\defval{kvseg/p140/768/p140-enc-fixed/fineweb-edu-100bt/384-512/lo}{+0.03576}
\defval{kvseg/p140/768/p140-enc-fixed/fineweb-edu-100bt/384-512/round}{0.8761}
\defval{kvseg/p140/768/p140-enc-fixed/fineweb-edu-100bt/896-1024/delta}{+0.03640}
\defval{kvseg/p140/768/p140-enc-fixed/fineweb-edu-100bt/896-1024/hi}{+0.03818}
\defval{kvseg/p140/768/p140-enc-fixed/fineweb-edu-100bt/896-1024/last}{0.9122}
\defval{kvseg/p140/768/p140-enc-fixed/fineweb-edu-100bt/896-1024/lo}{+0.03471}
\defval{kvseg/p140/768/p140-enc-fixed/fineweb-edu-100bt/896-1024/round}{0.8758}
\defval{kvseg/p140/768/p140-enc-fixed/fineweb-edu-100bt/all/delta}{+0.03696}
\defval{kvseg/p140/768/p140-enc-fixed/fineweb-edu-100bt/all/hi}{+0.03792}
\defval{kvseg/p140/768/p140-enc-fixed/fineweb-edu-100bt/all/last}{0.9402}
\defval{kvseg/p140/768/p140-enc-fixed/fineweb-edu-100bt/all/lo}{+0.03607}
\defval{kvseg/p140/768/p140-enc-fixed/fineweb-edu-100bt/all/round}{0.9033}
\defval{kvseg/p140/768/p140-enc-fixed/govreport-test/1-128/delta}{+0.03477}
\defval{kvseg/p140/768/p140-enc-fixed/govreport-test/1-128/hi}{+0.03616}
\defval{kvseg/p140/768/p140-enc-fixed/govreport-test/1-128/last}{0.9233}
\defval{kvseg/p140/768/p140-enc-fixed/govreport-test/1-128/lo}{+0.03340}
\defval{kvseg/p140/768/p140-enc-fixed/govreport-test/1-128/round}{0.8885}
\defval{kvseg/p140/768/p140-enc-fixed/govreport-test/128-256/delta}{+0.03388}
\defval{kvseg/p140/768/p140-enc-fixed/govreport-test/128-256/hi}{+0.03528}
\defval{kvseg/p140/768/p140-enc-fixed/govreport-test/128-256/last}{0.7974}
\defval{kvseg/p140/768/p140-enc-fixed/govreport-test/128-256/lo}{+0.03251}
\defval{kvseg/p140/768/p140-enc-fixed/govreport-test/128-256/round}{0.7636}
\defval{kvseg/p140/768/p140-enc-fixed/govreport-test/1920-2048/delta}{+0.04283}
\defval{kvseg/p140/768/p140-enc-fixed/govreport-test/1920-2048/hi}{+0.04514}
\defval{kvseg/p140/768/p140-enc-fixed/govreport-test/1920-2048/last}{0.7401}
\defval{kvseg/p140/768/p140-enc-fixed/govreport-test/1920-2048/lo}{+0.04074}
\defval{kvseg/p140/768/p140-enc-fixed/govreport-test/1920-2048/round}{0.6973}
\defval{kvseg/p140/768/p140-enc-fixed/govreport-test/384-512/delta}{+0.03475}
\defval{kvseg/p140/768/p140-enc-fixed/govreport-test/384-512/hi}{+0.03623}
\defval{kvseg/p140/768/p140-enc-fixed/govreport-test/384-512/last}{0.7595}
\defval{kvseg/p140/768/p140-enc-fixed/govreport-test/384-512/lo}{+0.03333}
\defval{kvseg/p140/768/p140-enc-fixed/govreport-test/384-512/round}{0.7248}
\defval{kvseg/p140/768/p140-enc-fixed/govreport-test/896-1024/delta}{+0.03766}
\defval{kvseg/p140/768/p140-enc-fixed/govreport-test/896-1024/hi}{+0.03928}
\defval{kvseg/p140/768/p140-enc-fixed/govreport-test/896-1024/last}{0.7426}
\defval{kvseg/p140/768/p140-enc-fixed/govreport-test/896-1024/lo}{+0.03608}
\defval{kvseg/p140/768/p140-enc-fixed/govreport-test/896-1024/round}{0.7050}
\defval{kvseg/p140/768/p140-enc-fixed/govreport-test/all/delta}{+0.03678}
\defval{kvseg/p140/768/p140-enc-fixed/govreport-test/all/hi}{+0.03758}
\defval{kvseg/p140/768/p140-enc-fixed/govreport-test/all/last}{0.7926}
\defval{kvseg/p140/768/p140-enc-fixed/govreport-test/all/lo}{+0.03599}
\defval{kvseg/p140/768/p140-enc-fixed/govreport-test/all/round}{0.7558}
\defval{kvseg/p140/768/p140-enc-fixed/longdocs/1-128/delta}{+0.05165}
\defval{kvseg/p140/768/p140-enc-fixed/longdocs/1-128/hi}{+0.05317}
\defval{kvseg/p140/768/p140-enc-fixed/longdocs/1-128/last}{1.3032}
\defval{kvseg/p140/768/p140-enc-fixed/longdocs/1-128/lo}{+0.05016}
\defval{kvseg/p140/768/p140-enc-fixed/longdocs/1-128/round}{1.2516}
\defval{kvseg/p140/768/p140-enc-fixed/longdocs/128-256/delta}{+0.04851}
\defval{kvseg/p140/768/p140-enc-fixed/longdocs/128-256/hi}{+0.05000}
\defval{kvseg/p140/768/p140-enc-fixed/longdocs/128-256/last}{1.1352}
\defval{kvseg/p140/768/p140-enc-fixed/longdocs/128-256/lo}{+0.04711}
\defval{kvseg/p140/768/p140-enc-fixed/longdocs/128-256/round}{1.0867}
\defval{kvseg/p140/768/p140-enc-fixed/longdocs/1920-2048/delta}{+0.06090}
\defval{kvseg/p140/768/p140-enc-fixed/longdocs/1920-2048/hi}{+0.06273}
\defval{kvseg/p140/768/p140-enc-fixed/longdocs/1920-2048/last}{1.0602}
\defval{kvseg/p140/768/p140-enc-fixed/longdocs/1920-2048/lo}{+0.05891}
\defval{kvseg/p140/768/p140-enc-fixed/longdocs/1920-2048/round}{0.9993}
\defval{kvseg/p140/768/p140-enc-fixed/longdocs/384-512/delta}{+0.04907}
\defval{kvseg/p140/768/p140-enc-fixed/longdocs/384-512/hi}{+0.05043}
\defval{kvseg/p140/768/p140-enc-fixed/longdocs/384-512/last}{1.0909}
\defval{kvseg/p140/768/p140-enc-fixed/longdocs/384-512/lo}{+0.04774}
\defval{kvseg/p140/768/p140-enc-fixed/longdocs/384-512/round}{1.0418}
\defval{kvseg/p140/768/p140-enc-fixed/longdocs/896-1024/delta}{+0.05285}
\defval{kvseg/p140/768/p140-enc-fixed/longdocs/896-1024/hi}{+0.05436}
\defval{kvseg/p140/768/p140-enc-fixed/longdocs/896-1024/last}{1.0787}
\defval{kvseg/p140/768/p140-enc-fixed/longdocs/896-1024/lo}{+0.05125}
\defval{kvseg/p140/768/p140-enc-fixed/longdocs/896-1024/round}{1.0259}
\defval{kvseg/p140/768/p140-enc-fixed/longdocs/all/delta}{+0.05259}
\defval{kvseg/p140/768/p140-enc-fixed/longdocs/all/hi}{+0.05352}
\defval{kvseg/p140/768/p140-enc-fixed/longdocs/all/last}{1.1336}
\defval{kvseg/p140/768/p140-enc-fixed/longdocs/all/lo}{+0.05159}
\defval{kvseg/p140/768/p140-enc-fixed/longdocs/all/round}{1.0811}
\defval{kvseg/p140/768/p140-enc-fixed/pg19-test/1-128/delta}{+0.05950}
\defval{kvseg/p140/768/p140-enc-fixed/pg19-test/1-128/hi}{+0.06252}
\defval{kvseg/p140/768/p140-enc-fixed/pg19-test/1-128/last}{1.6038}
\defval{kvseg/p140/768/p140-enc-fixed/pg19-test/1-128/lo}{+0.05666}
\defval{kvseg/p140/768/p140-enc-fixed/pg19-test/1-128/round}{1.5443}
\defval{kvseg/p140/768/p140-enc-fixed/pg19-test/128-256/delta}{+0.05604}
\defval{kvseg/p140/768/p140-enc-fixed/pg19-test/128-256/hi}{+0.05864}
\defval{kvseg/p140/768/p140-enc-fixed/pg19-test/128-256/last}{1.4404}
\defval{kvseg/p140/768/p140-enc-fixed/pg19-test/128-256/lo}{+0.05346}
\defval{kvseg/p140/768/p140-enc-fixed/pg19-test/128-256/round}{1.3844}
\defval{kvseg/p140/768/p140-enc-fixed/pg19-test/1920-2048/delta}{+0.06910}
\defval{kvseg/p140/768/p140-enc-fixed/pg19-test/1920-2048/hi}{+0.07230}
\defval{kvseg/p140/768/p140-enc-fixed/pg19-test/1920-2048/last}{1.3876}
\defval{kvseg/p140/768/p140-enc-fixed/pg19-test/1920-2048/lo}{+0.06619}
\defval{kvseg/p140/768/p140-enc-fixed/pg19-test/1920-2048/round}{1.3185}
\defval{kvseg/p140/768/p140-enc-fixed/pg19-test/384-512/delta}{+0.05569}
\defval{kvseg/p140/768/p140-enc-fixed/pg19-test/384-512/hi}{+0.05814}
\defval{kvseg/p140/768/p140-enc-fixed/pg19-test/384-512/last}{1.4028}
\defval{kvseg/p140/768/p140-enc-fixed/pg19-test/384-512/lo}{+0.05339}
\defval{kvseg/p140/768/p140-enc-fixed/pg19-test/384-512/round}{1.3471}
\defval{kvseg/p140/768/p140-enc-fixed/pg19-test/896-1024/delta}{+0.06224}
\defval{kvseg/p140/768/p140-enc-fixed/pg19-test/896-1024/hi}{+0.06495}
\defval{kvseg/p140/768/p140-enc-fixed/pg19-test/896-1024/last}{1.4061}
\defval{kvseg/p140/768/p140-enc-fixed/pg19-test/896-1024/lo}{+0.05957}
\defval{kvseg/p140/768/p140-enc-fixed/pg19-test/896-1024/round}{1.3438}
\defval{kvseg/p140/768/p140-enc-fixed/pg19-test/all/delta}{+0.06051}
\defval{kvseg/p140/768/p140-enc-fixed/pg19-test/all/hi}{+0.06213}
\defval{kvseg/p140/768/p140-enc-fixed/pg19-test/all/last}{1.4477}
\defval{kvseg/p140/768/p140-enc-fixed/pg19-test/all/lo}{+0.05881}
\defval{kvseg/p140/768/p140-enc-fixed/pg19-test/all/round}{1.3872}
\defval{kvseg/p140/768/p140-enc-fixed/pg19-val/1-128/delta}{+0.05385}
\defval{kvseg/p140/768/p140-enc-fixed/pg19-val/1-128/hi}{+0.05656}
\defval{kvseg/p140/768/p140-enc-fixed/pg19-val/1-128/last}{1.5434}
\defval{kvseg/p140/768/p140-enc-fixed/pg19-val/1-128/lo}{+0.05124}
\defval{kvseg/p140/768/p140-enc-fixed/pg19-val/1-128/round}{1.4896}
\defval{kvseg/p140/768/p140-enc-fixed/pg19-val/128-256/delta}{+0.05217}
\defval{kvseg/p140/768/p140-enc-fixed/pg19-val/128-256/hi}{+0.05447}
\defval{kvseg/p140/768/p140-enc-fixed/pg19-val/128-256/last}{1.3952}
\defval{kvseg/p140/768/p140-enc-fixed/pg19-val/128-256/lo}{+0.04996}
\defval{kvseg/p140/768/p140-enc-fixed/pg19-val/128-256/round}{1.3430}
\defval{kvseg/p140/768/p140-enc-fixed/pg19-val/1920-2048/delta}{+0.06415}
\defval{kvseg/p140/768/p140-enc-fixed/pg19-val/1920-2048/hi}{+0.06705}
\defval{kvseg/p140/768/p140-enc-fixed/pg19-val/1920-2048/last}{1.3416}
\defval{kvseg/p140/768/p140-enc-fixed/pg19-val/1920-2048/lo}{+0.06134}
\defval{kvseg/p140/768/p140-enc-fixed/pg19-val/1920-2048/round}{1.2775}
\defval{kvseg/p140/768/p140-enc-fixed/pg19-val/384-512/delta}{+0.05740}
\defval{kvseg/p140/768/p140-enc-fixed/pg19-val/384-512/hi}{+0.05980}
\defval{kvseg/p140/768/p140-enc-fixed/pg19-val/384-512/last}{1.3710}
\defval{kvseg/p140/768/p140-enc-fixed/pg19-val/384-512/lo}{+0.05482}
\defval{kvseg/p140/768/p140-enc-fixed/pg19-val/384-512/round}{1.3136}
\defval{kvseg/p140/768/p140-enc-fixed/pg19-val/896-1024/delta}{+0.05750}
\defval{kvseg/p140/768/p140-enc-fixed/pg19-val/896-1024/hi}{+0.05997}
\defval{kvseg/p140/768/p140-enc-fixed/pg19-val/896-1024/last}{1.3508}
\defval{kvseg/p140/768/p140-enc-fixed/pg19-val/896-1024/lo}{+0.05519}
\defval{kvseg/p140/768/p140-enc-fixed/pg19-val/896-1024/round}{1.2933}
\defval{kvseg/p140/768/p140-enc-fixed/pg19-val/all/delta}{+0.05702}
\defval{kvseg/p140/768/p140-enc-fixed/pg19-val/all/hi}{+0.05836}
\defval{kvseg/p140/768/p140-enc-fixed/pg19-val/all/last}{1.4003}
\defval{kvseg/p140/768/p140-enc-fixed/pg19-val/all/lo}{+0.05571}
\defval{kvseg/p140/768/p140-enc-fixed/pg19-val/all/round}{1.3433}
\defval{kvseg/p140/768/p140-enc/finepdfs-long/1-128/delta}{$-$0.00011}
\defval{kvseg/p140/768/p140-enc/finepdfs-long/1-128/hi}{+0.00028}
\defval{kvseg/p140/768/p140-enc/finepdfs-long/1-128/last}{1.4266}
\defval{kvseg/p140/768/p140-enc/finepdfs-long/1-128/lo}{$-$0.00049}
\defval{kvseg/p140/768/p140-enc/finepdfs-long/1-128/round}{1.4267}
\defval{kvseg/p140/768/p140-enc/finepdfs-long/128-256/delta}{$-$0.00080}
\defval{kvseg/p140/768/p140-enc/finepdfs-long/128-256/hi}{$-$0.00046}
\defval{kvseg/p140/768/p140-enc/finepdfs-long/128-256/last}{1.1987}
\defval{kvseg/p140/768/p140-enc/finepdfs-long/128-256/lo}{$-$0.00114}
\defval{kvseg/p140/768/p140-enc/finepdfs-long/128-256/round}{1.1995}
\defval{kvseg/p140/768/p140-enc/finepdfs-long/1920-2048/delta}{$-$0.00069}
\defval{kvseg/p140/768/p140-enc/finepdfs-long/1920-2048/hi}{$-$0.00019}
\defval{kvseg/p140/768/p140-enc/finepdfs-long/1920-2048/last}{1.0656}
\defval{kvseg/p140/768/p140-enc/finepdfs-long/1920-2048/lo}{$-$0.00120}
\defval{kvseg/p140/768/p140-enc/finepdfs-long/1920-2048/round}{1.0663}
\defval{kvseg/p140/768/p140-enc/finepdfs-long/384-512/delta}{$-$0.00091}
\defval{kvseg/p140/768/p140-enc/finepdfs-long/384-512/hi}{$-$0.00050}
\defval{kvseg/p140/768/p140-enc/finepdfs-long/384-512/last}{1.1409}
\defval{kvseg/p140/768/p140-enc/finepdfs-long/384-512/lo}{$-$0.00132}
\defval{kvseg/p140/768/p140-enc/finepdfs-long/384-512/round}{1.1418}
\defval{kvseg/p140/768/p140-enc/finepdfs-long/896-1024/delta}{$-$0.00074}
\defval{kvseg/p140/768/p140-enc/finepdfs-long/896-1024/hi}{$-$0.00034}
\defval{kvseg/p140/768/p140-enc/finepdfs-long/896-1024/last}{1.1202}
\defval{kvseg/p140/768/p140-enc/finepdfs-long/896-1024/lo}{$-$0.00115}
\defval{kvseg/p140/768/p140-enc/finepdfs-long/896-1024/round}{1.1209}
\defval{kvseg/p140/768/p140-enc/finepdfs-long/all/delta}{$-$0.00065}
\defval{kvseg/p140/768/p140-enc/finepdfs-long/all/hi}{$-$0.00046}
\defval{kvseg/p140/768/p140-enc/finepdfs-long/all/last}{1.1911}
\defval{kvseg/p140/768/p140-enc/finepdfs-long/all/lo}{$-$0.00085}
\defval{kvseg/p140/768/p140-enc/finepdfs-long/all/round}{1.1917}
\defval{kvseg/p140/768/p140-enc/fineweb-edu-100bt/1-128/delta}{$-$0.00016}
\defval{kvseg/p140/768/p140-enc/fineweb-edu-100bt/1-128/hi}{+0.00010}
\defval{kvseg/p140/768/p140-enc/fineweb-edu-100bt/1-128/last}{0.9967}
\defval{kvseg/p140/768/p140-enc/fineweb-edu-100bt/1-128/lo}{$-$0.00044}
\defval{kvseg/p140/768/p140-enc/fineweb-edu-100bt/1-128/round}{0.9969}
\defval{kvseg/p140/768/p140-enc/fineweb-edu-100bt/128-256/delta}{$-$0.00030}
\defval{kvseg/p140/768/p140-enc/fineweb-edu-100bt/128-256/hi}{$-$0.00007}
\defval{kvseg/p140/768/p140-enc/fineweb-edu-100bt/128-256/last}{0.8952}
\defval{kvseg/p140/768/p140-enc/fineweb-edu-100bt/128-256/lo}{$-$0.00053}
\defval{kvseg/p140/768/p140-enc/fineweb-edu-100bt/128-256/round}{0.8955}
\defval{kvseg/p140/768/p140-enc/fineweb-edu-100bt/1920-2048/delta}{$-$0.00029}
\defval{kvseg/p140/768/p140-enc/fineweb-edu-100bt/1920-2048/hi}{$-$0.00009}
\defval{kvseg/p140/768/p140-enc/fineweb-edu-100bt/1920-2048/last}{0.8732}
\defval{kvseg/p140/768/p140-enc/fineweb-edu-100bt/1920-2048/lo}{$-$0.00052}
\defval{kvseg/p140/768/p140-enc/fineweb-edu-100bt/1920-2048/round}{0.8735}
\defval{kvseg/p140/768/p140-enc/fineweb-edu-100bt/384-512/delta}{$-$0.00026}
\defval{kvseg/p140/768/p140-enc/fineweb-edu-100bt/384-512/hi}{+0.00002}
\defval{kvseg/p140/768/p140-enc/fineweb-edu-100bt/384-512/last}{0.8765}
\defval{kvseg/p140/768/p140-enc/fineweb-edu-100bt/384-512/lo}{$-$0.00051}
\defval{kvseg/p140/768/p140-enc/fineweb-edu-100bt/384-512/round}{0.8767}
\defval{kvseg/p140/768/p140-enc/fineweb-edu-100bt/896-1024/delta}{$-$0.00030}
\defval{kvseg/p140/768/p140-enc/fineweb-edu-100bt/896-1024/hi}{$-$0.00007}
\defval{kvseg/p140/768/p140-enc/fineweb-edu-100bt/896-1024/last}{0.8751}
\defval{kvseg/p140/768/p140-enc/fineweb-edu-100bt/896-1024/lo}{$-$0.00049}
\defval{kvseg/p140/768/p140-enc/fineweb-edu-100bt/896-1024/round}{0.8754}
\defval{kvseg/p140/768/p140-enc/fineweb-edu-100bt/all/delta}{$-$0.00026}
\defval{kvseg/p140/768/p140-enc/fineweb-edu-100bt/all/hi}{$-$0.00016}
\defval{kvseg/p140/768/p140-enc/fineweb-edu-100bt/all/last}{0.9031}
\defval{kvseg/p140/768/p140-enc/fineweb-edu-100bt/all/lo}{$-$0.00037}
\defval{kvseg/p140/768/p140-enc/fineweb-edu-100bt/all/round}{0.9034}
\defval{kvseg/p140/768/p140-enc/govreport-test/1-128/delta}{$-$0.00029}
\defval{kvseg/p140/768/p140-enc/govreport-test/1-128/hi}{$-$0.00008}
\defval{kvseg/p140/768/p140-enc/govreport-test/1-128/last}{0.8884}
\defval{kvseg/p140/768/p140-enc/govreport-test/1-128/lo}{$-$0.00052}
\defval{kvseg/p140/768/p140-enc/govreport-test/1-128/round}{0.8887}
\defval{kvseg/p140/768/p140-enc/govreport-test/128-256/delta}{$-$0.00023}
\defval{kvseg/p140/768/p140-enc/govreport-test/128-256/hi}{$-$0.00002}
\defval{kvseg/p140/768/p140-enc/govreport-test/128-256/last}{0.7632}
\defval{kvseg/p140/768/p140-enc/govreport-test/128-256/lo}{$-$0.00044}
\defval{kvseg/p140/768/p140-enc/govreport-test/128-256/round}{0.7634}
\defval{kvseg/p140/768/p140-enc/govreport-test/1920-2048/delta}{$-$0.00037}
\defval{kvseg/p140/768/p140-enc/govreport-test/1920-2048/hi}{$-$0.00014}
\defval{kvseg/p140/768/p140-enc/govreport-test/1920-2048/last}{0.6956}
\defval{kvseg/p140/768/p140-enc/govreport-test/1920-2048/lo}{$-$0.00060}
\defval{kvseg/p140/768/p140-enc/govreport-test/1920-2048/round}{0.6960}
\defval{kvseg/p140/768/p140-enc/govreport-test/384-512/delta}{$-$0.00060}
\defval{kvseg/p140/768/p140-enc/govreport-test/384-512/hi}{$-$0.00034}
\defval{kvseg/p140/768/p140-enc/govreport-test/384-512/last}{0.7246}
\defval{kvseg/p140/768/p140-enc/govreport-test/384-512/lo}{$-$0.00087}
\defval{kvseg/p140/768/p140-enc/govreport-test/384-512/round}{0.7252}
\defval{kvseg/p140/768/p140-enc/govreport-test/896-1024/delta}{$-$0.00028}
\defval{kvseg/p140/768/p140-enc/govreport-test/896-1024/hi}{$-$0.00005}
\defval{kvseg/p140/768/p140-enc/govreport-test/896-1024/last}{0.7055}
\defval{kvseg/p140/768/p140-enc/govreport-test/896-1024/lo}{$-$0.00052}
\defval{kvseg/p140/768/p140-enc/govreport-test/896-1024/round}{0.7058}
\defval{kvseg/p140/768/p140-enc/govreport-test/all/delta}{$-$0.00036}
\defval{kvseg/p140/768/p140-enc/govreport-test/all/hi}{$-$0.00024}
\defval{kvseg/p140/768/p140-enc/govreport-test/all/last}{0.7555}
\defval{kvseg/p140/768/p140-enc/govreport-test/all/lo}{$-$0.00047}
\defval{kvseg/p140/768/p140-enc/govreport-test/all/round}{0.7558}
\defval{kvseg/p140/768/p140-enc/longdocs/1-128/delta}{$-$0.00042}
\defval{kvseg/p140/768/p140-enc/longdocs/1-128/hi}{$-$0.00023}
\defval{kvseg/p140/768/p140-enc/longdocs/1-128/last}{1.2476}
\defval{kvseg/p140/768/p140-enc/longdocs/1-128/lo}{$-$0.00061}
\defval{kvseg/p140/768/p140-enc/longdocs/1-128/round}{1.2480}
\defval{kvseg/p140/768/p140-enc/longdocs/128-256/delta}{$-$0.00080}
\defval{kvseg/p140/768/p140-enc/longdocs/128-256/hi}{$-$0.00062}
\defval{kvseg/p140/768/p140-enc/longdocs/128-256/last}{1.0821}
\defval{kvseg/p140/768/p140-enc/longdocs/128-256/lo}{$-$0.00097}
\defval{kvseg/p140/768/p140-enc/longdocs/128-256/round}{1.0829}
\defval{kvseg/p140/768/p140-enc/longdocs/1920-2048/delta}{$-$0.00076}
\defval{kvseg/p140/768/p140-enc/longdocs/1920-2048/hi}{$-$0.00055}
\defval{kvseg/p140/768/p140-enc/longdocs/1920-2048/last}{0.9945}
\defval{kvseg/p140/768/p140-enc/longdocs/1920-2048/lo}{$-$0.00096}
\defval{kvseg/p140/768/p140-enc/longdocs/1920-2048/round}{0.9953}
\defval{kvseg/p140/768/p140-enc/longdocs/384-512/delta}{$-$0.00097}
\defval{kvseg/p140/768/p140-enc/longdocs/384-512/hi}{$-$0.00077}
\defval{kvseg/p140/768/p140-enc/longdocs/384-512/last}{1.0387}
\defval{kvseg/p140/768/p140-enc/longdocs/384-512/lo}{$-$0.00118}
\defval{kvseg/p140/768/p140-enc/longdocs/384-512/round}{1.0397}
\defval{kvseg/p140/768/p140-enc/longdocs/896-1024/delta}{$-$0.00075}
\defval{kvseg/p140/768/p140-enc/longdocs/896-1024/hi}{$-$0.00055}
\defval{kvseg/p140/768/p140-enc/longdocs/896-1024/last}{1.0233}
\defval{kvseg/p140/768/p140-enc/longdocs/896-1024/lo}{$-$0.00095}
\defval{kvseg/p140/768/p140-enc/longdocs/896-1024/round}{1.0241}
\defval{kvseg/p140/768/p140-enc/longdocs/all/delta}{$-$0.00074}
\defval{kvseg/p140/768/p140-enc/longdocs/all/hi}{$-$0.00065}
\defval{kvseg/p140/768/p140-enc/longdocs/all/last}{1.0772}
\defval{kvseg/p140/768/p140-enc/longdocs/all/lo}{$-$0.00083}
\defval{kvseg/p140/768/p140-enc/longdocs/all/round}{1.0780}
\defval{kvseg/p140/768/p140-enc/pg19-test/1-128/delta}{$-$0.00088}
\defval{kvseg/p140/768/p140-enc/pg19-test/1-128/hi}{$-$0.00048}
\defval{kvseg/p140/768/p140-enc/pg19-test/1-128/last}{1.5395}
\defval{kvseg/p140/768/p140-enc/pg19-test/1-128/lo}{$-$0.00128}
\defval{kvseg/p140/768/p140-enc/pg19-test/1-128/round}{1.5404}
\defval{kvseg/p140/768/p140-enc/pg19-test/128-256/delta}{$-$0.00154}
\defval{kvseg/p140/768/p140-enc/pg19-test/128-256/hi}{$-$0.00115}
\defval{kvseg/p140/768/p140-enc/pg19-test/128-256/last}{1.3782}
\defval{kvseg/p140/768/p140-enc/pg19-test/128-256/lo}{$-$0.00192}
\defval{kvseg/p140/768/p140-enc/pg19-test/128-256/round}{1.3797}
\defval{kvseg/p140/768/p140-enc/pg19-test/1920-2048/delta}{$-$0.00131}
\defval{kvseg/p140/768/p140-enc/pg19-test/1920-2048/hi}{$-$0.00095}
\defval{kvseg/p140/768/p140-enc/pg19-test/1920-2048/last}{1.3123}
\defval{kvseg/p140/768/p140-enc/pg19-test/1920-2048/lo}{$-$0.00165}
\defval{kvseg/p140/768/p140-enc/pg19-test/1920-2048/round}{1.3136}
\defval{kvseg/p140/768/p140-enc/pg19-test/384-512/delta}{$-$0.00150}
\defval{kvseg/p140/768/p140-enc/pg19-test/384-512/hi}{$-$0.00111}
\defval{kvseg/p140/768/p140-enc/pg19-test/384-512/last}{1.3442}
\defval{kvseg/p140/768/p140-enc/pg19-test/384-512/lo}{$-$0.00191}
\defval{kvseg/p140/768/p140-enc/pg19-test/384-512/round}{1.3457}
\defval{kvseg/p140/768/p140-enc/pg19-test/896-1024/delta}{$-$0.00136}
\defval{kvseg/p140/768/p140-enc/pg19-test/896-1024/hi}{$-$0.00093}
\defval{kvseg/p140/768/p140-enc/pg19-test/896-1024/last}{1.3403}
\defval{kvseg/p140/768/p140-enc/pg19-test/896-1024/lo}{$-$0.00175}
\defval{kvseg/p140/768/p140-enc/pg19-test/896-1024/round}{1.3416}
\defval{kvseg/p140/768/p140-enc/pg19-test/all/delta}{$-$0.00132}
\defval{kvseg/p140/768/p140-enc/pg19-test/all/hi}{$-$0.00115}
\defval{kvseg/p140/768/p140-enc/pg19-test/all/last}{1.3825}
\defval{kvseg/p140/768/p140-enc/pg19-test/all/lo}{$-$0.00150}
\defval{kvseg/p140/768/p140-enc/pg19-test/all/round}{1.3838}
\defval{kvseg/p140/768/p140-enc/pg19-val/1-128/delta}{$-$0.00082}
\defval{kvseg/p140/768/p140-enc/pg19-val/1-128/hi}{$-$0.00036}
\defval{kvseg/p140/768/p140-enc/pg19-val/1-128/last}{1.4859}
\defval{kvseg/p140/768/p140-enc/pg19-val/1-128/lo}{$-$0.00120}
\defval{kvseg/p140/768/p140-enc/pg19-val/1-128/round}{1.4867}
\defval{kvseg/p140/768/p140-enc/pg19-val/128-256/delta}{$-$0.00133}
\defval{kvseg/p140/768/p140-enc/pg19-val/128-256/hi}{$-$0.00095}
\defval{kvseg/p140/768/p140-enc/pg19-val/128-256/last}{1.3393}
\defval{kvseg/p140/768/p140-enc/pg19-val/128-256/lo}{$-$0.00172}
\defval{kvseg/p140/768/p140-enc/pg19-val/128-256/round}{1.3406}
\defval{kvseg/p140/768/p140-enc/pg19-val/1920-2048/delta}{$-$0.00148}
\defval{kvseg/p140/768/p140-enc/pg19-val/1920-2048/hi}{$-$0.00103}
\defval{kvseg/p140/768/p140-enc/pg19-val/1920-2048/last}{1.2725}
\defval{kvseg/p140/768/p140-enc/pg19-val/1920-2048/lo}{$-$0.00191}
\defval{kvseg/p140/768/p140-enc/pg19-val/1920-2048/round}{1.2740}
\defval{kvseg/p140/768/p140-enc/pg19-val/384-512/delta}{$-$0.00093}
\defval{kvseg/p140/768/p140-enc/pg19-val/384-512/hi}{$-$0.00058}
\defval{kvseg/p140/768/p140-enc/pg19-val/384-512/last}{1.3115}
\defval{kvseg/p140/768/p140-enc/pg19-val/384-512/lo}{$-$0.00127}
\defval{kvseg/p140/768/p140-enc/pg19-val/384-512/round}{1.3125}
\defval{kvseg/p140/768/p140-enc/pg19-val/896-1024/delta}{$-$0.00155}
\defval{kvseg/p140/768/p140-enc/pg19-val/896-1024/hi}{$-$0.00118}
\defval{kvseg/p140/768/p140-enc/pg19-val/896-1024/last}{1.2914}
\defval{kvseg/p140/768/p140-enc/pg19-val/896-1024/lo}{$-$0.00191}
\defval{kvseg/p140/768/p140-enc/pg19-val/896-1024/round}{1.2929}
\defval{kvseg/p140/768/p140-enc/pg19-val/all/delta}{$-$0.00122}
\defval{kvseg/p140/768/p140-enc/pg19-val/all/hi}{$-$0.00104}
\defval{kvseg/p140/768/p140-enc/pg19-val/all/last}{1.3400}
\defval{kvseg/p140/768/p140-enc/pg19-val/all/lo}{$-$0.00141}
\defval{kvseg/p140/768/p140-enc/pg19-val/all/round}{1.3412}
\defval{kvseg/x40/1024/ced-dec/finepdfs-long/1-128/delta}{$-$0.00013}
\defval{kvseg/x40/1024/ced-dec/finepdfs-long/1-128/hi}{+0.00002}
\defval{kvseg/x40/1024/ced-dec/finepdfs-long/1-128/last}{1.4196}
\defval{kvseg/x40/1024/ced-dec/finepdfs-long/1-128/lo}{$-$0.00027}
\defval{kvseg/x40/1024/ced-dec/finepdfs-long/1-128/round}{1.4197}
\defval{kvseg/x40/1024/ced-dec/finepdfs-long/128-256/delta}{$-$0.00004}
\defval{kvseg/x40/1024/ced-dec/finepdfs-long/128-256/hi}{+0.00008}
\defval{kvseg/x40/1024/ced-dec/finepdfs-long/128-256/last}{1.1894}
\defval{kvseg/x40/1024/ced-dec/finepdfs-long/128-256/lo}{$-$0.00017}
\defval{kvseg/x40/1024/ced-dec/finepdfs-long/128-256/round}{1.1895}
\defval{kvseg/x40/1024/ced-dec/finepdfs-long/1920-2048/delta}{+0.00007}
\defval{kvseg/x40/1024/ced-dec/finepdfs-long/1920-2048/hi}{+0.00020}
\defval{kvseg/x40/1024/ced-dec/finepdfs-long/1920-2048/last}{1.0601}
\defval{kvseg/x40/1024/ced-dec/finepdfs-long/1920-2048/lo}{$-$0.00004}
\defval{kvseg/x40/1024/ced-dec/finepdfs-long/1920-2048/round}{1.0601}
\defval{kvseg/x40/1024/ced-dec/finepdfs-long/384-512/delta}{+0.00009}
\defval{kvseg/x40/1024/ced-dec/finepdfs-long/384-512/hi}{+0.00021}
\defval{kvseg/x40/1024/ced-dec/finepdfs-long/384-512/last}{1.1307}
\defval{kvseg/x40/1024/ced-dec/finepdfs-long/384-512/lo}{$-$0.00001}
\defval{kvseg/x40/1024/ced-dec/finepdfs-long/384-512/round}{1.1306}
\defval{kvseg/x40/1024/ced-dec/finepdfs-long/896-1024/delta}{+0.00004}
\defval{kvseg/x40/1024/ced-dec/finepdfs-long/896-1024/hi}{+0.00015}
\defval{kvseg/x40/1024/ced-dec/finepdfs-long/896-1024/last}{1.1142}
\defval{kvseg/x40/1024/ced-dec/finepdfs-long/896-1024/lo}{$-$0.00007}
\defval{kvseg/x40/1024/ced-dec/finepdfs-long/896-1024/round}{1.1142}
\defval{kvseg/x40/1024/ced-dec/finepdfs-long/all/delta}{+0.00001}
\defval{kvseg/x40/1024/ced-dec/finepdfs-long/all/hi}{+0.00006}
\defval{kvseg/x40/1024/ced-dec/finepdfs-long/all/last}{1.1835}
\defval{kvseg/x40/1024/ced-dec/finepdfs-long/all/lo}{$-$0.00005}
\defval{kvseg/x40/1024/ced-dec/finepdfs-long/all/round}{1.1834}
\defval{kvseg/x40/1024/ced-dec/fineweb-edu-100bt/1-128/delta}{$-$0.00024}
\defval{kvseg/x40/1024/ced-dec/fineweb-edu-100bt/1-128/hi}{$-$0.00010}
\defval{kvseg/x40/1024/ced-dec/fineweb-edu-100bt/1-128/last}{0.9738}
\defval{kvseg/x40/1024/ced-dec/fineweb-edu-100bt/1-128/lo}{$-$0.00040}
\defval{kvseg/x40/1024/ced-dec/fineweb-edu-100bt/1-128/round}{0.9741}
\defval{kvseg/x40/1024/ced-dec/fineweb-edu-100bt/128-256/delta}{$-$0.00002}
\defval{kvseg/x40/1024/ced-dec/fineweb-edu-100bt/128-256/hi}{+0.00010}
\defval{kvseg/x40/1024/ced-dec/fineweb-edu-100bt/128-256/last}{0.8753}
\defval{kvseg/x40/1024/ced-dec/fineweb-edu-100bt/128-256/lo}{$-$0.00014}
\defval{kvseg/x40/1024/ced-dec/fineweb-edu-100bt/128-256/round}{0.8753}
\defval{kvseg/x40/1024/ced-dec/fineweb-edu-100bt/1920-2048/delta}{+0.00002}
\defval{kvseg/x40/1024/ced-dec/fineweb-edu-100bt/1920-2048/hi}{+0.00013}
\defval{kvseg/x40/1024/ced-dec/fineweb-edu-100bt/1920-2048/last}{0.8534}
\defval{kvseg/x40/1024/ced-dec/fineweb-edu-100bt/1920-2048/lo}{$-$0.00009}
\defval{kvseg/x40/1024/ced-dec/fineweb-edu-100bt/1920-2048/round}{0.8534}
\defval{kvseg/x40/1024/ced-dec/fineweb-edu-100bt/384-512/delta}{+0.00001}
\defval{kvseg/x40/1024/ced-dec/fineweb-edu-100bt/384-512/hi}{+0.00013}
\defval{kvseg/x40/1024/ced-dec/fineweb-edu-100bt/384-512/last}{0.8597}
\defval{kvseg/x40/1024/ced-dec/fineweb-edu-100bt/384-512/lo}{$-$0.00009}
\defval{kvseg/x40/1024/ced-dec/fineweb-edu-100bt/384-512/round}{0.8597}
\defval{kvseg/x40/1024/ced-dec/fineweb-edu-100bt/896-1024/delta}{+0.00006}
\defval{kvseg/x40/1024/ced-dec/fineweb-edu-100bt/896-1024/hi}{+0.00017}
\defval{kvseg/x40/1024/ced-dec/fineweb-edu-100bt/896-1024/last}{0.8566}
\defval{kvseg/x40/1024/ced-dec/fineweb-edu-100bt/896-1024/lo}{$-$0.00006}
\defval{kvseg/x40/1024/ced-dec/fineweb-edu-100bt/896-1024/round}{0.8565}
\defval{kvseg/x40/1024/ced-dec/fineweb-edu-100bt/all/delta}{$-$0.00004}
\defval{kvseg/x40/1024/ced-dec/fineweb-edu-100bt/all/hi}{+0.00002}
\defval{kvseg/x40/1024/ced-dec/fineweb-edu-100bt/all/last}{0.8836}
\defval{kvseg/x40/1024/ced-dec/fineweb-edu-100bt/all/lo}{$-$0.00009}
\defval{kvseg/x40/1024/ced-dec/fineweb-edu-100bt/all/round}{0.8836}
\defval{kvseg/x40/1024/ced-dec/govreport-test/1-128/delta}{+0.00021}
\defval{kvseg/x40/1024/ced-dec/govreport-test/1-128/hi}{+0.00039}
\defval{kvseg/x40/1024/ced-dec/govreport-test/1-128/last}{0.8718}
\defval{kvseg/x40/1024/ced-dec/govreport-test/1-128/lo}{+0.00008}
\defval{kvseg/x40/1024/ced-dec/govreport-test/1-128/round}{0.8716}
\defval{kvseg/x40/1024/ced-dec/govreport-test/128-256/delta}{+0.00019}
\defval{kvseg/x40/1024/ced-dec/govreport-test/128-256/hi}{+0.00028}
\defval{kvseg/x40/1024/ced-dec/govreport-test/128-256/last}{0.7496}
\defval{kvseg/x40/1024/ced-dec/govreport-test/128-256/lo}{+0.00009}
\defval{kvseg/x40/1024/ced-dec/govreport-test/128-256/round}{0.7494}
\defval{kvseg/x40/1024/ced-dec/govreport-test/1920-2048/delta}{+0.00012}
\defval{kvseg/x40/1024/ced-dec/govreport-test/1920-2048/hi}{+0.00020}
\defval{kvseg/x40/1024/ced-dec/govreport-test/1920-2048/last}{0.6867}
\defval{kvseg/x40/1024/ced-dec/govreport-test/1920-2048/lo}{+0.00003}
\defval{kvseg/x40/1024/ced-dec/govreport-test/1920-2048/round}{0.6866}
\defval{kvseg/x40/1024/ced-dec/govreport-test/384-512/delta}{+0.00018}
\defval{kvseg/x40/1024/ced-dec/govreport-test/384-512/hi}{+0.00027}
\defval{kvseg/x40/1024/ced-dec/govreport-test/384-512/last}{0.7125}
\defval{kvseg/x40/1024/ced-dec/govreport-test/384-512/lo}{+0.00010}
\defval{kvseg/x40/1024/ced-dec/govreport-test/384-512/round}{0.7123}
\defval{kvseg/x40/1024/ced-dec/govreport-test/896-1024/delta}{+0.00015}
\defval{kvseg/x40/1024/ced-dec/govreport-test/896-1024/hi}{+0.00022}
\defval{kvseg/x40/1024/ced-dec/govreport-test/896-1024/last}{0.6946}
\defval{kvseg/x40/1024/ced-dec/govreport-test/896-1024/lo}{+0.00006}
\defval{kvseg/x40/1024/ced-dec/govreport-test/896-1024/round}{0.6944}
\defval{kvseg/x40/1024/ced-dec/govreport-test/all/delta}{+0.00017}
\defval{kvseg/x40/1024/ced-dec/govreport-test/all/hi}{+0.00022}
\defval{kvseg/x40/1024/ced-dec/govreport-test/all/last}{0.7430}
\defval{kvseg/x40/1024/ced-dec/govreport-test/all/lo}{+0.00013}
\defval{kvseg/x40/1024/ced-dec/govreport-test/all/round}{0.7429}
\defval{kvseg/x40/1024/ced-dec/longdocs/1-128/delta}{$-$0.00006}
\defval{kvseg/x40/1024/ced-dec/longdocs/1-128/hi}{+0.00002}
\defval{kvseg/x40/1024/ced-dec/longdocs/1-128/last}{1.2310}
\defval{kvseg/x40/1024/ced-dec/longdocs/1-128/lo}{$-$0.00016}
\defval{kvseg/x40/1024/ced-dec/longdocs/1-128/round}{1.2311}
\defval{kvseg/x40/1024/ced-dec/longdocs/128-256/delta}{+0.00012}
\defval{kvseg/x40/1024/ced-dec/longdocs/128-256/hi}{+0.00018}
\defval{kvseg/x40/1024/ced-dec/longdocs/128-256/last}{1.0648}
\defval{kvseg/x40/1024/ced-dec/longdocs/128-256/lo}{+0.00005}
\defval{kvseg/x40/1024/ced-dec/longdocs/128-256/round}{1.0647}
\defval{kvseg/x40/1024/ced-dec/longdocs/1920-2048/delta}{+0.00015}
\defval{kvseg/x40/1024/ced-dec/longdocs/1920-2048/hi}{+0.00021}
\defval{kvseg/x40/1024/ced-dec/longdocs/1920-2048/last}{0.9806}
\defval{kvseg/x40/1024/ced-dec/longdocs/1920-2048/lo}{+0.00008}
\defval{kvseg/x40/1024/ced-dec/longdocs/1920-2048/round}{0.9805}
\defval{kvseg/x40/1024/ced-dec/longdocs/384-512/delta}{+0.00012}
\defval{kvseg/x40/1024/ced-dec/longdocs/384-512/hi}{+0.00019}
\defval{kvseg/x40/1024/ced-dec/longdocs/384-512/last}{1.0217}
\defval{kvseg/x40/1024/ced-dec/longdocs/384-512/lo}{+0.00005}
\defval{kvseg/x40/1024/ced-dec/longdocs/384-512/round}{1.0216}
\defval{kvseg/x40/1024/ced-dec/longdocs/896-1024/delta}{+0.00014}
\defval{kvseg/x40/1024/ced-dec/longdocs/896-1024/hi}{+0.00020}
\defval{kvseg/x40/1024/ced-dec/longdocs/896-1024/last}{1.0080}
\defval{kvseg/x40/1024/ced-dec/longdocs/896-1024/lo}{+0.00007}
\defval{kvseg/x40/1024/ced-dec/longdocs/896-1024/round}{1.0079}
\defval{kvseg/x40/1024/ced-dec/longdocs/all/delta}{+0.00009}
\defval{kvseg/x40/1024/ced-dec/longdocs/all/hi}{+0.00012}
\defval{kvseg/x40/1024/ced-dec/longdocs/all/last}{1.0612}
\defval{kvseg/x40/1024/ced-dec/longdocs/all/lo}{+0.00006}
\defval{kvseg/x40/1024/ced-dec/longdocs/all/round}{1.0611}
\defval{kvseg/x40/1024/ced-dec/pg19-test/1-128/delta}{$-$0.00036}
\defval{kvseg/x40/1024/ced-dec/pg19-test/1-128/hi}{$-$0.00021}
\defval{kvseg/x40/1024/ced-dec/pg19-test/1-128/last}{1.5135}
\defval{kvseg/x40/1024/ced-dec/pg19-test/1-128/lo}{$-$0.00051}
\defval{kvseg/x40/1024/ced-dec/pg19-test/1-128/round}{1.5139}
\defval{kvseg/x40/1024/ced-dec/pg19-test/128-256/delta}{+0.00017}
\defval{kvseg/x40/1024/ced-dec/pg19-test/128-256/hi}{+0.00032}
\defval{kvseg/x40/1024/ced-dec/pg19-test/128-256/last}{1.3485}
\defval{kvseg/x40/1024/ced-dec/pg19-test/128-256/lo}{+0.00003}
\defval{kvseg/x40/1024/ced-dec/pg19-test/128-256/round}{1.3483}
\defval{kvseg/x40/1024/ced-dec/pg19-test/1920-2048/delta}{+0.00026}
\defval{kvseg/x40/1024/ced-dec/pg19-test/1920-2048/hi}{+0.00043}
\defval{kvseg/x40/1024/ced-dec/pg19-test/1920-2048/last}{1.2838}
\defval{kvseg/x40/1024/ced-dec/pg19-test/1920-2048/lo}{+0.00011}
\defval{kvseg/x40/1024/ced-dec/pg19-test/1920-2048/round}{1.2836}
\defval{kvseg/x40/1024/ced-dec/pg19-test/384-512/delta}{+0.00005}
\defval{kvseg/x40/1024/ced-dec/pg19-test/384-512/hi}{+0.00019}
\defval{kvseg/x40/1024/ced-dec/pg19-test/384-512/last}{1.3142}
\defval{kvseg/x40/1024/ced-dec/pg19-test/384-512/lo}{$-$0.00009}
\defval{kvseg/x40/1024/ced-dec/pg19-test/384-512/round}{1.3141}
\defval{kvseg/x40/1024/ced-dec/pg19-test/896-1024/delta}{+0.00021}
\defval{kvseg/x40/1024/ced-dec/pg19-test/896-1024/hi}{+0.00036}
\defval{kvseg/x40/1024/ced-dec/pg19-test/896-1024/last}{1.3104}
\defval{kvseg/x40/1024/ced-dec/pg19-test/896-1024/lo}{+0.00007}
\defval{kvseg/x40/1024/ced-dec/pg19-test/896-1024/round}{1.3102}
\defval{kvseg/x40/1024/ced-dec/pg19-test/all/delta}{+0.00007}
\defval{kvseg/x40/1024/ced-dec/pg19-test/all/hi}{+0.00014}
\defval{kvseg/x40/1024/ced-dec/pg19-test/all/last}{1.3537}
\defval{kvseg/x40/1024/ced-dec/pg19-test/all/lo}{+0.00001}
\defval{kvseg/x40/1024/ced-dec/pg19-test/all/round}{1.3536}
\defval{kvseg/x40/1024/ced-dec/pg19-val/1-128/delta}{$-$0.00024}
\defval{kvseg/x40/1024/ced-dec/pg19-val/1-128/hi}{$-$0.00008}
\defval{kvseg/x40/1024/ced-dec/pg19-val/1-128/last}{1.4603}
\defval{kvseg/x40/1024/ced-dec/pg19-val/1-128/lo}{$-$0.00041}
\defval{kvseg/x40/1024/ced-dec/pg19-val/1-128/round}{1.4606}
\defval{kvseg/x40/1024/ced-dec/pg19-val/128-256/delta}{$-$0.00002}
\defval{kvseg/x40/1024/ced-dec/pg19-val/128-256/hi}{+0.00014}
\defval{kvseg/x40/1024/ced-dec/pg19-val/128-256/last}{1.3095}
\defval{kvseg/x40/1024/ced-dec/pg19-val/128-256/lo}{$-$0.00017}
\defval{kvseg/x40/1024/ced-dec/pg19-val/128-256/round}{1.3095}
\defval{kvseg/x40/1024/ced-dec/pg19-val/1920-2048/delta}{+0.00033}
\defval{kvseg/x40/1024/ced-dec/pg19-val/1920-2048/hi}{+0.00049}
\defval{kvseg/x40/1024/ced-dec/pg19-val/1920-2048/last}{1.2417}
\defval{kvseg/x40/1024/ced-dec/pg19-val/1920-2048/lo}{+0.00017}
\defval{kvseg/x40/1024/ced-dec/pg19-val/1920-2048/round}{1.2413}
\defval{kvseg/x40/1024/ced-dec/pg19-val/384-512/delta}{+0.00020}
\defval{kvseg/x40/1024/ced-dec/pg19-val/384-512/hi}{+0.00036}
\defval{kvseg/x40/1024/ced-dec/pg19-val/384-512/last}{1.2821}
\defval{kvseg/x40/1024/ced-dec/pg19-val/384-512/lo}{+0.00005}
\defval{kvseg/x40/1024/ced-dec/pg19-val/384-512/round}{1.2819}
\defval{kvseg/x40/1024/ced-dec/pg19-val/896-1024/delta}{+0.00007}
\defval{kvseg/x40/1024/ced-dec/pg19-val/896-1024/hi}{+0.00022}
\defval{kvseg/x40/1024/ced-dec/pg19-val/896-1024/last}{1.2587}
\defval{kvseg/x40/1024/ced-dec/pg19-val/896-1024/lo}{$-$0.00007}
\defval{kvseg/x40/1024/ced-dec/pg19-val/896-1024/round}{1.2586}
\defval{kvseg/x40/1024/ced-dec/pg19-val/all/delta}{+0.00007}
\defval{kvseg/x40/1024/ced-dec/pg19-val/all/hi}{+0.00014}
\defval{kvseg/x40/1024/ced-dec/pg19-val/all/last}{1.3104}
\defval{kvseg/x40/1024/ced-dec/pg19-val/all/lo}{+0.00001}
\defval{kvseg/x40/1024/ced-dec/pg19-val/all/round}{1.3103}
\defval{kvseg/x40/1024/ced-enc-fixed/finepdfs-long/1-128/delta}{+0.07030}
\defval{kvseg/x40/1024/ced-enc-fixed/finepdfs-long/1-128/hi}{+0.07556}
\defval{kvseg/x40/1024/ced-enc-fixed/finepdfs-long/1-128/last}{1.4803}
\defval{kvseg/x40/1024/ced-enc-fixed/finepdfs-long/1-128/lo}{+0.06552}
\defval{kvseg/x40/1024/ced-enc-fixed/finepdfs-long/1-128/round}{1.4100}
\defval{kvseg/x40/1024/ced-enc-fixed/finepdfs-long/128-256/delta}{+0.05831}
\defval{kvseg/x40/1024/ced-enc-fixed/finepdfs-long/128-256/hi}{+0.06198}
\defval{kvseg/x40/1024/ced-enc-fixed/finepdfs-long/128-256/last}{1.2308}
\defval{kvseg/x40/1024/ced-enc-fixed/finepdfs-long/128-256/lo}{+0.05437}
\defval{kvseg/x40/1024/ced-enc-fixed/finepdfs-long/128-256/round}{1.1724}
\defval{kvseg/x40/1024/ced-enc-fixed/finepdfs-long/1920-2048/delta}{+0.05417}
\defval{kvseg/x40/1024/ced-enc-fixed/finepdfs-long/1920-2048/hi}{+0.05774}
\defval{kvseg/x40/1024/ced-enc-fixed/finepdfs-long/1920-2048/last}{1.0859}
\defval{kvseg/x40/1024/ced-enc-fixed/finepdfs-long/1920-2048/lo}{+0.05045}
\defval{kvseg/x40/1024/ced-enc-fixed/finepdfs-long/1920-2048/round}{1.0317}
\defval{kvseg/x40/1024/ced-enc-fixed/finepdfs-long/384-512/delta}{+0.05645}
\defval{kvseg/x40/1024/ced-enc-fixed/finepdfs-long/384-512/hi}{+0.05979}
\defval{kvseg/x40/1024/ced-enc-fixed/finepdfs-long/384-512/last}{1.1625}
\defval{kvseg/x40/1024/ced-enc-fixed/finepdfs-long/384-512/lo}{+0.05303}
\defval{kvseg/x40/1024/ced-enc-fixed/finepdfs-long/384-512/round}{1.1060}
\defval{kvseg/x40/1024/ced-enc-fixed/finepdfs-long/896-1024/delta}{+0.05753}
\defval{kvseg/x40/1024/ced-enc-fixed/finepdfs-long/896-1024/hi}{+0.06150}
\defval{kvseg/x40/1024/ced-enc-fixed/finepdfs-long/896-1024/last}{1.1427}
\defval{kvseg/x40/1024/ced-enc-fixed/finepdfs-long/896-1024/lo}{+0.05354}
\defval{kvseg/x40/1024/ced-enc-fixed/finepdfs-long/896-1024/round}{1.0852}
\defval{kvseg/x40/1024/ced-enc-fixed/finepdfs-long/all/delta}{+0.05937}
\defval{kvseg/x40/1024/ced-enc-fixed/finepdfs-long/all/hi}{+0.06228}
\defval{kvseg/x40/1024/ced-enc-fixed/finepdfs-long/all/last}{1.2211}
\defval{kvseg/x40/1024/ced-enc-fixed/finepdfs-long/all/lo}{+0.05676}
\defval{kvseg/x40/1024/ced-enc-fixed/finepdfs-long/all/round}{1.1617}
\defval{kvseg/x40/1024/ced-enc-fixed/fineweb-edu-100bt/1-128/delta}{+0.02789}
\defval{kvseg/x40/1024/ced-enc-fixed/fineweb-edu-100bt/1-128/hi}{+0.02951}
\defval{kvseg/x40/1024/ced-enc-fixed/fineweb-edu-100bt/1-128/last}{0.9897}
\defval{kvseg/x40/1024/ced-enc-fixed/fineweb-edu-100bt/1-128/lo}{+0.02636}
\defval{kvseg/x40/1024/ced-enc-fixed/fineweb-edu-100bt/1-128/round}{0.9618}
\defval{kvseg/x40/1024/ced-enc-fixed/fineweb-edu-100bt/128-256/delta}{+0.02799}
\defval{kvseg/x40/1024/ced-enc-fixed/fineweb-edu-100bt/128-256/hi}{+0.02940}
\defval{kvseg/x40/1024/ced-enc-fixed/fineweb-edu-100bt/128-256/last}{0.8860}
\defval{kvseg/x40/1024/ced-enc-fixed/fineweb-edu-100bt/128-256/lo}{+0.02648}
\defval{kvseg/x40/1024/ced-enc-fixed/fineweb-edu-100bt/128-256/round}{0.8580}
\defval{kvseg/x40/1024/ced-enc-fixed/fineweb-edu-100bt/1920-2048/delta}{+0.02585}
\defval{kvseg/x40/1024/ced-enc-fixed/fineweb-edu-100bt/1920-2048/hi}{+0.02722}
\defval{kvseg/x40/1024/ced-enc-fixed/fineweb-edu-100bt/1920-2048/last}{0.8620}
\defval{kvseg/x40/1024/ced-enc-fixed/fineweb-edu-100bt/1920-2048/lo}{+0.02439}
\defval{kvseg/x40/1024/ced-enc-fixed/fineweb-edu-100bt/1920-2048/round}{0.8361}
\defval{kvseg/x40/1024/ced-enc-fixed/fineweb-edu-100bt/384-512/delta}{+0.02640}
\defval{kvseg/x40/1024/ced-enc-fixed/fineweb-edu-100bt/384-512/hi}{+0.02787}
\defval{kvseg/x40/1024/ced-enc-fixed/fineweb-edu-100bt/384-512/last}{0.8693}
\defval{kvseg/x40/1024/ced-enc-fixed/fineweb-edu-100bt/384-512/lo}{+0.02488}
\defval{kvseg/x40/1024/ced-enc-fixed/fineweb-edu-100bt/384-512/round}{0.8429}
\defval{kvseg/x40/1024/ced-enc-fixed/fineweb-edu-100bt/896-1024/delta}{+0.02560}
\defval{kvseg/x40/1024/ced-enc-fixed/fineweb-edu-100bt/896-1024/hi}{+0.02704}
\defval{kvseg/x40/1024/ced-enc-fixed/fineweb-edu-100bt/896-1024/last}{0.8653}
\defval{kvseg/x40/1024/ced-enc-fixed/fineweb-edu-100bt/896-1024/lo}{+0.02434}
\defval{kvseg/x40/1024/ced-enc-fixed/fineweb-edu-100bt/896-1024/round}{0.8397}
\defval{kvseg/x40/1024/ced-enc-fixed/fineweb-edu-100bt/all/delta}{+0.02674}
\defval{kvseg/x40/1024/ced-enc-fixed/fineweb-edu-100bt/all/hi}{+0.02750}
\defval{kvseg/x40/1024/ced-enc-fixed/fineweb-edu-100bt/all/last}{0.8943}
\defval{kvseg/x40/1024/ced-enc-fixed/fineweb-edu-100bt/all/lo}{+0.02604}
\defval{kvseg/x40/1024/ced-enc-fixed/fineweb-edu-100bt/all/round}{0.8675}
\defval{kvseg/x40/1024/ced-enc-fixed/govreport-test/1-128/delta}{+0.03263}
\defval{kvseg/x40/1024/ced-enc-fixed/govreport-test/1-128/hi}{+0.03392}
\defval{kvseg/x40/1024/ced-enc-fixed/govreport-test/1-128/last}{0.8941}
\defval{kvseg/x40/1024/ced-enc-fixed/govreport-test/1-128/lo}{+0.03133}
\defval{kvseg/x40/1024/ced-enc-fixed/govreport-test/1-128/round}{0.8614}
\defval{kvseg/x40/1024/ced-enc-fixed/govreport-test/128-256/delta}{+0.02989}
\defval{kvseg/x40/1024/ced-enc-fixed/govreport-test/128-256/hi}{+0.03121}
\defval{kvseg/x40/1024/ced-enc-fixed/govreport-test/128-256/last}{0.7642}
\defval{kvseg/x40/1024/ced-enc-fixed/govreport-test/128-256/lo}{+0.02861}
\defval{kvseg/x40/1024/ced-enc-fixed/govreport-test/128-256/round}{0.7343}
\defval{kvseg/x40/1024/ced-enc-fixed/govreport-test/1920-2048/delta}{+0.02861}
\defval{kvseg/x40/1024/ced-enc-fixed/govreport-test/1920-2048/hi}{+0.03000}
\defval{kvseg/x40/1024/ced-enc-fixed/govreport-test/1920-2048/last}{0.6977}
\defval{kvseg/x40/1024/ced-enc-fixed/govreport-test/1920-2048/lo}{+0.02727}
\defval{kvseg/x40/1024/ced-enc-fixed/govreport-test/1920-2048/round}{0.6691}
\defval{kvseg/x40/1024/ced-enc-fixed/govreport-test/384-512/delta}{+0.02716}
\defval{kvseg/x40/1024/ced-enc-fixed/govreport-test/384-512/hi}{+0.02843}
\defval{kvseg/x40/1024/ced-enc-fixed/govreport-test/384-512/last}{0.7229}
\defval{kvseg/x40/1024/ced-enc-fixed/govreport-test/384-512/lo}{+0.02595}
\defval{kvseg/x40/1024/ced-enc-fixed/govreport-test/384-512/round}{0.6957}
\defval{kvseg/x40/1024/ced-enc-fixed/govreport-test/896-1024/delta}{+0.02707}
\defval{kvseg/x40/1024/ced-enc-fixed/govreport-test/896-1024/hi}{+0.02845}
\defval{kvseg/x40/1024/ced-enc-fixed/govreport-test/896-1024/last}{0.7063}
\defval{kvseg/x40/1024/ced-enc-fixed/govreport-test/896-1024/lo}{+0.02575}
\defval{kvseg/x40/1024/ced-enc-fixed/govreport-test/896-1024/round}{0.6792}
\defval{kvseg/x40/1024/ced-enc-fixed/govreport-test/all/delta}{+0.02907}
\defval{kvseg/x40/1024/ced-enc-fixed/govreport-test/all/hi}{+0.02973}
\defval{kvseg/x40/1024/ced-enc-fixed/govreport-test/all/last}{0.7570}
\defval{kvseg/x40/1024/ced-enc-fixed/govreport-test/all/lo}{+0.02845}
\defval{kvseg/x40/1024/ced-enc-fixed/govreport-test/all/round}{0.7279}
\defval{kvseg/x40/1024/ced-enc-fixed/longdocs/1-128/delta}{+0.05076}
\defval{kvseg/x40/1024/ced-enc-fixed/longdocs/1-128/hi}{+0.05272}
\defval{kvseg/x40/1024/ced-enc-fixed/longdocs/1-128/last}{1.2697}
\defval{kvseg/x40/1024/ced-enc-fixed/longdocs/1-128/lo}{+0.04908}
\defval{kvseg/x40/1024/ced-enc-fixed/longdocs/1-128/round}{1.2189}
\defval{kvseg/x40/1024/ced-enc-fixed/longdocs/128-256/delta}{+0.04460}
\defval{kvseg/x40/1024/ced-enc-fixed/longdocs/128-256/hi}{+0.04620}
\defval{kvseg/x40/1024/ced-enc-fixed/longdocs/128-256/last}{1.0907}
\defval{kvseg/x40/1024/ced-enc-fixed/longdocs/128-256/lo}{+0.04305}
\defval{kvseg/x40/1024/ced-enc-fixed/longdocs/128-256/round}{1.0461}
\defval{kvseg/x40/1024/ced-enc-fixed/longdocs/1920-2048/delta}{+0.04246}
\defval{kvseg/x40/1024/ced-enc-fixed/longdocs/1920-2048/hi}{+0.04371}
\defval{kvseg/x40/1024/ced-enc-fixed/longdocs/1920-2048/last}{0.9994}
\defval{kvseg/x40/1024/ced-enc-fixed/longdocs/1920-2048/lo}{+0.04111}
\defval{kvseg/x40/1024/ced-enc-fixed/longdocs/1920-2048/round}{0.9570}
\defval{kvseg/x40/1024/ced-enc-fixed/longdocs/384-512/delta}{+0.04259}
\defval{kvseg/x40/1024/ced-enc-fixed/longdocs/384-512/hi}{+0.04397}
\defval{kvseg/x40/1024/ced-enc-fixed/longdocs/384-512/last}{1.0423}
\defval{kvseg/x40/1024/ced-enc-fixed/longdocs/384-512/lo}{+0.04129}
\defval{kvseg/x40/1024/ced-enc-fixed/longdocs/384-512/round}{0.9997}
\defval{kvseg/x40/1024/ced-enc-fixed/longdocs/896-1024/delta}{+0.04308}
\defval{kvseg/x40/1024/ced-enc-fixed/longdocs/896-1024/hi}{+0.04459}
\defval{kvseg/x40/1024/ced-enc-fixed/longdocs/896-1024/last}{1.0273}
\defval{kvseg/x40/1024/ced-enc-fixed/longdocs/896-1024/lo}{+0.04160}
\defval{kvseg/x40/1024/ced-enc-fixed/longdocs/896-1024/round}{0.9842}
\defval{kvseg/x40/1024/ced-enc-fixed/longdocs/all/delta}{+0.04470}
\defval{kvseg/x40/1024/ced-enc-fixed/longdocs/all/hi}{+0.04571}
\defval{kvseg/x40/1024/ced-enc-fixed/longdocs/all/last}{1.0859}
\defval{kvseg/x40/1024/ced-enc-fixed/longdocs/all/lo}{+0.04374}
\defval{kvseg/x40/1024/ced-enc-fixed/longdocs/all/round}{1.0412}
\defval{kvseg/x40/1024/ced-enc-fixed/pg19-test/1-128/delta}{+0.05513}
\defval{kvseg/x40/1024/ced-enc-fixed/pg19-test/1-128/hi}{+0.05878}
\defval{kvseg/x40/1024/ced-enc-fixed/pg19-test/1-128/last}{1.5518}
\defval{kvseg/x40/1024/ced-enc-fixed/pg19-test/1-128/lo}{+0.05185}
\defval{kvseg/x40/1024/ced-enc-fixed/pg19-test/1-128/round}{1.4967}
\defval{kvseg/x40/1024/ced-enc-fixed/pg19-test/128-256/delta}{+0.05008}
\defval{kvseg/x40/1024/ced-enc-fixed/pg19-test/128-256/hi}{+0.05304}
\defval{kvseg/x40/1024/ced-enc-fixed/pg19-test/128-256/last}{1.3735}
\defval{kvseg/x40/1024/ced-enc-fixed/pg19-test/128-256/lo}{+0.04709}
\defval{kvseg/x40/1024/ced-enc-fixed/pg19-test/128-256/round}{1.3234}
\defval{kvseg/x40/1024/ced-enc-fixed/pg19-test/1920-2048/delta}{+0.04906}
\defval{kvseg/x40/1024/ced-enc-fixed/pg19-test/1920-2048/hi}{+0.05181}
\defval{kvseg/x40/1024/ced-enc-fixed/pg19-test/1920-2048/last}{1.3060}
\defval{kvseg/x40/1024/ced-enc-fixed/pg19-test/1920-2048/lo}{+0.04645}
\defval{kvseg/x40/1024/ced-enc-fixed/pg19-test/1920-2048/round}{1.2570}
\defval{kvseg/x40/1024/ced-enc-fixed/pg19-test/384-512/delta}{+0.04914}
\defval{kvseg/x40/1024/ced-enc-fixed/pg19-test/384-512/hi}{+0.05168}
\defval{kvseg/x40/1024/ced-enc-fixed/pg19-test/384-512/last}{1.3372}
\defval{kvseg/x40/1024/ced-enc-fixed/pg19-test/384-512/lo}{+0.04652}
\defval{kvseg/x40/1024/ced-enc-fixed/pg19-test/384-512/round}{1.2881}
\defval{kvseg/x40/1024/ced-enc-fixed/pg19-test/896-1024/delta}{+0.04979}
\defval{kvseg/x40/1024/ced-enc-fixed/pg19-test/896-1024/hi}{+0.05208}
\defval{kvseg/x40/1024/ced-enc-fixed/pg19-test/896-1024/last}{1.3306}
\defval{kvseg/x40/1024/ced-enc-fixed/pg19-test/896-1024/lo}{+0.04730}
\defval{kvseg/x40/1024/ced-enc-fixed/pg19-test/896-1024/round}{1.2808}
\defval{kvseg/x40/1024/ced-enc-fixed/pg19-test/all/delta}{+0.05063}
\defval{kvseg/x40/1024/ced-enc-fixed/pg19-test/all/hi}{+0.05248}
\defval{kvseg/x40/1024/ced-enc-fixed/pg19-test/all/last}{1.3794}
\defval{kvseg/x40/1024/ced-enc-fixed/pg19-test/all/lo}{+0.04882}
\defval{kvseg/x40/1024/ced-enc-fixed/pg19-test/all/round}{1.3288}
\defval{kvseg/x40/1024/ced-enc-fixed/pg19-val/1-128/delta}{+0.04740}
\defval{kvseg/x40/1024/ced-enc-fixed/pg19-val/1-128/hi}{+0.05043}
\defval{kvseg/x40/1024/ced-enc-fixed/pg19-val/1-128/last}{1.4903}
\defval{kvseg/x40/1024/ced-enc-fixed/pg19-val/1-128/lo}{+0.04457}
\defval{kvseg/x40/1024/ced-enc-fixed/pg19-val/1-128/round}{1.4429}
\defval{kvseg/x40/1024/ced-enc-fixed/pg19-val/128-256/delta}{+0.04767}
\defval{kvseg/x40/1024/ced-enc-fixed/pg19-val/128-256/hi}{+0.05014}
\defval{kvseg/x40/1024/ced-enc-fixed/pg19-val/128-256/last}{1.3324}
\defval{kvseg/x40/1024/ced-enc-fixed/pg19-val/128-256/lo}{+0.04524}
\defval{kvseg/x40/1024/ced-enc-fixed/pg19-val/128-256/round}{1.2847}
\defval{kvseg/x40/1024/ced-enc-fixed/pg19-val/1920-2048/delta}{+0.04376}
\defval{kvseg/x40/1024/ced-enc-fixed/pg19-val/1920-2048/hi}{+0.04635}
\defval{kvseg/x40/1024/ced-enc-fixed/pg19-val/1920-2048/last}{1.2591}
\defval{kvseg/x40/1024/ced-enc-fixed/pg19-val/1920-2048/lo}{+0.04151}
\defval{kvseg/x40/1024/ced-enc-fixed/pg19-val/1920-2048/round}{1.2153}
\defval{kvseg/x40/1024/ced-enc-fixed/pg19-val/384-512/delta}{+0.04576}
\defval{kvseg/x40/1024/ced-enc-fixed/pg19-val/384-512/hi}{+0.04803}
\defval{kvseg/x40/1024/ced-enc-fixed/pg19-val/384-512/last}{1.3022}
\defval{kvseg/x40/1024/ced-enc-fixed/pg19-val/384-512/lo}{+0.04334}
\defval{kvseg/x40/1024/ced-enc-fixed/pg19-val/384-512/round}{1.2565}
\defval{kvseg/x40/1024/ced-enc-fixed/pg19-val/896-1024/delta}{+0.04508}
\defval{kvseg/x40/1024/ced-enc-fixed/pg19-val/896-1024/hi}{+0.04738}
\defval{kvseg/x40/1024/ced-enc-fixed/pg19-val/896-1024/last}{1.2784}
\defval{kvseg/x40/1024/ced-enc-fixed/pg19-val/896-1024/lo}{+0.04260}
\defval{kvseg/x40/1024/ced-enc-fixed/pg19-val/896-1024/round}{1.2333}
\defval{kvseg/x40/1024/ced-enc-fixed/pg19-val/all/delta}{+0.04594}
\defval{kvseg/x40/1024/ced-enc-fixed/pg19-val/all/hi}{+0.04741}
\defval{kvseg/x40/1024/ced-enc-fixed/pg19-val/all/last}{1.3324}
\defval{kvseg/x40/1024/ced-enc-fixed/pg19-val/all/lo}{+0.04449}
\defval{kvseg/x40/1024/ced-enc-fixed/pg19-val/all/round}{1.2864}
\defval{kvseg/x40/1024/ced-enc/finepdfs-long/1-128/delta}{$-$0.00036}
\defval{kvseg/x40/1024/ced-enc/finepdfs-long/1-128/hi}{+0.00008}
\defval{kvseg/x40/1024/ced-enc/finepdfs-long/1-128/last}{1.3984}
\defval{kvseg/x40/1024/ced-enc/finepdfs-long/1-128/lo}{$-$0.00080}
\defval{kvseg/x40/1024/ced-enc/finepdfs-long/1-128/round}{1.3987}
\defval{kvseg/x40/1024/ced-enc/finepdfs-long/128-256/delta}{$-$0.00034}
\defval{kvseg/x40/1024/ced-enc/finepdfs-long/128-256/hi}{+0.00005}
\defval{kvseg/x40/1024/ced-enc/finepdfs-long/128-256/last}{1.1656}
\defval{kvseg/x40/1024/ced-enc/finepdfs-long/128-256/lo}{$-$0.00075}
\defval{kvseg/x40/1024/ced-enc/finepdfs-long/128-256/round}{1.1660}
\defval{kvseg/x40/1024/ced-enc/finepdfs-long/1920-2048/delta}{$-$0.00065}
\defval{kvseg/x40/1024/ced-enc/finepdfs-long/1920-2048/hi}{$-$0.00020}
\defval{kvseg/x40/1024/ced-enc/finepdfs-long/1920-2048/last}{1.0260}
\defval{kvseg/x40/1024/ced-enc/finepdfs-long/1920-2048/lo}{$-$0.00115}
\defval{kvseg/x40/1024/ced-enc/finepdfs-long/1920-2048/round}{1.0266}
\defval{kvseg/x40/1024/ced-enc/finepdfs-long/384-512/delta}{$-$0.00080}
\defval{kvseg/x40/1024/ced-enc/finepdfs-long/384-512/hi}{$-$0.00037}
\defval{kvseg/x40/1024/ced-enc/finepdfs-long/384-512/last}{1.1004}
\defval{kvseg/x40/1024/ced-enc/finepdfs-long/384-512/lo}{$-$0.00122}
\defval{kvseg/x40/1024/ced-enc/finepdfs-long/384-512/round}{1.1012}
\defval{kvseg/x40/1024/ced-enc/finepdfs-long/896-1024/delta}{$-$0.00062}
\defval{kvseg/x40/1024/ced-enc/finepdfs-long/896-1024/hi}{$-$0.00027}
\defval{kvseg/x40/1024/ced-enc/finepdfs-long/896-1024/last}{1.0814}
\defval{kvseg/x40/1024/ced-enc/finepdfs-long/896-1024/lo}{$-$0.00100}
\defval{kvseg/x40/1024/ced-enc/finepdfs-long/896-1024/round}{1.0820}
\defval{kvseg/x40/1024/ced-enc/finepdfs-long/all/delta}{$-$0.00055}
\defval{kvseg/x40/1024/ced-enc/finepdfs-long/all/hi}{$-$0.00036}
\defval{kvseg/x40/1024/ced-enc/finepdfs-long/all/last}{1.1550}
\defval{kvseg/x40/1024/ced-enc/finepdfs-long/all/lo}{$-$0.00075}
\defval{kvseg/x40/1024/ced-enc/finepdfs-long/all/round}{1.1556}
\defval{kvseg/x40/1024/ced-enc/fineweb-edu-100bt/1-128/delta}{+0.00005}
\defval{kvseg/x40/1024/ced-enc/fineweb-edu-100bt/1-128/hi}{+0.00037}
\defval{kvseg/x40/1024/ced-enc/fineweb-edu-100bt/1-128/last}{0.9594}
\defval{kvseg/x40/1024/ced-enc/fineweb-edu-100bt/1-128/lo}{$-$0.00028}
\defval{kvseg/x40/1024/ced-enc/fineweb-edu-100bt/1-128/round}{0.9594}
\defval{kvseg/x40/1024/ced-enc/fineweb-edu-100bt/128-256/delta}{$-$0.00012}
\defval{kvseg/x40/1024/ced-enc/fineweb-edu-100bt/128-256/hi}{+0.00016}
\defval{kvseg/x40/1024/ced-enc/fineweb-edu-100bt/128-256/last}{0.8570}
\defval{kvseg/x40/1024/ced-enc/fineweb-edu-100bt/128-256/lo}{$-$0.00037}
\defval{kvseg/x40/1024/ced-enc/fineweb-edu-100bt/128-256/round}{0.8571}
\defval{kvseg/x40/1024/ced-enc/fineweb-edu-100bt/1920-2048/delta}{$-$0.00032}
\defval{kvseg/x40/1024/ced-enc/fineweb-edu-100bt/1920-2048/hi}{$-$0.00001}
\defval{kvseg/x40/1024/ced-enc/fineweb-edu-100bt/1920-2048/last}{0.8348}
\defval{kvseg/x40/1024/ced-enc/fineweb-edu-100bt/1920-2048/lo}{$-$0.00061}
\defval{kvseg/x40/1024/ced-enc/fineweb-edu-100bt/1920-2048/round}{0.8351}
\defval{kvseg/x40/1024/ced-enc/fineweb-edu-100bt/384-512/delta}{$-$0.00036}
\defval{kvseg/x40/1024/ced-enc/fineweb-edu-100bt/384-512/hi}{$-$0.00011}
\defval{kvseg/x40/1024/ced-enc/fineweb-edu-100bt/384-512/last}{0.8409}
\defval{kvseg/x40/1024/ced-enc/fineweb-edu-100bt/384-512/lo}{$-$0.00061}
\defval{kvseg/x40/1024/ced-enc/fineweb-edu-100bt/384-512/round}{0.8413}
\defval{kvseg/x40/1024/ced-enc/fineweb-edu-100bt/896-1024/delta}{$-$0.00024}
\defval{kvseg/x40/1024/ced-enc/fineweb-edu-100bt/896-1024/hi}{+0.00001}
\defval{kvseg/x40/1024/ced-enc/fineweb-edu-100bt/896-1024/last}{0.8376}
\defval{kvseg/x40/1024/ced-enc/fineweb-edu-100bt/896-1024/lo}{$-$0.00050}
\defval{kvseg/x40/1024/ced-enc/fineweb-edu-100bt/896-1024/round}{0.8378}
\defval{kvseg/x40/1024/ced-enc/fineweb-edu-100bt/all/delta}{$-$0.00020}
\defval{kvseg/x40/1024/ced-enc/fineweb-edu-100bt/all/hi}{$-$0.00007}
\defval{kvseg/x40/1024/ced-enc/fineweb-edu-100bt/all/last}{0.8657}
\defval{kvseg/x40/1024/ced-enc/fineweb-edu-100bt/all/lo}{$-$0.00033}
\defval{kvseg/x40/1024/ced-enc/fineweb-edu-100bt/all/round}{0.8659}
\defval{kvseg/x40/1024/ced-enc/govreport-test/1-128/delta}{$-$0.00004}
\defval{kvseg/x40/1024/ced-enc/govreport-test/1-128/hi}{+0.00020}
\defval{kvseg/x40/1024/ced-enc/govreport-test/1-128/last}{0.8602}
\defval{kvseg/x40/1024/ced-enc/govreport-test/1-128/lo}{$-$0.00028}
\defval{kvseg/x40/1024/ced-enc/govreport-test/1-128/round}{0.8603}
\defval{kvseg/x40/1024/ced-enc/govreport-test/128-256/delta}{$-$0.00009}
\defval{kvseg/x40/1024/ced-enc/govreport-test/128-256/hi}{+0.00014}
\defval{kvseg/x40/1024/ced-enc/govreport-test/128-256/last}{0.7318}
\defval{kvseg/x40/1024/ced-enc/govreport-test/128-256/lo}{$-$0.00032}
\defval{kvseg/x40/1024/ced-enc/govreport-test/128-256/round}{0.7319}
\defval{kvseg/x40/1024/ced-enc/govreport-test/1920-2048/delta}{$-$0.00011}
\defval{kvseg/x40/1024/ced-enc/govreport-test/1920-2048/hi}{+0.00011}
\defval{kvseg/x40/1024/ced-enc/govreport-test/1920-2048/last}{0.6660}
\defval{kvseg/x40/1024/ced-enc/govreport-test/1920-2048/lo}{$-$0.00036}
\defval{kvseg/x40/1024/ced-enc/govreport-test/1920-2048/round}{0.6661}
\defval{kvseg/x40/1024/ced-enc/govreport-test/384-512/delta}{$-$0.00027}
\defval{kvseg/x40/1024/ced-enc/govreport-test/384-512/hi}{$-$0.00003}
\defval{kvseg/x40/1024/ced-enc/govreport-test/384-512/last}{0.6934}
\defval{kvseg/x40/1024/ced-enc/govreport-test/384-512/lo}{$-$0.00052}
\defval{kvseg/x40/1024/ced-enc/govreport-test/384-512/round}{0.6937}
\defval{kvseg/x40/1024/ced-enc/govreport-test/896-1024/delta}{$-$0.00033}
\defval{kvseg/x40/1024/ced-enc/govreport-test/896-1024/hi}{$-$0.00012}
\defval{kvseg/x40/1024/ced-enc/govreport-test/896-1024/last}{0.6762}
\defval{kvseg/x40/1024/ced-enc/govreport-test/896-1024/lo}{$-$0.00053}
\defval{kvseg/x40/1024/ced-enc/govreport-test/896-1024/round}{0.6765}
\defval{kvseg/x40/1024/ced-enc/govreport-test/all/delta}{$-$0.00017}
\defval{kvseg/x40/1024/ced-enc/govreport-test/all/hi}{$-$0.00006}
\defval{kvseg/x40/1024/ced-enc/govreport-test/all/last}{0.7255}
\defval{kvseg/x40/1024/ced-enc/govreport-test/all/lo}{$-$0.00028}
\defval{kvseg/x40/1024/ced-enc/govreport-test/all/round}{0.7257}
\defval{kvseg/x40/1024/ced-enc/longdocs/1-128/delta}{$-$0.00056}
\defval{kvseg/x40/1024/ced-enc/longdocs/1-128/hi}{$-$0.00035}
\defval{kvseg/x40/1024/ced-enc/longdocs/1-128/last}{1.2124}
\defval{kvseg/x40/1024/ced-enc/longdocs/1-128/lo}{$-$0.00077}
\defval{kvseg/x40/1024/ced-enc/longdocs/1-128/round}{1.2130}
\defval{kvseg/x40/1024/ced-enc/longdocs/128-256/delta}{$-$0.00050}
\defval{kvseg/x40/1024/ced-enc/longdocs/128-256/hi}{$-$0.00029}
\defval{kvseg/x40/1024/ced-enc/longdocs/128-256/last}{1.0411}
\defval{kvseg/x40/1024/ced-enc/longdocs/128-256/lo}{$-$0.00070}
\defval{kvseg/x40/1024/ced-enc/longdocs/128-256/round}{1.0416}
\defval{kvseg/x40/1024/ced-enc/longdocs/1920-2048/delta}{$-$0.00055}
\defval{kvseg/x40/1024/ced-enc/longdocs/1920-2048/hi}{$-$0.00034}
\defval{kvseg/x40/1024/ced-enc/longdocs/1920-2048/last}{0.9519}
\defval{kvseg/x40/1024/ced-enc/longdocs/1920-2048/lo}{$-$0.00078}
\defval{kvseg/x40/1024/ced-enc/longdocs/1920-2048/round}{0.9524}
\defval{kvseg/x40/1024/ced-enc/longdocs/384-512/delta}{$-$0.00074}
\defval{kvseg/x40/1024/ced-enc/longdocs/384-512/hi}{$-$0.00053}
\defval{kvseg/x40/1024/ced-enc/longdocs/384-512/last}{0.9952}
\defval{kvseg/x40/1024/ced-enc/longdocs/384-512/lo}{$-$0.00095}
\defval{kvseg/x40/1024/ced-enc/longdocs/384-512/round}{0.9960}
\defval{kvseg/x40/1024/ced-enc/longdocs/896-1024/delta}{$-$0.00060}
\defval{kvseg/x40/1024/ced-enc/longdocs/896-1024/hi}{$-$0.00042}
\defval{kvseg/x40/1024/ced-enc/longdocs/896-1024/last}{0.9803}
\defval{kvseg/x40/1024/ced-enc/longdocs/896-1024/lo}{$-$0.00080}
\defval{kvseg/x40/1024/ced-enc/longdocs/896-1024/round}{0.9809}
\defval{kvseg/x40/1024/ced-enc/longdocs/all/delta}{$-$0.00059}
\defval{kvseg/x40/1024/ced-enc/longdocs/all/hi}{$-$0.00049}
\defval{kvseg/x40/1024/ced-enc/longdocs/all/last}{1.0362}
\defval{kvseg/x40/1024/ced-enc/longdocs/all/lo}{$-$0.00069}
\defval{kvseg/x40/1024/ced-enc/longdocs/all/round}{1.0368}
\defval{kvseg/x40/1024/ced-enc/pg19-test/1-128/delta}{$-$0.00143}
\defval{kvseg/x40/1024/ced-enc/pg19-test/1-128/hi}{$-$0.00099}
\defval{kvseg/x40/1024/ced-enc/pg19-test/1-128/last}{1.4883}
\defval{kvseg/x40/1024/ced-enc/pg19-test/1-128/lo}{$-$0.00188}
\defval{kvseg/x40/1024/ced-enc/pg19-test/1-128/round}{1.4897}
\defval{kvseg/x40/1024/ced-enc/pg19-test/128-256/delta}{$-$0.00117}
\defval{kvseg/x40/1024/ced-enc/pg19-test/128-256/hi}{$-$0.00075}
\defval{kvseg/x40/1024/ced-enc/pg19-test/128-256/last}{1.3171}
\defval{kvseg/x40/1024/ced-enc/pg19-test/128-256/lo}{$-$0.00158}
\defval{kvseg/x40/1024/ced-enc/pg19-test/128-256/round}{1.3183}
\defval{kvseg/x40/1024/ced-enc/pg19-test/1920-2048/delta}{$-$0.00103}
\defval{kvseg/x40/1024/ced-enc/pg19-test/1920-2048/hi}{$-$0.00060}
\defval{kvseg/x40/1024/ced-enc/pg19-test/1920-2048/last}{1.2500}
\defval{kvseg/x40/1024/ced-enc/pg19-test/1920-2048/lo}{$-$0.00148}
\defval{kvseg/x40/1024/ced-enc/pg19-test/1920-2048/round}{1.2510}
\defval{kvseg/x40/1024/ced-enc/pg19-test/384-512/delta}{$-$0.00128}
\defval{kvseg/x40/1024/ced-enc/pg19-test/384-512/hi}{$-$0.00082}
\defval{kvseg/x40/1024/ced-enc/pg19-test/384-512/last}{1.2819}
\defval{kvseg/x40/1024/ced-enc/pg19-test/384-512/lo}{$-$0.00171}
\defval{kvseg/x40/1024/ced-enc/pg19-test/384-512/round}{1.2832}
\defval{kvseg/x40/1024/ced-enc/pg19-test/896-1024/delta}{$-$0.00093}
\defval{kvseg/x40/1024/ced-enc/pg19-test/896-1024/hi}{$-$0.00051}
\defval{kvseg/x40/1024/ced-enc/pg19-test/896-1024/last}{1.2755}
\defval{kvseg/x40/1024/ced-enc/pg19-test/896-1024/lo}{$-$0.00137}
\defval{kvseg/x40/1024/ced-enc/pg19-test/896-1024/round}{1.2765}
\defval{kvseg/x40/1024/ced-enc/pg19-test/all/delta}{$-$0.00117}
\defval{kvseg/x40/1024/ced-enc/pg19-test/all/hi}{$-$0.00097}
\defval{kvseg/x40/1024/ced-enc/pg19-test/all/last}{1.3222}
\defval{kvseg/x40/1024/ced-enc/pg19-test/all/lo}{$-$0.00138}
\defval{kvseg/x40/1024/ced-enc/pg19-test/all/round}{1.3233}
\defval{kvseg/x40/1024/ced-enc/pg19-val/1-128/delta}{$-$0.00075}
\defval{kvseg/x40/1024/ced-enc/pg19-val/1-128/hi}{$-$0.00033}
\defval{kvseg/x40/1024/ced-enc/pg19-val/1-128/last}{1.4371}
\defval{kvseg/x40/1024/ced-enc/pg19-val/1-128/lo}{$-$0.00112}
\defval{kvseg/x40/1024/ced-enc/pg19-val/1-128/round}{1.4378}
\defval{kvseg/x40/1024/ced-enc/pg19-val/128-256/delta}{$-$0.00081}
\defval{kvseg/x40/1024/ced-enc/pg19-val/128-256/hi}{$-$0.00042}
\defval{kvseg/x40/1024/ced-enc/pg19-val/128-256/last}{1.2797}
\defval{kvseg/x40/1024/ced-enc/pg19-val/128-256/lo}{$-$0.00122}
\defval{kvseg/x40/1024/ced-enc/pg19-val/128-256/round}{1.2806}
\defval{kvseg/x40/1024/ced-enc/pg19-val/1920-2048/delta}{$-$0.00064}
\defval{kvseg/x40/1024/ced-enc/pg19-val/1920-2048/hi}{$-$0.00017}
\defval{kvseg/x40/1024/ced-enc/pg19-val/1920-2048/last}{1.2103}
\defval{kvseg/x40/1024/ced-enc/pg19-val/1920-2048/lo}{$-$0.00106}
\defval{kvseg/x40/1024/ced-enc/pg19-val/1920-2048/round}{1.2109}
\defval{kvseg/x40/1024/ced-enc/pg19-val/384-512/delta}{$-$0.00094}
\defval{kvseg/x40/1024/ced-enc/pg19-val/384-512/hi}{$-$0.00049}
\defval{kvseg/x40/1024/ced-enc/pg19-val/384-512/last}{1.2504}
\defval{kvseg/x40/1024/ced-enc/pg19-val/384-512/lo}{$-$0.00140}
\defval{kvseg/x40/1024/ced-enc/pg19-val/384-512/round}{1.2514}
\defval{kvseg/x40/1024/ced-enc/pg19-val/896-1024/delta}{$-$0.00102}
\defval{kvseg/x40/1024/ced-enc/pg19-val/896-1024/hi}{$-$0.00061}
\defval{kvseg/x40/1024/ced-enc/pg19-val/896-1024/last}{1.2277}
\defval{kvseg/x40/1024/ced-enc/pg19-val/896-1024/lo}{$-$0.00144}
\defval{kvseg/x40/1024/ced-enc/pg19-val/896-1024/round}{1.2287}
\defval{kvseg/x40/1024/ced-enc/pg19-val/all/delta}{$-$0.00083}
\defval{kvseg/x40/1024/ced-enc/pg19-val/all/hi}{$-$0.00062}
\defval{kvseg/x40/1024/ced-enc/pg19-val/all/last}{1.2809}
\defval{kvseg/x40/1024/ced-enc/pg19-val/all/lo}{$-$0.00103}
\defval{kvseg/x40/1024/ced-enc/pg19-val/all/round}{1.2818}
\defval{kvseg/x40/1024/dense-l25/finepdfs-long/1-128/delta}{$-$0.00048}
\defval{kvseg/x40/1024/dense-l25/finepdfs-long/1-128/hi}{$-$0.00009}
\defval{kvseg/x40/1024/dense-l25/finepdfs-long/1-128/last}{1.3969}
\defval{kvseg/x40/1024/dense-l25/finepdfs-long/1-128/lo}{$-$0.00085}
\defval{kvseg/x40/1024/dense-l25/finepdfs-long/1-128/round}{1.3974}
\defval{kvseg/x40/1024/dense-l25/finepdfs-long/128-256/delta}{$-$0.00004}
\defval{kvseg/x40/1024/dense-l25/finepdfs-long/128-256/hi}{+0.00034}
\defval{kvseg/x40/1024/dense-l25/finepdfs-long/128-256/last}{1.1657}
\defval{kvseg/x40/1024/dense-l25/finepdfs-long/128-256/lo}{$-$0.00044}
\defval{kvseg/x40/1024/dense-l25/finepdfs-long/128-256/round}{1.1658}
\defval{kvseg/x40/1024/dense-l25/finepdfs-long/1920-2048/delta}{+0.00070}
\defval{kvseg/x40/1024/dense-l25/finepdfs-long/1920-2048/hi}{+0.00113}
\defval{kvseg/x40/1024/dense-l25/finepdfs-long/1920-2048/last}{1.0286}
\defval{kvseg/x40/1024/dense-l25/finepdfs-long/1920-2048/lo}{+0.00026}
\defval{kvseg/x40/1024/dense-l25/finepdfs-long/1920-2048/round}{1.0279}
\defval{kvseg/x40/1024/dense-l25/finepdfs-long/384-512/delta}{+0.00052}
\defval{kvseg/x40/1024/dense-l25/finepdfs-long/384-512/hi}{+0.00097}
\defval{kvseg/x40/1024/dense-l25/finepdfs-long/384-512/last}{1.1055}
\defval{kvseg/x40/1024/dense-l25/finepdfs-long/384-512/lo}{+0.00008}
\defval{kvseg/x40/1024/dense-l25/finepdfs-long/384-512/round}{1.1050}
\defval{kvseg/x40/1024/dense-l25/finepdfs-long/896-1024/delta}{+0.00063}
\defval{kvseg/x40/1024/dense-l25/finepdfs-long/896-1024/hi}{+0.00110}
\defval{kvseg/x40/1024/dense-l25/finepdfs-long/896-1024/last}{1.0852}
\defval{kvseg/x40/1024/dense-l25/finepdfs-long/896-1024/lo}{+0.00019}
\defval{kvseg/x40/1024/dense-l25/finepdfs-long/896-1024/round}{1.0845}
\defval{kvseg/x40/1024/dense-l25/finepdfs-long/all/delta}{+0.00026}
\defval{kvseg/x40/1024/dense-l25/finepdfs-long/all/hi}{+0.00046}
\defval{kvseg/x40/1024/dense-l25/finepdfs-long/all/last}{1.1570}
\defval{kvseg/x40/1024/dense-l25/finepdfs-long/all/lo}{+0.00006}
\defval{kvseg/x40/1024/dense-l25/finepdfs-long/all/round}{1.1568}
\defval{kvseg/x40/1024/dense-l25/fineweb-edu-100bt/1-128/delta}{+0.00018}
\defval{kvseg/x40/1024/dense-l25/fineweb-edu-100bt/1-128/hi}{+0.00048}
\defval{kvseg/x40/1024/dense-l25/fineweb-edu-100bt/1-128/last}{0.9627}
\defval{kvseg/x40/1024/dense-l25/fineweb-edu-100bt/1-128/lo}{$-$0.00015}
\defval{kvseg/x40/1024/dense-l25/fineweb-edu-100bt/1-128/round}{0.9625}
\defval{kvseg/x40/1024/dense-l25/fineweb-edu-100bt/128-256/delta}{+0.00029}
\defval{kvseg/x40/1024/dense-l25/fineweb-edu-100bt/128-256/hi}{+0.00056}
\defval{kvseg/x40/1024/dense-l25/fineweb-edu-100bt/128-256/last}{0.8576}
\defval{kvseg/x40/1024/dense-l25/fineweb-edu-100bt/128-256/lo}{+0.00002}
\defval{kvseg/x40/1024/dense-l25/fineweb-edu-100bt/128-256/round}{0.8574}
\defval{kvseg/x40/1024/dense-l25/fineweb-edu-100bt/1920-2048/delta}{+0.00022}
\defval{kvseg/x40/1024/dense-l25/fineweb-edu-100bt/1920-2048/hi}{+0.00053}
\defval{kvseg/x40/1024/dense-l25/fineweb-edu-100bt/1920-2048/last}{0.8333}
\defval{kvseg/x40/1024/dense-l25/fineweb-edu-100bt/1920-2048/lo}{$-$0.00015}
\defval{kvseg/x40/1024/dense-l25/fineweb-edu-100bt/1920-2048/round}{0.8330}
\defval{kvseg/x40/1024/dense-l25/fineweb-edu-100bt/384-512/delta}{+0.00014}
\defval{kvseg/x40/1024/dense-l25/fineweb-edu-100bt/384-512/hi}{+0.00041}
\defval{kvseg/x40/1024/dense-l25/fineweb-edu-100bt/384-512/last}{0.8401}
\defval{kvseg/x40/1024/dense-l25/fineweb-edu-100bt/384-512/lo}{$-$0.00016}
\defval{kvseg/x40/1024/dense-l25/fineweb-edu-100bt/384-512/round}{0.8400}
\defval{kvseg/x40/1024/dense-l25/fineweb-edu-100bt/896-1024/delta}{+0.00029}
\defval{kvseg/x40/1024/dense-l25/fineweb-edu-100bt/896-1024/hi}{+0.00060}
\defval{kvseg/x40/1024/dense-l25/fineweb-edu-100bt/896-1024/last}{0.8357}
\defval{kvseg/x40/1024/dense-l25/fineweb-edu-100bt/896-1024/lo}{$-$0.00002}
\defval{kvseg/x40/1024/dense-l25/fineweb-edu-100bt/896-1024/round}{0.8354}
\defval{kvseg/x40/1024/dense-l25/fineweb-edu-100bt/all/delta}{+0.00022}
\defval{kvseg/x40/1024/dense-l25/fineweb-edu-100bt/all/hi}{+0.00035}
\defval{kvseg/x40/1024/dense-l25/fineweb-edu-100bt/all/last}{0.8657}
\defval{kvseg/x40/1024/dense-l25/fineweb-edu-100bt/all/lo}{+0.00008}
\defval{kvseg/x40/1024/dense-l25/fineweb-edu-100bt/all/round}{0.8655}
\defval{kvseg/x40/1024/dense-l25/govreport-test/1-128/delta}{+0.00020}
\defval{kvseg/x40/1024/dense-l25/govreport-test/1-128/hi}{+0.00047}
\defval{kvseg/x40/1024/dense-l25/govreport-test/1-128/last}{0.8598}
\defval{kvseg/x40/1024/dense-l25/govreport-test/1-128/lo}{$-$0.00004}
\defval{kvseg/x40/1024/dense-l25/govreport-test/1-128/round}{0.8596}
\defval{kvseg/x40/1024/dense-l25/govreport-test/128-256/delta}{+0.00029}
\defval{kvseg/x40/1024/dense-l25/govreport-test/128-256/hi}{+0.00057}
\defval{kvseg/x40/1024/dense-l25/govreport-test/128-256/last}{0.7313}
\defval{kvseg/x40/1024/dense-l25/govreport-test/128-256/lo}{+0.00003}
\defval{kvseg/x40/1024/dense-l25/govreport-test/128-256/round}{0.7310}
\defval{kvseg/x40/1024/dense-l25/govreport-test/1920-2048/delta}{+0.00025}
\defval{kvseg/x40/1024/dense-l25/govreport-test/1920-2048/hi}{+0.00054}
\defval{kvseg/x40/1024/dense-l25/govreport-test/1920-2048/last}{0.6624}
\defval{kvseg/x40/1024/dense-l25/govreport-test/1920-2048/lo}{$-$0.00005}
\defval{kvseg/x40/1024/dense-l25/govreport-test/1920-2048/round}{0.6622}
\defval{kvseg/x40/1024/dense-l25/govreport-test/384-512/delta}{+0.00031}
\defval{kvseg/x40/1024/dense-l25/govreport-test/384-512/hi}{+0.00053}
\defval{kvseg/x40/1024/dense-l25/govreport-test/384-512/last}{0.6912}
\defval{kvseg/x40/1024/dense-l25/govreport-test/384-512/lo}{+0.00008}
\defval{kvseg/x40/1024/dense-l25/govreport-test/384-512/round}{0.6909}
\defval{kvseg/x40/1024/dense-l25/govreport-test/896-1024/delta}{+0.00006}
\defval{kvseg/x40/1024/dense-l25/govreport-test/896-1024/hi}{+0.00035}
\defval{kvseg/x40/1024/dense-l25/govreport-test/896-1024/last}{0.6730}
\defval{kvseg/x40/1024/dense-l25/govreport-test/896-1024/lo}{$-$0.00022}
\defval{kvseg/x40/1024/dense-l25/govreport-test/896-1024/round}{0.6730}
\defval{kvseg/x40/1024/dense-l25/govreport-test/all/delta}{+0.00022}
\defval{kvseg/x40/1024/dense-l25/govreport-test/all/hi}{+0.00034}
\defval{kvseg/x40/1024/dense-l25/govreport-test/all/last}{0.7235}
\defval{kvseg/x40/1024/dense-l25/govreport-test/all/lo}{+0.00009}
\defval{kvseg/x40/1024/dense-l25/govreport-test/all/round}{0.7233}
\defval{kvseg/x40/1024/dense-l25/longdocs/1-128/delta}{$-$0.00004}
\defval{kvseg/x40/1024/dense-l25/longdocs/1-128/hi}{+0.00014}
\defval{kvseg/x40/1024/dense-l25/longdocs/1-128/last}{1.2112}
\defval{kvseg/x40/1024/dense-l25/longdocs/1-128/lo}{$-$0.00025}
\defval{kvseg/x40/1024/dense-l25/longdocs/1-128/round}{1.2112}
\defval{kvseg/x40/1024/dense-l25/longdocs/128-256/delta}{+0.00018}
\defval{kvseg/x40/1024/dense-l25/longdocs/128-256/hi}{+0.00038}
\defval{kvseg/x40/1024/dense-l25/longdocs/128-256/last}{1.0408}
\defval{kvseg/x40/1024/dense-l25/longdocs/128-256/lo}{$-$0.00001}
\defval{kvseg/x40/1024/dense-l25/longdocs/128-256/round}{1.0406}
\defval{kvseg/x40/1024/dense-l25/longdocs/1920-2048/delta}{+0.00037}
\defval{kvseg/x40/1024/dense-l25/longdocs/1920-2048/hi}{+0.00061}
\defval{kvseg/x40/1024/dense-l25/longdocs/1920-2048/last}{0.9512}
\defval{kvseg/x40/1024/dense-l25/longdocs/1920-2048/lo}{+0.00014}
\defval{kvseg/x40/1024/dense-l25/longdocs/1920-2048/round}{0.9508}
\defval{kvseg/x40/1024/dense-l25/longdocs/384-512/delta}{+0.00041}
\defval{kvseg/x40/1024/dense-l25/longdocs/384-512/hi}{+0.00063}
\defval{kvseg/x40/1024/dense-l25/longdocs/384-512/last}{0.9953}
\defval{kvseg/x40/1024/dense-l25/longdocs/384-512/lo}{+0.00020}
\defval{kvseg/x40/1024/dense-l25/longdocs/384-512/round}{0.9949}
\defval{kvseg/x40/1024/dense-l25/longdocs/896-1024/delta}{+0.00027}
\defval{kvseg/x40/1024/dense-l25/longdocs/896-1024/hi}{+0.00049}
\defval{kvseg/x40/1024/dense-l25/longdocs/896-1024/last}{0.9805}
\defval{kvseg/x40/1024/dense-l25/longdocs/896-1024/lo}{+0.00004}
\defval{kvseg/x40/1024/dense-l25/longdocs/896-1024/round}{0.9802}
\defval{kvseg/x40/1024/dense-l25/longdocs/all/delta}{+0.00024}
\defval{kvseg/x40/1024/dense-l25/longdocs/all/hi}{+0.00033}
\defval{kvseg/x40/1024/dense-l25/longdocs/all/last}{1.0358}
\defval{kvseg/x40/1024/dense-l25/longdocs/all/lo}{+0.00014}
\defval{kvseg/x40/1024/dense-l25/longdocs/all/round}{1.0356}
\defval{kvseg/x40/1024/dense-l25/pg19-test/1-128/delta}{+0.00008}
\defval{kvseg/x40/1024/dense-l25/pg19-test/1-128/hi}{+0.00050}
\defval{kvseg/x40/1024/dense-l25/pg19-test/1-128/last}{1.4863}
\defval{kvseg/x40/1024/dense-l25/pg19-test/1-128/lo}{$-$0.00037}
\defval{kvseg/x40/1024/dense-l25/pg19-test/1-128/round}{1.4862}
\defval{kvseg/x40/1024/dense-l25/pg19-test/128-256/delta}{+0.00027}
\defval{kvseg/x40/1024/dense-l25/pg19-test/128-256/hi}{+0.00066}
\defval{kvseg/x40/1024/dense-l25/pg19-test/128-256/last}{1.3168}
\defval{kvseg/x40/1024/dense-l25/pg19-test/128-256/lo}{$-$0.00013}
\defval{kvseg/x40/1024/dense-l25/pg19-test/128-256/round}{1.3165}
\defval{kvseg/x40/1024/dense-l25/pg19-test/1920-2048/delta}{+0.00022}
\defval{kvseg/x40/1024/dense-l25/pg19-test/1920-2048/hi}{+0.00064}
\defval{kvseg/x40/1024/dense-l25/pg19-test/1920-2048/last}{1.2498}
\defval{kvseg/x40/1024/dense-l25/pg19-test/1920-2048/lo}{$-$0.00025}
\defval{kvseg/x40/1024/dense-l25/pg19-test/1920-2048/round}{1.2496}
\defval{kvseg/x40/1024/dense-l25/pg19-test/384-512/delta}{+0.00044}
\defval{kvseg/x40/1024/dense-l25/pg19-test/384-512/hi}{+0.00092}
\defval{kvseg/x40/1024/dense-l25/pg19-test/384-512/last}{1.2802}
\defval{kvseg/x40/1024/dense-l25/pg19-test/384-512/lo}{+0.00000}
\defval{kvseg/x40/1024/dense-l25/pg19-test/384-512/round}{1.2797}
\defval{kvseg/x40/1024/dense-l25/pg19-test/896-1024/delta}{+0.00019}
\defval{kvseg/x40/1024/dense-l25/pg19-test/896-1024/hi}{+0.00066}
\defval{kvseg/x40/1024/dense-l25/pg19-test/896-1024/last}{1.2765}
\defval{kvseg/x40/1024/dense-l25/pg19-test/896-1024/lo}{$-$0.00026}
\defval{kvseg/x40/1024/dense-l25/pg19-test/896-1024/round}{1.2763}
\defval{kvseg/x40/1024/dense-l25/pg19-test/all/delta}{+0.00024}
\defval{kvseg/x40/1024/dense-l25/pg19-test/all/hi}{+0.00044}
\defval{kvseg/x40/1024/dense-l25/pg19-test/all/last}{1.3215}
\defval{kvseg/x40/1024/dense-l25/pg19-test/all/lo}{+0.00005}
\defval{kvseg/x40/1024/dense-l25/pg19-test/all/round}{1.3213}
\defval{kvseg/x40/1024/dense-l25/pg19-val/1-128/delta}{+0.00005}
\defval{kvseg/x40/1024/dense-l25/pg19-val/1-128/hi}{+0.00053}
\defval{kvseg/x40/1024/dense-l25/pg19-val/1-128/last}{1.4333}
\defval{kvseg/x40/1024/dense-l25/pg19-val/1-128/lo}{$-$0.00036}
\defval{kvseg/x40/1024/dense-l25/pg19-val/1-128/round}{1.4333}
\defval{kvseg/x40/1024/dense-l25/pg19-val/128-256/delta}{+0.00020}
\defval{kvseg/x40/1024/dense-l25/pg19-val/128-256/hi}{+0.00059}
\defval{kvseg/x40/1024/dense-l25/pg19-val/128-256/last}{1.2783}
\defval{kvseg/x40/1024/dense-l25/pg19-val/128-256/lo}{$-$0.00022}
\defval{kvseg/x40/1024/dense-l25/pg19-val/128-256/round}{1.2781}
\defval{kvseg/x40/1024/dense-l25/pg19-val/1920-2048/delta}{+0.00025}
\defval{kvseg/x40/1024/dense-l25/pg19-val/1920-2048/hi}{+0.00074}
\defval{kvseg/x40/1024/dense-l25/pg19-val/1920-2048/last}{1.2094}
\defval{kvseg/x40/1024/dense-l25/pg19-val/1920-2048/lo}{$-$0.00019}
\defval{kvseg/x40/1024/dense-l25/pg19-val/1920-2048/round}{1.2092}
\defval{kvseg/x40/1024/dense-l25/pg19-val/384-512/delta}{+0.00035}
\defval{kvseg/x40/1024/dense-l25/pg19-val/384-512/hi}{+0.00077}
\defval{kvseg/x40/1024/dense-l25/pg19-val/384-512/last}{1.2491}
\defval{kvseg/x40/1024/dense-l25/pg19-val/384-512/lo}{$-$0.00003}
\defval{kvseg/x40/1024/dense-l25/pg19-val/384-512/round}{1.2487}
\defval{kvseg/x40/1024/dense-l25/pg19-val/896-1024/delta}{+0.00071}
\defval{kvseg/x40/1024/dense-l25/pg19-val/896-1024/hi}{+0.00111}
\defval{kvseg/x40/1024/dense-l25/pg19-val/896-1024/last}{1.2258}
\defval{kvseg/x40/1024/dense-l25/pg19-val/896-1024/lo}{+0.00029}
\defval{kvseg/x40/1024/dense-l25/pg19-val/896-1024/round}{1.2250}
\defval{kvseg/x40/1024/dense-l25/pg19-val/all/delta}{+0.00031}
\defval{kvseg/x40/1024/dense-l25/pg19-val/all/hi}{+0.00052}
\defval{kvseg/x40/1024/dense-l25/pg19-val/all/last}{1.2791}
\defval{kvseg/x40/1024/dense-l25/pg19-val/all/lo}{+0.00012}
\defval{kvseg/x40/1024/dense-l25/pg19-val/all/round}{1.2788}
\defval{kvseg/x40/1024/dense-l47/finepdfs-long/1-128/delta}{$-$0.00010}
\defval{kvseg/x40/1024/dense-l47/finepdfs-long/1-128/hi}{+0.00027}
\defval{kvseg/x40/1024/dense-l47/finepdfs-long/1-128/last}{1.4006}
\defval{kvseg/x40/1024/dense-l47/finepdfs-long/1-128/lo}{$-$0.00047}
\defval{kvseg/x40/1024/dense-l47/finepdfs-long/1-128/round}{1.4007}
\defval{kvseg/x40/1024/dense-l47/finepdfs-long/128-256/delta}{$-$0.00036}
\defval{kvseg/x40/1024/dense-l47/finepdfs-long/128-256/hi}{$-$0.00003}
\defval{kvseg/x40/1024/dense-l47/finepdfs-long/128-256/last}{1.1753}
\defval{kvseg/x40/1024/dense-l47/finepdfs-long/128-256/lo}{$-$0.00071}
\defval{kvseg/x40/1024/dense-l47/finepdfs-long/128-256/round}{1.1757}
\defval{kvseg/x40/1024/dense-l47/finepdfs-long/1920-2048/delta}{$-$0.00017}
\defval{kvseg/x40/1024/dense-l47/finepdfs-long/1920-2048/hi}{+0.00015}
\defval{kvseg/x40/1024/dense-l47/finepdfs-long/1920-2048/last}{1.0415}
\defval{kvseg/x40/1024/dense-l47/finepdfs-long/1920-2048/lo}{$-$0.00050}
\defval{kvseg/x40/1024/dense-l47/finepdfs-long/1920-2048/round}{1.0417}
\defval{kvseg/x40/1024/dense-l47/finepdfs-long/384-512/delta}{$-$0.00027}
\defval{kvseg/x40/1024/dense-l47/finepdfs-long/384-512/hi}{+0.00008}
\defval{kvseg/x40/1024/dense-l47/finepdfs-long/384-512/last}{1.1146}
\defval{kvseg/x40/1024/dense-l47/finepdfs-long/384-512/lo}{$-$0.00063}
\defval{kvseg/x40/1024/dense-l47/finepdfs-long/384-512/round}{1.1149}
\defval{kvseg/x40/1024/dense-l47/finepdfs-long/896-1024/delta}{$-$0.00003}
\defval{kvseg/x40/1024/dense-l47/finepdfs-long/896-1024/hi}{+0.00035}
\defval{kvseg/x40/1024/dense-l47/finepdfs-long/896-1024/last}{1.0960}
\defval{kvseg/x40/1024/dense-l47/finepdfs-long/896-1024/lo}{$-$0.00043}
\defval{kvseg/x40/1024/dense-l47/finepdfs-long/896-1024/round}{1.0960}
\defval{kvseg/x40/1024/dense-l47/finepdfs-long/all/delta}{$-$0.00019}
\defval{kvseg/x40/1024/dense-l47/finepdfs-long/all/hi}{$-$0.00003}
\defval{kvseg/x40/1024/dense-l47/finepdfs-long/all/last}{1.1662}
\defval{kvseg/x40/1024/dense-l47/finepdfs-long/all/lo}{$-$0.00035}
\defval{kvseg/x40/1024/dense-l47/finepdfs-long/all/round}{1.1664}
\defval{kvseg/x40/1024/dense-l47/fineweb-edu-100bt/1-128/delta}{$-$0.00008}
\defval{kvseg/x40/1024/dense-l47/fineweb-edu-100bt/1-128/hi}{+0.00032}
\defval{kvseg/x40/1024/dense-l47/fineweb-edu-100bt/1-128/last}{0.9660}
\defval{kvseg/x40/1024/dense-l47/fineweb-edu-100bt/1-128/lo}{$-$0.00048}
\defval{kvseg/x40/1024/dense-l47/fineweb-edu-100bt/1-128/round}{0.9661}
\defval{kvseg/x40/1024/dense-l47/fineweb-edu-100bt/128-256/delta}{+0.00003}
\defval{kvseg/x40/1024/dense-l47/fineweb-edu-100bt/128-256/hi}{+0.00028}
\defval{kvseg/x40/1024/dense-l47/fineweb-edu-100bt/128-256/last}{0.8625}
\defval{kvseg/x40/1024/dense-l47/fineweb-edu-100bt/128-256/lo}{$-$0.00022}
\defval{kvseg/x40/1024/dense-l47/fineweb-edu-100bt/128-256/round}{0.8624}
\defval{kvseg/x40/1024/dense-l47/fineweb-edu-100bt/1920-2048/delta}{+0.00005}
\defval{kvseg/x40/1024/dense-l47/fineweb-edu-100bt/1920-2048/hi}{+0.00033}
\defval{kvseg/x40/1024/dense-l47/fineweb-edu-100bt/1920-2048/last}{0.8388}
\defval{kvseg/x40/1024/dense-l47/fineweb-edu-100bt/1920-2048/lo}{$-$0.00021}
\defval{kvseg/x40/1024/dense-l47/fineweb-edu-100bt/1920-2048/round}{0.8388}
\defval{kvseg/x40/1024/dense-l47/fineweb-edu-100bt/384-512/delta}{+0.00014}
\defval{kvseg/x40/1024/dense-l47/fineweb-edu-100bt/384-512/hi}{+0.00039}
\defval{kvseg/x40/1024/dense-l47/fineweb-edu-100bt/384-512/last}{0.8454}
\defval{kvseg/x40/1024/dense-l47/fineweb-edu-100bt/384-512/lo}{$-$0.00014}
\defval{kvseg/x40/1024/dense-l47/fineweb-edu-100bt/384-512/round}{0.8453}
\defval{kvseg/x40/1024/dense-l47/fineweb-edu-100bt/896-1024/delta}{$-$0.00015}
\defval{kvseg/x40/1024/dense-l47/fineweb-edu-100bt/896-1024/hi}{+0.00013}
\defval{kvseg/x40/1024/dense-l47/fineweb-edu-100bt/896-1024/last}{0.8406}
\defval{kvseg/x40/1024/dense-l47/fineweb-edu-100bt/896-1024/lo}{$-$0.00046}
\defval{kvseg/x40/1024/dense-l47/fineweb-edu-100bt/896-1024/round}{0.8408}
\defval{kvseg/x40/1024/dense-l47/fineweb-edu-100bt/all/delta}{$-$0.00000}
\defval{kvseg/x40/1024/dense-l47/fineweb-edu-100bt/all/hi}{+0.00012}
\defval{kvseg/x40/1024/dense-l47/fineweb-edu-100bt/all/last}{0.8705}
\defval{kvseg/x40/1024/dense-l47/fineweb-edu-100bt/all/lo}{$-$0.00015}
\defval{kvseg/x40/1024/dense-l47/fineweb-edu-100bt/all/round}{0.8705}
\defval{kvseg/x40/1024/dense-l47/govreport-test/1-128/delta}{+0.00003}
\defval{kvseg/x40/1024/dense-l47/govreport-test/1-128/hi}{+0.00026}
\defval{kvseg/x40/1024/dense-l47/govreport-test/1-128/last}{0.8647}
\defval{kvseg/x40/1024/dense-l47/govreport-test/1-128/lo}{$-$0.00017}
\defval{kvseg/x40/1024/dense-l47/govreport-test/1-128/round}{0.8647}
\defval{kvseg/x40/1024/dense-l47/govreport-test/128-256/delta}{+0.00016}
\defval{kvseg/x40/1024/dense-l47/govreport-test/128-256/hi}{+0.00038}
\defval{kvseg/x40/1024/dense-l47/govreport-test/128-256/last}{0.7366}
\defval{kvseg/x40/1024/dense-l47/govreport-test/128-256/lo}{$-$0.00007}
\defval{kvseg/x40/1024/dense-l47/govreport-test/128-256/round}{0.7364}
\defval{kvseg/x40/1024/dense-l47/govreport-test/1920-2048/delta}{+0.00015}
\defval{kvseg/x40/1024/dense-l47/govreport-test/1920-2048/hi}{+0.00042}
\defval{kvseg/x40/1024/dense-l47/govreport-test/1920-2048/last}{0.6692}
\defval{kvseg/x40/1024/dense-l47/govreport-test/1920-2048/lo}{$-$0.00012}
\defval{kvseg/x40/1024/dense-l47/govreport-test/1920-2048/round}{0.6691}
\defval{kvseg/x40/1024/dense-l47/govreport-test/384-512/delta}{$-$0.00003}
\defval{kvseg/x40/1024/dense-l47/govreport-test/384-512/hi}{+0.00020}
\defval{kvseg/x40/1024/dense-l47/govreport-test/384-512/last}{0.6977}
\defval{kvseg/x40/1024/dense-l47/govreport-test/384-512/lo}{$-$0.00026}
\defval{kvseg/x40/1024/dense-l47/govreport-test/384-512/round}{0.6977}
\defval{kvseg/x40/1024/dense-l47/govreport-test/896-1024/delta}{+0.00008}
\defval{kvseg/x40/1024/dense-l47/govreport-test/896-1024/hi}{+0.00029}
\defval{kvseg/x40/1024/dense-l47/govreport-test/896-1024/last}{0.6788}
\defval{kvseg/x40/1024/dense-l47/govreport-test/896-1024/lo}{$-$0.00014}
\defval{kvseg/x40/1024/dense-l47/govreport-test/896-1024/round}{0.6787}
\defval{kvseg/x40/1024/dense-l47/govreport-test/all/delta}{+0.00008}
\defval{kvseg/x40/1024/dense-l47/govreport-test/all/hi}{+0.00019}
\defval{kvseg/x40/1024/dense-l47/govreport-test/all/last}{0.7294}
\defval{kvseg/x40/1024/dense-l47/govreport-test/all/lo}{$-$0.00003}
\defval{kvseg/x40/1024/dense-l47/govreport-test/all/round}{0.7293}
\defval{kvseg/x40/1024/dense-l47/longdocs/1-128/delta}{$-$0.00003}
\defval{kvseg/x40/1024/dense-l47/longdocs/1-128/hi}{+0.00016}
\defval{kvseg/x40/1024/dense-l47/longdocs/1-128/last}{1.2160}
\defval{kvseg/x40/1024/dense-l47/longdocs/1-128/lo}{$-$0.00020}
\defval{kvseg/x40/1024/dense-l47/longdocs/1-128/round}{1.2160}
\defval{kvseg/x40/1024/dense-l47/longdocs/128-256/delta}{$-$0.00001}
\defval{kvseg/x40/1024/dense-l47/longdocs/128-256/hi}{+0.00016}
\defval{kvseg/x40/1024/dense-l47/longdocs/128-256/last}{1.0482}
\defval{kvseg/x40/1024/dense-l47/longdocs/128-256/lo}{$-$0.00019}
\defval{kvseg/x40/1024/dense-l47/longdocs/128-256/round}{1.0482}
\defval{kvseg/x40/1024/dense-l47/longdocs/1920-2048/delta}{$-$0.00003}
\defval{kvseg/x40/1024/dense-l47/longdocs/1920-2048/hi}{+0.00015}
\defval{kvseg/x40/1024/dense-l47/longdocs/1920-2048/last}{0.9601}
\defval{kvseg/x40/1024/dense-l47/longdocs/1920-2048/lo}{$-$0.00020}
\defval{kvseg/x40/1024/dense-l47/longdocs/1920-2048/round}{0.9601}
\defval{kvseg/x40/1024/dense-l47/longdocs/384-512/delta}{$-$0.00011}
\defval{kvseg/x40/1024/dense-l47/longdocs/384-512/hi}{+0.00006}
\defval{kvseg/x40/1024/dense-l47/longdocs/384-512/last}{1.0030}
\defval{kvseg/x40/1024/dense-l47/longdocs/384-512/lo}{$-$0.00029}
\defval{kvseg/x40/1024/dense-l47/longdocs/384-512/round}{1.0031}
\defval{kvseg/x40/1024/dense-l47/longdocs/896-1024/delta}{$-$0.00004}
\defval{kvseg/x40/1024/dense-l47/longdocs/896-1024/hi}{+0.00014}
\defval{kvseg/x40/1024/dense-l47/longdocs/896-1024/last}{0.9884}
\defval{kvseg/x40/1024/dense-l47/longdocs/896-1024/lo}{$-$0.00022}
\defval{kvseg/x40/1024/dense-l47/longdocs/896-1024/round}{0.9884}
\defval{kvseg/x40/1024/dense-l47/longdocs/all/delta}{$-$0.00004}
\defval{kvseg/x40/1024/dense-l47/longdocs/all/hi}{+0.00004}
\defval{kvseg/x40/1024/dense-l47/longdocs/all/last}{1.0431}
\defval{kvseg/x40/1024/dense-l47/longdocs/all/lo}{$-$0.00013}
\defval{kvseg/x40/1024/dense-l47/longdocs/all/round}{1.0432}
\defval{kvseg/x40/1024/dense-l47/pg19-test/1-128/delta}{$-$0.00002}
\defval{kvseg/x40/1024/dense-l47/pg19-test/1-128/hi}{+0.00036}
\defval{kvseg/x40/1024/dense-l47/pg19-test/1-128/last}{1.4920}
\defval{kvseg/x40/1024/dense-l47/pg19-test/1-128/lo}{$-$0.00041}
\defval{kvseg/x40/1024/dense-l47/pg19-test/1-128/round}{1.4920}
\defval{kvseg/x40/1024/dense-l47/pg19-test/128-256/delta}{+0.00011}
\defval{kvseg/x40/1024/dense-l47/pg19-test/128-256/hi}{+0.00046}
\defval{kvseg/x40/1024/dense-l47/pg19-test/128-256/last}{1.3247}
\defval{kvseg/x40/1024/dense-l47/pg19-test/128-256/lo}{$-$0.00024}
\defval{kvseg/x40/1024/dense-l47/pg19-test/128-256/round}{1.3246}
\defval{kvseg/x40/1024/dense-l47/pg19-test/1920-2048/delta}{$-$0.00012}
\defval{kvseg/x40/1024/dense-l47/pg19-test/1920-2048/hi}{+0.00023}
\defval{kvseg/x40/1024/dense-l47/pg19-test/1920-2048/last}{1.2576}
\defval{kvseg/x40/1024/dense-l47/pg19-test/1920-2048/lo}{$-$0.00047}
\defval{kvseg/x40/1024/dense-l47/pg19-test/1920-2048/round}{1.2577}
\defval{kvseg/x40/1024/dense-l47/pg19-test/384-512/delta}{$-$0.00006}
\defval{kvseg/x40/1024/dense-l47/pg19-test/384-512/hi}{+0.00033}
\defval{kvseg/x40/1024/dense-l47/pg19-test/384-512/last}{1.2881}
\defval{kvseg/x40/1024/dense-l47/pg19-test/384-512/lo}{$-$0.00044}
\defval{kvseg/x40/1024/dense-l47/pg19-test/384-512/round}{1.2881}
\defval{kvseg/x40/1024/dense-l47/pg19-test/896-1024/delta}{$-$0.00022}
\defval{kvseg/x40/1024/dense-l47/pg19-test/896-1024/hi}{+0.00014}
\defval{kvseg/x40/1024/dense-l47/pg19-test/896-1024/last}{1.2844}
\defval{kvseg/x40/1024/dense-l47/pg19-test/896-1024/lo}{$-$0.00058}
\defval{kvseg/x40/1024/dense-l47/pg19-test/896-1024/round}{1.2846}
\defval{kvseg/x40/1024/dense-l47/pg19-test/all/delta}{$-$0.00006}
\defval{kvseg/x40/1024/dense-l47/pg19-test/all/hi}{+0.00009}
\defval{kvseg/x40/1024/dense-l47/pg19-test/all/last}{1.3290}
\defval{kvseg/x40/1024/dense-l47/pg19-test/all/lo}{$-$0.00024}
\defval{kvseg/x40/1024/dense-l47/pg19-test/all/round}{1.3290}
\defval{kvseg/x40/1024/dense-l47/pg19-val/1-128/delta}{$-$0.00007}
\defval{kvseg/x40/1024/dense-l47/pg19-val/1-128/hi}{+0.00029}
\defval{kvseg/x40/1024/dense-l47/pg19-val/1-128/last}{1.4412}
\defval{kvseg/x40/1024/dense-l47/pg19-val/1-128/lo}{$-$0.00044}
\defval{kvseg/x40/1024/dense-l47/pg19-val/1-128/round}{1.4413}
\defval{kvseg/x40/1024/dense-l47/pg19-val/128-256/delta}{$-$0.00012}
\defval{kvseg/x40/1024/dense-l47/pg19-val/128-256/hi}{+0.00023}
\defval{kvseg/x40/1024/dense-l47/pg19-val/128-256/last}{1.2876}
\defval{kvseg/x40/1024/dense-l47/pg19-val/128-256/lo}{$-$0.00046}
\defval{kvseg/x40/1024/dense-l47/pg19-val/128-256/round}{1.2877}
\defval{kvseg/x40/1024/dense-l47/pg19-val/1920-2048/delta}{$-$0.00014}
\defval{kvseg/x40/1024/dense-l47/pg19-val/1920-2048/hi}{+0.00025}
\defval{kvseg/x40/1024/dense-l47/pg19-val/1920-2048/last}{1.2177}
\defval{kvseg/x40/1024/dense-l47/pg19-val/1920-2048/lo}{$-$0.00053}
\defval{kvseg/x40/1024/dense-l47/pg19-val/1920-2048/round}{1.2178}
\defval{kvseg/x40/1024/dense-l47/pg19-val/384-512/delta}{$-$0.00009}
\defval{kvseg/x40/1024/dense-l47/pg19-val/384-512/hi}{+0.00029}
\defval{kvseg/x40/1024/dense-l47/pg19-val/384-512/last}{1.2572}
\defval{kvseg/x40/1024/dense-l47/pg19-val/384-512/lo}{$-$0.00046}
\defval{kvseg/x40/1024/dense-l47/pg19-val/384-512/round}{1.2573}
\defval{kvseg/x40/1024/dense-l47/pg19-val/896-1024/delta}{$-$0.00006}
\defval{kvseg/x40/1024/dense-l47/pg19-val/896-1024/hi}{+0.00030}
\defval{kvseg/x40/1024/dense-l47/pg19-val/896-1024/last}{1.2348}
\defval{kvseg/x40/1024/dense-l47/pg19-val/896-1024/lo}{$-$0.00044}
\defval{kvseg/x40/1024/dense-l47/pg19-val/896-1024/round}{1.2349}
\defval{kvseg/x40/1024/dense-l47/pg19-val/all/delta}{$-$0.00010}
\defval{kvseg/x40/1024/dense-l47/pg19-val/all/hi}{+0.00007}
\defval{kvseg/x40/1024/dense-l47/pg19-val/all/last}{1.2876}
\defval{kvseg/x40/1024/dense-l47/pg19-val/all/lo}{$-$0.00025}
\defval{kvseg/x40/1024/dense-l47/pg19-val/all/round}{1.2877}
\defval{kvseg/x40/1024/yocou-loop/finepdfs-long/1-128/delta}{$-$0.00049}
\defval{kvseg/x40/1024/yocou-loop/finepdfs-long/1-128/hi}{$-$0.00014}
\defval{kvseg/x40/1024/yocou-loop/finepdfs-long/1-128/last}{1.3914}
\defval{kvseg/x40/1024/yocou-loop/finepdfs-long/1-128/lo}{$-$0.00082}
\defval{kvseg/x40/1024/yocou-loop/finepdfs-long/1-128/round}{1.3919}
\defval{kvseg/x40/1024/yocou-loop/finepdfs-long/128-256/delta}{$-$0.00045}
\defval{kvseg/x40/1024/yocou-loop/finepdfs-long/128-256/hi}{$-$0.00012}
\defval{kvseg/x40/1024/yocou-loop/finepdfs-long/128-256/last}{1.1626}
\defval{kvseg/x40/1024/yocou-loop/finepdfs-long/128-256/lo}{$-$0.00075}
\defval{kvseg/x40/1024/yocou-loop/finepdfs-long/128-256/round}{1.1630}
\defval{kvseg/x40/1024/yocou-loop/finepdfs-long/1920-2048/delta}{$-$0.00059}
\defval{kvseg/x40/1024/yocou-loop/finepdfs-long/1920-2048/hi}{$-$0.00020}
\defval{kvseg/x40/1024/yocou-loop/finepdfs-long/1920-2048/last}{1.0412}
\defval{kvseg/x40/1024/yocou-loop/finepdfs-long/1920-2048/lo}{$-$0.00100}
\defval{kvseg/x40/1024/yocou-loop/finepdfs-long/1920-2048/round}{1.0418}
\defval{kvseg/x40/1024/yocou-loop/finepdfs-long/384-512/delta}{$-$0.00019}
\defval{kvseg/x40/1024/yocou-loop/finepdfs-long/384-512/hi}{+0.00021}
\defval{kvseg/x40/1024/yocou-loop/finepdfs-long/384-512/last}{1.1093}
\defval{kvseg/x40/1024/yocou-loop/finepdfs-long/384-512/lo}{$-$0.00059}
\defval{kvseg/x40/1024/yocou-loop/finepdfs-long/384-512/round}{1.1095}
\defval{kvseg/x40/1024/yocou-loop/finepdfs-long/896-1024/delta}{$-$0.00033}
\defval{kvseg/x40/1024/yocou-loop/finepdfs-long/896-1024/hi}{$-$0.00001}
\defval{kvseg/x40/1024/yocou-loop/finepdfs-long/896-1024/last}{1.0941}
\defval{kvseg/x40/1024/yocou-loop/finepdfs-long/896-1024/lo}{$-$0.00067}
\defval{kvseg/x40/1024/yocou-loop/finepdfs-long/896-1024/round}{1.0944}
\defval{kvseg/x40/1024/yocou-loop/finepdfs-long/all/delta}{$-$0.00041}
\defval{kvseg/x40/1024/yocou-loop/finepdfs-long/all/hi}{$-$0.00025}
\defval{kvseg/x40/1024/yocou-loop/finepdfs-long/all/last}{1.1603}
\defval{kvseg/x40/1024/yocou-loop/finepdfs-long/all/lo}{$-$0.00057}
\defval{kvseg/x40/1024/yocou-loop/finepdfs-long/all/round}{1.1607}
\defval{kvseg/x40/1024/yocou-loop/fineweb-edu-100bt/1-128/delta}{+0.00006}
\defval{kvseg/x40/1024/yocou-loop/fineweb-edu-100bt/1-128/hi}{+0.00039}
\defval{kvseg/x40/1024/yocou-loop/fineweb-edu-100bt/1-128/last}{0.9585}
\defval{kvseg/x40/1024/yocou-loop/fineweb-edu-100bt/1-128/lo}{$-$0.00029}
\defval{kvseg/x40/1024/yocou-loop/fineweb-edu-100bt/1-128/round}{0.9584}
\defval{kvseg/x40/1024/yocou-loop/fineweb-edu-100bt/128-256/delta}{+0.00037}
\defval{kvseg/x40/1024/yocou-loop/fineweb-edu-100bt/128-256/hi}{+0.00064}
\defval{kvseg/x40/1024/yocou-loop/fineweb-edu-100bt/128-256/last}{0.8592}
\defval{kvseg/x40/1024/yocou-loop/fineweb-edu-100bt/128-256/lo}{+0.00011}
\defval{kvseg/x40/1024/yocou-loop/fineweb-edu-100bt/128-256/round}{0.8588}
\defval{kvseg/x40/1024/yocou-loop/fineweb-edu-100bt/1920-2048/delta}{$-$0.00015}
\defval{kvseg/x40/1024/yocou-loop/fineweb-edu-100bt/1920-2048/hi}{+0.00012}
\defval{kvseg/x40/1024/yocou-loop/fineweb-edu-100bt/1920-2048/last}{0.8402}
\defval{kvseg/x40/1024/yocou-loop/fineweb-edu-100bt/1920-2048/lo}{$-$0.00042}
\defval{kvseg/x40/1024/yocou-loop/fineweb-edu-100bt/1920-2048/round}{0.8404}
\defval{kvseg/x40/1024/yocou-loop/fineweb-edu-100bt/384-512/delta}{$-$0.00000}
\defval{kvseg/x40/1024/yocou-loop/fineweb-edu-100bt/384-512/hi}{+0.00027}
\defval{kvseg/x40/1024/yocou-loop/fineweb-edu-100bt/384-512/last}{0.8445}
\defval{kvseg/x40/1024/yocou-loop/fineweb-edu-100bt/384-512/lo}{$-$0.00029}
\defval{kvseg/x40/1024/yocou-loop/fineweb-edu-100bt/384-512/round}{0.8445}
\defval{kvseg/x40/1024/yocou-loop/fineweb-edu-100bt/896-1024/delta}{$-$0.00013}
\defval{kvseg/x40/1024/yocou-loop/fineweb-edu-100bt/896-1024/hi}{+0.00010}
\defval{kvseg/x40/1024/yocou-loop/fineweb-edu-100bt/896-1024/last}{0.8411}
\defval{kvseg/x40/1024/yocou-loop/fineweb-edu-100bt/896-1024/lo}{$-$0.00035}
\defval{kvseg/x40/1024/yocou-loop/fineweb-edu-100bt/896-1024/round}{0.8412}
\defval{kvseg/x40/1024/yocou-loop/fineweb-edu-100bt/all/delta}{+0.00003}
\defval{kvseg/x40/1024/yocou-loop/fineweb-edu-100bt/all/hi}{+0.00015}
\defval{kvseg/x40/1024/yocou-loop/fineweb-edu-100bt/all/last}{0.8685}
\defval{kvseg/x40/1024/yocou-loop/fineweb-edu-100bt/all/lo}{$-$0.00009}
\defval{kvseg/x40/1024/yocou-loop/fineweb-edu-100bt/all/round}{0.8685}
\defval{kvseg/x40/1024/yocou-loop/govreport-test/1-128/delta}{+0.00015}
\defval{kvseg/x40/1024/yocou-loop/govreport-test/1-128/hi}{+0.00039}
\defval{kvseg/x40/1024/yocou-loop/govreport-test/1-128/last}{0.8593}
\defval{kvseg/x40/1024/yocou-loop/govreport-test/1-128/lo}{$-$0.00010}
\defval{kvseg/x40/1024/yocou-loop/govreport-test/1-128/round}{0.8592}
\defval{kvseg/x40/1024/yocou-loop/govreport-test/128-256/delta}{$-$0.00016}
\defval{kvseg/x40/1024/yocou-loop/govreport-test/128-256/hi}{+0.00005}
\defval{kvseg/x40/1024/yocou-loop/govreport-test/128-256/last}{0.7341}
\defval{kvseg/x40/1024/yocou-loop/govreport-test/128-256/lo}{$-$0.00038}
\defval{kvseg/x40/1024/yocou-loop/govreport-test/128-256/round}{0.7343}
\defval{kvseg/x40/1024/yocou-loop/govreport-test/1920-2048/delta}{$-$0.00017}
\defval{kvseg/x40/1024/yocou-loop/govreport-test/1920-2048/hi}{+0.00005}
\defval{kvseg/x40/1024/yocou-loop/govreport-test/1920-2048/last}{0.6753}
\defval{kvseg/x40/1024/yocou-loop/govreport-test/1920-2048/lo}{$-$0.00038}
\defval{kvseg/x40/1024/yocou-loop/govreport-test/1920-2048/round}{0.6755}
\defval{kvseg/x40/1024/yocou-loop/govreport-test/384-512/delta}{$-$0.00010}
\defval{kvseg/x40/1024/yocou-loop/govreport-test/384-512/hi}{+0.00008}
\defval{kvseg/x40/1024/yocou-loop/govreport-test/384-512/last}{0.6978}
\defval{kvseg/x40/1024/yocou-loop/govreport-test/384-512/lo}{$-$0.00029}
\defval{kvseg/x40/1024/yocou-loop/govreport-test/384-512/round}{0.6979}
\defval{kvseg/x40/1024/yocou-loop/govreport-test/896-1024/delta}{$-$0.00024}
\defval{kvseg/x40/1024/yocou-loop/govreport-test/896-1024/hi}{$-$0.00005}
\defval{kvseg/x40/1024/yocou-loop/govreport-test/896-1024/last}{0.6837}
\defval{kvseg/x40/1024/yocou-loop/govreport-test/896-1024/lo}{$-$0.00045}
\defval{kvseg/x40/1024/yocou-loop/govreport-test/896-1024/round}{0.6840}
\defval{kvseg/x40/1024/yocou-loop/govreport-test/all/delta}{$-$0.00011}
\defval{kvseg/x40/1024/yocou-loop/govreport-test/all/hi}{$-$0.00001}
\defval{kvseg/x40/1024/yocou-loop/govreport-test/all/last}{0.7300}
\defval{kvseg/x40/1024/yocou-loop/govreport-test/all/lo}{$-$0.00021}
\defval{kvseg/x40/1024/yocou-loop/govreport-test/all/round}{0.7301}
\defval{kvseg/x40/1024/yocou-loop/longdocs/1-128/delta}{$-$0.00022}
\defval{kvseg/x40/1024/yocou-loop/longdocs/1-128/hi}{$-$0.00005}
\defval{kvseg/x40/1024/yocou-loop/longdocs/1-128/last}{1.2076}
\defval{kvseg/x40/1024/yocou-loop/longdocs/1-128/lo}{$-$0.00042}
\defval{kvseg/x40/1024/yocou-loop/longdocs/1-128/round}{1.2079}
\defval{kvseg/x40/1024/yocou-loop/longdocs/128-256/delta}{$-$0.00041}
\defval{kvseg/x40/1024/yocou-loop/longdocs/128-256/hi}{$-$0.00024}
\defval{kvseg/x40/1024/yocou-loop/longdocs/128-256/last}{1.0407}
\defval{kvseg/x40/1024/yocou-loop/longdocs/128-256/lo}{$-$0.00060}
\defval{kvseg/x40/1024/yocou-loop/longdocs/128-256/round}{1.0411}
\defval{kvseg/x40/1024/yocou-loop/longdocs/1920-2048/delta}{$-$0.00049}
\defval{kvseg/x40/1024/yocou-loop/longdocs/1920-2048/hi}{$-$0.00031}
\defval{kvseg/x40/1024/yocou-loop/longdocs/1920-2048/last}{0.9633}
\defval{kvseg/x40/1024/yocou-loop/longdocs/1920-2048/lo}{$-$0.00069}
\defval{kvseg/x40/1024/yocou-loop/longdocs/1920-2048/round}{0.9638}
\defval{kvseg/x40/1024/yocou-loop/longdocs/384-512/delta}{$-$0.00038}
\defval{kvseg/x40/1024/yocou-loop/longdocs/384-512/hi}{$-$0.00019}
\defval{kvseg/x40/1024/yocou-loop/longdocs/384-512/last}{1.0004}
\defval{kvseg/x40/1024/yocou-loop/longdocs/384-512/lo}{$-$0.00058}
\defval{kvseg/x40/1024/yocou-loop/longdocs/384-512/round}{1.0008}
\defval{kvseg/x40/1024/yocou-loop/longdocs/896-1024/delta}{$-$0.00040}
\defval{kvseg/x40/1024/yocou-loop/longdocs/896-1024/hi}{$-$0.00023}
\defval{kvseg/x40/1024/yocou-loop/longdocs/896-1024/last}{0.9899}
\defval{kvseg/x40/1024/yocou-loop/longdocs/896-1024/lo}{$-$0.00058}
\defval{kvseg/x40/1024/yocou-loop/longdocs/896-1024/round}{0.9903}
\defval{kvseg/x40/1024/yocou-loop/longdocs/all/delta}{$-$0.00038}
\defval{kvseg/x40/1024/yocou-loop/longdocs/all/hi}{$-$0.00030}
\defval{kvseg/x40/1024/yocou-loop/longdocs/all/last}{1.0404}
\defval{kvseg/x40/1024/yocou-loop/longdocs/all/lo}{$-$0.00047}
\defval{kvseg/x40/1024/yocou-loop/longdocs/all/round}{1.0408}
\defval{kvseg/x40/1024/yocou-loop/pg19-test/1-128/delta}{$-$0.00045}
\defval{kvseg/x40/1024/yocou-loop/pg19-test/1-128/hi}{$-$0.00004}
\defval{kvseg/x40/1024/yocou-loop/pg19-test/1-128/last}{1.4807}
\defval{kvseg/x40/1024/yocou-loop/pg19-test/1-128/lo}{$-$0.00086}
\defval{kvseg/x40/1024/yocou-loop/pg19-test/1-128/round}{1.4811}
\defval{kvseg/x40/1024/yocou-loop/pg19-test/128-256/delta}{$-$0.00070}
\defval{kvseg/x40/1024/yocou-loop/pg19-test/128-256/hi}{$-$0.00028}
\defval{kvseg/x40/1024/yocou-loop/pg19-test/128-256/last}{1.3157}
\defval{kvseg/x40/1024/yocou-loop/pg19-test/128-256/lo}{$-$0.00114}
\defval{kvseg/x40/1024/yocou-loop/pg19-test/128-256/round}{1.3164}
\defval{kvseg/x40/1024/yocou-loop/pg19-test/1920-2048/delta}{$-$0.00081}
\defval{kvseg/x40/1024/yocou-loop/pg19-test/1920-2048/hi}{$-$0.00049}
\defval{kvseg/x40/1024/yocou-loop/pg19-test/1920-2048/last}{1.2604}
\defval{kvseg/x40/1024/yocou-loop/pg19-test/1920-2048/lo}{$-$0.00119}
\defval{kvseg/x40/1024/yocou-loop/pg19-test/1920-2048/round}{1.2612}
\defval{kvseg/x40/1024/yocou-loop/pg19-test/384-512/delta}{$-$0.00092}
\defval{kvseg/x40/1024/yocou-loop/pg19-test/384-512/hi}{$-$0.00049}
\defval{kvseg/x40/1024/yocou-loop/pg19-test/384-512/last}{1.2846}
\defval{kvseg/x40/1024/yocou-loop/pg19-test/384-512/lo}{$-$0.00134}
\defval{kvseg/x40/1024/yocou-loop/pg19-test/384-512/round}{1.2855}
\defval{kvseg/x40/1024/yocou-loop/pg19-test/896-1024/delta}{$-$0.00067}
\defval{kvseg/x40/1024/yocou-loop/pg19-test/896-1024/hi}{$-$0.00029}
\defval{kvseg/x40/1024/yocou-loop/pg19-test/896-1024/last}{1.2850}
\defval{kvseg/x40/1024/yocou-loop/pg19-test/896-1024/lo}{$-$0.00105}
\defval{kvseg/x40/1024/yocou-loop/pg19-test/896-1024/round}{1.2856}
\defval{kvseg/x40/1024/yocou-loop/pg19-test/all/delta}{$-$0.00071}
\defval{kvseg/x40/1024/yocou-loop/pg19-test/all/hi}{$-$0.00054}
\defval{kvseg/x40/1024/yocou-loop/pg19-test/all/last}{1.3249}
\defval{kvseg/x40/1024/yocou-loop/pg19-test/all/lo}{$-$0.00088}
\defval{kvseg/x40/1024/yocou-loop/pg19-test/all/round}{1.3256}
\defval{kvseg/x40/1024/yocou-loop/pg19-val/1-128/delta}{$-$0.00054}
\defval{kvseg/x40/1024/yocou-loop/pg19-val/1-128/hi}{$-$0.00016}
\defval{kvseg/x40/1024/yocou-loop/pg19-val/1-128/last}{1.4304}
\defval{kvseg/x40/1024/yocou-loop/pg19-val/1-128/lo}{$-$0.00094}
\defval{kvseg/x40/1024/yocou-loop/pg19-val/1-128/round}{1.4309}
\defval{kvseg/x40/1024/yocou-loop/pg19-val/128-256/delta}{$-$0.00068}
\defval{kvseg/x40/1024/yocou-loop/pg19-val/128-256/hi}{$-$0.00030}
\defval{kvseg/x40/1024/yocou-loop/pg19-val/128-256/last}{1.2806}
\defval{kvseg/x40/1024/yocou-loop/pg19-val/128-256/lo}{$-$0.00106}
\defval{kvseg/x40/1024/yocou-loop/pg19-val/128-256/round}{1.2813}
\defval{kvseg/x40/1024/yocou-loop/pg19-val/1920-2048/delta}{$-$0.00064}
\defval{kvseg/x40/1024/yocou-loop/pg19-val/1920-2048/hi}{$-$0.00016}
\defval{kvseg/x40/1024/yocou-loop/pg19-val/1920-2048/last}{1.2182}
\defval{kvseg/x40/1024/yocou-loop/pg19-val/1920-2048/lo}{$-$0.00108}
\defval{kvseg/x40/1024/yocou-loop/pg19-val/1920-2048/round}{1.2188}
\defval{kvseg/x40/1024/yocou-loop/pg19-val/384-512/delta}{$-$0.00058}
\defval{kvseg/x40/1024/yocou-loop/pg19-val/384-512/hi}{$-$0.00018}
\defval{kvseg/x40/1024/yocou-loop/pg19-val/384-512/last}{1.2554}
\defval{kvseg/x40/1024/yocou-loop/pg19-val/384-512/lo}{$-$0.00098}
\defval{kvseg/x40/1024/yocou-loop/pg19-val/384-512/round}{1.2560}
\defval{kvseg/x40/1024/yocou-loop/pg19-val/896-1024/delta}{$-$0.00070}
\defval{kvseg/x40/1024/yocou-loop/pg19-val/896-1024/hi}{$-$0.00035}
\defval{kvseg/x40/1024/yocou-loop/pg19-val/896-1024/last}{1.2360}
\defval{kvseg/x40/1024/yocou-loop/pg19-val/896-1024/lo}{$-$0.00109}
\defval{kvseg/x40/1024/yocou-loop/pg19-val/896-1024/round}{1.2367}
\defval{kvseg/x40/1024/yocou-loop/pg19-val/all/delta}{$-$0.00063}
\defval{kvseg/x40/1024/yocou-loop/pg19-val/all/hi}{$-$0.00044}
\defval{kvseg/x40/1024/yocou-loop/pg19-val/all/last}{1.2840}
\defval{kvseg/x40/1024/yocou-loop/pg19-val/all/lo}{$-$0.00082}
\defval{kvseg/x40/1024/yocou-loop/pg19-val/all/round}{1.2846}
\defval{kvseg/x40/1536/ced-dec/finepdfs-long/1-128/delta}{$-$0.00020}
\defval{kvseg/x40/1536/ced-dec/finepdfs-long/1-128/hi}{+0.00009}
\defval{kvseg/x40/1536/ced-dec/finepdfs-long/1-128/last}{1.3446}
\defval{kvseg/x40/1536/ced-dec/finepdfs-long/1-128/lo}{$-$0.00052}
\defval{kvseg/x40/1536/ced-dec/finepdfs-long/1-128/round}{1.3448}
\defval{kvseg/x40/1536/ced-dec/finepdfs-long/128-256/delta}{+0.00015}
\defval{kvseg/x40/1536/ced-dec/finepdfs-long/128-256/hi}{+0.00034}
\defval{kvseg/x40/1536/ced-dec/finepdfs-long/128-256/last}{1.1116}
\defval{kvseg/x40/1536/ced-dec/finepdfs-long/128-256/lo}{$-$0.00005}
\defval{kvseg/x40/1536/ced-dec/finepdfs-long/128-256/round}{1.1114}
\defval{kvseg/x40/1536/ced-dec/finepdfs-long/1920-2048/delta}{+0.00011}
\defval{kvseg/x40/1536/ced-dec/finepdfs-long/1920-2048/hi}{+0.00034}
\defval{kvseg/x40/1536/ced-dec/finepdfs-long/1920-2048/last}{0.9879}
\defval{kvseg/x40/1536/ced-dec/finepdfs-long/1920-2048/lo}{$-$0.00014}
\defval{kvseg/x40/1536/ced-dec/finepdfs-long/1920-2048/round}{0.9878}
\defval{kvseg/x40/1536/ced-dec/finepdfs-long/384-512/delta}{+0.00024}
\defval{kvseg/x40/1536/ced-dec/finepdfs-long/384-512/hi}{+0.00043}
\defval{kvseg/x40/1536/ced-dec/finepdfs-long/384-512/last}{1.0542}
\defval{kvseg/x40/1536/ced-dec/finepdfs-long/384-512/lo}{+0.00007}
\defval{kvseg/x40/1536/ced-dec/finepdfs-long/384-512/round}{1.0540}
\defval{kvseg/x40/1536/ced-dec/finepdfs-long/896-1024/delta}{+0.00038}
\defval{kvseg/x40/1536/ced-dec/finepdfs-long/896-1024/hi}{+0.00058}
\defval{kvseg/x40/1536/ced-dec/finepdfs-long/896-1024/last}{1.0378}
\defval{kvseg/x40/1536/ced-dec/finepdfs-long/896-1024/lo}{+0.00018}
\defval{kvseg/x40/1536/ced-dec/finepdfs-long/896-1024/round}{1.0375}
\defval{kvseg/x40/1536/ced-dec/finepdfs-long/all/delta}{+0.00013}
\defval{kvseg/x40/1536/ced-dec/finepdfs-long/all/hi}{+0.00024}
\defval{kvseg/x40/1536/ced-dec/finepdfs-long/all/last}{1.1078}
\defval{kvseg/x40/1536/ced-dec/finepdfs-long/all/lo}{+0.00003}
\defval{kvseg/x40/1536/ced-dec/finepdfs-long/all/round}{1.1077}
\defval{kvseg/x40/1536/ced-dec/fineweb-edu-100bt/1-128/delta}{$-$0.00004}
\defval{kvseg/x40/1536/ced-dec/fineweb-edu-100bt/1-128/hi}{+0.00013}
\defval{kvseg/x40/1536/ced-dec/fineweb-edu-100bt/1-128/last}{0.9268}
\defval{kvseg/x40/1536/ced-dec/fineweb-edu-100bt/1-128/lo}{$-$0.00021}
\defval{kvseg/x40/1536/ced-dec/fineweb-edu-100bt/1-128/round}{0.9268}
\defval{kvseg/x40/1536/ced-dec/fineweb-edu-100bt/128-256/delta}{+0.00002}
\defval{kvseg/x40/1536/ced-dec/fineweb-edu-100bt/128-256/hi}{+0.00017}
\defval{kvseg/x40/1536/ced-dec/fineweb-edu-100bt/128-256/last}{0.8244}
\defval{kvseg/x40/1536/ced-dec/fineweb-edu-100bt/128-256/lo}{$-$0.00014}
\defval{kvseg/x40/1536/ced-dec/fineweb-edu-100bt/128-256/round}{0.8244}
\defval{kvseg/x40/1536/ced-dec/fineweb-edu-100bt/1920-2048/delta}{+0.00017}
\defval{kvseg/x40/1536/ced-dec/fineweb-edu-100bt/1920-2048/hi}{+0.00033}
\defval{kvseg/x40/1536/ced-dec/fineweb-edu-100bt/1920-2048/last}{0.8029}
\defval{kvseg/x40/1536/ced-dec/fineweb-edu-100bt/1920-2048/lo}{+0.00000}
\defval{kvseg/x40/1536/ced-dec/fineweb-edu-100bt/1920-2048/round}{0.8027}
\defval{kvseg/x40/1536/ced-dec/fineweb-edu-100bt/384-512/delta}{+0.00015}
\defval{kvseg/x40/1536/ced-dec/fineweb-edu-100bt/384-512/hi}{+0.00028}
\defval{kvseg/x40/1536/ced-dec/fineweb-edu-100bt/384-512/last}{0.8053}
\defval{kvseg/x40/1536/ced-dec/fineweb-edu-100bt/384-512/lo}{+0.00002}
\defval{kvseg/x40/1536/ced-dec/fineweb-edu-100bt/384-512/round}{0.8052}
\defval{kvseg/x40/1536/ced-dec/fineweb-edu-100bt/896-1024/delta}{+0.00000}
\defval{kvseg/x40/1536/ced-dec/fineweb-edu-100bt/896-1024/hi}{+0.00015}
\defval{kvseg/x40/1536/ced-dec/fineweb-edu-100bt/896-1024/last}{0.8044}
\defval{kvseg/x40/1536/ced-dec/fineweb-edu-100bt/896-1024/lo}{$-$0.00015}
\defval{kvseg/x40/1536/ced-dec/fineweb-edu-100bt/896-1024/round}{0.8044}
\defval{kvseg/x40/1536/ced-dec/fineweb-edu-100bt/all/delta}{+0.00006}
\defval{kvseg/x40/1536/ced-dec/fineweb-edu-100bt/all/hi}{+0.00013}
\defval{kvseg/x40/1536/ced-dec/fineweb-edu-100bt/all/last}{0.8326}
\defval{kvseg/x40/1536/ced-dec/fineweb-edu-100bt/all/lo}{$-$0.00001}
\defval{kvseg/x40/1536/ced-dec/fineweb-edu-100bt/all/round}{0.8325}
\defval{kvseg/x40/1536/ced-dec/govreport-test/1-128/delta}{+0.00008}
\defval{kvseg/x40/1536/ced-dec/govreport-test/1-128/hi}{+0.00024}
\defval{kvseg/x40/1536/ced-dec/govreport-test/1-128/last}{0.8306}
\defval{kvseg/x40/1536/ced-dec/govreport-test/1-128/lo}{$-$0.00006}
\defval{kvseg/x40/1536/ced-dec/govreport-test/1-128/round}{0.8305}
\defval{kvseg/x40/1536/ced-dec/govreport-test/128-256/delta}{+0.00018}
\defval{kvseg/x40/1536/ced-dec/govreport-test/128-256/hi}{+0.00029}
\defval{kvseg/x40/1536/ced-dec/govreport-test/128-256/last}{0.7053}
\defval{kvseg/x40/1536/ced-dec/govreport-test/128-256/lo}{+0.00006}
\defval{kvseg/x40/1536/ced-dec/govreport-test/128-256/round}{0.7052}
\defval{kvseg/x40/1536/ced-dec/govreport-test/1920-2048/delta}{+0.00010}
\defval{kvseg/x40/1536/ced-dec/govreport-test/1920-2048/hi}{+0.00020}
\defval{kvseg/x40/1536/ced-dec/govreport-test/1920-2048/last}{0.6433}
\defval{kvseg/x40/1536/ced-dec/govreport-test/1920-2048/lo}{$-$0.00001}
\defval{kvseg/x40/1536/ced-dec/govreport-test/1920-2048/round}{0.6432}
\defval{kvseg/x40/1536/ced-dec/govreport-test/384-512/delta}{+0.00006}
\defval{kvseg/x40/1536/ced-dec/govreport-test/384-512/hi}{+0.00018}
\defval{kvseg/x40/1536/ced-dec/govreport-test/384-512/last}{0.6682}
\defval{kvseg/x40/1536/ced-dec/govreport-test/384-512/lo}{$-$0.00005}
\defval{kvseg/x40/1536/ced-dec/govreport-test/384-512/round}{0.6681}
\defval{kvseg/x40/1536/ced-dec/govreport-test/896-1024/delta}{+0.00010}
\defval{kvseg/x40/1536/ced-dec/govreport-test/896-1024/hi}{+0.00021}
\defval{kvseg/x40/1536/ced-dec/govreport-test/896-1024/last}{0.6519}
\defval{kvseg/x40/1536/ced-dec/govreport-test/896-1024/lo}{$-$0.00002}
\defval{kvseg/x40/1536/ced-dec/govreport-test/896-1024/round}{0.6518}
\defval{kvseg/x40/1536/ced-dec/govreport-test/all/delta}{+0.00010}
\defval{kvseg/x40/1536/ced-dec/govreport-test/all/hi}{+0.00017}
\defval{kvseg/x40/1536/ced-dec/govreport-test/all/last}{0.6998}
\defval{kvseg/x40/1536/ced-dec/govreport-test/all/lo}{+0.00004}
\defval{kvseg/x40/1536/ced-dec/govreport-test/all/round}{0.6997}
\defval{kvseg/x40/1536/ced-dec/longdocs/1-128/delta}{$-$0.00006}
\defval{kvseg/x40/1536/ced-dec/longdocs/1-128/hi}{+0.00006}
\defval{kvseg/x40/1536/ced-dec/longdocs/1-128/last}{1.1698}
\defval{kvseg/x40/1536/ced-dec/longdocs/1-128/lo}{$-$0.00018}
\defval{kvseg/x40/1536/ced-dec/longdocs/1-128/round}{1.1699}
\defval{kvseg/x40/1536/ced-dec/longdocs/128-256/delta}{+0.00007}
\defval{kvseg/x40/1536/ced-dec/longdocs/128-256/hi}{+0.00018}
\defval{kvseg/x40/1536/ced-dec/longdocs/128-256/last}{0.9983}
\defval{kvseg/x40/1536/ced-dec/longdocs/128-256/lo}{$-$0.00003}
\defval{kvseg/x40/1536/ced-dec/longdocs/128-256/round}{0.9982}
\defval{kvseg/x40/1536/ced-dec/longdocs/1920-2048/delta}{+0.00010}
\defval{kvseg/x40/1536/ced-dec/longdocs/1920-2048/hi}{+0.00021}
\defval{kvseg/x40/1536/ced-dec/longdocs/1920-2048/last}{0.9163}
\defval{kvseg/x40/1536/ced-dec/longdocs/1920-2048/lo}{$-$0.00001}
\defval{kvseg/x40/1536/ced-dec/longdocs/1920-2048/round}{0.9162}
\defval{kvseg/x40/1536/ced-dec/longdocs/384-512/delta}{+0.00009}
\defval{kvseg/x40/1536/ced-dec/longdocs/384-512/hi}{+0.00019}
\defval{kvseg/x40/1536/ced-dec/longdocs/384-512/last}{0.9551}
\defval{kvseg/x40/1536/ced-dec/longdocs/384-512/lo}{$-$0.00001}
\defval{kvseg/x40/1536/ced-dec/longdocs/384-512/round}{0.9550}
\defval{kvseg/x40/1536/ced-dec/longdocs/896-1024/delta}{+0.00021}
\defval{kvseg/x40/1536/ced-dec/longdocs/896-1024/hi}{+0.00031}
\defval{kvseg/x40/1536/ced-dec/longdocs/896-1024/last}{0.9423}
\defval{kvseg/x40/1536/ced-dec/longdocs/896-1024/lo}{+0.00011}
\defval{kvseg/x40/1536/ced-dec/longdocs/896-1024/round}{0.9421}
\defval{kvseg/x40/1536/ced-dec/longdocs/all/delta}{+0.00008}
\defval{kvseg/x40/1536/ced-dec/longdocs/all/hi}{+0.00013}
\defval{kvseg/x40/1536/ced-dec/longdocs/all/last}{0.9964}
\defval{kvseg/x40/1536/ced-dec/longdocs/all/lo}{+0.00004}
\defval{kvseg/x40/1536/ced-dec/longdocs/all/round}{0.9963}
\defval{kvseg/x40/1536/ced-dec/pg19-test/1-128/delta}{$-$0.00012}
\defval{kvseg/x40/1536/ced-dec/pg19-test/1-128/hi}{+0.00011}
\defval{kvseg/x40/1536/ced-dec/pg19-test/1-128/last}{1.4399}
\defval{kvseg/x40/1536/ced-dec/pg19-test/1-128/lo}{$-$0.00034}
\defval{kvseg/x40/1536/ced-dec/pg19-test/1-128/round}{1.4400}
\defval{kvseg/x40/1536/ced-dec/pg19-test/128-256/delta}{$-$0.00013}
\defval{kvseg/x40/1536/ced-dec/pg19-test/128-256/hi}{+0.00010}
\defval{kvseg/x40/1536/ced-dec/pg19-test/128-256/last}{1.2642}
\defval{kvseg/x40/1536/ced-dec/pg19-test/128-256/lo}{$-$0.00034}
\defval{kvseg/x40/1536/ced-dec/pg19-test/128-256/round}{1.2643}
\defval{kvseg/x40/1536/ced-dec/pg19-test/1920-2048/delta}{+0.00010}
\defval{kvseg/x40/1536/ced-dec/pg19-test/1920-2048/hi}{+0.00030}
\defval{kvseg/x40/1536/ced-dec/pg19-test/1920-2048/last}{1.2003}
\defval{kvseg/x40/1536/ced-dec/pg19-test/1920-2048/lo}{$-$0.00011}
\defval{kvseg/x40/1536/ced-dec/pg19-test/1920-2048/round}{1.2002}
\defval{kvseg/x40/1536/ced-dec/pg19-test/384-512/delta}{$-$0.00003}
\defval{kvseg/x40/1536/ced-dec/pg19-test/384-512/hi}{+0.00020}
\defval{kvseg/x40/1536/ced-dec/pg19-test/384-512/last}{1.2286}
\defval{kvseg/x40/1536/ced-dec/pg19-test/384-512/lo}{$-$0.00024}
\defval{kvseg/x40/1536/ced-dec/pg19-test/384-512/round}{1.2286}
\defval{kvseg/x40/1536/ced-dec/pg19-test/896-1024/delta}{+0.00019}
\defval{kvseg/x40/1536/ced-dec/pg19-test/896-1024/hi}{+0.00041}
\defval{kvseg/x40/1536/ced-dec/pg19-test/896-1024/last}{1.2254}
\defval{kvseg/x40/1536/ced-dec/pg19-test/896-1024/lo}{$-$0.00002}
\defval{kvseg/x40/1536/ced-dec/pg19-test/896-1024/round}{1.2252}
\defval{kvseg/x40/1536/ced-dec/pg19-test/all/delta}{+0.00000}
\defval{kvseg/x40/1536/ced-dec/pg19-test/all/hi}{+0.00011}
\defval{kvseg/x40/1536/ced-dec/pg19-test/all/last}{1.2713}
\defval{kvseg/x40/1536/ced-dec/pg19-test/all/lo}{$-$0.00009}
\defval{kvseg/x40/1536/ced-dec/pg19-test/all/round}{1.2713}
\defval{kvseg/x40/1536/ced-dec/pg19-val/1-128/delta}{$-$0.00021}
\defval{kvseg/x40/1536/ced-dec/pg19-val/1-128/hi}{+0.00001}
\defval{kvseg/x40/1536/ced-dec/pg19-val/1-128/last}{1.3891}
\defval{kvseg/x40/1536/ced-dec/pg19-val/1-128/lo}{$-$0.00045}
\defval{kvseg/x40/1536/ced-dec/pg19-val/1-128/round}{1.3893}
\defval{kvseg/x40/1536/ced-dec/pg19-val/128-256/delta}{+0.00003}
\defval{kvseg/x40/1536/ced-dec/pg19-val/128-256/hi}{+0.00024}
\defval{kvseg/x40/1536/ced-dec/pg19-val/128-256/last}{1.2275}
\defval{kvseg/x40/1536/ced-dec/pg19-val/128-256/lo}{$-$0.00016}
\defval{kvseg/x40/1536/ced-dec/pg19-val/128-256/round}{1.2274}
\defval{kvseg/x40/1536/ced-dec/pg19-val/1920-2048/delta}{+0.00026}
\defval{kvseg/x40/1536/ced-dec/pg19-val/1920-2048/hi}{+0.00047}
\defval{kvseg/x40/1536/ced-dec/pg19-val/1920-2048/last}{1.1600}
\defval{kvseg/x40/1536/ced-dec/pg19-val/1920-2048/lo}{+0.00004}
\defval{kvseg/x40/1536/ced-dec/pg19-val/1920-2048/round}{1.1597}
\defval{kvseg/x40/1536/ced-dec/pg19-val/384-512/delta}{+0.00020}
\defval{kvseg/x40/1536/ced-dec/pg19-val/384-512/hi}{+0.00043}
\defval{kvseg/x40/1536/ced-dec/pg19-val/384-512/last}{1.2005}
\defval{kvseg/x40/1536/ced-dec/pg19-val/384-512/lo}{$-$0.00004}
\defval{kvseg/x40/1536/ced-dec/pg19-val/384-512/round}{1.2003}
\defval{kvseg/x40/1536/ced-dec/pg19-val/896-1024/delta}{+0.00001}
\defval{kvseg/x40/1536/ced-dec/pg19-val/896-1024/hi}{+0.00024}
\defval{kvseg/x40/1536/ced-dec/pg19-val/896-1024/last}{1.1759}
\defval{kvseg/x40/1536/ced-dec/pg19-val/896-1024/lo}{$-$0.00020}
\defval{kvseg/x40/1536/ced-dec/pg19-val/896-1024/round}{1.1758}
\defval{kvseg/x40/1536/ced-dec/pg19-val/all/delta}{+0.00006}
\defval{kvseg/x40/1536/ced-dec/pg19-val/all/hi}{+0.00017}
\defval{kvseg/x40/1536/ced-dec/pg19-val/all/last}{1.2305}
\defval{kvseg/x40/1536/ced-dec/pg19-val/all/lo}{$-$0.00004}
\defval{kvseg/x40/1536/ced-dec/pg19-val/all/round}{1.2304}
\defval{kvseg/x40/1536/ced-enc/finepdfs-long/1-128/delta}{$-$0.00042}
\defval{kvseg/x40/1536/ced-enc/finepdfs-long/1-128/hi}{+0.00005}
\defval{kvseg/x40/1536/ced-enc/finepdfs-long/1-128/last}{1.3470}
\defval{kvseg/x40/1536/ced-enc/finepdfs-long/1-128/lo}{$-$0.00088}
\defval{kvseg/x40/1536/ced-enc/finepdfs-long/1-128/round}{1.3475}
\defval{kvseg/x40/1536/ced-enc/finepdfs-long/128-256/delta}{+0.00011}
\defval{kvseg/x40/1536/ced-enc/finepdfs-long/128-256/hi}{+0.00048}
\defval{kvseg/x40/1536/ced-enc/finepdfs-long/128-256/last}{1.1057}
\defval{kvseg/x40/1536/ced-enc/finepdfs-long/128-256/lo}{$-$0.00023}
\defval{kvseg/x40/1536/ced-enc/finepdfs-long/128-256/round}{1.1056}
\defval{kvseg/x40/1536/ced-enc/finepdfs-long/1920-2048/delta}{$-$0.00022}
\defval{kvseg/x40/1536/ced-enc/finepdfs-long/1920-2048/hi}{+0.00025}
\defval{kvseg/x40/1536/ced-enc/finepdfs-long/1920-2048/last}{0.9664}
\defval{kvseg/x40/1536/ced-enc/finepdfs-long/1920-2048/lo}{$-$0.00071}
\defval{kvseg/x40/1536/ced-enc/finepdfs-long/1920-2048/round}{0.9666}
\defval{kvseg/x40/1536/ced-enc/finepdfs-long/384-512/delta}{$-$0.00006}
\defval{kvseg/x40/1536/ced-enc/finepdfs-long/384-512/hi}{+0.00037}
\defval{kvseg/x40/1536/ced-enc/finepdfs-long/384-512/last}{1.0418}
\defval{kvseg/x40/1536/ced-enc/finepdfs-long/384-512/lo}{$-$0.00049}
\defval{kvseg/x40/1536/ced-enc/finepdfs-long/384-512/round}{1.0419}
\defval{kvseg/x40/1536/ced-enc/finepdfs-long/896-1024/delta}{$-$0.00058}
\defval{kvseg/x40/1536/ced-enc/finepdfs-long/896-1024/hi}{$-$0.00016}
\defval{kvseg/x40/1536/ced-enc/finepdfs-long/896-1024/last}{1.0203}
\defval{kvseg/x40/1536/ced-enc/finepdfs-long/896-1024/lo}{$-$0.00100}
\defval{kvseg/x40/1536/ced-enc/finepdfs-long/896-1024/round}{1.0209}
\defval{kvseg/x40/1536/ced-enc/finepdfs-long/all/delta}{$-$0.00023}
\defval{kvseg/x40/1536/ced-enc/finepdfs-long/all/hi}{$-$0.00002}
\defval{kvseg/x40/1536/ced-enc/finepdfs-long/all/last}{1.0969}
\defval{kvseg/x40/1536/ced-enc/finepdfs-long/all/lo}{$-$0.00043}
\defval{kvseg/x40/1536/ced-enc/finepdfs-long/all/round}{1.0971}
\defval{kvseg/x40/1536/ced-enc/fineweb-edu-100bt/1-128/delta}{+0.00002}
\defval{kvseg/x40/1536/ced-enc/fineweb-edu-100bt/1-128/hi}{+0.00030}
\defval{kvseg/x40/1536/ced-enc/fineweb-edu-100bt/1-128/last}{0.9127}
\defval{kvseg/x40/1536/ced-enc/fineweb-edu-100bt/1-128/lo}{$-$0.00028}
\defval{kvseg/x40/1536/ced-enc/fineweb-edu-100bt/1-128/round}{0.9126}
\defval{kvseg/x40/1536/ced-enc/fineweb-edu-100bt/128-256/delta}{+0.00003}
\defval{kvseg/x40/1536/ced-enc/fineweb-edu-100bt/128-256/hi}{+0.00031}
\defval{kvseg/x40/1536/ced-enc/fineweb-edu-100bt/128-256/last}{0.8083}
\defval{kvseg/x40/1536/ced-enc/fineweb-edu-100bt/128-256/lo}{$-$0.00023}
\defval{kvseg/x40/1536/ced-enc/fineweb-edu-100bt/128-256/round}{0.8083}
\defval{kvseg/x40/1536/ced-enc/fineweb-edu-100bt/1920-2048/delta}{+0.00002}
\defval{kvseg/x40/1536/ced-enc/fineweb-edu-100bt/1920-2048/hi}{+0.00028}
\defval{kvseg/x40/1536/ced-enc/fineweb-edu-100bt/1920-2048/last}{0.7850}
\defval{kvseg/x40/1536/ced-enc/fineweb-edu-100bt/1920-2048/lo}{$-$0.00022}
\defval{kvseg/x40/1536/ced-enc/fineweb-edu-100bt/1920-2048/round}{0.7850}
\defval{kvseg/x40/1536/ced-enc/fineweb-edu-100bt/384-512/delta}{$-$0.00009}
\defval{kvseg/x40/1536/ced-enc/fineweb-edu-100bt/384-512/hi}{+0.00016}
\defval{kvseg/x40/1536/ced-enc/fineweb-edu-100bt/384-512/last}{0.7885}
\defval{kvseg/x40/1536/ced-enc/fineweb-edu-100bt/384-512/lo}{$-$0.00036}
\defval{kvseg/x40/1536/ced-enc/fineweb-edu-100bt/384-512/round}{0.7886}
\defval{kvseg/x40/1536/ced-enc/fineweb-edu-100bt/896-1024/delta}{$-$0.00002}
\defval{kvseg/x40/1536/ced-enc/fineweb-edu-100bt/896-1024/hi}{+0.00021}
\defval{kvseg/x40/1536/ced-enc/fineweb-edu-100bt/896-1024/last}{0.7867}
\defval{kvseg/x40/1536/ced-enc/fineweb-edu-100bt/896-1024/lo}{$-$0.00025}
\defval{kvseg/x40/1536/ced-enc/fineweb-edu-100bt/896-1024/round}{0.7868}
\defval{kvseg/x40/1536/ced-enc/fineweb-edu-100bt/all/delta}{$-$0.00001}
\defval{kvseg/x40/1536/ced-enc/fineweb-edu-100bt/all/hi}{+0.00011}
\defval{kvseg/x40/1536/ced-enc/fineweb-edu-100bt/all/last}{0.8161}
\defval{kvseg/x40/1536/ced-enc/fineweb-edu-100bt/all/lo}{$-$0.00012}
\defval{kvseg/x40/1536/ced-enc/fineweb-edu-100bt/all/round}{0.8161}
\defval{kvseg/x40/1536/ced-enc/govreport-test/1-128/delta}{+0.00020}
\defval{kvseg/x40/1536/ced-enc/govreport-test/1-128/hi}{+0.00047}
\defval{kvseg/x40/1536/ced-enc/govreport-test/1-128/last}{0.8185}
\defval{kvseg/x40/1536/ced-enc/govreport-test/1-128/lo}{$-$0.00005}
\defval{kvseg/x40/1536/ced-enc/govreport-test/1-128/round}{0.8183}
\defval{kvseg/x40/1536/ced-enc/govreport-test/128-256/delta}{$-$0.00017}
\defval{kvseg/x40/1536/ced-enc/govreport-test/128-256/hi}{+0.00011}
\defval{kvseg/x40/1536/ced-enc/govreport-test/128-256/last}{0.6896}
\defval{kvseg/x40/1536/ced-enc/govreport-test/128-256/lo}{$-$0.00043}
\defval{kvseg/x40/1536/ced-enc/govreport-test/128-256/round}{0.6898}
\defval{kvseg/x40/1536/ced-enc/govreport-test/1920-2048/delta}{$-$0.00000}
\defval{kvseg/x40/1536/ced-enc/govreport-test/1920-2048/hi}{+0.00027}
\defval{kvseg/x40/1536/ced-enc/govreport-test/1920-2048/last}{0.6242}
\defval{kvseg/x40/1536/ced-enc/govreport-test/1920-2048/lo}{$-$0.00031}
\defval{kvseg/x40/1536/ced-enc/govreport-test/1920-2048/round}{0.6242}
\defval{kvseg/x40/1536/ced-enc/govreport-test/384-512/delta}{$-$0.00041}
\defval{kvseg/x40/1536/ced-enc/govreport-test/384-512/hi}{$-$0.00013}
\defval{kvseg/x40/1536/ced-enc/govreport-test/384-512/last}{0.6515}
\defval{kvseg/x40/1536/ced-enc/govreport-test/384-512/lo}{$-$0.00070}
\defval{kvseg/x40/1536/ced-enc/govreport-test/384-512/round}{0.6519}
\defval{kvseg/x40/1536/ced-enc/govreport-test/896-1024/delta}{$-$0.00040}
\defval{kvseg/x40/1536/ced-enc/govreport-test/896-1024/hi}{$-$0.00018}
\defval{kvseg/x40/1536/ced-enc/govreport-test/896-1024/last}{0.6323}
\defval{kvseg/x40/1536/ced-enc/govreport-test/896-1024/lo}{$-$0.00062}
\defval{kvseg/x40/1536/ced-enc/govreport-test/896-1024/round}{0.6327}
\defval{kvseg/x40/1536/ced-enc/govreport-test/all/delta}{$-$0.00016}
\defval{kvseg/x40/1536/ced-enc/govreport-test/all/hi}{$-$0.00004}
\defval{kvseg/x40/1536/ced-enc/govreport-test/all/last}{0.6832}
\defval{kvseg/x40/1536/ced-enc/govreport-test/all/lo}{$-$0.00027}
\defval{kvseg/x40/1536/ced-enc/govreport-test/all/round}{0.6834}
\defval{kvseg/x40/1536/ced-enc/longdocs/1-128/delta}{$-$0.00037}
\defval{kvseg/x40/1536/ced-enc/longdocs/1-128/hi}{$-$0.00014}
\defval{kvseg/x40/1536/ced-enc/longdocs/1-128/last}{1.1604}
\defval{kvseg/x40/1536/ced-enc/longdocs/1-128/lo}{$-$0.00059}
\defval{kvseg/x40/1536/ced-enc/longdocs/1-128/round}{1.1607}
\defval{kvseg/x40/1536/ced-enc/longdocs/128-256/delta}{$-$0.00020}
\defval{kvseg/x40/1536/ced-enc/longdocs/128-256/hi}{+0.00001}
\defval{kvseg/x40/1536/ced-enc/longdocs/128-256/last}{0.9838}
\defval{kvseg/x40/1536/ced-enc/longdocs/128-256/lo}{$-$0.00041}
\defval{kvseg/x40/1536/ced-enc/longdocs/128-256/round}{0.9840}
\defval{kvseg/x40/1536/ced-enc/longdocs/1920-2048/delta}{$-$0.00003}
\defval{kvseg/x40/1536/ced-enc/longdocs/1920-2048/hi}{+0.00022}
\defval{kvseg/x40/1536/ced-enc/longdocs/1920-2048/last}{0.8951}
\defval{kvseg/x40/1536/ced-enc/longdocs/1920-2048/lo}{$-$0.00027}
\defval{kvseg/x40/1536/ced-enc/longdocs/1920-2048/round}{0.8952}
\defval{kvseg/x40/1536/ced-enc/longdocs/384-512/delta}{$-$0.00028}
\defval{kvseg/x40/1536/ced-enc/longdocs/384-512/hi}{$-$0.00007}
\defval{kvseg/x40/1536/ced-enc/longdocs/384-512/last}{0.9369}
\defval{kvseg/x40/1536/ced-enc/longdocs/384-512/lo}{$-$0.00050}
\defval{kvseg/x40/1536/ced-enc/longdocs/384-512/round}{0.9372}
\defval{kvseg/x40/1536/ced-enc/longdocs/896-1024/delta}{$-$0.00041}
\defval{kvseg/x40/1536/ced-enc/longdocs/896-1024/hi}{$-$0.00021}
\defval{kvseg/x40/1536/ced-enc/longdocs/896-1024/last}{0.9218}
\defval{kvseg/x40/1536/ced-enc/longdocs/896-1024/lo}{$-$0.00063}
\defval{kvseg/x40/1536/ced-enc/longdocs/896-1024/round}{0.9222}
\defval{kvseg/x40/1536/ced-enc/longdocs/all/delta}{$-$0.00026}
\defval{kvseg/x40/1536/ced-enc/longdocs/all/hi}{$-$0.00017}
\defval{kvseg/x40/1536/ced-enc/longdocs/all/last}{0.9796}
\defval{kvseg/x40/1536/ced-enc/longdocs/all/lo}{$-$0.00036}
\defval{kvseg/x40/1536/ced-enc/longdocs/all/round}{0.9799}
\defval{kvseg/x40/1536/ced-enc/pg19-test/1-128/delta}{$-$0.00108}
\defval{kvseg/x40/1536/ced-enc/pg19-test/1-128/hi}{$-$0.00063}
\defval{kvseg/x40/1536/ced-enc/pg19-test/1-128/last}{1.4220}
\defval{kvseg/x40/1536/ced-enc/pg19-test/1-128/lo}{$-$0.00156}
\defval{kvseg/x40/1536/ced-enc/pg19-test/1-128/round}{1.4231}
\defval{kvseg/x40/1536/ced-enc/pg19-test/128-256/delta}{$-$0.00055}
\defval{kvseg/x40/1536/ced-enc/pg19-test/128-256/hi}{$-$0.00007}
\defval{kvseg/x40/1536/ced-enc/pg19-test/128-256/last}{1.2430}
\defval{kvseg/x40/1536/ced-enc/pg19-test/128-256/lo}{$-$0.00101}
\defval{kvseg/x40/1536/ced-enc/pg19-test/128-256/round}{1.2435}
\defval{kvseg/x40/1536/ced-enc/pg19-test/1920-2048/delta}{+0.00013}
\defval{kvseg/x40/1536/ced-enc/pg19-test/1920-2048/hi}{+0.00068}
\defval{kvseg/x40/1536/ced-enc/pg19-test/1920-2048/last}{1.1767}
\defval{kvseg/x40/1536/ced-enc/pg19-test/1920-2048/lo}{$-$0.00040}
\defval{kvseg/x40/1536/ced-enc/pg19-test/1920-2048/round}{1.1765}
\defval{kvseg/x40/1536/ced-enc/pg19-test/384-512/delta}{$-$0.00034}
\defval{kvseg/x40/1536/ced-enc/pg19-test/384-512/hi}{+0.00011}
\defval{kvseg/x40/1536/ced-enc/pg19-test/384-512/last}{1.2029}
\defval{kvseg/x40/1536/ced-enc/pg19-test/384-512/lo}{$-$0.00074}
\defval{kvseg/x40/1536/ced-enc/pg19-test/384-512/round}{1.2032}
\defval{kvseg/x40/1536/ced-enc/pg19-test/896-1024/delta}{$-$0.00028}
\defval{kvseg/x40/1536/ced-enc/pg19-test/896-1024/hi}{+0.00018}
\defval{kvseg/x40/1536/ced-enc/pg19-test/896-1024/last}{1.2007}
\defval{kvseg/x40/1536/ced-enc/pg19-test/896-1024/lo}{$-$0.00078}
\defval{kvseg/x40/1536/ced-enc/pg19-test/896-1024/round}{1.2010}
\defval{kvseg/x40/1536/ced-enc/pg19-test/all/delta}{$-$0.00042}
\defval{kvseg/x40/1536/ced-enc/pg19-test/all/hi}{$-$0.00019}
\defval{kvseg/x40/1536/ced-enc/pg19-test/all/last}{1.2486}
\defval{kvseg/x40/1536/ced-enc/pg19-test/all/lo}{$-$0.00064}
\defval{kvseg/x40/1536/ced-enc/pg19-test/all/round}{1.2491}
\defval{kvseg/x40/1536/ced-enc/pg19-val/1-128/delta}{$-$0.00142}
\defval{kvseg/x40/1536/ced-enc/pg19-val/1-128/hi}{$-$0.00096}
\defval{kvseg/x40/1536/ced-enc/pg19-val/1-128/last}{1.3730}
\defval{kvseg/x40/1536/ced-enc/pg19-val/1-128/lo}{$-$0.00187}
\defval{kvseg/x40/1536/ced-enc/pg19-val/1-128/round}{1.3744}
\defval{kvseg/x40/1536/ced-enc/pg19-val/128-256/delta}{$-$0.00055}
\defval{kvseg/x40/1536/ced-enc/pg19-val/128-256/hi}{$-$0.00018}
\defval{kvseg/x40/1536/ced-enc/pg19-val/128-256/last}{1.2060}
\defval{kvseg/x40/1536/ced-enc/pg19-val/128-256/lo}{$-$0.00097}
\defval{kvseg/x40/1536/ced-enc/pg19-val/128-256/round}{1.2066}
\defval{kvseg/x40/1536/ced-enc/pg19-val/1920-2048/delta}{+0.00009}
\defval{kvseg/x40/1536/ced-enc/pg19-val/1920-2048/hi}{+0.00053}
\defval{kvseg/x40/1536/ced-enc/pg19-val/1920-2048/last}{1.1342}
\defval{kvseg/x40/1536/ced-enc/pg19-val/1920-2048/lo}{$-$0.00036}
\defval{kvseg/x40/1536/ced-enc/pg19-val/1920-2048/round}{1.1342}
\defval{kvseg/x40/1536/ced-enc/pg19-val/384-512/delta}{$-$0.00021}
\defval{kvseg/x40/1536/ced-enc/pg19-val/384-512/hi}{+0.00020}
\defval{kvseg/x40/1536/ced-enc/pg19-val/384-512/last}{1.1753}
\defval{kvseg/x40/1536/ced-enc/pg19-val/384-512/lo}{$-$0.00064}
\defval{kvseg/x40/1536/ced-enc/pg19-val/384-512/round}{1.1756}
\defval{kvseg/x40/1536/ced-enc/pg19-val/896-1024/delta}{$-$0.00026}
\defval{kvseg/x40/1536/ced-enc/pg19-val/896-1024/hi}{+0.00017}
\defval{kvseg/x40/1536/ced-enc/pg19-val/896-1024/last}{1.1515}
\defval{kvseg/x40/1536/ced-enc/pg19-val/896-1024/lo}{$-$0.00071}
\defval{kvseg/x40/1536/ced-enc/pg19-val/896-1024/round}{1.1518}
\defval{kvseg/x40/1536/ced-enc/pg19-val/all/delta}{$-$0.00047}
\defval{kvseg/x40/1536/ced-enc/pg19-val/all/hi}{$-$0.00027}
\defval{kvseg/x40/1536/ced-enc/pg19-val/all/last}{1.2079}
\defval{kvseg/x40/1536/ced-enc/pg19-val/all/lo}{$-$0.00066}
\defval{kvseg/x40/1536/ced-enc/pg19-val/all/round}{1.2084}
\defval{kvseg/x40/1536/dense-l25/finepdfs-long/1-128/delta}{$-$0.00034}
\defval{kvseg/x40/1536/dense-l25/finepdfs-long/1-128/hi}{+0.00009}
\defval{kvseg/x40/1536/dense-l25/finepdfs-long/1-128/last}{1.3250}
\defval{kvseg/x40/1536/dense-l25/finepdfs-long/1-128/lo}{$-$0.00072}
\defval{kvseg/x40/1536/dense-l25/finepdfs-long/1-128/round}{1.3253}
\defval{kvseg/x40/1536/dense-l25/finepdfs-long/128-256/delta}{$-$0.00018}
\defval{kvseg/x40/1536/dense-l25/finepdfs-long/128-256/hi}{+0.00023}
\defval{kvseg/x40/1536/dense-l25/finepdfs-long/128-256/last}{1.0956}
\defval{kvseg/x40/1536/dense-l25/finepdfs-long/128-256/lo}{$-$0.00056}
\defval{kvseg/x40/1536/dense-l25/finepdfs-long/128-256/round}{1.0958}
\defval{kvseg/x40/1536/dense-l25/finepdfs-long/1920-2048/delta}{+0.00013}
\defval{kvseg/x40/1536/dense-l25/finepdfs-long/1920-2048/hi}{+0.00054}
\defval{kvseg/x40/1536/dense-l25/finepdfs-long/1920-2048/last}{0.9618}
\defval{kvseg/x40/1536/dense-l25/finepdfs-long/1920-2048/lo}{$-$0.00027}
\defval{kvseg/x40/1536/dense-l25/finepdfs-long/1920-2048/round}{0.9617}
\defval{kvseg/x40/1536/dense-l25/finepdfs-long/384-512/delta}{+0.00016}
\defval{kvseg/x40/1536/dense-l25/finepdfs-long/384-512/hi}{+0.00053}
\defval{kvseg/x40/1536/dense-l25/finepdfs-long/384-512/last}{1.0331}
\defval{kvseg/x40/1536/dense-l25/finepdfs-long/384-512/lo}{$-$0.00023}
\defval{kvseg/x40/1536/dense-l25/finepdfs-long/384-512/round}{1.0330}
\defval{kvseg/x40/1536/dense-l25/finepdfs-long/896-1024/delta}{+0.00004}
\defval{kvseg/x40/1536/dense-l25/finepdfs-long/896-1024/hi}{+0.00049}
\defval{kvseg/x40/1536/dense-l25/finepdfs-long/896-1024/last}{1.0155}
\defval{kvseg/x40/1536/dense-l25/finepdfs-long/896-1024/lo}{$-$0.00041}
\defval{kvseg/x40/1536/dense-l25/finepdfs-long/896-1024/round}{1.0155}
\defval{kvseg/x40/1536/dense-l25/finepdfs-long/all/delta}{$-$0.00004}
\defval{kvseg/x40/1536/dense-l25/finepdfs-long/all/hi}{+0.00016}
\defval{kvseg/x40/1536/dense-l25/finepdfs-long/all/last}{1.0869}
\defval{kvseg/x40/1536/dense-l25/finepdfs-long/all/lo}{$-$0.00024}
\defval{kvseg/x40/1536/dense-l25/finepdfs-long/all/round}{1.0869}
\defval{kvseg/x40/1536/dense-l25/fineweb-edu-100bt/1-128/delta}{+0.00018}
\defval{kvseg/x40/1536/dense-l25/fineweb-edu-100bt/1-128/hi}{+0.00050}
\defval{kvseg/x40/1536/dense-l25/fineweb-edu-100bt/1-128/last}{0.9155}
\defval{kvseg/x40/1536/dense-l25/fineweb-edu-100bt/1-128/lo}{$-$0.00016}
\defval{kvseg/x40/1536/dense-l25/fineweb-edu-100bt/1-128/round}{0.9153}
\defval{kvseg/x40/1536/dense-l25/fineweb-edu-100bt/128-256/delta}{$-$0.00003}
\defval{kvseg/x40/1536/dense-l25/fineweb-edu-100bt/128-256/hi}{+0.00029}
\defval{kvseg/x40/1536/dense-l25/fineweb-edu-100bt/128-256/last}{0.8095}
\defval{kvseg/x40/1536/dense-l25/fineweb-edu-100bt/128-256/lo}{$-$0.00037}
\defval{kvseg/x40/1536/dense-l25/fineweb-edu-100bt/128-256/round}{0.8095}
\defval{kvseg/x40/1536/dense-l25/fineweb-edu-100bt/1920-2048/delta}{+0.00020}
\defval{kvseg/x40/1536/dense-l25/fineweb-edu-100bt/1920-2048/hi}{+0.00044}
\defval{kvseg/x40/1536/dense-l25/fineweb-edu-100bt/1920-2048/last}{0.7831}
\defval{kvseg/x40/1536/dense-l25/fineweb-edu-100bt/1920-2048/lo}{$-$0.00004}
\defval{kvseg/x40/1536/dense-l25/fineweb-edu-100bt/1920-2048/round}{0.7829}
\defval{kvseg/x40/1536/dense-l25/fineweb-edu-100bt/384-512/delta}{+0.00013}
\defval{kvseg/x40/1536/dense-l25/fineweb-edu-100bt/384-512/hi}{+0.00037}
\defval{kvseg/x40/1536/dense-l25/fineweb-edu-100bt/384-512/last}{0.7881}
\defval{kvseg/x40/1536/dense-l25/fineweb-edu-100bt/384-512/lo}{$-$0.00011}
\defval{kvseg/x40/1536/dense-l25/fineweb-edu-100bt/384-512/round}{0.7880}
\defval{kvseg/x40/1536/dense-l25/fineweb-edu-100bt/896-1024/delta}{+0.00014}
\defval{kvseg/x40/1536/dense-l25/fineweb-edu-100bt/896-1024/hi}{+0.00040}
\defval{kvseg/x40/1536/dense-l25/fineweb-edu-100bt/896-1024/last}{0.7867}
\defval{kvseg/x40/1536/dense-l25/fineweb-edu-100bt/896-1024/lo}{$-$0.00012}
\defval{kvseg/x40/1536/dense-l25/fineweb-edu-100bt/896-1024/round}{0.7865}
\defval{kvseg/x40/1536/dense-l25/fineweb-edu-100bt/all/delta}{+0.00012}
\defval{kvseg/x40/1536/dense-l25/fineweb-edu-100bt/all/hi}{+0.00024}
\defval{kvseg/x40/1536/dense-l25/fineweb-edu-100bt/all/last}{0.8164}
\defval{kvseg/x40/1536/dense-l25/fineweb-edu-100bt/all/lo}{+0.00001}
\defval{kvseg/x40/1536/dense-l25/fineweb-edu-100bt/all/round}{0.8162}
\defval{kvseg/x40/1536/dense-l25/govreport-test/1-128/delta}{+0.00004}
\defval{kvseg/x40/1536/dense-l25/govreport-test/1-128/hi}{+0.00038}
\defval{kvseg/x40/1536/dense-l25/govreport-test/1-128/last}{0.8195}
\defval{kvseg/x40/1536/dense-l25/govreport-test/1-128/lo}{$-$0.00037}
\defval{kvseg/x40/1536/dense-l25/govreport-test/1-128/round}{0.8195}
\defval{kvseg/x40/1536/dense-l25/govreport-test/128-256/delta}{+0.00002}
\defval{kvseg/x40/1536/dense-l25/govreport-test/128-256/hi}{+0.00028}
\defval{kvseg/x40/1536/dense-l25/govreport-test/128-256/last}{0.6903}
\defval{kvseg/x40/1536/dense-l25/govreport-test/128-256/lo}{$-$0.00025}
\defval{kvseg/x40/1536/dense-l25/govreport-test/128-256/round}{0.6903}
\defval{kvseg/x40/1536/dense-l25/govreport-test/1920-2048/delta}{+0.00033}
\defval{kvseg/x40/1536/dense-l25/govreport-test/1920-2048/hi}{+0.00056}
\defval{kvseg/x40/1536/dense-l25/govreport-test/1920-2048/last}{0.6211}
\defval{kvseg/x40/1536/dense-l25/govreport-test/1920-2048/lo}{+0.00011}
\defval{kvseg/x40/1536/dense-l25/govreport-test/1920-2048/round}{0.6208}
\defval{kvseg/x40/1536/dense-l25/govreport-test/384-512/delta}{$-$0.00006}
\defval{kvseg/x40/1536/dense-l25/govreport-test/384-512/hi}{+0.00023}
\defval{kvseg/x40/1536/dense-l25/govreport-test/384-512/last}{0.6499}
\defval{kvseg/x40/1536/dense-l25/govreport-test/384-512/lo}{$-$0.00032}
\defval{kvseg/x40/1536/dense-l25/govreport-test/384-512/round}{0.6500}
\defval{kvseg/x40/1536/dense-l25/govreport-test/896-1024/delta}{+0.00015}
\defval{kvseg/x40/1536/dense-l25/govreport-test/896-1024/hi}{+0.00037}
\defval{kvseg/x40/1536/dense-l25/govreport-test/896-1024/last}{0.6312}
\defval{kvseg/x40/1536/dense-l25/govreport-test/896-1024/lo}{$-$0.00006}
\defval{kvseg/x40/1536/dense-l25/govreport-test/896-1024/round}{0.6310}
\defval{kvseg/x40/1536/dense-l25/govreport-test/all/delta}{+0.00010}
\defval{kvseg/x40/1536/dense-l25/govreport-test/all/hi}{+0.00022}
\defval{kvseg/x40/1536/dense-l25/govreport-test/all/last}{0.6824}
\defval{kvseg/x40/1536/dense-l25/govreport-test/all/lo}{$-$0.00004}
\defval{kvseg/x40/1536/dense-l25/govreport-test/all/round}{0.6823}
\defval{kvseg/x40/1536/dense-l25/longdocs/1-128/delta}{$-$0.00027}
\defval{kvseg/x40/1536/dense-l25/longdocs/1-128/hi}{$-$0.00004}
\defval{kvseg/x40/1536/dense-l25/longdocs/1-128/last}{1.1510}
\defval{kvseg/x40/1536/dense-l25/longdocs/1-128/lo}{$-$0.00053}
\defval{kvseg/x40/1536/dense-l25/longdocs/1-128/round}{1.1513}
\defval{kvseg/x40/1536/dense-l25/longdocs/128-256/delta}{$-$0.00015}
\defval{kvseg/x40/1536/dense-l25/longdocs/128-256/hi}{+0.00008}
\defval{kvseg/x40/1536/dense-l25/longdocs/128-256/last}{0.9797}
\defval{kvseg/x40/1536/dense-l25/longdocs/128-256/lo}{$-$0.00034}
\defval{kvseg/x40/1536/dense-l25/longdocs/128-256/round}{0.9798}
\defval{kvseg/x40/1536/dense-l25/longdocs/1920-2048/delta}{+0.00013}
\defval{kvseg/x40/1536/dense-l25/longdocs/1920-2048/hi}{+0.00032}
\defval{kvseg/x40/1536/dense-l25/longdocs/1920-2048/last}{0.8905}
\defval{kvseg/x40/1536/dense-l25/longdocs/1920-2048/lo}{$-$0.00009}
\defval{kvseg/x40/1536/dense-l25/longdocs/1920-2048/round}{0.8904}
\defval{kvseg/x40/1536/dense-l25/longdocs/384-512/delta}{$-$0.00002}
\defval{kvseg/x40/1536/dense-l25/longdocs/384-512/hi}{+0.00018}
\defval{kvseg/x40/1536/dense-l25/longdocs/384-512/last}{0.9328}
\defval{kvseg/x40/1536/dense-l25/longdocs/384-512/lo}{$-$0.00023}
\defval{kvseg/x40/1536/dense-l25/longdocs/384-512/round}{0.9329}
\defval{kvseg/x40/1536/dense-l25/longdocs/896-1024/delta}{+0.00020}
\defval{kvseg/x40/1536/dense-l25/longdocs/896-1024/hi}{+0.00040}
\defval{kvseg/x40/1536/dense-l25/longdocs/896-1024/last}{0.9178}
\defval{kvseg/x40/1536/dense-l25/longdocs/896-1024/lo}{$-$0.00001}
\defval{kvseg/x40/1536/dense-l25/longdocs/896-1024/round}{0.9176}
\defval{kvseg/x40/1536/dense-l25/longdocs/all/delta}{$-$0.00002}
\defval{kvseg/x40/1536/dense-l25/longdocs/all/hi}{+0.00007}
\defval{kvseg/x40/1536/dense-l25/longdocs/all/last}{0.9744}
\defval{kvseg/x40/1536/dense-l25/longdocs/all/lo}{$-$0.00013}
\defval{kvseg/x40/1536/dense-l25/longdocs/all/round}{0.9744}
\defval{kvseg/x40/1536/dense-l25/pg19-test/1-128/delta}{$-$0.00060}
\defval{kvseg/x40/1536/dense-l25/pg19-test/1-128/hi}{$-$0.00015}
\defval{kvseg/x40/1536/dense-l25/pg19-test/1-128/last}{1.4118}
\defval{kvseg/x40/1536/dense-l25/pg19-test/1-128/lo}{$-$0.00105}
\defval{kvseg/x40/1536/dense-l25/pg19-test/1-128/round}{1.4124}
\defval{kvseg/x40/1536/dense-l25/pg19-test/128-256/delta}{$-$0.00034}
\defval{kvseg/x40/1536/dense-l25/pg19-test/128-256/hi}{+0.00010}
\defval{kvseg/x40/1536/dense-l25/pg19-test/128-256/last}{1.2385}
\defval{kvseg/x40/1536/dense-l25/pg19-test/128-256/lo}{$-$0.00076}
\defval{kvseg/x40/1536/dense-l25/pg19-test/128-256/round}{1.2389}
\defval{kvseg/x40/1536/dense-l25/pg19-test/1920-2048/delta}{$-$0.00015}
\defval{kvseg/x40/1536/dense-l25/pg19-test/1920-2048/hi}{+0.00024}
\defval{kvseg/x40/1536/dense-l25/pg19-test/1920-2048/last}{1.1701}
\defval{kvseg/x40/1536/dense-l25/pg19-test/1920-2048/lo}{$-$0.00058}
\defval{kvseg/x40/1536/dense-l25/pg19-test/1920-2048/round}{1.1702}
\defval{kvseg/x40/1536/dense-l25/pg19-test/384-512/delta}{$-$0.00015}
\defval{kvseg/x40/1536/dense-l25/pg19-test/384-512/hi}{+0.00027}
\defval{kvseg/x40/1536/dense-l25/pg19-test/384-512/last}{1.2001}
\defval{kvseg/x40/1536/dense-l25/pg19-test/384-512/lo}{$-$0.00058}
\defval{kvseg/x40/1536/dense-l25/pg19-test/384-512/round}{1.2002}
\defval{kvseg/x40/1536/dense-l25/pg19-test/896-1024/delta}{+0.00041}
\defval{kvseg/x40/1536/dense-l25/pg19-test/896-1024/hi}{+0.00086}
\defval{kvseg/x40/1536/dense-l25/pg19-test/896-1024/last}{1.1937}
\defval{kvseg/x40/1536/dense-l25/pg19-test/896-1024/lo}{$-$0.00002}
\defval{kvseg/x40/1536/dense-l25/pg19-test/896-1024/round}{1.1933}
\defval{kvseg/x40/1536/dense-l25/pg19-test/all/delta}{$-$0.00016}
\defval{kvseg/x40/1536/dense-l25/pg19-test/all/hi}{+0.00003}
\defval{kvseg/x40/1536/dense-l25/pg19-test/all/last}{1.2424}
\defval{kvseg/x40/1536/dense-l25/pg19-test/all/lo}{$-$0.00037}
\defval{kvseg/x40/1536/dense-l25/pg19-test/all/round}{1.2426}
\defval{kvseg/x40/1536/dense-l25/pg19-val/1-128/delta}{$-$0.00090}
\defval{kvseg/x40/1536/dense-l25/pg19-val/1-128/hi}{$-$0.00051}
\defval{kvseg/x40/1536/dense-l25/pg19-val/1-128/last}{1.3648}
\defval{kvseg/x40/1536/dense-l25/pg19-val/1-128/lo}{$-$0.00130}
\defval{kvseg/x40/1536/dense-l25/pg19-val/1-128/round}{1.3657}
\defval{kvseg/x40/1536/dense-l25/pg19-val/128-256/delta}{$-$0.00051}
\defval{kvseg/x40/1536/dense-l25/pg19-val/128-256/hi}{$-$0.00015}
\defval{kvseg/x40/1536/dense-l25/pg19-val/128-256/last}{1.2027}
\defval{kvseg/x40/1536/dense-l25/pg19-val/128-256/lo}{$-$0.00086}
\defval{kvseg/x40/1536/dense-l25/pg19-val/128-256/round}{1.2032}
\defval{kvseg/x40/1536/dense-l25/pg19-val/1920-2048/delta}{$-$0.00021}
\defval{kvseg/x40/1536/dense-l25/pg19-val/1920-2048/hi}{+0.00024}
\defval{kvseg/x40/1536/dense-l25/pg19-val/1920-2048/last}{1.1303}
\defval{kvseg/x40/1536/dense-l25/pg19-val/1920-2048/lo}{$-$0.00061}
\defval{kvseg/x40/1536/dense-l25/pg19-val/1920-2048/round}{1.1305}
\defval{kvseg/x40/1536/dense-l25/pg19-val/384-512/delta}{$-$0.00016}
\defval{kvseg/x40/1536/dense-l25/pg19-val/384-512/hi}{+0.00022}
\defval{kvseg/x40/1536/dense-l25/pg19-val/384-512/last}{1.1726}
\defval{kvseg/x40/1536/dense-l25/pg19-val/384-512/lo}{$-$0.00054}
\defval{kvseg/x40/1536/dense-l25/pg19-val/384-512/round}{1.1727}
\defval{kvseg/x40/1536/dense-l25/pg19-val/896-1024/delta}{+0.00001}
\defval{kvseg/x40/1536/dense-l25/pg19-val/896-1024/hi}{+0.00042}
\defval{kvseg/x40/1536/dense-l25/pg19-val/896-1024/last}{1.1473}
\defval{kvseg/x40/1536/dense-l25/pg19-val/896-1024/lo}{$-$0.00039}
\defval{kvseg/x40/1536/dense-l25/pg19-val/896-1024/round}{1.1473}
\defval{kvseg/x40/1536/dense-l25/pg19-val/all/delta}{$-$0.00036}
\defval{kvseg/x40/1536/dense-l25/pg19-val/all/hi}{$-$0.00016}
\defval{kvseg/x40/1536/dense-l25/pg19-val/all/last}{1.2034}
\defval{kvseg/x40/1536/dense-l25/pg19-val/all/lo}{$-$0.00055}
\defval{kvseg/x40/1536/dense-l25/pg19-val/all/round}{1.2038}
\defval{kvseg/x40/768/ced-dec/finepdfs-long/1-128/delta}{+0.00015}
\defval{kvseg/x40/768/ced-dec/finepdfs-long/1-128/hi}{+0.00027}
\defval{kvseg/x40/768/ced-dec/finepdfs-long/1-128/last}{1.4590}
\defval{kvseg/x40/768/ced-dec/finepdfs-long/1-128/lo}{+0.00002}
\defval{kvseg/x40/768/ced-dec/finepdfs-long/1-128/round}{1.4589}
\defval{kvseg/x40/768/ced-dec/finepdfs-long/128-256/delta}{+0.00009}
\defval{kvseg/x40/768/ced-dec/finepdfs-long/128-256/hi}{+0.00020}
\defval{kvseg/x40/768/ced-dec/finepdfs-long/128-256/last}{1.2376}
\defval{kvseg/x40/768/ced-dec/finepdfs-long/128-256/lo}{$-$0.00000}
\defval{kvseg/x40/768/ced-dec/finepdfs-long/128-256/round}{1.2375}
\defval{kvseg/x40/768/ced-dec/finepdfs-long/1920-2048/delta}{+0.00020}
\defval{kvseg/x40/768/ced-dec/finepdfs-long/1920-2048/hi}{+0.00028}
\defval{kvseg/x40/768/ced-dec/finepdfs-long/1920-2048/last}{1.1141}
\defval{kvseg/x40/768/ced-dec/finepdfs-long/1920-2048/lo}{+0.00011}
\defval{kvseg/x40/768/ced-dec/finepdfs-long/1920-2048/round}{1.1139}
\defval{kvseg/x40/768/ced-dec/finepdfs-long/384-512/delta}{+0.00010}
\defval{kvseg/x40/768/ced-dec/finepdfs-long/384-512/hi}{+0.00020}
\defval{kvseg/x40/768/ced-dec/finepdfs-long/384-512/last}{1.1823}
\defval{kvseg/x40/768/ced-dec/finepdfs-long/384-512/lo}{+0.00001}
\defval{kvseg/x40/768/ced-dec/finepdfs-long/384-512/round}{1.1822}
\defval{kvseg/x40/768/ced-dec/finepdfs-long/896-1024/delta}{+0.00016}
\defval{kvseg/x40/768/ced-dec/finepdfs-long/896-1024/hi}{+0.00025}
\defval{kvseg/x40/768/ced-dec/finepdfs-long/896-1024/last}{1.1666}
\defval{kvseg/x40/768/ced-dec/finepdfs-long/896-1024/lo}{+0.00007}
\defval{kvseg/x40/768/ced-dec/finepdfs-long/896-1024/round}{1.1664}
\defval{kvseg/x40/768/ced-dec/finepdfs-long/all/delta}{+0.00014}
\defval{kvseg/x40/768/ced-dec/finepdfs-long/all/hi}{+0.00018}
\defval{kvseg/x40/768/ced-dec/finepdfs-long/all/last}{1.2325}
\defval{kvseg/x40/768/ced-dec/finepdfs-long/all/lo}{+0.00010}
\defval{kvseg/x40/768/ced-dec/finepdfs-long/all/round}{1.2324}
\defval{kvseg/x40/768/ced-dec/fineweb-edu-100bt/1-128/delta}{$-$0.00001}
\defval{kvseg/x40/768/ced-dec/fineweb-edu-100bt/1-128/hi}{+0.00010}
\defval{kvseg/x40/768/ced-dec/fineweb-edu-100bt/1-128/last}{1.0168}
\defval{kvseg/x40/768/ced-dec/fineweb-edu-100bt/1-128/lo}{$-$0.00014}
\defval{kvseg/x40/768/ced-dec/fineweb-edu-100bt/1-128/round}{1.0168}
\defval{kvseg/x40/768/ced-dec/fineweb-edu-100bt/128-256/delta}{$-$0.00001}
\defval{kvseg/x40/768/ced-dec/fineweb-edu-100bt/128-256/hi}{+0.00009}
\defval{kvseg/x40/768/ced-dec/fineweb-edu-100bt/128-256/last}{0.9217}
\defval{kvseg/x40/768/ced-dec/fineweb-edu-100bt/128-256/lo}{$-$0.00011}
\defval{kvseg/x40/768/ced-dec/fineweb-edu-100bt/128-256/round}{0.9217}
\defval{kvseg/x40/768/ced-dec/fineweb-edu-100bt/1920-2048/delta}{+0.00007}
\defval{kvseg/x40/768/ced-dec/fineweb-edu-100bt/1920-2048/hi}{+0.00016}
\defval{kvseg/x40/768/ced-dec/fineweb-edu-100bt/1920-2048/last}{0.9011}
\defval{kvseg/x40/768/ced-dec/fineweb-edu-100bt/1920-2048/lo}{$-$0.00002}
\defval{kvseg/x40/768/ced-dec/fineweb-edu-100bt/1920-2048/round}{0.9010}
\defval{kvseg/x40/768/ced-dec/fineweb-edu-100bt/384-512/delta}{$-$0.00002}
\defval{kvseg/x40/768/ced-dec/fineweb-edu-100bt/384-512/hi}{+0.00007}
\defval{kvseg/x40/768/ced-dec/fineweb-edu-100bt/384-512/last}{0.9061}
\defval{kvseg/x40/768/ced-dec/fineweb-edu-100bt/384-512/lo}{$-$0.00011}
\defval{kvseg/x40/768/ced-dec/fineweb-edu-100bt/384-512/round}{0.9061}
\defval{kvseg/x40/768/ced-dec/fineweb-edu-100bt/896-1024/delta}{+0.00009}
\defval{kvseg/x40/768/ced-dec/fineweb-edu-100bt/896-1024/hi}{+0.00018}
\defval{kvseg/x40/768/ced-dec/fineweb-edu-100bt/896-1024/last}{0.9005}
\defval{kvseg/x40/768/ced-dec/fineweb-edu-100bt/896-1024/lo}{+0.00000}
\defval{kvseg/x40/768/ced-dec/fineweb-edu-100bt/896-1024/round}{0.9004}
\defval{kvseg/x40/768/ced-dec/fineweb-edu-100bt/all/delta}{+0.00002}
\defval{kvseg/x40/768/ced-dec/fineweb-edu-100bt/all/hi}{+0.00007}
\defval{kvseg/x40/768/ced-dec/fineweb-edu-100bt/all/last}{0.9290}
\defval{kvseg/x40/768/ced-dec/fineweb-edu-100bt/all/lo}{$-$0.00002}
\defval{kvseg/x40/768/ced-dec/fineweb-edu-100bt/all/round}{0.9290}
\defval{kvseg/x40/768/ced-dec/govreport-test/1-128/delta}{+0.00011}
\defval{kvseg/x40/768/ced-dec/govreport-test/1-128/hi}{+0.00020}
\defval{kvseg/x40/768/ced-dec/govreport-test/1-128/last}{0.9087}
\defval{kvseg/x40/768/ced-dec/govreport-test/1-128/lo}{+0.00002}
\defval{kvseg/x40/768/ced-dec/govreport-test/1-128/round}{0.9086}
\defval{kvseg/x40/768/ced-dec/govreport-test/128-256/delta}{+0.00013}
\defval{kvseg/x40/768/ced-dec/govreport-test/128-256/hi}{+0.00020}
\defval{kvseg/x40/768/ced-dec/govreport-test/128-256/last}{0.7884}
\defval{kvseg/x40/768/ced-dec/govreport-test/128-256/lo}{+0.00007}
\defval{kvseg/x40/768/ced-dec/govreport-test/128-256/round}{0.7882}
\defval{kvseg/x40/768/ced-dec/govreport-test/1920-2048/delta}{+0.00010}
\defval{kvseg/x40/768/ced-dec/govreport-test/1920-2048/hi}{+0.00017}
\defval{kvseg/x40/768/ced-dec/govreport-test/1920-2048/last}{0.7258}
\defval{kvseg/x40/768/ced-dec/govreport-test/1920-2048/lo}{+0.00003}
\defval{kvseg/x40/768/ced-dec/govreport-test/1920-2048/round}{0.7257}
\defval{kvseg/x40/768/ced-dec/govreport-test/384-512/delta}{+0.00009}
\defval{kvseg/x40/768/ced-dec/govreport-test/384-512/hi}{+0.00015}
\defval{kvseg/x40/768/ced-dec/govreport-test/384-512/last}{0.7521}
\defval{kvseg/x40/768/ced-dec/govreport-test/384-512/lo}{+0.00002}
\defval{kvseg/x40/768/ced-dec/govreport-test/384-512/round}{0.7520}
\defval{kvseg/x40/768/ced-dec/govreport-test/896-1024/delta}{+0.00004}
\defval{kvseg/x40/768/ced-dec/govreport-test/896-1024/hi}{+0.00010}
\defval{kvseg/x40/768/ced-dec/govreport-test/896-1024/last}{0.7335}
\defval{kvseg/x40/768/ced-dec/govreport-test/896-1024/lo}{$-$0.00002}
\defval{kvseg/x40/768/ced-dec/govreport-test/896-1024/round}{0.7335}
\defval{kvseg/x40/768/ced-dec/govreport-test/all/delta}{+0.00009}
\defval{kvseg/x40/768/ced-dec/govreport-test/all/hi}{+0.00013}
\defval{kvseg/x40/768/ced-dec/govreport-test/all/last}{0.7817}
\defval{kvseg/x40/768/ced-dec/govreport-test/all/lo}{+0.00007}
\defval{kvseg/x40/768/ced-dec/govreport-test/all/round}{0.7816}
\defval{kvseg/x40/768/ced-dec/longdocs/1-128/delta}{+0.00001}
\defval{kvseg/x40/768/ced-dec/longdocs/1-128/hi}{+0.00007}
\defval{kvseg/x40/768/ced-dec/longdocs/1-128/last}{1.2741}
\defval{kvseg/x40/768/ced-dec/longdocs/1-128/lo}{$-$0.00005}
\defval{kvseg/x40/768/ced-dec/longdocs/1-128/round}{1.2741}
\defval{kvseg/x40/768/ced-dec/longdocs/128-256/delta}{+0.00009}
\defval{kvseg/x40/768/ced-dec/longdocs/128-256/hi}{+0.00014}
\defval{kvseg/x40/768/ced-dec/longdocs/128-256/last}{1.1141}
\defval{kvseg/x40/768/ced-dec/longdocs/128-256/lo}{+0.00004}
\defval{kvseg/x40/768/ced-dec/longdocs/128-256/round}{1.1140}
\defval{kvseg/x40/768/ced-dec/longdocs/1920-2048/delta}{+0.00016}
\defval{kvseg/x40/768/ced-dec/longdocs/1920-2048/hi}{+0.00021}
\defval{kvseg/x40/768/ced-dec/longdocs/1920-2048/last}{1.0334}
\defval{kvseg/x40/768/ced-dec/longdocs/1920-2048/lo}{+0.00011}
\defval{kvseg/x40/768/ced-dec/longdocs/1920-2048/round}{1.0333}
\defval{kvseg/x40/768/ced-dec/longdocs/384-512/delta}{+0.00008}
\defval{kvseg/x40/768/ced-dec/longdocs/384-512/hi}{+0.00013}
\defval{kvseg/x40/768/ced-dec/longdocs/384-512/last}{1.0736}
\defval{kvseg/x40/768/ced-dec/longdocs/384-512/lo}{+0.00003}
\defval{kvseg/x40/768/ced-dec/longdocs/384-512/round}{1.0735}
\defval{kvseg/x40/768/ced-dec/longdocs/896-1024/delta}{+0.00014}
\defval{kvseg/x40/768/ced-dec/longdocs/896-1024/hi}{+0.00019}
\defval{kvseg/x40/768/ced-dec/longdocs/896-1024/last}{1.0593}
\defval{kvseg/x40/768/ced-dec/longdocs/896-1024/lo}{+0.00009}
\defval{kvseg/x40/768/ced-dec/longdocs/896-1024/round}{1.0592}
\defval{kvseg/x40/768/ced-dec/longdocs/all/delta}{+0.00010}
\defval{kvseg/x40/768/ced-dec/longdocs/all/hi}{+0.00012}
\defval{kvseg/x40/768/ced-dec/longdocs/all/last}{1.1109}
\defval{kvseg/x40/768/ced-dec/longdocs/all/lo}{+0.00007}
\defval{kvseg/x40/768/ced-dec/longdocs/all/round}{1.1108}
\defval{kvseg/x40/768/ced-dec/pg19-test/1-128/delta}{$-$0.00024}
\defval{kvseg/x40/768/ced-dec/pg19-test/1-128/hi}{$-$0.00012}
\defval{kvseg/x40/768/ced-dec/pg19-test/1-128/last}{1.5684}
\defval{kvseg/x40/768/ced-dec/pg19-test/1-128/lo}{$-$0.00036}
\defval{kvseg/x40/768/ced-dec/pg19-test/1-128/round}{1.5686}
\defval{kvseg/x40/768/ced-dec/pg19-test/128-256/delta}{+0.00002}
\defval{kvseg/x40/768/ced-dec/pg19-test/128-256/hi}{+0.00014}
\defval{kvseg/x40/768/ced-dec/pg19-test/128-256/last}{1.4123}
\defval{kvseg/x40/768/ced-dec/pg19-test/128-256/lo}{$-$0.00009}
\defval{kvseg/x40/768/ced-dec/pg19-test/128-256/round}{1.4123}
\defval{kvseg/x40/768/ced-dec/pg19-test/1920-2048/delta}{+0.00021}
\defval{kvseg/x40/768/ced-dec/pg19-test/1920-2048/hi}{+0.00031}
\defval{kvseg/x40/768/ced-dec/pg19-test/1920-2048/last}{1.3534}
\defval{kvseg/x40/768/ced-dec/pg19-test/1920-2048/lo}{+0.00010}
\defval{kvseg/x40/768/ced-dec/pg19-test/1920-2048/round}{1.3531}
\defval{kvseg/x40/768/ced-dec/pg19-test/384-512/delta}{+0.00006}
\defval{kvseg/x40/768/ced-dec/pg19-test/384-512/hi}{+0.00017}
\defval{kvseg/x40/768/ced-dec/pg19-test/384-512/last}{1.3823}
\defval{kvseg/x40/768/ced-dec/pg19-test/384-512/lo}{$-$0.00005}
\defval{kvseg/x40/768/ced-dec/pg19-test/384-512/round}{1.3822}
\defval{kvseg/x40/768/ced-dec/pg19-test/896-1024/delta}{+0.00024}
\defval{kvseg/x40/768/ced-dec/pg19-test/896-1024/hi}{+0.00035}
\defval{kvseg/x40/768/ced-dec/pg19-test/896-1024/last}{1.3766}
\defval{kvseg/x40/768/ced-dec/pg19-test/896-1024/lo}{+0.00014}
\defval{kvseg/x40/768/ced-dec/pg19-test/896-1024/round}{1.3764}
\defval{kvseg/x40/768/ced-dec/pg19-test/all/delta}{+0.00006}
\defval{kvseg/x40/768/ced-dec/pg19-test/all/hi}{+0.00011}
\defval{kvseg/x40/768/ced-dec/pg19-test/all/last}{1.4182}
\defval{kvseg/x40/768/ced-dec/pg19-test/all/lo}{+0.00001}
\defval{kvseg/x40/768/ced-dec/pg19-test/all/round}{1.4182}
\defval{kvseg/x40/768/ced-dec/pg19-val/1-128/delta}{$-$0.00020}
\defval{kvseg/x40/768/ced-dec/pg19-val/1-128/hi}{$-$0.00008}
\defval{kvseg/x40/768/ced-dec/pg19-val/1-128/last}{1.5132}
\defval{kvseg/x40/768/ced-dec/pg19-val/1-128/lo}{$-$0.00033}
\defval{kvseg/x40/768/ced-dec/pg19-val/1-128/round}{1.5134}
\defval{kvseg/x40/768/ced-dec/pg19-val/128-256/delta}{+0.00003}
\defval{kvseg/x40/768/ced-dec/pg19-val/128-256/hi}{+0.00015}
\defval{kvseg/x40/768/ced-dec/pg19-val/128-256/last}{1.3731}
\defval{kvseg/x40/768/ced-dec/pg19-val/128-256/lo}{$-$0.00007}
\defval{kvseg/x40/768/ced-dec/pg19-val/128-256/round}{1.3731}
\defval{kvseg/x40/768/ced-dec/pg19-val/1920-2048/delta}{+0.00019}
\defval{kvseg/x40/768/ced-dec/pg19-val/1920-2048/hi}{+0.00028}
\defval{kvseg/x40/768/ced-dec/pg19-val/1920-2048/last}{1.3112}
\defval{kvseg/x40/768/ced-dec/pg19-val/1920-2048/lo}{+0.00009}
\defval{kvseg/x40/768/ced-dec/pg19-val/1920-2048/round}{1.3110}
\defval{kvseg/x40/768/ced-dec/pg19-val/384-512/delta}{$-$0.00004}
\defval{kvseg/x40/768/ced-dec/pg19-val/384-512/hi}{+0.00007}
\defval{kvseg/x40/768/ced-dec/pg19-val/384-512/last}{1.3474}
\defval{kvseg/x40/768/ced-dec/pg19-val/384-512/lo}{$-$0.00014}
\defval{kvseg/x40/768/ced-dec/pg19-val/384-512/round}{1.3474}
\defval{kvseg/x40/768/ced-dec/pg19-val/896-1024/delta}{+0.00009}
\defval{kvseg/x40/768/ced-dec/pg19-val/896-1024/hi}{+0.00020}
\defval{kvseg/x40/768/ced-dec/pg19-val/896-1024/last}{1.3277}
\defval{kvseg/x40/768/ced-dec/pg19-val/896-1024/lo}{$-$0.00001}
\defval{kvseg/x40/768/ced-dec/pg19-val/896-1024/round}{1.3276}
\defval{kvseg/x40/768/ced-dec/pg19-val/all/delta}{+0.00002}
\defval{kvseg/x40/768/ced-dec/pg19-val/all/hi}{+0.00007}
\defval{kvseg/x40/768/ced-dec/pg19-val/all/last}{1.3744}
\defval{kvseg/x40/768/ced-dec/pg19-val/all/lo}{$-$0.00004}
\defval{kvseg/x40/768/ced-dec/pg19-val/all/round}{1.3744}
\defval{kvseg/x40/768/ced-enc/finepdfs-long/1-128/delta}{$-$0.00010}
\defval{kvseg/x40/768/ced-enc/finepdfs-long/1-128/hi}{+0.00031}
\defval{kvseg/x40/768/ced-enc/finepdfs-long/1-128/last}{1.4390}
\defval{kvseg/x40/768/ced-enc/finepdfs-long/1-128/lo}{$-$0.00051}
\defval{kvseg/x40/768/ced-enc/finepdfs-long/1-128/round}{1.4391}
\defval{kvseg/x40/768/ced-enc/finepdfs-long/128-256/delta}{$-$0.00075}
\defval{kvseg/x40/768/ced-enc/finepdfs-long/128-256/hi}{$-$0.00038}
\defval{kvseg/x40/768/ced-enc/finepdfs-long/128-256/last}{1.2126}
\defval{kvseg/x40/768/ced-enc/finepdfs-long/128-256/lo}{$-$0.00110}
\defval{kvseg/x40/768/ced-enc/finepdfs-long/128-256/round}{1.2134}
\defval{kvseg/x40/768/ced-enc/finepdfs-long/1920-2048/delta}{$-$0.00072}
\defval{kvseg/x40/768/ced-enc/finepdfs-long/1920-2048/hi}{$-$0.00031}
\defval{kvseg/x40/768/ced-enc/finepdfs-long/1920-2048/last}{1.0771}
\defval{kvseg/x40/768/ced-enc/finepdfs-long/1920-2048/lo}{$-$0.00118}
\defval{kvseg/x40/768/ced-enc/finepdfs-long/1920-2048/round}{1.0779}
\defval{kvseg/x40/768/ced-enc/finepdfs-long/384-512/delta}{$-$0.00067}
\defval{kvseg/x40/768/ced-enc/finepdfs-long/384-512/hi}{$-$0.00027}
\defval{kvseg/x40/768/ced-enc/finepdfs-long/384-512/last}{1.1531}
\defval{kvseg/x40/768/ced-enc/finepdfs-long/384-512/lo}{$-$0.00107}
\defval{kvseg/x40/768/ced-enc/finepdfs-long/384-512/round}{1.1538}
\defval{kvseg/x40/768/ced-enc/finepdfs-long/896-1024/delta}{$-$0.00069}
\defval{kvseg/x40/768/ced-enc/finepdfs-long/896-1024/hi}{$-$0.00022}
\defval{kvseg/x40/768/ced-enc/finepdfs-long/896-1024/last}{1.1355}
\defval{kvseg/x40/768/ced-enc/finepdfs-long/896-1024/lo}{$-$0.00110}
\defval{kvseg/x40/768/ced-enc/finepdfs-long/896-1024/round}{1.1362}
\defval{kvseg/x40/768/ced-enc/finepdfs-long/all/delta}{$-$0.00059}
\defval{kvseg/x40/768/ced-enc/finepdfs-long/all/hi}{$-$0.00040}
\defval{kvseg/x40/768/ced-enc/finepdfs-long/all/last}{1.2041}
\defval{kvseg/x40/768/ced-enc/finepdfs-long/all/lo}{$-$0.00077}
\defval{kvseg/x40/768/ced-enc/finepdfs-long/all/round}{1.2047}
\defval{kvseg/x40/768/ced-enc/fineweb-edu-100bt/1-128/delta}{$-$0.00013}
\defval{kvseg/x40/768/ced-enc/fineweb-edu-100bt/1-128/hi}{+0.00014}
\defval{kvseg/x40/768/ced-enc/fineweb-edu-100bt/1-128/last}{1.0027}
\defval{kvseg/x40/768/ced-enc/fineweb-edu-100bt/1-128/lo}{$-$0.00041}
\defval{kvseg/x40/768/ced-enc/fineweb-edu-100bt/1-128/round}{1.0028}
\defval{kvseg/x40/768/ced-enc/fineweb-edu-100bt/128-256/delta}{$-$0.00019}
\defval{kvseg/x40/768/ced-enc/fineweb-edu-100bt/128-256/hi}{+0.00008}
\defval{kvseg/x40/768/ced-enc/fineweb-edu-100bt/128-256/last}{0.9059}
\defval{kvseg/x40/768/ced-enc/fineweb-edu-100bt/128-256/lo}{$-$0.00047}
\defval{kvseg/x40/768/ced-enc/fineweb-edu-100bt/128-256/round}{0.9061}
\defval{kvseg/x40/768/ced-enc/fineweb-edu-100bt/1920-2048/delta}{$-$0.00020}
\defval{kvseg/x40/768/ced-enc/fineweb-edu-100bt/1920-2048/hi}{+0.00008}
\defval{kvseg/x40/768/ced-enc/fineweb-edu-100bt/1920-2048/last}{0.8797}
\defval{kvseg/x40/768/ced-enc/fineweb-edu-100bt/1920-2048/lo}{$-$0.00045}
\defval{kvseg/x40/768/ced-enc/fineweb-edu-100bt/1920-2048/round}{0.8799}
\defval{kvseg/x40/768/ced-enc/fineweb-edu-100bt/384-512/delta}{$-$0.00009}
\defval{kvseg/x40/768/ced-enc/fineweb-edu-100bt/384-512/hi}{+0.00018}
\defval{kvseg/x40/768/ced-enc/fineweb-edu-100bt/384-512/last}{0.8868}
\defval{kvseg/x40/768/ced-enc/fineweb-edu-100bt/384-512/lo}{$-$0.00035}
\defval{kvseg/x40/768/ced-enc/fineweb-edu-100bt/384-512/round}{0.8869}
\defval{kvseg/x40/768/ced-enc/fineweb-edu-100bt/896-1024/delta}{$-$0.00008}
\defval{kvseg/x40/768/ced-enc/fineweb-edu-100bt/896-1024/hi}{+0.00014}
\defval{kvseg/x40/768/ced-enc/fineweb-edu-100bt/896-1024/last}{0.8826}
\defval{kvseg/x40/768/ced-enc/fineweb-edu-100bt/896-1024/lo}{$-$0.00030}
\defval{kvseg/x40/768/ced-enc/fineweb-edu-100bt/896-1024/round}{0.8827}
\defval{kvseg/x40/768/ced-enc/fineweb-edu-100bt/all/delta}{$-$0.00014}
\defval{kvseg/x40/768/ced-enc/fineweb-edu-100bt/all/hi}{$-$0.00002}
\defval{kvseg/x40/768/ced-enc/fineweb-edu-100bt/all/last}{0.9113}
\defval{kvseg/x40/768/ced-enc/fineweb-edu-100bt/all/lo}{$-$0.00025}
\defval{kvseg/x40/768/ced-enc/fineweb-edu-100bt/all/round}{0.9115}
\defval{kvseg/x40/768/ced-enc/govreport-test/1-128/delta}{$-$0.00017}
\defval{kvseg/x40/768/ced-enc/govreport-test/1-128/hi}{+0.00004}
\defval{kvseg/x40/768/ced-enc/govreport-test/1-128/last}{0.8937}
\defval{kvseg/x40/768/ced-enc/govreport-test/1-128/lo}{$-$0.00039}
\defval{kvseg/x40/768/ced-enc/govreport-test/1-128/round}{0.8939}
\defval{kvseg/x40/768/ced-enc/govreport-test/128-256/delta}{+0.00004}
\defval{kvseg/x40/768/ced-enc/govreport-test/128-256/hi}{+0.00027}
\defval{kvseg/x40/768/ced-enc/govreport-test/128-256/last}{0.7706}
\defval{kvseg/x40/768/ced-enc/govreport-test/128-256/lo}{$-$0.00020}
\defval{kvseg/x40/768/ced-enc/govreport-test/128-256/round}{0.7706}
\defval{kvseg/x40/768/ced-enc/govreport-test/1920-2048/delta}{$-$0.00020}
\defval{kvseg/x40/768/ced-enc/govreport-test/1920-2048/hi}{+0.00003}
\defval{kvseg/x40/768/ced-enc/govreport-test/1920-2048/last}{0.7036}
\defval{kvseg/x40/768/ced-enc/govreport-test/1920-2048/lo}{$-$0.00046}
\defval{kvseg/x40/768/ced-enc/govreport-test/1920-2048/round}{0.7038}
\defval{kvseg/x40/768/ced-enc/govreport-test/384-512/delta}{$-$0.00034}
\defval{kvseg/x40/768/ced-enc/govreport-test/384-512/hi}{$-$0.00010}
\defval{kvseg/x40/768/ced-enc/govreport-test/384-512/last}{0.7317}
\defval{kvseg/x40/768/ced-enc/govreport-test/384-512/lo}{$-$0.00061}
\defval{kvseg/x40/768/ced-enc/govreport-test/384-512/round}{0.7320}
\defval{kvseg/x40/768/ced-enc/govreport-test/896-1024/delta}{$-$0.00020}
\defval{kvseg/x40/768/ced-enc/govreport-test/896-1024/hi}{+0.00004}
\defval{kvseg/x40/768/ced-enc/govreport-test/896-1024/last}{0.7146}
\defval{kvseg/x40/768/ced-enc/govreport-test/896-1024/lo}{$-$0.00045}
\defval{kvseg/x40/768/ced-enc/govreport-test/896-1024/round}{0.7148}
\defval{kvseg/x40/768/ced-enc/govreport-test/all/delta}{$-$0.00018}
\defval{kvseg/x40/768/ced-enc/govreport-test/all/hi}{$-$0.00008}
\defval{kvseg/x40/768/ced-enc/govreport-test/all/last}{0.7628}
\defval{kvseg/x40/768/ced-enc/govreport-test/all/lo}{$-$0.00029}
\defval{kvseg/x40/768/ced-enc/govreport-test/all/round}{0.7630}
\defval{kvseg/x40/768/ced-enc/longdocs/1-128/delta}{$-$0.00023}
\defval{kvseg/x40/768/ced-enc/longdocs/1-128/hi}{$-$0.00005}
\defval{kvseg/x40/768/ced-enc/longdocs/1-128/last}{1.2548}
\defval{kvseg/x40/768/ced-enc/longdocs/1-128/lo}{$-$0.00043}
\defval{kvseg/x40/768/ced-enc/longdocs/1-128/round}{1.2550}
\defval{kvseg/x40/768/ced-enc/longdocs/128-256/delta}{$-$0.00049}
\defval{kvseg/x40/768/ced-enc/longdocs/128-256/hi}{$-$0.00032}
\defval{kvseg/x40/768/ced-enc/longdocs/128-256/last}{1.0911}
\defval{kvseg/x40/768/ced-enc/longdocs/128-256/lo}{$-$0.00067}
\defval{kvseg/x40/768/ced-enc/longdocs/128-256/round}{1.0916}
\defval{kvseg/x40/768/ced-enc/longdocs/1920-2048/delta}{$-$0.00062}
\defval{kvseg/x40/768/ced-enc/longdocs/1920-2048/hi}{$-$0.00042}
\defval{kvseg/x40/768/ced-enc/longdocs/1920-2048/last}{1.0041}
\defval{kvseg/x40/768/ced-enc/longdocs/1920-2048/lo}{$-$0.00082}
\defval{kvseg/x40/768/ced-enc/longdocs/1920-2048/round}{1.0047}
\defval{kvseg/x40/768/ced-enc/longdocs/384-512/delta}{$-$0.00054}
\defval{kvseg/x40/768/ced-enc/longdocs/384-512/hi}{$-$0.00036}
\defval{kvseg/x40/768/ced-enc/longdocs/384-512/last}{1.0480}
\defval{kvseg/x40/768/ced-enc/longdocs/384-512/lo}{$-$0.00071}
\defval{kvseg/x40/768/ced-enc/longdocs/384-512/round}{1.0486}
\defval{kvseg/x40/768/ced-enc/longdocs/896-1024/delta}{$-$0.00054}
\defval{kvseg/x40/768/ced-enc/longdocs/896-1024/hi}{$-$0.00034}
\defval{kvseg/x40/768/ced-enc/longdocs/896-1024/last}{1.0341}
\defval{kvseg/x40/768/ced-enc/longdocs/896-1024/lo}{$-$0.00075}
\defval{kvseg/x40/768/ced-enc/longdocs/896-1024/round}{1.0347}
\defval{kvseg/x40/768/ced-enc/longdocs/all/delta}{$-$0.00048}
\defval{kvseg/x40/768/ced-enc/longdocs/all/hi}{$-$0.00040}
\defval{kvseg/x40/768/ced-enc/longdocs/all/last}{1.0864}
\defval{kvseg/x40/768/ced-enc/longdocs/all/lo}{$-$0.00057}
\defval{kvseg/x40/768/ced-enc/longdocs/all/round}{1.0869}
\defval{kvseg/x40/768/ced-enc/pg19-test/1-128/delta}{$-$0.00044}
\defval{kvseg/x40/768/ced-enc/pg19-test/1-128/hi}{$-$0.00004}
\defval{kvseg/x40/768/ced-enc/pg19-test/1-128/last}{1.5441}
\defval{kvseg/x40/768/ced-enc/pg19-test/1-128/lo}{$-$0.00088}
\defval{kvseg/x40/768/ced-enc/pg19-test/1-128/round}{1.5446}
\defval{kvseg/x40/768/ced-enc/pg19-test/128-256/delta}{$-$0.00091}
\defval{kvseg/x40/768/ced-enc/pg19-test/128-256/hi}{$-$0.00055}
\defval{kvseg/x40/768/ced-enc/pg19-test/128-256/last}{1.3845}
\defval{kvseg/x40/768/ced-enc/pg19-test/128-256/lo}{$-$0.00129}
\defval{kvseg/x40/768/ced-enc/pg19-test/128-256/round}{1.3854}
\defval{kvseg/x40/768/ced-enc/pg19-test/1920-2048/delta}{$-$0.00105}
\defval{kvseg/x40/768/ced-enc/pg19-test/1920-2048/hi}{$-$0.00066}
\defval{kvseg/x40/768/ced-enc/pg19-test/1920-2048/last}{1.3221}
\defval{kvseg/x40/768/ced-enc/pg19-test/1920-2048/lo}{$-$0.00142}
\defval{kvseg/x40/768/ced-enc/pg19-test/1920-2048/round}{1.3232}
\defval{kvseg/x40/768/ced-enc/pg19-test/384-512/delta}{$-$0.00067}
\defval{kvseg/x40/768/ced-enc/pg19-test/384-512/hi}{$-$0.00032}
\defval{kvseg/x40/768/ced-enc/pg19-test/384-512/last}{1.3536}
\defval{kvseg/x40/768/ced-enc/pg19-test/384-512/lo}{$-$0.00100}
\defval{kvseg/x40/768/ced-enc/pg19-test/384-512/round}{1.3543}
\defval{kvseg/x40/768/ced-enc/pg19-test/896-1024/delta}{$-$0.00083}
\defval{kvseg/x40/768/ced-enc/pg19-test/896-1024/hi}{$-$0.00042}
\defval{kvseg/x40/768/ced-enc/pg19-test/896-1024/last}{1.3490}
\defval{kvseg/x40/768/ced-enc/pg19-test/896-1024/lo}{$-$0.00122}
\defval{kvseg/x40/768/ced-enc/pg19-test/896-1024/round}{1.3498}
\defval{kvseg/x40/768/ced-enc/pg19-test/all/delta}{$-$0.00078}
\defval{kvseg/x40/768/ced-enc/pg19-test/all/hi}{$-$0.00063}
\defval{kvseg/x40/768/ced-enc/pg19-test/all/last}{1.3903}
\defval{kvseg/x40/768/ced-enc/pg19-test/all/lo}{$-$0.00097}
\defval{kvseg/x40/768/ced-enc/pg19-test/all/round}{1.3911}
\defval{kvseg/x40/768/ced-enc/pg19-val/1-128/delta}{$-$0.00026}
\defval{kvseg/x40/768/ced-enc/pg19-val/1-128/hi}{+0.00023}
\defval{kvseg/x40/768/ced-enc/pg19-val/1-128/last}{1.4896}
\defval{kvseg/x40/768/ced-enc/pg19-val/1-128/lo}{$-$0.00074}
\defval{kvseg/x40/768/ced-enc/pg19-val/1-128/round}{1.4899}
\defval{kvseg/x40/768/ced-enc/pg19-val/128-256/delta}{$-$0.00097}
\defval{kvseg/x40/768/ced-enc/pg19-val/128-256/hi}{$-$0.00061}
\defval{kvseg/x40/768/ced-enc/pg19-val/128-256/last}{1.3480}
\defval{kvseg/x40/768/ced-enc/pg19-val/128-256/lo}{$-$0.00135}
\defval{kvseg/x40/768/ced-enc/pg19-val/128-256/round}{1.3489}
\defval{kvseg/x40/768/ced-enc/pg19-val/1920-2048/delta}{$-$0.00077}
\defval{kvseg/x40/768/ced-enc/pg19-val/1920-2048/hi}{$-$0.00032}
\defval{kvseg/x40/768/ced-enc/pg19-val/1920-2048/last}{1.2870}
\defval{kvseg/x40/768/ced-enc/pg19-val/1920-2048/lo}{$-$0.00118}
\defval{kvseg/x40/768/ced-enc/pg19-val/1920-2048/round}{1.2878}
\defval{kvseg/x40/768/ced-enc/pg19-val/384-512/delta}{$-$0.00085}
\defval{kvseg/x40/768/ced-enc/pg19-val/384-512/hi}{$-$0.00051}
\defval{kvseg/x40/768/ced-enc/pg19-val/384-512/last}{1.3236}
\defval{kvseg/x40/768/ced-enc/pg19-val/384-512/lo}{$-$0.00120}
\defval{kvseg/x40/768/ced-enc/pg19-val/384-512/round}{1.3244}
\defval{kvseg/x40/768/ced-enc/pg19-val/896-1024/delta}{$-$0.00071}
\defval{kvseg/x40/768/ced-enc/pg19-val/896-1024/hi}{$-$0.00033}
\defval{kvseg/x40/768/ced-enc/pg19-val/896-1024/last}{1.3009}
\defval{kvseg/x40/768/ced-enc/pg19-val/896-1024/lo}{$-$0.00111}
\defval{kvseg/x40/768/ced-enc/pg19-val/896-1024/round}{1.3016}
\defval{kvseg/x40/768/ced-enc/pg19-val/all/delta}{$-$0.00071}
\defval{kvseg/x40/768/ced-enc/pg19-val/all/hi}{$-$0.00052}
\defval{kvseg/x40/768/ced-enc/pg19-val/all/last}{1.3497}
\defval{kvseg/x40/768/ced-enc/pg19-val/all/lo}{$-$0.00089}
\defval{kvseg/x40/768/ced-enc/pg19-val/all/round}{1.3504}
\defval{kvseg/x40/768/dense-l25/finepdfs-long/1-128/delta}{+0.00029}
\defval{kvseg/x40/768/dense-l25/finepdfs-long/1-128/hi}{+0.00074}
\defval{kvseg/x40/768/dense-l25/finepdfs-long/1-128/last}{1.4500}
\defval{kvseg/x40/768/dense-l25/finepdfs-long/1-128/lo}{$-$0.00012}
\defval{kvseg/x40/768/dense-l25/finepdfs-long/1-128/round}{1.4497}
\defval{kvseg/x40/768/dense-l25/finepdfs-long/128-256/delta}{$-$0.00013}
\defval{kvseg/x40/768/dense-l25/finepdfs-long/128-256/hi}{+0.00021}
\defval{kvseg/x40/768/dense-l25/finepdfs-long/128-256/last}{1.2222}
\defval{kvseg/x40/768/dense-l25/finepdfs-long/128-256/lo}{$-$0.00048}
\defval{kvseg/x40/768/dense-l25/finepdfs-long/128-256/round}{1.2223}
\defval{kvseg/x40/768/dense-l25/finepdfs-long/1920-2048/delta}{$-$0.00016}
\defval{kvseg/x40/768/dense-l25/finepdfs-long/1920-2048/hi}{+0.00029}
\defval{kvseg/x40/768/dense-l25/finepdfs-long/1920-2048/last}{1.0849}
\defval{kvseg/x40/768/dense-l25/finepdfs-long/1920-2048/lo}{$-$0.00062}
\defval{kvseg/x40/768/dense-l25/finepdfs-long/1920-2048/round}{1.0851}
\defval{kvseg/x40/768/dense-l25/finepdfs-long/384-512/delta}{$-$0.00012}
\defval{kvseg/x40/768/dense-l25/finepdfs-long/384-512/hi}{+0.00030}
\defval{kvseg/x40/768/dense-l25/finepdfs-long/384-512/last}{1.1626}
\defval{kvseg/x40/768/dense-l25/finepdfs-long/384-512/lo}{$-$0.00053}
\defval{kvseg/x40/768/dense-l25/finepdfs-long/384-512/round}{1.1627}
\defval{kvseg/x40/768/dense-l25/finepdfs-long/896-1024/delta}{$-$0.00065}
\defval{kvseg/x40/768/dense-l25/finepdfs-long/896-1024/hi}{$-$0.00027}
\defval{kvseg/x40/768/dense-l25/finepdfs-long/896-1024/last}{1.1407}
\defval{kvseg/x40/768/dense-l25/finepdfs-long/896-1024/lo}{$-$0.00101}
\defval{kvseg/x40/768/dense-l25/finepdfs-long/896-1024/round}{1.1413}
\defval{kvseg/x40/768/dense-l25/finepdfs-long/all/delta}{$-$0.00016}
\defval{kvseg/x40/768/dense-l25/finepdfs-long/all/hi}{+0.00005}
\defval{kvseg/x40/768/dense-l25/finepdfs-long/all/last}{1.2127}
\defval{kvseg/x40/768/dense-l25/finepdfs-long/all/lo}{$-$0.00037}
\defval{kvseg/x40/768/dense-l25/finepdfs-long/all/round}{1.2129}
\defval{kvseg/x40/768/dense-l25/fineweb-edu-100bt/1-128/delta}{+0.00008}
\defval{kvseg/x40/768/dense-l25/fineweb-edu-100bt/1-128/hi}{+0.00040}
\defval{kvseg/x40/768/dense-l25/fineweb-edu-100bt/1-128/last}{1.0048}
\defval{kvseg/x40/768/dense-l25/fineweb-edu-100bt/1-128/lo}{$-$0.00023}
\defval{kvseg/x40/768/dense-l25/fineweb-edu-100bt/1-128/round}{1.0047}
\defval{kvseg/x40/768/dense-l25/fineweb-edu-100bt/128-256/delta}{$-$0.00003}
\defval{kvseg/x40/768/dense-l25/fineweb-edu-100bt/128-256/hi}{+0.00022}
\defval{kvseg/x40/768/dense-l25/fineweb-edu-100bt/128-256/last}{0.9059}
\defval{kvseg/x40/768/dense-l25/fineweb-edu-100bt/128-256/lo}{$-$0.00029}
\defval{kvseg/x40/768/dense-l25/fineweb-edu-100bt/128-256/round}{0.9059}
\defval{kvseg/x40/768/dense-l25/fineweb-edu-100bt/1920-2048/delta}{$-$0.00011}
\defval{kvseg/x40/768/dense-l25/fineweb-edu-100bt/1920-2048/hi}{+0.00011}
\defval{kvseg/x40/768/dense-l25/fineweb-edu-100bt/1920-2048/last}{0.8790}
\defval{kvseg/x40/768/dense-l25/fineweb-edu-100bt/1920-2048/lo}{$-$0.00032}
\defval{kvseg/x40/768/dense-l25/fineweb-edu-100bt/1920-2048/round}{0.8791}
\defval{kvseg/x40/768/dense-l25/fineweb-edu-100bt/384-512/delta}{+0.00006}
\defval{kvseg/x40/768/dense-l25/fineweb-edu-100bt/384-512/hi}{+0.00039}
\defval{kvseg/x40/768/dense-l25/fineweb-edu-100bt/384-512/last}{0.8866}
\defval{kvseg/x40/768/dense-l25/fineweb-edu-100bt/384-512/lo}{$-$0.00024}
\defval{kvseg/x40/768/dense-l25/fineweb-edu-100bt/384-512/round}{0.8865}
\defval{kvseg/x40/768/dense-l25/fineweb-edu-100bt/896-1024/delta}{$-$0.00027}
\defval{kvseg/x40/768/dense-l25/fineweb-edu-100bt/896-1024/hi}{$-$0.00001}
\defval{kvseg/x40/768/dense-l25/fineweb-edu-100bt/896-1024/last}{0.8823}
\defval{kvseg/x40/768/dense-l25/fineweb-edu-100bt/896-1024/lo}{$-$0.00053}
\defval{kvseg/x40/768/dense-l25/fineweb-edu-100bt/896-1024/round}{0.8825}
\defval{kvseg/x40/768/dense-l25/fineweb-edu-100bt/all/delta}{$-$0.00005}
\defval{kvseg/x40/768/dense-l25/fineweb-edu-100bt/all/hi}{+0.00007}
\defval{kvseg/x40/768/dense-l25/fineweb-edu-100bt/all/last}{0.9115}
\defval{kvseg/x40/768/dense-l25/fineweb-edu-100bt/all/lo}{$-$0.00018}
\defval{kvseg/x40/768/dense-l25/fineweb-edu-100bt/all/round}{0.9116}
\defval{kvseg/x40/768/dense-l25/govreport-test/1-128/delta}{$-$0.00002}
\defval{kvseg/x40/768/dense-l25/govreport-test/1-128/hi}{+0.00019}
\defval{kvseg/x40/768/dense-l25/govreport-test/1-128/last}{0.8957}
\defval{kvseg/x40/768/dense-l25/govreport-test/1-128/lo}{$-$0.00022}
\defval{kvseg/x40/768/dense-l25/govreport-test/1-128/round}{0.8957}
\defval{kvseg/x40/768/dense-l25/govreport-test/128-256/delta}{$-$0.00007}
\defval{kvseg/x40/768/dense-l25/govreport-test/128-256/hi}{+0.00019}
\defval{kvseg/x40/768/dense-l25/govreport-test/128-256/last}{0.7704}
\defval{kvseg/x40/768/dense-l25/govreport-test/128-256/lo}{$-$0.00031}
\defval{kvseg/x40/768/dense-l25/govreport-test/128-256/round}{0.7705}
\defval{kvseg/x40/768/dense-l25/govreport-test/1920-2048/delta}{$-$0.00014}
\defval{kvseg/x40/768/dense-l25/govreport-test/1920-2048/hi}{+0.00012}
\defval{kvseg/x40/768/dense-l25/govreport-test/1920-2048/last}{0.7016}
\defval{kvseg/x40/768/dense-l25/govreport-test/1920-2048/lo}{$-$0.00039}
\defval{kvseg/x40/768/dense-l25/govreport-test/1920-2048/round}{0.7017}
\defval{kvseg/x40/768/dense-l25/govreport-test/384-512/delta}{$-$0.00021}
\defval{kvseg/x40/768/dense-l25/govreport-test/384-512/hi}{+0.00001}
\defval{kvseg/x40/768/dense-l25/govreport-test/384-512/last}{0.7305}
\defval{kvseg/x40/768/dense-l25/govreport-test/384-512/lo}{$-$0.00043}
\defval{kvseg/x40/768/dense-l25/govreport-test/384-512/round}{0.7307}
\defval{kvseg/x40/768/dense-l25/govreport-test/896-1024/delta}{$-$0.00015}
\defval{kvseg/x40/768/dense-l25/govreport-test/896-1024/hi}{+0.00008}
\defval{kvseg/x40/768/dense-l25/govreport-test/896-1024/last}{0.7118}
\defval{kvseg/x40/768/dense-l25/govreport-test/896-1024/lo}{$-$0.00039}
\defval{kvseg/x40/768/dense-l25/govreport-test/896-1024/round}{0.7120}
\defval{kvseg/x40/768/dense-l25/govreport-test/all/delta}{$-$0.00012}
\defval{kvseg/x40/768/dense-l25/govreport-test/all/hi}{$-$0.00001}
\defval{kvseg/x40/768/dense-l25/govreport-test/all/last}{0.7620}
\defval{kvseg/x40/768/dense-l25/govreport-test/all/lo}{$-$0.00022}
\defval{kvseg/x40/768/dense-l25/govreport-test/all/round}{0.7621}
\defval{kvseg/x40/768/dense-l25/longdocs/1-128/delta}{$-$0.00002}
\defval{kvseg/x40/768/dense-l25/longdocs/1-128/hi}{+0.00018}
\defval{kvseg/x40/768/dense-l25/longdocs/1-128/last}{1.2627}
\defval{kvseg/x40/768/dense-l25/longdocs/1-128/lo}{$-$0.00019}
\defval{kvseg/x40/768/dense-l25/longdocs/1-128/round}{1.2627}
\defval{kvseg/x40/768/dense-l25/longdocs/128-256/delta}{$-$0.00022}
\defval{kvseg/x40/768/dense-l25/longdocs/128-256/hi}{$-$0.00003}
\defval{kvseg/x40/768/dense-l25/longdocs/128-256/last}{1.0962}
\defval{kvseg/x40/768/dense-l25/longdocs/128-256/lo}{$-$0.00041}
\defval{kvseg/x40/768/dense-l25/longdocs/128-256/round}{1.0964}
\defval{kvseg/x40/768/dense-l25/longdocs/1920-2048/delta}{$-$0.00033}
\defval{kvseg/x40/768/dense-l25/longdocs/1920-2048/hi}{$-$0.00013}
\defval{kvseg/x40/768/dense-l25/longdocs/1920-2048/last}{1.0064}
\defval{kvseg/x40/768/dense-l25/longdocs/1920-2048/lo}{$-$0.00053}
\defval{kvseg/x40/768/dense-l25/longdocs/1920-2048/round}{1.0068}
\defval{kvseg/x40/768/dense-l25/longdocs/384-512/delta}{$-$0.00028}
\defval{kvseg/x40/768/dense-l25/longdocs/384-512/hi}{$-$0.00008}
\defval{kvseg/x40/768/dense-l25/longdocs/384-512/last}{1.0513}
\defval{kvseg/x40/768/dense-l25/longdocs/384-512/lo}{$-$0.00047}
\defval{kvseg/x40/768/dense-l25/longdocs/384-512/round}{1.0516}
\defval{kvseg/x40/768/dense-l25/longdocs/896-1024/delta}{$-$0.00037}
\defval{kvseg/x40/768/dense-l25/longdocs/896-1024/hi}{$-$0.00020}
\defval{kvseg/x40/768/dense-l25/longdocs/896-1024/last}{1.0347}
\defval{kvseg/x40/768/dense-l25/longdocs/896-1024/lo}{$-$0.00056}
\defval{kvseg/x40/768/dense-l25/longdocs/896-1024/round}{1.0350}
\defval{kvseg/x40/768/dense-l25/longdocs/all/delta}{$-$0.00024}
\defval{kvseg/x40/768/dense-l25/longdocs/all/hi}{$-$0.00015}
\defval{kvseg/x40/768/dense-l25/longdocs/all/last}{1.0903}
\defval{kvseg/x40/768/dense-l25/longdocs/all/lo}{$-$0.00033}
\defval{kvseg/x40/768/dense-l25/longdocs/all/round}{1.0905}
\defval{kvseg/x40/768/dense-l25/pg19-test/1-128/delta}{$-$0.00031}
\defval{kvseg/x40/768/dense-l25/pg19-test/1-128/hi}{+0.00010}
\defval{kvseg/x40/768/dense-l25/pg19-test/1-128/last}{1.5567}
\defval{kvseg/x40/768/dense-l25/pg19-test/1-128/lo}{$-$0.00069}
\defval{kvseg/x40/768/dense-l25/pg19-test/1-128/round}{1.5570}
\defval{kvseg/x40/768/dense-l25/pg19-test/128-256/delta}{$-$0.00049}
\defval{kvseg/x40/768/dense-l25/pg19-test/128-256/hi}{$-$0.00010}
\defval{kvseg/x40/768/dense-l25/pg19-test/128-256/last}{1.3920}
\defval{kvseg/x40/768/dense-l25/pg19-test/128-256/lo}{$-$0.00087}
\defval{kvseg/x40/768/dense-l25/pg19-test/128-256/round}{1.3925}
\defval{kvseg/x40/768/dense-l25/pg19-test/1920-2048/delta}{$-$0.00075}
\defval{kvseg/x40/768/dense-l25/pg19-test/1920-2048/hi}{$-$0.00037}
\defval{kvseg/x40/768/dense-l25/pg19-test/1920-2048/last}{1.3249}
\defval{kvseg/x40/768/dense-l25/pg19-test/1920-2048/lo}{$-$0.00112}
\defval{kvseg/x40/768/dense-l25/pg19-test/1920-2048/round}{1.3256}
\defval{kvseg/x40/768/dense-l25/pg19-test/384-512/delta}{$-$0.00052}
\defval{kvseg/x40/768/dense-l25/pg19-test/384-512/hi}{$-$0.00015}
\defval{kvseg/x40/768/dense-l25/pg19-test/384-512/last}{1.3567}
\defval{kvseg/x40/768/dense-l25/pg19-test/384-512/lo}{$-$0.00087}
\defval{kvseg/x40/768/dense-l25/pg19-test/384-512/round}{1.3573}
\defval{kvseg/x40/768/dense-l25/pg19-test/896-1024/delta}{$-$0.00040}
\defval{kvseg/x40/768/dense-l25/pg19-test/896-1024/hi}{$-$0.00004}
\defval{kvseg/x40/768/dense-l25/pg19-test/896-1024/last}{1.3493}
\defval{kvseg/x40/768/dense-l25/pg19-test/896-1024/lo}{$-$0.00076}
\defval{kvseg/x40/768/dense-l25/pg19-test/896-1024/round}{1.3497}
\defval{kvseg/x40/768/dense-l25/pg19-test/all/delta}{$-$0.00049}
\defval{kvseg/x40/768/dense-l25/pg19-test/all/hi}{$-$0.00033}
\defval{kvseg/x40/768/dense-l25/pg19-test/all/last}{1.3956}
\defval{kvseg/x40/768/dense-l25/pg19-test/all/lo}{$-$0.00066}
\defval{kvseg/x40/768/dense-l25/pg19-test/all/round}{1.3961}
\defval{kvseg/x40/768/dense-l25/pg19-val/1-128/delta}{$-$0.00017}
\defval{kvseg/x40/768/dense-l25/pg19-val/1-128/hi}{+0.00022}
\defval{kvseg/x40/768/dense-l25/pg19-val/1-128/last}{1.4993}
\defval{kvseg/x40/768/dense-l25/pg19-val/1-128/lo}{$-$0.00060}
\defval{kvseg/x40/768/dense-l25/pg19-val/1-128/round}{1.4994}
\defval{kvseg/x40/768/dense-l25/pg19-val/128-256/delta}{$-$0.00077}
\defval{kvseg/x40/768/dense-l25/pg19-val/128-256/hi}{$-$0.00043}
\defval{kvseg/x40/768/dense-l25/pg19-val/128-256/last}{1.3507}
\defval{kvseg/x40/768/dense-l25/pg19-val/128-256/lo}{$-$0.00109}
\defval{kvseg/x40/768/dense-l25/pg19-val/128-256/round}{1.3515}
\defval{kvseg/x40/768/dense-l25/pg19-val/1920-2048/delta}{$-$0.00078}
\defval{kvseg/x40/768/dense-l25/pg19-val/1920-2048/hi}{$-$0.00039}
\defval{kvseg/x40/768/dense-l25/pg19-val/1920-2048/last}{1.2825}
\defval{kvseg/x40/768/dense-l25/pg19-val/1920-2048/lo}{$-$0.00122}
\defval{kvseg/x40/768/dense-l25/pg19-val/1920-2048/round}{1.2833}
\defval{kvseg/x40/768/dense-l25/pg19-val/384-512/delta}{$-$0.00051}
\defval{kvseg/x40/768/dense-l25/pg19-val/384-512/hi}{$-$0.00013}
\defval{kvseg/x40/768/dense-l25/pg19-val/384-512/last}{1.3212}
\defval{kvseg/x40/768/dense-l25/pg19-val/384-512/lo}{$-$0.00084}
\defval{kvseg/x40/768/dense-l25/pg19-val/384-512/round}{1.3217}
\defval{kvseg/x40/768/dense-l25/pg19-val/896-1024/delta}{$-$0.00038}
\defval{kvseg/x40/768/dense-l25/pg19-val/896-1024/hi}{+0.00002}
\defval{kvseg/x40/768/dense-l25/pg19-val/896-1024/last}{1.2985}
\defval{kvseg/x40/768/dense-l25/pg19-val/896-1024/lo}{$-$0.00076}
\defval{kvseg/x40/768/dense-l25/pg19-val/896-1024/round}{1.2989}
\defval{kvseg/x40/768/dense-l25/pg19-val/all/delta}{$-$0.00052}
\defval{kvseg/x40/768/dense-l25/pg19-val/all/hi}{$-$0.00033}
\defval{kvseg/x40/768/dense-l25/pg19-val/all/last}{1.3503}
\defval{kvseg/x40/768/dense-l25/pg19-val/all/lo}{$-$0.00069}
\defval{kvseg/x40/768/dense-l25/pg19-val/all/round}{1.3509}
\defval{kvseg/x40/768/dense-l47/finepdfs-long/1-128/delta}{$-$0.00041}
\defval{kvseg/x40/768/dense-l47/finepdfs-long/1-128/hi}{$-$0.00010}
\defval{kvseg/x40/768/dense-l47/finepdfs-long/1-128/last}{1.4525}
\defval{kvseg/x40/768/dense-l47/finepdfs-long/1-128/lo}{$-$0.00074}
\defval{kvseg/x40/768/dense-l47/finepdfs-long/1-128/round}{1.4530}
\defval{kvseg/x40/768/dense-l47/finepdfs-long/128-256/delta}{$-$0.00032}
\defval{kvseg/x40/768/dense-l47/finepdfs-long/128-256/hi}{$-$0.00008}
\defval{kvseg/x40/768/dense-l47/finepdfs-long/128-256/last}{1.2264}
\defval{kvseg/x40/768/dense-l47/finepdfs-long/128-256/lo}{$-$0.00060}
\defval{kvseg/x40/768/dense-l47/finepdfs-long/128-256/round}{1.2267}
\defval{kvseg/x40/768/dense-l47/finepdfs-long/1920-2048/delta}{$-$0.00054}
\defval{kvseg/x40/768/dense-l47/finepdfs-long/1920-2048/hi}{$-$0.00023}
\defval{kvseg/x40/768/dense-l47/finepdfs-long/1920-2048/last}{1.0951}
\defval{kvseg/x40/768/dense-l47/finepdfs-long/1920-2048/lo}{$-$0.00085}
\defval{kvseg/x40/768/dense-l47/finepdfs-long/1920-2048/round}{1.0956}
\defval{kvseg/x40/768/dense-l47/finepdfs-long/384-512/delta}{$-$0.00060}
\defval{kvseg/x40/768/dense-l47/finepdfs-long/384-512/hi}{$-$0.00033}
\defval{kvseg/x40/768/dense-l47/finepdfs-long/384-512/last}{1.1660}
\defval{kvseg/x40/768/dense-l47/finepdfs-long/384-512/lo}{$-$0.00090}
\defval{kvseg/x40/768/dense-l47/finepdfs-long/384-512/round}{1.1666}
\defval{kvseg/x40/768/dense-l47/finepdfs-long/896-1024/delta}{$-$0.00054}
\defval{kvseg/x40/768/dense-l47/finepdfs-long/896-1024/hi}{$-$0.00021}
\defval{kvseg/x40/768/dense-l47/finepdfs-long/896-1024/last}{1.1497}
\defval{kvseg/x40/768/dense-l47/finepdfs-long/896-1024/lo}{$-$0.00086}
\defval{kvseg/x40/768/dense-l47/finepdfs-long/896-1024/round}{1.1503}
\defval{kvseg/x40/768/dense-l47/finepdfs-long/all/delta}{$-$0.00048}
\defval{kvseg/x40/768/dense-l47/finepdfs-long/all/hi}{$-$0.00036}
\defval{kvseg/x40/768/dense-l47/finepdfs-long/all/last}{1.2186}
\defval{kvseg/x40/768/dense-l47/finepdfs-long/all/lo}{$-$0.00061}
\defval{kvseg/x40/768/dense-l47/finepdfs-long/all/round}{1.2191}
\defval{kvseg/x40/768/dense-l47/fineweb-edu-100bt/1-128/delta}{$-$0.00012}
\defval{kvseg/x40/768/dense-l47/fineweb-edu-100bt/1-128/hi}{+0.00019}
\defval{kvseg/x40/768/dense-l47/fineweb-edu-100bt/1-128/last}{1.0076}
\defval{kvseg/x40/768/dense-l47/fineweb-edu-100bt/1-128/lo}{$-$0.00044}
\defval{kvseg/x40/768/dense-l47/fineweb-edu-100bt/1-128/round}{1.0077}
\defval{kvseg/x40/768/dense-l47/fineweb-edu-100bt/128-256/delta}{$-$0.00013}
\defval{kvseg/x40/768/dense-l47/fineweb-edu-100bt/128-256/hi}{+0.00011}
\defval{kvseg/x40/768/dense-l47/fineweb-edu-100bt/128-256/last}{0.9097}
\defval{kvseg/x40/768/dense-l47/fineweb-edu-100bt/128-256/lo}{$-$0.00039}
\defval{kvseg/x40/768/dense-l47/fineweb-edu-100bt/128-256/round}{0.9098}
\defval{kvseg/x40/768/dense-l47/fineweb-edu-100bt/1920-2048/delta}{$-$0.00002}
\defval{kvseg/x40/768/dense-l47/fineweb-edu-100bt/1920-2048/hi}{+0.00023}
\defval{kvseg/x40/768/dense-l47/fineweb-edu-100bt/1920-2048/last}{0.8868}
\defval{kvseg/x40/768/dense-l47/fineweb-edu-100bt/1920-2048/lo}{$-$0.00029}
\defval{kvseg/x40/768/dense-l47/fineweb-edu-100bt/1920-2048/round}{0.8868}
\defval{kvseg/x40/768/dense-l47/fineweb-edu-100bt/384-512/delta}{$-$0.00015}
\defval{kvseg/x40/768/dense-l47/fineweb-edu-100bt/384-512/hi}{+0.00009}
\defval{kvseg/x40/768/dense-l47/fineweb-edu-100bt/384-512/last}{0.8924}
\defval{kvseg/x40/768/dense-l47/fineweb-edu-100bt/384-512/lo}{$-$0.00036}
\defval{kvseg/x40/768/dense-l47/fineweb-edu-100bt/384-512/round}{0.8925}
\defval{kvseg/x40/768/dense-l47/fineweb-edu-100bt/896-1024/delta}{$-$0.00020}
\defval{kvseg/x40/768/dense-l47/fineweb-edu-100bt/896-1024/hi}{+0.00002}
\defval{kvseg/x40/768/dense-l47/fineweb-edu-100bt/896-1024/last}{0.8877}
\defval{kvseg/x40/768/dense-l47/fineweb-edu-100bt/896-1024/lo}{$-$0.00043}
\defval{kvseg/x40/768/dense-l47/fineweb-edu-100bt/896-1024/round}{0.8879}
\defval{kvseg/x40/768/dense-l47/fineweb-edu-100bt/all/delta}{$-$0.00012}
\defval{kvseg/x40/768/dense-l47/fineweb-edu-100bt/all/hi}{$-$0.00002}
\defval{kvseg/x40/768/dense-l47/fineweb-edu-100bt/all/last}{0.9167}
\defval{kvseg/x40/768/dense-l47/fineweb-edu-100bt/all/lo}{$-$0.00024}
\defval{kvseg/x40/768/dense-l47/fineweb-edu-100bt/all/round}{0.9168}
\defval{kvseg/x40/768/dense-l47/govreport-test/1-128/delta}{+0.00014}
\defval{kvseg/x40/768/dense-l47/govreport-test/1-128/hi}{+0.00034}
\defval{kvseg/x40/768/dense-l47/govreport-test/1-128/last}{0.9010}
\defval{kvseg/x40/768/dense-l47/govreport-test/1-128/lo}{$-$0.00009}
\defval{kvseg/x40/768/dense-l47/govreport-test/1-128/round}{0.9009}
\defval{kvseg/x40/768/dense-l47/govreport-test/128-256/delta}{$-$0.00006}
\defval{kvseg/x40/768/dense-l47/govreport-test/128-256/hi}{+0.00016}
\defval{kvseg/x40/768/dense-l47/govreport-test/128-256/last}{0.7766}
\defval{kvseg/x40/768/dense-l47/govreport-test/128-256/lo}{$-$0.00026}
\defval{kvseg/x40/768/dense-l47/govreport-test/128-256/round}{0.7766}
\defval{kvseg/x40/768/dense-l47/govreport-test/1920-2048/delta}{$-$0.00025}
\defval{kvseg/x40/768/dense-l47/govreport-test/1920-2048/hi}{$-$0.00005}
\defval{kvseg/x40/768/dense-l47/govreport-test/1920-2048/last}{0.7083}
\defval{kvseg/x40/768/dense-l47/govreport-test/1920-2048/lo}{$-$0.00045}
\defval{kvseg/x40/768/dense-l47/govreport-test/1920-2048/round}{0.7086}
\defval{kvseg/x40/768/dense-l47/govreport-test/384-512/delta}{$-$0.00026}
\defval{kvseg/x40/768/dense-l47/govreport-test/384-512/hi}{$-$0.00005}
\defval{kvseg/x40/768/dense-l47/govreport-test/384-512/last}{0.7369}
\defval{kvseg/x40/768/dense-l47/govreport-test/384-512/lo}{$-$0.00049}
\defval{kvseg/x40/768/dense-l47/govreport-test/384-512/round}{0.7371}
\defval{kvseg/x40/768/dense-l47/govreport-test/896-1024/delta}{$-$0.00021}
\defval{kvseg/x40/768/dense-l47/govreport-test/896-1024/hi}{+0.00001}
\defval{kvseg/x40/768/dense-l47/govreport-test/896-1024/last}{0.7176}
\defval{kvseg/x40/768/dense-l47/govreport-test/896-1024/lo}{$-$0.00042}
\defval{kvseg/x40/768/dense-l47/govreport-test/896-1024/round}{0.7178}
\defval{kvseg/x40/768/dense-l47/govreport-test/all/delta}{$-$0.00013}
\defval{kvseg/x40/768/dense-l47/govreport-test/all/hi}{$-$0.00003}
\defval{kvseg/x40/768/dense-l47/govreport-test/all/last}{0.7681}
\defval{kvseg/x40/768/dense-l47/govreport-test/all/lo}{$-$0.00022}
\defval{kvseg/x40/768/dense-l47/govreport-test/all/round}{0.7682}
\defval{kvseg/x40/768/dense-l47/longdocs/1-128/delta}{$-$0.00025}
\defval{kvseg/x40/768/dense-l47/longdocs/1-128/hi}{$-$0.00008}
\defval{kvseg/x40/768/dense-l47/longdocs/1-128/last}{1.2663}
\defval{kvseg/x40/768/dense-l47/longdocs/1-128/lo}{$-$0.00041}
\defval{kvseg/x40/768/dense-l47/longdocs/1-128/round}{1.2665}
\defval{kvseg/x40/768/dense-l47/longdocs/128-256/delta}{$-$0.00027}
\defval{kvseg/x40/768/dense-l47/longdocs/128-256/hi}{$-$0.00013}
\defval{kvseg/x40/768/dense-l47/longdocs/128-256/last}{1.1011}
\defval{kvseg/x40/768/dense-l47/longdocs/128-256/lo}{$-$0.00042}
\defval{kvseg/x40/768/dense-l47/longdocs/128-256/round}{1.1013}
\defval{kvseg/x40/768/dense-l47/longdocs/1920-2048/delta}{$-$0.00052}
\defval{kvseg/x40/768/dense-l47/longdocs/1920-2048/hi}{$-$0.00036}
\defval{kvseg/x40/768/dense-l47/longdocs/1920-2048/last}{1.0131}
\defval{kvseg/x40/768/dense-l47/longdocs/1920-2048/lo}{$-$0.00068}
\defval{kvseg/x40/768/dense-l47/longdocs/1920-2048/round}{1.0136}
\defval{kvseg/x40/768/dense-l47/longdocs/384-512/delta}{$-$0.00049}
\defval{kvseg/x40/768/dense-l47/longdocs/384-512/hi}{$-$0.00034}
\defval{kvseg/x40/768/dense-l47/longdocs/384-512/last}{1.0560}
\defval{kvseg/x40/768/dense-l47/longdocs/384-512/lo}{$-$0.00065}
\defval{kvseg/x40/768/dense-l47/longdocs/384-512/round}{1.0565}
\defval{kvseg/x40/768/dense-l47/longdocs/896-1024/delta}{$-$0.00037}
\defval{kvseg/x40/768/dense-l47/longdocs/896-1024/hi}{$-$0.00020}
\defval{kvseg/x40/768/dense-l47/longdocs/896-1024/last}{1.0409}
\defval{kvseg/x40/768/dense-l47/longdocs/896-1024/lo}{$-$0.00053}
\defval{kvseg/x40/768/dense-l47/longdocs/896-1024/round}{1.0412}
\defval{kvseg/x40/768/dense-l47/longdocs/all/delta}{$-$0.00038}
\defval{kvseg/x40/768/dense-l47/longdocs/all/hi}{$-$0.00031}
\defval{kvseg/x40/768/dense-l47/longdocs/all/last}{1.0955}
\defval{kvseg/x40/768/dense-l47/longdocs/all/lo}{$-$0.00045}
\defval{kvseg/x40/768/dense-l47/longdocs/all/round}{1.0958}
\defval{kvseg/x40/768/dense-l47/pg19-test/1-128/delta}{$-$0.00059}
\defval{kvseg/x40/768/dense-l47/pg19-test/1-128/hi}{$-$0.00025}
\defval{kvseg/x40/768/dense-l47/pg19-test/1-128/last}{1.5589}
\defval{kvseg/x40/768/dense-l47/pg19-test/1-128/lo}{$-$0.00098}
\defval{kvseg/x40/768/dense-l47/pg19-test/1-128/round}{1.5595}
\defval{kvseg/x40/768/dense-l47/pg19-test/128-256/delta}{$-$0.00049}
\defval{kvseg/x40/768/dense-l47/pg19-test/128-256/hi}{$-$0.00017}
\defval{kvseg/x40/768/dense-l47/pg19-test/128-256/last}{1.3958}
\defval{kvseg/x40/768/dense-l47/pg19-test/128-256/lo}{$-$0.00079}
\defval{kvseg/x40/768/dense-l47/pg19-test/128-256/round}{1.3963}
\defval{kvseg/x40/768/dense-l47/pg19-test/1920-2048/delta}{$-$0.00085}
\defval{kvseg/x40/768/dense-l47/pg19-test/1920-2048/hi}{$-$0.00048}
\defval{kvseg/x40/768/dense-l47/pg19-test/1920-2048/last}{1.3281}
\defval{kvseg/x40/768/dense-l47/pg19-test/1920-2048/lo}{$-$0.00121}
\defval{kvseg/x40/768/dense-l47/pg19-test/1920-2048/round}{1.3289}
\defval{kvseg/x40/768/dense-l47/pg19-test/384-512/delta}{$-$0.00069}
\defval{kvseg/x40/768/dense-l47/pg19-test/384-512/hi}{$-$0.00037}
\defval{kvseg/x40/768/dense-l47/pg19-test/384-512/last}{1.3603}
\defval{kvseg/x40/768/dense-l47/pg19-test/384-512/lo}{$-$0.00099}
\defval{kvseg/x40/768/dense-l47/pg19-test/384-512/round}{1.3610}
\defval{kvseg/x40/768/dense-l47/pg19-test/896-1024/delta}{$-$0.00041}
\defval{kvseg/x40/768/dense-l47/pg19-test/896-1024/hi}{$-$0.00004}
\defval{kvseg/x40/768/dense-l47/pg19-test/896-1024/last}{1.3533}
\defval{kvseg/x40/768/dense-l47/pg19-test/896-1024/lo}{$-$0.00078}
\defval{kvseg/x40/768/dense-l47/pg19-test/896-1024/round}{1.3537}
\defval{kvseg/x40/768/dense-l47/pg19-test/all/delta}{$-$0.00061}
\defval{kvseg/x40/768/dense-l47/pg19-test/all/hi}{$-$0.00045}
\defval{kvseg/x40/768/dense-l47/pg19-test/all/last}{1.3989}
\defval{kvseg/x40/768/dense-l47/pg19-test/all/lo}{$-$0.00076}
\defval{kvseg/x40/768/dense-l47/pg19-test/all/round}{1.3995}
\defval{kvseg/x40/768/dense-l47/pg19-val/1-128/delta}{$-$0.00074}
\defval{kvseg/x40/768/dense-l47/pg19-val/1-128/hi}{$-$0.00040}
\defval{kvseg/x40/768/dense-l47/pg19-val/1-128/last}{1.5030}
\defval{kvseg/x40/768/dense-l47/pg19-val/1-128/lo}{$-$0.00105}
\defval{kvseg/x40/768/dense-l47/pg19-val/1-128/round}{1.5037}
\defval{kvseg/x40/768/dense-l47/pg19-val/128-256/delta}{$-$0.00072}
\defval{kvseg/x40/768/dense-l47/pg19-val/128-256/hi}{$-$0.00042}
\defval{kvseg/x40/768/dense-l47/pg19-val/128-256/last}{1.3560}
\defval{kvseg/x40/768/dense-l47/pg19-val/128-256/lo}{$-$0.00101}
\defval{kvseg/x40/768/dense-l47/pg19-val/128-256/round}{1.3567}
\defval{kvseg/x40/768/dense-l47/pg19-val/1920-2048/delta}{$-$0.00067}
\defval{kvseg/x40/768/dense-l47/pg19-val/1920-2048/hi}{$-$0.00031}
\defval{kvseg/x40/768/dense-l47/pg19-val/1920-2048/last}{1.2879}
\defval{kvseg/x40/768/dense-l47/pg19-val/1920-2048/lo}{$-$0.00101}
\defval{kvseg/x40/768/dense-l47/pg19-val/1920-2048/round}{1.2886}
\defval{kvseg/x40/768/dense-l47/pg19-val/384-512/delta}{$-$0.00046}
\defval{kvseg/x40/768/dense-l47/pg19-val/384-512/hi}{$-$0.00014}
\defval{kvseg/x40/768/dense-l47/pg19-val/384-512/last}{1.3254}
\defval{kvseg/x40/768/dense-l47/pg19-val/384-512/lo}{$-$0.00075}
\defval{kvseg/x40/768/dense-l47/pg19-val/384-512/round}{1.3258}
\defval{kvseg/x40/768/dense-l47/pg19-val/896-1024/delta}{$-$0.00079}
\defval{kvseg/x40/768/dense-l47/pg19-val/896-1024/hi}{$-$0.00048}
\defval{kvseg/x40/768/dense-l47/pg19-val/896-1024/last}{1.3031}
\defval{kvseg/x40/768/dense-l47/pg19-val/896-1024/lo}{$-$0.00113}
\defval{kvseg/x40/768/dense-l47/pg19-val/896-1024/round}{1.3039}
\defval{kvseg/x40/768/dense-l47/pg19-val/all/delta}{$-$0.00068}
\defval{kvseg/x40/768/dense-l47/pg19-val/all/hi}{$-$0.00053}
\defval{kvseg/x40/768/dense-l47/pg19-val/all/last}{1.3550}
\defval{kvseg/x40/768/dense-l47/pg19-val/all/lo}{$-$0.00083}
\defval{kvseg/x40/768/dense-l47/pg19-val/all/round}{1.3556}
\defval{kvseg/x40/768/yocou-loop/finepdfs-long/1-128/delta}{$-$0.00057}
\defval{kvseg/x40/768/yocou-loop/finepdfs-long/1-128/hi}{$-$0.00023}
\defval{kvseg/x40/768/yocou-loop/finepdfs-long/1-128/last}{1.4532}
\defval{kvseg/x40/768/yocou-loop/finepdfs-long/1-128/lo}{$-$0.00093}
\defval{kvseg/x40/768/yocou-loop/finepdfs-long/1-128/round}{1.4538}
\defval{kvseg/x40/768/yocou-loop/finepdfs-long/128-256/delta}{$-$0.00053}
\defval{kvseg/x40/768/yocou-loop/finepdfs-long/128-256/hi}{$-$0.00025}
\defval{kvseg/x40/768/yocou-loop/finepdfs-long/128-256/last}{1.2228}
\defval{kvseg/x40/768/yocou-loop/finepdfs-long/128-256/lo}{$-$0.00083}
\defval{kvseg/x40/768/yocou-loop/finepdfs-long/128-256/round}{1.2233}
\defval{kvseg/x40/768/yocou-loop/finepdfs-long/1920-2048/delta}{$-$0.00071}
\defval{kvseg/x40/768/yocou-loop/finepdfs-long/1920-2048/hi}{$-$0.00040}
\defval{kvseg/x40/768/yocou-loop/finepdfs-long/1920-2048/last}{1.0974}
\defval{kvseg/x40/768/yocou-loop/finepdfs-long/1920-2048/lo}{$-$0.00106}
\defval{kvseg/x40/768/yocou-loop/finepdfs-long/1920-2048/round}{1.0981}
\defval{kvseg/x40/768/yocou-loop/finepdfs-long/384-512/delta}{$-$0.00046}
\defval{kvseg/x40/768/yocou-loop/finepdfs-long/384-512/hi}{$-$0.00014}
\defval{kvseg/x40/768/yocou-loop/finepdfs-long/384-512/last}{1.1671}
\defval{kvseg/x40/768/yocou-loop/finepdfs-long/384-512/lo}{$-$0.00082}
\defval{kvseg/x40/768/yocou-loop/finepdfs-long/384-512/round}{1.1676}
\defval{kvseg/x40/768/yocou-loop/finepdfs-long/896-1024/delta}{$-$0.00037}
\defval{kvseg/x40/768/yocou-loop/finepdfs-long/896-1024/hi}{$-$0.00005}
\defval{kvseg/x40/768/yocou-loop/finepdfs-long/896-1024/last}{1.1508}
\defval{kvseg/x40/768/yocou-loop/finepdfs-long/896-1024/lo}{$-$0.00070}
\defval{kvseg/x40/768/yocou-loop/finepdfs-long/896-1024/round}{1.1511}
\defval{kvseg/x40/768/yocou-loop/finepdfs-long/all/delta}{$-$0.00053}
\defval{kvseg/x40/768/yocou-loop/finepdfs-long/all/hi}{$-$0.00037}
\defval{kvseg/x40/768/yocou-loop/finepdfs-long/all/last}{1.2189}
\defval{kvseg/x40/768/yocou-loop/finepdfs-long/all/lo}{$-$0.00070}
\defval{kvseg/x40/768/yocou-loop/finepdfs-long/all/round}{1.2194}
\defval{kvseg/x40/768/yocou-loop/fineweb-edu-100bt/1-128/delta}{$-$0.00023}
\defval{kvseg/x40/768/yocou-loop/fineweb-edu-100bt/1-128/hi}{+0.00013}
\defval{kvseg/x40/768/yocou-loop/fineweb-edu-100bt/1-128/last}{1.0019}
\defval{kvseg/x40/768/yocou-loop/fineweb-edu-100bt/1-128/lo}{$-$0.00057}
\defval{kvseg/x40/768/yocou-loop/fineweb-edu-100bt/1-128/round}{1.0021}
\defval{kvseg/x40/768/yocou-loop/fineweb-edu-100bt/128-256/delta}{$-$0.00007}
\defval{kvseg/x40/768/yocou-loop/fineweb-edu-100bt/128-256/hi}{+0.00019}
\defval{kvseg/x40/768/yocou-loop/fineweb-edu-100bt/128-256/last}{0.9049}
\defval{kvseg/x40/768/yocou-loop/fineweb-edu-100bt/128-256/lo}{$-$0.00029}
\defval{kvseg/x40/768/yocou-loop/fineweb-edu-100bt/128-256/round}{0.9049}
\defval{kvseg/x40/768/yocou-loop/fineweb-edu-100bt/1920-2048/delta}{$-$0.00031}
\defval{kvseg/x40/768/yocou-loop/fineweb-edu-100bt/1920-2048/hi}{$-$0.00008}
\defval{kvseg/x40/768/yocou-loop/fineweb-edu-100bt/1920-2048/last}{0.8841}
\defval{kvseg/x40/768/yocou-loop/fineweb-edu-100bt/1920-2048/lo}{$-$0.00054}
\defval{kvseg/x40/768/yocou-loop/fineweb-edu-100bt/1920-2048/round}{0.8844}
\defval{kvseg/x40/768/yocou-loop/fineweb-edu-100bt/384-512/delta}{$-$0.00020}
\defval{kvseg/x40/768/yocou-loop/fineweb-edu-100bt/384-512/hi}{+0.00006}
\defval{kvseg/x40/768/yocou-loop/fineweb-edu-100bt/384-512/last}{0.8894}
\defval{kvseg/x40/768/yocou-loop/fineweb-edu-100bt/384-512/lo}{$-$0.00044}
\defval{kvseg/x40/768/yocou-loop/fineweb-edu-100bt/384-512/round}{0.8896}
\defval{kvseg/x40/768/yocou-loop/fineweb-edu-100bt/896-1024/delta}{$-$0.00016}
\defval{kvseg/x40/768/yocou-loop/fineweb-edu-100bt/896-1024/hi}{+0.00011}
\defval{kvseg/x40/768/yocou-loop/fineweb-edu-100bt/896-1024/last}{0.8861}
\defval{kvseg/x40/768/yocou-loop/fineweb-edu-100bt/896-1024/lo}{$-$0.00040}
\defval{kvseg/x40/768/yocou-loop/fineweb-edu-100bt/896-1024/round}{0.8862}
\defval{kvseg/x40/768/yocou-loop/fineweb-edu-100bt/all/delta}{$-$0.00019}
\defval{kvseg/x40/768/yocou-loop/fineweb-edu-100bt/all/hi}{$-$0.00008}
\defval{kvseg/x40/768/yocou-loop/fineweb-edu-100bt/all/last}{0.9131}
\defval{kvseg/x40/768/yocou-loop/fineweb-edu-100bt/all/lo}{$-$0.00031}
\defval{kvseg/x40/768/yocou-loop/fineweb-edu-100bt/all/round}{0.9133}
\defval{kvseg/x40/768/yocou-loop/govreport-test/1-128/delta}{$-$0.00001}
\defval{kvseg/x40/768/yocou-loop/govreport-test/1-128/hi}{+0.00024}
\defval{kvseg/x40/768/yocou-loop/govreport-test/1-128/last}{0.8948}
\defval{kvseg/x40/768/yocou-loop/govreport-test/1-128/lo}{$-$0.00024}
\defval{kvseg/x40/768/yocou-loop/govreport-test/1-128/round}{0.8948}
\defval{kvseg/x40/768/yocou-loop/govreport-test/128-256/delta}{+0.00003}
\defval{kvseg/x40/768/yocou-loop/govreport-test/128-256/hi}{+0.00024}
\defval{kvseg/x40/768/yocou-loop/govreport-test/128-256/last}{0.7719}
\defval{kvseg/x40/768/yocou-loop/govreport-test/128-256/lo}{$-$0.00018}
\defval{kvseg/x40/768/yocou-loop/govreport-test/128-256/round}{0.7719}
\defval{kvseg/x40/768/yocou-loop/govreport-test/1920-2048/delta}{$-$0.00020}
\defval{kvseg/x40/768/yocou-loop/govreport-test/1920-2048/hi}{+0.00002}
\defval{kvseg/x40/768/yocou-loop/govreport-test/1920-2048/last}{0.7129}
\defval{kvseg/x40/768/yocou-loop/govreport-test/1920-2048/lo}{$-$0.00045}
\defval{kvseg/x40/768/yocou-loop/govreport-test/1920-2048/round}{0.7131}
\defval{kvseg/x40/768/yocou-loop/govreport-test/384-512/delta}{$-$0.00023}
\defval{kvseg/x40/768/yocou-loop/govreport-test/384-512/hi}{$-$0.00003}
\defval{kvseg/x40/768/yocou-loop/govreport-test/384-512/last}{0.7368}
\defval{kvseg/x40/768/yocou-loop/govreport-test/384-512/lo}{$-$0.00046}
\defval{kvseg/x40/768/yocou-loop/govreport-test/384-512/round}{0.7370}
\defval{kvseg/x40/768/yocou-loop/govreport-test/896-1024/delta}{$-$0.00014}
\defval{kvseg/x40/768/yocou-loop/govreport-test/896-1024/hi}{+0.00006}
\defval{kvseg/x40/768/yocou-loop/govreport-test/896-1024/last}{0.7206}
\defval{kvseg/x40/768/yocou-loop/govreport-test/896-1024/lo}{$-$0.00035}
\defval{kvseg/x40/768/yocou-loop/govreport-test/896-1024/round}{0.7207}
\defval{kvseg/x40/768/yocou-loop/govreport-test/all/delta}{$-$0.00011}
\defval{kvseg/x40/768/yocou-loop/govreport-test/all/hi}{$-$0.00001}
\defval{kvseg/x40/768/yocou-loop/govreport-test/all/last}{0.7674}
\defval{kvseg/x40/768/yocou-loop/govreport-test/all/lo}{$-$0.00022}
\defval{kvseg/x40/768/yocou-loop/govreport-test/all/round}{0.7675}
\defval{kvseg/x40/768/yocou-loop/longdocs/1-128/delta}{$-$0.00037}
\defval{kvseg/x40/768/yocou-loop/longdocs/1-128/hi}{$-$0.00020}
\defval{kvseg/x40/768/yocou-loop/longdocs/1-128/last}{1.2600}
\defval{kvseg/x40/768/yocou-loop/longdocs/1-128/lo}{$-$0.00056}
\defval{kvseg/x40/768/yocou-loop/longdocs/1-128/round}{1.2604}
\defval{kvseg/x40/768/yocou-loop/longdocs/128-256/delta}{$-$0.00034}
\defval{kvseg/x40/768/yocou-loop/longdocs/128-256/hi}{$-$0.00017}
\defval{kvseg/x40/768/yocou-loop/longdocs/128-256/last}{1.0955}
\defval{kvseg/x40/768/yocou-loop/longdocs/128-256/lo}{$-$0.00051}
\defval{kvseg/x40/768/yocou-loop/longdocs/128-256/round}{1.0958}
\defval{kvseg/x40/768/yocou-loop/longdocs/1920-2048/delta}{$-$0.00054}
\defval{kvseg/x40/768/yocou-loop/longdocs/1920-2048/hi}{$-$0.00038}
\defval{kvseg/x40/768/yocou-loop/longdocs/1920-2048/last}{1.0162}
\defval{kvseg/x40/768/yocou-loop/longdocs/1920-2048/lo}{$-$0.00071}
\defval{kvseg/x40/768/yocou-loop/longdocs/1920-2048/round}{1.0168}
\defval{kvseg/x40/768/yocou-loop/longdocs/384-512/delta}{$-$0.00050}
\defval{kvseg/x40/768/yocou-loop/longdocs/384-512/hi}{$-$0.00031}
\defval{kvseg/x40/768/yocou-loop/longdocs/384-512/last}{1.0554}
\defval{kvseg/x40/768/yocou-loop/longdocs/384-512/lo}{$-$0.00070}
\defval{kvseg/x40/768/yocou-loop/longdocs/384-512/round}{1.0559}
\defval{kvseg/x40/768/yocou-loop/longdocs/896-1024/delta}{$-$0.00040}
\defval{kvseg/x40/768/yocou-loop/longdocs/896-1024/hi}{$-$0.00024}
\defval{kvseg/x40/768/yocou-loop/longdocs/896-1024/last}{1.0430}
\defval{kvseg/x40/768/yocou-loop/longdocs/896-1024/lo}{$-$0.00057}
\defval{kvseg/x40/768/yocou-loop/longdocs/896-1024/round}{1.0434}
\defval{kvseg/x40/768/yocou-loop/longdocs/all/delta}{$-$0.00043}
\defval{kvseg/x40/768/yocou-loop/longdocs/all/hi}{$-$0.00035}
\defval{kvseg/x40/768/yocou-loop/longdocs/all/last}{1.0940}
\defval{kvseg/x40/768/yocou-loop/longdocs/all/lo}{$-$0.00051}
\defval{kvseg/x40/768/yocou-loop/longdocs/all/round}{1.0945}
\defval{kvseg/x40/768/yocou-loop/pg19-test/1-128/delta}{$-$0.00063}
\defval{kvseg/x40/768/yocou-loop/pg19-test/1-128/hi}{$-$0.00024}
\defval{kvseg/x40/768/yocou-loop/pg19-test/1-128/last}{1.5458}
\defval{kvseg/x40/768/yocou-loop/pg19-test/1-128/lo}{$-$0.00104}
\defval{kvseg/x40/768/yocou-loop/pg19-test/1-128/round}{1.5464}
\defval{kvseg/x40/768/yocou-loop/pg19-test/128-256/delta}{$-$0.00061}
\defval{kvseg/x40/768/yocou-loop/pg19-test/128-256/hi}{$-$0.00022}
\defval{kvseg/x40/768/yocou-loop/pg19-test/128-256/last}{1.3871}
\defval{kvseg/x40/768/yocou-loop/pg19-test/128-256/lo}{$-$0.00103}
\defval{kvseg/x40/768/yocou-loop/pg19-test/128-256/round}{1.3877}
\defval{kvseg/x40/768/yocou-loop/pg19-test/1920-2048/delta}{$-$0.00080}
\defval{kvseg/x40/768/yocou-loop/pg19-test/1920-2048/hi}{$-$0.00048}
\defval{kvseg/x40/768/yocou-loop/pg19-test/1920-2048/last}{1.3301}
\defval{kvseg/x40/768/yocou-loop/pg19-test/1920-2048/lo}{$-$0.00116}
\defval{kvseg/x40/768/yocou-loop/pg19-test/1920-2048/round}{1.3309}
\defval{kvseg/x40/768/yocou-loop/pg19-test/384-512/delta}{$-$0.00089}
\defval{kvseg/x40/768/yocou-loop/pg19-test/384-512/hi}{$-$0.00045}
\defval{kvseg/x40/768/yocou-loop/pg19-test/384-512/last}{1.3575}
\defval{kvseg/x40/768/yocou-loop/pg19-test/384-512/lo}{$-$0.00131}
\defval{kvseg/x40/768/yocou-loop/pg19-test/384-512/round}{1.3584}
\defval{kvseg/x40/768/yocou-loop/pg19-test/896-1024/delta}{$-$0.00077}
\defval{kvseg/x40/768/yocou-loop/pg19-test/896-1024/hi}{$-$0.00044}
\defval{kvseg/x40/768/yocou-loop/pg19-test/896-1024/last}{1.3556}
\defval{kvseg/x40/768/yocou-loop/pg19-test/896-1024/lo}{$-$0.00112}
\defval{kvseg/x40/768/yocou-loop/pg19-test/896-1024/round}{1.3563}
\defval{kvseg/x40/768/yocou-loop/pg19-test/all/delta}{$-$0.00074}
\defval{kvseg/x40/768/yocou-loop/pg19-test/all/hi}{$-$0.00057}
\defval{kvseg/x40/768/yocou-loop/pg19-test/all/last}{1.3948}
\defval{kvseg/x40/768/yocou-loop/pg19-test/all/lo}{$-$0.00093}
\defval{kvseg/x40/768/yocou-loop/pg19-test/all/round}{1.3956}
\defval{kvseg/x40/768/yocou-loop/pg19-val/1-128/delta}{$-$0.00047}
\defval{kvseg/x40/768/yocou-loop/pg19-val/1-128/hi}{$-$0.00005}
\defval{kvseg/x40/768/yocou-loop/pg19-val/1-128/last}{1.4915}
\defval{kvseg/x40/768/yocou-loop/pg19-val/1-128/lo}{$-$0.00087}
\defval{kvseg/x40/768/yocou-loop/pg19-val/1-128/round}{1.4920}
\defval{kvseg/x40/768/yocou-loop/pg19-val/128-256/delta}{$-$0.00045}
\defval{kvseg/x40/768/yocou-loop/pg19-val/128-256/hi}{$-$0.00006}
\defval{kvseg/x40/768/yocou-loop/pg19-val/128-256/last}{1.3488}
\defval{kvseg/x40/768/yocou-loop/pg19-val/128-256/lo}{$-$0.00084}
\defval{kvseg/x40/768/yocou-loop/pg19-val/128-256/round}{1.3492}
\defval{kvseg/x40/768/yocou-loop/pg19-val/1920-2048/delta}{$-$0.00079}
\defval{kvseg/x40/768/yocou-loop/pg19-val/1920-2048/hi}{$-$0.00042}
\defval{kvseg/x40/768/yocou-loop/pg19-val/1920-2048/last}{1.2885}
\defval{kvseg/x40/768/yocou-loop/pg19-val/1920-2048/lo}{$-$0.00115}
\defval{kvseg/x40/768/yocou-loop/pg19-val/1920-2048/round}{1.2893}
\defval{kvseg/x40/768/yocou-loop/pg19-val/384-512/delta}{$-$0.00059}
\defval{kvseg/x40/768/yocou-loop/pg19-val/384-512/hi}{$-$0.00028}
\defval{kvseg/x40/768/yocou-loop/pg19-val/384-512/last}{1.3262}
\defval{kvseg/x40/768/yocou-loop/pg19-val/384-512/lo}{$-$0.00092}
\defval{kvseg/x40/768/yocou-loop/pg19-val/384-512/round}{1.3267}
\defval{kvseg/x40/768/yocou-loop/pg19-val/896-1024/delta}{$-$0.00041}
\defval{kvseg/x40/768/yocou-loop/pg19-val/896-1024/hi}{$-$0.00004}
\defval{kvseg/x40/768/yocou-loop/pg19-val/896-1024/last}{1.3058}
\defval{kvseg/x40/768/yocou-loop/pg19-val/896-1024/lo}{$-$0.00076}
\defval{kvseg/x40/768/yocou-loop/pg19-val/896-1024/round}{1.3063}
\defval{kvseg/x40/768/yocou-loop/pg19-val/all/delta}{$-$0.00054}
\defval{kvseg/x40/768/yocou-loop/pg19-val/all/hi}{$-$0.00036}
\defval{kvseg/x40/768/yocou-loop/pg19-val/all/last}{1.3521}
\defval{kvseg/x40/768/yocou-loop/pg19-val/all/lo}{$-$0.00069}
\defval{kvseg/x40/768/yocou-loop/pg19-val/all/round}{1.3526}
\defval{kvshare/long/1024/ced-dec/delta}{$-$0.0001}
\defval{kvshare/long/1024/ced-dec/last}{0.8296}
\defval{kvshare/long/1024/ced-dec/round}{0.8297}
\defval{kvshare/long/1024/ced-enc/delta}{$-$0.0001}
\defval{kvshare/long/1024/ced-enc/last}{0.8123}
\defval{kvshare/long/1024/ced-enc/round}{0.8124}
\defval{kvshare/long/1024/dense-l47/delta}{0.0000}
\defval{kvshare/long/1024/dense-l47/last}{0.8177}
\defval{kvshare/long/1024/dense-l47/round}{0.8177}
\defpending{kvshare/p140/768/p140-dec-fixed/delta}{p140b-d0}{unified eval (--parts kv): last-round KV vs per-round KV}
\defpending{kvshare/p140/768/p140-dec-fixed/last}{p140b-d0}{unified eval (--parts kv): last-round KV vs per-round KV}
\defpending{kvshare/p140/768/p140-dec-fixed/round}{p140b-d0}{unified eval (--parts kv): last-round KV vs per-round KV}
\defpending{kvshare/p140/768/p140-enc-fixed/delta}{p140b-e0}{unified eval (--parts kv): last-round KV vs per-round KV}
\defpending{kvshare/p140/768/p140-enc-fixed/last}{p140b-e0}{unified eval (--parts kv): last-round KV vs per-round KV}
\defpending{kvshare/p140/768/p140-enc-fixed/round}{p140b-e0}{unified eval (--parts kv): last-round KV vs per-round KV}
\defpending{kvshare/p140/768/p140-enc/delta}{p140b-e1}{unified eval (--parts kv): last-round KV vs per-round KV}
\defpending{kvshare/p140/768/p140-enc/last}{p140b-e1}{unified eval (--parts kv): last-round KV vs per-round KV}
\defpending{kvshare/p140/768/p140-enc/round}{p140b-e1}{unified eval (--parts kv): last-round KV vs per-round KV}
\defval{kvshare/x40/1024/ced-dec/delta}{0.0000}
\defval{kvshare/x40/1024/ced-dec/last}{0.8648}
\defval{kvshare/x40/1024/ced-dec/round}{0.8648}
\defval{kvshare/x40/1024/ced-enc-fixed/delta}{+0.0317}
\defval{kvshare/x40/1024/ced-enc-fixed/last}{0.8801}
\defval{kvshare/x40/1024/ced-enc-fixed/round}{0.8484}
\defval{kvshare/x40/1024/ced-enc/delta}{$-$0.0002}
\defval{kvshare/x40/1024/ced-enc/last}{0.8463}
\defval{kvshare/x40/1024/ced-enc/round}{0.8465}
\defval{kvshare/x40/1024/dense-l25/delta}{+0.0003}
\defval{kvshare/x40/1024/dense-l25/last}{0.8460}
\defval{kvshare/x40/1024/dense-l25/round}{0.8457}
\defval{kvshare/x40/1024/dense-l47/delta}{0.0000}
\defval{kvshare/x40/1024/dense-l47/last}{0.8510}
\defval{kvshare/x40/1024/dense-l47/round}{0.8510}
\defval{kvshare/x40/1024/yocou-loop/delta}{$-$0.0001}
\defval{kvshare/x40/1024/yocou-loop/last}{0.8502}
\defval{kvshare/x40/1024/yocou-loop/round}{0.8503}
\defval{kvshare/x40/1536/ced-dec/delta}{0.0000}
\defval{kvshare/x40/1536/ced-dec/last}{0.8138}
\defval{kvshare/x40/1536/ced-dec/round}{0.8138}
\defval{kvshare/x40/1536/ced-enc/delta}{0.0000}
\defval{kvshare/x40/1536/ced-enc/last}{0.7965}
\defval{kvshare/x40/1536/ced-enc/round}{0.7965}
\defval{kvshare/x40/1536/dense-l25/delta}{+0.0001}
\defval{kvshare/x40/1536/dense-l25/last}{0.7963}
\defval{kvshare/x40/1536/dense-l25/round}{0.7962}
\defval{kvshare/x40/768/ced-dec/delta}{0.0000}
\defval{kvshare/x40/768/ced-dec/last}{0.9108}
\defval{kvshare/x40/768/ced-dec/round}{0.9108}
\defval{kvshare/x40/768/ced-enc/delta}{$-$0.0002}
\defval{kvshare/x40/768/ced-enc/last}{0.8919}
\defval{kvshare/x40/768/ced-enc/round}{0.8921}
\defval{kvshare/x40/768/dense-l25/delta}{$-$0.0001}
\defval{kvshare/x40/768/dense-l25/last}{0.8917}
\defval{kvshare/x40/768/dense-l25/round}{0.8918}
\defval{kvshare/x40/768/dense-l47/delta}{$-$0.0001}
\defval{kvshare/x40/768/dense-l47/last}{0.8975}
\defval{kvshare/x40/768/dense-l47/round}{0.8976}
\defval{kvshare/x40/768/yocou-loop/delta}{$-$0.0002}
\defval{kvshare/x40/768/yocou-loop/last}{0.8951}
\defval{kvshare/x40/768/yocou-loop/round}{0.8953}
\defval{kvshare64/long/1024/ced-enc/delta}{+0.0000}
\defval{kvshare64/long/1024/ced-enc/last}{0.8213}
\defval{kvshare64/long/1024/ced-enc/round}{0.8213}
\defval{kvshare64/long/1024/ced-enc/windows}{64}
\defval{kvshare64/p140/768/p140-enc-fixed/delta}{+0.0412}
\defval{kvshare64/p140/768/p140-enc-fixed/last}{0.9331}
\defval{kvshare64/p140/768/p140-enc-fixed/round}{0.8919}
\defval{kvshare64/p140/768/p140-enc-fixed/windows}{64}
\defval{kvshare64/p140/768/p140-enc/delta}{$-$0.0002}
\defval{kvshare64/p140/768/p140-enc/last}{0.8907}
\defval{kvshare64/p140/768/p140-enc/round}{0.8909}
\defval{kvshare64/p140/768/p140-enc/windows}{64}
\defval{lit/agentic/crossturn}{55}
\defval{lit/agentic/turn}{90}
\defval{lit/benice/fail-min}{770}
\defval{lit/benice/ok-max}{370}
\defval{lit/mooncake/max}{55}
\defval{lit/parcae/core/parcae}{20.00}
\defval{lit/parcae/core/transformer}{17.46}
\defval{lit/parcae/coreext/parcae}{12.75}
\defval{lit/parcae/coreext/transformer}{11.71}
\defval{lit/parcae/tokens}{29.6}
\defval{lit/prefixcache/a}{62}
\defval{lit/prefixcache/b}{54}
\defval{lit/v41/active-decode}{16}
\defval{lit/v41/active-prefill}{8}
\defval{lit/v41/hidden}{5120}
\defval{lit/v41/kv}{890}
\defval{lit/v41/layers}{40}
\defval{lit/v41/window}{128}
\defval{long/d1-last-eval-tokens}{29.6}
\defval{params/long/1024/ced-dec}{157.3}
\defval{params/long/1024/ced-enc}{157.3}
\defval{params/long/1024/ced-noloop}{155.2}
\defval{params/long/1024/dense-l47}{153.1}
\defval{params/long/1024/dense-noloop}{151.0}
\defval{params/p140/768/p140-dec-fixed}{44.8}
\defval{params/p140/768/p140-enc}{44.8}
\defval{params/p140/768/p140-enc-fixed}{44.8}
\defval{params/p140/768/p140-noloop}{43.7}
\defval{params/x40/1024/ced-dec}{157.3}
\defval{params/x40/1024/ced-enc}{157.3}
\defval{params/x40/1024/ced-enc-fixed}{157.3}
\defval{params/x40/1024/ced-noloop}{155.2}
\defval{params/x40/1024/dense-l25}{153.1}
\defval{params/x40/1024/dense-l47}{153.1}
\defval{params/x40/1024/dense-noloop}{151.0}
\defval{params/x40/1024/injection}{2.1}
\defval{params/x40/1024/yocou-loop}{155.2}
\defval{params/x40/1024/yocou-noloop}{153.1}
\defval{params/x40/1536/ced-dec}{353.9}
\defval{params/x40/1536/ced-enc}{353.9}
\defval{params/x40/1536/ced-noloop}{349.2}
\defval{params/x40/1536/dense-l25}{344.5}
\defval{params/x40/1536/dense-noloop}{339.8}
\defval{params/x40/768/ced-dec}{88.5}
\defval{params/x40/768/ced-enc}{88.5}
\defval{params/x40/768/ced-noloop}{87.3}
\defval{params/x40/768/dense-l25}{86.1}
\defval{params/x40/768/dense-l47}{86.1}
\defval{params/x40/768/dense-noloop}{85.0}
\defval{params/x40/768/yocou-loop}{87.3}
\defval{params/x40/768/yocou-noloop}{86.1}
\defval{passes/x40/1024/ced-dec/decode}{40}
\defval{passes/x40/1024/ced-dec/prefill}{6}
\defval{passes/x40/1024/ced-enc-fixed/decode}{40}
\defval{passes/x40/1024/ced-enc-fixed/prefill}{34}
\defval{passes/x40/1024/ced-enc/decode}{40}
\defval{passes/x40/1024/ced-enc/prefill}{34}
\defval{passes/x40/1024/ced-noloop/decode}{12}
\defval{passes/x40/1024/ced-noloop/prefill}{6}
\defval{passes/x40/1024/dense-l25/decode}{40}
\defval{passes/x40/1024/dense-l25/prefill}{40}
\defval{passes/x40/1024/dense-l47/decode}{40}
\defval{passes/x40/1024/dense-l47/prefill}{40}
\defval{passes/x40/1024/dense-noloop/decode}{12}
\defval{passes/x40/1024/dense-noloop/prefill}{12}
\defval{passes/x40/1024/yocou-loop/decode}{40}
\defval{passes/x40/1024/yocou-loop/prefill}{34}
\defval{passes/x40/1024/yocou-noloop/decode}{12}
\defval{passes/x40/1024/yocou-noloop/prefill}{6}
\defval{pos/long/1024/ced-dec/finepdfs-long/c1-128}{1.3704}
\defval{pos/long/1024/ced-dec/finepdfs-long/c1024-2048}{1.0201}
\defval{pos/long/1024/ced-dec/finepdfs-long/c128-256}{1.1291}
\defval{pos/long/1024/ced-dec/finepdfs-long/c256-512}{1.0825}
\defval{pos/long/1024/ced-dec/finepdfs-long/c512-1024}{1.0423}
\defval{pos/long/1024/ced-dec/finepdfs-long/last1024}{1.0201}
\defval{pos/long/1024/ced-dec/finepdfs-long/last128}{1.0140}
\defval{pos/long/1024/ced-dec/finepdfs-long/last256}{1.0121}
\defval{pos/long/1024/ced-dec/finepdfs-long/last512}{1.0137}
\defval{pos/long/1024/ced-dec/finepdfs-long/mean}{1.0624}
\defval{pos/long/1024/ced-dec/fineweb-edu-100bt/c1-128}{0.8737}
\defval{pos/long/1024/ced-dec/fineweb-edu-100bt/c1024-2048}{0.8366}
\defval{pos/long/1024/ced-dec/fineweb-edu-100bt/c128-256}{0.8178}
\defval{pos/long/1024/ced-dec/fineweb-edu-100bt/c256-512}{0.8128}
\defval{pos/long/1024/ced-dec/fineweb-edu-100bt/c512-1024}{0.8163}
\defval{pos/long/1024/ced-dec/fineweb-edu-100bt/last1024}{0.8222}
\defval{pos/long/1024/ced-dec/fineweb-edu-100bt/last128}{0.8197}
\defval{pos/long/1024/ced-dec/fineweb-edu-100bt/last256}{0.8207}
\defval{pos/long/1024/ced-dec/fineweb-edu-100bt/last512}{0.8206}
\defval{pos/long/1024/ced-dec/fineweb-edu-100bt/mean}{0.8297}
\defval{pos/long/1024/ced-dec/govreport-test/c1-128}{0.8499}
\defval{pos/long/1024/ced-dec/govreport-test/c1024-2048}{0.6512}
\defval{pos/long/1024/ced-dec/govreport-test/c128-256}{0.7173}
\defval{pos/long/1024/ced-dec/govreport-test/c256-512}{0.6877}
\defval{pos/long/1024/ced-dec/govreport-test/c512-1024}{0.6664}
\defval{pos/long/1024/ced-dec/govreport-test/last1024}{0.6512}
\defval{pos/long/1024/ced-dec/govreport-test/last128}{0.6456}
\defval{pos/long/1024/ced-dec/govreport-test/last256}{0.6459}
\defval{pos/long/1024/ced-dec/govreport-test/last512}{0.6482}
\defval{pos/long/1024/ced-dec/govreport-test/mean}{0.6760}
\defval{pos/long/1024/ced-dec/longdocs/c1-128}{1.2256}
\defval{pos/long/1024/ced-dec/longdocs/c1024-2048}{0.9662}
\defval{pos/long/1024/ced-dec/longdocs/c128-256}{1.0476}
\defval{pos/long/1024/ced-dec/longdocs/c256-512}{1.0102}
\defval{pos/long/1024/ced-dec/longdocs/c512-1024}{0.9825}
\defval{pos/long/1024/ced-dec/longdocs/last1024}{0.9662}
\defval{pos/long/1024/ced-dec/longdocs/last128}{0.9618}
\defval{pos/long/1024/ced-dec/longdocs/last256}{0.9611}
\defval{pos/long/1024/ced-dec/longdocs/last512}{0.9628}
\defval{pos/long/1024/ced-dec/longdocs/mean}{0.9971}
\defval{pos/long/1024/ced-dec/pg19-test/c1-128}{1.4581}
\defval{pos/long/1024/ced-dec/pg19-test/c1024-2048}{1.2299}
\defval{pos/long/1024/ced-dec/pg19-test/c128-256}{1.2975}
\defval{pos/long/1024/ced-dec/pg19-test/c256-512}{1.2616}
\defval{pos/long/1024/ced-dec/pg19-test/c512-1024}{1.2393}
\defval{pos/long/1024/ced-dec/pg19-test/last1024}{1.2299}
\defval{pos/long/1024/ced-dec/pg19-test/last128}{1.2298}
\defval{pos/long/1024/ced-dec/pg19-test/last256}{1.2288}
\defval{pos/long/1024/ced-dec/pg19-test/last512}{1.2293}
\defval{pos/long/1024/ced-dec/pg19-test/mean}{1.2546}
\defval{pos/long/1024/ced-dec/pg19-val/c1-128}{1.4180}
\defval{pos/long/1024/ced-dec/pg19-val/c1024-2048}{1.1949}
\defval{pos/long/1024/ced-dec/pg19-val/c128-256}{1.2562}
\defval{pos/long/1024/ced-dec/pg19-val/c256-512}{1.2277}
\defval{pos/long/1024/ced-dec/pg19-val/c512-1024}{1.2030}
\defval{pos/long/1024/ced-dec/pg19-val/last1024}{1.1949}
\defval{pos/long/1024/ced-dec/pg19-val/last128}{1.1869}
\defval{pos/long/1024/ced-dec/pg19-val/last256}{1.1857}
\defval{pos/long/1024/ced-dec/pg19-val/last512}{1.1887}
\defval{pos/long/1024/ced-dec/pg19-val/mean}{1.2187}
\defval{pos/long/1024/ced-enc/finepdfs-long/c1-128}{1.3563}
\defval{pos/long/1024/ced-enc/finepdfs-long/c1024-2048}{0.9877}
\defval{pos/long/1024/ced-enc/finepdfs-long/c128-256}{1.1009}
\defval{pos/long/1024/ced-enc/finepdfs-long/c256-512}{1.0518}
\defval{pos/long/1024/ced-enc/finepdfs-long/c512-1024}{1.0102}
\defval{pos/long/1024/ced-enc/finepdfs-long/last1024}{0.9877}
\defval{pos/long/1024/ced-enc/finepdfs-long/last128}{0.9802}
\defval{pos/long/1024/ced-enc/finepdfs-long/last256}{0.9790}
\defval{pos/long/1024/ced-enc/finepdfs-long/last512}{0.9810}
\defval{pos/long/1024/ced-enc/finepdfs-long/mean}{1.0318}
\defval{pos/long/1024/ced-enc/fineweb-edu-100bt/c1-128}{0.8598}
\defval{pos/long/1024/ced-enc/fineweb-edu-100bt/c1024-2048}{0.8169}
\defval{pos/long/1024/ced-enc/fineweb-edu-100bt/c128-256}{0.8015}
\defval{pos/long/1024/ced-enc/fineweb-edu-100bt/c256-512}{0.7956}
\defval{pos/long/1024/ced-enc/fineweb-edu-100bt/c512-1024}{0.7982}
\defval{pos/long/1024/ced-enc/fineweb-edu-100bt/last1024}{0.8044}
\defval{pos/long/1024/ced-enc/fineweb-edu-100bt/last128}{0.8009}
\defval{pos/long/1024/ced-enc/fineweb-edu-100bt/last256}{0.8026}
\defval{pos/long/1024/ced-enc/fineweb-edu-100bt/last512}{0.8027}
\defval{pos/long/1024/ced-enc/fineweb-edu-100bt/mean}{0.8124}
\defval{pos/long/1024/ced-enc/govreport-test/c1-128}{0.8380}
\defval{pos/long/1024/ced-enc/govreport-test/c1024-2048}{0.6316}
\defval{pos/long/1024/ced-enc/govreport-test/c128-256}{0.7020}
\defval{pos/long/1024/ced-enc/govreport-test/c256-512}{0.6704}
\defval{pos/long/1024/ced-enc/govreport-test/c512-1024}{0.6483}
\defval{pos/long/1024/ced-enc/govreport-test/last1024}{0.6316}
\defval{pos/long/1024/ced-enc/govreport-test/last128}{0.6253}
\defval{pos/long/1024/ced-enc/govreport-test/last256}{0.6258}
\defval{pos/long/1024/ced-enc/govreport-test/last512}{0.6283}
\defval{pos/long/1024/ced-enc/govreport-test/mean}{0.6579}
\defval{pos/long/1024/ced-enc/longdocs/c1-128}{1.2140}
\defval{pos/long/1024/ced-enc/longdocs/c1024-2048}{0.9398}
\defval{pos/long/1024/ced-enc/longdocs/c128-256}{1.0248}
\defval{pos/long/1024/ced-enc/longdocs/c256-512}{0.9848}
\defval{pos/long/1024/ced-enc/longdocs/c512-1024}{0.9563}
\defval{pos/long/1024/ced-enc/longdocs/last1024}{0.9398}
\defval{pos/long/1024/ced-enc/longdocs/last128}{0.9348}
\defval{pos/long/1024/ced-enc/longdocs/last256}{0.9343}
\defval{pos/long/1024/ced-enc/longdocs/last512}{0.9361}
\defval{pos/long/1024/ced-enc/longdocs/mean}{0.9721}
\defval{pos/long/1024/ced-enc/pg19-test/c1-128}{1.4494}
\defval{pos/long/1024/ced-enc/pg19-test/c1024-2048}{1.2024}
\defval{pos/long/1024/ced-enc/pg19-test/c128-256}{1.2723}
\defval{pos/long/1024/ced-enc/pg19-test/c256-512}{1.2332}
\defval{pos/long/1024/ced-enc/pg19-test/c512-1024}{1.2109}
\defval{pos/long/1024/ced-enc/pg19-test/last1024}{1.2024}
\defval{pos/long/1024/ced-enc/pg19-test/last128}{1.2026}
\defval{pos/long/1024/ced-enc/pg19-test/last256}{1.2014}
\defval{pos/long/1024/ced-enc/pg19-test/last512}{1.2018}
\defval{pos/long/1024/ced-enc/pg19-test/mean}{1.2281}
\defval{pos/long/1024/ced-enc/pg19-val/c1-128}{1.4102}
\defval{pos/long/1024/ced-enc/pg19-val/c1024-2048}{1.1686}
\defval{pos/long/1024/ced-enc/pg19-val/c128-256}{1.2306}
\defval{pos/long/1024/ced-enc/pg19-val/c256-512}{1.2009}
\defval{pos/long/1024/ced-enc/pg19-val/c512-1024}{1.1756}
\defval{pos/long/1024/ced-enc/pg19-val/last1024}{1.1686}
\defval{pos/long/1024/ced-enc/pg19-val/last128}{1.1592}
\defval{pos/long/1024/ced-enc/pg19-val/last256}{1.1587}
\defval{pos/long/1024/ced-enc/pg19-val/last512}{1.1627}
\defval{pos/long/1024/ced-enc/pg19-val/mean}{1.1933}
\defval{pos/long/1024/ced-noloop/finepdfs-long/c1-128}{1.3716}
\defval{pos/long/1024/ced-noloop/finepdfs-long/c1024-2048}{1.0330}
\defval{pos/long/1024/ced-noloop/finepdfs-long/c128-256}{1.1378}
\defval{pos/long/1024/ced-noloop/finepdfs-long/c256-512}{1.0927}
\defval{pos/long/1024/ced-noloop/finepdfs-long/c512-1024}{1.0539}
\defval{pos/long/1024/ced-noloop/finepdfs-long/last1024}{1.0330}
\defval{pos/long/1024/ced-noloop/finepdfs-long/last128}{1.0263}
\defval{pos/long/1024/ced-noloop/finepdfs-long/last256}{1.0252}
\defval{pos/long/1024/ced-noloop/finepdfs-long/last512}{1.0269}
\defval{pos/long/1024/ced-noloop/finepdfs-long/mean}{1.0737}
\defval{pos/long/1024/ced-noloop/fineweb-edu-100bt/c1-128}{0.8842}
\defval{pos/long/1024/ced-noloop/fineweb-edu-100bt/c1024-2048}{0.8481}
\defval{pos/long/1024/ced-noloop/fineweb-edu-100bt/c128-256}{0.8284}
\defval{pos/long/1024/ced-noloop/fineweb-edu-100bt/c256-512}{0.8236}
\defval{pos/long/1024/ced-noloop/fineweb-edu-100bt/c512-1024}{0.8270}
\defval{pos/long/1024/ced-noloop/fineweb-edu-100bt/last1024}{0.8331}
\defval{pos/long/1024/ced-noloop/fineweb-edu-100bt/last128}{0.8297}
\defval{pos/long/1024/ced-noloop/fineweb-edu-100bt/last256}{0.8310}
\defval{pos/long/1024/ced-noloop/fineweb-edu-100bt/last512}{0.8314}
\defval{pos/long/1024/ced-noloop/fineweb-edu-100bt/mean}{0.8405}
\defval{pos/long/1024/ced-noloop/govreport-test/c1-128}{0.8571}
\defval{pos/long/1024/ced-noloop/govreport-test/c1024-2048}{0.6597}
\defval{pos/long/1024/ced-noloop/govreport-test/c128-256}{0.7251}
\defval{pos/long/1024/ced-noloop/govreport-test/c256-512}{0.6957}
\defval{pos/long/1024/ced-noloop/govreport-test/c512-1024}{0.6751}
\defval{pos/long/1024/ced-noloop/govreport-test/last1024}{0.6597}
\defval{pos/long/1024/ced-noloop/govreport-test/last128}{0.6542}
\defval{pos/long/1024/ced-noloop/govreport-test/last256}{0.6542}
\defval{pos/long/1024/ced-noloop/govreport-test/last512}{0.6566}
\defval{pos/long/1024/ced-noloop/govreport-test/mean}{0.6844}
\defval{pos/long/1024/ced-noloop/longdocs/c1-128}{1.2319}
\defval{pos/long/1024/ced-noloop/longdocs/c1024-2048}{0.9797}
\defval{pos/long/1024/ced-noloop/longdocs/c128-256}{1.0579}
\defval{pos/long/1024/ced-noloop/longdocs/c256-512}{1.0219}
\defval{pos/long/1024/ced-noloop/longdocs/c512-1024}{0.9955}
\defval{pos/long/1024/ced-noloop/longdocs/last1024}{0.9797}
\defval{pos/long/1024/ced-noloop/longdocs/last128}{0.9750}
\defval{pos/long/1024/ced-noloop/longdocs/last256}{0.9745}
\defval{pos/long/1024/ced-noloop/longdocs/last512}{0.9762}
\defval{pos/long/1024/ced-noloop/longdocs/mean}{1.0096}
\defval{pos/long/1024/ced-noloop/pg19-test/c1-128}{1.4686}
\defval{pos/long/1024/ced-noloop/pg19-test/c1024-2048}{1.2488}
\defval{pos/long/1024/ced-noloop/pg19-test/c128-256}{1.3119}
\defval{pos/long/1024/ced-noloop/pg19-test/c256-512}{1.2784}
\defval{pos/long/1024/ced-noloop/pg19-test/c512-1024}{1.2578}
\defval{pos/long/1024/ced-noloop/pg19-test/last1024}{1.2488}
\defval{pos/long/1024/ced-noloop/pg19-test/last128}{1.2483}
\defval{pos/long/1024/ced-noloop/pg19-test/last256}{1.2477}
\defval{pos/long/1024/ced-noloop/pg19-test/last512}{1.2481}
\defval{pos/long/1024/ced-noloop/pg19-test/mean}{1.2724}
\defval{pos/long/1024/ced-noloop/pg19-val/c1-128}{1.4266}
\defval{pos/long/1024/ced-noloop/pg19-val/c1024-2048}{1.2127}
\defval{pos/long/1024/ced-noloop/pg19-val/c128-256}{1.2698}
\defval{pos/long/1024/ced-noloop/pg19-val/c256-512}{1.2432}
\defval{pos/long/1024/ced-noloop/pg19-val/c512-1024}{1.2201}
\defval{pos/long/1024/ced-noloop/pg19-val/last1024}{1.2127}
\defval{pos/long/1024/ced-noloop/pg19-val/last128}{1.2060}
\defval{pos/long/1024/ced-noloop/pg19-val/last256}{1.2040}
\defval{pos/long/1024/ced-noloop/pg19-val/last512}{1.2073}
\defval{pos/long/1024/ced-noloop/pg19-val/mean}{1.2352}
\defval{pos/long/1024/dense-l47/finepdfs-long/c1-128}{1.3676}
\defval{pos/long/1024/dense-l47/finepdfs-long/c1024-2048}{1.0004}
\defval{pos/long/1024/dense-l47/finepdfs-long/c128-256}{1.1160}
\defval{pos/long/1024/dense-l47/finepdfs-long/c256-512}{1.0665}
\defval{pos/long/1024/dense-l47/finepdfs-long/c512-1024}{1.0237}
\defval{pos/long/1024/dense-l47/finepdfs-long/last1024}{1.0004}
\defval{pos/long/1024/dense-l47/finepdfs-long/last128}{0.9931}
\defval{pos/long/1024/dense-l47/finepdfs-long/last256}{0.9918}
\defval{pos/long/1024/dense-l47/finepdfs-long/last512}{0.9938}
\defval{pos/long/1024/dense-l47/finepdfs-long/mean}{1.0450}
\defval{pos/long/1024/dense-l47/fineweb-edu-100bt/c1-128}{0.8658}
\defval{pos/long/1024/dense-l47/fineweb-edu-100bt/c1024-2048}{0.8215}
\defval{pos/long/1024/dense-l47/fineweb-edu-100bt/c128-256}{0.8068}
\defval{pos/long/1024/dense-l47/fineweb-edu-100bt/c256-512}{0.8008}
\defval{pos/long/1024/dense-l47/fineweb-edu-100bt/c512-1024}{0.8038}
\defval{pos/long/1024/dense-l47/fineweb-edu-100bt/last1024}{0.8092}
\defval{pos/long/1024/dense-l47/fineweb-edu-100bt/last128}{0.8061}
\defval{pos/long/1024/dense-l47/fineweb-edu-100bt/last256}{0.8077}
\defval{pos/long/1024/dense-l47/fineweb-edu-100bt/last512}{0.8074}
\defval{pos/long/1024/dense-l47/fineweb-edu-100bt/mean}{0.8178}
\defval{pos/long/1024/dense-l47/govreport-test/c1-128}{0.8427}
\defval{pos/long/1024/dense-l47/govreport-test/c1024-2048}{0.6342}
\defval{pos/long/1024/dense-l47/govreport-test/c128-256}{0.7067}
\defval{pos/long/1024/dense-l47/govreport-test/c256-512}{0.6743}
\defval{pos/long/1024/dense-l47/govreport-test/c512-1024}{0.6513}
\defval{pos/long/1024/dense-l47/govreport-test/last1024}{0.6342}
\defval{pos/long/1024/dense-l47/govreport-test/last128}{0.6285}
\defval{pos/long/1024/dense-l47/govreport-test/last256}{0.6287}
\defval{pos/long/1024/dense-l47/govreport-test/last512}{0.6310}
\defval{pos/long/1024/dense-l47/govreport-test/mean}{0.6610}
\defval{pos/long/1024/dense-l47/longdocs/c1-128}{1.2229}
\defval{pos/long/1024/dense-l47/longdocs/c1024-2048}{0.9480}
\defval{pos/long/1024/dense-l47/longdocs/c128-256}{1.0364}
\defval{pos/long/1024/dense-l47/longdocs/c256-512}{0.9954}
\defval{pos/long/1024/dense-l47/longdocs/c512-1024}{0.9655}
\defval{pos/long/1024/dense-l47/longdocs/last1024}{0.9480}
\defval{pos/long/1024/dense-l47/longdocs/last128}{0.9429}
\defval{pos/long/1024/dense-l47/longdocs/last256}{0.9425}
\defval{pos/long/1024/dense-l47/longdocs/last512}{0.9443}
\defval{pos/long/1024/dense-l47/longdocs/mean}{0.9810}
\defval{pos/long/1024/dense-l47/pg19-test/c1-128}{1.4600}
\defval{pos/long/1024/dense-l47/pg19-test/c1024-2048}{1.2116}
\defval{pos/long/1024/dense-l47/pg19-test/c128-256}{1.2874}
\defval{pos/long/1024/dense-l47/pg19-test/c256-512}{1.2465}
\defval{pos/long/1024/dense-l47/pg19-test/c512-1024}{1.2221}
\defval{pos/long/1024/dense-l47/pg19-test/last1024}{1.2116}
\defval{pos/long/1024/dense-l47/pg19-test/last128}{1.2111}
\defval{pos/long/1024/dense-l47/pg19-test/last256}{1.2104}
\defval{pos/long/1024/dense-l47/pg19-test/last512}{1.2110}
\defval{pos/long/1024/dense-l47/pg19-test/mean}{1.2388}
\defval{pos/long/1024/dense-l47/pg19-val/c1-128}{1.4214}
\defval{pos/long/1024/dense-l47/pg19-val/c1024-2048}{1.1773}
\defval{pos/long/1024/dense-l47/pg19-val/c128-256}{1.2477}
\defval{pos/long/1024/dense-l47/pg19-val/c256-512}{1.2147}
\defval{pos/long/1024/dense-l47/pg19-val/c512-1024}{1.1863}
\defval{pos/long/1024/dense-l47/pg19-val/last1024}{1.1773}
\defval{pos/long/1024/dense-l47/pg19-val/last128}{1.1688}
\defval{pos/long/1024/dense-l47/pg19-val/last256}{1.1667}
\defval{pos/long/1024/dense-l47/pg19-val/last512}{1.1710}
\defval{pos/long/1024/dense-l47/pg19-val/mean}{1.2038}
\defval{pos/long/1024/dense-noloop/finepdfs-long/c1-128}{1.3816}
\defval{pos/long/1024/dense-noloop/finepdfs-long/c1024-2048}{1.0367}
\defval{pos/long/1024/dense-noloop/finepdfs-long/c128-256}{1.1471}
\defval{pos/long/1024/dense-noloop/finepdfs-long/c256-512}{1.1004}
\defval{pos/long/1024/dense-noloop/finepdfs-long/c512-1024}{1.0595}
\defval{pos/long/1024/dense-noloop/finepdfs-long/last1024}{1.0367}
\defval{pos/long/1024/dense-noloop/finepdfs-long/last128}{1.0298}
\defval{pos/long/1024/dense-noloop/finepdfs-long/last256}{1.0286}
\defval{pos/long/1024/dense-noloop/finepdfs-long/last512}{1.0305}
\defval{pos/long/1024/dense-noloop/finepdfs-long/mean}{1.0791}
\defval{pos/long/1024/dense-noloop/fineweb-edu-100bt/c1-128}{0.8858}
\defval{pos/long/1024/dense-noloop/fineweb-edu-100bt/c1024-2048}{0.8463}
\defval{pos/long/1024/dense-noloop/fineweb-edu-100bt/c128-256}{0.8296}
\defval{pos/long/1024/dense-noloop/fineweb-edu-100bt/c256-512}{0.8233}
\defval{pos/long/1024/dense-noloop/fineweb-edu-100bt/c512-1024}{0.8272}
\defval{pos/long/1024/dense-noloop/fineweb-edu-100bt/last1024}{0.8326}
\defval{pos/long/1024/dense-noloop/fineweb-edu-100bt/last128}{0.8294}
\defval{pos/long/1024/dense-noloop/fineweb-edu-100bt/last256}{0.8312}
\defval{pos/long/1024/dense-noloop/fineweb-edu-100bt/last512}{0.8309}
\defval{pos/long/1024/dense-noloop/fineweb-edu-100bt/mean}{0.8406}
\defval{pos/long/1024/dense-noloop/govreport-test/c1-128}{0.8596}
\defval{pos/long/1024/dense-noloop/govreport-test/c1024-2048}{0.6571}
\defval{pos/long/1024/dense-noloop/govreport-test/c128-256}{0.7260}
\defval{pos/long/1024/dense-noloop/govreport-test/c256-512}{0.6957}
\defval{pos/long/1024/dense-noloop/govreport-test/c512-1024}{0.6736}
\defval{pos/long/1024/dense-noloop/govreport-test/last1024}{0.6571}
\defval{pos/long/1024/dense-noloop/govreport-test/last128}{0.6510}
\defval{pos/long/1024/dense-noloop/govreport-test/last256}{0.6514}
\defval{pos/long/1024/dense-noloop/govreport-test/last512}{0.6538}
\defval{pos/long/1024/dense-noloop/govreport-test/mean}{0.6829}
\defval{pos/long/1024/dense-noloop/longdocs/c1-128}{1.2389}
\defval{pos/long/1024/dense-noloop/longdocs/c1024-2048}{0.9804}
\defval{pos/long/1024/dense-noloop/longdocs/c128-256}{1.0640}
\defval{pos/long/1024/dense-noloop/longdocs/c256-512}{1.0258}
\defval{pos/long/1024/dense-noloop/longdocs/c512-1024}{0.9975}
\defval{pos/long/1024/dense-noloop/longdocs/last1024}{0.9804}
\defval{pos/long/1024/dense-noloop/longdocs/last128}{0.9753}
\defval{pos/long/1024/dense-noloop/longdocs/last256}{0.9750}
\defval{pos/long/1024/dense-noloop/longdocs/last512}{0.9768}
\defval{pos/long/1024/dense-noloop/longdocs/mean}{1.0118}
\defval{pos/long/1024/dense-noloop/pg19-test/c1-128}{1.4771}
\defval{pos/long/1024/dense-noloop/pg19-test/c1024-2048}{1.2498}
\defval{pos/long/1024/dense-noloop/pg19-test/c128-256}{1.3199}
\defval{pos/long/1024/dense-noloop/pg19-test/c256-512}{1.2826}
\defval{pos/long/1024/dense-noloop/pg19-test/c512-1024}{1.2598}
\defval{pos/long/1024/dense-noloop/pg19-test/last1024}{1.2498}
\defval{pos/long/1024/dense-noloop/pg19-test/last128}{1.2492}
\defval{pos/long/1024/dense-noloop/pg19-test/last256}{1.2485}
\defval{pos/long/1024/dense-noloop/pg19-test/last512}{1.2490}
\defval{pos/long/1024/dense-noloop/pg19-test/mean}{1.2750}
\defval{pos/long/1024/dense-noloop/pg19-val/c1-128}{1.4357}
\defval{pos/long/1024/dense-noloop/pg19-val/c1024-2048}{1.2143}
\defval{pos/long/1024/dense-noloop/pg19-val/c128-256}{1.2761}
\defval{pos/long/1024/dense-noloop/pg19-val/c256-512}{1.2476}
\defval{pos/long/1024/dense-noloop/pg19-val/c512-1024}{1.2233}
\defval{pos/long/1024/dense-noloop/pg19-val/last1024}{1.2143}
\defval{pos/long/1024/dense-noloop/pg19-val/last128}{1.2050}
\defval{pos/long/1024/dense-noloop/pg19-val/last256}{1.2036}
\defval{pos/long/1024/dense-noloop/pg19-val/last512}{1.2084}
\defval{pos/long/1024/dense-noloop/pg19-val/mean}{1.2384}
\defval{pos/p140/768/p140-dec-fixed/finepdfs-long/c1-128}{1.4370}
\defval{pos/p140/768/p140-dec-fixed/finepdfs-long/c1024-2048}{1.1164}
\defval{pos/p140/768/p140-dec-fixed/finepdfs-long/c128-256}{1.2100}
\defval{pos/p140/768/p140-dec-fixed/finepdfs-long/c256-512}{1.1704}
\defval{pos/p140/768/p140-dec-fixed/finepdfs-long/c512-1024}{1.1347}
\defval{pos/p140/768/p140-dec-fixed/finepdfs-long/last1024}{1.1164}
\defval{pos/p140/768/p140-dec-fixed/finepdfs-long/last128}{1.1102}
\defval{pos/p140/768/p140-dec-fixed/finepdfs-long/last256}{1.1088}
\defval{pos/p140/768/p140-dec-fixed/finepdfs-long/last512}{1.1107}
\defval{pos/p140/768/p140-dec-fixed/finepdfs-long/mean}{1.1539}
\defval{pos/p140/768/p140-dec-fixed/fineweb-edu-100bt/c1-128}{0.9343}
\defval{pos/p140/768/p140-dec-fixed/fineweb-edu-100bt/c1024-2048}{0.9043}
\defval{pos/p140/768/p140-dec-fixed/fineweb-edu-100bt/c128-256}{0.8809}
\defval{pos/p140/768/p140-dec-fixed/fineweb-edu-100bt/c256-512}{0.8754}
\defval{pos/p140/768/p140-dec-fixed/fineweb-edu-100bt/c512-1024}{0.8802}
\defval{pos/p140/768/p140-dec-fixed/fineweb-edu-100bt/last1024}{0.8864}
\defval{pos/p140/768/p140-dec-fixed/fineweb-edu-100bt/last128}{0.8843}
\defval{pos/p140/768/p140-dec-fixed/fineweb-edu-100bt/last256}{0.8849}
\defval{pos/p140/768/p140-dec-fixed/fineweb-edu-100bt/last512}{0.8850}
\defval{pos/p140/768/p140-dec-fixed/fineweb-edu-100bt/mean}{0.8934}
\defval{pos/p140/768/p140-dec-fixed/govreport-test/c1-128}{0.8987}
\defval{pos/p140/768/p140-dec-fixed/govreport-test/c1024-2048}{0.7086}
\defval{pos/p140/768/p140-dec-fixed/govreport-test/c128-256}{0.7694}
\defval{pos/p140/768/p140-dec-fixed/govreport-test/c256-512}{0.7418}
\defval{pos/p140/768/p140-dec-fixed/govreport-test/c512-1024}{0.7229}
\defval{pos/p140/768/p140-dec-fixed/govreport-test/last1024}{0.7086}
\defval{pos/p140/768/p140-dec-fixed/govreport-test/last128}{0.7031}
\defval{pos/p140/768/p140-dec-fixed/govreport-test/last256}{0.7033}
\defval{pos/p140/768/p140-dec-fixed/govreport-test/last512}{0.7056}
\defval{pos/p140/768/p140-dec-fixed/govreport-test/mean}{0.7320}
\defval{pos/p140/768/p140-dec-fixed/longdocs/c1-128}{1.2911}
\defval{pos/p140/768/p140-dec-fixed/longdocs/c1024-2048}{1.0547}
\defval{pos/p140/768/p140-dec-fixed/longdocs/c128-256}{1.1252}
\defval{pos/p140/768/p140-dec-fixed/longdocs/c256-512}{1.0929}
\defval{pos/p140/768/p140-dec-fixed/longdocs/c512-1024}{1.0687}
\defval{pos/p140/768/p140-dec-fixed/longdocs/last1024}{1.0547}
\defval{pos/p140/768/p140-dec-fixed/longdocs/last128}{1.0503}
\defval{pos/p140/768/p140-dec-fixed/longdocs/last256}{1.0499}
\defval{pos/p140/768/p140-dec-fixed/longdocs/last512}{1.0516}
\defval{pos/p140/768/p140-dec-fixed/longdocs/mean}{1.0821}
\defval{pos/p140/768/p140-dec-fixed/pg19-test/c1-128}{1.5394}
\defval{pos/p140/768/p140-dec-fixed/pg19-test/c1024-2048}{1.3416}
\defval{pos/p140/768/p140-dec-fixed/pg19-test/c128-256}{1.3972}
\defval{pos/p140/768/p140-dec-fixed/pg19-test/c256-512}{1.3677}
\defval{pos/p140/768/p140-dec-fixed/pg19-test/c512-1024}{1.3489}
\defval{pos/p140/768/p140-dec-fixed/pg19-test/last1024}{1.3416}
\defval{pos/p140/768/p140-dec-fixed/pg19-test/last128}{1.3420}
\defval{pos/p140/768/p140-dec-fixed/pg19-test/last256}{1.3415}
\defval{pos/p140/768/p140-dec-fixed/pg19-test/last512}{1.3417}
\defval{pos/p140/768/p140-dec-fixed/pg19-test/mean}{1.3625}
\defval{pos/p140/768/p140-dec-fixed/pg19-val/c1-128}{1.4956}
\defval{pos/p140/768/p140-dec-fixed/pg19-val/c1024-2048}{1.3038}
\defval{pos/p140/768/p140-dec-fixed/pg19-val/c128-256}{1.3516}
\defval{pos/p140/768/p140-dec-fixed/pg19-val/c256-512}{1.3279}
\defval{pos/p140/768/p140-dec-fixed/pg19-val/c512-1024}{1.3094}
\defval{pos/p140/768/p140-dec-fixed/pg19-val/last1024}{1.3038}
\defval{pos/p140/768/p140-dec-fixed/pg19-val/last128}{1.2950}
\defval{pos/p140/768/p140-dec-fixed/pg19-val/last256}{1.2943}
\defval{pos/p140/768/p140-dec-fixed/pg19-val/last512}{1.2976}
\defval{pos/p140/768/p140-dec-fixed/pg19-val/mean}{1.3231}
\defval{pos/p140/768/p140-enc-fixed/finepdfs-long/c1-128}{1.4271}
\defval{pos/p140/768/p140-enc-fixed/finepdfs-long/c1024-2048}{1.0837}
\defval{pos/p140/768/p140-enc-fixed/finepdfs-long/c128-256}{1.1917}
\defval{pos/p140/768/p140-enc-fixed/finepdfs-long/c256-512}{1.1461}
\defval{pos/p140/768/p140-enc-fixed/finepdfs-long/c512-1024}{1.1051}
\defval{pos/p140/768/p140-enc-fixed/finepdfs-long/last1024}{1.0837}
\defval{pos/p140/768/p140-enc-fixed/finepdfs-long/last128}{1.0772}
\defval{pos/p140/768/p140-enc-fixed/finepdfs-long/last256}{1.0758}
\defval{pos/p140/768/p140-enc-fixed/finepdfs-long/last512}{1.0776}
\defval{pos/p140/768/p140-enc-fixed/finepdfs-long/mean}{1.1253}
\defval{pos/p140/768/p140-enc-fixed/fineweb-edu-100bt/c1-128}{0.9282}
\defval{pos/p140/768/p140-enc-fixed/fineweb-edu-100bt/c1024-2048}{0.8922}
\defval{pos/p140/768/p140-enc-fixed/fineweb-edu-100bt/c128-256}{0.8722}
\defval{pos/p140/768/p140-enc-fixed/fineweb-edu-100bt/c256-512}{0.8663}
\defval{pos/p140/768/p140-enc-fixed/fineweb-edu-100bt/c512-1024}{0.8705}
\defval{pos/p140/768/p140-enc-fixed/fineweb-edu-100bt/last1024}{0.8763}
\defval{pos/p140/768/p140-enc-fixed/fineweb-edu-100bt/last128}{0.8734}
\defval{pos/p140/768/p140-enc-fixed/fineweb-edu-100bt/last256}{0.8747}
\defval{pos/p140/768/p140-enc-fixed/fineweb-edu-100bt/last512}{0.8749}
\defval{pos/p140/768/p140-enc-fixed/fineweb-edu-100bt/mean}{0.8841}
\defval{pos/p140/768/p140-enc-fixed/govreport-test/c1-128}{0.8937}
\defval{pos/p140/768/p140-enc-fixed/govreport-test/c1024-2048}{0.6930}
\defval{pos/p140/768/p140-enc-fixed/govreport-test/c128-256}{0.7614}
\defval{pos/p140/768/p140-enc-fixed/govreport-test/c256-512}{0.7309}
\defval{pos/p140/768/p140-enc-fixed/govreport-test/c512-1024}{0.7093}
\defval{pos/p140/768/p140-enc-fixed/govreport-test/last1024}{0.6930}
\defval{pos/p140/768/p140-enc-fixed/govreport-test/last128}{0.6868}
\defval{pos/p140/768/p140-enc-fixed/govreport-test/last256}{0.6876}
\defval{pos/p140/768/p140-enc-fixed/govreport-test/last512}{0.6899}
\defval{pos/p140/768/p140-enc-fixed/govreport-test/mean}{0.7186}
\defval{pos/p140/768/p140-enc-fixed/longdocs/c1-128}{1.2840}
\defval{pos/p140/768/p140-enc-fixed/longdocs/c1024-2048}{1.0325}
\defval{pos/p140/768/p140-enc-fixed/longdocs/c128-256}{1.1128}
\defval{pos/p140/768/p140-enc-fixed/longdocs/c256-512}{1.0762}
\defval{pos/p140/768/p140-enc-fixed/longdocs/c512-1024}{1.0486}
\defval{pos/p140/768/p140-enc-fixed/longdocs/last1024}{1.0325}
\defval{pos/p140/768/p140-enc-fixed/longdocs/last128}{1.0279}
\defval{pos/p140/768/p140-enc-fixed/longdocs/last256}{1.0276}
\defval{pos/p140/768/p140-enc-fixed/longdocs/last512}{1.0292}
\defval{pos/p140/768/p140-enc-fixed/longdocs/mean}{1.0628}
\defval{pos/p140/768/p140-enc-fixed/pg19-test/c1-128}{1.5330}
\defval{pos/p140/768/p140-enc-fixed/pg19-test/c1024-2048}{1.3233}
\defval{pos/p140/768/p140-enc-fixed/pg19-test/c128-256}{1.3864}
\defval{pos/p140/768/p140-enc-fixed/pg19-test/c256-512}{1.3528}
\defval{pos/p140/768/p140-enc-fixed/pg19-test/c512-1024}{1.3318}
\defval{pos/p140/768/p140-enc-fixed/pg19-test/last1024}{1.3233}
\defval{pos/p140/768/p140-enc-fixed/pg19-test/last128}{1.3238}
\defval{pos/p140/768/p140-enc-fixed/pg19-test/last256}{1.3229}
\defval{pos/p140/768/p140-enc-fixed/pg19-test/last512}{1.3231}
\defval{pos/p140/768/p140-enc-fixed/pg19-test/mean}{1.3461}
\defval{pos/p140/768/p140-enc-fixed/pg19-val/c1-128}{1.4895}
\defval{pos/p140/768/p140-enc-fixed/pg19-val/c1024-2048}{1.2860}
\defval{pos/p140/768/p140-enc-fixed/pg19-val/c128-256}{1.3431}
\defval{pos/p140/768/p140-enc-fixed/pg19-val/c256-512}{1.3154}
\defval{pos/p140/768/p140-enc-fixed/pg19-val/c512-1024}{1.2931}
\defval{pos/p140/768/p140-enc-fixed/pg19-val/last1024}{1.2860}
\defval{pos/p140/768/p140-enc-fixed/pg19-val/last128}{1.2775}
\defval{pos/p140/768/p140-enc-fixed/pg19-val/last256}{1.2760}
\defval{pos/p140/768/p140-enc-fixed/pg19-val/last512}{1.2798}
\defval{pos/p140/768/p140-enc-fixed/pg19-val/mean}{1.3076}
\defval{pos/p140/768/p140-enc/finepdfs-long/c1-128}{1.4197}
\defval{pos/p140/768/p140-enc/finepdfs-long/c1024-2048}{1.0777}
\defval{pos/p140/768/p140-enc/finepdfs-long/c128-256}{1.1853}
\defval{pos/p140/768/p140-enc/finepdfs-long/c256-512}{1.1399}
\defval{pos/p140/768/p140-enc/finepdfs-long/c512-1024}{1.0993}
\defval{pos/p140/768/p140-enc/finepdfs-long/last1024}{1.0777}
\defval{pos/p140/768/p140-enc/finepdfs-long/last128}{1.0702}
\defval{pos/p140/768/p140-enc/finepdfs-long/last256}{1.0694}
\defval{pos/p140/768/p140-enc/finepdfs-long/last512}{1.0713}
\defval{pos/p140/768/p140-enc/finepdfs-long/mean}{1.1193}
\defval{pos/p140/768/p140-enc/fineweb-edu-100bt/c1-128}{0.9282}
\defval{pos/p140/768/p140-enc/fineweb-edu-100bt/c1024-2048}{0.8923}
\defval{pos/p140/768/p140-enc/fineweb-edu-100bt/c128-256}{0.8719}
\defval{pos/p140/768/p140-enc/fineweb-edu-100bt/c256-512}{0.8668}
\defval{pos/p140/768/p140-enc/fineweb-edu-100bt/c512-1024}{0.8700}
\defval{pos/p140/768/p140-enc/fineweb-edu-100bt/last1024}{0.8765}
\defval{pos/p140/768/p140-enc/fineweb-edu-100bt/last128}{0.8735}
\defval{pos/p140/768/p140-enc/fineweb-edu-100bt/last256}{0.8750}
\defval{pos/p140/768/p140-enc/fineweb-edu-100bt/last512}{0.8746}
\defval{pos/p140/768/p140-enc/fineweb-edu-100bt/mean}{0.8841}
\defval{pos/p140/768/p140-enc/govreport-test/c1-128}{0.8949}
\defval{pos/p140/768/p140-enc/govreport-test/c1024-2048}{0.6930}
\defval{pos/p140/768/p140-enc/govreport-test/c128-256}{0.7616}
\defval{pos/p140/768/p140-enc/govreport-test/c256-512}{0.7311}
\defval{pos/p140/768/p140-enc/govreport-test/c512-1024}{0.7097}
\defval{pos/p140/768/p140-enc/govreport-test/last1024}{0.6930}
\defval{pos/p140/768/p140-enc/govreport-test/last128}{0.6869}
\defval{pos/p140/768/p140-enc/govreport-test/last256}{0.6871}
\defval{pos/p140/768/p140-enc/govreport-test/last512}{0.6896}
\defval{pos/p140/768/p140-enc/govreport-test/mean}{0.7188}
\defval{pos/p140/768/p140-enc/longdocs/c1-128}{1.2808}
\defval{pos/p140/768/p140-enc/longdocs/c1024-2048}{1.0299}
\defval{pos/p140/768/p140-enc/longdocs/c128-256}{1.1094}
\defval{pos/p140/768/p140-enc/longdocs/c256-512}{1.0732}
\defval{pos/p140/768/p140-enc/longdocs/c512-1024}{1.0462}
\defval{pos/p140/768/p140-enc/longdocs/last1024}{1.0299}
\defval{pos/p140/768/p140-enc/longdocs/last128}{1.0246}
\defval{pos/p140/768/p140-enc/longdocs/last256}{1.0243}
\defval{pos/p140/768/p140-enc/longdocs/last512}{1.0262}
\defval{pos/p140/768/p140-enc/longdocs/mean}{1.0601}
\defval{pos/p140/768/p140-enc/pg19-test/c1-128}{1.5295}
\defval{pos/p140/768/p140-enc/pg19-test/c1024-2048}{1.3212}
\defval{pos/p140/768/p140-enc/pg19-test/c128-256}{1.3822}
\defval{pos/p140/768/p140-enc/pg19-test/c256-512}{1.3496}
\defval{pos/p140/768/p140-enc/pg19-test/c512-1024}{1.3298}
\defval{pos/p140/768/p140-enc/pg19-test/last1024}{1.3212}
\defval{pos/p140/768/p140-enc/pg19-test/last128}{1.3205}
\defval{pos/p140/768/p140-enc/pg19-test/last256}{1.3199}
\defval{pos/p140/768/p140-enc/pg19-test/last512}{1.3205}
\defval{pos/p140/768/p140-enc/pg19-test/mean}{1.3437}
\defval{pos/p140/768/p140-enc/pg19-val/c1-128}{1.4867}
\defval{pos/p140/768/p140-enc/pg19-val/c1024-2048}{1.2834}
\defval{pos/p140/768/p140-enc/pg19-val/c128-256}{1.3407}
\defval{pos/p140/768/p140-enc/pg19-val/c256-512}{1.3125}
\defval{pos/p140/768/p140-enc/pg19-val/c512-1024}{1.2911}
\defval{pos/p140/768/p140-enc/pg19-val/last1024}{1.2834}
\defval{pos/p140/768/p140-enc/pg19-val/last128}{1.2740}
\defval{pos/p140/768/p140-enc/pg19-val/last256}{1.2732}
\defval{pos/p140/768/p140-enc/pg19-val/last512}{1.2774}
\defval{pos/p140/768/p140-enc/pg19-val/mean}{1.3052}
\defval{pos/p140/768/p140-noloop/finepdfs-long/c1-128}{1.4501}
\defval{pos/p140/768/p140-noloop/finepdfs-long/c1024-2048}{1.1393}
\defval{pos/p140/768/p140-noloop/finepdfs-long/c128-256}{1.2315}
\defval{pos/p140/768/p140-noloop/finepdfs-long/c256-512}{1.1922}
\defval{pos/p140/768/p140-noloop/finepdfs-long/c512-1024}{1.1565}
\defval{pos/p140/768/p140-noloop/finepdfs-long/last1024}{1.1393}
\defval{pos/p140/768/p140-noloop/finepdfs-long/last128}{1.1338}
\defval{pos/p140/768/p140-noloop/finepdfs-long/last256}{1.1322}
\defval{pos/p140/768/p140-noloop/finepdfs-long/last512}{1.1339}
\defval{pos/p140/768/p140-noloop/finepdfs-long/mean}{1.1757}
\defval{pos/p140/768/p140-noloop/fineweb-edu-100bt/c1-128}{0.9549}
\defval{pos/p140/768/p140-noloop/fineweb-edu-100bt/c1024-2048}{0.9274}
\defval{pos/p140/768/p140-noloop/fineweb-edu-100bt/c128-256}{0.9024}
\defval{pos/p140/768/p140-noloop/fineweb-edu-100bt/c256-512}{0.8984}
\defval{pos/p140/768/p140-noloop/fineweb-edu-100bt/c512-1024}{0.9034}
\defval{pos/p140/768/p140-noloop/fineweb-edu-100bt/last1024}{0.9090}
\defval{pos/p140/768/p140-noloop/fineweb-edu-100bt/last128}{0.9073}
\defval{pos/p140/768/p140-noloop/fineweb-edu-100bt/last256}{0.9080}
\defval{pos/p140/768/p140-noloop/fineweb-edu-100bt/last512}{0.9078}
\defval{pos/p140/768/p140-noloop/fineweb-edu-100bt/mean}{0.9159}
\defval{pos/p140/768/p140-noloop/govreport-test/c1-128}{0.9166}
\defval{pos/p140/768/p140-noloop/govreport-test/c1024-2048}{0.7288}
\defval{pos/p140/768/p140-noloop/govreport-test/c128-256}{0.7896}
\defval{pos/p140/768/p140-noloop/govreport-test/c256-512}{0.7622}
\defval{pos/p140/768/p140-noloop/govreport-test/c512-1024}{0.7434}
\defval{pos/p140/768/p140-noloop/govreport-test/last1024}{0.7288}
\defval{pos/p140/768/p140-noloop/govreport-test/last128}{0.7235}
\defval{pos/p140/768/p140-noloop/govreport-test/last256}{0.7237}
\defval{pos/p140/768/p140-noloop/govreport-test/last512}{0.7260}
\defval{pos/p140/768/p140-noloop/govreport-test/mean}{0.7521}
\defval{pos/p140/768/p140-noloop/longdocs/c1-128}{1.3088}
\defval{pos/p140/768/p140-noloop/longdocs/c1024-2048}{1.0799}
\defval{pos/p140/768/p140-noloop/longdocs/c128-256}{1.1494}
\defval{pos/p140/768/p140-noloop/longdocs/c256-512}{1.1176}
\defval{pos/p140/768/p140-noloop/longdocs/c512-1024}{1.0937}
\defval{pos/p140/768/p140-noloop/longdocs/last1024}{1.0799}
\defval{pos/p140/768/p140-noloop/longdocs/last128}{1.0755}
\defval{pos/p140/768/p140-noloop/longdocs/last256}{1.0751}
\defval{pos/p140/768/p140-noloop/longdocs/last512}{1.0768}
\defval{pos/p140/768/p140-noloop/longdocs/mean}{1.1067}
\defval{pos/p140/768/p140-noloop/pg19-test/c1-128}{1.5612}
\defval{pos/p140/768/p140-noloop/pg19-test/c1024-2048}{1.3741}
\defval{pos/p140/768/p140-noloop/pg19-test/c128-256}{1.4280}
\defval{pos/p140/768/p140-noloop/pg19-test/c256-512}{1.3998}
\defval{pos/p140/768/p140-noloop/pg19-test/c512-1024}{1.3817}
\defval{pos/p140/768/p140-noloop/pg19-test/last1024}{1.3741}
\defval{pos/p140/768/p140-noloop/pg19-test/last128}{1.3738}
\defval{pos/p140/768/p140-noloop/pg19-test/last256}{1.3734}
\defval{pos/p140/768/p140-noloop/pg19-test/last512}{1.3738}
\defval{pos/p140/768/p140-noloop/pg19-test/mean}{1.3942}
\defval{pos/p140/768/p140-noloop/pg19-val/c1-128}{1.5153}
\defval{pos/p140/768/p140-noloop/pg19-val/c1024-2048}{1.3340}
\defval{pos/p140/768/p140-noloop/pg19-val/c128-256}{1.3814}
\defval{pos/p140/768/p140-noloop/pg19-val/c256-512}{1.3595}
\defval{pos/p140/768/p140-noloop/pg19-val/c512-1024}{1.3396}
\defval{pos/p140/768/p140-noloop/pg19-val/last1024}{1.3340}
\defval{pos/p140/768/p140-noloop/pg19-val/last128}{1.3241}
\defval{pos/p140/768/p140-noloop/pg19-val/last256}{1.3237}
\defval{pos/p140/768/p140-noloop/pg19-val/last512}{1.3278}
\defval{pos/p140/768/p140-noloop/pg19-val/mean}{1.3529}
\defval{pos/x40/1024/ced-dec/finepdfs-long/c1-128}{1.4135}
\defval{pos/x40/1024/ced-dec/finepdfs-long/c1024-2048}{1.0707}
\defval{pos/x40/1024/ced-dec/finepdfs-long/c128-256}{1.1747}
\defval{pos/x40/1024/ced-dec/finepdfs-long/c256-512}{1.1303}
\defval{pos/x40/1024/ced-dec/finepdfs-long/c512-1024}{1.0912}
\defval{pos/x40/1024/ced-dec/finepdfs-long/last1024}{1.0707}
\defval{pos/x40/1024/ced-dec/finepdfs-long/last128}{1.0635}
\defval{pos/x40/1024/ced-dec/finepdfs-long/last256}{1.0629}
\defval{pos/x40/1024/ced-dec/finepdfs-long/last512}{1.0645}
\defval{pos/x40/1024/ced-dec/finepdfs-long/mean}{1.1115}
\defval{pos/x40/1024/ced-dec/fineweb-edu-100bt/c1-128}{0.9067}
\defval{pos/x40/1024/ced-dec/fineweb-edu-100bt/c1024-2048}{0.8739}
\defval{pos/x40/1024/ced-dec/fineweb-edu-100bt/c128-256}{0.8522}
\defval{pos/x40/1024/ced-dec/fineweb-edu-100bt/c256-512}{0.8481}
\defval{pos/x40/1024/ced-dec/fineweb-edu-100bt/c512-1024}{0.8513}
\defval{pos/x40/1024/ced-dec/fineweb-edu-100bt/last1024}{0.8573}
\defval{pos/x40/1024/ced-dec/fineweb-edu-100bt/last128}{0.8534}
\defval{pos/x40/1024/ced-dec/fineweb-edu-100bt/last256}{0.8551}
\defval{pos/x40/1024/ced-dec/fineweb-edu-100bt/last512}{0.8557}
\defval{pos/x40/1024/ced-dec/fineweb-edu-100bt/mean}{0.8648}
\defval{pos/x40/1024/ced-dec/govreport-test/c1-128}{0.8774}
\defval{pos/x40/1024/ced-dec/govreport-test/c1024-2048}{0.6827}
\defval{pos/x40/1024/ced-dec/govreport-test/c128-256}{0.7465}
\defval{pos/x40/1024/ced-dec/govreport-test/c256-512}{0.7180}
\defval{pos/x40/1024/ced-dec/govreport-test/c512-1024}{0.6976}
\defval{pos/x40/1024/ced-dec/govreport-test/last1024}{0.6827}
\defval{pos/x40/1024/ced-dec/govreport-test/last128}{0.6772}
\defval{pos/x40/1024/ced-dec/govreport-test/last256}{0.6773}
\defval{pos/x40/1024/ced-dec/govreport-test/last512}{0.6797}
\defval{pos/x40/1024/ced-dec/govreport-test/mean}{0.7069}
\defval{pos/x40/1024/ced-dec/longdocs/c1-128}{1.2639}
\defval{pos/x40/1024/ced-dec/longdocs/c1024-2048}{1.0130}
\defval{pos/x40/1024/ced-dec/longdocs/c128-256}{1.0906}
\defval{pos/x40/1024/ced-dec/longdocs/c256-512}{1.0547}
\defval{pos/x40/1024/ced-dec/longdocs/c512-1024}{1.0282}
\defval{pos/x40/1024/ced-dec/longdocs/last1024}{1.0130}
\defval{pos/x40/1024/ced-dec/longdocs/last128}{1.0084}
\defval{pos/x40/1024/ced-dec/longdocs/last256}{1.0081}
\defval{pos/x40/1024/ced-dec/longdocs/last512}{1.0097}
\defval{pos/x40/1024/ced-dec/longdocs/mean}{1.0426}
\defval{pos/x40/1024/ced-dec/pg19-test/c1-128}{1.5025}
\defval{pos/x40/1024/ced-dec/pg19-test/c1024-2048}{1.2881}
\defval{pos/x40/1024/ced-dec/pg19-test/c128-256}{1.3515}
\defval{pos/x40/1024/ced-dec/pg19-test/c256-512}{1.3171}
\defval{pos/x40/1024/ced-dec/pg19-test/c512-1024}{1.2962}
\defval{pos/x40/1024/ced-dec/pg19-test/last1024}{1.2881}
\defval{pos/x40/1024/ced-dec/pg19-test/last128}{1.2888}
\defval{pos/x40/1024/ced-dec/pg19-test/last256}{1.2876}
\defval{pos/x40/1024/ced-dec/pg19-test/last512}{1.2879}
\defval{pos/x40/1024/ced-dec/pg19-test/mean}{1.3110}
\defval{pos/x40/1024/ced-dec/pg19-val/c1-128}{1.4606}
\defval{pos/x40/1024/ced-dec/pg19-val/c1024-2048}{1.2501}
\defval{pos/x40/1024/ced-dec/pg19-val/c128-256}{1.3095}
\defval{pos/x40/1024/ced-dec/pg19-val/c256-512}{1.2808}
\defval{pos/x40/1024/ced-dec/pg19-val/c512-1024}{1.2581}
\defval{pos/x40/1024/ced-dec/pg19-val/last1024}{1.2501}
\defval{pos/x40/1024/ced-dec/pg19-val/last128}{1.2413}
\defval{pos/x40/1024/ced-dec/pg19-val/last256}{1.2411}
\defval{pos/x40/1024/ced-dec/pg19-val/last512}{1.2443}
\defval{pos/x40/1024/ced-dec/pg19-val/mean}{1.2728}
\defval{pos/x40/1024/ced-enc-fixed/finepdfs-long/c1-128}{1.4024}
\defval{pos/x40/1024/ced-enc-fixed/finepdfs-long/c1024-2048}{1.0410}
\defval{pos/x40/1024/ced-enc-fixed/finepdfs-long/c128-256}{1.1558}
\defval{pos/x40/1024/ced-enc-fixed/finepdfs-long/c256-512}{1.1075}
\defval{pos/x40/1024/ced-enc-fixed/finepdfs-long/c512-1024}{1.0646}
\defval{pos/x40/1024/ced-enc-fixed/finepdfs-long/last1024}{1.0410}
\defval{pos/x40/1024/ced-enc-fixed/finepdfs-long/last128}{1.0336}
\defval{pos/x40/1024/ced-enc-fixed/finepdfs-long/last256}{1.0325}
\defval{pos/x40/1024/ced-enc-fixed/finepdfs-long/last512}{1.0342}
\defval{pos/x40/1024/ced-enc-fixed/finepdfs-long/mean}{1.0853}
\defval{pos/x40/1024/ced-enc-fixed/fineweb-edu-100bt/c1-128}{0.8933}
\defval{pos/x40/1024/ced-enc-fixed/fineweb-edu-100bt/c1024-2048}{0.8549}
\defval{pos/x40/1024/ced-enc-fixed/fineweb-edu-100bt/c128-256}{0.8360}
\defval{pos/x40/1024/ced-enc-fixed/fineweb-edu-100bt/c256-512}{0.8323}
\defval{pos/x40/1024/ced-enc-fixed/fineweb-edu-100bt/c512-1024}{0.8343}
\defval{pos/x40/1024/ced-enc-fixed/fineweb-edu-100bt/last1024}{0.8405}
\defval{pos/x40/1024/ced-enc-fixed/fineweb-edu-100bt/last128}{0.8362}
\defval{pos/x40/1024/ced-enc-fixed/fineweb-edu-100bt/last256}{0.8383}
\defval{pos/x40/1024/ced-enc-fixed/fineweb-edu-100bt/last512}{0.8389}
\defval{pos/x40/1024/ced-enc-fixed/fineweb-edu-100bt/mean}{0.8484}
\defval{pos/x40/1024/ced-enc-fixed/govreport-test/c1-128}{0.8666}
\defval{pos/x40/1024/ced-enc-fixed/govreport-test/c1024-2048}{0.6654}
\defval{pos/x40/1024/ced-enc-fixed/govreport-test/c128-256}{0.7322}
\defval{pos/x40/1024/ced-enc-fixed/govreport-test/c256-512}{0.7023}
\defval{pos/x40/1024/ced-enc-fixed/govreport-test/c512-1024}{0.6812}
\defval{pos/x40/1024/ced-enc-fixed/govreport-test/last1024}{0.6654}
\defval{pos/x40/1024/ced-enc-fixed/govreport-test/last128}{0.6602}
\defval{pos/x40/1024/ced-enc-fixed/govreport-test/last256}{0.6600}
\defval{pos/x40/1024/ced-enc-fixed/govreport-test/last512}{0.6623}
\defval{pos/x40/1024/ced-enc-fixed/govreport-test/mean}{0.6907}
\defval{pos/x40/1024/ced-enc-fixed/longdocs/c1-128}{1.2507}
\defval{pos/x40/1024/ced-enc-fixed/longdocs/c1024-2048}{0.9885}
\defval{pos/x40/1024/ced-enc-fixed/longdocs/c128-256}{1.0711}
\defval{pos/x40/1024/ced-enc-fixed/longdocs/c256-512}{1.0334}
\defval{pos/x40/1024/ced-enc-fixed/longdocs/c512-1024}{1.0054}
\defval{pos/x40/1024/ced-enc-fixed/longdocs/last1024}{0.9885}
\defval{pos/x40/1024/ced-enc-fixed/longdocs/last128}{0.9837}
\defval{pos/x40/1024/ced-enc-fixed/longdocs/last256}{0.9832}
\defval{pos/x40/1024/ced-enc-fixed/longdocs/last512}{0.9848}
\defval{pos/x40/1024/ced-enc-fixed/longdocs/mean}{1.0199}
\defval{pos/x40/1024/ced-enc-fixed/pg19-test/c1-128}{1.4849}
\defval{pos/x40/1024/ced-enc-fixed/pg19-test/c1024-2048}{1.2615}
\defval{pos/x40/1024/ced-enc-fixed/pg19-test/c128-256}{1.3264}
\defval{pos/x40/1024/ced-enc-fixed/pg19-test/c256-512}{1.2917}
\defval{pos/x40/1024/ced-enc-fixed/pg19-test/c512-1024}{1.2708}
\defval{pos/x40/1024/ced-enc-fixed/pg19-test/last1024}{1.2615}
\defval{pos/x40/1024/ced-enc-fixed/pg19-test/last128}{1.2611}
\defval{pos/x40/1024/ced-enc-fixed/pg19-test/last256}{1.2605}
\defval{pos/x40/1024/ced-enc-fixed/pg19-test/last512}{1.2608}
\defval{pos/x40/1024/ced-enc-fixed/pg19-test/mean}{1.2856}
\defval{pos/x40/1024/ced-enc-fixed/pg19-val/c1-128}{1.4431}
\defval{pos/x40/1024/ced-enc-fixed/pg19-val/c1024-2048}{1.2263}
\defval{pos/x40/1024/ced-enc-fixed/pg19-val/c128-256}{1.2848}
\defval{pos/x40/1024/ced-enc-fixed/pg19-val/c256-512}{1.2567}
\defval{pos/x40/1024/ced-enc-fixed/pg19-val/c512-1024}{1.2334}
\defval{pos/x40/1024/ced-enc-fixed/pg19-val/last1024}{1.2263}
\defval{pos/x40/1024/ced-enc-fixed/pg19-val/last128}{1.2154}
\defval{pos/x40/1024/ced-enc-fixed/pg19-val/last256}{1.2151}
\defval{pos/x40/1024/ced-enc-fixed/pg19-val/last512}{1.2195}
\defval{pos/x40/1024/ced-enc-fixed/pg19-val/mean}{1.2490}
\defval{pos/x40/1024/ced-enc/finepdfs-long/c1-128}{1.3911}
\defval{pos/x40/1024/ced-enc/finepdfs-long/c1024-2048}{1.0362}
\defval{pos/x40/1024/ced-enc/finepdfs-long/c128-256}{1.1498}
\defval{pos/x40/1024/ced-enc/finepdfs-long/c256-512}{1.1019}
\defval{pos/x40/1024/ced-enc/finepdfs-long/c512-1024}{1.0595}
\defval{pos/x40/1024/ced-enc/finepdfs-long/last1024}{1.0362}
\defval{pos/x40/1024/ced-enc/finepdfs-long/last128}{1.0288}
\defval{pos/x40/1024/ced-enc/finepdfs-long/last256}{1.0276}
\defval{pos/x40/1024/ced-enc/finepdfs-long/last512}{1.0294}
\defval{pos/x40/1024/ced-enc/finepdfs-long/mean}{1.0798}
\defval{pos/x40/1024/ced-enc/fineweb-edu-100bt/c1-128}{0.8909}
\defval{pos/x40/1024/ced-enc/fineweb-edu-100bt/c1024-2048}{0.8532}
\defval{pos/x40/1024/ced-enc/fineweb-edu-100bt/c128-256}{0.8347}
\defval{pos/x40/1024/ced-enc/fineweb-edu-100bt/c256-512}{0.8304}
\defval{pos/x40/1024/ced-enc/fineweb-edu-100bt/c512-1024}{0.8322}
\defval{pos/x40/1024/ced-enc/fineweb-edu-100bt/last1024}{0.8387}
\defval{pos/x40/1024/ced-enc/fineweb-edu-100bt/last128}{0.8351}
\defval{pos/x40/1024/ced-enc/fineweb-edu-100bt/last256}{0.8365}
\defval{pos/x40/1024/ced-enc/fineweb-edu-100bt/last512}{0.8370}
\defval{pos/x40/1024/ced-enc/fineweb-edu-100bt/mean}{0.8465}
\defval{pos/x40/1024/ced-enc/govreport-test/c1-128}{0.8650}
\defval{pos/x40/1024/ced-enc/govreport-test/c1024-2048}{0.6627}
\defval{pos/x40/1024/ced-enc/govreport-test/c128-256}{0.7307}
\defval{pos/x40/1024/ced-enc/govreport-test/c256-512}{0.7001}
\defval{pos/x40/1024/ced-enc/govreport-test/c512-1024}{0.6788}
\defval{pos/x40/1024/ced-enc/govreport-test/last1024}{0.6627}
\defval{pos/x40/1024/ced-enc/govreport-test/last128}{0.6572}
\defval{pos/x40/1024/ced-enc/govreport-test/last256}{0.6570}
\defval{pos/x40/1024/ced-enc/govreport-test/last512}{0.6596}
\defval{pos/x40/1024/ced-enc/govreport-test/mean}{0.6882}
\defval{pos/x40/1024/ced-enc/longdocs/c1-128}{1.2441}
\defval{pos/x40/1024/ced-enc/longdocs/c1024-2048}{0.9841}
\defval{pos/x40/1024/ced-enc/longdocs/c128-256}{1.0671}
\defval{pos/x40/1024/ced-enc/longdocs/c256-512}{1.0293}
\defval{pos/x40/1024/ced-enc/longdocs/c512-1024}{1.0012}
\defval{pos/x40/1024/ced-enc/longdocs/last1024}{0.9841}
\defval{pos/x40/1024/ced-enc/longdocs/last128}{0.9793}
\defval{pos/x40/1024/ced-enc/longdocs/last256}{0.9787}
\defval{pos/x40/1024/ced-enc/longdocs/last512}{0.9804}
\defval{pos/x40/1024/ced-enc/longdocs/mean}{1.0155}
\defval{pos/x40/1024/ced-enc/pg19-test/c1-128}{1.4779}
\defval{pos/x40/1024/ced-enc/pg19-test/c1024-2048}{1.2558}
\defval{pos/x40/1024/ced-enc/pg19-test/c128-256}{1.3218}
\defval{pos/x40/1024/ced-enc/pg19-test/c256-512}{1.2870}
\defval{pos/x40/1024/ced-enc/pg19-test/c512-1024}{1.2655}
\defval{pos/x40/1024/ced-enc/pg19-test/last1024}{1.2558}
\defval{pos/x40/1024/ced-enc/pg19-test/last128}{1.2559}
\defval{pos/x40/1024/ced-enc/pg19-test/last256}{1.2548}
\defval{pos/x40/1024/ced-enc/pg19-test/last512}{1.2552}
\defval{pos/x40/1024/ced-enc/pg19-test/mean}{1.2801}
\defval{pos/x40/1024/ced-enc/pg19-val/c1-128}{1.4379}
\defval{pos/x40/1024/ced-enc/pg19-val/c1024-2048}{1.2208}
\defval{pos/x40/1024/ced-enc/pg19-val/c128-256}{1.2806}
\defval{pos/x40/1024/ced-enc/pg19-val/c256-512}{1.2515}
\defval{pos/x40/1024/ced-enc/pg19-val/c512-1024}{1.2293}
\defval{pos/x40/1024/ced-enc/pg19-val/last1024}{1.2208}
\defval{pos/x40/1024/ced-enc/pg19-val/last128}{1.2110}
\defval{pos/x40/1024/ced-enc/pg19-val/last256}{1.2101}
\defval{pos/x40/1024/ced-enc/pg19-val/last512}{1.2142}
\defval{pos/x40/1024/ced-enc/pg19-val/mean}{1.2440}
\defval{pos/x40/1024/ced-noloop/finepdfs-long/c1-128}{1.4204}
\defval{pos/x40/1024/ced-noloop/finepdfs-long/c1024-2048}{1.0767}
\defval{pos/x40/1024/ced-noloop/finepdfs-long/c128-256}{1.1851}
\defval{pos/x40/1024/ced-noloop/finepdfs-long/c256-512}{1.1390}
\defval{pos/x40/1024/ced-noloop/finepdfs-long/c512-1024}{1.0984}
\defval{pos/x40/1024/ced-noloop/finepdfs-long/last1024}{1.0767}
\defval{pos/x40/1024/ced-noloop/finepdfs-long/last128}{1.0697}
\defval{pos/x40/1024/ced-noloop/finepdfs-long/last256}{1.0686}
\defval{pos/x40/1024/ced-noloop/finepdfs-long/last512}{1.0705}
\defval{pos/x40/1024/ced-noloop/finepdfs-long/mean}{1.1185}
\defval{pos/x40/1024/ced-noloop/fineweb-edu-100bt/c1-128}{0.9139}
\defval{pos/x40/1024/ced-noloop/fineweb-edu-100bt/c1024-2048}{0.8804}
\defval{pos/x40/1024/ced-noloop/fineweb-edu-100bt/c128-256}{0.8593}
\defval{pos/x40/1024/ced-noloop/fineweb-edu-100bt/c256-512}{0.8549}
\defval{pos/x40/1024/ced-noloop/fineweb-edu-100bt/c512-1024}{0.8579}
\defval{pos/x40/1024/ced-noloop/fineweb-edu-100bt/last1024}{0.8644}
\defval{pos/x40/1024/ced-noloop/fineweb-edu-100bt/last128}{0.8603}
\defval{pos/x40/1024/ced-noloop/fineweb-edu-100bt/last256}{0.8620}
\defval{pos/x40/1024/ced-noloop/fineweb-edu-100bt/last512}{0.8629}
\defval{pos/x40/1024/ced-noloop/fineweb-edu-100bt/mean}{0.8716}
\defval{pos/x40/1024/ced-noloop/govreport-test/c1-128}{0.8826}
\defval{pos/x40/1024/ced-noloop/govreport-test/c1024-2048}{0.6869}
\defval{pos/x40/1024/ced-noloop/govreport-test/c128-256}{0.7509}
\defval{pos/x40/1024/ced-noloop/govreport-test/c256-512}{0.7221}
\defval{pos/x40/1024/ced-noloop/govreport-test/c512-1024}{0.7022}
\defval{pos/x40/1024/ced-noloop/govreport-test/last1024}{0.6869}
\defval{pos/x40/1024/ced-noloop/govreport-test/last128}{0.6811}
\defval{pos/x40/1024/ced-noloop/govreport-test/last256}{0.6815}
\defval{pos/x40/1024/ced-noloop/govreport-test/last512}{0.6840}
\defval{pos/x40/1024/ced-noloop/govreport-test/mean}{0.7113}
\defval{pos/x40/1024/ced-noloop/longdocs/c1-128}{1.2718}
\defval{pos/x40/1024/ced-noloop/longdocs/c1024-2048}{1.0200}
\defval{pos/x40/1024/ced-noloop/longdocs/c128-256}{1.0995}
\defval{pos/x40/1024/ced-noloop/longdocs/c256-512}{1.0631}
\defval{pos/x40/1024/ced-noloop/longdocs/c512-1024}{1.0360}
\defval{pos/x40/1024/ced-noloop/longdocs/last1024}{1.0200}
\defval{pos/x40/1024/ced-noloop/longdocs/last128}{1.0149}
\defval{pos/x40/1024/ced-noloop/longdocs/last256}{1.0148}
\defval{pos/x40/1024/ced-noloop/longdocs/last512}{1.0165}
\defval{pos/x40/1024/ced-noloop/longdocs/mean}{1.0501}
\defval{pos/x40/1024/ced-noloop/pg19-test/c1-128}{1.5140}
\defval{pos/x40/1024/ced-noloop/pg19-test/c1024-2048}{1.2989}
\defval{pos/x40/1024/ced-noloop/pg19-test/c128-256}{1.3634}
\defval{pos/x40/1024/ced-noloop/pg19-test/c256-512}{1.3294}
\defval{pos/x40/1024/ced-noloop/pg19-test/c512-1024}{1.3080}
\defval{pos/x40/1024/ced-noloop/pg19-test/last1024}{1.2989}
\defval{pos/x40/1024/ced-noloop/pg19-test/last128}{1.2981}
\defval{pos/x40/1024/ced-noloop/pg19-test/last256}{1.2980}
\defval{pos/x40/1024/ced-noloop/pg19-test/last512}{1.2982}
\defval{pos/x40/1024/ced-noloop/pg19-test/mean}{1.3224}
\defval{pos/x40/1024/ced-noloop/pg19-val/c1-128}{1.4694}
\defval{pos/x40/1024/ced-noloop/pg19-val/c1024-2048}{1.2616}
\defval{pos/x40/1024/ced-noloop/pg19-val/c128-256}{1.3192}
\defval{pos/x40/1024/ced-noloop/pg19-val/c256-512}{1.2911}
\defval{pos/x40/1024/ced-noloop/pg19-val/c512-1024}{1.2696}
\defval{pos/x40/1024/ced-noloop/pg19-val/last1024}{1.2616}
\defval{pos/x40/1024/ced-noloop/pg19-val/last128}{1.2519}
\defval{pos/x40/1024/ced-noloop/pg19-val/last256}{1.2510}
\defval{pos/x40/1024/ced-noloop/pg19-val/last512}{1.2549}
\defval{pos/x40/1024/ced-noloop/pg19-val/mean}{1.2838}
\defval{pos/x40/1024/dense-l25/finepdfs-long/c1-128}{1.3910}
\defval{pos/x40/1024/dense-l25/finepdfs-long/c1024-2048}{1.0394}
\defval{pos/x40/1024/dense-l25/finepdfs-long/c128-256}{1.1507}
\defval{pos/x40/1024/dense-l25/finepdfs-long/c256-512}{1.1044}
\defval{pos/x40/1024/dense-l25/finepdfs-long/c512-1024}{1.0625}
\defval{pos/x40/1024/dense-l25/finepdfs-long/last1024}{1.0394}
\defval{pos/x40/1024/dense-l25/finepdfs-long/last128}{1.0311}
\defval{pos/x40/1024/dense-l25/finepdfs-long/last256}{1.0305}
\defval{pos/x40/1024/dense-l25/finepdfs-long/last512}{1.0325}
\defval{pos/x40/1024/dense-l25/finepdfs-long/mean}{1.0825}
\defval{pos/x40/1024/dense-l25/fineweb-edu-100bt/c1-128}{0.8931}
\defval{pos/x40/1024/dense-l25/fineweb-edu-100bt/c1024-2048}{0.8512}
\defval{pos/x40/1024/dense-l25/fineweb-edu-100bt/c128-256}{0.8344}
\defval{pos/x40/1024/dense-l25/fineweb-edu-100bt/c256-512}{0.8288}
\defval{pos/x40/1024/dense-l25/fineweb-edu-100bt/c512-1024}{0.8310}
\defval{pos/x40/1024/dense-l25/fineweb-edu-100bt/last1024}{0.8374}
\defval{pos/x40/1024/dense-l25/fineweb-edu-100bt/last128}{0.8331}
\defval{pos/x40/1024/dense-l25/fineweb-edu-100bt/last256}{0.8354}
\defval{pos/x40/1024/dense-l25/fineweb-edu-100bt/last512}{0.8363}
\defval{pos/x40/1024/dense-l25/fineweb-edu-100bt/mean}{0.8457}
\defval{pos/x40/1024/dense-l25/govreport-test/c1-128}{0.8652}
\defval{pos/x40/1024/dense-l25/govreport-test/c1024-2048}{0.6589}
\defval{pos/x40/1024/dense-l25/govreport-test/c128-256}{0.7299}
\defval{pos/x40/1024/dense-l25/govreport-test/c256-512}{0.6980}
\defval{pos/x40/1024/dense-l25/govreport-test/c512-1024}{0.6757}
\defval{pos/x40/1024/dense-l25/govreport-test/last1024}{0.6589}
\defval{pos/x40/1024/dense-l25/govreport-test/last128}{0.6533}
\defval{pos/x40/1024/dense-l25/govreport-test/last256}{0.6534}
\defval{pos/x40/1024/dense-l25/govreport-test/last512}{0.6556}
\defval{pos/x40/1024/dense-l25/govreport-test/mean}{0.6852}
\defval{pos/x40/1024/dense-l25/longdocs/c1-128}{1.2433}
\defval{pos/x40/1024/dense-l25/longdocs/c1024-2048}{0.9834}
\defval{pos/x40/1024/dense-l25/longdocs/c128-256}{1.0666}
\defval{pos/x40/1024/dense-l25/longdocs/c256-512}{1.0287}
\defval{pos/x40/1024/dense-l25/longdocs/c512-1024}{1.0005}
\defval{pos/x40/1024/dense-l25/longdocs/last1024}{0.9834}
\defval{pos/x40/1024/dense-l25/longdocs/last128}{0.9782}
\defval{pos/x40/1024/dense-l25/longdocs/last256}{0.9779}
\defval{pos/x40/1024/dense-l25/longdocs/last512}{0.9797}
\defval{pos/x40/1024/dense-l25/longdocs/mean}{1.0148}
\defval{pos/x40/1024/dense-l25/pg19-test/c1-128}{1.4755}
\defval{pos/x40/1024/dense-l25/pg19-test/c1024-2048}{1.2544}
\defval{pos/x40/1024/dense-l25/pg19-test/c128-256}{1.3203}
\defval{pos/x40/1024/dense-l25/pg19-test/c256-512}{1.2851}
\defval{pos/x40/1024/dense-l25/pg19-test/c512-1024}{1.2636}
\defval{pos/x40/1024/dense-l25/pg19-test/last1024}{1.2544}
\defval{pos/x40/1024/dense-l25/pg19-test/last128}{1.2543}
\defval{pos/x40/1024/dense-l25/pg19-test/last256}{1.2535}
\defval{pos/x40/1024/dense-l25/pg19-test/last512}{1.2540}
\defval{pos/x40/1024/dense-l25/pg19-test/mean}{1.2785}
\defval{pos/x40/1024/dense-l25/pg19-val/c1-128}{1.4333}
\defval{pos/x40/1024/dense-l25/pg19-val/c1024-2048}{1.2188}
\defval{pos/x40/1024/dense-l25/pg19-val/c128-256}{1.2781}
\defval{pos/x40/1024/dense-l25/pg19-val/c256-512}{1.2492}
\defval{pos/x40/1024/dense-l25/pg19-val/c512-1024}{1.2263}
\defval{pos/x40/1024/dense-l25/pg19-val/last1024}{1.2188}
\defval{pos/x40/1024/dense-l25/pg19-val/last128}{1.2092}
\defval{pos/x40/1024/dense-l25/pg19-val/last256}{1.2083}
\defval{pos/x40/1024/dense-l25/pg19-val/last512}{1.2124}
\defval{pos/x40/1024/dense-l25/pg19-val/mean}{1.2415}
\defval{pos/x40/1024/dense-l47/finepdfs-long/c1-128}{1.3942}
\defval{pos/x40/1024/dense-l47/finepdfs-long/c1024-2048}{1.0516}
\defval{pos/x40/1024/dense-l47/finepdfs-long/c128-256}{1.1608}
\defval{pos/x40/1024/dense-l47/finepdfs-long/c256-512}{1.1145}
\defval{pos/x40/1024/dense-l47/finepdfs-long/c512-1024}{1.0736}
\defval{pos/x40/1024/dense-l47/finepdfs-long/last1024}{1.0516}
\defval{pos/x40/1024/dense-l47/finepdfs-long/last128}{1.0445}
\defval{pos/x40/1024/dense-l47/finepdfs-long/last256}{1.0438}
\defval{pos/x40/1024/dense-l47/finepdfs-long/last512}{1.0452}
\defval{pos/x40/1024/dense-l47/finepdfs-long/mean}{1.0935}
\defval{pos/x40/1024/dense-l47/fineweb-edu-100bt/c1-128}{0.8978}
\defval{pos/x40/1024/dense-l47/fineweb-edu-100bt/c1024-2048}{0.8570}
\defval{pos/x40/1024/dense-l47/fineweb-edu-100bt/c128-256}{0.8392}
\defval{pos/x40/1024/dense-l47/fineweb-edu-100bt/c256-512}{0.8343}
\defval{pos/x40/1024/dense-l47/fineweb-edu-100bt/c512-1024}{0.8364}
\defval{pos/x40/1024/dense-l47/fineweb-edu-100bt/last1024}{0.8430}
\defval{pos/x40/1024/dense-l47/fineweb-edu-100bt/last128}{0.8388}
\defval{pos/x40/1024/dense-l47/fineweb-edu-100bt/last256}{0.8406}
\defval{pos/x40/1024/dense-l47/fineweb-edu-100bt/last512}{0.8411}
\defval{pos/x40/1024/dense-l47/fineweb-edu-100bt/mean}{0.8511}
\defval{pos/x40/1024/dense-l47/govreport-test/c1-128}{0.8691}
\defval{pos/x40/1024/dense-l47/govreport-test/c1024-2048}{0.6653}
\defval{pos/x40/1024/dense-l47/govreport-test/c128-256}{0.7344}
\defval{pos/x40/1024/dense-l47/govreport-test/c256-512}{0.7036}
\defval{pos/x40/1024/dense-l47/govreport-test/c512-1024}{0.6817}
\defval{pos/x40/1024/dense-l47/govreport-test/last1024}{0.6653}
\defval{pos/x40/1024/dense-l47/govreport-test/last128}{0.6600}
\defval{pos/x40/1024/dense-l47/govreport-test/last256}{0.6599}
\defval{pos/x40/1024/dense-l47/govreport-test/last512}{0.6622}
\defval{pos/x40/1024/dense-l47/govreport-test/mean}{0.6912}
\defval{pos/x40/1024/dense-l47/longdocs/c1-128}{1.2474}
\defval{pos/x40/1024/dense-l47/longdocs/c1024-2048}{0.9925}
\defval{pos/x40/1024/dense-l47/longdocs/c128-256}{1.0741}
\defval{pos/x40/1024/dense-l47/longdocs/c256-512}{1.0367}
\defval{pos/x40/1024/dense-l47/longdocs/c512-1024}{1.0090}
\defval{pos/x40/1024/dense-l47/longdocs/last1024}{0.9925}
\defval{pos/x40/1024/dense-l47/longdocs/last128}{0.9878}
\defval{pos/x40/1024/dense-l47/longdocs/last256}{0.9874}
\defval{pos/x40/1024/dense-l47/longdocs/last512}{0.9891}
\defval{pos/x40/1024/dense-l47/longdocs/mean}{1.0232}
\defval{pos/x40/1024/dense-l47/pg19-test/c1-128}{1.4805}
\defval{pos/x40/1024/dense-l47/pg19-test/c1024-2048}{1.2632}
\defval{pos/x40/1024/dense-l47/pg19-test/c128-256}{1.3281}
\defval{pos/x40/1024/dense-l47/pg19-test/c256-512}{1.2931}
\defval{pos/x40/1024/dense-l47/pg19-test/c512-1024}{1.2720}
\defval{pos/x40/1024/dense-l47/pg19-test/last1024}{1.2632}
\defval{pos/x40/1024/dense-l47/pg19-test/last128}{1.2630}
\defval{pos/x40/1024/dense-l47/pg19-test/last256}{1.2624}
\defval{pos/x40/1024/dense-l47/pg19-test/last512}{1.2629}
\defval{pos/x40/1024/dense-l47/pg19-test/mean}{1.2868}
\defval{pos/x40/1024/dense-l47/pg19-val/c1-128}{1.4414}
\defval{pos/x40/1024/dense-l47/pg19-val/c1024-2048}{1.2274}
\defval{pos/x40/1024/dense-l47/pg19-val/c128-256}{1.2878}
\defval{pos/x40/1024/dense-l47/pg19-val/c256-512}{1.2572}
\defval{pos/x40/1024/dense-l47/pg19-val/c512-1024}{1.2361}
\defval{pos/x40/1024/dense-l47/pg19-val/last1024}{1.2274}
\defval{pos/x40/1024/dense-l47/pg19-val/last128}{1.2178}
\defval{pos/x40/1024/dense-l47/pg19-val/last256}{1.2178}
\defval{pos/x40/1024/dense-l47/pg19-val/last512}{1.2214}
\defval{pos/x40/1024/dense-l47/pg19-val/mean}{1.2504}
\defval{pos/x40/1024/dense-noloop/finepdfs-long/c1-128}{1.4241}
\defval{pos/x40/1024/dense-noloop/finepdfs-long/c1024-2048}{1.0853}
\defval{pos/x40/1024/dense-noloop/finepdfs-long/c128-256}{1.1912}
\defval{pos/x40/1024/dense-noloop/finepdfs-long/c256-512}{1.1476}
\defval{pos/x40/1024/dense-noloop/finepdfs-long/c512-1024}{1.1067}
\defval{pos/x40/1024/dense-noloop/finepdfs-long/last1024}{1.0853}
\defval{pos/x40/1024/dense-noloop/finepdfs-long/last128}{1.0788}
\defval{pos/x40/1024/dense-noloop/finepdfs-long/last256}{1.0775}
\defval{pos/x40/1024/dense-noloop/finepdfs-long/last512}{1.0792}
\defval{pos/x40/1024/dense-noloop/finepdfs-long/mean}{1.1265}
\defval{pos/x40/1024/dense-noloop/fineweb-edu-100bt/c1-128}{0.9190}
\defval{pos/x40/1024/dense-noloop/fineweb-edu-100bt/c1024-2048}{0.8829}
\defval{pos/x40/1024/dense-noloop/fineweb-edu-100bt/c128-256}{0.8631}
\defval{pos/x40/1024/dense-noloop/fineweb-edu-100bt/c256-512}{0.8569}
\defval{pos/x40/1024/dense-noloop/fineweb-edu-100bt/c512-1024}{0.8601}
\defval{pos/x40/1024/dense-noloop/fineweb-edu-100bt/last1024}{0.8670}
\defval{pos/x40/1024/dense-noloop/fineweb-edu-100bt/last128}{0.8628}
\defval{pos/x40/1024/dense-noloop/fineweb-edu-100bt/last256}{0.8652}
\defval{pos/x40/1024/dense-noloop/fineweb-edu-100bt/last512}{0.8658}
\defval{pos/x40/1024/dense-noloop/fineweb-edu-100bt/mean}{0.8746}
\defval{pos/x40/1024/dense-noloop/govreport-test/c1-128}{0.8868}
\defval{pos/x40/1024/dense-noloop/govreport-test/c1024-2048}{0.6870}
\defval{pos/x40/1024/dense-noloop/govreport-test/c128-256}{0.7542}
\defval{pos/x40/1024/dense-noloop/govreport-test/c256-512}{0.7239}
\defval{pos/x40/1024/dense-noloop/govreport-test/c512-1024}{0.7027}
\defval{pos/x40/1024/dense-noloop/govreport-test/last1024}{0.6870}
\defval{pos/x40/1024/dense-noloop/govreport-test/last128}{0.6816}
\defval{pos/x40/1024/dense-noloop/govreport-test/last256}{0.6817}
\defval{pos/x40/1024/dense-noloop/govreport-test/last512}{0.6839}
\defval{pos/x40/1024/dense-noloop/govreport-test/mean}{0.7122}
\defval{pos/x40/1024/dense-noloop/longdocs/c1-128}{1.2757}
\defval{pos/x40/1024/dense-noloop/longdocs/c1024-2048}{1.0263}
\defval{pos/x40/1024/dense-noloop/longdocs/c128-256}{1.1054}
\defval{pos/x40/1024/dense-noloop/longdocs/c256-512}{1.0695}
\defval{pos/x40/1024/dense-noloop/longdocs/c512-1024}{1.0420}
\defval{pos/x40/1024/dense-noloop/longdocs/last1024}{1.0263}
\defval{pos/x40/1024/dense-noloop/longdocs/last128}{1.0218}
\defval{pos/x40/1024/dense-noloop/longdocs/last256}{1.0213}
\defval{pos/x40/1024/dense-noloop/longdocs/last512}{1.0230}
\defval{pos/x40/1024/dense-noloop/longdocs/mean}{1.0562}
\defval{pos/x40/1024/dense-noloop/pg19-test/c1-128}{1.5178}
\defval{pos/x40/1024/dense-noloop/pg19-test/c1024-2048}{1.3092}
\defval{pos/x40/1024/dense-noloop/pg19-test/c128-256}{1.3719}
\defval{pos/x40/1024/dense-noloop/pg19-test/c256-512}{1.3383}
\defval{pos/x40/1024/dense-noloop/pg19-test/c512-1024}{1.3171}
\defval{pos/x40/1024/dense-noloop/pg19-test/last1024}{1.3092}
\defval{pos/x40/1024/dense-noloop/pg19-test/last128}{1.3094}
\defval{pos/x40/1024/dense-noloop/pg19-test/last256}{1.3086}
\defval{pos/x40/1024/dense-noloop/pg19-test/last512}{1.3090}
\defval{pos/x40/1024/dense-noloop/pg19-test/mean}{1.3317}
\defval{pos/x40/1024/dense-noloop/pg19-val/c1-128}{1.4719}
\defval{pos/x40/1024/dense-noloop/pg19-val/c1024-2048}{1.2702}
\defval{pos/x40/1024/dense-noloop/pg19-val/c128-256}{1.3262}
\defval{pos/x40/1024/dense-noloop/pg19-val/c256-512}{1.2991}
\defval{pos/x40/1024/dense-noloop/pg19-val/c512-1024}{1.2779}
\defval{pos/x40/1024/dense-noloop/pg19-val/last1024}{1.2702}
\defval{pos/x40/1024/dense-noloop/pg19-val/last128}{1.2619}
\defval{pos/x40/1024/dense-noloop/pg19-val/last256}{1.2605}
\defval{pos/x40/1024/dense-noloop/pg19-val/last512}{1.2638}
\defval{pos/x40/1024/dense-noloop/pg19-val/mean}{1.2918}
\defval{pos/x40/1024/yocou-loop/finepdfs-long/c1-128}{1.3837}
\defval{pos/x40/1024/yocou-loop/finepdfs-long/c1024-2048}{1.0506}
\defval{pos/x40/1024/yocou-loop/finepdfs-long/c128-256}{1.1493}
\defval{pos/x40/1024/yocou-loop/finepdfs-long/c256-512}{1.1077}
\defval{pos/x40/1024/yocou-loop/finepdfs-long/c512-1024}{1.0701}
\defval{pos/x40/1024/yocou-loop/finepdfs-long/last1024}{1.0506}
\defval{pos/x40/1024/yocou-loop/finepdfs-long/last128}{1.0444}
\defval{pos/x40/1024/yocou-loop/finepdfs-long/last256}{1.0431}
\defval{pos/x40/1024/yocou-loop/finepdfs-long/last512}{1.0445}
\defval{pos/x40/1024/yocou-loop/finepdfs-long/mean}{1.0899}
\defval{pos/x40/1024/yocou-loop/fineweb-edu-100bt/c1-128}{0.8914}
\defval{pos/x40/1024/yocou-loop/fineweb-edu-100bt/c1024-2048}{0.8603}
\defval{pos/x40/1024/yocou-loop/fineweb-edu-100bt/c128-256}{0.8367}
\defval{pos/x40/1024/yocou-loop/fineweb-edu-100bt/c256-512}{0.8333}
\defval{pos/x40/1024/yocou-loop/fineweb-edu-100bt/c512-1024}{0.8373}
\defval{pos/x40/1024/yocou-loop/fineweb-edu-100bt/last1024}{0.8434}
\defval{pos/x40/1024/yocou-loop/fineweb-edu-100bt/last128}{0.8404}
\defval{pos/x40/1024/yocou-loop/fineweb-edu-100bt/last256}{0.8416}
\defval{pos/x40/1024/yocou-loop/fineweb-edu-100bt/last512}{0.8421}
\defval{pos/x40/1024/yocou-loop/fineweb-edu-100bt/mean}{0.8503}
\defval{pos/x40/1024/yocou-loop/govreport-test/c1-128}{0.8646}
\defval{pos/x40/1024/yocou-loop/govreport-test/c1024-2048}{0.6718}
\defval{pos/x40/1024/yocou-loop/govreport-test/c128-256}{0.7320}
\defval{pos/x40/1024/yocou-loop/govreport-test/c256-512}{0.7044}
\defval{pos/x40/1024/yocou-loop/govreport-test/c512-1024}{0.6856}
\defval{pos/x40/1024/yocou-loop/govreport-test/last1024}{0.6718}
\defval{pos/x40/1024/yocou-loop/govreport-test/last128}{0.6673}
\defval{pos/x40/1024/yocou-loop/govreport-test/last256}{0.6671}
\defval{pos/x40/1024/yocou-loop/govreport-test/last512}{0.6692}
\defval{pos/x40/1024/yocou-loop/govreport-test/mean}{0.6951}
\defval{pos/x40/1024/yocou-loop/longdocs/c1-128}{1.2391}
\defval{pos/x40/1024/yocou-loop/longdocs/c1024-2048}{0.9948}
\defval{pos/x40/1024/yocou-loop/longdocs/c128-256}{1.0668}
\defval{pos/x40/1024/yocou-loop/longdocs/c256-512}{1.0334}
\defval{pos/x40/1024/yocou-loop/longdocs/c512-1024}{1.0089}
\defval{pos/x40/1024/yocou-loop/longdocs/last1024}{0.9948}
\defval{pos/x40/1024/yocou-loop/longdocs/last128}{0.9907}
\defval{pos/x40/1024/yocou-loop/longdocs/last256}{0.9901}
\defval{pos/x40/1024/yocou-loop/longdocs/last512}{0.9916}
\defval{pos/x40/1024/yocou-loop/longdocs/mean}{1.0229}
\defval{pos/x40/1024/yocou-loop/pg19-test/c1-128}{1.4708}
\defval{pos/x40/1024/yocou-loop/pg19-test/c1024-2048}{1.2644}
\defval{pos/x40/1024/yocou-loop/pg19-test/c128-256}{1.3201}
\defval{pos/x40/1024/yocou-loop/pg19-test/c256-512}{1.2891}
\defval{pos/x40/1024/yocou-loop/pg19-test/c512-1024}{1.2714}
\defval{pos/x40/1024/yocou-loop/pg19-test/last1024}{1.2644}
\defval{pos/x40/1024/yocou-loop/pg19-test/last128}{1.2644}
\defval{pos/x40/1024/yocou-loop/pg19-test/last256}{1.2636}
\defval{pos/x40/1024/yocou-loop/pg19-test/last512}{1.2641}
\defval{pos/x40/1024/yocou-loop/pg19-test/mean}{1.2856}
\defval{pos/x40/1024/yocou-loop/pg19-val/c1-128}{1.4310}
\defval{pos/x40/1024/yocou-loop/pg19-val/c1024-2048}{1.2290}
\defval{pos/x40/1024/yocou-loop/pg19-val/c128-256}{1.2814}
\defval{pos/x40/1024/yocou-loop/pg19-val/c256-512}{1.2555}
\defval{pos/x40/1024/yocou-loop/pg19-val/c512-1024}{1.2354}
\defval{pos/x40/1024/yocou-loop/pg19-val/last1024}{1.2290}
\defval{pos/x40/1024/yocou-loop/pg19-val/last128}{1.2189}
\defval{pos/x40/1024/yocou-loop/pg19-val/last256}{1.2186}
\defval{pos/x40/1024/yocou-loop/pg19-val/last512}{1.2225}
\defval{pos/x40/1024/yocou-loop/pg19-val/mean}{1.2498}
\defval{pos/x40/1024/yocou-noloop/finepdfs-long/c1-128}{1.4056}
\defval{pos/x40/1024/yocou-noloop/finepdfs-long/c1024-2048}{1.0769}
\defval{pos/x40/1024/yocou-noloop/finepdfs-long/c128-256}{1.1755}
\defval{pos/x40/1024/yocou-noloop/finepdfs-long/c256-512}{1.1341}
\defval{pos/x40/1024/yocou-noloop/finepdfs-long/c512-1024}{1.0965}
\defval{pos/x40/1024/yocou-noloop/finepdfs-long/last1024}{1.0769}
\defval{pos/x40/1024/yocou-noloop/finepdfs-long/last128}{1.0711}
\defval{pos/x40/1024/yocou-noloop/finepdfs-long/last256}{1.0699}
\defval{pos/x40/1024/yocou-noloop/finepdfs-long/last512}{1.0712}
\defval{pos/x40/1024/yocou-noloop/finepdfs-long/mean}{1.1160}
\defval{pos/x40/1024/yocou-noloop/fineweb-edu-100bt/c1-128}{0.9148}
\defval{pos/x40/1024/yocou-noloop/fineweb-edu-100bt/c1024-2048}{0.8845}
\defval{pos/x40/1024/yocou-noloop/fineweb-edu-100bt/c128-256}{0.8612}
\defval{pos/x40/1024/yocou-noloop/fineweb-edu-100bt/c256-512}{0.8571}
\defval{pos/x40/1024/yocou-noloop/fineweb-edu-100bt/c512-1024}{0.8616}
\defval{pos/x40/1024/yocou-noloop/fineweb-edu-100bt/last1024}{0.8672}
\defval{pos/x40/1024/yocou-noloop/fineweb-edu-100bt/last128}{0.8636}
\defval{pos/x40/1024/yocou-noloop/fineweb-edu-100bt/last256}{0.8648}
\defval{pos/x40/1024/yocou-noloop/fineweb-edu-100bt/last512}{0.8655}
\defval{pos/x40/1024/yocou-noloop/fineweb-edu-100bt/mean}{0.8743}
\defval{pos/x40/1024/yocou-noloop/govreport-test/c1-128}{0.8832}
\defval{pos/x40/1024/yocou-noloop/govreport-test/c1024-2048}{0.6923}
\defval{pos/x40/1024/yocou-noloop/govreport-test/c128-256}{0.7539}
\defval{pos/x40/1024/yocou-noloop/govreport-test/c256-512}{0.7259}
\defval{pos/x40/1024/yocou-noloop/govreport-test/c512-1024}{0.7067}
\defval{pos/x40/1024/yocou-noloop/govreport-test/last1024}{0.6923}
\defval{pos/x40/1024/yocou-noloop/govreport-test/last128}{0.6875}
\defval{pos/x40/1024/yocou-noloop/govreport-test/last256}{0.6875}
\defval{pos/x40/1024/yocou-noloop/govreport-test/last512}{0.6895}
\defval{pos/x40/1024/yocou-noloop/govreport-test/mean}{0.7158}
\defval{pos/x40/1024/yocou-noloop/longdocs/c1-128}{1.2644}
\defval{pos/x40/1024/yocou-noloop/longdocs/c1024-2048}{1.0237}
\defval{pos/x40/1024/yocou-noloop/longdocs/c128-256}{1.0960}
\defval{pos/x40/1024/yocou-noloop/longdocs/c256-512}{1.0627}
\defval{pos/x40/1024/yocou-noloop/longdocs/c512-1024}{1.0381}
\defval{pos/x40/1024/yocou-noloop/longdocs/last1024}{1.0237}
\defval{pos/x40/1024/yocou-noloop/longdocs/last128}{1.0197}
\defval{pos/x40/1024/yocou-noloop/longdocs/last256}{1.0192}
\defval{pos/x40/1024/yocou-noloop/longdocs/last512}{1.0206}
\defval{pos/x40/1024/yocou-noloop/longdocs/mean}{1.0518}
\defval{pos/x40/1024/yocou-noloop/pg19-test/c1-128}{1.5059}
\defval{pos/x40/1024/yocou-noloop/pg19-test/c1024-2048}{1.3044}
\defval{pos/x40/1024/yocou-noloop/pg19-test/c128-256}{1.3594}
\defval{pos/x40/1024/yocou-noloop/pg19-test/c256-512}{1.3293}
\defval{pos/x40/1024/yocou-noloop/pg19-test/c512-1024}{1.3116}
\defval{pos/x40/1024/yocou-noloop/pg19-test/last1024}{1.3044}
\defval{pos/x40/1024/yocou-noloop/pg19-test/last128}{1.3047}
\defval{pos/x40/1024/yocou-noloop/pg19-test/last256}{1.3039}
\defval{pos/x40/1024/yocou-noloop/pg19-test/last512}{1.3041}
\defval{pos/x40/1024/yocou-noloop/pg19-test/mean}{1.3253}
\defval{pos/x40/1024/yocou-noloop/pg19-val/c1-128}{1.4630}
\defval{pos/x40/1024/yocou-noloop/pg19-val/c1024-2048}{1.2666}
\defval{pos/x40/1024/yocou-noloop/pg19-val/c128-256}{1.3167}
\defval{pos/x40/1024/yocou-noloop/pg19-val/c256-512}{1.2916}
\defval{pos/x40/1024/yocou-noloop/pg19-val/c512-1024}{1.2733}
\defval{pos/x40/1024/yocou-noloop/pg19-val/last1024}{1.2666}
\defval{pos/x40/1024/yocou-noloop/pg19-val/last128}{1.2568}
\defval{pos/x40/1024/yocou-noloop/pg19-val/last256}{1.2559}
\defval{pos/x40/1024/yocou-noloop/pg19-val/last512}{1.2599}
\defval{pos/x40/1024/yocou-noloop/pg19-val/mean}{1.2868}
\defval{pos/x40/1536/ced-dec/finepdfs-long/c1-128}{1.3372}
\defval{pos/x40/1536/ced-dec/finepdfs-long/c1024-2048}{0.9952}
\defval{pos/x40/1536/ced-dec/finepdfs-long/c128-256}{1.0986}
\defval{pos/x40/1536/ced-dec/finepdfs-long/c256-512}{1.0548}
\defval{pos/x40/1536/ced-dec/finepdfs-long/c512-1024}{1.0167}
\defval{pos/x40/1536/ced-dec/finepdfs-long/last1024}{0.9952}
\defval{pos/x40/1536/ced-dec/finepdfs-long/last128}{0.9885}
\defval{pos/x40/1536/ced-dec/finepdfs-long/last256}{0.9876}
\defval{pos/x40/1536/ced-dec/finepdfs-long/last512}{0.9892}
\defval{pos/x40/1536/ced-dec/finepdfs-long/mean}{1.0362}
\defval{pos/x40/1536/ced-dec/fineweb-edu-100bt/c1-128}{0.8583}
\defval{pos/x40/1536/ced-dec/fineweb-edu-100bt/c1024-2048}{0.8199}
\defval{pos/x40/1536/ced-dec/fineweb-edu-100bt/c128-256}{0.8025}
\defval{pos/x40/1536/ced-dec/fineweb-edu-100bt/c256-512}{0.7968}
\defval{pos/x40/1536/ced-dec/fineweb-edu-100bt/c512-1024}{0.8004}
\defval{pos/x40/1536/ced-dec/fineweb-edu-100bt/last1024}{0.8059}
\defval{pos/x40/1536/ced-dec/fineweb-edu-100bt/last128}{0.8028}
\defval{pos/x40/1536/ced-dec/fineweb-edu-100bt/last256}{0.8040}
\defval{pos/x40/1536/ced-dec/fineweb-edu-100bt/last512}{0.8043}
\defval{pos/x40/1536/ced-dec/fineweb-edu-100bt/mean}{0.8138}
\defval{pos/x40/1536/ced-dec/govreport-test/c1-128}{0.8366}
\defval{pos/x40/1536/ced-dec/govreport-test/c1024-2048}{0.6397}
\defval{pos/x40/1536/ced-dec/govreport-test/c128-256}{0.7039}
\defval{pos/x40/1536/ced-dec/govreport-test/c256-512}{0.6749}
\defval{pos/x40/1536/ced-dec/govreport-test/c512-1024}{0.6547}
\defval{pos/x40/1536/ced-dec/govreport-test/last1024}{0.6397}
\defval{pos/x40/1536/ced-dec/govreport-test/last128}{0.6345}
\defval{pos/x40/1536/ced-dec/govreport-test/last256}{0.6345}
\defval{pos/x40/1536/ced-dec/govreport-test/last512}{0.6367}
\defval{pos/x40/1536/ced-dec/govreport-test/mean}{0.6641}
\defval{pos/x40/1536/ced-dec/longdocs/c1-128}{1.2001}
\defval{pos/x40/1536/ced-dec/longdocs/c1024-2048}{0.9456}
\defval{pos/x40/1536/ced-dec/longdocs/c128-256}{1.0230}
\defval{pos/x40/1536/ced-dec/longdocs/c256-512}{0.9870}
\defval{pos/x40/1536/ced-dec/longdocs/c512-1024}{0.9612}
\defval{pos/x40/1536/ced-dec/longdocs/last1024}{0.9456}
\defval{pos/x40/1536/ced-dec/longdocs/last128}{0.9413}
\defval{pos/x40/1536/ced-dec/longdocs/last256}{0.9407}
\defval{pos/x40/1536/ced-dec/longdocs/last512}{0.9422}
\defval{pos/x40/1536/ced-dec/longdocs/mean}{0.9754}
\defval{pos/x40/1536/ced-dec/pg19-test/c1-128}{1.4281}
\defval{pos/x40/1536/ced-dec/pg19-test/c1024-2048}{1.2040}
\defval{pos/x40/1536/ced-dec/pg19-test/c128-256}{1.2675}
\defval{pos/x40/1536/ced-dec/pg19-test/c256-512}{1.2322}
\defval{pos/x40/1536/ced-dec/pg19-test/c512-1024}{1.2126}
\defval{pos/x40/1536/ced-dec/pg19-test/last1024}{1.2040}
\defval{pos/x40/1536/ced-dec/pg19-test/last128}{1.2047}
\defval{pos/x40/1536/ced-dec/pg19-test/last256}{1.2033}
\defval{pos/x40/1536/ced-dec/pg19-test/last512}{1.2034}
\defval{pos/x40/1536/ced-dec/pg19-test/mean}{1.2276}
\defval{pos/x40/1536/ced-dec/pg19-val/c1-128}{1.3894}
\defval{pos/x40/1536/ced-dec/pg19-val/c1024-2048}{1.1690}
\defval{pos/x40/1536/ced-dec/pg19-val/c128-256}{1.2274}
\defval{pos/x40/1536/ced-dec/pg19-val/c256-512}{1.1996}
\defval{pos/x40/1536/ced-dec/pg19-val/c512-1024}{1.1765}
\defval{pos/x40/1536/ced-dec/pg19-val/last1024}{1.1690}
\defval{pos/x40/1536/ced-dec/pg19-val/last128}{1.1597}
\defval{pos/x40/1536/ced-dec/pg19-val/last256}{1.1592}
\defval{pos/x40/1536/ced-dec/pg19-val/last512}{1.1631}
\defval{pos/x40/1536/ced-dec/pg19-val/mean}{1.1921}
\defval{pos/x40/1536/ced-enc/finepdfs-long/c1-128}{1.3402}
\defval{pos/x40/1536/ced-enc/finepdfs-long/c1024-2048}{0.9753}
\defval{pos/x40/1536/ced-enc/finepdfs-long/c128-256}{1.0912}
\defval{pos/x40/1536/ced-enc/finepdfs-long/c256-512}{1.0428}
\defval{pos/x40/1536/ced-enc/finepdfs-long/c512-1024}{0.9994}
\defval{pos/x40/1536/ced-enc/finepdfs-long/last1024}{0.9753}
\defval{pos/x40/1536/ced-enc/finepdfs-long/last128}{0.9668}
\defval{pos/x40/1536/ced-enc/finepdfs-long/last256}{0.9662}
\defval{pos/x40/1536/ced-enc/finepdfs-long/last512}{0.9682}
\defval{pos/x40/1536/ced-enc/finepdfs-long/mean}{1.0202}
\defval{pos/x40/1536/ced-enc/fineweb-edu-100bt/c1-128}{0.8448}
\defval{pos/x40/1536/ced-enc/fineweb-edu-100bt/c1024-2048}{0.8001}
\defval{pos/x40/1536/ced-enc/fineweb-edu-100bt/c128-256}{0.7868}
\defval{pos/x40/1536/ced-enc/fineweb-edu-100bt/c256-512}{0.7801}
\defval{pos/x40/1536/ced-enc/fineweb-edu-100bt/c512-1024}{0.7817}
\defval{pos/x40/1536/ced-enc/fineweb-edu-100bt/last1024}{0.7884}
\defval{pos/x40/1536/ced-enc/fineweb-edu-100bt/last128}{0.7850}
\defval{pos/x40/1536/ced-enc/fineweb-edu-100bt/last256}{0.7861}
\defval{pos/x40/1536/ced-enc/fineweb-edu-100bt/last512}{0.7869}
\defval{pos/x40/1536/ced-enc/fineweb-edu-100bt/mean}{0.7966}
\defval{pos/x40/1536/ced-enc/govreport-test/c1-128}{0.8247}
\defval{pos/x40/1536/ced-enc/govreport-test/c1024-2048}{0.6207}
\defval{pos/x40/1536/ced-enc/govreport-test/c128-256}{0.6893}
\defval{pos/x40/1536/ced-enc/govreport-test/c256-512}{0.6582}
\defval{pos/x40/1536/ced-enc/govreport-test/c512-1024}{0.6366}
\defval{pos/x40/1536/ced-enc/govreport-test/last1024}{0.6207}
\defval{pos/x40/1536/ced-enc/govreport-test/last128}{0.6153}
\defval{pos/x40/1536/ced-enc/govreport-test/last256}{0.6154}
\defval{pos/x40/1536/ced-enc/govreport-test/last512}{0.6176}
\defval{pos/x40/1536/ced-enc/govreport-test/mean}{0.6463}
\defval{pos/x40/1536/ced-enc/longdocs/c1-128}{1.1918}
\defval{pos/x40/1536/ced-enc/longdocs/c1024-2048}{0.9240}
\defval{pos/x40/1536/ced-enc/longdocs/c128-256}{1.0084}
\defval{pos/x40/1536/ced-enc/longdocs/c256-512}{0.9697}
\defval{pos/x40/1536/ced-enc/longdocs/c512-1024}{0.9411}
\defval{pos/x40/1536/ced-enc/longdocs/last1024}{0.9240}
\defval{pos/x40/1536/ced-enc/longdocs/last128}{0.9191}
\defval{pos/x40/1536/ced-enc/longdocs/last256}{0.9185}
\defval{pos/x40/1536/ced-enc/longdocs/last512}{0.9203}
\defval{pos/x40/1536/ced-enc/longdocs/mean}{0.9560}
\defval{pos/x40/1536/ced-enc/pg19-test/c1-128}{1.4121}
\defval{pos/x40/1536/ced-enc/pg19-test/c1024-2048}{1.1783}
\defval{pos/x40/1536/ced-enc/pg19-test/c128-256}{1.2457}
\defval{pos/x40/1536/ced-enc/pg19-test/c256-512}{1.2093}
\defval{pos/x40/1536/ced-enc/pg19-test/c512-1024}{1.1876}
\defval{pos/x40/1536/ced-enc/pg19-test/last1024}{1.1783}
\defval{pos/x40/1536/ced-enc/pg19-test/last128}{1.1789}
\defval{pos/x40/1536/ced-enc/pg19-test/last256}{1.1773}
\defval{pos/x40/1536/ced-enc/pg19-test/last512}{1.1777}
\defval{pos/x40/1536/ced-enc/pg19-test/mean}{1.2033}
\defval{pos/x40/1536/ced-enc/pg19-val/c1-128}{1.3745}
\defval{pos/x40/1536/ced-enc/pg19-val/c1024-2048}{1.1444}
\defval{pos/x40/1536/ced-enc/pg19-val/c128-256}{1.2067}
\defval{pos/x40/1536/ced-enc/pg19-val/c256-512}{1.1771}
\defval{pos/x40/1536/ced-enc/pg19-val/c512-1024}{1.1522}
\defval{pos/x40/1536/ced-enc/pg19-val/last1024}{1.1444}
\defval{pos/x40/1536/ced-enc/pg19-val/last128}{1.1342}
\defval{pos/x40/1536/ced-enc/pg19-val/last256}{1.1342}
\defval{pos/x40/1536/ced-enc/pg19-val/last512}{1.1384}
\defval{pos/x40/1536/ced-enc/pg19-val/mean}{1.1687}
\defval{pos/x40/1536/ced-noloop/finepdfs-long/c1-128}{1.3439}
\defval{pos/x40/1536/ced-noloop/finepdfs-long/c1024-2048}{1.0047}
\defval{pos/x40/1536/ced-noloop/finepdfs-long/c128-256}{1.1116}
\defval{pos/x40/1536/ced-noloop/finepdfs-long/c256-512}{1.0660}
\defval{pos/x40/1536/ced-noloop/finepdfs-long/c512-1024}{1.0265}
\defval{pos/x40/1536/ced-noloop/finepdfs-long/last1024}{1.0047}
\defval{pos/x40/1536/ced-noloop/finepdfs-long/last128}{0.9980}
\defval{pos/x40/1536/ced-noloop/finepdfs-long/last256}{0.9968}
\defval{pos/x40/1536/ced-noloop/finepdfs-long/last512}{0.9985}
\defval{pos/x40/1536/ced-noloop/finepdfs-long/mean}{1.0460}
\defval{pos/x40/1536/ced-noloop/fineweb-edu-100bt/c1-128}{0.8692}
\defval{pos/x40/1536/ced-noloop/fineweb-edu-100bt/c1024-2048}{0.8312}
\defval{pos/x40/1536/ced-noloop/fineweb-edu-100bt/c128-256}{0.8138}
\defval{pos/x40/1536/ced-noloop/fineweb-edu-100bt/c256-512}{0.8074}
\defval{pos/x40/1536/ced-noloop/fineweb-edu-100bt/c512-1024}{0.8109}
\defval{pos/x40/1536/ced-noloop/fineweb-edu-100bt/last1024}{0.8170}
\defval{pos/x40/1536/ced-noloop/fineweb-edu-100bt/last128}{0.8140}
\defval{pos/x40/1536/ced-noloop/fineweb-edu-100bt/last256}{0.8157}
\defval{pos/x40/1536/ced-noloop/fineweb-edu-100bt/last512}{0.8157}
\defval{pos/x40/1536/ced-noloop/fineweb-edu-100bt/mean}{0.8247}
\defval{pos/x40/1536/ced-noloop/govreport-test/c1-128}{0.8447}
\defval{pos/x40/1536/ced-noloop/govreport-test/c1024-2048}{0.6485}
\defval{pos/x40/1536/ced-noloop/govreport-test/c128-256}{0.7131}
\defval{pos/x40/1536/ced-noloop/govreport-test/c256-512}{0.6838}
\defval{pos/x40/1536/ced-noloop/govreport-test/c512-1024}{0.6633}
\defval{pos/x40/1536/ced-noloop/govreport-test/last1024}{0.6485}
\defval{pos/x40/1536/ced-noloop/govreport-test/last128}{0.6432}
\defval{pos/x40/1536/ced-noloop/govreport-test/last256}{0.6432}
\defval{pos/x40/1536/ced-noloop/govreport-test/last512}{0.6456}
\defval{pos/x40/1536/ced-noloop/govreport-test/mean}{0.6728}
\defval{pos/x40/1536/ced-noloop/longdocs/c1-128}{1.2102}
\defval{pos/x40/1536/ced-noloop/longdocs/c1024-2048}{0.9589}
\defval{pos/x40/1536/ced-noloop/longdocs/c128-256}{1.0377}
\defval{pos/x40/1536/ced-noloop/longdocs/c256-512}{1.0012}
\defval{pos/x40/1536/ced-noloop/longdocs/c512-1024}{0.9747}
\defval{pos/x40/1536/ced-noloop/longdocs/last1024}{0.9589}
\defval{pos/x40/1536/ced-noloop/longdocs/last128}{0.9542}
\defval{pos/x40/1536/ced-noloop/longdocs/last256}{0.9537}
\defval{pos/x40/1536/ced-noloop/longdocs/last512}{0.9555}
\defval{pos/x40/1536/ced-noloop/longdocs/mean}{0.9888}
\defval{pos/x40/1536/ced-noloop/pg19-test/c1-128}{1.4435}
\defval{pos/x40/1536/ced-noloop/pg19-test/c1024-2048}{1.2257}
\defval{pos/x40/1536/ced-noloop/pg19-test/c128-256}{1.2895}
\defval{pos/x40/1536/ced-noloop/pg19-test/c256-512}{1.2549}
\defval{pos/x40/1536/ced-noloop/pg19-test/c512-1024}{1.2345}
\defval{pos/x40/1536/ced-noloop/pg19-test/last1024}{1.2257}
\defval{pos/x40/1536/ced-noloop/pg19-test/last128}{1.2252}
\defval{pos/x40/1536/ced-noloop/pg19-test/last256}{1.2245}
\defval{pos/x40/1536/ced-noloop/pg19-test/last512}{1.2250}
\defval{pos/x40/1536/ced-noloop/pg19-test/mean}{1.2491}
\defval{pos/x40/1536/ced-noloop/pg19-val/c1-128}{1.4031}
\defval{pos/x40/1536/ced-noloop/pg19-val/c1024-2048}{1.1893}
\defval{pos/x40/1536/ced-noloop/pg19-val/c128-256}{1.2492}
\defval{pos/x40/1536/ced-noloop/pg19-val/c256-512}{1.2206}
\defval{pos/x40/1536/ced-noloop/pg19-val/c512-1024}{1.1968}
\defval{pos/x40/1536/ced-noloop/pg19-val/last1024}{1.1893}
\defval{pos/x40/1536/ced-noloop/pg19-val/last128}{1.1796}
\defval{pos/x40/1536/ced-noloop/pg19-val/last256}{1.1790}
\defval{pos/x40/1536/ced-noloop/pg19-val/last512}{1.1829}
\defval{pos/x40/1536/ced-noloop/pg19-val/mean}{1.2121}
\defval{pos/x40/1536/dense-l25/finepdfs-long/c1-128}{1.3192}
\defval{pos/x40/1536/dense-l25/finepdfs-long/c1024-2048}{0.9699}
\defval{pos/x40/1536/dense-l25/finepdfs-long/c128-256}{1.0803}
\defval{pos/x40/1536/dense-l25/finepdfs-long/c256-512}{1.0337}
\defval{pos/x40/1536/dense-l25/finepdfs-long/c512-1024}{0.9926}
\defval{pos/x40/1536/dense-l25/finepdfs-long/last1024}{0.9699}
\defval{pos/x40/1536/dense-l25/finepdfs-long/last128}{0.9626}
\defval{pos/x40/1536/dense-l25/finepdfs-long/last256}{0.9613}
\defval{pos/x40/1536/dense-l25/finepdfs-long/last512}{0.9633}
\defval{pos/x40/1536/dense-l25/finepdfs-long/mean}{1.0126}
\defval{pos/x40/1536/dense-l25/fineweb-edu-100bt/c1-128}{0.8466}
\defval{pos/x40/1536/dense-l25/fineweb-edu-100bt/c1024-2048}{0.7981}
\defval{pos/x40/1536/dense-l25/fineweb-edu-100bt/c128-256}{0.7873}
\defval{pos/x40/1536/dense-l25/fineweb-edu-100bt/c256-512}{0.7794}
\defval{pos/x40/1536/dense-l25/fineweb-edu-100bt/c512-1024}{0.7810}
\defval{pos/x40/1536/dense-l25/fineweb-edu-100bt/last1024}{0.7874}
\defval{pos/x40/1536/dense-l25/fineweb-edu-100bt/last128}{0.7829}
\defval{pos/x40/1536/dense-l25/fineweb-edu-100bt/last256}{0.7849}
\defval{pos/x40/1536/dense-l25/fineweb-edu-100bt/last512}{0.7858}
\defval{pos/x40/1536/dense-l25/fineweb-edu-100bt/mean}{0.7963}
\defval{pos/x40/1536/dense-l25/govreport-test/c1-128}{0.8252}
\defval{pos/x40/1536/dense-l25/govreport-test/c1024-2048}{0.6178}
\defval{pos/x40/1536/dense-l25/govreport-test/c128-256}{0.6888}
\defval{pos/x40/1536/dense-l25/govreport-test/c256-512}{0.6568}
\defval{pos/x40/1536/dense-l25/govreport-test/c512-1024}{0.6340}
\defval{pos/x40/1536/dense-l25/govreport-test/last1024}{0.6178}
\defval{pos/x40/1536/dense-l25/govreport-test/last128}{0.6121}
\defval{pos/x40/1536/dense-l25/govreport-test/last256}{0.6121}
\defval{pos/x40/1536/dense-l25/govreport-test/last512}{0.6145}
\defval{pos/x40/1536/dense-l25/govreport-test/mean}{0.6441}
\defval{pos/x40/1536/dense-l25/longdocs/c1-128}{1.1819}
\defval{pos/x40/1536/dense-l25/longdocs/c1024-2048}{0.9195}
\defval{pos/x40/1536/dense-l25/longdocs/c128-256}{1.0031}
\defval{pos/x40/1536/dense-l25/longdocs/c256-512}{0.9648}
\defval{pos/x40/1536/dense-l25/longdocs/c512-1024}{0.9364}
\defval{pos/x40/1536/dense-l25/longdocs/last1024}{0.9195}
\defval{pos/x40/1536/dense-l25/longdocs/last128}{0.9146}
\defval{pos/x40/1536/dense-l25/longdocs/last256}{0.9140}
\defval{pos/x40/1536/dense-l25/longdocs/last512}{0.9157}
\defval{pos/x40/1536/dense-l25/longdocs/mean}{0.9510}
\defval{pos/x40/1536/dense-l25/pg19-test/c1-128}{1.4027}
\defval{pos/x40/1536/dense-l25/pg19-test/c1024-2048}{1.1730}
\defval{pos/x40/1536/dense-l25/pg19-test/c128-256}{1.2412}
\defval{pos/x40/1536/dense-l25/pg19-test/c256-512}{1.2050}
\defval{pos/x40/1536/dense-l25/pg19-test/c512-1024}{1.1829}
\defval{pos/x40/1536/dense-l25/pg19-test/last1024}{1.1730}
\defval{pos/x40/1536/dense-l25/pg19-test/last128}{1.1727}
\defval{pos/x40/1536/dense-l25/pg19-test/last256}{1.1718}
\defval{pos/x40/1536/dense-l25/pg19-test/last512}{1.1721}
\defval{pos/x40/1536/dense-l25/pg19-test/mean}{1.1980}
\defval{pos/x40/1536/dense-l25/pg19-val/c1-128}{1.3657}
\defval{pos/x40/1536/dense-l25/pg19-val/c1024-2048}{1.1402}
\defval{pos/x40/1536/dense-l25/pg19-val/c128-256}{1.2032}
\defval{pos/x40/1536/dense-l25/pg19-val/c256-512}{1.1731}
\defval{pos/x40/1536/dense-l25/pg19-val/c512-1024}{1.1476}
\defval{pos/x40/1536/dense-l25/pg19-val/last1024}{1.1402}
\defval{pos/x40/1536/dense-l25/pg19-val/last128}{1.1306}
\defval{pos/x40/1536/dense-l25/pg19-val/last256}{1.1302}
\defval{pos/x40/1536/dense-l25/pg19-val/last512}{1.1344}
\defval{pos/x40/1536/dense-l25/pg19-val/mean}{1.1642}
\defval{pos/x40/1536/dense-noloop/finepdfs-long/c1-128}{1.3572}
\defval{pos/x40/1536/dense-noloop/finepdfs-long/c1024-2048}{1.0151}
\defval{pos/x40/1536/dense-noloop/finepdfs-long/c128-256}{1.1231}
\defval{pos/x40/1536/dense-noloop/finepdfs-long/c256-512}{1.0776}
\defval{pos/x40/1536/dense-noloop/finepdfs-long/c512-1024}{1.0372}
\defval{pos/x40/1536/dense-noloop/finepdfs-long/last1024}{1.0151}
\defval{pos/x40/1536/dense-noloop/finepdfs-long/last128}{1.0076}
\defval{pos/x40/1536/dense-noloop/finepdfs-long/last256}{1.0067}
\defval{pos/x40/1536/dense-noloop/finepdfs-long/last512}{1.0086}
\defval{pos/x40/1536/dense-noloop/finepdfs-long/mean}{1.0569}
\defval{pos/x40/1536/dense-noloop/fineweb-edu-100bt/c1-128}{0.8740}
\defval{pos/x40/1536/dense-noloop/fineweb-edu-100bt/c1024-2048}{0.8314}
\defval{pos/x40/1536/dense-noloop/fineweb-edu-100bt/c128-256}{0.8157}
\defval{pos/x40/1536/dense-noloop/fineweb-edu-100bt/c256-512}{0.8092}
\defval{pos/x40/1536/dense-noloop/fineweb-edu-100bt/c512-1024}{0.8123}
\defval{pos/x40/1536/dense-noloop/fineweb-edu-100bt/last1024}{0.8182}
\defval{pos/x40/1536/dense-noloop/fineweb-edu-100bt/last128}{0.8144}
\defval{pos/x40/1536/dense-noloop/fineweb-edu-100bt/last256}{0.8164}
\defval{pos/x40/1536/dense-noloop/fineweb-edu-100bt/last512}{0.8170}
\defval{pos/x40/1536/dense-noloop/fineweb-edu-100bt/mean}{0.8265}
\defval{pos/x40/1536/dense-noloop/govreport-test/c1-128}{0.8478}
\defval{pos/x40/1536/dense-noloop/govreport-test/c1024-2048}{0.6461}
\defval{pos/x40/1536/dense-noloop/govreport-test/c128-256}{0.7141}
\defval{pos/x40/1536/dense-noloop/govreport-test/c256-512}{0.6840}
\defval{pos/x40/1536/dense-noloop/govreport-test/c512-1024}{0.6623}
\defval{pos/x40/1536/dense-noloop/govreport-test/last1024}{0.6461}
\defval{pos/x40/1536/dense-noloop/govreport-test/last128}{0.6405}
\defval{pos/x40/1536/dense-noloop/govreport-test/last256}{0.6404}
\defval{pos/x40/1536/dense-noloop/govreport-test/last512}{0.6428}
\defval{pos/x40/1536/dense-noloop/govreport-test/mean}{0.6717}
\defval{pos/x40/1536/dense-noloop/longdocs/c1-128}{1.2177}
\defval{pos/x40/1536/dense-noloop/longdocs/c1024-2048}{0.9625}
\defval{pos/x40/1536/dense-noloop/longdocs/c128-256}{1.0435}
\defval{pos/x40/1536/dense-noloop/longdocs/c256-512}{1.0063}
\defval{pos/x40/1536/dense-noloop/longdocs/c512-1024}{0.9788}
\defval{pos/x40/1536/dense-noloop/longdocs/last1024}{0.9625}
\defval{pos/x40/1536/dense-noloop/longdocs/last128}{0.9576}
\defval{pos/x40/1536/dense-noloop/longdocs/last256}{0.9572}
\defval{pos/x40/1536/dense-noloop/longdocs/last512}{0.9589}
\defval{pos/x40/1536/dense-noloop/longdocs/mean}{0.9931}
\defval{pos/x40/1536/dense-noloop/pg19-test/c1-128}{1.4499}
\defval{pos/x40/1536/dense-noloop/pg19-test/c1024-2048}{1.2287}
\defval{pos/x40/1536/dense-noloop/pg19-test/c128-256}{1.2942}
\defval{pos/x40/1536/dense-noloop/pg19-test/c256-512}{1.2586}
\defval{pos/x40/1536/dense-noloop/pg19-test/c512-1024}{1.2373}
\defval{pos/x40/1536/dense-noloop/pg19-test/last1024}{1.2287}
\defval{pos/x40/1536/dense-noloop/pg19-test/last128}{1.2288}
\defval{pos/x40/1536/dense-noloop/pg19-test/last256}{1.2279}
\defval{pos/x40/1536/dense-noloop/pg19-test/last512}{1.2282}
\defval{pos/x40/1536/dense-noloop/pg19-test/mean}{1.2524}
\defval{pos/x40/1536/dense-noloop/pg19-val/c1-128}{1.4088}
\defval{pos/x40/1536/dense-noloop/pg19-val/c1024-2048}{1.1941}
\defval{pos/x40/1536/dense-noloop/pg19-val/c128-256}{1.2528}
\defval{pos/x40/1536/dense-noloop/pg19-val/c256-512}{1.2237}
\defval{pos/x40/1536/dense-noloop/pg19-val/c512-1024}{1.2014}
\defval{pos/x40/1536/dense-noloop/pg19-val/last1024}{1.1941}
\defval{pos/x40/1536/dense-noloop/pg19-val/last128}{1.1865}
\defval{pos/x40/1536/dense-noloop/pg19-val/last256}{1.1849}
\defval{pos/x40/1536/dense-noloop/pg19-val/last512}{1.1884}
\defval{pos/x40/1536/dense-noloop/pg19-val/mean}{1.2167}
\defval{pos/x40/768/ced-dec/finepdfs-long/c1-128}{1.4527}
\defval{pos/x40/768/ced-dec/finepdfs-long/c1024-2048}{1.1240}
\defval{pos/x40/768/ced-dec/finepdfs-long/c128-256}{1.2242}
\defval{pos/x40/768/ced-dec/finepdfs-long/c256-512}{1.1818}
\defval{pos/x40/768/ced-dec/finepdfs-long/c512-1024}{1.1433}
\defval{pos/x40/768/ced-dec/finepdfs-long/last1024}{1.1240}
\defval{pos/x40/768/ced-dec/finepdfs-long/last128}{1.1177}
\defval{pos/x40/768/ced-dec/finepdfs-long/last256}{1.1170}
\defval{pos/x40/768/ced-dec/finepdfs-long/last512}{1.1184}
\defval{pos/x40/768/ced-dec/finepdfs-long/mean}{1.1632}
\defval{pos/x40/768/ced-dec/fineweb-edu-100bt/c1-128}{0.9503}
\defval{pos/x40/768/ced-dec/fineweb-edu-100bt/c1024-2048}{0.9218}
\defval{pos/x40/768/ced-dec/fineweb-edu-100bt/c128-256}{0.8984}
\defval{pos/x40/768/ced-dec/fineweb-edu-100bt/c256-512}{0.8933}
\defval{pos/x40/768/ced-dec/fineweb-edu-100bt/c512-1024}{0.8976}
\defval{pos/x40/768/ced-dec/fineweb-edu-100bt/last1024}{0.9039}
\defval{pos/x40/768/ced-dec/fineweb-edu-100bt/last128}{0.9010}
\defval{pos/x40/768/ced-dec/fineweb-edu-100bt/last256}{0.9023}
\defval{pos/x40/768/ced-dec/fineweb-edu-100bt/last512}{0.9023}
\defval{pos/x40/768/ced-dec/fineweb-edu-100bt/mean}{0.9108}
\defval{pos/x40/768/ced-dec/govreport-test/c1-128}{0.9137}
\defval{pos/x40/768/ced-dec/govreport-test/c1024-2048}{0.7212}
\defval{pos/x40/768/ced-dec/govreport-test/c128-256}{0.7853}
\defval{pos/x40/768/ced-dec/govreport-test/c256-512}{0.7565}
\defval{pos/x40/768/ced-dec/govreport-test/c512-1024}{0.7366}
\defval{pos/x40/768/ced-dec/govreport-test/last1024}{0.7212}
\defval{pos/x40/768/ced-dec/govreport-test/last128}{0.7159}
\defval{pos/x40/768/ced-dec/govreport-test/last256}{0.7161}
\defval{pos/x40/768/ced-dec/govreport-test/last512}{0.7183}
\defval{pos/x40/768/ced-dec/govreport-test/mean}{0.7455}
\defval{pos/x40/768/ced-dec/longdocs/c1-128}{1.3070}
\defval{pos/x40/768/ced-dec/longdocs/c1024-2048}{1.0664}
\defval{pos/x40/768/ced-dec/longdocs/c128-256}{1.1417}
\defval{pos/x40/768/ced-dec/longdocs/c256-512}{1.1073}
\defval{pos/x40/768/ced-dec/longdocs/c512-1024}{1.0813}
\defval{pos/x40/768/ced-dec/longdocs/last1024}{1.0664}
\defval{pos/x40/768/ced-dec/longdocs/last128}{1.0621}
\defval{pos/x40/768/ced-dec/longdocs/last256}{1.0619}
\defval{pos/x40/768/ced-dec/longdocs/last512}{1.0633}
\defval{pos/x40/768/ced-dec/longdocs/mean}{1.0950}
\defval{pos/x40/768/ced-dec/pg19-test/c1-128}{1.5563}
\defval{pos/x40/768/ced-dec/pg19-test/c1024-2048}{1.3564}
\defval{pos/x40/768/ced-dec/pg19-test/c128-256}{1.4167}
\defval{pos/x40/768/ced-dec/pg19-test/c256-512}{1.3848}
\defval{pos/x40/768/ced-dec/pg19-test/c512-1024}{1.3643}
\defval{pos/x40/768/ced-dec/pg19-test/last1024}{1.3564}
\defval{pos/x40/768/ced-dec/pg19-test/last128}{1.3570}
\defval{pos/x40/768/ced-dec/pg19-test/last256}{1.3564}
\defval{pos/x40/768/ced-dec/pg19-test/last512}{1.3563}
\defval{pos/x40/768/ced-dec/pg19-test/mean}{1.3782}
\defval{pos/x40/768/ced-dec/pg19-val/c1-128}{1.5135}
\defval{pos/x40/768/ced-dec/pg19-val/c1024-2048}{1.3183}
\defval{pos/x40/768/ced-dec/pg19-val/c128-256}{1.3732}
\defval{pos/x40/768/ced-dec/pg19-val/c256-512}{1.3462}
\defval{pos/x40/768/ced-dec/pg19-val/c512-1024}{1.3253}
\defval{pos/x40/768/ced-dec/pg19-val/last1024}{1.3183}
\defval{pos/x40/768/ced-dec/pg19-val/last128}{1.3111}
\defval{pos/x40/768/ced-dec/pg19-val/last256}{1.3104}
\defval{pos/x40/768/ced-dec/pg19-val/last512}{1.3127}
\defval{pos/x40/768/ced-dec/pg19-val/mean}{1.3391}
\defval{pos/x40/768/ced-enc/finepdfs-long/c1-128}{1.4322}
\defval{pos/x40/768/ced-enc/finepdfs-long/c1024-2048}{1.0898}
\defval{pos/x40/768/ced-enc/finepdfs-long/c128-256}{1.1990}
\defval{pos/x40/768/ced-enc/finepdfs-long/c256-512}{1.1539}
\defval{pos/x40/768/ced-enc/finepdfs-long/c512-1024}{1.1123}
\defval{pos/x40/768/ced-enc/finepdfs-long/last1024}{1.0898}
\defval{pos/x40/768/ced-enc/finepdfs-long/last128}{1.0830}
\defval{pos/x40/768/ced-enc/finepdfs-long/last256}{1.0815}
\defval{pos/x40/768/ced-enc/finepdfs-long/last512}{1.0832}
\defval{pos/x40/768/ced-enc/finepdfs-long/mean}{1.1320}
\defval{pos/x40/768/ced-enc/fineweb-edu-100bt/c1-128}{0.9343}
\defval{pos/x40/768/ced-enc/fineweb-edu-100bt/c1024-2048}{0.9011}
\defval{pos/x40/768/ced-enc/fineweb-edu-100bt/c128-256}{0.8812}
\defval{pos/x40/768/ced-enc/fineweb-edu-100bt/c256-512}{0.8746}
\defval{pos/x40/768/ced-enc/fineweb-edu-100bt/c512-1024}{0.8783}
\defval{pos/x40/768/ced-enc/fineweb-edu-100bt/last1024}{0.8839}
\defval{pos/x40/768/ced-enc/fineweb-edu-100bt/last128}{0.8799}
\defval{pos/x40/768/ced-enc/fineweb-edu-100bt/last256}{0.8826}
\defval{pos/x40/768/ced-enc/fineweb-edu-100bt/last512}{0.8825}
\defval{pos/x40/768/ced-enc/fineweb-edu-100bt/mean}{0.8922}
\defval{pos/x40/768/ced-enc/govreport-test/c1-128}{0.8996}
\defval{pos/x40/768/ced-enc/govreport-test/c1024-2048}{0.7009}
\defval{pos/x40/768/ced-enc/govreport-test/c128-256}{0.7676}
\defval{pos/x40/768/ced-enc/govreport-test/c256-512}{0.7382}
\defval{pos/x40/768/ced-enc/govreport-test/c512-1024}{0.7173}
\defval{pos/x40/768/ced-enc/govreport-test/last1024}{0.7009}
\defval{pos/x40/768/ced-enc/govreport-test/last128}{0.6946}
\defval{pos/x40/768/ced-enc/govreport-test/last256}{0.6951}
\defval{pos/x40/768/ced-enc/govreport-test/last512}{0.6977}
\defval{pos/x40/768/ced-enc/govreport-test/mean}{0.7262}
\defval{pos/x40/768/ced-enc/longdocs/c1-128}{1.2872}
\defval{pos/x40/768/ced-enc/longdocs/c1024-2048}{1.0395}
\defval{pos/x40/768/ced-enc/longdocs/c128-256}{1.1183}
\defval{pos/x40/768/ced-enc/longdocs/c256-512}{1.0830}
\defval{pos/x40/768/ced-enc/longdocs/c512-1024}{1.0560}
\defval{pos/x40/768/ced-enc/longdocs/last1024}{1.0395}
\defval{pos/x40/768/ced-enc/longdocs/last128}{1.0345}
\defval{pos/x40/768/ced-enc/longdocs/last256}{1.0342}
\defval{pos/x40/768/ced-enc/longdocs/last512}{1.0359}
\defval{pos/x40/768/ced-enc/longdocs/mean}{1.0695}
\defval{pos/x40/768/ced-enc/pg19-test/c1-128}{1.5316}
\defval{pos/x40/768/ced-enc/pg19-test/c1024-2048}{1.3302}
\defval{pos/x40/768/ced-enc/pg19-test/c128-256}{1.3892}
\defval{pos/x40/768/ced-enc/pg19-test/c256-512}{1.3580}
\defval{pos/x40/768/ced-enc/pg19-test/c512-1024}{1.3387}
\defval{pos/x40/768/ced-enc/pg19-test/last1024}{1.3302}
\defval{pos/x40/768/ced-enc/pg19-test/last128}{1.3299}
\defval{pos/x40/768/ced-enc/pg19-test/last256}{1.3296}
\defval{pos/x40/768/ced-enc/pg19-test/last512}{1.3298}
\defval{pos/x40/768/ced-enc/pg19-test/mean}{1.3520}
\defval{pos/x40/768/ced-enc/pg19-val/c1-128}{1.4901}
\defval{pos/x40/768/ced-enc/pg19-val/c1024-2048}{1.2951}
\defval{pos/x40/768/ced-enc/pg19-val/c128-256}{1.3491}
\defval{pos/x40/768/ced-enc/pg19-val/c256-512}{1.3223}
\defval{pos/x40/768/ced-enc/pg19-val/c512-1024}{1.3001}
\defval{pos/x40/768/ced-enc/pg19-val/last1024}{1.2951}
\defval{pos/x40/768/ced-enc/pg19-val/last128}{1.2879}
\defval{pos/x40/768/ced-enc/pg19-val/last256}{1.2852}
\defval{pos/x40/768/ced-enc/pg19-val/last512}{1.2889}
\defval{pos/x40/768/ced-enc/pg19-val/mean}{1.3153}
\defval{pos/x40/768/ced-noloop/finepdfs-long/c1-128}{1.4590}
\defval{pos/x40/768/ced-noloop/finepdfs-long/c1024-2048}{1.1290}
\defval{pos/x40/768/ced-noloop/finepdfs-long/c128-256}{1.2319}
\defval{pos/x40/768/ced-noloop/finepdfs-long/c256-512}{1.1884}
\defval{pos/x40/768/ced-noloop/finepdfs-long/c512-1024}{1.1493}
\defval{pos/x40/768/ced-noloop/finepdfs-long/last1024}{1.1290}
\defval{pos/x40/768/ced-noloop/finepdfs-long/last128}{1.1234}
\defval{pos/x40/768/ced-noloop/finepdfs-long/last256}{1.1217}
\defval{pos/x40/768/ced-noloop/finepdfs-long/last512}{1.1231}
\defval{pos/x40/768/ced-noloop/finepdfs-long/mean}{1.1688}
\defval{pos/x40/768/ced-noloop/fineweb-edu-100bt/c1-128}{0.9543}
\defval{pos/x40/768/ced-noloop/fineweb-edu-100bt/c1024-2048}{0.9247}
\defval{pos/x40/768/ced-noloop/fineweb-edu-100bt/c128-256}{0.9024}
\defval{pos/x40/768/ced-noloop/fineweb-edu-100bt/c256-512}{0.8970}
\defval{pos/x40/768/ced-noloop/fineweb-edu-100bt/c512-1024}{0.9006}
\defval{pos/x40/768/ced-noloop/fineweb-edu-100bt/last1024}{0.9072}
\defval{pos/x40/768/ced-noloop/fineweb-edu-100bt/last128}{0.9047}
\defval{pos/x40/768/ced-noloop/fineweb-edu-100bt/last256}{0.9057}
\defval{pos/x40/768/ced-noloop/fineweb-edu-100bt/last512}{0.9060}
\defval{pos/x40/768/ced-noloop/fineweb-edu-100bt/mean}{0.9142}
\defval{pos/x40/768/ced-noloop/govreport-test/c1-128}{0.9156}
\defval{pos/x40/768/ced-noloop/govreport-test/c1024-2048}{0.7237}
\defval{pos/x40/768/ced-noloop/govreport-test/c128-256}{0.7870}
\defval{pos/x40/768/ced-noloop/govreport-test/c256-512}{0.7586}
\defval{pos/x40/768/ced-noloop/govreport-test/c512-1024}{0.7387}
\defval{pos/x40/768/ced-noloop/govreport-test/last1024}{0.7237}
\defval{pos/x40/768/ced-noloop/govreport-test/last128}{0.7181}
\defval{pos/x40/768/ced-noloop/govreport-test/last256}{0.7182}
\defval{pos/x40/768/ced-noloop/govreport-test/last512}{0.7207}
\defval{pos/x40/768/ced-noloop/govreport-test/mean}{0.7477}
\defval{pos/x40/768/ced-noloop/longdocs/c1-128}{1.3131}
\defval{pos/x40/768/ced-noloop/longdocs/c1024-2048}{1.0725}
\defval{pos/x40/768/ced-noloop/longdocs/c128-256}{1.1485}
\defval{pos/x40/768/ced-noloop/longdocs/c256-512}{1.1139}
\defval{pos/x40/768/ced-noloop/longdocs/c512-1024}{1.0878}
\defval{pos/x40/768/ced-noloop/longdocs/last1024}{1.0725}
\defval{pos/x40/768/ced-noloop/longdocs/last128}{1.0681}
\defval{pos/x40/768/ced-noloop/longdocs/last256}{1.0676}
\defval{pos/x40/768/ced-noloop/longdocs/last512}{1.0692}
\defval{pos/x40/768/ced-noloop/longdocs/mean}{1.1013}
\defval{pos/x40/768/ced-noloop/pg19-test/c1-128}{1.5664}
\defval{pos/x40/768/ced-noloop/pg19-test/c1024-2048}{1.3672}
\defval{pos/x40/768/ced-noloop/pg19-test/c128-256}{1.4277}
\defval{pos/x40/768/ced-noloop/pg19-test/c256-512}{1.3960}
\defval{pos/x40/768/ced-noloop/pg19-test/c512-1024}{1.3756}
\defval{pos/x40/768/ced-noloop/pg19-test/last1024}{1.3672}
\defval{pos/x40/768/ced-noloop/pg19-test/last128}{1.3671}
\defval{pos/x40/768/ced-noloop/pg19-test/last256}{1.3668}
\defval{pos/x40/768/ced-noloop/pg19-test/last512}{1.3669}
\defval{pos/x40/768/ced-noloop/pg19-test/mean}{1.3891}
\defval{pos/x40/768/ced-noloop/pg19-val/c1-128}{1.5235}
\defval{pos/x40/768/ced-noloop/pg19-val/c1024-2048}{1.3278}
\defval{pos/x40/768/ced-noloop/pg19-val/c128-256}{1.3808}
\defval{pos/x40/768/ced-noloop/pg19-val/c256-512}{1.3556}
\defval{pos/x40/768/ced-noloop/pg19-val/c512-1024}{1.3356}
\defval{pos/x40/768/ced-noloop/pg19-val/last1024}{1.3278}
\defval{pos/x40/768/ced-noloop/pg19-val/last128}{1.3195}
\defval{pos/x40/768/ced-noloop/pg19-val/last256}{1.3182}
\defval{pos/x40/768/ced-noloop/pg19-val/last512}{1.3214}
\defval{pos/x40/768/ced-noloop/pg19-val/mean}{1.3487}
\defval{pos/x40/768/dense-l25/finepdfs-long/c1-128}{1.4440}
\defval{pos/x40/768/dense-l25/finepdfs-long/c1024-2048}{1.0959}
\defval{pos/x40/768/dense-l25/finepdfs-long/c128-256}{1.2074}
\defval{pos/x40/768/dense-l25/finepdfs-long/c256-512}{1.1617}
\defval{pos/x40/768/dense-l25/finepdfs-long/c512-1024}{1.1188}
\defval{pos/x40/768/dense-l25/finepdfs-long/last1024}{1.0959}
\defval{pos/x40/768/dense-l25/finepdfs-long/last128}{1.0892}
\defval{pos/x40/768/dense-l25/finepdfs-long/last256}{1.0880}
\defval{pos/x40/768/dense-l25/finepdfs-long/last512}{1.0895}
\defval{pos/x40/768/dense-l25/finepdfs-long/mean}{1.1389}
\defval{pos/x40/768/dense-l25/fineweb-edu-100bt/c1-128}{0.9367}
\defval{pos/x40/768/dense-l25/fineweb-edu-100bt/c1024-2048}{0.8987}
\defval{pos/x40/768/dense-l25/fineweb-edu-100bt/c128-256}{0.8817}
\defval{pos/x40/768/dense-l25/fineweb-edu-100bt/c256-512}{0.8742}
\defval{pos/x40/768/dense-l25/fineweb-edu-100bt/c512-1024}{0.8776}
\defval{pos/x40/768/dense-l25/fineweb-edu-100bt/last1024}{0.8836}
\defval{pos/x40/768/dense-l25/fineweb-edu-100bt/last128}{0.8792}
\defval{pos/x40/768/dense-l25/fineweb-edu-100bt/last256}{0.8819}
\defval{pos/x40/768/dense-l25/fineweb-edu-100bt/last512}{0.8818}
\defval{pos/x40/768/dense-l25/fineweb-edu-100bt/mean}{0.8919}
\defval{pos/x40/768/dense-l25/govreport-test/c1-128}{0.9010}
\defval{pos/x40/768/dense-l25/govreport-test/c1024-2048}{0.6975}
\defval{pos/x40/768/dense-l25/govreport-test/c128-256}{0.7673}
\defval{pos/x40/768/dense-l25/govreport-test/c256-512}{0.7366}
\defval{pos/x40/768/dense-l25/govreport-test/c512-1024}{0.7144}
\defval{pos/x40/768/dense-l25/govreport-test/last1024}{0.6975}
\defval{pos/x40/768/dense-l25/govreport-test/last128}{0.6917}
\defval{pos/x40/768/dense-l25/govreport-test/last256}{0.6920}
\defval{pos/x40/768/dense-l25/govreport-test/last512}{0.6943}
\defval{pos/x40/768/dense-l25/govreport-test/mean}{0.7236}
\defval{pos/x40/768/dense-l25/longdocs/c1-128}{1.2960}
\defval{pos/x40/768/dense-l25/longdocs/c1024-2048}{1.0406}
\defval{pos/x40/768/dense-l25/longdocs/c128-256}{1.1233}
\defval{pos/x40/768/dense-l25/longdocs/c256-512}{1.0865}
\defval{pos/x40/768/dense-l25/longdocs/c512-1024}{1.0578}
\defval{pos/x40/768/dense-l25/longdocs/last1024}{1.0406}
\defval{pos/x40/768/dense-l25/longdocs/last128}{1.0357}
\defval{pos/x40/768/dense-l25/longdocs/last256}{1.0354}
\defval{pos/x40/768/dense-l25/longdocs/last512}{1.0370}
\defval{pos/x40/768/dense-l25/longdocs/mean}{1.0718}
\defval{pos/x40/768/dense-l25/pg19-test/c1-128}{1.5448}
\defval{pos/x40/768/dense-l25/pg19-test/c1024-2048}{1.3308}
\defval{pos/x40/768/dense-l25/pg19-test/c128-256}{1.3962}
\defval{pos/x40/768/dense-l25/pg19-test/c256-512}{1.3626}
\defval{pos/x40/768/dense-l25/pg19-test/c512-1024}{1.3406}
\defval{pos/x40/768/dense-l25/pg19-test/last1024}{1.3308}
\defval{pos/x40/768/dense-l25/pg19-test/last128}{1.3303}
\defval{pos/x40/768/dense-l25/pg19-test/last256}{1.3300}
\defval{pos/x40/768/dense-l25/pg19-test/last512}{1.3302}
\defval{pos/x40/768/dense-l25/pg19-test/mean}{1.3546}
\defval{pos/x40/768/dense-l25/pg19-val/c1-128}{1.4995}
\defval{pos/x40/768/dense-l25/pg19-val/c1024-2048}{1.2909}
\defval{pos/x40/768/dense-l25/pg19-val/c128-256}{1.3516}
\defval{pos/x40/768/dense-l25/pg19-val/c256-512}{1.3214}
\defval{pos/x40/768/dense-l25/pg19-val/c512-1024}{1.2990}
\defval{pos/x40/768/dense-l25/pg19-val/last1024}{1.2909}
\defval{pos/x40/768/dense-l25/pg19-val/last128}{1.2833}
\defval{pos/x40/768/dense-l25/pg19-val/last256}{1.2815}
\defval{pos/x40/768/dense-l25/pg19-val/last512}{1.2847}
\defval{pos/x40/768/dense-l25/pg19-val/mean}{1.3135}
\defval{pos/x40/768/dense-l47/finepdfs-long/c1-128}{1.4459}
\defval{pos/x40/768/dense-l47/finepdfs-long/c1024-2048}{1.1049}
\defval{pos/x40/768/dense-l47/finepdfs-long/c128-256}{1.2123}
\defval{pos/x40/768/dense-l47/finepdfs-long/c256-512}{1.1670}
\defval{pos/x40/768/dense-l47/finepdfs-long/c512-1024}{1.1260}
\defval{pos/x40/768/dense-l47/finepdfs-long/last1024}{1.1049}
\defval{pos/x40/768/dense-l47/finepdfs-long/last128}{1.0992}
\defval{pos/x40/768/dense-l47/finepdfs-long/last256}{1.0974}
\defval{pos/x40/768/dense-l47/finepdfs-long/last512}{1.0988}
\defval{pos/x40/768/dense-l47/finepdfs-long/mean}{1.1463}
\defval{pos/x40/768/dense-l47/fineweb-edu-100bt/c1-128}{0.9411}
\defval{pos/x40/768/dense-l47/fineweb-edu-100bt/c1024-2048}{0.9069}
\defval{pos/x40/768/dense-l47/fineweb-edu-100bt/c128-256}{0.8858}
\defval{pos/x40/768/dense-l47/fineweb-edu-100bt/c256-512}{0.8801}
\defval{pos/x40/768/dense-l47/fineweb-edu-100bt/c512-1024}{0.8830}
\defval{pos/x40/768/dense-l47/fineweb-edu-100bt/last1024}{0.8902}
\defval{pos/x40/768/dense-l47/fineweb-edu-100bt/last128}{0.8868}
\defval{pos/x40/768/dense-l47/fineweb-edu-100bt/last256}{0.8885}
\defval{pos/x40/768/dense-l47/fineweb-edu-100bt/last512}{0.8885}
\defval{pos/x40/768/dense-l47/fineweb-edu-100bt/mean}{0.8976}
\defval{pos/x40/768/dense-l47/govreport-test/c1-128}{0.9068}
\defval{pos/x40/768/dense-l47/govreport-test/c1024-2048}{0.7047}
\defval{pos/x40/768/dense-l47/govreport-test/c128-256}{0.7728}
\defval{pos/x40/768/dense-l47/govreport-test/c256-512}{0.7426}
\defval{pos/x40/768/dense-l47/govreport-test/c512-1024}{0.7210}
\defval{pos/x40/768/dense-l47/govreport-test/last1024}{0.7047}
\defval{pos/x40/768/dense-l47/govreport-test/last128}{0.6988}
\defval{pos/x40/768/dense-l47/govreport-test/last256}{0.6992}
\defval{pos/x40/768/dense-l47/govreport-test/last512}{0.7016}
\defval{pos/x40/768/dense-l47/govreport-test/mean}{0.7303}
\defval{pos/x40/768/dense-l47/longdocs/c1-128}{1.2997}
\defval{pos/x40/768/dense-l47/longdocs/c1024-2048}{1.0471}
\defval{pos/x40/768/dense-l47/longdocs/c128-256}{1.1282}
\defval{pos/x40/768/dense-l47/longdocs/c256-512}{1.0913}
\defval{pos/x40/768/dense-l47/longdocs/c512-1024}{1.0632}
\defval{pos/x40/768/dense-l47/longdocs/last1024}{1.0471}
\defval{pos/x40/768/dense-l47/longdocs/last128}{1.0427}
\defval{pos/x40/768/dense-l47/longdocs/last256}{1.0422}
\defval{pos/x40/768/dense-l47/longdocs/last512}{1.0437}
\defval{pos/x40/768/dense-l47/longdocs/mean}{1.0775}
\defval{pos/x40/768/dense-l47/pg19-test/c1-128}{1.5481}
\defval{pos/x40/768/dense-l47/pg19-test/c1024-2048}{1.3341}
\defval{pos/x40/768/dense-l47/pg19-test/c128-256}{1.4005}
\defval{pos/x40/768/dense-l47/pg19-test/c256-512}{1.3655}
\defval{pos/x40/768/dense-l47/pg19-test/c512-1024}{1.3429}
\defval{pos/x40/768/dense-l47/pg19-test/last1024}{1.3341}
\defval{pos/x40/768/dense-l47/pg19-test/last128}{1.3344}
\defval{pos/x40/768/dense-l47/pg19-test/last256}{1.3338}
\defval{pos/x40/768/dense-l47/pg19-test/last512}{1.3339}
\defval{pos/x40/768/dense-l47/pg19-test/mean}{1.3577}
\defval{pos/x40/768/dense-l47/pg19-val/c1-128}{1.5038}
\defval{pos/x40/768/dense-l47/pg19-val/c1024-2048}{1.2957}
\defval{pos/x40/768/dense-l47/pg19-val/c128-256}{1.3568}
\defval{pos/x40/768/dense-l47/pg19-val/c256-512}{1.3264}
\defval{pos/x40/768/dense-l47/pg19-val/c512-1024}{1.3034}
\defval{pos/x40/768/dense-l47/pg19-val/last1024}{1.2957}
\defval{pos/x40/768/dense-l47/pg19-val/last128}{1.2887}
\defval{pos/x40/768/dense-l47/pg19-val/last256}{1.2870}
\defval{pos/x40/768/dense-l47/pg19-val/last512}{1.2895}
\defval{pos/x40/768/dense-l47/pg19-val/mean}{1.3182}
\defval{pos/x40/768/dense-noloop/finepdfs-long/c1-128}{1.4704}
\defval{pos/x40/768/dense-noloop/finepdfs-long/c1024-2048}{1.1347}
\defval{pos/x40/768/dense-noloop/finepdfs-long/c128-256}{1.2402}
\defval{pos/x40/768/dense-noloop/finepdfs-long/c256-512}{1.1962}
\defval{pos/x40/768/dense-noloop/finepdfs-long/c512-1024}{1.1560}
\defval{pos/x40/768/dense-noloop/finepdfs-long/last1024}{1.1347}
\defval{pos/x40/768/dense-noloop/finepdfs-long/last128}{1.1282}
\defval{pos/x40/768/dense-noloop/finepdfs-long/last256}{1.1266}
\defval{pos/x40/768/dense-noloop/finepdfs-long/last512}{1.1284}
\defval{pos/x40/768/dense-noloop/finepdfs-long/mean}{1.1756}
\defval{pos/x40/768/dense-noloop/fineweb-edu-100bt/c1-128}{0.9593}
\defval{pos/x40/768/dense-noloop/fineweb-edu-100bt/c1024-2048}{0.9258}
\defval{pos/x40/768/dense-noloop/fineweb-edu-100bt/c128-256}{0.9059}
\defval{pos/x40/768/dense-noloop/fineweb-edu-100bt/c256-512}{0.8982}
\defval{pos/x40/768/dense-noloop/fineweb-edu-100bt/c512-1024}{0.9017}
\defval{pos/x40/768/dense-noloop/fineweb-edu-100bt/last1024}{0.9087}
\defval{pos/x40/768/dense-noloop/fineweb-edu-100bt/last128}{0.9051}
\defval{pos/x40/768/dense-noloop/fineweb-edu-100bt/last256}{0.9069}
\defval{pos/x40/768/dense-noloop/fineweb-edu-100bt/last512}{0.9071}
\defval{pos/x40/768/dense-noloop/fineweb-edu-100bt/mean}{0.9163}
\defval{pos/x40/768/dense-noloop/govreport-test/c1-128}{0.9203}
\defval{pos/x40/768/dense-noloop/govreport-test/c1024-2048}{0.7214}
\defval{pos/x40/768/dense-noloop/govreport-test/c128-256}{0.7895}
\defval{pos/x40/768/dense-noloop/govreport-test/c256-512}{0.7595}
\defval{pos/x40/768/dense-noloop/govreport-test/c512-1024}{0.7379}
\defval{pos/x40/768/dense-noloop/govreport-test/last1024}{0.7214}
\defval{pos/x40/768/dense-noloop/govreport-test/last128}{0.7152}
\defval{pos/x40/768/dense-noloop/govreport-test/last256}{0.7157}
\defval{pos/x40/768/dense-noloop/govreport-test/last512}{0.7181}
\defval{pos/x40/768/dense-noloop/govreport-test/mean}{0.7469}
\defval{pos/x40/768/dense-noloop/longdocs/c1-128}{1.3201}
\defval{pos/x40/768/dense-noloop/longdocs/c1024-2048}{1.0745}
\defval{pos/x40/768/dense-noloop/longdocs/c128-256}{1.1531}
\defval{pos/x40/768/dense-noloop/longdocs/c256-512}{1.1179}
\defval{pos/x40/768/dense-noloop/longdocs/c512-1024}{1.0908}
\defval{pos/x40/768/dense-noloop/longdocs/last1024}{1.0745}
\defval{pos/x40/768/dense-noloop/longdocs/last128}{1.0696}
\defval{pos/x40/768/dense-noloop/longdocs/last256}{1.0693}
\defval{pos/x40/768/dense-noloop/longdocs/last512}{1.0710}
\defval{pos/x40/768/dense-noloop/longdocs/mean}{1.1043}
\defval{pos/x40/768/dense-noloop/pg19-test/c1-128}{1.5714}
\defval{pos/x40/768/dense-noloop/pg19-test/c1024-2048}{1.3702}
\defval{pos/x40/768/dense-noloop/pg19-test/c128-256}{1.4308}
\defval{pos/x40/768/dense-noloop/pg19-test/c256-512}{1.3991}
\defval{pos/x40/768/dense-noloop/pg19-test/c512-1024}{1.3789}
\defval{pos/x40/768/dense-noloop/pg19-test/last1024}{1.3702}
\defval{pos/x40/768/dense-noloop/pg19-test/last128}{1.3698}
\defval{pos/x40/768/dense-noloop/pg19-test/last256}{1.3695}
\defval{pos/x40/768/dense-noloop/pg19-test/last512}{1.3698}
\defval{pos/x40/768/dense-noloop/pg19-test/mean}{1.3923}
\defval{pos/x40/768/dense-noloop/pg19-val/c1-128}{1.5268}
\defval{pos/x40/768/dense-noloop/pg19-val/c1024-2048}{1.3313}
\defval{pos/x40/768/dense-noloop/pg19-val/c128-256}{1.3851}
\defval{pos/x40/768/dense-noloop/pg19-val/c256-512}{1.3596}
\defval{pos/x40/768/dense-noloop/pg19-val/c512-1024}{1.3381}
\defval{pos/x40/768/dense-noloop/pg19-val/last1024}{1.3313}
\defval{pos/x40/768/dense-noloop/pg19-val/last128}{1.3243}
\defval{pos/x40/768/dense-noloop/pg19-val/last256}{1.3224}
\defval{pos/x40/768/dense-noloop/pg19-val/last512}{1.3249}
\defval{pos/x40/768/dense-noloop/pg19-val/mean}{1.3521}
\defval{pos/x40/768/yocou-loop/finepdfs-long/c1-128}{1.4464}
\defval{pos/x40/768/yocou-loop/finepdfs-long/c1024-2048}{1.1073}
\defval{pos/x40/768/yocou-loop/finepdfs-long/c128-256}{1.2088}
\defval{pos/x40/768/yocou-loop/finepdfs-long/c256-512}{1.1666}
\defval{pos/x40/768/yocou-loop/finepdfs-long/c512-1024}{1.1277}
\defval{pos/x40/768/yocou-loop/finepdfs-long/last1024}{1.1073}
\defval{pos/x40/768/yocou-loop/finepdfs-long/last128}{1.1015}
\defval{pos/x40/768/yocou-loop/finepdfs-long/last256}{1.0999}
\defval{pos/x40/768/yocou-loop/finepdfs-long/last512}{1.1012}
\defval{pos/x40/768/yocou-loop/finepdfs-long/mean}{1.1476}
\defval{pos/x40/768/yocou-loop/fineweb-edu-100bt/c1-128}{0.9348}
\defval{pos/x40/768/yocou-loop/fineweb-edu-100bt/c1024-2048}{0.9067}
\defval{pos/x40/768/yocou-loop/fineweb-edu-100bt/c128-256}{0.8814}
\defval{pos/x40/768/yocou-loop/fineweb-edu-100bt/c256-512}{0.8775}
\defval{pos/x40/768/yocou-loop/fineweb-edu-100bt/c512-1024}{0.8831}
\defval{pos/x40/768/yocou-loop/fineweb-edu-100bt/last1024}{0.8882}
\defval{pos/x40/768/yocou-loop/fineweb-edu-100bt/last128}{0.8845}
\defval{pos/x40/768/yocou-loop/fineweb-edu-100bt/last256}{0.8858}
\defval{pos/x40/768/yocou-loop/fineweb-edu-100bt/last512}{0.8867}
\defval{pos/x40/768/yocou-loop/fineweb-edu-100bt/mean}{0.8953}
\defval{pos/x40/768/yocou-loop/govreport-test/c1-128}{0.9007}
\defval{pos/x40/768/yocou-loop/govreport-test/c1024-2048}{0.7093}
\defval{pos/x40/768/yocou-loop/govreport-test/c128-256}{0.7698}
\defval{pos/x40/768/yocou-loop/govreport-test/c256-512}{0.7426}
\defval{pos/x40/768/yocou-loop/govreport-test/c512-1024}{0.7239}
\defval{pos/x40/768/yocou-loop/govreport-test/last1024}{0.7093}
\defval{pos/x40/768/yocou-loop/govreport-test/last128}{0.7034}
\defval{pos/x40/768/yocou-loop/govreport-test/last256}{0.7040}
\defval{pos/x40/768/yocou-loop/govreport-test/last512}{0.7064}
\defval{pos/x40/768/yocou-loop/govreport-test/mean}{0.7328}
\defval{pos/x40/768/yocou-loop/longdocs/c1-128}{1.2930}
\defval{pos/x40/768/yocou-loop/longdocs/c1024-2048}{1.0500}
\defval{pos/x40/768/yocou-loop/longdocs/c128-256}{1.1226}
\defval{pos/x40/768/yocou-loop/longdocs/c256-512}{1.0898}
\defval{pos/x40/768/yocou-loop/longdocs/c512-1024}{1.0650}
\defval{pos/x40/768/yocou-loop/longdocs/last1024}{1.0500}
\defval{pos/x40/768/yocou-loop/longdocs/last128}{1.0455}
\defval{pos/x40/768/yocou-loop/longdocs/last256}{1.0451}
\defval{pos/x40/768/yocou-loop/longdocs/last512}{1.0466}
\defval{pos/x40/768/yocou-loop/longdocs/mean}{1.0785}
\defval{pos/x40/768/yocou-loop/pg19-test/c1-128}{1.5338}
\defval{pos/x40/768/yocou-loop/pg19-test/c1024-2048}{1.3358}
\defval{pos/x40/768/yocou-loop/pg19-test/c128-256}{1.3903}
\defval{pos/x40/768/yocou-loop/pg19-test/c256-512}{1.3615}
\defval{pos/x40/768/yocou-loop/pg19-test/c512-1024}{1.3438}
\defval{pos/x40/768/yocou-loop/pg19-test/last1024}{1.3358}
\defval{pos/x40/768/yocou-loop/pg19-test/last128}{1.3358}
\defval{pos/x40/768/yocou-loop/pg19-test/last256}{1.3351}
\defval{pos/x40/768/yocou-loop/pg19-test/last512}{1.3354}
\defval{pos/x40/768/yocou-loop/pg19-test/mean}{1.3567}
\defval{pos/x40/768/yocou-loop/pg19-val/c1-128}{1.4921}
\defval{pos/x40/768/yocou-loop/pg19-val/c1024-2048}{1.2993}
\defval{pos/x40/768/yocou-loop/pg19-val/c128-256}{1.3493}
\defval{pos/x40/768/yocou-loop/pg19-val/c256-512}{1.3250}
\defval{pos/x40/768/yocou-loop/pg19-val/c512-1024}{1.3048}
\defval{pos/x40/768/yocou-loop/pg19-val/last1024}{1.2993}
\defval{pos/x40/768/yocou-loop/pg19-val/last128}{1.2894}
\defval{pos/x40/768/yocou-loop/pg19-val/last256}{1.2891}
\defval{pos/x40/768/yocou-loop/pg19-val/last512}{1.2930}
\defval{pos/x40/768/yocou-loop/pg19-val/mean}{1.3190}
\defval{pos/x40/768/yocou-noloop/finepdfs-long/c1-128}{1.4618}
\defval{pos/x40/768/yocou-noloop/finepdfs-long/c1024-2048}{1.1374}
\defval{pos/x40/768/yocou-noloop/finepdfs-long/c128-256}{1.2349}
\defval{pos/x40/768/yocou-noloop/finepdfs-long/c256-512}{1.1934}
\defval{pos/x40/768/yocou-noloop/finepdfs-long/c512-1024}{1.1560}
\defval{pos/x40/768/yocou-noloop/finepdfs-long/last1024}{1.1374}
\defval{pos/x40/768/yocou-noloop/finepdfs-long/last128}{1.1313}
\defval{pos/x40/768/yocou-noloop/finepdfs-long/last256}{1.1300}
\defval{pos/x40/768/yocou-noloop/finepdfs-long/last512}{1.1316}
\defval{pos/x40/768/yocou-noloop/finepdfs-long/mean}{1.1757}
\defval{pos/x40/768/yocou-noloop/fineweb-edu-100bt/c1-128}{0.9559}
\defval{pos/x40/768/yocou-noloop/fineweb-edu-100bt/c1024-2048}{0.9288}
\defval{pos/x40/768/yocou-noloop/fineweb-edu-100bt/c128-256}{0.9045}
\defval{pos/x40/768/yocou-noloop/fineweb-edu-100bt/c256-512}{0.8997}
\defval{pos/x40/768/yocou-noloop/fineweb-edu-100bt/c512-1024}{0.9042}
\defval{pos/x40/768/yocou-noloop/fineweb-edu-100bt/last1024}{0.9104}
\defval{pos/x40/768/yocou-noloop/fineweb-edu-100bt/last128}{0.9061}
\defval{pos/x40/768/yocou-noloop/fineweb-edu-100bt/last256}{0.9076}
\defval{pos/x40/768/yocou-noloop/fineweb-edu-100bt/last512}{0.9087}
\defval{pos/x40/768/yocou-noloop/fineweb-edu-100bt/mean}{0.9172}
\defval{pos/x40/768/yocou-noloop/govreport-test/c1-128}{0.9172}
\defval{pos/x40/768/yocou-noloop/govreport-test/c1024-2048}{0.7278}
\defval{pos/x40/768/yocou-noloop/govreport-test/c128-256}{0.7890}
\defval{pos/x40/768/yocou-noloop/govreport-test/c256-512}{0.7618}
\defval{pos/x40/768/yocou-noloop/govreport-test/c512-1024}{0.7427}
\defval{pos/x40/768/yocou-noloop/govreport-test/last1024}{0.7278}
\defval{pos/x40/768/yocou-noloop/govreport-test/last128}{0.7221}
\defval{pos/x40/768/yocou-noloop/govreport-test/last256}{0.7227}
\defval{pos/x40/768/yocou-noloop/govreport-test/last512}{0.7249}
\defval{pos/x40/768/yocou-noloop/govreport-test/mean}{0.7514}
\defval{pos/x40/768/yocou-noloop/longdocs/c1-128}{1.3141}
\defval{pos/x40/768/yocou-noloop/longdocs/c1024-2048}{1.0781}
\defval{pos/x40/768/yocou-noloop/longdocs/c128-256}{1.1497}
\defval{pos/x40/768/yocou-noloop/longdocs/c256-512}{1.1172}
\defval{pos/x40/768/yocou-noloop/longdocs/c512-1024}{1.0925}
\defval{pos/x40/768/yocou-noloop/longdocs/last1024}{1.0781}
\defval{pos/x40/768/yocou-noloop/longdocs/last128}{1.0736}
\defval{pos/x40/768/yocou-noloop/longdocs/last256}{1.0734}
\defval{pos/x40/768/yocou-noloop/longdocs/last512}{1.0749}
\defval{pos/x40/768/yocou-noloop/longdocs/mean}{1.1058}
\defval{pos/x40/768/yocou-noloop/pg19-test/c1-128}{1.5651}
\defval{pos/x40/768/yocou-noloop/pg19-test/c1024-2048}{1.3716}
\defval{pos/x40/768/yocou-noloop/pg19-test/c128-256}{1.4263}
\defval{pos/x40/768/yocou-noloop/pg19-test/c256-512}{1.3976}
\defval{pos/x40/768/yocou-noloop/pg19-test/c512-1024}{1.3791}
\defval{pos/x40/768/yocou-noloop/pg19-test/last1024}{1.3716}
\defval{pos/x40/768/yocou-noloop/pg19-test/last128}{1.3718}
\defval{pos/x40/768/yocou-noloop/pg19-test/last256}{1.3713}
\defval{pos/x40/768/yocou-noloop/pg19-test/last512}{1.3715}
\defval{pos/x40/768/yocou-noloop/pg19-test/mean}{1.3922}
\defval{pos/x40/768/yocou-noloop/pg19-val/c1-128}{1.5196}
\defval{pos/x40/768/yocou-noloop/pg19-val/c1024-2048}{1.3323}
\defval{pos/x40/768/yocou-noloop/pg19-val/c128-256}{1.3803}
\defval{pos/x40/768/yocou-noloop/pg19-val/c256-512}{1.3576}
\defval{pos/x40/768/yocou-noloop/pg19-val/c512-1024}{1.3377}
\defval{pos/x40/768/yocou-noloop/pg19-val/last1024}{1.3323}
\defval{pos/x40/768/yocou-noloop/pg19-val/last128}{1.3242}
\defval{pos/x40/768/yocou-noloop/pg19-val/last256}{1.3234}
\defval{pos/x40/768/yocou-noloop/pg19-val/last512}{1.3262}
\defval{pos/x40/768/yocou-noloop/pg19-val/mean}{1.3515}
\defval{posabs/long/1024/ced-dec-minus-ced-enc/longdocs/c1-128}{0.0116}
\defval{posabs/long/1024/ced-dec-minus-ced-enc/longdocs/c1-128/hi}{0.0125}
\defval{posabs/long/1024/ced-dec-minus-ced-enc/longdocs/c1-128/lo}{0.0107}
\defval{posabs/long/1024/ced-dec-minus-ced-enc/longdocs/c1024-2048}{0.0264}
\defval{posabs/long/1024/ced-dec-minus-ced-enc/longdocs/c1024-2048/hi}{0.0271}
\defval{posabs/long/1024/ced-dec-minus-ced-enc/longdocs/c1024-2048/lo}{0.0256}
\defval{posabs/long/1024/ced-dec-minus-ced-enc/longdocs/c128-256}{0.0229}
\defval{posabs/long/1024/ced-dec-minus-ced-enc/longdocs/c128-256/hi}{0.0238}
\defval{posabs/long/1024/ced-dec-minus-ced-enc/longdocs/c128-256/lo}{0.0220}
\defval{posabs/long/1024/ced-dec-minus-ced-enc/longdocs/c256-512}{0.0254}
\defval{posabs/long/1024/ced-dec-minus-ced-enc/longdocs/c256-512/hi}{0.0262}
\defval{posabs/long/1024/ced-dec-minus-ced-enc/longdocs/c256-512/lo}{0.0246}
\defval{posabs/long/1024/ced-dec-minus-ced-enc/longdocs/c512-1024}{0.0262}
\defval{posabs/long/1024/ced-dec-minus-ced-enc/longdocs/c512-1024/hi}{0.0270}
\defval{posabs/long/1024/ced-dec-minus-ced-enc/longdocs/c512-1024/lo}{0.0254}
\defval{posabs/long/1024/ced-dec-minus-ced-enc/longdocs/mean}{0.0251}
\defval{posabs/long/1024/ced-dec-minus-ced-enc/longdocs/mean/hi}{0.0257}
\defval{posabs/long/1024/ced-dec-minus-ced-enc/longdocs/mean/lo}{0.0244}
\defval{posabs/long/1024/ced-dec-minus-dense-l47/longdocs/c1-128}{0.0027}
\defval{posabs/long/1024/ced-dec-minus-dense-l47/longdocs/c1-128/hi}{0.0036}
\defval{posabs/long/1024/ced-dec-minus-dense-l47/longdocs/c1-128/lo}{0.0018}
\defval{posabs/long/1024/ced-dec-minus-dense-l47/longdocs/c1024-2048}{0.0183}
\defval{posabs/long/1024/ced-dec-minus-dense-l47/longdocs/c1024-2048/hi}{0.0190}
\defval{posabs/long/1024/ced-dec-minus-dense-l47/longdocs/c1024-2048/lo}{0.0175}
\defval{posabs/long/1024/ced-dec-minus-dense-l47/longdocs/c128-256}{0.0113}
\defval{posabs/long/1024/ced-dec-minus-dense-l47/longdocs/c128-256/hi}{0.0121}
\defval{posabs/long/1024/ced-dec-minus-dense-l47/longdocs/c128-256/lo}{0.0104}
\defval{posabs/long/1024/ced-dec-minus-dense-l47/longdocs/c256-512}{0.0148}
\defval{posabs/long/1024/ced-dec-minus-dense-l47/longdocs/c256-512/hi}{0.0155}
\defval{posabs/long/1024/ced-dec-minus-dense-l47/longdocs/c256-512/lo}{0.0140}
\defval{posabs/long/1024/ced-dec-minus-dense-l47/longdocs/c512-1024}{0.0170}
\defval{posabs/long/1024/ced-dec-minus-dense-l47/longdocs/c512-1024/hi}{0.0177}
\defval{posabs/long/1024/ced-dec-minus-dense-l47/longdocs/c512-1024/lo}{0.0162}
\defval{posabs/long/1024/ced-dec-minus-dense-l47/longdocs/mean}{0.0161}
\defval{posabs/long/1024/ced-dec-minus-dense-l47/longdocs/mean/hi}{0.0167}
\defval{posabs/long/1024/ced-dec-minus-dense-l47/longdocs/mean/lo}{0.0154}
\defval{posabs/long/1024/ced-enc-minus-dense-l47/longdocs/c1-128}{0.0089}
\defval{posabs/long/1024/ced-enc-minus-dense-l47/longdocs/c1-128/hi}{0.0097}
\defval{posabs/long/1024/ced-enc-minus-dense-l47/longdocs/c1-128/lo}{0.0081}
\defval{posabs/long/1024/ced-enc-minus-dense-l47/longdocs/c1024-2048}{0.0081}
\defval{posabs/long/1024/ced-enc-minus-dense-l47/longdocs/c1024-2048/hi}{0.0086}
\defval{posabs/long/1024/ced-enc-minus-dense-l47/longdocs/c1024-2048/lo}{0.0076}
\defval{posabs/long/1024/ced-enc-minus-dense-l47/longdocs/c128-256}{0.0116}
\defval{posabs/long/1024/ced-enc-minus-dense-l47/longdocs/c128-256/hi}{0.0123}
\defval{posabs/long/1024/ced-enc-minus-dense-l47/longdocs/c128-256/lo}{0.0109}
\defval{posabs/long/1024/ced-enc-minus-dense-l47/longdocs/c256-512}{0.0106}
\defval{posabs/long/1024/ced-enc-minus-dense-l47/longdocs/c256-512/hi}{0.0112}
\defval{posabs/long/1024/ced-enc-minus-dense-l47/longdocs/c256-512/lo}{0.0100}
\defval{posabs/long/1024/ced-enc-minus-dense-l47/longdocs/c512-1024}{0.0092}
\defval{posabs/long/1024/ced-enc-minus-dense-l47/longdocs/c512-1024/hi}{0.0097}
\defval{posabs/long/1024/ced-enc-minus-dense-l47/longdocs/c512-1024/lo}{0.0086}
\defval{posabs/long/1024/ced-enc-minus-dense-l47/longdocs/mean}{0.0090}
\defval{posabs/long/1024/ced-enc-minus-dense-l47/longdocs/mean/hi}{0.0094}
\defval{posabs/long/1024/ced-enc-minus-dense-l47/longdocs/mean/lo}{0.0085}
\defval{posabs/long/1024/ced-noloop-minus-ced-dec/longdocs/c1-128}{0.0063}
\defval{posabs/long/1024/ced-noloop-minus-ced-dec/longdocs/c1-128/hi}{0.0070}
\defval{posabs/long/1024/ced-noloop-minus-ced-dec/longdocs/c1-128/lo}{0.0056}
\defval{posabs/long/1024/ced-noloop-minus-ced-dec/longdocs/c1024-2048}{0.0135}
\defval{posabs/long/1024/ced-noloop-minus-ced-dec/longdocs/c1024-2048/hi}{0.0140}
\defval{posabs/long/1024/ced-noloop-minus-ced-dec/longdocs/c1024-2048/lo}{0.0130}
\defval{posabs/long/1024/ced-noloop-minus-ced-dec/longdocs/c128-256}{0.0103}
\defval{posabs/long/1024/ced-noloop-minus-ced-dec/longdocs/c128-256/hi}{0.0110}
\defval{posabs/long/1024/ced-noloop-minus-ced-dec/longdocs/c128-256/lo}{0.0096}
\defval{posabs/long/1024/ced-noloop-minus-ced-dec/longdocs/c256-512}{0.0117}
\defval{posabs/long/1024/ced-noloop-minus-ced-dec/longdocs/c256-512/hi}{0.0123}
\defval{posabs/long/1024/ced-noloop-minus-ced-dec/longdocs/c256-512/lo}{0.0112}
\defval{posabs/long/1024/ced-noloop-minus-ced-dec/longdocs/c512-1024}{0.0130}
\defval{posabs/long/1024/ced-noloop-minus-ced-dec/longdocs/c512-1024/hi}{0.0135}
\defval{posabs/long/1024/ced-noloop-minus-ced-dec/longdocs/c512-1024/lo}{0.0125}
\defval{posabs/long/1024/ced-noloop-minus-ced-dec/longdocs/mean}{0.0125}
\defval{posabs/long/1024/ced-noloop-minus-ced-dec/longdocs/mean/hi}{0.0129}
\defval{posabs/long/1024/ced-noloop-minus-ced-dec/longdocs/mean/lo}{0.0120}
\defval{posabs/long/1024/ced-noloop-minus-ced-enc/longdocs/c1-128}{0.0179}
\defval{posabs/long/1024/ced-noloop-minus-ced-enc/longdocs/c1-128/hi}{0.0189}
\defval{posabs/long/1024/ced-noloop-minus-ced-enc/longdocs/c1-128/lo}{0.0169}
\defval{posabs/long/1024/ced-noloop-minus-ced-enc/longdocs/c1024-2048}{0.0399}
\defval{posabs/long/1024/ced-noloop-minus-ced-enc/longdocs/c1024-2048/hi}{0.0407}
\defval{posabs/long/1024/ced-noloop-minus-ced-enc/longdocs/c1024-2048/lo}{0.0390}
\defval{posabs/long/1024/ced-noloop-minus-ced-enc/longdocs/c128-256}{0.0332}
\defval{posabs/long/1024/ced-noloop-minus-ced-enc/longdocs/c128-256/hi}{0.0341}
\defval{posabs/long/1024/ced-noloop-minus-ced-enc/longdocs/c128-256/lo}{0.0323}
\defval{posabs/long/1024/ced-noloop-minus-ced-enc/longdocs/c256-512}{0.0371}
\defval{posabs/long/1024/ced-noloop-minus-ced-enc/longdocs/c256-512/hi}{0.0379}
\defval{posabs/long/1024/ced-noloop-minus-ced-enc/longdocs/c256-512/lo}{0.0362}
\defval{posabs/long/1024/ced-noloop-minus-ced-enc/longdocs/c512-1024}{0.0391}
\defval{posabs/long/1024/ced-noloop-minus-ced-enc/longdocs/c512-1024/hi}{0.0400}
\defval{posabs/long/1024/ced-noloop-minus-ced-enc/longdocs/c512-1024/lo}{0.0383}
\defval{posabs/long/1024/ced-noloop-minus-ced-enc/longdocs/mean}{0.0376}
\defval{posabs/long/1024/ced-noloop-minus-ced-enc/longdocs/mean/hi}{0.0383}
\defval{posabs/long/1024/ced-noloop-minus-ced-enc/longdocs/mean/lo}{0.0368}
\defval{posabs/long/1024/ced-noloop-minus-dense-l47/longdocs/c1-128}{0.0090}
\defval{posabs/long/1024/ced-noloop-minus-dense-l47/longdocs/c1-128/hi}{0.0099}
\defval{posabs/long/1024/ced-noloop-minus-dense-l47/longdocs/c1-128/lo}{0.0081}
\defval{posabs/long/1024/ced-noloop-minus-dense-l47/longdocs/c1024-2048}{0.0318}
\defval{posabs/long/1024/ced-noloop-minus-dense-l47/longdocs/c1024-2048/hi}{0.0326}
\defval{posabs/long/1024/ced-noloop-minus-dense-l47/longdocs/c1024-2048/lo}{0.0310}
\defval{posabs/long/1024/ced-noloop-minus-dense-l47/longdocs/c128-256}{0.0216}
\defval{posabs/long/1024/ced-noloop-minus-dense-l47/longdocs/c128-256/hi}{0.0224}
\defval{posabs/long/1024/ced-noloop-minus-dense-l47/longdocs/c128-256/lo}{0.0207}
\defval{posabs/long/1024/ced-noloop-minus-dense-l47/longdocs/c256-512}{0.0265}
\defval{posabs/long/1024/ced-noloop-minus-dense-l47/longdocs/c256-512/hi}{0.0273}
\defval{posabs/long/1024/ced-noloop-minus-dense-l47/longdocs/c256-512/lo}{0.0257}
\defval{posabs/long/1024/ced-noloop-minus-dense-l47/longdocs/c512-1024}{0.0299}
\defval{posabs/long/1024/ced-noloop-minus-dense-l47/longdocs/c512-1024/hi}{0.0307}
\defval{posabs/long/1024/ced-noloop-minus-dense-l47/longdocs/c512-1024/lo}{0.0292}
\defval{posabs/long/1024/ced-noloop-minus-dense-l47/longdocs/mean}{0.0286}
\defval{posabs/long/1024/ced-noloop-minus-dense-l47/longdocs/mean/hi}{0.0292}
\defval{posabs/long/1024/ced-noloop-minus-dense-l47/longdocs/mean/lo}{0.0279}
\defval{posabs/p140/768/p140-dec-fixed-minus-p140-enc-fixed/longdocs/c1-128}{0.0071}
\defval{posabs/p140/768/p140-dec-fixed-minus-p140-enc-fixed/longdocs/c1-128/hi}{0.0079}
\defval{posabs/p140/768/p140-dec-fixed-minus-p140-enc-fixed/longdocs/c1-128/lo}{0.0063}
\defval{posabs/p140/768/p140-dec-fixed-minus-p140-enc-fixed/longdocs/c1024-2048}{0.0221}
\defval{posabs/p140/768/p140-dec-fixed-minus-p140-enc-fixed/longdocs/c1024-2048/hi}{0.0229}
\defval{posabs/p140/768/p140-dec-fixed-minus-p140-enc-fixed/longdocs/c1024-2048/lo}{0.0214}
\defval{posabs/p140/768/p140-dec-fixed-minus-p140-enc-fixed/longdocs/c128-256}{0.0123}
\defval{posabs/p140/768/p140-dec-fixed-minus-p140-enc-fixed/longdocs/c128-256/hi}{0.0131}
\defval{posabs/p140/768/p140-dec-fixed-minus-p140-enc-fixed/longdocs/c128-256/lo}{0.0116}
\defval{posabs/p140/768/p140-dec-fixed-minus-p140-enc-fixed/longdocs/c256-512}{0.0167}
\defval{posabs/p140/768/p140-dec-fixed-minus-p140-enc-fixed/longdocs/c256-512/hi}{0.0173}
\defval{posabs/p140/768/p140-dec-fixed-minus-p140-enc-fixed/longdocs/c256-512/lo}{0.0160}
\defval{posabs/p140/768/p140-dec-fixed-minus-p140-enc-fixed/longdocs/c512-1024}{0.0200}
\defval{posabs/p140/768/p140-dec-fixed-minus-p140-enc-fixed/longdocs/c512-1024/hi}{0.0208}
\defval{posabs/p140/768/p140-dec-fixed-minus-p140-enc-fixed/longdocs/c512-1024/lo}{0.0193}
\defval{posabs/p140/768/p140-dec-fixed-minus-p140-enc-fixed/longdocs/mean}{0.0194}
\defval{posabs/p140/768/p140-dec-fixed-minus-p140-enc-fixed/longdocs/mean/hi}{0.0200}
\defval{posabs/p140/768/p140-dec-fixed-minus-p140-enc-fixed/longdocs/mean/lo}{0.0188}
\defval{posabs/p140/768/p140-enc-fixed-minus-p140-enc/longdocs/c1-128}{0.0032}
\defval{posabs/p140/768/p140-enc-fixed-minus-p140-enc/longdocs/c1-128/hi}{0.0039}
\defval{posabs/p140/768/p140-enc-fixed-minus-p140-enc/longdocs/c1-128/lo}{0.0026}
\defval{posabs/p140/768/p140-enc-fixed-minus-p140-enc/longdocs/c1024-2048}{0.0027}
\defval{posabs/p140/768/p140-enc-fixed-minus-p140-enc/longdocs/c1024-2048/hi}{0.0032}
\defval{posabs/p140/768/p140-enc-fixed-minus-p140-enc/longdocs/c1024-2048/lo}{0.0021}
\defval{posabs/p140/768/p140-enc-fixed-minus-p140-enc/longdocs/c128-256}{0.0035}
\defval{posabs/p140/768/p140-enc-fixed-minus-p140-enc/longdocs/c128-256/hi}{0.0041}
\defval{posabs/p140/768/p140-enc-fixed-minus-p140-enc/longdocs/c128-256/lo}{0.0028}
\defval{posabs/p140/768/p140-enc-fixed-minus-p140-enc/longdocs/c256-512}{0.0030}
\defval{posabs/p140/768/p140-enc-fixed-minus-p140-enc/longdocs/c256-512/hi}{0.0035}
\defval{posabs/p140/768/p140-enc-fixed-minus-p140-enc/longdocs/c256-512/lo}{0.0024}
\defval{posabs/p140/768/p140-enc-fixed-minus-p140-enc/longdocs/c512-1024}{0.0024}
\defval{posabs/p140/768/p140-enc-fixed-minus-p140-enc/longdocs/c512-1024/hi}{0.0030}
\defval{posabs/p140/768/p140-enc-fixed-minus-p140-enc/longdocs/c512-1024/lo}{0.0019}
\defval{posabs/p140/768/p140-enc-fixed-minus-p140-enc/longdocs/mean}{0.0027}
\defval{posabs/p140/768/p140-enc-fixed-minus-p140-enc/longdocs/mean/hi}{0.0032}
\defval{posabs/p140/768/p140-enc-fixed-minus-p140-enc/longdocs/mean/lo}{0.0022}
\defval{posabs/p140/768/p140-noloop-minus-p140-enc/longdocs/c1-128}{0.0280}
\defval{posabs/p140/768/p140-noloop-minus-p140-enc/longdocs/c1-128/hi}{0.0287}
\defval{posabs/p140/768/p140-noloop-minus-p140-enc/longdocs/c1-128/lo}{0.0272}
\defval{posabs/p140/768/p140-noloop-minus-p140-enc/longdocs/c1024-2048}{0.0500}
\defval{posabs/p140/768/p140-noloop-minus-p140-enc/longdocs/c1024-2048/hi}{0.0510}
\defval{posabs/p140/768/p140-noloop-minus-p140-enc/longdocs/c1024-2048/lo}{0.0489}
\defval{posabs/p140/768/p140-noloop-minus-p140-enc/longdocs/c128-256}{0.0400}
\defval{posabs/p140/768/p140-noloop-minus-p140-enc/longdocs/c128-256/hi}{0.0409}
\defval{posabs/p140/768/p140-noloop-minus-p140-enc/longdocs/c128-256/lo}{0.0390}
\defval{posabs/p140/768/p140-noloop-minus-p140-enc/longdocs/c256-512}{0.0444}
\defval{posabs/p140/768/p140-noloop-minus-p140-enc/longdocs/c256-512/hi}{0.0453}
\defval{posabs/p140/768/p140-noloop-minus-p140-enc/longdocs/c256-512/lo}{0.0436}
\defval{posabs/p140/768/p140-noloop-minus-p140-enc/longdocs/c512-1024}{0.0475}
\defval{posabs/p140/768/p140-noloop-minus-p140-enc/longdocs/c512-1024/hi}{0.0485}
\defval{posabs/p140/768/p140-noloop-minus-p140-enc/longdocs/c512-1024/lo}{0.0466}
\defval{posabs/p140/768/p140-noloop-minus-p140-enc/longdocs/mean}{0.0467}
\defval{posabs/p140/768/p140-noloop-minus-p140-enc/longdocs/mean/hi}{0.0475}
\defval{posabs/p140/768/p140-noloop-minus-p140-enc/longdocs/mean/lo}{0.0458}
\defval{posabs/x40/1024/ced-dec-minus-ced-enc/longdocs/c1-128}{0.0198}
\defval{posabs/x40/1024/ced-dec-minus-ced-enc/longdocs/c1-128/hi}{0.0205}
\defval{posabs/x40/1024/ced-dec-minus-ced-enc/longdocs/c1-128/lo}{0.0191}
\defval{posabs/x40/1024/ced-dec-minus-ced-enc/longdocs/c1024-2048}{0.0288}
\defval{posabs/x40/1024/ced-dec-minus-ced-enc/longdocs/c1024-2048/hi}{0.0295}
\defval{posabs/x40/1024/ced-dec-minus-ced-enc/longdocs/c1024-2048/lo}{0.0282}
\defval{posabs/x40/1024/ced-dec-minus-ced-enc/longdocs/c128-256}{0.0235}
\defval{posabs/x40/1024/ced-dec-minus-ced-enc/longdocs/c128-256/hi}{0.0242}
\defval{posabs/x40/1024/ced-dec-minus-ced-enc/longdocs/c128-256/lo}{0.0227}
\defval{posabs/x40/1024/ced-dec-minus-ced-enc/longdocs/c256-512}{0.0254}
\defval{posabs/x40/1024/ced-dec-minus-ced-enc/longdocs/c256-512/hi}{0.0260}
\defval{posabs/x40/1024/ced-dec-minus-ced-enc/longdocs/c256-512/lo}{0.0248}
\defval{posabs/x40/1024/ced-dec-minus-ced-enc/longdocs/c512-1024}{0.0270}
\defval{posabs/x40/1024/ced-dec-minus-ced-enc/longdocs/c512-1024/hi}{0.0276}
\defval{posabs/x40/1024/ced-dec-minus-ced-enc/longdocs/c512-1024/lo}{0.0264}
\defval{posabs/x40/1024/ced-dec-minus-ced-enc/longdocs/mean}{0.0271}
\defval{posabs/x40/1024/ced-dec-minus-ced-enc/longdocs/mean/hi}{0.0276}
\defval{posabs/x40/1024/ced-dec-minus-ced-enc/longdocs/mean/lo}{0.0266}
\defval{posabs/x40/1024/ced-dec-minus-dense-l25/longdocs/c1-128}{0.0206}
\defval{posabs/x40/1024/ced-dec-minus-dense-l25/longdocs/c1-128/hi}{0.0215}
\defval{posabs/x40/1024/ced-dec-minus-dense-l25/longdocs/c1-128/lo}{0.0197}
\defval{posabs/x40/1024/ced-dec-minus-dense-l25/longdocs/c1024-2048}{0.0295}
\defval{posabs/x40/1024/ced-dec-minus-dense-l25/longdocs/c1024-2048/hi}{0.0305}
\defval{posabs/x40/1024/ced-dec-minus-dense-l25/longdocs/c1024-2048/lo}{0.0287}
\defval{posabs/x40/1024/ced-dec-minus-dense-l25/longdocs/c128-256}{0.0240}
\defval{posabs/x40/1024/ced-dec-minus-dense-l25/longdocs/c128-256/hi}{0.0249}
\defval{posabs/x40/1024/ced-dec-minus-dense-l25/longdocs/c128-256/lo}{0.0230}
\defval{posabs/x40/1024/ced-dec-minus-dense-l25/longdocs/c256-512}{0.0260}
\defval{posabs/x40/1024/ced-dec-minus-dense-l25/longdocs/c256-512/hi}{0.0268}
\defval{posabs/x40/1024/ced-dec-minus-dense-l25/longdocs/c256-512/lo}{0.0251}
\defval{posabs/x40/1024/ced-dec-minus-dense-l25/longdocs/c512-1024}{0.0277}
\defval{posabs/x40/1024/ced-dec-minus-dense-l25/longdocs/c512-1024/hi}{0.0285}
\defval{posabs/x40/1024/ced-dec-minus-dense-l25/longdocs/c512-1024/lo}{0.0269}
\defval{posabs/x40/1024/ced-dec-minus-dense-l25/longdocs/mean}{0.0277}
\defval{posabs/x40/1024/ced-dec-minus-dense-l25/longdocs/mean/hi}{0.0285}
\defval{posabs/x40/1024/ced-dec-minus-dense-l25/longdocs/mean/lo}{0.0270}
\defval{posabs/x40/1024/ced-enc-fixed-minus-ced-enc/longdocs/c1-128}{0.0066}
\defval{posabs/x40/1024/ced-enc-fixed-minus-ced-enc/longdocs/c1-128/hi}{0.0072}
\defval{posabs/x40/1024/ced-enc-fixed-minus-ced-enc/longdocs/c1-128/lo}{0.0059}
\defval{posabs/x40/1024/ced-enc-fixed-minus-ced-enc/longdocs/c1024-2048}{0.0044}
\defval{posabs/x40/1024/ced-enc-fixed-minus-ced-enc/longdocs/c1024-2048/hi}{0.0049}
\defval{posabs/x40/1024/ced-enc-fixed-minus-ced-enc/longdocs/c1024-2048/lo}{0.0040}
\defval{posabs/x40/1024/ced-enc-fixed-minus-ced-enc/longdocs/c128-256}{0.0040}
\defval{posabs/x40/1024/ced-enc-fixed-minus-ced-enc/longdocs/c128-256/hi}{0.0047}
\defval{posabs/x40/1024/ced-enc-fixed-minus-ced-enc/longdocs/c128-256/lo}{0.0034}
\defval{posabs/x40/1024/ced-enc-fixed-minus-ced-enc/longdocs/c256-512}{0.0042}
\defval{posabs/x40/1024/ced-enc-fixed-minus-ced-enc/longdocs/c256-512/hi}{0.0046}
\defval{posabs/x40/1024/ced-enc-fixed-minus-ced-enc/longdocs/c256-512/lo}{0.0037}
\defval{posabs/x40/1024/ced-enc-fixed-minus-ced-enc/longdocs/c512-1024}{0.0042}
\defval{posabs/x40/1024/ced-enc-fixed-minus-ced-enc/longdocs/c512-1024/hi}{0.0047}
\defval{posabs/x40/1024/ced-enc-fixed-minus-ced-enc/longdocs/c512-1024/lo}{0.0037}
\defval{posabs/x40/1024/ced-enc-fixed-minus-ced-enc/longdocs/mean}{0.0045}
\defval{posabs/x40/1024/ced-enc-fixed-minus-ced-enc/longdocs/mean/hi}{0.0048}
\defval{posabs/x40/1024/ced-enc-fixed-minus-ced-enc/longdocs/mean/lo}{0.0041}
\defval{posabs/x40/1024/ced-enc-minus-dense-l25/longdocs/c1-128}{0.0008}
\defval{posabs/x40/1024/ced-enc-minus-dense-l25/longdocs/c1-128/hi}{0.0016}
\defval{posabs/x40/1024/ced-enc-minus-dense-l25/longdocs/c1-128/lo}{0.0000}
\defval{posabs/x40/1024/ced-enc-minus-dense-l25/longdocs/c1024-2048}{0.0007}
\defval{posabs/x40/1024/ced-enc-minus-dense-l25/longdocs/c1024-2048/hi}{0.0015}
\defval{posabs/x40/1024/ced-enc-minus-dense-l25/longdocs/c1024-2048/lo}{0.0000}
\defval{posabs/x40/1024/ced-enc-minus-dense-l25/longdocs/c128-256}{0.0005}
\defval{posabs/x40/1024/ced-enc-minus-dense-l25/longdocs/c128-256/hi}{0.0012}
\defval{posabs/x40/1024/ced-enc-minus-dense-l25/longdocs/c128-256/lo}{0.0000}
\defval{posabs/x40/1024/ced-enc-minus-dense-l25/longdocs/c256-512}{0.0005}
\defval{posabs/x40/1024/ced-enc-minus-dense-l25/longdocs/c256-512/hi}{0.0012}
\defval{posabs/x40/1024/ced-enc-minus-dense-l25/longdocs/c256-512/lo}{0.0000}
\defval{posabs/x40/1024/ced-enc-minus-dense-l25/longdocs/c512-1024}{0.0007}
\defval{posabs/x40/1024/ced-enc-minus-dense-l25/longdocs/c512-1024/hi}{0.0014}
\defval{posabs/x40/1024/ced-enc-minus-dense-l25/longdocs/c512-1024/lo}{0.0000}
\defval{posabs/x40/1024/ced-enc-minus-dense-l25/longdocs/mean}{0.0007}
\defval{posabs/x40/1024/ced-enc-minus-dense-l25/longdocs/mean/hi}{0.0013}
\defval{posabs/x40/1024/ced-enc-minus-dense-l25/longdocs/mean/lo}{0.0000}
\defval{posabs/x40/1024/ced-enc-minus-yocou-loop/longdocs/c1-128}{0.0050}
\defval{posabs/x40/1024/ced-enc-minus-yocou-loop/longdocs/c1-128/hi}{0.0056}
\defval{posabs/x40/1024/ced-enc-minus-yocou-loop/longdocs/c1-128/lo}{0.0043}
\defval{posabs/x40/1024/ced-enc-minus-yocou-loop/longdocs/c1024-2048}{0.0107}
\defval{posabs/x40/1024/ced-enc-minus-yocou-loop/longdocs/c1024-2048/hi}{0.0113}
\defval{posabs/x40/1024/ced-enc-minus-yocou-loop/longdocs/c1024-2048/lo}{0.0101}
\defval{posabs/x40/1024/ced-enc-minus-yocou-loop/longdocs/c128-256}{0.0003}
\defval{posabs/x40/1024/ced-enc-minus-yocou-loop/longdocs/c128-256/hi}{0.0009}
\defval{posabs/x40/1024/ced-enc-minus-yocou-loop/longdocs/c128-256/lo}{0.0000}
\defval{posabs/x40/1024/ced-enc-minus-yocou-loop/longdocs/c256-512}{0.0041}
\defval{posabs/x40/1024/ced-enc-minus-yocou-loop/longdocs/c256-512/hi}{0.0046}
\defval{posabs/x40/1024/ced-enc-minus-yocou-loop/longdocs/c256-512/lo}{0.0035}
\defval{posabs/x40/1024/ced-enc-minus-yocou-loop/longdocs/c512-1024}{0.0077}
\defval{posabs/x40/1024/ced-enc-minus-yocou-loop/longdocs/c512-1024/hi}{0.0083}
\defval{posabs/x40/1024/ced-enc-minus-yocou-loop/longdocs/c512-1024/lo}{0.0072}
\defval{posabs/x40/1024/ced-enc-minus-yocou-loop/longdocs/mean}{0.0074}
\defval{posabs/x40/1024/ced-enc-minus-yocou-loop/longdocs/mean/hi}{0.0079}
\defval{posabs/x40/1024/ced-enc-minus-yocou-loop/longdocs/mean/lo}{0.0070}
\defval{posabs/x40/1024/ced-noloop-minus-ced-dec/longdocs/c1-128}{0.0079}
\defval{posabs/x40/1024/ced-noloop-minus-ced-dec/longdocs/c1-128/hi}{0.0086}
\defval{posabs/x40/1024/ced-noloop-minus-ced-dec/longdocs/c1-128/lo}{0.0071}
\defval{posabs/x40/1024/ced-noloop-minus-ced-dec/longdocs/c1024-2048}{0.0070}
\defval{posabs/x40/1024/ced-noloop-minus-ced-dec/longdocs/c1024-2048/hi}{0.0076}
\defval{posabs/x40/1024/ced-noloop-minus-ced-dec/longdocs/c1024-2048/lo}{0.0064}
\defval{posabs/x40/1024/ced-noloop-minus-ced-dec/longdocs/c128-256}{0.0089}
\defval{posabs/x40/1024/ced-noloop-minus-ced-dec/longdocs/c128-256/hi}{0.0096}
\defval{posabs/x40/1024/ced-noloop-minus-ced-dec/longdocs/c128-256/lo}{0.0082}
\defval{posabs/x40/1024/ced-noloop-minus-ced-dec/longdocs/c256-512}{0.0084}
\defval{posabs/x40/1024/ced-noloop-minus-ced-dec/longdocs/c256-512/hi}{0.0090}
\defval{posabs/x40/1024/ced-noloop-minus-ced-dec/longdocs/c256-512/lo}{0.0078}
\defval{posabs/x40/1024/ced-noloop-minus-ced-dec/longdocs/c512-1024}{0.0079}
\defval{posabs/x40/1024/ced-noloop-minus-ced-dec/longdocs/c512-1024/hi}{0.0085}
\defval{posabs/x40/1024/ced-noloop-minus-ced-dec/longdocs/c512-1024/lo}{0.0072}
\defval{posabs/x40/1024/ced-noloop-minus-ced-dec/longdocs/mean}{0.0076}
\defval{posabs/x40/1024/ced-noloop-minus-ced-dec/longdocs/mean/hi}{0.0081}
\defval{posabs/x40/1024/ced-noloop-minus-ced-dec/longdocs/mean/lo}{0.0070}
\defval{posabs/x40/1024/ced-noloop-minus-ced-enc/longdocs/c1-128}{0.0277}
\defval{posabs/x40/1024/ced-noloop-minus-ced-enc/longdocs/c1-128/hi}{0.0284}
\defval{posabs/x40/1024/ced-noloop-minus-ced-enc/longdocs/c1-128/lo}{0.0269}
\defval{posabs/x40/1024/ced-noloop-minus-ced-enc/longdocs/c1024-2048}{0.0359}
\defval{posabs/x40/1024/ced-noloop-minus-ced-enc/longdocs/c1024-2048/hi}{0.0367}
\defval{posabs/x40/1024/ced-noloop-minus-ced-enc/longdocs/c1024-2048/lo}{0.0351}
\defval{posabs/x40/1024/ced-noloop-minus-ced-enc/longdocs/c128-256}{0.0323}
\defval{posabs/x40/1024/ced-noloop-minus-ced-enc/longdocs/c128-256/hi}{0.0332}
\defval{posabs/x40/1024/ced-noloop-minus-ced-enc/longdocs/c128-256/lo}{0.0314}
\defval{posabs/x40/1024/ced-noloop-minus-ced-enc/longdocs/c256-512}{0.0338}
\defval{posabs/x40/1024/ced-noloop-minus-ced-enc/longdocs/c256-512/hi}{0.0345}
\defval{posabs/x40/1024/ced-noloop-minus-ced-enc/longdocs/c256-512/lo}{0.0331}
\defval{posabs/x40/1024/ced-noloop-minus-ced-enc/longdocs/c512-1024}{0.0349}
\defval{posabs/x40/1024/ced-noloop-minus-ced-enc/longdocs/c512-1024/hi}{0.0355}
\defval{posabs/x40/1024/ced-noloop-minus-ced-enc/longdocs/c512-1024/lo}{0.0341}
\defval{posabs/x40/1024/ced-noloop-minus-ced-enc/longdocs/mean}{0.0346}
\defval{posabs/x40/1024/ced-noloop-minus-ced-enc/longdocs/mean/hi}{0.0353}
\defval{posabs/x40/1024/ced-noloop-minus-ced-enc/longdocs/mean/lo}{0.0340}
\defval{posabs/x40/1024/ced-noloop-minus-dense-noloop/longdocs/c1-128}{0.0039}
\defval{posabs/x40/1024/ced-noloop-minus-dense-noloop/longdocs/c1-128/hi}{0.0049}
\defval{posabs/x40/1024/ced-noloop-minus-dense-noloop/longdocs/c1-128/lo}{0.0030}
\defval{posabs/x40/1024/ced-noloop-minus-dense-noloop/longdocs/c1024-2048}{0.0063}
\defval{posabs/x40/1024/ced-noloop-minus-dense-noloop/longdocs/c1024-2048/hi}{0.0072}
\defval{posabs/x40/1024/ced-noloop-minus-dense-noloop/longdocs/c1024-2048/lo}{0.0055}
\defval{posabs/x40/1024/ced-noloop-minus-dense-noloop/longdocs/c128-256}{0.0060}
\defval{posabs/x40/1024/ced-noloop-minus-dense-noloop/longdocs/c128-256/hi}{0.0069}
\defval{posabs/x40/1024/ced-noloop-minus-dense-noloop/longdocs/c128-256/lo}{0.0050}
\defval{posabs/x40/1024/ced-noloop-minus-dense-noloop/longdocs/c256-512}{0.0064}
\defval{posabs/x40/1024/ced-noloop-minus-dense-noloop/longdocs/c256-512/hi}{0.0072}
\defval{posabs/x40/1024/ced-noloop-minus-dense-noloop/longdocs/c256-512/lo}{0.0057}
\defval{posabs/x40/1024/ced-noloop-minus-dense-noloop/longdocs/c512-1024}{0.0060}
\defval{posabs/x40/1024/ced-noloop-minus-dense-noloop/longdocs/c512-1024/hi}{0.0068}
\defval{posabs/x40/1024/ced-noloop-minus-dense-noloop/longdocs/c512-1024/lo}{0.0052}
\defval{posabs/x40/1024/ced-noloop-minus-dense-noloop/longdocs/mean}{0.0061}
\defval{posabs/x40/1024/ced-noloop-minus-dense-noloop/longdocs/mean/hi}{0.0068}
\defval{posabs/x40/1024/ced-noloop-minus-dense-noloop/longdocs/mean/lo}{0.0053}
\defval{posabs/x40/1024/ced-noloop-minus-yocou-noloop/longdocs/c1-128}{0.0074}
\defval{posabs/x40/1024/ced-noloop-minus-yocou-noloop/longdocs/c1-128/hi}{0.0081}
\defval{posabs/x40/1024/ced-noloop-minus-yocou-noloop/longdocs/c1-128/lo}{0.0067}
\defval{posabs/x40/1024/ced-noloop-minus-yocou-noloop/longdocs/c1024-2048}{0.0037}
\defval{posabs/x40/1024/ced-noloop-minus-yocou-noloop/longdocs/c1024-2048/hi}{0.0043}
\defval{posabs/x40/1024/ced-noloop-minus-yocou-noloop/longdocs/c1024-2048/lo}{0.0031}
\defval{posabs/x40/1024/ced-noloop-minus-yocou-noloop/longdocs/c128-256}{0.0035}
\defval{posabs/x40/1024/ced-noloop-minus-yocou-noloop/longdocs/c128-256/hi}{0.0042}
\defval{posabs/x40/1024/ced-noloop-minus-yocou-noloop/longdocs/c128-256/lo}{0.0028}
\defval{posabs/x40/1024/ced-noloop-minus-yocou-noloop/longdocs/c256-512}{0.0004}
\defval{posabs/x40/1024/ced-noloop-minus-yocou-noloop/longdocs/c256-512/hi}{0.0009}
\defval{posabs/x40/1024/ced-noloop-minus-yocou-noloop/longdocs/c256-512/lo}{0.0000}
\defval{posabs/x40/1024/ced-noloop-minus-yocou-noloop/longdocs/c512-1024}{0.0021}
\defval{posabs/x40/1024/ced-noloop-minus-yocou-noloop/longdocs/c512-1024/hi}{0.0027}
\defval{posabs/x40/1024/ced-noloop-minus-yocou-noloop/longdocs/c512-1024/lo}{0.0015}
\defval{posabs/x40/1024/ced-noloop-minus-yocou-noloop/longdocs/mean}{0.0017}
\defval{posabs/x40/1024/ced-noloop-minus-yocou-noloop/longdocs/mean/hi}{0.0022}
\defval{posabs/x40/1024/ced-noloop-minus-yocou-noloop/longdocs/mean/lo}{0.0011}
\defval{posabs/x40/1024/dense-l47-minus-dense-l25/longdocs/c1-128}{0.0041}
\defval{posabs/x40/1024/dense-l47-minus-dense-l25/longdocs/c1-128/hi}{0.0048}
\defval{posabs/x40/1024/dense-l47-minus-dense-l25/longdocs/c1-128/lo}{0.0033}
\defval{posabs/x40/1024/dense-l47-minus-dense-l25/longdocs/c1024-2048}{0.0091}
\defval{posabs/x40/1024/dense-l47-minus-dense-l25/longdocs/c1024-2048/hi}{0.0097}
\defval{posabs/x40/1024/dense-l47-minus-dense-l25/longdocs/c1024-2048/lo}{0.0086}
\defval{posabs/x40/1024/dense-l47-minus-dense-l25/longdocs/c128-256}{0.0075}
\defval{posabs/x40/1024/dense-l47-minus-dense-l25/longdocs/c128-256/hi}{0.0082}
\defval{posabs/x40/1024/dense-l47-minus-dense-l25/longdocs/c128-256/lo}{0.0068}
\defval{posabs/x40/1024/dense-l47-minus-dense-l25/longdocs/c256-512}{0.0079}
\defval{posabs/x40/1024/dense-l47-minus-dense-l25/longdocs/c256-512/hi}{0.0085}
\defval{posabs/x40/1024/dense-l47-minus-dense-l25/longdocs/c256-512/lo}{0.0073}
\defval{posabs/x40/1024/dense-l47-minus-dense-l25/longdocs/c512-1024}{0.0085}
\defval{posabs/x40/1024/dense-l47-minus-dense-l25/longdocs/c512-1024/hi}{0.0090}
\defval{posabs/x40/1024/dense-l47-minus-dense-l25/longdocs/c512-1024/lo}{0.0079}
\defval{posabs/x40/1024/dense-l47-minus-dense-l25/longdocs/mean}{0.0084}
\defval{posabs/x40/1024/dense-l47-minus-dense-l25/longdocs/mean/hi}{0.0088}
\defval{posabs/x40/1024/dense-l47-minus-dense-l25/longdocs/mean/lo}{0.0079}
\defval{posabs/x40/1024/dense-noloop-minus-dense-l25/longdocs/c1-128}{0.0324}
\defval{posabs/x40/1024/dense-noloop-minus-dense-l25/longdocs/c1-128/hi}{0.0332}
\defval{posabs/x40/1024/dense-noloop-minus-dense-l25/longdocs/c1-128/lo}{0.0315}
\defval{posabs/x40/1024/dense-noloop-minus-dense-l25/longdocs/c1024-2048}{0.0429}
\defval{posabs/x40/1024/dense-noloop-minus-dense-l25/longdocs/c1024-2048/hi}{0.0438}
\defval{posabs/x40/1024/dense-noloop-minus-dense-l25/longdocs/c1024-2048/lo}{0.0420}
\defval{posabs/x40/1024/dense-noloop-minus-dense-l25/longdocs/c128-256}{0.0388}
\defval{posabs/x40/1024/dense-noloop-minus-dense-l25/longdocs/c128-256/hi}{0.0397}
\defval{posabs/x40/1024/dense-noloop-minus-dense-l25/longdocs/c128-256/lo}{0.0379}
\defval{posabs/x40/1024/dense-noloop-minus-dense-l25/longdocs/c256-512}{0.0408}
\defval{posabs/x40/1024/dense-noloop-minus-dense-l25/longdocs/c256-512/hi}{0.0415}
\defval{posabs/x40/1024/dense-noloop-minus-dense-l25/longdocs/c256-512/lo}{0.0400}
\defval{posabs/x40/1024/dense-noloop-minus-dense-l25/longdocs/c512-1024}{0.0416}
\defval{posabs/x40/1024/dense-noloop-minus-dense-l25/longdocs/c512-1024/hi}{0.0423}
\defval{posabs/x40/1024/dense-noloop-minus-dense-l25/longdocs/c512-1024/lo}{0.0408}
\defval{posabs/x40/1024/dense-noloop-minus-dense-l25/longdocs/mean}{0.0414}
\defval{posabs/x40/1024/dense-noloop-minus-dense-l25/longdocs/mean/hi}{0.0421}
\defval{posabs/x40/1024/dense-noloop-minus-dense-l25/longdocs/mean/lo}{0.0407}
\defval{posabs/x40/1024/yocou-loop-minus-ced-enc/longdocs/c1-128}{0.0050}
\defval{posabs/x40/1024/yocou-loop-minus-ced-enc/longdocs/c1-128/hi}{0.0056}
\defval{posabs/x40/1024/yocou-loop-minus-ced-enc/longdocs/c1-128/lo}{0.0043}
\defval{posabs/x40/1024/yocou-loop-minus-ced-enc/longdocs/c1024-2048}{0.0107}
\defval{posabs/x40/1024/yocou-loop-minus-ced-enc/longdocs/c1024-2048/hi}{0.0113}
\defval{posabs/x40/1024/yocou-loop-minus-ced-enc/longdocs/c1024-2048/lo}{0.0101}
\defval{posabs/x40/1024/yocou-loop-minus-ced-enc/longdocs/c128-256}{0.0003}
\defval{posabs/x40/1024/yocou-loop-minus-ced-enc/longdocs/c128-256/hi}{0.0009}
\defval{posabs/x40/1024/yocou-loop-minus-ced-enc/longdocs/c128-256/lo}{0.0000}
\defval{posabs/x40/1024/yocou-loop-minus-ced-enc/longdocs/c256-512}{0.0041}
\defval{posabs/x40/1024/yocou-loop-minus-ced-enc/longdocs/c256-512/hi}{0.0046}
\defval{posabs/x40/1024/yocou-loop-minus-ced-enc/longdocs/c256-512/lo}{0.0035}
\defval{posabs/x40/1024/yocou-loop-minus-ced-enc/longdocs/c512-1024}{0.0077}
\defval{posabs/x40/1024/yocou-loop-minus-ced-enc/longdocs/c512-1024/hi}{0.0083}
\defval{posabs/x40/1024/yocou-loop-minus-ced-enc/longdocs/c512-1024/lo}{0.0072}
\defval{posabs/x40/1024/yocou-loop-minus-ced-enc/longdocs/mean}{0.0074}
\defval{posabs/x40/1024/yocou-loop-minus-ced-enc/longdocs/mean/hi}{0.0079}
\defval{posabs/x40/1024/yocou-loop-minus-ced-enc/longdocs/mean/lo}{0.0070}
\defval{posabs/x40/1024/yocou-loop-minus-dense-l25/longdocs/c1-128}{0.0042}
\defval{posabs/x40/1024/yocou-loop-minus-dense-l25/longdocs/c1-128/hi}{0.0049}
\defval{posabs/x40/1024/yocou-loop-minus-dense-l25/longdocs/c1-128/lo}{0.0034}
\defval{posabs/x40/1024/yocou-loop-minus-dense-l25/longdocs/c1024-2048}{0.0114}
\defval{posabs/x40/1024/yocou-loop-minus-dense-l25/longdocs/c1024-2048/hi}{0.0121}
\defval{posabs/x40/1024/yocou-loop-minus-dense-l25/longdocs/c1024-2048/lo}{0.0106}
\defval{posabs/x40/1024/yocou-loop-minus-dense-l25/longdocs/c128-256}{0.0002}
\defval{posabs/x40/1024/yocou-loop-minus-dense-l25/longdocs/c128-256/hi}{0.0010}
\defval{posabs/x40/1024/yocou-loop-minus-dense-l25/longdocs/c128-256/lo}{0.0000}
\defval{posabs/x40/1024/yocou-loop-minus-dense-l25/longdocs/c256-512}{0.0046}
\defval{posabs/x40/1024/yocou-loop-minus-dense-l25/longdocs/c256-512/hi}{0.0053}
\defval{posabs/x40/1024/yocou-loop-minus-dense-l25/longdocs/c256-512/lo}{0.0039}
\defval{posabs/x40/1024/yocou-loop-minus-dense-l25/longdocs/c512-1024}{0.0084}
\defval{posabs/x40/1024/yocou-loop-minus-dense-l25/longdocs/c512-1024/hi}{0.0091}
\defval{posabs/x40/1024/yocou-loop-minus-dense-l25/longdocs/c512-1024/lo}{0.0077}
\defval{posabs/x40/1024/yocou-loop-minus-dense-l25/longdocs/mean}{0.0081}
\defval{posabs/x40/1024/yocou-loop-minus-dense-l25/longdocs/mean/hi}{0.0087}
\defval{posabs/x40/1024/yocou-loop-minus-dense-l25/longdocs/mean/lo}{0.0075}
\defval{posabs/x40/1024/yocou-noloop-minus-ced-noloop/longdocs/c1-128}{0.0074}
\defval{posabs/x40/1024/yocou-noloop-minus-ced-noloop/longdocs/c1-128/hi}{0.0081}
\defval{posabs/x40/1024/yocou-noloop-minus-ced-noloop/longdocs/c1-128/lo}{0.0067}
\defval{posabs/x40/1024/yocou-noloop-minus-ced-noloop/longdocs/c1024-2048}{0.0037}
\defval{posabs/x40/1024/yocou-noloop-minus-ced-noloop/longdocs/c1024-2048/hi}{0.0043}
\defval{posabs/x40/1024/yocou-noloop-minus-ced-noloop/longdocs/c1024-2048/lo}{0.0031}
\defval{posabs/x40/1024/yocou-noloop-minus-ced-noloop/longdocs/c128-256}{0.0035}
\defval{posabs/x40/1024/yocou-noloop-minus-ced-noloop/longdocs/c128-256/hi}{0.0042}
\defval{posabs/x40/1024/yocou-noloop-minus-ced-noloop/longdocs/c128-256/lo}{0.0028}
\defval{posabs/x40/1024/yocou-noloop-minus-ced-noloop/longdocs/c256-512}{0.0004}
\defval{posabs/x40/1024/yocou-noloop-minus-ced-noloop/longdocs/c256-512/hi}{0.0009}
\defval{posabs/x40/1024/yocou-noloop-minus-ced-noloop/longdocs/c256-512/lo}{0.0000}
\defval{posabs/x40/1024/yocou-noloop-minus-ced-noloop/longdocs/c512-1024}{0.0021}
\defval{posabs/x40/1024/yocou-noloop-minus-ced-noloop/longdocs/c512-1024/hi}{0.0027}
\defval{posabs/x40/1024/yocou-noloop-minus-ced-noloop/longdocs/c512-1024/lo}{0.0015}
\defval{posabs/x40/1024/yocou-noloop-minus-ced-noloop/longdocs/mean}{0.0017}
\defval{posabs/x40/1024/yocou-noloop-minus-ced-noloop/longdocs/mean/hi}{0.0022}
\defval{posabs/x40/1024/yocou-noloop-minus-ced-noloop/longdocs/mean/lo}{0.0011}
\defval{posabs/x40/1024/yocou-noloop-minus-dense-noloop/longdocs/c1-128}{0.0113}
\defval{posabs/x40/1024/yocou-noloop-minus-dense-noloop/longdocs/c1-128/hi}{0.0122}
\defval{posabs/x40/1024/yocou-noloop-minus-dense-noloop/longdocs/c1-128/lo}{0.0105}
\defval{posabs/x40/1024/yocou-noloop-minus-dense-noloop/longdocs/c1024-2048}{0.0026}
\defval{posabs/x40/1024/yocou-noloop-minus-dense-noloop/longdocs/c1024-2048/hi}{0.0034}
\defval{posabs/x40/1024/yocou-noloop-minus-dense-noloop/longdocs/c1024-2048/lo}{0.0018}
\defval{posabs/x40/1024/yocou-noloop-minus-dense-noloop/longdocs/c128-256}{0.0094}
\defval{posabs/x40/1024/yocou-noloop-minus-dense-noloop/longdocs/c128-256/hi}{0.0103}
\defval{posabs/x40/1024/yocou-noloop-minus-dense-noloop/longdocs/c128-256/lo}{0.0086}
\defval{posabs/x40/1024/yocou-noloop-minus-dense-noloop/longdocs/c256-512}{0.0068}
\defval{posabs/x40/1024/yocou-noloop-minus-dense-noloop/longdocs/c256-512/hi}{0.0076}
\defval{posabs/x40/1024/yocou-noloop-minus-dense-noloop/longdocs/c256-512/lo}{0.0060}
\defval{posabs/x40/1024/yocou-noloop-minus-dense-noloop/longdocs/c512-1024}{0.0039}
\defval{posabs/x40/1024/yocou-noloop-minus-dense-noloop/longdocs/c512-1024/hi}{0.0047}
\defval{posabs/x40/1024/yocou-noloop-minus-dense-noloop/longdocs/c512-1024/lo}{0.0031}
\defval{posabs/x40/1024/yocou-noloop-minus-dense-noloop/longdocs/mean}{0.0044}
\defval{posabs/x40/1024/yocou-noloop-minus-dense-noloop/longdocs/mean/hi}{0.0052}
\defval{posabs/x40/1024/yocou-noloop-minus-dense-noloop/longdocs/mean/lo}{0.0037}
\defval{posabs/x40/1024/yocou-noloop-minus-yocou-loop/longdocs/c1-128}{0.0252}
\defval{posabs/x40/1024/yocou-noloop-minus-yocou-loop/longdocs/c1-128/hi}{0.0260}
\defval{posabs/x40/1024/yocou-noloop-minus-yocou-loop/longdocs/c1-128/lo}{0.0245}
\defval{posabs/x40/1024/yocou-noloop-minus-yocou-loop/longdocs/c1024-2048}{0.0289}
\defval{posabs/x40/1024/yocou-noloop-minus-yocou-loop/longdocs/c1024-2048/hi}{0.0296}
\defval{posabs/x40/1024/yocou-noloop-minus-yocou-loop/longdocs/c1024-2048/lo}{0.0284}
\defval{posabs/x40/1024/yocou-noloop-minus-yocou-loop/longdocs/c128-256}{0.0291}
\defval{posabs/x40/1024/yocou-noloop-minus-yocou-loop/longdocs/c128-256/hi}{0.0299}
\defval{posabs/x40/1024/yocou-noloop-minus-yocou-loop/longdocs/c128-256/lo}{0.0284}
\defval{posabs/x40/1024/yocou-noloop-minus-yocou-loop/longdocs/c256-512}{0.0294}
\defval{posabs/x40/1024/yocou-noloop-minus-yocou-loop/longdocs/c256-512/hi}{0.0300}
\defval{posabs/x40/1024/yocou-noloop-minus-yocou-loop/longdocs/c256-512/lo}{0.0287}
\defval{posabs/x40/1024/yocou-noloop-minus-yocou-loop/longdocs/c512-1024}{0.0293}
\defval{posabs/x40/1024/yocou-noloop-minus-yocou-loop/longdocs/c512-1024/hi}{0.0299}
\defval{posabs/x40/1024/yocou-noloop-minus-yocou-loop/longdocs/c512-1024/lo}{0.0287}
\defval{posabs/x40/1024/yocou-noloop-minus-yocou-loop/longdocs/mean}{0.0289}
\defval{posabs/x40/1024/yocou-noloop-minus-yocou-loop/longdocs/mean/hi}{0.0294}
\defval{posabs/x40/1024/yocou-noloop-minus-yocou-loop/longdocs/mean/lo}{0.0284}
\defval{posdiff/long/1024/ced-dec-minus-ced-enc/finepdfs-long/c1-128}{+0.0141}
\defval{posdiff/long/1024/ced-dec-minus-ced-enc/finepdfs-long/c1-128/hi}{+0.0161}
\defval{posdiff/long/1024/ced-dec-minus-ced-enc/finepdfs-long/c1-128/lo}{+0.0122}
\defval{posdiff/long/1024/ced-dec-minus-ced-enc/finepdfs-long/c1024-2048}{+0.0324}
\defval{posdiff/long/1024/ced-dec-minus-ced-enc/finepdfs-long/c1024-2048/hi}{+0.0344}
\defval{posdiff/long/1024/ced-dec-minus-ced-enc/finepdfs-long/c1024-2048/lo}{+0.0304}
\defval{posdiff/long/1024/ced-dec-minus-ced-enc/finepdfs-long/c128-256}{+0.0281}
\defval{posdiff/long/1024/ced-dec-minus-ced-enc/finepdfs-long/c128-256/hi}{+0.0302}
\defval{posdiff/long/1024/ced-dec-minus-ced-enc/finepdfs-long/c128-256/lo}{+0.0258}
\defval{posdiff/long/1024/ced-dec-minus-ced-enc/finepdfs-long/c256-512}{+0.0307}
\defval{posdiff/long/1024/ced-dec-minus-ced-enc/finepdfs-long/c256-512/hi}{+0.0325}
\defval{posdiff/long/1024/ced-dec-minus-ced-enc/finepdfs-long/c256-512/lo}{+0.0289}
\defval{posdiff/long/1024/ced-dec-minus-ced-enc/finepdfs-long/c512-1024}{+0.0321}
\defval{posdiff/long/1024/ced-dec-minus-ced-enc/finepdfs-long/c512-1024/hi}{+0.0340}
\defval{posdiff/long/1024/ced-dec-minus-ced-enc/finepdfs-long/c512-1024/lo}{+0.0300}
\defval{posdiff/long/1024/ced-dec-minus-ced-enc/finepdfs-long/mean}{+0.0307}
\defval{posdiff/long/1024/ced-dec-minus-ced-enc/finepdfs-long/mean/hi}{+0.0324}
\defval{posdiff/long/1024/ced-dec-minus-ced-enc/finepdfs-long/mean/lo}{+0.0289}
\defval{posdiff/long/1024/ced-dec-minus-ced-enc/fineweb-edu-100bt/c1-128}{+0.0139}
\defval{posdiff/long/1024/ced-dec-minus-ced-enc/fineweb-edu-100bt/c1-128/hi}{+0.0150}
\defval{posdiff/long/1024/ced-dec-minus-ced-enc/fineweb-edu-100bt/c1-128/lo}{+0.0128}
\defval{posdiff/long/1024/ced-dec-minus-ced-enc/fineweb-edu-100bt/c1024-2048}{+0.0197}
\defval{posdiff/long/1024/ced-dec-minus-ced-enc/fineweb-edu-100bt/c1024-2048/hi}{+0.0208}
\defval{posdiff/long/1024/ced-dec-minus-ced-enc/fineweb-edu-100bt/c1024-2048/lo}{+0.0186}
\defval{posdiff/long/1024/ced-dec-minus-ced-enc/fineweb-edu-100bt/c128-256}{+0.0163}
\defval{posdiff/long/1024/ced-dec-minus-ced-enc/fineweb-edu-100bt/c128-256/hi}{+0.0175}
\defval{posdiff/long/1024/ced-dec-minus-ced-enc/fineweb-edu-100bt/c128-256/lo}{+0.0150}
\defval{posdiff/long/1024/ced-dec-minus-ced-enc/fineweb-edu-100bt/c256-512}{+0.0172}
\defval{posdiff/long/1024/ced-dec-minus-ced-enc/fineweb-edu-100bt/c256-512/hi}{+0.0183}
\defval{posdiff/long/1024/ced-dec-minus-ced-enc/fineweb-edu-100bt/c256-512/lo}{+0.0163}
\defval{posdiff/long/1024/ced-dec-minus-ced-enc/fineweb-edu-100bt/c512-1024}{+0.0181}
\defval{posdiff/long/1024/ced-dec-minus-ced-enc/fineweb-edu-100bt/c512-1024/hi}{+0.0191}
\defval{posdiff/long/1024/ced-dec-minus-ced-enc/fineweb-edu-100bt/c512-1024/lo}{+0.0171}
\defval{posdiff/long/1024/ced-dec-minus-ced-enc/fineweb-edu-100bt/mean}{+0.0173}
\defval{posdiff/long/1024/ced-dec-minus-ced-enc/fineweb-edu-100bt/mean/hi}{+0.0178}
\defval{posdiff/long/1024/ced-dec-minus-ced-enc/fineweb-edu-100bt/mean/lo}{+0.0167}
\defval{posdiff/long/1024/ced-dec-minus-ced-enc/govreport-test/c1-128}{+0.0118}
\defval{posdiff/long/1024/ced-dec-minus-ced-enc/govreport-test/c1-128/hi}{+0.0128}
\defval{posdiff/long/1024/ced-dec-minus-ced-enc/govreport-test/c1-128/lo}{+0.0109}
\defval{posdiff/long/1024/ced-dec-minus-ced-enc/govreport-test/c1024-2048}{+0.0195}
\defval{posdiff/long/1024/ced-dec-minus-ced-enc/govreport-test/c1024-2048/hi}{+0.0201}
\defval{posdiff/long/1024/ced-dec-minus-ced-enc/govreport-test/c1024-2048/lo}{+0.0190}
\defval{posdiff/long/1024/ced-dec-minus-ced-enc/govreport-test/c128-256}{+0.0153}
\defval{posdiff/long/1024/ced-dec-minus-ced-enc/govreport-test/c128-256/hi}{+0.0161}
\defval{posdiff/long/1024/ced-dec-minus-ced-enc/govreport-test/c128-256/lo}{+0.0145}
\defval{posdiff/long/1024/ced-dec-minus-ced-enc/govreport-test/c256-512}{+0.0172}
\defval{posdiff/long/1024/ced-dec-minus-ced-enc/govreport-test/c256-512/hi}{+0.0179}
\defval{posdiff/long/1024/ced-dec-minus-ced-enc/govreport-test/c256-512/lo}{+0.0166}
\defval{posdiff/long/1024/ced-dec-minus-ced-enc/govreport-test/c512-1024}{+0.0181}
\defval{posdiff/long/1024/ced-dec-minus-ced-enc/govreport-test/c512-1024/hi}{+0.0188}
\defval{posdiff/long/1024/ced-dec-minus-ced-enc/govreport-test/c512-1024/lo}{+0.0175}
\defval{posdiff/long/1024/ced-dec-minus-ced-enc/govreport-test/mean}{+0.0181}
\defval{posdiff/long/1024/ced-dec-minus-ced-enc/govreport-test/mean/hi}{+0.0186}
\defval{posdiff/long/1024/ced-dec-minus-ced-enc/govreport-test/mean/lo}{+0.0177}
\defval{posdiff/long/1024/ced-dec-minus-ced-enc/longdocs/c1-128}{+0.0116}
\defval{posdiff/long/1024/ced-dec-minus-ced-enc/longdocs/c1-128/hi}{+0.0125}
\defval{posdiff/long/1024/ced-dec-minus-ced-enc/longdocs/c1-128/lo}{+0.0107}
\defval{posdiff/long/1024/ced-dec-minus-ced-enc/longdocs/c1024-2048}{+0.0264}
\defval{posdiff/long/1024/ced-dec-minus-ced-enc/longdocs/c1024-2048/hi}{+0.0271}
\defval{posdiff/long/1024/ced-dec-minus-ced-enc/longdocs/c1024-2048/lo}{+0.0256}
\defval{posdiff/long/1024/ced-dec-minus-ced-enc/longdocs/c128-256}{+0.0229}
\defval{posdiff/long/1024/ced-dec-minus-ced-enc/longdocs/c128-256/hi}{+0.0238}
\defval{posdiff/long/1024/ced-dec-minus-ced-enc/longdocs/c128-256/lo}{+0.0220}
\defval{posdiff/long/1024/ced-dec-minus-ced-enc/longdocs/c256-512}{+0.0254}
\defval{posdiff/long/1024/ced-dec-minus-ced-enc/longdocs/c256-512/hi}{+0.0262}
\defval{posdiff/long/1024/ced-dec-minus-ced-enc/longdocs/c256-512/lo}{+0.0246}
\defval{posdiff/long/1024/ced-dec-minus-ced-enc/longdocs/c512-1024}{+0.0262}
\defval{posdiff/long/1024/ced-dec-minus-ced-enc/longdocs/c512-1024/hi}{+0.0270}
\defval{posdiff/long/1024/ced-dec-minus-ced-enc/longdocs/c512-1024/lo}{+0.0254}
\defval{posdiff/long/1024/ced-dec-minus-ced-enc/longdocs/mean}{+0.0251}
\defval{posdiff/long/1024/ced-dec-minus-ced-enc/longdocs/mean/hi}{+0.0257}
\defval{posdiff/long/1024/ced-dec-minus-ced-enc/longdocs/mean/lo}{+0.0244}
\defval{posdiff/long/1024/ced-dec-minus-ced-enc/pg19-test/c1-128}{+0.0088}
\defval{posdiff/long/1024/ced-dec-minus-ced-enc/pg19-test/c1-128/hi}{+0.0104}
\defval{posdiff/long/1024/ced-dec-minus-ced-enc/pg19-test/c1-128/lo}{+0.0073}
\defval{posdiff/long/1024/ced-dec-minus-ced-enc/pg19-test/c1024-2048}{+0.0275}
\defval{posdiff/long/1024/ced-dec-minus-ced-enc/pg19-test/c1024-2048/hi}{+0.0287}
\defval{posdiff/long/1024/ced-dec-minus-ced-enc/pg19-test/c1024-2048/lo}{+0.0264}
\defval{posdiff/long/1024/ced-dec-minus-ced-enc/pg19-test/c128-256}{+0.0252}
\defval{posdiff/long/1024/ced-dec-minus-ced-enc/pg19-test/c128-256/hi}{+0.0268}
\defval{posdiff/long/1024/ced-dec-minus-ced-enc/pg19-test/c128-256/lo}{+0.0236}
\defval{posdiff/long/1024/ced-dec-minus-ced-enc/pg19-test/c256-512}{+0.0283}
\defval{posdiff/long/1024/ced-dec-minus-ced-enc/pg19-test/c256-512/hi}{+0.0295}
\defval{posdiff/long/1024/ced-dec-minus-ced-enc/pg19-test/c256-512/lo}{+0.0270}
\defval{posdiff/long/1024/ced-dec-minus-ced-enc/pg19-test/c512-1024}{+0.0285}
\defval{posdiff/long/1024/ced-dec-minus-ced-enc/pg19-test/c512-1024/hi}{+0.0297}
\defval{posdiff/long/1024/ced-dec-minus-ced-enc/pg19-test/c512-1024/lo}{+0.0273}
\defval{posdiff/long/1024/ced-dec-minus-ced-enc/pg19-test/mean}{+0.0265}
\defval{posdiff/long/1024/ced-dec-minus-ced-enc/pg19-test/mean/hi}{+0.0275}
\defval{posdiff/long/1024/ced-dec-minus-ced-enc/pg19-test/mean/lo}{+0.0255}
\defval{posdiff/long/1024/ced-dec-minus-ced-enc/pg19-val/c1-128}{+0.0078}
\defval{posdiff/long/1024/ced-dec-minus-ced-enc/pg19-val/c1-128/hi}{+0.0106}
\defval{posdiff/long/1024/ced-dec-minus-ced-enc/pg19-val/c1-128/lo}{+0.0050}
\defval{posdiff/long/1024/ced-dec-minus-ced-enc/pg19-val/c1024-2048}{+0.0263}
\defval{posdiff/long/1024/ced-dec-minus-ced-enc/pg19-val/c1024-2048/hi}{+0.0275}
\defval{posdiff/long/1024/ced-dec-minus-ced-enc/pg19-val/c1024-2048/lo}{+0.0252}
\defval{posdiff/long/1024/ced-dec-minus-ced-enc/pg19-val/c128-256}{+0.0257}
\defval{posdiff/long/1024/ced-dec-minus-ced-enc/pg19-val/c128-256/hi}{+0.0284}
\defval{posdiff/long/1024/ced-dec-minus-ced-enc/pg19-val/c128-256/lo}{+0.0226}
\defval{posdiff/long/1024/ced-dec-minus-ced-enc/pg19-val/c256-512}{+0.0269}
\defval{posdiff/long/1024/ced-dec-minus-ced-enc/pg19-val/c256-512/hi}{+0.0290}
\defval{posdiff/long/1024/ced-dec-minus-ced-enc/pg19-val/c256-512/lo}{+0.0248}
\defval{posdiff/long/1024/ced-dec-minus-ced-enc/pg19-val/c512-1024}{+0.0273}
\defval{posdiff/long/1024/ced-dec-minus-ced-enc/pg19-val/c512-1024/hi}{+0.0287}
\defval{posdiff/long/1024/ced-dec-minus-ced-enc/pg19-val/c512-1024/lo}{+0.0259}
\defval{posdiff/long/1024/ced-dec-minus-ced-enc/pg19-val/mean}{+0.0255}
\defval{posdiff/long/1024/ced-dec-minus-ced-enc/pg19-val/mean/hi}{+0.0264}
\defval{posdiff/long/1024/ced-dec-minus-ced-enc/pg19-val/mean/lo}{+0.0246}
\defval{posdiff/long/1024/ced-dec-minus-dense-l47/finepdfs-long/c1-128}{+0.0028}
\defval{posdiff/long/1024/ced-dec-minus-dense-l47/finepdfs-long/c1-128/hi}{+0.0048}
\defval{posdiff/long/1024/ced-dec-minus-dense-l47/finepdfs-long/c1-128/lo}{+0.0008}
\defval{posdiff/long/1024/ced-dec-minus-dense-l47/finepdfs-long/c1024-2048}{+0.0197}
\defval{posdiff/long/1024/ced-dec-minus-dense-l47/finepdfs-long/c1024-2048/hi}{+0.0216}
\defval{posdiff/long/1024/ced-dec-minus-dense-l47/finepdfs-long/c1024-2048/lo}{+0.0176}
\defval{posdiff/long/1024/ced-dec-minus-dense-l47/finepdfs-long/c128-256}{+0.0131}
\defval{posdiff/long/1024/ced-dec-minus-dense-l47/finepdfs-long/c128-256/hi}{+0.0150}
\defval{posdiff/long/1024/ced-dec-minus-dense-l47/finepdfs-long/c128-256/lo}{+0.0112}
\defval{posdiff/long/1024/ced-dec-minus-dense-l47/finepdfs-long/c256-512}{+0.0160}
\defval{posdiff/long/1024/ced-dec-minus-dense-l47/finepdfs-long/c256-512/hi}{+0.0178}
\defval{posdiff/long/1024/ced-dec-minus-dense-l47/finepdfs-long/c256-512/lo}{+0.0141}
\defval{posdiff/long/1024/ced-dec-minus-dense-l47/finepdfs-long/c512-1024}{+0.0186}
\defval{posdiff/long/1024/ced-dec-minus-dense-l47/finepdfs-long/c512-1024/hi}{+0.0204}
\defval{posdiff/long/1024/ced-dec-minus-dense-l47/finepdfs-long/c512-1024/lo}{+0.0166}
\defval{posdiff/long/1024/ced-dec-minus-dense-l47/finepdfs-long/mean}{+0.0174}
\defval{posdiff/long/1024/ced-dec-minus-dense-l47/finepdfs-long/mean/hi}{+0.0192}
\defval{posdiff/long/1024/ced-dec-minus-dense-l47/finepdfs-long/mean/lo}{+0.0156}
\defval{posdiff/long/1024/ced-dec-minus-dense-l47/fineweb-edu-100bt/c1-128}{+0.0079}
\defval{posdiff/long/1024/ced-dec-minus-dense-l47/fineweb-edu-100bt/c1-128/hi}{+0.0090}
\defval{posdiff/long/1024/ced-dec-minus-dense-l47/fineweb-edu-100bt/c1-128/lo}{+0.0069}
\defval{posdiff/long/1024/ced-dec-minus-dense-l47/fineweb-edu-100bt/c1024-2048}{+0.0151}
\defval{posdiff/long/1024/ced-dec-minus-dense-l47/fineweb-edu-100bt/c1024-2048/hi}{+0.0163}
\defval{posdiff/long/1024/ced-dec-minus-dense-l47/fineweb-edu-100bt/c1024-2048/lo}{+0.0139}
\defval{posdiff/long/1024/ced-dec-minus-dense-l47/fineweb-edu-100bt/c128-256}{+0.0110}
\defval{posdiff/long/1024/ced-dec-minus-dense-l47/fineweb-edu-100bt/c128-256/hi}{+0.0121}
\defval{posdiff/long/1024/ced-dec-minus-dense-l47/fineweb-edu-100bt/c128-256/lo}{+0.0099}
\defval{posdiff/long/1024/ced-dec-minus-dense-l47/fineweb-edu-100bt/c256-512}{+0.0121}
\defval{posdiff/long/1024/ced-dec-minus-dense-l47/fineweb-edu-100bt/c256-512/hi}{+0.0131}
\defval{posdiff/long/1024/ced-dec-minus-dense-l47/fineweb-edu-100bt/c256-512/lo}{+0.0111}
\defval{posdiff/long/1024/ced-dec-minus-dense-l47/fineweb-edu-100bt/c512-1024}{+0.0125}
\defval{posdiff/long/1024/ced-dec-minus-dense-l47/fineweb-edu-100bt/c512-1024/hi}{+0.0135}
\defval{posdiff/long/1024/ced-dec-minus-dense-l47/fineweb-edu-100bt/c512-1024/lo}{+0.0116}
\defval{posdiff/long/1024/ced-dec-minus-dense-l47/fineweb-edu-100bt/mean}{+0.0120}
\defval{posdiff/long/1024/ced-dec-minus-dense-l47/fineweb-edu-100bt/mean/hi}{+0.0125}
\defval{posdiff/long/1024/ced-dec-minus-dense-l47/fineweb-edu-100bt/mean/lo}{+0.0115}
\defval{posdiff/long/1024/ced-dec-minus-dense-l47/govreport-test/c1-128}{+0.0071}
\defval{posdiff/long/1024/ced-dec-minus-dense-l47/govreport-test/c1-128/hi}{+0.0080}
\defval{posdiff/long/1024/ced-dec-minus-dense-l47/govreport-test/c1-128/lo}{+0.0063}
\defval{posdiff/long/1024/ced-dec-minus-dense-l47/govreport-test/c1024-2048}{+0.0169}
\defval{posdiff/long/1024/ced-dec-minus-dense-l47/govreport-test/c1024-2048/hi}{+0.0176}
\defval{posdiff/long/1024/ced-dec-minus-dense-l47/govreport-test/c1024-2048/lo}{+0.0164}
\defval{posdiff/long/1024/ced-dec-minus-dense-l47/govreport-test/c128-256}{+0.0106}
\defval{posdiff/long/1024/ced-dec-minus-dense-l47/govreport-test/c128-256/hi}{+0.0114}
\defval{posdiff/long/1024/ced-dec-minus-dense-l47/govreport-test/c128-256/lo}{+0.0097}
\defval{posdiff/long/1024/ced-dec-minus-dense-l47/govreport-test/c256-512}{+0.0133}
\defval{posdiff/long/1024/ced-dec-minus-dense-l47/govreport-test/c256-512/hi}{+0.0140}
\defval{posdiff/long/1024/ced-dec-minus-dense-l47/govreport-test/c256-512/lo}{+0.0126}
\defval{posdiff/long/1024/ced-dec-minus-dense-l47/govreport-test/c512-1024}{+0.0151}
\defval{posdiff/long/1024/ced-dec-minus-dense-l47/govreport-test/c512-1024/hi}{+0.0157}
\defval{posdiff/long/1024/ced-dec-minus-dense-l47/govreport-test/c512-1024/lo}{+0.0145}
\defval{posdiff/long/1024/ced-dec-minus-dense-l47/govreport-test/mean}{+0.0150}
\defval{posdiff/long/1024/ced-dec-minus-dense-l47/govreport-test/mean/hi}{+0.0155}
\defval{posdiff/long/1024/ced-dec-minus-dense-l47/govreport-test/mean/lo}{+0.0146}
\defval{posdiff/long/1024/ced-dec-minus-dense-l47/longdocs/c1-128}{+0.0027}
\defval{posdiff/long/1024/ced-dec-minus-dense-l47/longdocs/c1-128/hi}{+0.0036}
\defval{posdiff/long/1024/ced-dec-minus-dense-l47/longdocs/c1-128/lo}{+0.0018}
\defval{posdiff/long/1024/ced-dec-minus-dense-l47/longdocs/c1024-2048}{+0.0183}
\defval{posdiff/long/1024/ced-dec-minus-dense-l47/longdocs/c1024-2048/hi}{+0.0190}
\defval{posdiff/long/1024/ced-dec-minus-dense-l47/longdocs/c1024-2048/lo}{+0.0175}
\defval{posdiff/long/1024/ced-dec-minus-dense-l47/longdocs/c128-256}{+0.0113}
\defval{posdiff/long/1024/ced-dec-minus-dense-l47/longdocs/c128-256/hi}{+0.0121}
\defval{posdiff/long/1024/ced-dec-minus-dense-l47/longdocs/c128-256/lo}{+0.0104}
\defval{posdiff/long/1024/ced-dec-minus-dense-l47/longdocs/c256-512}{+0.0148}
\defval{posdiff/long/1024/ced-dec-minus-dense-l47/longdocs/c256-512/hi}{+0.0155}
\defval{posdiff/long/1024/ced-dec-minus-dense-l47/longdocs/c256-512/lo}{+0.0140}
\defval{posdiff/long/1024/ced-dec-minus-dense-l47/longdocs/c512-1024}{+0.0170}
\defval{posdiff/long/1024/ced-dec-minus-dense-l47/longdocs/c512-1024/hi}{+0.0177}
\defval{posdiff/long/1024/ced-dec-minus-dense-l47/longdocs/c512-1024/lo}{+0.0162}
\defval{posdiff/long/1024/ced-dec-minus-dense-l47/longdocs/mean}{+0.0161}
\defval{posdiff/long/1024/ced-dec-minus-dense-l47/longdocs/mean/hi}{+0.0167}
\defval{posdiff/long/1024/ced-dec-minus-dense-l47/longdocs/mean/lo}{+0.0154}
\defval{posdiff/long/1024/ced-dec-minus-dense-l47/pg19-test/c1-128}{$-$0.0018}
\defval{posdiff/long/1024/ced-dec-minus-dense-l47/pg19-test/c1-128/hi}{$-$0.0003}
\defval{posdiff/long/1024/ced-dec-minus-dense-l47/pg19-test/c1-128/lo}{$-$0.0032}
\defval{posdiff/long/1024/ced-dec-minus-dense-l47/pg19-test/c1024-2048}{+0.0183}
\defval{posdiff/long/1024/ced-dec-minus-dense-l47/pg19-test/c1024-2048/hi}{+0.0195}
\defval{posdiff/long/1024/ced-dec-minus-dense-l47/pg19-test/c1024-2048/lo}{+0.0171}
\defval{posdiff/long/1024/ced-dec-minus-dense-l47/pg19-test/c128-256}{+0.0101}
\defval{posdiff/long/1024/ced-dec-minus-dense-l47/pg19-test/c128-256/hi}{+0.0116}
\defval{posdiff/long/1024/ced-dec-minus-dense-l47/pg19-test/c128-256/lo}{+0.0086}
\defval{posdiff/long/1024/ced-dec-minus-dense-l47/pg19-test/c256-512}{+0.0151}
\defval{posdiff/long/1024/ced-dec-minus-dense-l47/pg19-test/c256-512/hi}{+0.0163}
\defval{posdiff/long/1024/ced-dec-minus-dense-l47/pg19-test/c256-512/lo}{+0.0138}
\defval{posdiff/long/1024/ced-dec-minus-dense-l47/pg19-test/c512-1024}{+0.0173}
\defval{posdiff/long/1024/ced-dec-minus-dense-l47/pg19-test/c512-1024/hi}{+0.0185}
\defval{posdiff/long/1024/ced-dec-minus-dense-l47/pg19-test/c512-1024/lo}{+0.0160}
\defval{posdiff/long/1024/ced-dec-minus-dense-l47/pg19-test/mean}{+0.0159}
\defval{posdiff/long/1024/ced-dec-minus-dense-l47/pg19-test/mean/hi}{+0.0169}
\defval{posdiff/long/1024/ced-dec-minus-dense-l47/pg19-test/mean/lo}{+0.0148}
\defval{posdiff/long/1024/ced-dec-minus-dense-l47/pg19-val/c1-128}{$-$0.0034}
\defval{posdiff/long/1024/ced-dec-minus-dense-l47/pg19-val/c1-128/hi}{$-$0.0007}
\defval{posdiff/long/1024/ced-dec-minus-dense-l47/pg19-val/c1-128/lo}{$-$0.0062}
\defval{posdiff/long/1024/ced-dec-minus-dense-l47/pg19-val/c1024-2048}{+0.0176}
\defval{posdiff/long/1024/ced-dec-minus-dense-l47/pg19-val/c1024-2048/hi}{+0.0188}
\defval{posdiff/long/1024/ced-dec-minus-dense-l47/pg19-val/c1024-2048/lo}{+0.0163}
\defval{posdiff/long/1024/ced-dec-minus-dense-l47/pg19-val/c128-256}{+0.0086}
\defval{posdiff/long/1024/ced-dec-minus-dense-l47/pg19-val/c128-256/hi}{+0.0115}
\defval{posdiff/long/1024/ced-dec-minus-dense-l47/pg19-val/c128-256/lo}{+0.0057}
\defval{posdiff/long/1024/ced-dec-minus-dense-l47/pg19-val/c256-512}{+0.0131}
\defval{posdiff/long/1024/ced-dec-minus-dense-l47/pg19-val/c256-512/hi}{+0.0153}
\defval{posdiff/long/1024/ced-dec-minus-dense-l47/pg19-val/c256-512/lo}{+0.0109}
\defval{posdiff/long/1024/ced-dec-minus-dense-l47/pg19-val/c512-1024}{+0.0167}
\defval{posdiff/long/1024/ced-dec-minus-dense-l47/pg19-val/c512-1024/hi}{+0.0182}
\defval{posdiff/long/1024/ced-dec-minus-dense-l47/pg19-val/c512-1024/lo}{+0.0151}
\defval{posdiff/long/1024/ced-dec-minus-dense-l47/pg19-val/mean}{+0.0149}
\defval{posdiff/long/1024/ced-dec-minus-dense-l47/pg19-val/mean/hi}{+0.0159}
\defval{posdiff/long/1024/ced-dec-minus-dense-l47/pg19-val/mean/lo}{+0.0139}
\defval{posdiff/long/1024/ced-enc-minus-dense-l47/finepdfs-long/c1-128}{$-$0.0113}
\defval{posdiff/long/1024/ced-enc-minus-dense-l47/finepdfs-long/c1-128/hi}{$-$0.0098}
\defval{posdiff/long/1024/ced-enc-minus-dense-l47/finepdfs-long/c1-128/lo}{$-$0.0131}
\defval{posdiff/long/1024/ced-enc-minus-dense-l47/finepdfs-long/c1024-2048}{$-$0.0127}
\defval{posdiff/long/1024/ced-enc-minus-dense-l47/finepdfs-long/c1024-2048/hi}{$-$0.0115}
\defval{posdiff/long/1024/ced-enc-minus-dense-l47/finepdfs-long/c1024-2048/lo}{$-$0.0141}
\defval{posdiff/long/1024/ced-enc-minus-dense-l47/finepdfs-long/c128-256}{$-$0.0151}
\defval{posdiff/long/1024/ced-enc-minus-dense-l47/finepdfs-long/c128-256/hi}{$-$0.0133}
\defval{posdiff/long/1024/ced-enc-minus-dense-l47/finepdfs-long/c128-256/lo}{$-$0.0166}
\defval{posdiff/long/1024/ced-enc-minus-dense-l47/finepdfs-long/c256-512}{$-$0.0148}
\defval{posdiff/long/1024/ced-enc-minus-dense-l47/finepdfs-long/c256-512/hi}{$-$0.0135}
\defval{posdiff/long/1024/ced-enc-minus-dense-l47/finepdfs-long/c256-512/lo}{$-$0.0161}
\defval{posdiff/long/1024/ced-enc-minus-dense-l47/finepdfs-long/c512-1024}{$-$0.0135}
\defval{posdiff/long/1024/ced-enc-minus-dense-l47/finepdfs-long/c512-1024/hi}{$-$0.0121}
\defval{posdiff/long/1024/ced-enc-minus-dense-l47/finepdfs-long/c512-1024/lo}{$-$0.0148}
\defval{posdiff/long/1024/ced-enc-minus-dense-l47/finepdfs-long/mean}{$-$0.0132}
\defval{posdiff/long/1024/ced-enc-minus-dense-l47/finepdfs-long/mean/hi}{$-$0.0121}
\defval{posdiff/long/1024/ced-enc-minus-dense-l47/finepdfs-long/mean/lo}{$-$0.0144}
\defval{posdiff/long/1024/ced-enc-minus-dense-l47/fineweb-edu-100bt/c1-128}{$-$0.0060}
\defval{posdiff/long/1024/ced-enc-minus-dense-l47/fineweb-edu-100bt/c1-128/hi}{$-$0.0049}
\defval{posdiff/long/1024/ced-enc-minus-dense-l47/fineweb-edu-100bt/c1-128/lo}{$-$0.0070}
\defval{posdiff/long/1024/ced-enc-minus-dense-l47/fineweb-edu-100bt/c1024-2048}{$-$0.0046}
\defval{posdiff/long/1024/ced-enc-minus-dense-l47/fineweb-edu-100bt/c1024-2048/hi}{$-$0.0036}
\defval{posdiff/long/1024/ced-enc-minus-dense-l47/fineweb-edu-100bt/c1024-2048/lo}{$-$0.0056}
\defval{posdiff/long/1024/ced-enc-minus-dense-l47/fineweb-edu-100bt/c128-256}{$-$0.0053}
\defval{posdiff/long/1024/ced-enc-minus-dense-l47/fineweb-edu-100bt/c128-256/hi}{$-$0.0041}
\defval{posdiff/long/1024/ced-enc-minus-dense-l47/fineweb-edu-100bt/c128-256/lo}{$-$0.0062}
\defval{posdiff/long/1024/ced-enc-minus-dense-l47/fineweb-edu-100bt/c256-512}{$-$0.0052}
\defval{posdiff/long/1024/ced-enc-minus-dense-l47/fineweb-edu-100bt/c256-512/hi}{$-$0.0043}
\defval{posdiff/long/1024/ced-enc-minus-dense-l47/fineweb-edu-100bt/c256-512/lo}{$-$0.0060}
\defval{posdiff/long/1024/ced-enc-minus-dense-l47/fineweb-edu-100bt/c512-1024}{$-$0.0056}
\defval{posdiff/long/1024/ced-enc-minus-dense-l47/fineweb-edu-100bt/c512-1024/hi}{$-$0.0048}
\defval{posdiff/long/1024/ced-enc-minus-dense-l47/fineweb-edu-100bt/c512-1024/lo}{$-$0.0065}
\defval{posdiff/long/1024/ced-enc-minus-dense-l47/fineweb-edu-100bt/mean}{$-$0.0053}
\defval{posdiff/long/1024/ced-enc-minus-dense-l47/fineweb-edu-100bt/mean/hi}{$-$0.0048}
\defval{posdiff/long/1024/ced-enc-minus-dense-l47/fineweb-edu-100bt/mean/lo}{$-$0.0057}
\defval{posdiff/long/1024/ced-enc-minus-dense-l47/govreport-test/c1-128}{$-$0.0047}
\defval{posdiff/long/1024/ced-enc-minus-dense-l47/govreport-test/c1-128/hi}{$-$0.0040}
\defval{posdiff/long/1024/ced-enc-minus-dense-l47/govreport-test/c1-128/lo}{$-$0.0055}
\defval{posdiff/long/1024/ced-enc-minus-dense-l47/govreport-test/c1024-2048}{$-$0.0026}
\defval{posdiff/long/1024/ced-enc-minus-dense-l47/govreport-test/c1024-2048/hi}{$-$0.0021}
\defval{posdiff/long/1024/ced-enc-minus-dense-l47/govreport-test/c1024-2048/lo}{$-$0.0030}
\defval{posdiff/long/1024/ced-enc-minus-dense-l47/govreport-test/c128-256}{$-$0.0048}
\defval{posdiff/long/1024/ced-enc-minus-dense-l47/govreport-test/c128-256/hi}{$-$0.0040}
\defval{posdiff/long/1024/ced-enc-minus-dense-l47/govreport-test/c128-256/lo}{$-$0.0055}
\defval{posdiff/long/1024/ced-enc-minus-dense-l47/govreport-test/c256-512}{$-$0.0039}
\defval{posdiff/long/1024/ced-enc-minus-dense-l47/govreport-test/c256-512/hi}{$-$0.0033}
\defval{posdiff/long/1024/ced-enc-minus-dense-l47/govreport-test/c256-512/lo}{$-$0.0045}
\defval{posdiff/long/1024/ced-enc-minus-dense-l47/govreport-test/c512-1024}{$-$0.0030}
\defval{posdiff/long/1024/ced-enc-minus-dense-l47/govreport-test/c512-1024/hi}{$-$0.0025}
\defval{posdiff/long/1024/ced-enc-minus-dense-l47/govreport-test/c512-1024/lo}{$-$0.0035}
\defval{posdiff/long/1024/ced-enc-minus-dense-l47/govreport-test/mean}{$-$0.0031}
\defval{posdiff/long/1024/ced-enc-minus-dense-l47/govreport-test/mean/hi}{$-$0.0027}
\defval{posdiff/long/1024/ced-enc-minus-dense-l47/govreport-test/mean/lo}{$-$0.0035}
\defval{posdiff/long/1024/ced-enc-minus-dense-l47/longdocs/c1-128}{$-$0.0089}
\defval{posdiff/long/1024/ced-enc-minus-dense-l47/longdocs/c1-128/hi}{$-$0.0081}
\defval{posdiff/long/1024/ced-enc-minus-dense-l47/longdocs/c1-128/lo}{$-$0.0097}
\defval{posdiff/long/1024/ced-enc-minus-dense-l47/longdocs/c1024-2048}{$-$0.0081}
\defval{posdiff/long/1024/ced-enc-minus-dense-l47/longdocs/c1024-2048/hi}{$-$0.0076}
\defval{posdiff/long/1024/ced-enc-minus-dense-l47/longdocs/c1024-2048/lo}{$-$0.0086}
\defval{posdiff/long/1024/ced-enc-minus-dense-l47/longdocs/c128-256}{$-$0.0116}
\defval{posdiff/long/1024/ced-enc-minus-dense-l47/longdocs/c128-256/hi}{$-$0.0109}
\defval{posdiff/long/1024/ced-enc-minus-dense-l47/longdocs/c128-256/lo}{$-$0.0123}
\defval{posdiff/long/1024/ced-enc-minus-dense-l47/longdocs/c256-512}{$-$0.0106}
\defval{posdiff/long/1024/ced-enc-minus-dense-l47/longdocs/c256-512/hi}{$-$0.0100}
\defval{posdiff/long/1024/ced-enc-minus-dense-l47/longdocs/c256-512/lo}{$-$0.0112}
\defval{posdiff/long/1024/ced-enc-minus-dense-l47/longdocs/c512-1024}{$-$0.0092}
\defval{posdiff/long/1024/ced-enc-minus-dense-l47/longdocs/c512-1024/hi}{$-$0.0086}
\defval{posdiff/long/1024/ced-enc-minus-dense-l47/longdocs/c512-1024/lo}{$-$0.0097}
\defval{posdiff/long/1024/ced-enc-minus-dense-l47/longdocs/mean}{$-$0.0090}
\defval{posdiff/long/1024/ced-enc-minus-dense-l47/longdocs/mean/hi}{$-$0.0085}
\defval{posdiff/long/1024/ced-enc-minus-dense-l47/longdocs/mean/lo}{$-$0.0094}
\defval{posdiff/long/1024/ced-enc-minus-dense-l47/pg19-test/c1-128}{$-$0.0106}
\defval{posdiff/long/1024/ced-enc-minus-dense-l47/pg19-test/c1-128/hi}{$-$0.0093}
\defval{posdiff/long/1024/ced-enc-minus-dense-l47/pg19-test/c1-128/lo}{$-$0.0120}
\defval{posdiff/long/1024/ced-enc-minus-dense-l47/pg19-test/c1024-2048}{$-$0.0092}
\defval{posdiff/long/1024/ced-enc-minus-dense-l47/pg19-test/c1024-2048/hi}{$-$0.0082}
\defval{posdiff/long/1024/ced-enc-minus-dense-l47/pg19-test/c1024-2048/lo}{$-$0.0103}
\defval{posdiff/long/1024/ced-enc-minus-dense-l47/pg19-test/c128-256}{$-$0.0151}
\defval{posdiff/long/1024/ced-enc-minus-dense-l47/pg19-test/c128-256/hi}{$-$0.0137}
\defval{posdiff/long/1024/ced-enc-minus-dense-l47/pg19-test/c128-256/lo}{$-$0.0165}
\defval{posdiff/long/1024/ced-enc-minus-dense-l47/pg19-test/c256-512}{$-$0.0132}
\defval{posdiff/long/1024/ced-enc-minus-dense-l47/pg19-test/c256-512/hi}{$-$0.0121}
\defval{posdiff/long/1024/ced-enc-minus-dense-l47/pg19-test/c256-512/lo}{$-$0.0143}
\defval{posdiff/long/1024/ced-enc-minus-dense-l47/pg19-test/c512-1024}{$-$0.0112}
\defval{posdiff/long/1024/ced-enc-minus-dense-l47/pg19-test/c512-1024/hi}{$-$0.0101}
\defval{posdiff/long/1024/ced-enc-minus-dense-l47/pg19-test/c512-1024/lo}{$-$0.0123}
\defval{posdiff/long/1024/ced-enc-minus-dense-l47/pg19-test/mean}{$-$0.0107}
\defval{posdiff/long/1024/ced-enc-minus-dense-l47/pg19-test/mean/hi}{$-$0.0098}
\defval{posdiff/long/1024/ced-enc-minus-dense-l47/pg19-test/mean/lo}{$-$0.0115}
\defval{posdiff/long/1024/ced-enc-minus-dense-l47/pg19-val/c1-128}{$-$0.0112}
\defval{posdiff/long/1024/ced-enc-minus-dense-l47/pg19-val/c1-128/hi}{$-$0.0085}
\defval{posdiff/long/1024/ced-enc-minus-dense-l47/pg19-val/c1-128/lo}{$-$0.0139}
\defval{posdiff/long/1024/ced-enc-minus-dense-l47/pg19-val/c1024-2048}{$-$0.0087}
\defval{posdiff/long/1024/ced-enc-minus-dense-l47/pg19-val/c1024-2048/hi}{$-$0.0078}
\defval{posdiff/long/1024/ced-enc-minus-dense-l47/pg19-val/c1024-2048/lo}{$-$0.0099}
\defval{posdiff/long/1024/ced-enc-minus-dense-l47/pg19-val/c128-256}{$-$0.0171}
\defval{posdiff/long/1024/ced-enc-minus-dense-l47/pg19-val/c128-256/hi}{$-$0.0143}
\defval{posdiff/long/1024/ced-enc-minus-dense-l47/pg19-val/c128-256/lo}{$-$0.0197}
\defval{posdiff/long/1024/ced-enc-minus-dense-l47/pg19-val/c256-512}{$-$0.0138}
\defval{posdiff/long/1024/ced-enc-minus-dense-l47/pg19-val/c256-512/hi}{$-$0.0119}
\defval{posdiff/long/1024/ced-enc-minus-dense-l47/pg19-val/c256-512/lo}{$-$0.0157}
\defval{posdiff/long/1024/ced-enc-minus-dense-l47/pg19-val/c512-1024}{$-$0.0107}
\defval{posdiff/long/1024/ced-enc-minus-dense-l47/pg19-val/c512-1024/hi}{$-$0.0092}
\defval{posdiff/long/1024/ced-enc-minus-dense-l47/pg19-val/c512-1024/lo}{$-$0.0120}
\defval{posdiff/long/1024/ced-enc-minus-dense-l47/pg19-val/mean}{$-$0.0105}
\defval{posdiff/long/1024/ced-enc-minus-dense-l47/pg19-val/mean/hi}{$-$0.0097}
\defval{posdiff/long/1024/ced-enc-minus-dense-l47/pg19-val/mean/lo}{$-$0.0114}
\defval{posdiff/long/1024/ced-noloop-minus-ced-dec/finepdfs-long/c1-128}{+0.0012}
\defval{posdiff/long/1024/ced-noloop-minus-ced-dec/finepdfs-long/c1-128/hi}{+0.0026}
\defval{posdiff/long/1024/ced-noloop-minus-ced-dec/finepdfs-long/c1-128/lo}{$-$0.0003}
\defval{posdiff/long/1024/ced-noloop-minus-ced-dec/finepdfs-long/c1024-2048}{+0.0130}
\defval{posdiff/long/1024/ced-noloop-minus-ced-dec/finepdfs-long/c1024-2048/hi}{+0.0142}
\defval{posdiff/long/1024/ced-noloop-minus-ced-dec/finepdfs-long/c1024-2048/lo}{+0.0118}
\defval{posdiff/long/1024/ced-noloop-minus-ced-dec/finepdfs-long/c128-256}{+0.0087}
\defval{posdiff/long/1024/ced-noloop-minus-ced-dec/finepdfs-long/c128-256/hi}{+0.0101}
\defval{posdiff/long/1024/ced-noloop-minus-ced-dec/finepdfs-long/c128-256/lo}{+0.0073}
\defval{posdiff/long/1024/ced-noloop-minus-ced-dec/finepdfs-long/c256-512}{+0.0102}
\defval{posdiff/long/1024/ced-noloop-minus-ced-dec/finepdfs-long/c256-512/hi}{+0.0113}
\defval{posdiff/long/1024/ced-noloop-minus-ced-dec/finepdfs-long/c256-512/lo}{+0.0090}
\defval{posdiff/long/1024/ced-noloop-minus-ced-dec/finepdfs-long/c512-1024}{+0.0116}
\defval{posdiff/long/1024/ced-noloop-minus-ced-dec/finepdfs-long/c512-1024/hi}{+0.0128}
\defval{posdiff/long/1024/ced-noloop-minus-ced-dec/finepdfs-long/c512-1024/lo}{+0.0105}
\defval{posdiff/long/1024/ced-noloop-minus-ced-dec/finepdfs-long/mean}{+0.0113}
\defval{posdiff/long/1024/ced-noloop-minus-ced-dec/finepdfs-long/mean/hi}{+0.0123}
\defval{posdiff/long/1024/ced-noloop-minus-ced-dec/finepdfs-long/mean/lo}{+0.0103}
\defval{posdiff/long/1024/ced-noloop-minus-ced-dec/fineweb-edu-100bt/c1-128}{+0.0105}
\defval{posdiff/long/1024/ced-noloop-minus-ced-dec/fineweb-edu-100bt/c1-128/hi}{+0.0115}
\defval{posdiff/long/1024/ced-noloop-minus-ced-dec/fineweb-edu-100bt/c1-128/lo}{+0.0094}
\defval{posdiff/long/1024/ced-noloop-minus-ced-dec/fineweb-edu-100bt/c1024-2048}{+0.0115}
\defval{posdiff/long/1024/ced-noloop-minus-ced-dec/fineweb-edu-100bt/c1024-2048/hi}{+0.0124}
\defval{posdiff/long/1024/ced-noloop-minus-ced-dec/fineweb-edu-100bt/c1024-2048/lo}{+0.0105}
\defval{posdiff/long/1024/ced-noloop-minus-ced-dec/fineweb-edu-100bt/c128-256}{+0.0106}
\defval{posdiff/long/1024/ced-noloop-minus-ced-dec/fineweb-edu-100bt/c128-256/hi}{+0.0117}
\defval{posdiff/long/1024/ced-noloop-minus-ced-dec/fineweb-edu-100bt/c128-256/lo}{+0.0096}
\defval{posdiff/long/1024/ced-noloop-minus-ced-dec/fineweb-edu-100bt/c256-512}{+0.0108}
\defval{posdiff/long/1024/ced-noloop-minus-ced-dec/fineweb-edu-100bt/c256-512/hi}{+0.0116}
\defval{posdiff/long/1024/ced-noloop-minus-ced-dec/fineweb-edu-100bt/c256-512/lo}{+0.0098}
\defval{posdiff/long/1024/ced-noloop-minus-ced-dec/fineweb-edu-100bt/c512-1024}{+0.0107}
\defval{posdiff/long/1024/ced-noloop-minus-ced-dec/fineweb-edu-100bt/c512-1024/hi}{+0.0117}
\defval{posdiff/long/1024/ced-noloop-minus-ced-dec/fineweb-edu-100bt/c512-1024/lo}{+0.0098}
\defval{posdiff/long/1024/ced-noloop-minus-ced-dec/fineweb-edu-100bt/mean}{+0.0108}
\defval{posdiff/long/1024/ced-noloop-minus-ced-dec/fineweb-edu-100bt/mean/hi}{+0.0113}
\defval{posdiff/long/1024/ced-noloop-minus-ced-dec/fineweb-edu-100bt/mean/lo}{+0.0104}
\defval{posdiff/long/1024/ced-noloop-minus-ced-dec/govreport-test/c1-128}{+0.0072}
\defval{posdiff/long/1024/ced-noloop-minus-ced-dec/govreport-test/c1-128/hi}{+0.0080}
\defval{posdiff/long/1024/ced-noloop-minus-ced-dec/govreport-test/c1-128/lo}{+0.0063}
\defval{posdiff/long/1024/ced-noloop-minus-ced-dec/govreport-test/c1024-2048}{+0.0086}
\defval{posdiff/long/1024/ced-noloop-minus-ced-dec/govreport-test/c1024-2048/hi}{+0.0091}
\defval{posdiff/long/1024/ced-noloop-minus-ced-dec/govreport-test/c1024-2048/lo}{+0.0081}
\defval{posdiff/long/1024/ced-noloop-minus-ced-dec/govreport-test/c128-256}{+0.0078}
\defval{posdiff/long/1024/ced-noloop-minus-ced-dec/govreport-test/c128-256/hi}{+0.0086}
\defval{posdiff/long/1024/ced-noloop-minus-ced-dec/govreport-test/c128-256/lo}{+0.0070}
\defval{posdiff/long/1024/ced-noloop-minus-ced-dec/govreport-test/c256-512}{+0.0081}
\defval{posdiff/long/1024/ced-noloop-minus-ced-dec/govreport-test/c256-512/hi}{+0.0087}
\defval{posdiff/long/1024/ced-noloop-minus-ced-dec/govreport-test/c256-512/lo}{+0.0074}
\defval{posdiff/long/1024/ced-noloop-minus-ced-dec/govreport-test/c512-1024}{+0.0088}
\defval{posdiff/long/1024/ced-noloop-minus-ced-dec/govreport-test/c512-1024/hi}{+0.0093}
\defval{posdiff/long/1024/ced-noloop-minus-ced-dec/govreport-test/c512-1024/lo}{+0.0082}
\defval{posdiff/long/1024/ced-noloop-minus-ced-dec/govreport-test/mean}{+0.0084}
\defval{posdiff/long/1024/ced-noloop-minus-ced-dec/govreport-test/mean/hi}{+0.0088}
\defval{posdiff/long/1024/ced-noloop-minus-ced-dec/govreport-test/mean/lo}{+0.0080}
\defval{posdiff/long/1024/ced-noloop-minus-ced-dec/longdocs/c1-128}{+0.0063}
\defval{posdiff/long/1024/ced-noloop-minus-ced-dec/longdocs/c1-128/hi}{+0.0070}
\defval{posdiff/long/1024/ced-noloop-minus-ced-dec/longdocs/c1-128/lo}{+0.0056}
\defval{posdiff/long/1024/ced-noloop-minus-ced-dec/longdocs/c1024-2048}{+0.0135}
\defval{posdiff/long/1024/ced-noloop-minus-ced-dec/longdocs/c1024-2048/hi}{+0.0140}
\defval{posdiff/long/1024/ced-noloop-minus-ced-dec/longdocs/c1024-2048/lo}{+0.0130}
\defval{posdiff/long/1024/ced-noloop-minus-ced-dec/longdocs/c128-256}{+0.0103}
\defval{posdiff/long/1024/ced-noloop-minus-ced-dec/longdocs/c128-256/hi}{+0.0110}
\defval{posdiff/long/1024/ced-noloop-minus-ced-dec/longdocs/c128-256/lo}{+0.0096}
\defval{posdiff/long/1024/ced-noloop-minus-ced-dec/longdocs/c256-512}{+0.0117}
\defval{posdiff/long/1024/ced-noloop-minus-ced-dec/longdocs/c256-512/hi}{+0.0123}
\defval{posdiff/long/1024/ced-noloop-minus-ced-dec/longdocs/c256-512/lo}{+0.0112}
\defval{posdiff/long/1024/ced-noloop-minus-ced-dec/longdocs/c512-1024}{+0.0130}
\defval{posdiff/long/1024/ced-noloop-minus-ced-dec/longdocs/c512-1024/hi}{+0.0135}
\defval{posdiff/long/1024/ced-noloop-minus-ced-dec/longdocs/c512-1024/lo}{+0.0125}
\defval{posdiff/long/1024/ced-noloop-minus-ced-dec/longdocs/mean}{+0.0125}
\defval{posdiff/long/1024/ced-noloop-minus-ced-dec/longdocs/mean/hi}{+0.0129}
\defval{posdiff/long/1024/ced-noloop-minus-ced-dec/longdocs/mean/lo}{+0.0120}
\defval{posdiff/long/1024/ced-noloop-minus-ced-dec/pg19-test/c1-128}{+0.0104}
\defval{posdiff/long/1024/ced-noloop-minus-ced-dec/pg19-test/c1-128/hi}{+0.0117}
\defval{posdiff/long/1024/ced-noloop-minus-ced-dec/pg19-test/c1-128/lo}{+0.0091}
\defval{posdiff/long/1024/ced-noloop-minus-ced-dec/pg19-test/c1024-2048}{+0.0189}
\defval{posdiff/long/1024/ced-noloop-minus-ced-dec/pg19-test/c1024-2048/hi}{+0.0198}
\defval{posdiff/long/1024/ced-noloop-minus-ced-dec/pg19-test/c1024-2048/lo}{+0.0180}
\defval{posdiff/long/1024/ced-noloop-minus-ced-dec/pg19-test/c128-256}{+0.0144}
\defval{posdiff/long/1024/ced-noloop-minus-ced-dec/pg19-test/c128-256/hi}{+0.0157}
\defval{posdiff/long/1024/ced-noloop-minus-ced-dec/pg19-test/c128-256/lo}{+0.0132}
\defval{posdiff/long/1024/ced-noloop-minus-ced-dec/pg19-test/c256-512}{+0.0168}
\defval{posdiff/long/1024/ced-noloop-minus-ced-dec/pg19-test/c256-512/hi}{+0.0177}
\defval{posdiff/long/1024/ced-noloop-minus-ced-dec/pg19-test/c256-512/lo}{+0.0159}
\defval{posdiff/long/1024/ced-noloop-minus-ced-dec/pg19-test/c512-1024}{+0.0185}
\defval{posdiff/long/1024/ced-noloop-minus-ced-dec/pg19-test/c512-1024/hi}{+0.0195}
\defval{posdiff/long/1024/ced-noloop-minus-ced-dec/pg19-test/c512-1024/lo}{+0.0176}
\defval{posdiff/long/1024/ced-noloop-minus-ced-dec/pg19-test/mean}{+0.0177}
\defval{posdiff/long/1024/ced-noloop-minus-ced-dec/pg19-test/mean/hi}{+0.0185}
\defval{posdiff/long/1024/ced-noloop-minus-ced-dec/pg19-test/mean/lo}{+0.0170}
\defval{posdiff/long/1024/ced-noloop-minus-ced-dec/pg19-val/c1-128}{+0.0085}
\defval{posdiff/long/1024/ced-noloop-minus-ced-dec/pg19-val/c1-128/hi}{+0.0112}
\defval{posdiff/long/1024/ced-noloop-minus-ced-dec/pg19-val/c1-128/lo}{+0.0061}
\defval{posdiff/long/1024/ced-noloop-minus-ced-dec/pg19-val/c1024-2048}{+0.0178}
\defval{posdiff/long/1024/ced-noloop-minus-ced-dec/pg19-val/c1024-2048/hi}{+0.0189}
\defval{posdiff/long/1024/ced-noloop-minus-ced-dec/pg19-val/c1024-2048/lo}{+0.0167}
\defval{posdiff/long/1024/ced-noloop-minus-ced-dec/pg19-val/c128-256}{+0.0136}
\defval{posdiff/long/1024/ced-noloop-minus-ced-dec/pg19-val/c128-256/hi}{+0.0159}
\defval{posdiff/long/1024/ced-noloop-minus-ced-dec/pg19-val/c128-256/lo}{+0.0111}
\defval{posdiff/long/1024/ced-noloop-minus-ced-dec/pg19-val/c256-512}{+0.0154}
\defval{posdiff/long/1024/ced-noloop-minus-ced-dec/pg19-val/c256-512/hi}{+0.0171}
\defval{posdiff/long/1024/ced-noloop-minus-ced-dec/pg19-val/c256-512/lo}{+0.0137}
\defval{posdiff/long/1024/ced-noloop-minus-ced-dec/pg19-val/c512-1024}{+0.0171}
\defval{posdiff/long/1024/ced-noloop-minus-ced-dec/pg19-val/c512-1024/hi}{+0.0184}
\defval{posdiff/long/1024/ced-noloop-minus-ced-dec/pg19-val/c512-1024/lo}{+0.0160}
\defval{posdiff/long/1024/ced-noloop-minus-ced-dec/pg19-val/mean}{+0.0165}
\defval{posdiff/long/1024/ced-noloop-minus-ced-dec/pg19-val/mean/hi}{+0.0173}
\defval{posdiff/long/1024/ced-noloop-minus-ced-dec/pg19-val/mean/lo}{+0.0157}
\defval{posdiff/long/1024/ced-noloop-minus-ced-enc/finepdfs-long/c1-128}{+0.0153}
\defval{posdiff/long/1024/ced-noloop-minus-ced-enc/finepdfs-long/c1-128/hi}{+0.0175}
\defval{posdiff/long/1024/ced-noloop-minus-ced-enc/finepdfs-long/c1-128/lo}{+0.0133}
\defval{posdiff/long/1024/ced-noloop-minus-ced-enc/finepdfs-long/c1024-2048}{+0.0453}
\defval{posdiff/long/1024/ced-noloop-minus-ced-enc/finepdfs-long/c1024-2048/hi}{+0.0477}
\defval{posdiff/long/1024/ced-noloop-minus-ced-enc/finepdfs-long/c1024-2048/lo}{+0.0427}
\defval{posdiff/long/1024/ced-noloop-minus-ced-enc/finepdfs-long/c128-256}{+0.0369}
\defval{posdiff/long/1024/ced-noloop-minus-ced-enc/finepdfs-long/c128-256/hi}{+0.0391}
\defval{posdiff/long/1024/ced-noloop-minus-ced-enc/finepdfs-long/c128-256/lo}{+0.0347}
\defval{posdiff/long/1024/ced-noloop-minus-ced-enc/finepdfs-long/c256-512}{+0.0409}
\defval{posdiff/long/1024/ced-noloop-minus-ced-enc/finepdfs-long/c256-512/hi}{+0.0430}
\defval{posdiff/long/1024/ced-noloop-minus-ced-enc/finepdfs-long/c256-512/lo}{+0.0389}
\defval{posdiff/long/1024/ced-noloop-minus-ced-enc/finepdfs-long/c512-1024}{+0.0437}
\defval{posdiff/long/1024/ced-noloop-minus-ced-enc/finepdfs-long/c512-1024/hi}{+0.0458}
\defval{posdiff/long/1024/ced-noloop-minus-ced-enc/finepdfs-long/c512-1024/lo}{+0.0414}
\defval{posdiff/long/1024/ced-noloop-minus-ced-enc/finepdfs-long/mean}{+0.0419}
\defval{posdiff/long/1024/ced-noloop-minus-ced-enc/finepdfs-long/mean/hi}{+0.0439}
\defval{posdiff/long/1024/ced-noloop-minus-ced-enc/finepdfs-long/mean/lo}{+0.0398}
\defval{posdiff/long/1024/ced-noloop-minus-ced-enc/fineweb-edu-100bt/c1-128}{+0.0244}
\defval{posdiff/long/1024/ced-noloop-minus-ced-enc/fineweb-edu-100bt/c1-128/hi}{+0.0256}
\defval{posdiff/long/1024/ced-noloop-minus-ced-enc/fineweb-edu-100bt/c1-128/lo}{+0.0232}
\defval{posdiff/long/1024/ced-noloop-minus-ced-enc/fineweb-edu-100bt/c1024-2048}{+0.0311}
\defval{posdiff/long/1024/ced-noloop-minus-ced-enc/fineweb-edu-100bt/c1024-2048/hi}{+0.0324}
\defval{posdiff/long/1024/ced-noloop-minus-ced-enc/fineweb-edu-100bt/c1024-2048/lo}{+0.0299}
\defval{posdiff/long/1024/ced-noloop-minus-ced-enc/fineweb-edu-100bt/c128-256}{+0.0269}
\defval{posdiff/long/1024/ced-noloop-minus-ced-enc/fineweb-edu-100bt/c128-256/hi}{+0.0281}
\defval{posdiff/long/1024/ced-noloop-minus-ced-enc/fineweb-edu-100bt/c128-256/lo}{+0.0256}
\defval{posdiff/long/1024/ced-noloop-minus-ced-enc/fineweb-edu-100bt/c256-512}{+0.0280}
\defval{posdiff/long/1024/ced-noloop-minus-ced-enc/fineweb-edu-100bt/c256-512/hi}{+0.0292}
\defval{posdiff/long/1024/ced-noloop-minus-ced-enc/fineweb-edu-100bt/c256-512/lo}{+0.0269}
\defval{posdiff/long/1024/ced-noloop-minus-ced-enc/fineweb-edu-100bt/c512-1024}{+0.0288}
\defval{posdiff/long/1024/ced-noloop-minus-ced-enc/fineweb-edu-100bt/c512-1024/hi}{+0.0299}
\defval{posdiff/long/1024/ced-noloop-minus-ced-enc/fineweb-edu-100bt/c512-1024/lo}{+0.0277}
\defval{posdiff/long/1024/ced-noloop-minus-ced-enc/fineweb-edu-100bt/mean}{+0.0281}
\defval{posdiff/long/1024/ced-noloop-minus-ced-enc/fineweb-edu-100bt/mean/hi}{+0.0288}
\defval{posdiff/long/1024/ced-noloop-minus-ced-enc/fineweb-edu-100bt/mean/lo}{+0.0274}
\defval{posdiff/long/1024/ced-noloop-minus-ced-enc/govreport-test/c1-128}{+0.0190}
\defval{posdiff/long/1024/ced-noloop-minus-ced-enc/govreport-test/c1-128/hi}{+0.0198}
\defval{posdiff/long/1024/ced-noloop-minus-ced-enc/govreport-test/c1-128/lo}{+0.0181}
\defval{posdiff/long/1024/ced-noloop-minus-ced-enc/govreport-test/c1024-2048}{+0.0281}
\defval{posdiff/long/1024/ced-noloop-minus-ced-enc/govreport-test/c1024-2048/hi}{+0.0288}
\defval{posdiff/long/1024/ced-noloop-minus-ced-enc/govreport-test/c1024-2048/lo}{+0.0274}
\defval{posdiff/long/1024/ced-noloop-minus-ced-enc/govreport-test/c128-256}{+0.0231}
\defval{posdiff/long/1024/ced-noloop-minus-ced-enc/govreport-test/c128-256/hi}{+0.0241}
\defval{posdiff/long/1024/ced-noloop-minus-ced-enc/govreport-test/c128-256/lo}{+0.0222}
\defval{posdiff/long/1024/ced-noloop-minus-ced-enc/govreport-test/c256-512}{+0.0253}
\defval{posdiff/long/1024/ced-noloop-minus-ced-enc/govreport-test/c256-512/hi}{+0.0260}
\defval{posdiff/long/1024/ced-noloop-minus-ced-enc/govreport-test/c256-512/lo}{+0.0246}
\defval{posdiff/long/1024/ced-noloop-minus-ced-enc/govreport-test/c512-1024}{+0.0269}
\defval{posdiff/long/1024/ced-noloop-minus-ced-enc/govreport-test/c512-1024/hi}{+0.0275}
\defval{posdiff/long/1024/ced-noloop-minus-ced-enc/govreport-test/c512-1024/lo}{+0.0261}
\defval{posdiff/long/1024/ced-noloop-minus-ced-enc/govreport-test/mean}{+0.0265}
\defval{posdiff/long/1024/ced-noloop-minus-ced-enc/govreport-test/mean/hi}{+0.0271}
\defval{posdiff/long/1024/ced-noloop-minus-ced-enc/govreport-test/mean/lo}{+0.0260}
\defval{posdiff/long/1024/ced-noloop-minus-ced-enc/longdocs/c1-128}{+0.0179}
\defval{posdiff/long/1024/ced-noloop-minus-ced-enc/longdocs/c1-128/hi}{+0.0189}
\defval{posdiff/long/1024/ced-noloop-minus-ced-enc/longdocs/c1-128/lo}{+0.0169}
\defval{posdiff/long/1024/ced-noloop-minus-ced-enc/longdocs/c1024-2048}{+0.0399}
\defval{posdiff/long/1024/ced-noloop-minus-ced-enc/longdocs/c1024-2048/hi}{+0.0407}
\defval{posdiff/long/1024/ced-noloop-minus-ced-enc/longdocs/c1024-2048/lo}{+0.0390}
\defval{posdiff/long/1024/ced-noloop-minus-ced-enc/longdocs/c128-256}{+0.0332}
\defval{posdiff/long/1024/ced-noloop-minus-ced-enc/longdocs/c128-256/hi}{+0.0341}
\defval{posdiff/long/1024/ced-noloop-minus-ced-enc/longdocs/c128-256/lo}{+0.0323}
\defval{posdiff/long/1024/ced-noloop-minus-ced-enc/longdocs/c256-512}{+0.0371}
\defval{posdiff/long/1024/ced-noloop-minus-ced-enc/longdocs/c256-512/hi}{+0.0379}
\defval{posdiff/long/1024/ced-noloop-minus-ced-enc/longdocs/c256-512/lo}{+0.0362}
\defval{posdiff/long/1024/ced-noloop-minus-ced-enc/longdocs/c512-1024}{+0.0391}
\defval{posdiff/long/1024/ced-noloop-minus-ced-enc/longdocs/c512-1024/hi}{+0.0400}
\defval{posdiff/long/1024/ced-noloop-minus-ced-enc/longdocs/c512-1024/lo}{+0.0383}
\defval{posdiff/long/1024/ced-noloop-minus-ced-enc/longdocs/mean}{+0.0376}
\defval{posdiff/long/1024/ced-noloop-minus-ced-enc/longdocs/mean/hi}{+0.0383}
\defval{posdiff/long/1024/ced-noloop-minus-ced-enc/longdocs/mean/lo}{+0.0368}
\defval{posdiff/long/1024/ced-noloop-minus-ced-enc/pg19-test/c1-128}{+0.0192}
\defval{posdiff/long/1024/ced-noloop-minus-ced-enc/pg19-test/c1-128/hi}{+0.0209}
\defval{posdiff/long/1024/ced-noloop-minus-ced-enc/pg19-test/c1-128/lo}{+0.0175}
\defval{posdiff/long/1024/ced-noloop-minus-ced-enc/pg19-test/c1024-2048}{+0.0464}
\defval{posdiff/long/1024/ced-noloop-minus-ced-enc/pg19-test/c1024-2048/hi}{+0.0477}
\defval{posdiff/long/1024/ced-noloop-minus-ced-enc/pg19-test/c1024-2048/lo}{+0.0452}
\defval{posdiff/long/1024/ced-noloop-minus-ced-enc/pg19-test/c128-256}{+0.0396}
\defval{posdiff/long/1024/ced-noloop-minus-ced-enc/pg19-test/c128-256/hi}{+0.0413}
\defval{posdiff/long/1024/ced-noloop-minus-ced-enc/pg19-test/c128-256/lo}{+0.0381}
\defval{posdiff/long/1024/ced-noloop-minus-ced-enc/pg19-test/c256-512}{+0.0451}
\defval{posdiff/long/1024/ced-noloop-minus-ced-enc/pg19-test/c256-512/hi}{+0.0465}
\defval{posdiff/long/1024/ced-noloop-minus-ced-enc/pg19-test/c256-512/lo}{+0.0437}
\defval{posdiff/long/1024/ced-noloop-minus-ced-enc/pg19-test/c512-1024}{+0.0469}
\defval{posdiff/long/1024/ced-noloop-minus-ced-enc/pg19-test/c512-1024/hi}{+0.0483}
\defval{posdiff/long/1024/ced-noloop-minus-ced-enc/pg19-test/c512-1024/lo}{+0.0457}
\defval{posdiff/long/1024/ced-noloop-minus-ced-enc/pg19-test/mean}{+0.0443}
\defval{posdiff/long/1024/ced-noloop-minus-ced-enc/pg19-test/mean/hi}{+0.0455}
\defval{posdiff/long/1024/ced-noloop-minus-ced-enc/pg19-test/mean/lo}{+0.0432}
\defval{posdiff/long/1024/ced-noloop-minus-ced-enc/pg19-val/c1-128}{+0.0163}
\defval{posdiff/long/1024/ced-noloop-minus-ced-enc/pg19-val/c1-128/hi}{+0.0194}
\defval{posdiff/long/1024/ced-noloop-minus-ced-enc/pg19-val/c1-128/lo}{+0.0135}
\defval{posdiff/long/1024/ced-noloop-minus-ced-enc/pg19-val/c1024-2048}{+0.0441}
\defval{posdiff/long/1024/ced-noloop-minus-ced-enc/pg19-val/c1024-2048/hi}{+0.0454}
\defval{posdiff/long/1024/ced-noloop-minus-ced-enc/pg19-val/c1024-2048/lo}{+0.0429}
\defval{posdiff/long/1024/ced-noloop-minus-ced-enc/pg19-val/c128-256}{+0.0392}
\defval{posdiff/long/1024/ced-noloop-minus-ced-enc/pg19-val/c128-256/hi}{+0.0421}
\defval{posdiff/long/1024/ced-noloop-minus-ced-enc/pg19-val/c128-256/lo}{+0.0363}
\defval{posdiff/long/1024/ced-noloop-minus-ced-enc/pg19-val/c256-512}{+0.0423}
\defval{posdiff/long/1024/ced-noloop-minus-ced-enc/pg19-val/c256-512/hi}{+0.0444}
\defval{posdiff/long/1024/ced-noloop-minus-ced-enc/pg19-val/c256-512/lo}{+0.0401}
\defval{posdiff/long/1024/ced-noloop-minus-ced-enc/pg19-val/c512-1024}{+0.0445}
\defval{posdiff/long/1024/ced-noloop-minus-ced-enc/pg19-val/c512-1024/hi}{+0.0460}
\defval{posdiff/long/1024/ced-noloop-minus-ced-enc/pg19-val/c512-1024/lo}{+0.0432}
\defval{posdiff/long/1024/ced-noloop-minus-ced-enc/pg19-val/mean}{+0.0419}
\defval{posdiff/long/1024/ced-noloop-minus-ced-enc/pg19-val/mean/hi}{+0.0429}
\defval{posdiff/long/1024/ced-noloop-minus-ced-enc/pg19-val/mean/lo}{+0.0410}
\defval{posdiff/long/1024/ced-noloop-minus-dense-l47/finepdfs-long/c1-128}{+0.0040}
\defval{posdiff/long/1024/ced-noloop-minus-dense-l47/finepdfs-long/c1-128/hi}{+0.0061}
\defval{posdiff/long/1024/ced-noloop-minus-dense-l47/finepdfs-long/c1-128/lo}{+0.0018}
\defval{posdiff/long/1024/ced-noloop-minus-dense-l47/finepdfs-long/c1024-2048}{+0.0326}
\defval{posdiff/long/1024/ced-noloop-minus-dense-l47/finepdfs-long/c1024-2048/hi}{+0.0345}
\defval{posdiff/long/1024/ced-noloop-minus-dense-l47/finepdfs-long/c1024-2048/lo}{+0.0305}
\defval{posdiff/long/1024/ced-noloop-minus-dense-l47/finepdfs-long/c128-256}{+0.0218}
\defval{posdiff/long/1024/ced-noloop-minus-dense-l47/finepdfs-long/c128-256/hi}{+0.0239}
\defval{posdiff/long/1024/ced-noloop-minus-dense-l47/finepdfs-long/c128-256/lo}{+0.0199}
\defval{posdiff/long/1024/ced-noloop-minus-dense-l47/finepdfs-long/c256-512}{+0.0262}
\defval{posdiff/long/1024/ced-noloop-minus-dense-l47/finepdfs-long/c256-512/hi}{+0.0280}
\defval{posdiff/long/1024/ced-noloop-minus-dense-l47/finepdfs-long/c256-512/lo}{+0.0244}
\defval{posdiff/long/1024/ced-noloop-minus-dense-l47/finepdfs-long/c512-1024}{+0.0302}
\defval{posdiff/long/1024/ced-noloop-minus-dense-l47/finepdfs-long/c512-1024/hi}{+0.0319}
\defval{posdiff/long/1024/ced-noloop-minus-dense-l47/finepdfs-long/c512-1024/lo}{+0.0283}
\defval{posdiff/long/1024/ced-noloop-minus-dense-l47/finepdfs-long/mean}{+0.0287}
\defval{posdiff/long/1024/ced-noloop-minus-dense-l47/finepdfs-long/mean/hi}{+0.0304}
\defval{posdiff/long/1024/ced-noloop-minus-dense-l47/finepdfs-long/mean/lo}{+0.0270}
\defval{posdiff/long/1024/ced-noloop-minus-dense-l47/fineweb-edu-100bt/c1-128}{+0.0184}
\defval{posdiff/long/1024/ced-noloop-minus-dense-l47/fineweb-edu-100bt/c1-128/hi}{+0.0196}
\defval{posdiff/long/1024/ced-noloop-minus-dense-l47/fineweb-edu-100bt/c1-128/lo}{+0.0174}
\defval{posdiff/long/1024/ced-noloop-minus-dense-l47/fineweb-edu-100bt/c1024-2048}{+0.0265}
\defval{posdiff/long/1024/ced-noloop-minus-dense-l47/fineweb-edu-100bt/c1024-2048/hi}{+0.0278}
\defval{posdiff/long/1024/ced-noloop-minus-dense-l47/fineweb-edu-100bt/c1024-2048/lo}{+0.0253}
\defval{posdiff/long/1024/ced-noloop-minus-dense-l47/fineweb-edu-100bt/c128-256}{+0.0216}
\defval{posdiff/long/1024/ced-noloop-minus-dense-l47/fineweb-edu-100bt/c128-256/hi}{+0.0229}
\defval{posdiff/long/1024/ced-noloop-minus-dense-l47/fineweb-edu-100bt/c128-256/lo}{+0.0204}
\defval{posdiff/long/1024/ced-noloop-minus-dense-l47/fineweb-edu-100bt/c256-512}{+0.0228}
\defval{posdiff/long/1024/ced-noloop-minus-dense-l47/fineweb-edu-100bt/c256-512/hi}{+0.0239}
\defval{posdiff/long/1024/ced-noloop-minus-dense-l47/fineweb-edu-100bt/c256-512/lo}{+0.0217}
\defval{posdiff/long/1024/ced-noloop-minus-dense-l47/fineweb-edu-100bt/c512-1024}{+0.0232}
\defval{posdiff/long/1024/ced-noloop-minus-dense-l47/fineweb-edu-100bt/c512-1024/hi}{+0.0243}
\defval{posdiff/long/1024/ced-noloop-minus-dense-l47/fineweb-edu-100bt/c512-1024/lo}{+0.0222}
\defval{posdiff/long/1024/ced-noloop-minus-dense-l47/fineweb-edu-100bt/mean}{+0.0228}
\defval{posdiff/long/1024/ced-noloop-minus-dense-l47/fineweb-edu-100bt/mean/hi}{+0.0234}
\defval{posdiff/long/1024/ced-noloop-minus-dense-l47/fineweb-edu-100bt/mean/lo}{+0.0222}
\defval{posdiff/long/1024/ced-noloop-minus-dense-l47/govreport-test/c1-128}{+0.0143}
\defval{posdiff/long/1024/ced-noloop-minus-dense-l47/govreport-test/c1-128/hi}{+0.0151}
\defval{posdiff/long/1024/ced-noloop-minus-dense-l47/govreport-test/c1-128/lo}{+0.0134}
\defval{posdiff/long/1024/ced-noloop-minus-dense-l47/govreport-test/c1024-2048}{+0.0255}
\defval{posdiff/long/1024/ced-noloop-minus-dense-l47/govreport-test/c1024-2048/hi}{+0.0262}
\defval{posdiff/long/1024/ced-noloop-minus-dense-l47/govreport-test/c1024-2048/lo}{+0.0249}
\defval{posdiff/long/1024/ced-noloop-minus-dense-l47/govreport-test/c128-256}{+0.0184}
\defval{posdiff/long/1024/ced-noloop-minus-dense-l47/govreport-test/c128-256/hi}{+0.0193}
\defval{posdiff/long/1024/ced-noloop-minus-dense-l47/govreport-test/c128-256/lo}{+0.0175}
\defval{posdiff/long/1024/ced-noloop-minus-dense-l47/govreport-test/c256-512}{+0.0214}
\defval{posdiff/long/1024/ced-noloop-minus-dense-l47/govreport-test/c256-512/hi}{+0.0221}
\defval{posdiff/long/1024/ced-noloop-minus-dense-l47/govreport-test/c256-512/lo}{+0.0206}
\defval{posdiff/long/1024/ced-noloop-minus-dense-l47/govreport-test/c512-1024}{+0.0238}
\defval{posdiff/long/1024/ced-noloop-minus-dense-l47/govreport-test/c512-1024/hi}{+0.0245}
\defval{posdiff/long/1024/ced-noloop-minus-dense-l47/govreport-test/c512-1024/lo}{+0.0232}
\defval{posdiff/long/1024/ced-noloop-minus-dense-l47/govreport-test/mean}{+0.0234}
\defval{posdiff/long/1024/ced-noloop-minus-dense-l47/govreport-test/mean/hi}{+0.0240}
\defval{posdiff/long/1024/ced-noloop-minus-dense-l47/govreport-test/mean/lo}{+0.0229}
\defval{posdiff/long/1024/ced-noloop-minus-dense-l47/longdocs/c1-128}{+0.0090}
\defval{posdiff/long/1024/ced-noloop-minus-dense-l47/longdocs/c1-128/hi}{+0.0099}
\defval{posdiff/long/1024/ced-noloop-minus-dense-l47/longdocs/c1-128/lo}{+0.0081}
\defval{posdiff/long/1024/ced-noloop-minus-dense-l47/longdocs/c1024-2048}{+0.0318}
\defval{posdiff/long/1024/ced-noloop-minus-dense-l47/longdocs/c1024-2048/hi}{+0.0326}
\defval{posdiff/long/1024/ced-noloop-minus-dense-l47/longdocs/c1024-2048/lo}{+0.0310}
\defval{posdiff/long/1024/ced-noloop-minus-dense-l47/longdocs/c128-256}{+0.0216}
\defval{posdiff/long/1024/ced-noloop-minus-dense-l47/longdocs/c128-256/hi}{+0.0224}
\defval{posdiff/long/1024/ced-noloop-minus-dense-l47/longdocs/c128-256/lo}{+0.0207}
\defval{posdiff/long/1024/ced-noloop-minus-dense-l47/longdocs/c256-512}{+0.0265}
\defval{posdiff/long/1024/ced-noloop-minus-dense-l47/longdocs/c256-512/hi}{+0.0273}
\defval{posdiff/long/1024/ced-noloop-minus-dense-l47/longdocs/c256-512/lo}{+0.0257}
\defval{posdiff/long/1024/ced-noloop-minus-dense-l47/longdocs/c512-1024}{+0.0299}
\defval{posdiff/long/1024/ced-noloop-minus-dense-l47/longdocs/c512-1024/hi}{+0.0307}
\defval{posdiff/long/1024/ced-noloop-minus-dense-l47/longdocs/c512-1024/lo}{+0.0292}
\defval{posdiff/long/1024/ced-noloop-minus-dense-l47/longdocs/mean}{+0.0286}
\defval{posdiff/long/1024/ced-noloop-minus-dense-l47/longdocs/mean/hi}{+0.0292}
\defval{posdiff/long/1024/ced-noloop-minus-dense-l47/longdocs/mean/lo}{+0.0279}
\defval{posdiff/long/1024/ced-noloop-minus-dense-l47/pg19-test/c1-128}{+0.0086}
\defval{posdiff/long/1024/ced-noloop-minus-dense-l47/pg19-test/c1-128/hi}{+0.0102}
\defval{posdiff/long/1024/ced-noloop-minus-dense-l47/pg19-test/c1-128/lo}{+0.0071}
\defval{posdiff/long/1024/ced-noloop-minus-dense-l47/pg19-test/c1024-2048}{+0.0372}
\defval{posdiff/long/1024/ced-noloop-minus-dense-l47/pg19-test/c1024-2048/hi}{+0.0386}
\defval{posdiff/long/1024/ced-noloop-minus-dense-l47/pg19-test/c1024-2048/lo}{+0.0358}
\defval{posdiff/long/1024/ced-noloop-minus-dense-l47/pg19-test/c128-256}{+0.0245}
\defval{posdiff/long/1024/ced-noloop-minus-dense-l47/pg19-test/c128-256/hi}{+0.0261}
\defval{posdiff/long/1024/ced-noloop-minus-dense-l47/pg19-test/c128-256/lo}{+0.0230}
\defval{posdiff/long/1024/ced-noloop-minus-dense-l47/pg19-test/c256-512}{+0.0319}
\defval{posdiff/long/1024/ced-noloop-minus-dense-l47/pg19-test/c256-512/hi}{+0.0333}
\defval{posdiff/long/1024/ced-noloop-minus-dense-l47/pg19-test/c256-512/lo}{+0.0305}
\defval{posdiff/long/1024/ced-noloop-minus-dense-l47/pg19-test/c512-1024}{+0.0358}
\defval{posdiff/long/1024/ced-noloop-minus-dense-l47/pg19-test/c512-1024/hi}{+0.0371}
\defval{posdiff/long/1024/ced-noloop-minus-dense-l47/pg19-test/c512-1024/lo}{+0.0345}
\defval{posdiff/long/1024/ced-noloop-minus-dense-l47/pg19-test/mean}{+0.0336}
\defval{posdiff/long/1024/ced-noloop-minus-dense-l47/pg19-test/mean/hi}{+0.0348}
\defval{posdiff/long/1024/ced-noloop-minus-dense-l47/pg19-test/mean/lo}{+0.0324}
\defval{posdiff/long/1024/ced-noloop-minus-dense-l47/pg19-val/c1-128}{+0.0051}
\defval{posdiff/long/1024/ced-noloop-minus-dense-l47/pg19-val/c1-128/hi}{+0.0082}
\defval{posdiff/long/1024/ced-noloop-minus-dense-l47/pg19-val/c1-128/lo}{+0.0023}
\defval{posdiff/long/1024/ced-noloop-minus-dense-l47/pg19-val/c1024-2048}{+0.0354}
\defval{posdiff/long/1024/ced-noloop-minus-dense-l47/pg19-val/c1024-2048/hi}{+0.0367}
\defval{posdiff/long/1024/ced-noloop-minus-dense-l47/pg19-val/c1024-2048/lo}{+0.0340}
\defval{posdiff/long/1024/ced-noloop-minus-dense-l47/pg19-val/c128-256}{+0.0221}
\defval{posdiff/long/1024/ced-noloop-minus-dense-l47/pg19-val/c128-256/hi}{+0.0249}
\defval{posdiff/long/1024/ced-noloop-minus-dense-l47/pg19-val/c128-256/lo}{+0.0193}
\defval{posdiff/long/1024/ced-noloop-minus-dense-l47/pg19-val/c256-512}{+0.0285}
\defval{posdiff/long/1024/ced-noloop-minus-dense-l47/pg19-val/c256-512/hi}{+0.0307}
\defval{posdiff/long/1024/ced-noloop-minus-dense-l47/pg19-val/c256-512/lo}{+0.0262}
\defval{posdiff/long/1024/ced-noloop-minus-dense-l47/pg19-val/c512-1024}{+0.0338}
\defval{posdiff/long/1024/ced-noloop-minus-dense-l47/pg19-val/c512-1024/hi}{+0.0355}
\defval{posdiff/long/1024/ced-noloop-minus-dense-l47/pg19-val/c512-1024/lo}{+0.0323}
\defval{posdiff/long/1024/ced-noloop-minus-dense-l47/pg19-val/mean}{+0.0314}
\defval{posdiff/long/1024/ced-noloop-minus-dense-l47/pg19-val/mean/hi}{+0.0325}
\defval{posdiff/long/1024/ced-noloop-minus-dense-l47/pg19-val/mean/lo}{+0.0303}
\defval{posdiff/p140/768/p140-dec-fixed-minus-p140-enc-fixed/finepdfs-long/c1-128}{+0.0100}
\defval{posdiff/p140/768/p140-dec-fixed-minus-p140-enc-fixed/finepdfs-long/c1-128/hi}{+0.0116}
\defval{posdiff/p140/768/p140-dec-fixed-minus-p140-enc-fixed/finepdfs-long/c1-128/lo}{+0.0083}
\defval{posdiff/p140/768/p140-dec-fixed-minus-p140-enc-fixed/finepdfs-long/c1024-2048}{+0.0327}
\defval{posdiff/p140/768/p140-dec-fixed-minus-p140-enc-fixed/finepdfs-long/c1024-2048/hi}{+0.0347}
\defval{posdiff/p140/768/p140-dec-fixed-minus-p140-enc-fixed/finepdfs-long/c1024-2048/lo}{+0.0308}
\defval{posdiff/p140/768/p140-dec-fixed-minus-p140-enc-fixed/finepdfs-long/c128-256}{+0.0182}
\defval{posdiff/p140/768/p140-dec-fixed-minus-p140-enc-fixed/finepdfs-long/c128-256/hi}{+0.0199}
\defval{posdiff/p140/768/p140-dec-fixed-minus-p140-enc-fixed/finepdfs-long/c128-256/lo}{+0.0167}
\defval{posdiff/p140/768/p140-dec-fixed-minus-p140-enc-fixed/finepdfs-long/c256-512}{+0.0243}
\defval{posdiff/p140/768/p140-dec-fixed-minus-p140-enc-fixed/finepdfs-long/c256-512/hi}{+0.0259}
\defval{posdiff/p140/768/p140-dec-fixed-minus-p140-enc-fixed/finepdfs-long/c256-512/lo}{+0.0228}
\defval{posdiff/p140/768/p140-dec-fixed-minus-p140-enc-fixed/finepdfs-long/c512-1024}{+0.0296}
\defval{posdiff/p140/768/p140-dec-fixed-minus-p140-enc-fixed/finepdfs-long/c512-1024/hi}{+0.0315}
\defval{posdiff/p140/768/p140-dec-fixed-minus-p140-enc-fixed/finepdfs-long/c512-1024/lo}{+0.0278}
\defval{posdiff/p140/768/p140-dec-fixed-minus-p140-enc-fixed/finepdfs-long/mean}{+0.0285}
\defval{posdiff/p140/768/p140-dec-fixed-minus-p140-enc-fixed/finepdfs-long/mean/hi}{+0.0301}
\defval{posdiff/p140/768/p140-dec-fixed-minus-p140-enc-fixed/finepdfs-long/mean/lo}{+0.0270}
\defval{posdiff/p140/768/p140-dec-fixed-minus-p140-enc-fixed/fineweb-edu-100bt/c1-128}{+0.0061}
\defval{posdiff/p140/768/p140-dec-fixed-minus-p140-enc-fixed/fineweb-edu-100bt/c1-128/hi}{+0.0071}
\defval{posdiff/p140/768/p140-dec-fixed-minus-p140-enc-fixed/fineweb-edu-100bt/c1-128/lo}{+0.0050}
\defval{posdiff/p140/768/p140-dec-fixed-minus-p140-enc-fixed/fineweb-edu-100bt/c1024-2048}{+0.0121}
\defval{posdiff/p140/768/p140-dec-fixed-minus-p140-enc-fixed/fineweb-edu-100bt/c1024-2048/hi}{+0.0133}
\defval{posdiff/p140/768/p140-dec-fixed-minus-p140-enc-fixed/fineweb-edu-100bt/c1024-2048/lo}{+0.0111}
\defval{posdiff/p140/768/p140-dec-fixed-minus-p140-enc-fixed/fineweb-edu-100bt/c128-256}{+0.0087}
\defval{posdiff/p140/768/p140-dec-fixed-minus-p140-enc-fixed/fineweb-edu-100bt/c128-256/hi}{+0.0098}
\defval{posdiff/p140/768/p140-dec-fixed-minus-p140-enc-fixed/fineweb-edu-100bt/c128-256/lo}{+0.0076}
\defval{posdiff/p140/768/p140-dec-fixed-minus-p140-enc-fixed/fineweb-edu-100bt/c256-512}{+0.0091}
\defval{posdiff/p140/768/p140-dec-fixed-minus-p140-enc-fixed/fineweb-edu-100bt/c256-512/hi}{+0.0102}
\defval{posdiff/p140/768/p140-dec-fixed-minus-p140-enc-fixed/fineweb-edu-100bt/c256-512/lo}{+0.0080}
\defval{posdiff/p140/768/p140-dec-fixed-minus-p140-enc-fixed/fineweb-edu-100bt/c512-1024}{+0.0097}
\defval{posdiff/p140/768/p140-dec-fixed-minus-p140-enc-fixed/fineweb-edu-100bt/c512-1024/hi}{+0.0107}
\defval{posdiff/p140/768/p140-dec-fixed-minus-p140-enc-fixed/fineweb-edu-100bt/c512-1024/lo}{+0.0087}
\defval{posdiff/p140/768/p140-dec-fixed-minus-p140-enc-fixed/fineweb-edu-100bt/mean}{+0.0093}
\defval{posdiff/p140/768/p140-dec-fixed-minus-p140-enc-fixed/fineweb-edu-100bt/mean/hi}{+0.0099}
\defval{posdiff/p140/768/p140-dec-fixed-minus-p140-enc-fixed/fineweb-edu-100bt/mean/lo}{+0.0088}
\defval{posdiff/p140/768/p140-dec-fixed-minus-p140-enc-fixed/govreport-test/c1-128}{+0.0050}
\defval{posdiff/p140/768/p140-dec-fixed-minus-p140-enc-fixed/govreport-test/c1-128/hi}{+0.0059}
\defval{posdiff/p140/768/p140-dec-fixed-minus-p140-enc-fixed/govreport-test/c1-128/lo}{+0.0042}
\defval{posdiff/p140/768/p140-dec-fixed-minus-p140-enc-fixed/govreport-test/c1024-2048}{+0.0156}
\defval{posdiff/p140/768/p140-dec-fixed-minus-p140-enc-fixed/govreport-test/c1024-2048/hi}{+0.0164}
\defval{posdiff/p140/768/p140-dec-fixed-minus-p140-enc-fixed/govreport-test/c1024-2048/lo}{+0.0149}
\defval{posdiff/p140/768/p140-dec-fixed-minus-p140-enc-fixed/govreport-test/c128-256}{+0.0080}
\defval{posdiff/p140/768/p140-dec-fixed-minus-p140-enc-fixed/govreport-test/c128-256/hi}{+0.0089}
\defval{posdiff/p140/768/p140-dec-fixed-minus-p140-enc-fixed/govreport-test/c128-256/lo}{+0.0072}
\defval{posdiff/p140/768/p140-dec-fixed-minus-p140-enc-fixed/govreport-test/c256-512}{+0.0109}
\defval{posdiff/p140/768/p140-dec-fixed-minus-p140-enc-fixed/govreport-test/c256-512/hi}{+0.0117}
\defval{posdiff/p140/768/p140-dec-fixed-minus-p140-enc-fixed/govreport-test/c256-512/lo}{+0.0102}
\defval{posdiff/p140/768/p140-dec-fixed-minus-p140-enc-fixed/govreport-test/c512-1024}{+0.0136}
\defval{posdiff/p140/768/p140-dec-fixed-minus-p140-enc-fixed/govreport-test/c512-1024/hi}{+0.0143}
\defval{posdiff/p140/768/p140-dec-fixed-minus-p140-enc-fixed/govreport-test/c512-1024/lo}{+0.0128}
\defval{posdiff/p140/768/p140-dec-fixed-minus-p140-enc-fixed/govreport-test/mean}{+0.0134}
\defval{posdiff/p140/768/p140-dec-fixed-minus-p140-enc-fixed/govreport-test/mean/hi}{+0.0140}
\defval{posdiff/p140/768/p140-dec-fixed-minus-p140-enc-fixed/govreport-test/mean/lo}{+0.0128}
\defval{posdiff/p140/768/p140-dec-fixed-minus-p140-enc-fixed/longdocs/c1-128}{+0.0071}
\defval{posdiff/p140/768/p140-dec-fixed-minus-p140-enc-fixed/longdocs/c1-128/hi}{+0.0079}
\defval{posdiff/p140/768/p140-dec-fixed-minus-p140-enc-fixed/longdocs/c1-128/lo}{+0.0063}
\defval{posdiff/p140/768/p140-dec-fixed-minus-p140-enc-fixed/longdocs/c1024-2048}{+0.0221}
\defval{posdiff/p140/768/p140-dec-fixed-minus-p140-enc-fixed/longdocs/c1024-2048/hi}{+0.0229}
\defval{posdiff/p140/768/p140-dec-fixed-minus-p140-enc-fixed/longdocs/c1024-2048/lo}{+0.0214}
\defval{posdiff/p140/768/p140-dec-fixed-minus-p140-enc-fixed/longdocs/c128-256}{+0.0123}
\defval{posdiff/p140/768/p140-dec-fixed-minus-p140-enc-fixed/longdocs/c128-256/hi}{+0.0131}
\defval{posdiff/p140/768/p140-dec-fixed-minus-p140-enc-fixed/longdocs/c128-256/lo}{+0.0116}
\defval{posdiff/p140/768/p140-dec-fixed-minus-p140-enc-fixed/longdocs/c256-512}{+0.0167}
\defval{posdiff/p140/768/p140-dec-fixed-minus-p140-enc-fixed/longdocs/c256-512/hi}{+0.0173}
\defval{posdiff/p140/768/p140-dec-fixed-minus-p140-enc-fixed/longdocs/c256-512/lo}{+0.0160}
\defval{posdiff/p140/768/p140-dec-fixed-minus-p140-enc-fixed/longdocs/c512-1024}{+0.0200}
\defval{posdiff/p140/768/p140-dec-fixed-minus-p140-enc-fixed/longdocs/c512-1024/hi}{+0.0208}
\defval{posdiff/p140/768/p140-dec-fixed-minus-p140-enc-fixed/longdocs/c512-1024/lo}{+0.0193}
\defval{posdiff/p140/768/p140-dec-fixed-minus-p140-enc-fixed/longdocs/mean}{+0.0194}
\defval{posdiff/p140/768/p140-dec-fixed-minus-p140-enc-fixed/longdocs/mean/hi}{+0.0200}
\defval{posdiff/p140/768/p140-dec-fixed-minus-p140-enc-fixed/longdocs/mean/lo}{+0.0188}
\defval{posdiff/p140/768/p140-dec-fixed-minus-p140-enc-fixed/pg19-test/c1-128}{+0.0063}
\defval{posdiff/p140/768/p140-dec-fixed-minus-p140-enc-fixed/pg19-test/c1-128/hi}{+0.0076}
\defval{posdiff/p140/768/p140-dec-fixed-minus-p140-enc-fixed/pg19-test/c1-128/lo}{+0.0051}
\defval{posdiff/p140/768/p140-dec-fixed-minus-p140-enc-fixed/pg19-test/c1024-2048}{+0.0184}
\defval{posdiff/p140/768/p140-dec-fixed-minus-p140-enc-fixed/pg19-test/c1024-2048/hi}{+0.0196}
\defval{posdiff/p140/768/p140-dec-fixed-minus-p140-enc-fixed/pg19-test/c1024-2048/lo}{+0.0172}
\defval{posdiff/p140/768/p140-dec-fixed-minus-p140-enc-fixed/pg19-test/c128-256}{+0.0108}
\defval{posdiff/p140/768/p140-dec-fixed-minus-p140-enc-fixed/pg19-test/c128-256/hi}{+0.0122}
\defval{posdiff/p140/768/p140-dec-fixed-minus-p140-enc-fixed/pg19-test/c128-256/lo}{+0.0094}
\defval{posdiff/p140/768/p140-dec-fixed-minus-p140-enc-fixed/pg19-test/c256-512}{+0.0149}
\defval{posdiff/p140/768/p140-dec-fixed-minus-p140-enc-fixed/pg19-test/c256-512/hi}{+0.0162}
\defval{posdiff/p140/768/p140-dec-fixed-minus-p140-enc-fixed/pg19-test/c256-512/lo}{+0.0137}
\defval{posdiff/p140/768/p140-dec-fixed-minus-p140-enc-fixed/pg19-test/c512-1024}{+0.0170}
\defval{posdiff/p140/768/p140-dec-fixed-minus-p140-enc-fixed/pg19-test/c512-1024/hi}{+0.0183}
\defval{posdiff/p140/768/p140-dec-fixed-minus-p140-enc-fixed/pg19-test/c512-1024/lo}{+0.0158}
\defval{posdiff/p140/768/p140-dec-fixed-minus-p140-enc-fixed/pg19-test/mean}{+0.0164}
\defval{posdiff/p140/768/p140-dec-fixed-minus-p140-enc-fixed/pg19-test/mean/hi}{+0.0175}
\defval{posdiff/p140/768/p140-dec-fixed-minus-p140-enc-fixed/pg19-test/mean/lo}{+0.0155}
\defval{posdiff/p140/768/p140-dec-fixed-minus-p140-enc-fixed/pg19-val/c1-128}{+0.0061}
\defval{posdiff/p140/768/p140-dec-fixed-minus-p140-enc-fixed/pg19-val/c1-128/hi}{+0.0087}
\defval{posdiff/p140/768/p140-dec-fixed-minus-p140-enc-fixed/pg19-val/c1-128/lo}{+0.0037}
\defval{posdiff/p140/768/p140-dec-fixed-minus-p140-enc-fixed/pg19-val/c1024-2048}{+0.0178}
\defval{posdiff/p140/768/p140-dec-fixed-minus-p140-enc-fixed/pg19-val/c1024-2048/hi}{+0.0191}
\defval{posdiff/p140/768/p140-dec-fixed-minus-p140-enc-fixed/pg19-val/c1024-2048/lo}{+0.0164}
\defval{posdiff/p140/768/p140-dec-fixed-minus-p140-enc-fixed/pg19-val/c128-256}{+0.0086}
\defval{posdiff/p140/768/p140-dec-fixed-minus-p140-enc-fixed/pg19-val/c128-256/hi}{+0.0112}
\defval{posdiff/p140/768/p140-dec-fixed-minus-p140-enc-fixed/pg19-val/c128-256/lo}{+0.0061}
\defval{posdiff/p140/768/p140-dec-fixed-minus-p140-enc-fixed/pg19-val/c256-512}{+0.0125}
\defval{posdiff/p140/768/p140-dec-fixed-minus-p140-enc-fixed/pg19-val/c256-512/hi}{+0.0146}
\defval{posdiff/p140/768/p140-dec-fixed-minus-p140-enc-fixed/pg19-val/c256-512/lo}{+0.0104}
\defval{posdiff/p140/768/p140-dec-fixed-minus-p140-enc-fixed/pg19-val/c512-1024}{+0.0163}
\defval{posdiff/p140/768/p140-dec-fixed-minus-p140-enc-fixed/pg19-val/c512-1024/hi}{+0.0180}
\defval{posdiff/p140/768/p140-dec-fixed-minus-p140-enc-fixed/pg19-val/c512-1024/lo}{+0.0146}
\defval{posdiff/p140/768/p140-dec-fixed-minus-p140-enc-fixed/pg19-val/mean}{+0.0155}
\defval{posdiff/p140/768/p140-dec-fixed-minus-p140-enc-fixed/pg19-val/mean/hi}{+0.0166}
\defval{posdiff/p140/768/p140-dec-fixed-minus-p140-enc-fixed/pg19-val/mean/lo}{+0.0144}
\defval{posdiff/p140/768/p140-enc-fixed-minus-p140-enc/finepdfs-long/c1-128}{+0.0074}
\defval{posdiff/p140/768/p140-enc-fixed-minus-p140-enc/finepdfs-long/c1-128/hi}{+0.0088}
\defval{posdiff/p140/768/p140-enc-fixed-minus-p140-enc/finepdfs-long/c1-128/lo}{+0.0059}
\defval{posdiff/p140/768/p140-enc-fixed-minus-p140-enc/finepdfs-long/c1024-2048}{+0.0059}
\defval{posdiff/p140/768/p140-enc-fixed-minus-p140-enc/finepdfs-long/c1024-2048/hi}{+0.0074}
\defval{posdiff/p140/768/p140-enc-fixed-minus-p140-enc/finepdfs-long/c1024-2048/lo}{+0.0044}
\defval{posdiff/p140/768/p140-enc-fixed-minus-p140-enc/finepdfs-long/c128-256}{+0.0064}
\defval{posdiff/p140/768/p140-enc-fixed-minus-p140-enc/finepdfs-long/c128-256/hi}{+0.0077}
\defval{posdiff/p140/768/p140-enc-fixed-minus-p140-enc/finepdfs-long/c128-256/lo}{+0.0051}
\defval{posdiff/p140/768/p140-enc-fixed-minus-p140-enc/finepdfs-long/c256-512}{+0.0061}
\defval{posdiff/p140/768/p140-enc-fixed-minus-p140-enc/finepdfs-long/c256-512/hi}{+0.0073}
\defval{posdiff/p140/768/p140-enc-fixed-minus-p140-enc/finepdfs-long/c256-512/lo}{+0.0050}
\defval{posdiff/p140/768/p140-enc-fixed-minus-p140-enc/finepdfs-long/c512-1024}{+0.0058}
\defval{posdiff/p140/768/p140-enc-fixed-minus-p140-enc/finepdfs-long/c512-1024/hi}{+0.0071}
\defval{posdiff/p140/768/p140-enc-fixed-minus-p140-enc/finepdfs-long/c512-1024/lo}{+0.0045}
\defval{posdiff/p140/768/p140-enc-fixed-minus-p140-enc/finepdfs-long/mean}{+0.0060}
\defval{posdiff/p140/768/p140-enc-fixed-minus-p140-enc/finepdfs-long/mean/hi}{+0.0073}
\defval{posdiff/p140/768/p140-enc-fixed-minus-p140-enc/finepdfs-long/mean/lo}{+0.0048}
\defval{posdiff/p140/768/p140-enc-fixed-minus-p140-enc/fineweb-edu-100bt/c1-128}{+0.0001}
\defval{posdiff/p140/768/p140-enc-fixed-minus-p140-enc/fineweb-edu-100bt/c1-128/hi}{+0.0011}
\defval{posdiff/p140/768/p140-enc-fixed-minus-p140-enc/fineweb-edu-100bt/c1-128/lo}{$-$0.0010}
\defval{posdiff/p140/768/p140-enc-fixed-minus-p140-enc/fineweb-edu-100bt/c1024-2048}{$-$0.0001}
\defval{posdiff/p140/768/p140-enc-fixed-minus-p140-enc/fineweb-edu-100bt/c1024-2048/hi}{+0.0009}
\defval{posdiff/p140/768/p140-enc-fixed-minus-p140-enc/fineweb-edu-100bt/c1024-2048/lo}{$-$0.0010}
\defval{posdiff/p140/768/p140-enc-fixed-minus-p140-enc/fineweb-edu-100bt/c128-256}{+0.0002}
\defval{posdiff/p140/768/p140-enc-fixed-minus-p140-enc/fineweb-edu-100bt/c128-256/hi}{+0.0013}
\defval{posdiff/p140/768/p140-enc-fixed-minus-p140-enc/fineweb-edu-100bt/c128-256/lo}{$-$0.0008}
\defval{posdiff/p140/768/p140-enc-fixed-minus-p140-enc/fineweb-edu-100bt/c256-512}{$-$0.0005}
\defval{posdiff/p140/768/p140-enc-fixed-minus-p140-enc/fineweb-edu-100bt/c256-512/hi}{+0.0004}
\defval{posdiff/p140/768/p140-enc-fixed-minus-p140-enc/fineweb-edu-100bt/c256-512/lo}{$-$0.0014}
\defval{posdiff/p140/768/p140-enc-fixed-minus-p140-enc/fineweb-edu-100bt/c512-1024}{+0.0004}
\defval{posdiff/p140/768/p140-enc-fixed-minus-p140-enc/fineweb-edu-100bt/c512-1024/hi}{+0.0012}
\defval{posdiff/p140/768/p140-enc-fixed-minus-p140-enc/fineweb-edu-100bt/c512-1024/lo}{$-$0.0004}
\defval{posdiff/p140/768/p140-enc-fixed-minus-p140-enc/fineweb-edu-100bt/mean}{+0.0000}
\defval{posdiff/p140/768/p140-enc-fixed-minus-p140-enc/fineweb-edu-100bt/mean/hi}{+0.0005}
\defval{posdiff/p140/768/p140-enc-fixed-minus-p140-enc/fineweb-edu-100bt/mean/lo}{$-$0.0004}
\defval{posdiff/p140/768/p140-enc-fixed-minus-p140-enc/govreport-test/c1-128}{$-$0.0012}
\defval{posdiff/p140/768/p140-enc-fixed-minus-p140-enc/govreport-test/c1-128/hi}{$-$0.0004}
\defval{posdiff/p140/768/p140-enc-fixed-minus-p140-enc/govreport-test/c1-128/lo}{$-$0.0020}
\defval{posdiff/p140/768/p140-enc-fixed-minus-p140-enc/govreport-test/c1024-2048}{+0.0000}
\defval{posdiff/p140/768/p140-enc-fixed-minus-p140-enc/govreport-test/c1024-2048/hi}{+0.0005}
\defval{posdiff/p140/768/p140-enc-fixed-minus-p140-enc/govreport-test/c1024-2048/lo}{$-$0.0005}
\defval{posdiff/p140/768/p140-enc-fixed-minus-p140-enc/govreport-test/c128-256}{$-$0.0002}
\defval{posdiff/p140/768/p140-enc-fixed-minus-p140-enc/govreport-test/c128-256/hi}{+0.0006}
\defval{posdiff/p140/768/p140-enc-fixed-minus-p140-enc/govreport-test/c128-256/lo}{$-$0.0010}
\defval{posdiff/p140/768/p140-enc-fixed-minus-p140-enc/govreport-test/c256-512}{$-$0.0003}
\defval{posdiff/p140/768/p140-enc-fixed-minus-p140-enc/govreport-test/c256-512/hi}{+0.0004}
\defval{posdiff/p140/768/p140-enc-fixed-minus-p140-enc/govreport-test/c256-512/lo}{$-$0.0009}
\defval{posdiff/p140/768/p140-enc-fixed-minus-p140-enc/govreport-test/c512-1024}{$-$0.0004}
\defval{posdiff/p140/768/p140-enc-fixed-minus-p140-enc/govreport-test/c512-1024/hi}{+0.0002}
\defval{posdiff/p140/768/p140-enc-fixed-minus-p140-enc/govreport-test/c512-1024/lo}{$-$0.0010}
\defval{posdiff/p140/768/p140-enc-fixed-minus-p140-enc/govreport-test/mean}{$-$0.0002}
\defval{posdiff/p140/768/p140-enc-fixed-minus-p140-enc/govreport-test/mean/hi}{+0.0002}
\defval{posdiff/p140/768/p140-enc-fixed-minus-p140-enc/govreport-test/mean/lo}{$-$0.0006}
\defval{posdiff/p140/768/p140-enc-fixed-minus-p140-enc/longdocs/c1-128}{+0.0032}
\defval{posdiff/p140/768/p140-enc-fixed-minus-p140-enc/longdocs/c1-128/hi}{+0.0039}
\defval{posdiff/p140/768/p140-enc-fixed-minus-p140-enc/longdocs/c1-128/lo}{+0.0026}
\defval{posdiff/p140/768/p140-enc-fixed-minus-p140-enc/longdocs/c1024-2048}{+0.0027}
\defval{posdiff/p140/768/p140-enc-fixed-minus-p140-enc/longdocs/c1024-2048/hi}{+0.0032}
\defval{posdiff/p140/768/p140-enc-fixed-minus-p140-enc/longdocs/c1024-2048/lo}{+0.0021}
\defval{posdiff/p140/768/p140-enc-fixed-minus-p140-enc/longdocs/c128-256}{+0.0035}
\defval{posdiff/p140/768/p140-enc-fixed-minus-p140-enc/longdocs/c128-256/hi}{+0.0041}
\defval{posdiff/p140/768/p140-enc-fixed-minus-p140-enc/longdocs/c128-256/lo}{+0.0028}
\defval{posdiff/p140/768/p140-enc-fixed-minus-p140-enc/longdocs/c256-512}{+0.0030}
\defval{posdiff/p140/768/p140-enc-fixed-minus-p140-enc/longdocs/c256-512/hi}{+0.0035}
\defval{posdiff/p140/768/p140-enc-fixed-minus-p140-enc/longdocs/c256-512/lo}{+0.0024}
\defval{posdiff/p140/768/p140-enc-fixed-minus-p140-enc/longdocs/c512-1024}{+0.0024}
\defval{posdiff/p140/768/p140-enc-fixed-minus-p140-enc/longdocs/c512-1024/hi}{+0.0030}
\defval{posdiff/p140/768/p140-enc-fixed-minus-p140-enc/longdocs/c512-1024/lo}{+0.0019}
\defval{posdiff/p140/768/p140-enc-fixed-minus-p140-enc/longdocs/mean}{+0.0027}
\defval{posdiff/p140/768/p140-enc-fixed-minus-p140-enc/longdocs/mean/hi}{+0.0032}
\defval{posdiff/p140/768/p140-enc-fixed-minus-p140-enc/longdocs/mean/lo}{+0.0022}
\defval{posdiff/p140/768/p140-enc-fixed-minus-p140-enc/pg19-test/c1-128}{+0.0035}
\defval{posdiff/p140/768/p140-enc-fixed-minus-p140-enc/pg19-test/c1-128/hi}{+0.0048}
\defval{posdiff/p140/768/p140-enc-fixed-minus-p140-enc/pg19-test/c1-128/lo}{+0.0022}
\defval{posdiff/p140/768/p140-enc-fixed-minus-p140-enc/pg19-test/c1024-2048}{+0.0021}
\defval{posdiff/p140/768/p140-enc-fixed-minus-p140-enc/pg19-test/c1024-2048/hi}{+0.0029}
\defval{posdiff/p140/768/p140-enc-fixed-minus-p140-enc/pg19-test/c1024-2048/lo}{+0.0013}
\defval{posdiff/p140/768/p140-enc-fixed-minus-p140-enc/pg19-test/c128-256}{+0.0042}
\defval{posdiff/p140/768/p140-enc-fixed-minus-p140-enc/pg19-test/c128-256/hi}{+0.0054}
\defval{posdiff/p140/768/p140-enc-fixed-minus-p140-enc/pg19-test/c128-256/lo}{+0.0030}
\defval{posdiff/p140/768/p140-enc-fixed-minus-p140-enc/pg19-test/c256-512}{+0.0031}
\defval{posdiff/p140/768/p140-enc-fixed-minus-p140-enc/pg19-test/c256-512/hi}{+0.0041}
\defval{posdiff/p140/768/p140-enc-fixed-minus-p140-enc/pg19-test/c256-512/lo}{+0.0022}
\defval{posdiff/p140/768/p140-enc-fixed-minus-p140-enc/pg19-test/c512-1024}{+0.0020}
\defval{posdiff/p140/768/p140-enc-fixed-minus-p140-enc/pg19-test/c512-1024/hi}{+0.0029}
\defval{posdiff/p140/768/p140-enc-fixed-minus-p140-enc/pg19-test/c512-1024/lo}{+0.0010}
\defval{posdiff/p140/768/p140-enc-fixed-minus-p140-enc/pg19-test/mean}{+0.0024}
\defval{posdiff/p140/768/p140-enc-fixed-minus-p140-enc/pg19-test/mean/hi}{+0.0031}
\defval{posdiff/p140/768/p140-enc-fixed-minus-p140-enc/pg19-test/mean/lo}{+0.0017}
\defval{posdiff/p140/768/p140-enc-fixed-minus-p140-enc/pg19-val/c1-128}{+0.0028}
\defval{posdiff/p140/768/p140-enc-fixed-minus-p140-enc/pg19-val/c1-128/hi}{+0.0052}
\defval{posdiff/p140/768/p140-enc-fixed-minus-p140-enc/pg19-val/c1-128/lo}{+0.0005}
\defval{posdiff/p140/768/p140-enc-fixed-minus-p140-enc/pg19-val/c1024-2048}{+0.0025}
\defval{posdiff/p140/768/p140-enc-fixed-minus-p140-enc/pg19-val/c1024-2048/hi}{+0.0035}
\defval{posdiff/p140/768/p140-enc-fixed-minus-p140-enc/pg19-val/c1024-2048/lo}{+0.0014}
\defval{posdiff/p140/768/p140-enc-fixed-minus-p140-enc/pg19-val/c128-256}{+0.0023}
\defval{posdiff/p140/768/p140-enc-fixed-minus-p140-enc/pg19-val/c128-256/hi}{+0.0046}
\defval{posdiff/p140/768/p140-enc-fixed-minus-p140-enc/pg19-val/c128-256/lo}{+0.0000}
\defval{posdiff/p140/768/p140-enc-fixed-minus-p140-enc/pg19-val/c256-512}{+0.0029}
\defval{posdiff/p140/768/p140-enc-fixed-minus-p140-enc/pg19-val/c256-512/hi}{+0.0047}
\defval{posdiff/p140/768/p140-enc-fixed-minus-p140-enc/pg19-val/c256-512/lo}{+0.0012}
\defval{posdiff/p140/768/p140-enc-fixed-minus-p140-enc/pg19-val/c512-1024}{+0.0019}
\defval{posdiff/p140/768/p140-enc-fixed-minus-p140-enc/pg19-val/c512-1024/hi}{+0.0032}
\defval{posdiff/p140/768/p140-enc-fixed-minus-p140-enc/pg19-val/c512-1024/lo}{+0.0005}
\defval{posdiff/p140/768/p140-enc-fixed-minus-p140-enc/pg19-val/mean}{+0.0024}
\defval{posdiff/p140/768/p140-enc-fixed-minus-p140-enc/pg19-val/mean/hi}{+0.0032}
\defval{posdiff/p140/768/p140-enc-fixed-minus-p140-enc/pg19-val/mean/lo}{+0.0016}
\defval{posdiff/p140/768/p140-noloop-minus-p140-enc/finepdfs-long/c1-128}{+0.0304}
\defval{posdiff/p140/768/p140-noloop-minus-p140-enc/finepdfs-long/c1-128/hi}{+0.0321}
\defval{posdiff/p140/768/p140-noloop-minus-p140-enc/finepdfs-long/c1-128/lo}{+0.0287}
\defval{posdiff/p140/768/p140-noloop-minus-p140-enc/finepdfs-long/c1024-2048}{+0.0616}
\defval{posdiff/p140/768/p140-noloop-minus-p140-enc/finepdfs-long/c1024-2048/hi}{+0.0645}
\defval{posdiff/p140/768/p140-noloop-minus-p140-enc/finepdfs-long/c1024-2048/lo}{+0.0584}
\defval{posdiff/p140/768/p140-noloop-minus-p140-enc/finepdfs-long/c128-256}{+0.0462}
\defval{posdiff/p140/768/p140-noloop-minus-p140-enc/finepdfs-long/c128-256/hi}{+0.0483}
\defval{posdiff/p140/768/p140-noloop-minus-p140-enc/finepdfs-long/c128-256/lo}{+0.0441}
\defval{posdiff/p140/768/p140-noloop-minus-p140-enc/finepdfs-long/c256-512}{+0.0522}
\defval{posdiff/p140/768/p140-noloop-minus-p140-enc/finepdfs-long/c256-512/hi}{+0.0544}
\defval{posdiff/p140/768/p140-noloop-minus-p140-enc/finepdfs-long/c256-512/lo}{+0.0500}
\defval{posdiff/p140/768/p140-noloop-minus-p140-enc/finepdfs-long/c512-1024}{+0.0572}
\defval{posdiff/p140/768/p140-noloop-minus-p140-enc/finepdfs-long/c512-1024/hi}{+0.0597}
\defval{posdiff/p140/768/p140-noloop-minus-p140-enc/finepdfs-long/c512-1024/lo}{+0.0547}
\defval{posdiff/p140/768/p140-noloop-minus-p140-enc/finepdfs-long/mean}{+0.0564}
\defval{posdiff/p140/768/p140-noloop-minus-p140-enc/finepdfs-long/mean/hi}{+0.0586}
\defval{posdiff/p140/768/p140-noloop-minus-p140-enc/finepdfs-long/mean/lo}{+0.0540}
\defval{posdiff/p140/768/p140-noloop-minus-p140-enc/fineweb-edu-100bt/c1-128}{+0.0268}
\defval{posdiff/p140/768/p140-noloop-minus-p140-enc/fineweb-edu-100bt/c1-128/hi}{+0.0280}
\defval{posdiff/p140/768/p140-noloop-minus-p140-enc/fineweb-edu-100bt/c1-128/lo}{+0.0254}
\defval{posdiff/p140/768/p140-noloop-minus-p140-enc/fineweb-edu-100bt/c1024-2048}{+0.0351}
\defval{posdiff/p140/768/p140-noloop-minus-p140-enc/fineweb-edu-100bt/c1024-2048/hi}{+0.0366}
\defval{posdiff/p140/768/p140-noloop-minus-p140-enc/fineweb-edu-100bt/c1024-2048/lo}{+0.0337}
\defval{posdiff/p140/768/p140-noloop-minus-p140-enc/fineweb-edu-100bt/c128-256}{+0.0305}
\defval{posdiff/p140/768/p140-noloop-minus-p140-enc/fineweb-edu-100bt/c128-256/hi}{+0.0318}
\defval{posdiff/p140/768/p140-noloop-minus-p140-enc/fineweb-edu-100bt/c128-256/lo}{+0.0291}
\defval{posdiff/p140/768/p140-noloop-minus-p140-enc/fineweb-edu-100bt/c256-512}{+0.0316}
\defval{posdiff/p140/768/p140-noloop-minus-p140-enc/fineweb-edu-100bt/c256-512/hi}{+0.0329}
\defval{posdiff/p140/768/p140-noloop-minus-p140-enc/fineweb-edu-100bt/c256-512/lo}{+0.0304}
\defval{posdiff/p140/768/p140-noloop-minus-p140-enc/fineweb-edu-100bt/c512-1024}{+0.0334}
\defval{posdiff/p140/768/p140-noloop-minus-p140-enc/fineweb-edu-100bt/c512-1024/hi}{+0.0345}
\defval{posdiff/p140/768/p140-noloop-minus-p140-enc/fineweb-edu-100bt/c512-1024/lo}{+0.0322}
\defval{posdiff/p140/768/p140-noloop-minus-p140-enc/fineweb-edu-100bt/mean}{+0.0318}
\defval{posdiff/p140/768/p140-noloop-minus-p140-enc/fineweb-edu-100bt/mean/hi}{+0.0325}
\defval{posdiff/p140/768/p140-noloop-minus-p140-enc/fineweb-edu-100bt/mean/lo}{+0.0311}
\defval{posdiff/p140/768/p140-noloop-minus-p140-enc/govreport-test/c1-128}{+0.0217}
\defval{posdiff/p140/768/p140-noloop-minus-p140-enc/govreport-test/c1-128/hi}{+0.0227}
\defval{posdiff/p140/768/p140-noloop-minus-p140-enc/govreport-test/c1-128/lo}{+0.0208}
\defval{posdiff/p140/768/p140-noloop-minus-p140-enc/govreport-test/c1024-2048}{+0.0358}
\defval{posdiff/p140/768/p140-noloop-minus-p140-enc/govreport-test/c1024-2048/hi}{+0.0366}
\defval{posdiff/p140/768/p140-noloop-minus-p140-enc/govreport-test/c1024-2048/lo}{+0.0350}
\defval{posdiff/p140/768/p140-noloop-minus-p140-enc/govreport-test/c128-256}{+0.0280}
\defval{posdiff/p140/768/p140-noloop-minus-p140-enc/govreport-test/c128-256/hi}{+0.0289}
\defval{posdiff/p140/768/p140-noloop-minus-p140-enc/govreport-test/c128-256/lo}{+0.0270}
\defval{posdiff/p140/768/p140-noloop-minus-p140-enc/govreport-test/c256-512}{+0.0311}
\defval{posdiff/p140/768/p140-noloop-minus-p140-enc/govreport-test/c256-512/hi}{+0.0319}
\defval{posdiff/p140/768/p140-noloop-minus-p140-enc/govreport-test/c256-512/lo}{+0.0303}
\defval{posdiff/p140/768/p140-noloop-minus-p140-enc/govreport-test/c512-1024}{+0.0337}
\defval{posdiff/p140/768/p140-noloop-minus-p140-enc/govreport-test/c512-1024/hi}{+0.0345}
\defval{posdiff/p140/768/p140-noloop-minus-p140-enc/govreport-test/c512-1024/lo}{+0.0329}
\defval{posdiff/p140/768/p140-noloop-minus-p140-enc/govreport-test/mean}{+0.0333}
\defval{posdiff/p140/768/p140-noloop-minus-p140-enc/govreport-test/mean/hi}{+0.0340}
\defval{posdiff/p140/768/p140-noloop-minus-p140-enc/govreport-test/mean/lo}{+0.0327}
\defval{posdiff/p140/768/p140-noloop-minus-p140-enc/longdocs/c1-128}{+0.0280}
\defval{posdiff/p140/768/p140-noloop-minus-p140-enc/longdocs/c1-128/hi}{+0.0287}
\defval{posdiff/p140/768/p140-noloop-minus-p140-enc/longdocs/c1-128/lo}{+0.0272}
\defval{posdiff/p140/768/p140-noloop-minus-p140-enc/longdocs/c1024-2048}{+0.0500}
\defval{posdiff/p140/768/p140-noloop-minus-p140-enc/longdocs/c1024-2048/hi}{+0.0510}
\defval{posdiff/p140/768/p140-noloop-minus-p140-enc/longdocs/c1024-2048/lo}{+0.0489}
\defval{posdiff/p140/768/p140-noloop-minus-p140-enc/longdocs/c128-256}{+0.0400}
\defval{posdiff/p140/768/p140-noloop-minus-p140-enc/longdocs/c128-256/hi}{+0.0409}
\defval{posdiff/p140/768/p140-noloop-minus-p140-enc/longdocs/c128-256/lo}{+0.0390}
\defval{posdiff/p140/768/p140-noloop-minus-p140-enc/longdocs/c256-512}{+0.0444}
\defval{posdiff/p140/768/p140-noloop-minus-p140-enc/longdocs/c256-512/hi}{+0.0453}
\defval{posdiff/p140/768/p140-noloop-minus-p140-enc/longdocs/c256-512/lo}{+0.0436}
\defval{posdiff/p140/768/p140-noloop-minus-p140-enc/longdocs/c512-1024}{+0.0475}
\defval{posdiff/p140/768/p140-noloop-minus-p140-enc/longdocs/c512-1024/hi}{+0.0485}
\defval{posdiff/p140/768/p140-noloop-minus-p140-enc/longdocs/c512-1024/lo}{+0.0466}
\defval{posdiff/p140/768/p140-noloop-minus-p140-enc/longdocs/mean}{+0.0467}
\defval{posdiff/p140/768/p140-noloop-minus-p140-enc/longdocs/mean/hi}{+0.0475}
\defval{posdiff/p140/768/p140-noloop-minus-p140-enc/longdocs/mean/lo}{+0.0458}
\defval{posdiff/p140/768/p140-noloop-minus-p140-enc/pg19-test/c1-128}{+0.0317}
\defval{posdiff/p140/768/p140-noloop-minus-p140-enc/pg19-test/c1-128/hi}{+0.0331}
\defval{posdiff/p140/768/p140-noloop-minus-p140-enc/pg19-test/c1-128/lo}{+0.0303}
\defval{posdiff/p140/768/p140-noloop-minus-p140-enc/pg19-test/c1024-2048}{+0.0529}
\defval{posdiff/p140/768/p140-noloop-minus-p140-enc/pg19-test/c1024-2048/hi}{+0.0543}
\defval{posdiff/p140/768/p140-noloop-minus-p140-enc/pg19-test/c1024-2048/lo}{+0.0515}
\defval{posdiff/p140/768/p140-noloop-minus-p140-enc/pg19-test/c128-256}{+0.0458}
\defval{posdiff/p140/768/p140-noloop-minus-p140-enc/pg19-test/c128-256/hi}{+0.0476}
\defval{posdiff/p140/768/p140-noloop-minus-p140-enc/pg19-test/c128-256/lo}{+0.0441}
\defval{posdiff/p140/768/p140-noloop-minus-p140-enc/pg19-test/c256-512}{+0.0502}
\defval{posdiff/p140/768/p140-noloop-minus-p140-enc/pg19-test/c256-512/hi}{+0.0517}
\defval{posdiff/p140/768/p140-noloop-minus-p140-enc/pg19-test/c256-512/lo}{+0.0486}
\defval{posdiff/p140/768/p140-noloop-minus-p140-enc/pg19-test/c512-1024}{+0.0519}
\defval{posdiff/p140/768/p140-noloop-minus-p140-enc/pg19-test/c512-1024/hi}{+0.0534}
\defval{posdiff/p140/768/p140-noloop-minus-p140-enc/pg19-test/c512-1024/lo}{+0.0504}
\defval{posdiff/p140/768/p140-noloop-minus-p140-enc/pg19-test/mean}{+0.0506}
\defval{posdiff/p140/768/p140-noloop-minus-p140-enc/pg19-test/mean/hi}{+0.0517}
\defval{posdiff/p140/768/p140-noloop-minus-p140-enc/pg19-test/mean/lo}{+0.0493}
\defval{posdiff/p140/768/p140-noloop-minus-p140-enc/pg19-val/c1-128}{+0.0286}
\defval{posdiff/p140/768/p140-noloop-minus-p140-enc/pg19-val/c1-128/hi}{+0.0312}
\defval{posdiff/p140/768/p140-noloop-minus-p140-enc/pg19-val/c1-128/lo}{+0.0260}
\defval{posdiff/p140/768/p140-noloop-minus-p140-enc/pg19-val/c1024-2048}{+0.0506}
\defval{posdiff/p140/768/p140-noloop-minus-p140-enc/pg19-val/c1024-2048/hi}{+0.0521}
\defval{posdiff/p140/768/p140-noloop-minus-p140-enc/pg19-val/c1024-2048/lo}{+0.0492}
\defval{posdiff/p140/768/p140-noloop-minus-p140-enc/pg19-val/c128-256}{+0.0407}
\defval{posdiff/p140/768/p140-noloop-minus-p140-enc/pg19-val/c128-256/hi}{+0.0432}
\defval{posdiff/p140/768/p140-noloop-minus-p140-enc/pg19-val/c128-256/lo}{+0.0379}
\defval{posdiff/p140/768/p140-noloop-minus-p140-enc/pg19-val/c256-512}{+0.0470}
\defval{posdiff/p140/768/p140-noloop-minus-p140-enc/pg19-val/c256-512/hi}{+0.0493}
\defval{posdiff/p140/768/p140-noloop-minus-p140-enc/pg19-val/c256-512/lo}{+0.0448}
\defval{posdiff/p140/768/p140-noloop-minus-p140-enc/pg19-val/c512-1024}{+0.0485}
\defval{posdiff/p140/768/p140-noloop-minus-p140-enc/pg19-val/c512-1024/hi}{+0.0503}
\defval{posdiff/p140/768/p140-noloop-minus-p140-enc/pg19-val/c512-1024/lo}{+0.0467}
\defval{posdiff/p140/768/p140-noloop-minus-p140-enc/pg19-val/mean}{+0.0476}
\defval{posdiff/p140/768/p140-noloop-minus-p140-enc/pg19-val/mean/hi}{+0.0487}
\defval{posdiff/p140/768/p140-noloop-minus-p140-enc/pg19-val/mean/lo}{+0.0466}
\defval{posdiff/x40/1024/ced-dec-minus-ced-enc/finepdfs-long/c1-128}{+0.0224}
\defval{posdiff/x40/1024/ced-dec-minus-ced-enc/finepdfs-long/c1-128/hi}{+0.0239}
\defval{posdiff/x40/1024/ced-dec-minus-ced-enc/finepdfs-long/c1-128/lo}{+0.0208}
\defval{posdiff/x40/1024/ced-dec-minus-ced-enc/finepdfs-long/c1024-2048}{+0.0345}
\defval{posdiff/x40/1024/ced-dec-minus-ced-enc/finepdfs-long/c1024-2048/hi}{+0.0362}
\defval{posdiff/x40/1024/ced-dec-minus-ced-enc/finepdfs-long/c1024-2048/lo}{+0.0329}
\defval{posdiff/x40/1024/ced-dec-minus-ced-enc/finepdfs-long/c128-256}{+0.0249}
\defval{posdiff/x40/1024/ced-dec-minus-ced-enc/finepdfs-long/c128-256/hi}{+0.0265}
\defval{posdiff/x40/1024/ced-dec-minus-ced-enc/finepdfs-long/c128-256/lo}{+0.0233}
\defval{posdiff/x40/1024/ced-dec-minus-ced-enc/finepdfs-long/c256-512}{+0.0284}
\defval{posdiff/x40/1024/ced-dec-minus-ced-enc/finepdfs-long/c256-512/hi}{+0.0298}
\defval{posdiff/x40/1024/ced-dec-minus-ced-enc/finepdfs-long/c256-512/lo}{+0.0270}
\defval{posdiff/x40/1024/ced-dec-minus-ced-enc/finepdfs-long/c512-1024}{+0.0316}
\defval{posdiff/x40/1024/ced-dec-minus-ced-enc/finepdfs-long/c512-1024/hi}{+0.0331}
\defval{posdiff/x40/1024/ced-dec-minus-ced-enc/finepdfs-long/c512-1024/lo}{+0.0302}
\defval{posdiff/x40/1024/ced-dec-minus-ced-enc/finepdfs-long/mean}{+0.0316}
\defval{posdiff/x40/1024/ced-dec-minus-ced-enc/finepdfs-long/mean/hi}{+0.0329}
\defval{posdiff/x40/1024/ced-dec-minus-ced-enc/finepdfs-long/mean/lo}{+0.0304}
\defval{posdiff/x40/1024/ced-dec-minus-ced-enc/fineweb-edu-100bt/c1-128}{+0.0158}
\defval{posdiff/x40/1024/ced-dec-minus-ced-enc/fineweb-edu-100bt/c1-128/hi}{+0.0168}
\defval{posdiff/x40/1024/ced-dec-minus-ced-enc/fineweb-edu-100bt/c1-128/lo}{+0.0147}
\defval{posdiff/x40/1024/ced-dec-minus-ced-enc/fineweb-edu-100bt/c1024-2048}{+0.0207}
\defval{posdiff/x40/1024/ced-dec-minus-ced-enc/fineweb-edu-100bt/c1024-2048/hi}{+0.0219}
\defval{posdiff/x40/1024/ced-dec-minus-ced-enc/fineweb-edu-100bt/c1024-2048/lo}{+0.0196}
\defval{posdiff/x40/1024/ced-dec-minus-ced-enc/fineweb-edu-100bt/c128-256}{+0.0175}
\defval{posdiff/x40/1024/ced-dec-minus-ced-enc/fineweb-edu-100bt/c128-256/hi}{+0.0187}
\defval{posdiff/x40/1024/ced-dec-minus-ced-enc/fineweb-edu-100bt/c128-256/lo}{+0.0165}
\defval{posdiff/x40/1024/ced-dec-minus-ced-enc/fineweb-edu-100bt/c256-512}{+0.0177}
\defval{posdiff/x40/1024/ced-dec-minus-ced-enc/fineweb-edu-100bt/c256-512/hi}{+0.0186}
\defval{posdiff/x40/1024/ced-dec-minus-ced-enc/fineweb-edu-100bt/c256-512/lo}{+0.0168}
\defval{posdiff/x40/1024/ced-dec-minus-ced-enc/fineweb-edu-100bt/c512-1024}{+0.0190}
\defval{posdiff/x40/1024/ced-dec-minus-ced-enc/fineweb-edu-100bt/c512-1024/hi}{+0.0200}
\defval{posdiff/x40/1024/ced-dec-minus-ced-enc/fineweb-edu-100bt/c512-1024/lo}{+0.0180}
\defval{posdiff/x40/1024/ced-dec-minus-ced-enc/fineweb-edu-100bt/mean}{+0.0183}
\defval{posdiff/x40/1024/ced-dec-minus-ced-enc/fineweb-edu-100bt/mean/hi}{+0.0188}
\defval{posdiff/x40/1024/ced-dec-minus-ced-enc/fineweb-edu-100bt/mean/lo}{+0.0178}
\defval{posdiff/x40/1024/ced-dec-minus-ced-enc/govreport-test/c1-128}{+0.0124}
\defval{posdiff/x40/1024/ced-dec-minus-ced-enc/govreport-test/c1-128/hi}{+0.0133}
\defval{posdiff/x40/1024/ced-dec-minus-ced-enc/govreport-test/c1-128/lo}{+0.0116}
\defval{posdiff/x40/1024/ced-dec-minus-ced-enc/govreport-test/c1024-2048}{+0.0200}
\defval{posdiff/x40/1024/ced-dec-minus-ced-enc/govreport-test/c1024-2048/hi}{+0.0206}
\defval{posdiff/x40/1024/ced-dec-minus-ced-enc/govreport-test/c1024-2048/lo}{+0.0194}
\defval{posdiff/x40/1024/ced-dec-minus-ced-enc/govreport-test/c128-256}{+0.0158}
\defval{posdiff/x40/1024/ced-dec-minus-ced-enc/govreport-test/c128-256/hi}{+0.0167}
\defval{posdiff/x40/1024/ced-dec-minus-ced-enc/govreport-test/c128-256/lo}{+0.0150}
\defval{posdiff/x40/1024/ced-dec-minus-ced-enc/govreport-test/c256-512}{+0.0179}
\defval{posdiff/x40/1024/ced-dec-minus-ced-enc/govreport-test/c256-512/hi}{+0.0186}
\defval{posdiff/x40/1024/ced-dec-minus-ced-enc/govreport-test/c256-512/lo}{+0.0172}
\defval{posdiff/x40/1024/ced-dec-minus-ced-enc/govreport-test/c512-1024}{+0.0188}
\defval{posdiff/x40/1024/ced-dec-minus-ced-enc/govreport-test/c512-1024/hi}{+0.0194}
\defval{posdiff/x40/1024/ced-dec-minus-ced-enc/govreport-test/c512-1024/lo}{+0.0181}
\defval{posdiff/x40/1024/ced-dec-minus-ced-enc/govreport-test/mean}{+0.0187}
\defval{posdiff/x40/1024/ced-dec-minus-ced-enc/govreport-test/mean/hi}{+0.0192}
\defval{posdiff/x40/1024/ced-dec-minus-ced-enc/govreport-test/mean/lo}{+0.0182}
\defval{posdiff/x40/1024/ced-dec-minus-ced-enc/longdocs/c1-128}{+0.0198}
\defval{posdiff/x40/1024/ced-dec-minus-ced-enc/longdocs/c1-128/hi}{+0.0205}
\defval{posdiff/x40/1024/ced-dec-minus-ced-enc/longdocs/c1-128/lo}{+0.0191}
\defval{posdiff/x40/1024/ced-dec-minus-ced-enc/longdocs/c1024-2048}{+0.0288}
\defval{posdiff/x40/1024/ced-dec-minus-ced-enc/longdocs/c1024-2048/hi}{+0.0295}
\defval{posdiff/x40/1024/ced-dec-minus-ced-enc/longdocs/c1024-2048/lo}{+0.0282}
\defval{posdiff/x40/1024/ced-dec-minus-ced-enc/longdocs/c128-256}{+0.0235}
\defval{posdiff/x40/1024/ced-dec-minus-ced-enc/longdocs/c128-256/hi}{+0.0242}
\defval{posdiff/x40/1024/ced-dec-minus-ced-enc/longdocs/c128-256/lo}{+0.0227}
\defval{posdiff/x40/1024/ced-dec-minus-ced-enc/longdocs/c256-512}{+0.0254}
\defval{posdiff/x40/1024/ced-dec-minus-ced-enc/longdocs/c256-512/hi}{+0.0260}
\defval{posdiff/x40/1024/ced-dec-minus-ced-enc/longdocs/c256-512/lo}{+0.0248}
\defval{posdiff/x40/1024/ced-dec-minus-ced-enc/longdocs/c512-1024}{+0.0270}
\defval{posdiff/x40/1024/ced-dec-minus-ced-enc/longdocs/c512-1024/hi}{+0.0276}
\defval{posdiff/x40/1024/ced-dec-minus-ced-enc/longdocs/c512-1024/lo}{+0.0264}
\defval{posdiff/x40/1024/ced-dec-minus-ced-enc/longdocs/mean}{+0.0271}
\defval{posdiff/x40/1024/ced-dec-minus-ced-enc/longdocs/mean/hi}{+0.0276}
\defval{posdiff/x40/1024/ced-dec-minus-ced-enc/longdocs/mean/lo}{+0.0266}
\defval{posdiff/x40/1024/ced-dec-minus-ced-enc/pg19-test/c1-128}{+0.0246}
\defval{posdiff/x40/1024/ced-dec-minus-ced-enc/pg19-test/c1-128/hi}{+0.0257}
\defval{posdiff/x40/1024/ced-dec-minus-ced-enc/pg19-test/c1-128/lo}{+0.0234}
\defval{posdiff/x40/1024/ced-dec-minus-ced-enc/pg19-test/c1024-2048}{+0.0323}
\defval{posdiff/x40/1024/ced-dec-minus-ced-enc/pg19-test/c1024-2048/hi}{+0.0333}
\defval{posdiff/x40/1024/ced-dec-minus-ced-enc/pg19-test/c1024-2048/lo}{+0.0312}
\defval{posdiff/x40/1024/ced-dec-minus-ced-enc/pg19-test/c128-256}{+0.0297}
\defval{posdiff/x40/1024/ced-dec-minus-ced-enc/pg19-test/c128-256/hi}{+0.0309}
\defval{posdiff/x40/1024/ced-dec-minus-ced-enc/pg19-test/c128-256/lo}{+0.0284}
\defval{posdiff/x40/1024/ced-dec-minus-ced-enc/pg19-test/c256-512}{+0.0300}
\defval{posdiff/x40/1024/ced-dec-minus-ced-enc/pg19-test/c256-512/hi}{+0.0312}
\defval{posdiff/x40/1024/ced-dec-minus-ced-enc/pg19-test/c256-512/lo}{+0.0289}
\defval{posdiff/x40/1024/ced-dec-minus-ced-enc/pg19-test/c512-1024}{+0.0307}
\defval{posdiff/x40/1024/ced-dec-minus-ced-enc/pg19-test/c512-1024/hi}{+0.0318}
\defval{posdiff/x40/1024/ced-dec-minus-ced-enc/pg19-test/c512-1024/lo}{+0.0297}
\defval{posdiff/x40/1024/ced-dec-minus-ced-enc/pg19-test/mean}{+0.0310}
\defval{posdiff/x40/1024/ced-dec-minus-ced-enc/pg19-test/mean/hi}{+0.0319}
\defval{posdiff/x40/1024/ced-dec-minus-ced-enc/pg19-test/mean/lo}{+0.0301}
\defval{posdiff/x40/1024/ced-dec-minus-ced-enc/pg19-val/c1-128}{+0.0227}
\defval{posdiff/x40/1024/ced-dec-minus-ced-enc/pg19-val/c1-128/hi}{+0.0251}
\defval{posdiff/x40/1024/ced-dec-minus-ced-enc/pg19-val/c1-128/lo}{+0.0204}
\defval{posdiff/x40/1024/ced-dec-minus-ced-enc/pg19-val/c1024-2048}{+0.0293}
\defval{posdiff/x40/1024/ced-dec-minus-ced-enc/pg19-val/c1024-2048/hi}{+0.0305}
\defval{posdiff/x40/1024/ced-dec-minus-ced-enc/pg19-val/c1024-2048/lo}{+0.0282}
\defval{posdiff/x40/1024/ced-dec-minus-ced-enc/pg19-val/c128-256}{+0.0289}
\defval{posdiff/x40/1024/ced-dec-minus-ced-enc/pg19-val/c128-256/hi}{+0.0318}
\defval{posdiff/x40/1024/ced-dec-minus-ced-enc/pg19-val/c128-256/lo}{+0.0260}
\defval{posdiff/x40/1024/ced-dec-minus-ced-enc/pg19-val/c256-512}{+0.0293}
\defval{posdiff/x40/1024/ced-dec-minus-ced-enc/pg19-val/c256-512/hi}{+0.0312}
\defval{posdiff/x40/1024/ced-dec-minus-ced-enc/pg19-val/c256-512/lo}{+0.0274}
\defval{posdiff/x40/1024/ced-dec-minus-ced-enc/pg19-val/c512-1024}{+0.0289}
\defval{posdiff/x40/1024/ced-dec-minus-ced-enc/pg19-val/c512-1024/hi}{+0.0303}
\defval{posdiff/x40/1024/ced-dec-minus-ced-enc/pg19-val/c512-1024/lo}{+0.0275}
\defval{posdiff/x40/1024/ced-dec-minus-ced-enc/pg19-val/mean}{+0.0287}
\defval{posdiff/x40/1024/ced-dec-minus-ced-enc/pg19-val/mean/hi}{+0.0295}
\defval{posdiff/x40/1024/ced-dec-minus-ced-enc/pg19-val/mean/lo}{+0.0279}
\defval{posdiff/x40/1024/ced-dec-minus-dense-l25/finepdfs-long/c1-128}{+0.0225}
\defval{posdiff/x40/1024/ced-dec-minus-dense-l25/finepdfs-long/c1-128/hi}{+0.0243}
\defval{posdiff/x40/1024/ced-dec-minus-dense-l25/finepdfs-long/c1-128/lo}{+0.0207}
\defval{posdiff/x40/1024/ced-dec-minus-dense-l25/finepdfs-long/c1024-2048}{+0.0313}
\defval{posdiff/x40/1024/ced-dec-minus-dense-l25/finepdfs-long/c1024-2048/hi}{+0.0331}
\defval{posdiff/x40/1024/ced-dec-minus-dense-l25/finepdfs-long/c1024-2048/lo}{+0.0296}
\defval{posdiff/x40/1024/ced-dec-minus-dense-l25/finepdfs-long/c128-256}{+0.0240}
\defval{posdiff/x40/1024/ced-dec-minus-dense-l25/finepdfs-long/c128-256/hi}{+0.0260}
\defval{posdiff/x40/1024/ced-dec-minus-dense-l25/finepdfs-long/c128-256/lo}{+0.0221}
\defval{posdiff/x40/1024/ced-dec-minus-dense-l25/finepdfs-long/c256-512}{+0.0259}
\defval{posdiff/x40/1024/ced-dec-minus-dense-l25/finepdfs-long/c256-512/hi}{+0.0276}
\defval{posdiff/x40/1024/ced-dec-minus-dense-l25/finepdfs-long/c256-512/lo}{+0.0242}
\defval{posdiff/x40/1024/ced-dec-minus-dense-l25/finepdfs-long/c512-1024}{+0.0286}
\defval{posdiff/x40/1024/ced-dec-minus-dense-l25/finepdfs-long/c512-1024/hi}{+0.0304}
\defval{posdiff/x40/1024/ced-dec-minus-dense-l25/finepdfs-long/c512-1024/lo}{+0.0269}
\defval{posdiff/x40/1024/ced-dec-minus-dense-l25/finepdfs-long/mean}{+0.0289}
\defval{posdiff/x40/1024/ced-dec-minus-dense-l25/finepdfs-long/mean/hi}{+0.0305}
\defval{posdiff/x40/1024/ced-dec-minus-dense-l25/finepdfs-long/mean/lo}{+0.0274}
\defval{posdiff/x40/1024/ced-dec-minus-dense-l25/fineweb-edu-100bt/c1-128}{+0.0136}
\defval{posdiff/x40/1024/ced-dec-minus-dense-l25/fineweb-edu-100bt/c1-128/hi}{+0.0147}
\defval{posdiff/x40/1024/ced-dec-minus-dense-l25/fineweb-edu-100bt/c1-128/lo}{+0.0124}
\defval{posdiff/x40/1024/ced-dec-minus-dense-l25/fineweb-edu-100bt/c1024-2048}{+0.0227}
\defval{posdiff/x40/1024/ced-dec-minus-dense-l25/fineweb-edu-100bt/c1024-2048/hi}{+0.0239}
\defval{posdiff/x40/1024/ced-dec-minus-dense-l25/fineweb-edu-100bt/c1024-2048/lo}{+0.0216}
\defval{posdiff/x40/1024/ced-dec-minus-dense-l25/fineweb-edu-100bt/c128-256}{+0.0178}
\defval{posdiff/x40/1024/ced-dec-minus-dense-l25/fineweb-edu-100bt/c128-256/hi}{+0.0190}
\defval{posdiff/x40/1024/ced-dec-minus-dense-l25/fineweb-edu-100bt/c128-256/lo}{+0.0167}
\defval{posdiff/x40/1024/ced-dec-minus-dense-l25/fineweb-edu-100bt/c256-512}{+0.0193}
\defval{posdiff/x40/1024/ced-dec-minus-dense-l25/fineweb-edu-100bt/c256-512/hi}{+0.0203}
\defval{posdiff/x40/1024/ced-dec-minus-dense-l25/fineweb-edu-100bt/c256-512/lo}{+0.0183}
\defval{posdiff/x40/1024/ced-dec-minus-dense-l25/fineweb-edu-100bt/c512-1024}{+0.0203}
\defval{posdiff/x40/1024/ced-dec-minus-dense-l25/fineweb-edu-100bt/c512-1024/hi}{+0.0213}
\defval{posdiff/x40/1024/ced-dec-minus-dense-l25/fineweb-edu-100bt/c512-1024/lo}{+0.0193}
\defval{posdiff/x40/1024/ced-dec-minus-dense-l25/fineweb-edu-100bt/mean}{+0.0191}
\defval{posdiff/x40/1024/ced-dec-minus-dense-l25/fineweb-edu-100bt/mean/hi}{+0.0196}
\defval{posdiff/x40/1024/ced-dec-minus-dense-l25/fineweb-edu-100bt/mean/lo}{+0.0186}
\defval{posdiff/x40/1024/ced-dec-minus-dense-l25/govreport-test/c1-128}{+0.0122}
\defval{posdiff/x40/1024/ced-dec-minus-dense-l25/govreport-test/c1-128/hi}{+0.0131}
\defval{posdiff/x40/1024/ced-dec-minus-dense-l25/govreport-test/c1-128/lo}{+0.0113}
\defval{posdiff/x40/1024/ced-dec-minus-dense-l25/govreport-test/c1024-2048}{+0.0238}
\defval{posdiff/x40/1024/ced-dec-minus-dense-l25/govreport-test/c1024-2048/hi}{+0.0246}
\defval{posdiff/x40/1024/ced-dec-minus-dense-l25/govreport-test/c1024-2048/lo}{+0.0232}
\defval{posdiff/x40/1024/ced-dec-minus-dense-l25/govreport-test/c128-256}{+0.0167}
\defval{posdiff/x40/1024/ced-dec-minus-dense-l25/govreport-test/c128-256/hi}{+0.0175}
\defval{posdiff/x40/1024/ced-dec-minus-dense-l25/govreport-test/c128-256/lo}{+0.0159}
\defval{posdiff/x40/1024/ced-dec-minus-dense-l25/govreport-test/c256-512}{+0.0200}
\defval{posdiff/x40/1024/ced-dec-minus-dense-l25/govreport-test/c256-512/hi}{+0.0207}
\defval{posdiff/x40/1024/ced-dec-minus-dense-l25/govreport-test/c256-512/lo}{+0.0192}
\defval{posdiff/x40/1024/ced-dec-minus-dense-l25/govreport-test/c512-1024}{+0.0220}
\defval{posdiff/x40/1024/ced-dec-minus-dense-l25/govreport-test/c512-1024/hi}{+0.0226}
\defval{posdiff/x40/1024/ced-dec-minus-dense-l25/govreport-test/c512-1024/lo}{+0.0213}
\defval{posdiff/x40/1024/ced-dec-minus-dense-l25/govreport-test/mean}{+0.0217}
\defval{posdiff/x40/1024/ced-dec-minus-dense-l25/govreport-test/mean/hi}{+0.0222}
\defval{posdiff/x40/1024/ced-dec-minus-dense-l25/govreport-test/mean/lo}{+0.0212}
\defval{posdiff/x40/1024/ced-dec-minus-dense-l25/longdocs/c1-128}{+0.0206}
\defval{posdiff/x40/1024/ced-dec-minus-dense-l25/longdocs/c1-128/hi}{+0.0215}
\defval{posdiff/x40/1024/ced-dec-minus-dense-l25/longdocs/c1-128/lo}{+0.0197}
\defval{posdiff/x40/1024/ced-dec-minus-dense-l25/longdocs/c1024-2048}{+0.0295}
\defval{posdiff/x40/1024/ced-dec-minus-dense-l25/longdocs/c1024-2048/hi}{+0.0305}
\defval{posdiff/x40/1024/ced-dec-minus-dense-l25/longdocs/c1024-2048/lo}{+0.0287}
\defval{posdiff/x40/1024/ced-dec-minus-dense-l25/longdocs/c128-256}{+0.0240}
\defval{posdiff/x40/1024/ced-dec-minus-dense-l25/longdocs/c128-256/hi}{+0.0249}
\defval{posdiff/x40/1024/ced-dec-minus-dense-l25/longdocs/c128-256/lo}{+0.0230}
\defval{posdiff/x40/1024/ced-dec-minus-dense-l25/longdocs/c256-512}{+0.0260}
\defval{posdiff/x40/1024/ced-dec-minus-dense-l25/longdocs/c256-512/hi}{+0.0268}
\defval{posdiff/x40/1024/ced-dec-minus-dense-l25/longdocs/c256-512/lo}{+0.0251}
\defval{posdiff/x40/1024/ced-dec-minus-dense-l25/longdocs/c512-1024}{+0.0277}
\defval{posdiff/x40/1024/ced-dec-minus-dense-l25/longdocs/c512-1024/hi}{+0.0285}
\defval{posdiff/x40/1024/ced-dec-minus-dense-l25/longdocs/c512-1024/lo}{+0.0269}
\defval{posdiff/x40/1024/ced-dec-minus-dense-l25/longdocs/mean}{+0.0277}
\defval{posdiff/x40/1024/ced-dec-minus-dense-l25/longdocs/mean/hi}{+0.0285}
\defval{posdiff/x40/1024/ced-dec-minus-dense-l25/longdocs/mean/lo}{+0.0270}
\defval{posdiff/x40/1024/ced-dec-minus-dense-l25/pg19-test/c1-128}{+0.0270}
\defval{posdiff/x40/1024/ced-dec-minus-dense-l25/pg19-test/c1-128/hi}{+0.0288}
\defval{posdiff/x40/1024/ced-dec-minus-dense-l25/pg19-test/c1-128/lo}{+0.0251}
\defval{posdiff/x40/1024/ced-dec-minus-dense-l25/pg19-test/c1024-2048}{+0.0336}
\defval{posdiff/x40/1024/ced-dec-minus-dense-l25/pg19-test/c1024-2048/hi}{+0.0353}
\defval{posdiff/x40/1024/ced-dec-minus-dense-l25/pg19-test/c1024-2048/lo}{+0.0318}
\defval{posdiff/x40/1024/ced-dec-minus-dense-l25/pg19-test/c128-256}{+0.0312}
\defval{posdiff/x40/1024/ced-dec-minus-dense-l25/pg19-test/c128-256/hi}{+0.0331}
\defval{posdiff/x40/1024/ced-dec-minus-dense-l25/pg19-test/c128-256/lo}{+0.0292}
\defval{posdiff/x40/1024/ced-dec-minus-dense-l25/pg19-test/c256-512}{+0.0320}
\defval{posdiff/x40/1024/ced-dec-minus-dense-l25/pg19-test/c256-512/hi}{+0.0337}
\defval{posdiff/x40/1024/ced-dec-minus-dense-l25/pg19-test/c256-512/lo}{+0.0303}
\defval{posdiff/x40/1024/ced-dec-minus-dense-l25/pg19-test/c512-1024}{+0.0326}
\defval{posdiff/x40/1024/ced-dec-minus-dense-l25/pg19-test/c512-1024/hi}{+0.0342}
\defval{posdiff/x40/1024/ced-dec-minus-dense-l25/pg19-test/c512-1024/lo}{+0.0309}
\defval{posdiff/x40/1024/ced-dec-minus-dense-l25/pg19-test/mean}{+0.0326}
\defval{posdiff/x40/1024/ced-dec-minus-dense-l25/pg19-test/mean/hi}{+0.0341}
\defval{posdiff/x40/1024/ced-dec-minus-dense-l25/pg19-test/mean/lo}{+0.0310}
\defval{posdiff/x40/1024/ced-dec-minus-dense-l25/pg19-val/c1-128}{+0.0273}
\defval{posdiff/x40/1024/ced-dec-minus-dense-l25/pg19-val/c1-128/hi}{+0.0301}
\defval{posdiff/x40/1024/ced-dec-minus-dense-l25/pg19-val/c1-128/lo}{+0.0246}
\defval{posdiff/x40/1024/ced-dec-minus-dense-l25/pg19-val/c1024-2048}{+0.0313}
\defval{posdiff/x40/1024/ced-dec-minus-dense-l25/pg19-val/c1024-2048/hi}{+0.0328}
\defval{posdiff/x40/1024/ced-dec-minus-dense-l25/pg19-val/c1024-2048/lo}{+0.0297}
\defval{posdiff/x40/1024/ced-dec-minus-dense-l25/pg19-val/c128-256}{+0.0314}
\defval{posdiff/x40/1024/ced-dec-minus-dense-l25/pg19-val/c128-256/hi}{+0.0341}
\defval{posdiff/x40/1024/ced-dec-minus-dense-l25/pg19-val/c128-256/lo}{+0.0283}
\defval{posdiff/x40/1024/ced-dec-minus-dense-l25/pg19-val/c256-512}{+0.0316}
\defval{posdiff/x40/1024/ced-dec-minus-dense-l25/pg19-val/c256-512/hi}{+0.0336}
\defval{posdiff/x40/1024/ced-dec-minus-dense-l25/pg19-val/c256-512/lo}{+0.0295}
\defval{posdiff/x40/1024/ced-dec-minus-dense-l25/pg19-val/c512-1024}{+0.0319}
\defval{posdiff/x40/1024/ced-dec-minus-dense-l25/pg19-val/c512-1024/hi}{+0.0336}
\defval{posdiff/x40/1024/ced-dec-minus-dense-l25/pg19-val/c512-1024/lo}{+0.0301}
\defval{posdiff/x40/1024/ced-dec-minus-dense-l25/pg19-val/mean}{+0.0313}
\defval{posdiff/x40/1024/ced-dec-minus-dense-l25/pg19-val/mean/hi}{+0.0324}
\defval{posdiff/x40/1024/ced-dec-minus-dense-l25/pg19-val/mean/lo}{+0.0300}
\defval{posdiff/x40/1024/ced-enc-fixed-minus-ced-enc/finepdfs-long/c1-128}{+0.0113}
\defval{posdiff/x40/1024/ced-enc-fixed-minus-ced-enc/finepdfs-long/c1-128/hi}{+0.0128}
\defval{posdiff/x40/1024/ced-enc-fixed-minus-ced-enc/finepdfs-long/c1-128/lo}{+0.0101}
\defval{posdiff/x40/1024/ced-enc-fixed-minus-ced-enc/finepdfs-long/c1024-2048}{+0.0048}
\defval{posdiff/x40/1024/ced-enc-fixed-minus-ced-enc/finepdfs-long/c1024-2048/hi}{+0.0058}
\defval{posdiff/x40/1024/ced-enc-fixed-minus-ced-enc/finepdfs-long/c1024-2048/lo}{+0.0038}
\defval{posdiff/x40/1024/ced-enc-fixed-minus-ced-enc/finepdfs-long/c128-256}{+0.0060}
\defval{posdiff/x40/1024/ced-enc-fixed-minus-ced-enc/finepdfs-long/c128-256/hi}{+0.0073}
\defval{posdiff/x40/1024/ced-enc-fixed-minus-ced-enc/finepdfs-long/c128-256/lo}{+0.0048}
\defval{posdiff/x40/1024/ced-enc-fixed-minus-ced-enc/finepdfs-long/c256-512}{+0.0056}
\defval{posdiff/x40/1024/ced-enc-fixed-minus-ced-enc/finepdfs-long/c256-512/hi}{+0.0067}
\defval{posdiff/x40/1024/ced-enc-fixed-minus-ced-enc/finepdfs-long/c256-512/lo}{+0.0045}
\defval{posdiff/x40/1024/ced-enc-fixed-minus-ced-enc/finepdfs-long/c512-1024}{+0.0051}
\defval{posdiff/x40/1024/ced-enc-fixed-minus-ced-enc/finepdfs-long/c512-1024/hi}{+0.0061}
\defval{posdiff/x40/1024/ced-enc-fixed-minus-ced-enc/finepdfs-long/c512-1024/lo}{+0.0040}
\defval{posdiff/x40/1024/ced-enc-fixed-minus-ced-enc/finepdfs-long/mean}{+0.0055}
\defval{posdiff/x40/1024/ced-enc-fixed-minus-ced-enc/finepdfs-long/mean/hi}{+0.0063}
\defval{posdiff/x40/1024/ced-enc-fixed-minus-ced-enc/finepdfs-long/mean/lo}{+0.0046}
\defval{posdiff/x40/1024/ced-enc-fixed-minus-ced-enc/fineweb-edu-100bt/c1-128}{+0.0024}
\defval{posdiff/x40/1024/ced-enc-fixed-minus-ced-enc/fineweb-edu-100bt/c1-128/hi}{+0.0032}
\defval{posdiff/x40/1024/ced-enc-fixed-minus-ced-enc/fineweb-edu-100bt/c1-128/lo}{+0.0015}
\defval{posdiff/x40/1024/ced-enc-fixed-minus-ced-enc/fineweb-edu-100bt/c1024-2048}{+0.0017}
\defval{posdiff/x40/1024/ced-enc-fixed-minus-ced-enc/fineweb-edu-100bt/c1024-2048/hi}{+0.0026}
\defval{posdiff/x40/1024/ced-enc-fixed-minus-ced-enc/fineweb-edu-100bt/c1024-2048/lo}{+0.0008}
\defval{posdiff/x40/1024/ced-enc-fixed-minus-ced-enc/fineweb-edu-100bt/c128-256}{+0.0014}
\defval{posdiff/x40/1024/ced-enc-fixed-minus-ced-enc/fineweb-edu-100bt/c128-256/hi}{+0.0022}
\defval{posdiff/x40/1024/ced-enc-fixed-minus-ced-enc/fineweb-edu-100bt/c128-256/lo}{+0.0005}
\defval{posdiff/x40/1024/ced-enc-fixed-minus-ced-enc/fineweb-edu-100bt/c256-512}{+0.0019}
\defval{posdiff/x40/1024/ced-enc-fixed-minus-ced-enc/fineweb-edu-100bt/c256-512/hi}{+0.0026}
\defval{posdiff/x40/1024/ced-enc-fixed-minus-ced-enc/fineweb-edu-100bt/c256-512/lo}{+0.0011}
\defval{posdiff/x40/1024/ced-enc-fixed-minus-ced-enc/fineweb-edu-100bt/c512-1024}{+0.0021}
\defval{posdiff/x40/1024/ced-enc-fixed-minus-ced-enc/fineweb-edu-100bt/c512-1024/hi}{+0.0028}
\defval{posdiff/x40/1024/ced-enc-fixed-minus-ced-enc/fineweb-edu-100bt/c512-1024/lo}{+0.0013}
\defval{posdiff/x40/1024/ced-enc-fixed-minus-ced-enc/fineweb-edu-100bt/mean}{+0.0019}
\defval{posdiff/x40/1024/ced-enc-fixed-minus-ced-enc/fineweb-edu-100bt/mean/hi}{+0.0023}
\defval{posdiff/x40/1024/ced-enc-fixed-minus-ced-enc/fineweb-edu-100bt/mean/lo}{+0.0015}
\defval{posdiff/x40/1024/ced-enc-fixed-minus-ced-enc/govreport-test/c1-128}{+0.0016}
\defval{posdiff/x40/1024/ced-enc-fixed-minus-ced-enc/govreport-test/c1-128/hi}{+0.0023}
\defval{posdiff/x40/1024/ced-enc-fixed-minus-ced-enc/govreport-test/c1-128/lo}{+0.0008}
\defval{posdiff/x40/1024/ced-enc-fixed-minus-ced-enc/govreport-test/c1024-2048}{+0.0027}
\defval{posdiff/x40/1024/ced-enc-fixed-minus-ced-enc/govreport-test/c1024-2048/hi}{+0.0032}
\defval{posdiff/x40/1024/ced-enc-fixed-minus-ced-enc/govreport-test/c1024-2048/lo}{+0.0023}
\defval{posdiff/x40/1024/ced-enc-fixed-minus-ced-enc/govreport-test/c128-256}{+0.0016}
\defval{posdiff/x40/1024/ced-enc-fixed-minus-ced-enc/govreport-test/c128-256/hi}{+0.0023}
\defval{posdiff/x40/1024/ced-enc-fixed-minus-ced-enc/govreport-test/c128-256/lo}{+0.0009}
\defval{posdiff/x40/1024/ced-enc-fixed-minus-ced-enc/govreport-test/c256-512}{+0.0022}
\defval{posdiff/x40/1024/ced-enc-fixed-minus-ced-enc/govreport-test/c256-512/hi}{+0.0027}
\defval{posdiff/x40/1024/ced-enc-fixed-minus-ced-enc/govreport-test/c256-512/lo}{+0.0017}
\defval{posdiff/x40/1024/ced-enc-fixed-minus-ced-enc/govreport-test/c512-1024}{+0.0023}
\defval{posdiff/x40/1024/ced-enc-fixed-minus-ced-enc/govreport-test/c512-1024/hi}{+0.0028}
\defval{posdiff/x40/1024/ced-enc-fixed-minus-ced-enc/govreport-test/c512-1024/lo}{+0.0019}
\defval{posdiff/x40/1024/ced-enc-fixed-minus-ced-enc/govreport-test/mean}{+0.0024}
\defval{posdiff/x40/1024/ced-enc-fixed-minus-ced-enc/govreport-test/mean/hi}{+0.0028}
\defval{posdiff/x40/1024/ced-enc-fixed-minus-ced-enc/govreport-test/mean/lo}{+0.0021}
\defval{posdiff/x40/1024/ced-enc-fixed-minus-ced-enc/longdocs/c1-128}{+0.0066}
\defval{posdiff/x40/1024/ced-enc-fixed-minus-ced-enc/longdocs/c1-128/hi}{+0.0072}
\defval{posdiff/x40/1024/ced-enc-fixed-minus-ced-enc/longdocs/c1-128/lo}{+0.0059}
\defval{posdiff/x40/1024/ced-enc-fixed-minus-ced-enc/longdocs/c1024-2048}{+0.0044}
\defval{posdiff/x40/1024/ced-enc-fixed-minus-ced-enc/longdocs/c1024-2048/hi}{+0.0049}
\defval{posdiff/x40/1024/ced-enc-fixed-minus-ced-enc/longdocs/c1024-2048/lo}{+0.0040}
\defval{posdiff/x40/1024/ced-enc-fixed-minus-ced-enc/longdocs/c128-256}{+0.0040}
\defval{posdiff/x40/1024/ced-enc-fixed-minus-ced-enc/longdocs/c128-256/hi}{+0.0047}
\defval{posdiff/x40/1024/ced-enc-fixed-minus-ced-enc/longdocs/c128-256/lo}{+0.0034}
\defval{posdiff/x40/1024/ced-enc-fixed-minus-ced-enc/longdocs/c256-512}{+0.0042}
\defval{posdiff/x40/1024/ced-enc-fixed-minus-ced-enc/longdocs/c256-512/hi}{+0.0046}
\defval{posdiff/x40/1024/ced-enc-fixed-minus-ced-enc/longdocs/c256-512/lo}{+0.0037}
\defval{posdiff/x40/1024/ced-enc-fixed-minus-ced-enc/longdocs/c512-1024}{+0.0042}
\defval{posdiff/x40/1024/ced-enc-fixed-minus-ced-enc/longdocs/c512-1024/hi}{+0.0047}
\defval{posdiff/x40/1024/ced-enc-fixed-minus-ced-enc/longdocs/c512-1024/lo}{+0.0037}
\defval{posdiff/x40/1024/ced-enc-fixed-minus-ced-enc/longdocs/mean}{+0.0045}
\defval{posdiff/x40/1024/ced-enc-fixed-minus-ced-enc/longdocs/mean/hi}{+0.0048}
\defval{posdiff/x40/1024/ced-enc-fixed-minus-ced-enc/longdocs/mean/lo}{+0.0041}
\defval{posdiff/x40/1024/ced-enc-fixed-minus-ced-enc/pg19-test/c1-128}{+0.0070}
\defval{posdiff/x40/1024/ced-enc-fixed-minus-ced-enc/pg19-test/c1-128/hi}{+0.0080}
\defval{posdiff/x40/1024/ced-enc-fixed-minus-ced-enc/pg19-test/c1-128/lo}{+0.0058}
\defval{posdiff/x40/1024/ced-enc-fixed-minus-ced-enc/pg19-test/c1024-2048}{+0.0057}
\defval{posdiff/x40/1024/ced-enc-fixed-minus-ced-enc/pg19-test/c1024-2048/hi}{+0.0064}
\defval{posdiff/x40/1024/ced-enc-fixed-minus-ced-enc/pg19-test/c1024-2048/lo}{+0.0049}
\defval{posdiff/x40/1024/ced-enc-fixed-minus-ced-enc/pg19-test/c128-256}{+0.0046}
\defval{posdiff/x40/1024/ced-enc-fixed-minus-ced-enc/pg19-test/c128-256/hi}{+0.0056}
\defval{posdiff/x40/1024/ced-enc-fixed-minus-ced-enc/pg19-test/c128-256/lo}{+0.0034}
\defval{posdiff/x40/1024/ced-enc-fixed-minus-ced-enc/pg19-test/c256-512}{+0.0047}
\defval{posdiff/x40/1024/ced-enc-fixed-minus-ced-enc/pg19-test/c256-512/hi}{+0.0055}
\defval{posdiff/x40/1024/ced-enc-fixed-minus-ced-enc/pg19-test/c256-512/lo}{+0.0038}
\defval{posdiff/x40/1024/ced-enc-fixed-minus-ced-enc/pg19-test/c512-1024}{+0.0053}
\defval{posdiff/x40/1024/ced-enc-fixed-minus-ced-enc/pg19-test/c512-1024/hi}{+0.0060}
\defval{posdiff/x40/1024/ced-enc-fixed-minus-ced-enc/pg19-test/c512-1024/lo}{+0.0045}
\defval{posdiff/x40/1024/ced-enc-fixed-minus-ced-enc/pg19-test/mean}{+0.0055}
\defval{posdiff/x40/1024/ced-enc-fixed-minus-ced-enc/pg19-test/mean/hi}{+0.0060}
\defval{posdiff/x40/1024/ced-enc-fixed-minus-ced-enc/pg19-test/mean/lo}{+0.0048}
\defval{posdiff/x40/1024/ced-enc-fixed-minus-ced-enc/pg19-val/c1-128}{+0.0052}
\defval{posdiff/x40/1024/ced-enc-fixed-minus-ced-enc/pg19-val/c1-128/hi}{+0.0074}
\defval{posdiff/x40/1024/ced-enc-fixed-minus-ced-enc/pg19-val/c1-128/lo}{+0.0030}
\defval{posdiff/x40/1024/ced-enc-fixed-minus-ced-enc/pg19-val/c1024-2048}{+0.0054}
\defval{posdiff/x40/1024/ced-enc-fixed-minus-ced-enc/pg19-val/c1024-2048/hi}{+0.0063}
\defval{posdiff/x40/1024/ced-enc-fixed-minus-ced-enc/pg19-val/c1024-2048/lo}{+0.0046}
\defval{posdiff/x40/1024/ced-enc-fixed-minus-ced-enc/pg19-val/c128-256}{+0.0042}
\defval{posdiff/x40/1024/ced-enc-fixed-minus-ced-enc/pg19-val/c128-256/hi}{+0.0063}
\defval{posdiff/x40/1024/ced-enc-fixed-minus-ced-enc/pg19-val/c128-256/lo}{+0.0021}
\defval{posdiff/x40/1024/ced-enc-fixed-minus-ced-enc/pg19-val/c256-512}{+0.0051}
\defval{posdiff/x40/1024/ced-enc-fixed-minus-ced-enc/pg19-val/c256-512/hi}{+0.0070}
\defval{posdiff/x40/1024/ced-enc-fixed-minus-ced-enc/pg19-val/c256-512/lo}{+0.0035}
\defval{posdiff/x40/1024/ced-enc-fixed-minus-ced-enc/pg19-val/c512-1024}{+0.0042}
\defval{posdiff/x40/1024/ced-enc-fixed-minus-ced-enc/pg19-val/c512-1024/hi}{+0.0053}
\defval{posdiff/x40/1024/ced-enc-fixed-minus-ced-enc/pg19-val/c512-1024/lo}{+0.0030}
\defval{posdiff/x40/1024/ced-enc-fixed-minus-ced-enc/pg19-val/mean}{+0.0050}
\defval{posdiff/x40/1024/ced-enc-fixed-minus-ced-enc/pg19-val/mean/hi}{+0.0057}
\defval{posdiff/x40/1024/ced-enc-fixed-minus-ced-enc/pg19-val/mean/lo}{+0.0044}
\defval{posdiff/x40/1024/ced-enc-minus-dense-l25/finepdfs-long/c1-128}{+0.0001}
\defval{posdiff/x40/1024/ced-enc-minus-dense-l25/finepdfs-long/c1-128/hi}{+0.0016}
\defval{posdiff/x40/1024/ced-enc-minus-dense-l25/finepdfs-long/c1-128/lo}{$-$0.0014}
\defval{posdiff/x40/1024/ced-enc-minus-dense-l25/finepdfs-long/c1024-2048}{$-$0.0032}
\defval{posdiff/x40/1024/ced-enc-minus-dense-l25/finepdfs-long/c1024-2048/hi}{$-$0.0015}
\defval{posdiff/x40/1024/ced-enc-minus-dense-l25/finepdfs-long/c1024-2048/lo}{$-$0.0049}
\defval{posdiff/x40/1024/ced-enc-minus-dense-l25/finepdfs-long/c128-256}{$-$0.0009}
\defval{posdiff/x40/1024/ced-enc-minus-dense-l25/finepdfs-long/c128-256/hi}{+0.0006}
\defval{posdiff/x40/1024/ced-enc-minus-dense-l25/finepdfs-long/c128-256/lo}{$-$0.0024}
\defval{posdiff/x40/1024/ced-enc-minus-dense-l25/finepdfs-long/c256-512}{$-$0.0025}
\defval{posdiff/x40/1024/ced-enc-minus-dense-l25/finepdfs-long/c256-512/hi}{$-$0.0012}
\defval{posdiff/x40/1024/ced-enc-minus-dense-l25/finepdfs-long/c256-512/lo}{$-$0.0038}
\defval{posdiff/x40/1024/ced-enc-minus-dense-l25/finepdfs-long/c512-1024}{$-$0.0030}
\defval{posdiff/x40/1024/ced-enc-minus-dense-l25/finepdfs-long/c512-1024/hi}{$-$0.0015}
\defval{posdiff/x40/1024/ced-enc-minus-dense-l25/finepdfs-long/c512-1024/lo}{$-$0.0045}
\defval{posdiff/x40/1024/ced-enc-minus-dense-l25/finepdfs-long/mean}{$-$0.0027}
\defval{posdiff/x40/1024/ced-enc-minus-dense-l25/finepdfs-long/mean/hi}{$-$0.0014}
\defval{posdiff/x40/1024/ced-enc-minus-dense-l25/finepdfs-long/mean/lo}{$-$0.0040}
\defval{posdiff/x40/1024/ced-enc-minus-dense-l25/fineweb-edu-100bt/c1-128}{$-$0.0022}
\defval{posdiff/x40/1024/ced-enc-minus-dense-l25/fineweb-edu-100bt/c1-128/hi}{$-$0.0012}
\defval{posdiff/x40/1024/ced-enc-minus-dense-l25/fineweb-edu-100bt/c1-128/lo}{$-$0.0032}
\defval{posdiff/x40/1024/ced-enc-minus-dense-l25/fineweb-edu-100bt/c1024-2048}{+0.0020}
\defval{posdiff/x40/1024/ced-enc-minus-dense-l25/fineweb-edu-100bt/c1024-2048/hi}{+0.0030}
\defval{posdiff/x40/1024/ced-enc-minus-dense-l25/fineweb-edu-100bt/c1024-2048/lo}{+0.0010}
\defval{posdiff/x40/1024/ced-enc-minus-dense-l25/fineweb-edu-100bt/c128-256}{+0.0003}
\defval{posdiff/x40/1024/ced-enc-minus-dense-l25/fineweb-edu-100bt/c128-256/hi}{+0.0013}
\defval{posdiff/x40/1024/ced-enc-minus-dense-l25/fineweb-edu-100bt/c128-256/lo}{$-$0.0007}
\defval{posdiff/x40/1024/ced-enc-minus-dense-l25/fineweb-edu-100bt/c256-512}{+0.0016}
\defval{posdiff/x40/1024/ced-enc-minus-dense-l25/fineweb-edu-100bt/c256-512/hi}{+0.0024}
\defval{posdiff/x40/1024/ced-enc-minus-dense-l25/fineweb-edu-100bt/c256-512/lo}{+0.0008}
\defval{posdiff/x40/1024/ced-enc-minus-dense-l25/fineweb-edu-100bt/c512-1024}{+0.0012}
\defval{posdiff/x40/1024/ced-enc-minus-dense-l25/fineweb-edu-100bt/c512-1024/hi}{+0.0021}
\defval{posdiff/x40/1024/ced-enc-minus-dense-l25/fineweb-edu-100bt/c512-1024/lo}{+0.0004}
\defval{posdiff/x40/1024/ced-enc-minus-dense-l25/fineweb-edu-100bt/mean}{+0.0008}
\defval{posdiff/x40/1024/ced-enc-minus-dense-l25/fineweb-edu-100bt/mean/hi}{+0.0012}
\defval{posdiff/x40/1024/ced-enc-minus-dense-l25/fineweb-edu-100bt/mean/lo}{+0.0004}
\defval{posdiff/x40/1024/ced-enc-minus-dense-l25/govreport-test/c1-128}{$-$0.0002}
\defval{posdiff/x40/1024/ced-enc-minus-dense-l25/govreport-test/c1-128/hi}{+0.0005}
\defval{posdiff/x40/1024/ced-enc-minus-dense-l25/govreport-test/c1-128/lo}{$-$0.0009}
\defval{posdiff/x40/1024/ced-enc-minus-dense-l25/govreport-test/c1024-2048}{+0.0038}
\defval{posdiff/x40/1024/ced-enc-minus-dense-l25/govreport-test/c1024-2048/hi}{+0.0043}
\defval{posdiff/x40/1024/ced-enc-minus-dense-l25/govreport-test/c1024-2048/lo}{+0.0033}
\defval{posdiff/x40/1024/ced-enc-minus-dense-l25/govreport-test/c128-256}{+0.0008}
\defval{posdiff/x40/1024/ced-enc-minus-dense-l25/govreport-test/c128-256/hi}{+0.0016}
\defval{posdiff/x40/1024/ced-enc-minus-dense-l25/govreport-test/c128-256/lo}{+0.0001}
\defval{posdiff/x40/1024/ced-enc-minus-dense-l25/govreport-test/c256-512}{+0.0021}
\defval{posdiff/x40/1024/ced-enc-minus-dense-l25/govreport-test/c256-512/hi}{+0.0026}
\defval{posdiff/x40/1024/ced-enc-minus-dense-l25/govreport-test/c256-512/lo}{+0.0015}
\defval{posdiff/x40/1024/ced-enc-minus-dense-l25/govreport-test/c512-1024}{+0.0032}
\defval{posdiff/x40/1024/ced-enc-minus-dense-l25/govreport-test/c512-1024/hi}{+0.0037}
\defval{posdiff/x40/1024/ced-enc-minus-dense-l25/govreport-test/c512-1024/lo}{+0.0027}
\defval{posdiff/x40/1024/ced-enc-minus-dense-l25/govreport-test/mean}{+0.0030}
\defval{posdiff/x40/1024/ced-enc-minus-dense-l25/govreport-test/mean/hi}{+0.0034}
\defval{posdiff/x40/1024/ced-enc-minus-dense-l25/govreport-test/mean/lo}{+0.0026}
\defval{posdiff/x40/1024/ced-enc-minus-dense-l25/longdocs/c1-128}{+0.0008}
\defval{posdiff/x40/1024/ced-enc-minus-dense-l25/longdocs/c1-128/hi}{+0.0016}
\defval{posdiff/x40/1024/ced-enc-minus-dense-l25/longdocs/c1-128/lo}{$-$0.0000}
\defval{posdiff/x40/1024/ced-enc-minus-dense-l25/longdocs/c1024-2048}{+0.0007}
\defval{posdiff/x40/1024/ced-enc-minus-dense-l25/longdocs/c1024-2048/hi}{+0.0015}
\defval{posdiff/x40/1024/ced-enc-minus-dense-l25/longdocs/c1024-2048/lo}{$-$0.0001}
\defval{posdiff/x40/1024/ced-enc-minus-dense-l25/longdocs/c128-256}{+0.0005}
\defval{posdiff/x40/1024/ced-enc-minus-dense-l25/longdocs/c128-256/hi}{+0.0012}
\defval{posdiff/x40/1024/ced-enc-minus-dense-l25/longdocs/c128-256/lo}{$-$0.0003}
\defval{posdiff/x40/1024/ced-enc-minus-dense-l25/longdocs/c256-512}{+0.0005}
\defval{posdiff/x40/1024/ced-enc-minus-dense-l25/longdocs/c256-512/hi}{+0.0012}
\defval{posdiff/x40/1024/ced-enc-minus-dense-l25/longdocs/c256-512/lo}{$-$0.0002}
\defval{posdiff/x40/1024/ced-enc-minus-dense-l25/longdocs/c512-1024}{+0.0007}
\defval{posdiff/x40/1024/ced-enc-minus-dense-l25/longdocs/c512-1024/hi}{+0.0014}
\defval{posdiff/x40/1024/ced-enc-minus-dense-l25/longdocs/c512-1024/lo}{$-$0.0000}
\defval{posdiff/x40/1024/ced-enc-minus-dense-l25/longdocs/mean}{+0.0007}
\defval{posdiff/x40/1024/ced-enc-minus-dense-l25/longdocs/mean/hi}{+0.0013}
\defval{posdiff/x40/1024/ced-enc-minus-dense-l25/longdocs/mean/lo}{+0.0000}
\defval{posdiff/x40/1024/ced-enc-minus-dense-l25/pg19-test/c1-128}{+0.0024}
\defval{posdiff/x40/1024/ced-enc-minus-dense-l25/pg19-test/c1-128/hi}{+0.0040}
\defval{posdiff/x40/1024/ced-enc-minus-dense-l25/pg19-test/c1-128/lo}{+0.0009}
\defval{posdiff/x40/1024/ced-enc-minus-dense-l25/pg19-test/c1024-2048}{+0.0014}
\defval{posdiff/x40/1024/ced-enc-minus-dense-l25/pg19-test/c1024-2048/hi}{+0.0028}
\defval{posdiff/x40/1024/ced-enc-minus-dense-l25/pg19-test/c1024-2048/lo}{$-$0.0002}
\defval{posdiff/x40/1024/ced-enc-minus-dense-l25/pg19-test/c128-256}{+0.0015}
\defval{posdiff/x40/1024/ced-enc-minus-dense-l25/pg19-test/c128-256/hi}{+0.0031}
\defval{posdiff/x40/1024/ced-enc-minus-dense-l25/pg19-test/c128-256/lo}{+0.0001}
\defval{posdiff/x40/1024/ced-enc-minus-dense-l25/pg19-test/c256-512}{+0.0020}
\defval{posdiff/x40/1024/ced-enc-minus-dense-l25/pg19-test/c256-512/hi}{+0.0033}
\defval{posdiff/x40/1024/ced-enc-minus-dense-l25/pg19-test/c256-512/lo}{+0.0006}
\defval{posdiff/x40/1024/ced-enc-minus-dense-l25/pg19-test/c512-1024}{+0.0019}
\defval{posdiff/x40/1024/ced-enc-minus-dense-l25/pg19-test/c512-1024/hi}{+0.0032}
\defval{posdiff/x40/1024/ced-enc-minus-dense-l25/pg19-test/c512-1024/lo}{+0.0005}
\defval{posdiff/x40/1024/ced-enc-minus-dense-l25/pg19-test/mean}{+0.0016}
\defval{posdiff/x40/1024/ced-enc-minus-dense-l25/pg19-test/mean/hi}{+0.0029}
\defval{posdiff/x40/1024/ced-enc-minus-dense-l25/pg19-test/mean/lo}{+0.0003}
\defval{posdiff/x40/1024/ced-enc-minus-dense-l25/pg19-val/c1-128}{+0.0046}
\defval{posdiff/x40/1024/ced-enc-minus-dense-l25/pg19-val/c1-128/hi}{+0.0069}
\defval{posdiff/x40/1024/ced-enc-minus-dense-l25/pg19-val/c1-128/lo}{+0.0021}
\defval{posdiff/x40/1024/ced-enc-minus-dense-l25/pg19-val/c1024-2048}{+0.0020}
\defval{posdiff/x40/1024/ced-enc-minus-dense-l25/pg19-val/c1024-2048/hi}{+0.0034}
\defval{posdiff/x40/1024/ced-enc-minus-dense-l25/pg19-val/c1024-2048/lo}{+0.0004}
\defval{posdiff/x40/1024/ced-enc-minus-dense-l25/pg19-val/c128-256}{+0.0025}
\defval{posdiff/x40/1024/ced-enc-minus-dense-l25/pg19-val/c128-256/hi}{+0.0049}
\defval{posdiff/x40/1024/ced-enc-minus-dense-l25/pg19-val/c128-256/lo}{$-$0.0002}
\defval{posdiff/x40/1024/ced-enc-minus-dense-l25/pg19-val/c256-512}{+0.0023}
\defval{posdiff/x40/1024/ced-enc-minus-dense-l25/pg19-val/c256-512/hi}{+0.0042}
\defval{posdiff/x40/1024/ced-enc-minus-dense-l25/pg19-val/c256-512/lo}{+0.0003}
\defval{posdiff/x40/1024/ced-enc-minus-dense-l25/pg19-val/c512-1024}{+0.0030}
\defval{posdiff/x40/1024/ced-enc-minus-dense-l25/pg19-val/c512-1024/hi}{+0.0045}
\defval{posdiff/x40/1024/ced-enc-minus-dense-l25/pg19-val/c512-1024/lo}{+0.0014}
\defval{posdiff/x40/1024/ced-enc-minus-dense-l25/pg19-val/mean}{+0.0025}
\defval{posdiff/x40/1024/ced-enc-minus-dense-l25/pg19-val/mean/hi}{+0.0035}
\defval{posdiff/x40/1024/ced-enc-minus-dense-l25/pg19-val/mean/lo}{+0.0012}
\defval{posdiff/x40/1024/ced-enc-minus-yocou-loop/finepdfs-long/c1-128}{+0.0074}
\defval{posdiff/x40/1024/ced-enc-minus-yocou-loop/finepdfs-long/c1-128/hi}{+0.0087}
\defval{posdiff/x40/1024/ced-enc-minus-yocou-loop/finepdfs-long/c1-128/lo}{+0.0060}
\defval{posdiff/x40/1024/ced-enc-minus-yocou-loop/finepdfs-long/c1024-2048}{$-$0.0144}
\defval{posdiff/x40/1024/ced-enc-minus-yocou-loop/finepdfs-long/c1024-2048/hi}{$-$0.0130}
\defval{posdiff/x40/1024/ced-enc-minus-yocou-loop/finepdfs-long/c1024-2048/lo}{$-$0.0159}
\defval{posdiff/x40/1024/ced-enc-minus-yocou-loop/finepdfs-long/c128-256}{+0.0005}
\defval{posdiff/x40/1024/ced-enc-minus-yocou-loop/finepdfs-long/c128-256/hi}{+0.0018}
\defval{posdiff/x40/1024/ced-enc-minus-yocou-loop/finepdfs-long/c128-256/lo}{$-$0.0008}
\defval{posdiff/x40/1024/ced-enc-minus-yocou-loop/finepdfs-long/c256-512}{$-$0.0058}
\defval{posdiff/x40/1024/ced-enc-minus-yocou-loop/finepdfs-long/c256-512/hi}{$-$0.0046}
\defval{posdiff/x40/1024/ced-enc-minus-yocou-loop/finepdfs-long/c256-512/lo}{$-$0.0068}
\defval{posdiff/x40/1024/ced-enc-minus-yocou-loop/finepdfs-long/c512-1024}{$-$0.0106}
\defval{posdiff/x40/1024/ced-enc-minus-yocou-loop/finepdfs-long/c512-1024/hi}{$-$0.0092}
\defval{posdiff/x40/1024/ced-enc-minus-yocou-loop/finepdfs-long/c512-1024/lo}{$-$0.0120}
\defval{posdiff/x40/1024/ced-enc-minus-yocou-loop/finepdfs-long/mean}{$-$0.0100}
\defval{posdiff/x40/1024/ced-enc-minus-yocou-loop/finepdfs-long/mean/hi}{$-$0.0090}
\defval{posdiff/x40/1024/ced-enc-minus-yocou-loop/finepdfs-long/mean/lo}{$-$0.0111}
\defval{posdiff/x40/1024/ced-enc-minus-yocou-loop/fineweb-edu-100bt/c1-128}{$-$0.0005}
\defval{posdiff/x40/1024/ced-enc-minus-yocou-loop/fineweb-edu-100bt/c1-128/hi}{+0.0004}
\defval{posdiff/x40/1024/ced-enc-minus-yocou-loop/fineweb-edu-100bt/c1-128/lo}{$-$0.0015}
\defval{posdiff/x40/1024/ced-enc-minus-yocou-loop/fineweb-edu-100bt/c1024-2048}{$-$0.0071}
\defval{posdiff/x40/1024/ced-enc-minus-yocou-loop/fineweb-edu-100bt/c1024-2048/hi}{$-$0.0061}
\defval{posdiff/x40/1024/ced-enc-minus-yocou-loop/fineweb-edu-100bt/c1024-2048/lo}{$-$0.0081}
\defval{posdiff/x40/1024/ced-enc-minus-yocou-loop/fineweb-edu-100bt/c128-256}{$-$0.0020}
\defval{posdiff/x40/1024/ced-enc-minus-yocou-loop/fineweb-edu-100bt/c128-256/hi}{$-$0.0009}
\defval{posdiff/x40/1024/ced-enc-minus-yocou-loop/fineweb-edu-100bt/c128-256/lo}{$-$0.0029}
\defval{posdiff/x40/1024/ced-enc-minus-yocou-loop/fineweb-edu-100bt/c256-512}{$-$0.0029}
\defval{posdiff/x40/1024/ced-enc-minus-yocou-loop/fineweb-edu-100bt/c256-512/hi}{$-$0.0021}
\defval{posdiff/x40/1024/ced-enc-minus-yocou-loop/fineweb-edu-100bt/c256-512/lo}{$-$0.0039}
\defval{posdiff/x40/1024/ced-enc-minus-yocou-loop/fineweb-edu-100bt/c512-1024}{$-$0.0051}
\defval{posdiff/x40/1024/ced-enc-minus-yocou-loop/fineweb-edu-100bt/c512-1024/hi}{$-$0.0043}
\defval{posdiff/x40/1024/ced-enc-minus-yocou-loop/fineweb-edu-100bt/c512-1024/lo}{$-$0.0060}
\defval{posdiff/x40/1024/ced-enc-minus-yocou-loop/fineweb-edu-100bt/mean}{$-$0.0038}
\defval{posdiff/x40/1024/ced-enc-minus-yocou-loop/fineweb-edu-100bt/mean/hi}{$-$0.0034}
\defval{posdiff/x40/1024/ced-enc-minus-yocou-loop/fineweb-edu-100bt/mean/lo}{$-$0.0043}
\defval{posdiff/x40/1024/ced-enc-minus-yocou-loop/govreport-test/c1-128}{+0.0004}
\defval{posdiff/x40/1024/ced-enc-minus-yocou-loop/govreport-test/c1-128/hi}{+0.0012}
\defval{posdiff/x40/1024/ced-enc-minus-yocou-loop/govreport-test/c1-128/lo}{$-$0.0003}
\defval{posdiff/x40/1024/ced-enc-minus-yocou-loop/govreport-test/c1024-2048}{$-$0.0091}
\defval{posdiff/x40/1024/ced-enc-minus-yocou-loop/govreport-test/c1024-2048/hi}{$-$0.0086}
\defval{posdiff/x40/1024/ced-enc-minus-yocou-loop/govreport-test/c1024-2048/lo}{$-$0.0097}
\defval{posdiff/x40/1024/ced-enc-minus-yocou-loop/govreport-test/c128-256}{$-$0.0013}
\defval{posdiff/x40/1024/ced-enc-minus-yocou-loop/govreport-test/c128-256/hi}{$-$0.0005}
\defval{posdiff/x40/1024/ced-enc-minus-yocou-loop/govreport-test/c128-256/lo}{$-$0.0021}
\defval{posdiff/x40/1024/ced-enc-minus-yocou-loop/govreport-test/c256-512}{$-$0.0044}
\defval{posdiff/x40/1024/ced-enc-minus-yocou-loop/govreport-test/c256-512/hi}{$-$0.0038}
\defval{posdiff/x40/1024/ced-enc-minus-yocou-loop/govreport-test/c256-512/lo}{$-$0.0049}
\defval{posdiff/x40/1024/ced-enc-minus-yocou-loop/govreport-test/c512-1024}{$-$0.0068}
\defval{posdiff/x40/1024/ced-enc-minus-yocou-loop/govreport-test/c512-1024/hi}{$-$0.0062}
\defval{posdiff/x40/1024/ced-enc-minus-yocou-loop/govreport-test/c512-1024/lo}{$-$0.0074}
\defval{posdiff/x40/1024/ced-enc-minus-yocou-loop/govreport-test/mean}{$-$0.0069}
\defval{posdiff/x40/1024/ced-enc-minus-yocou-loop/govreport-test/mean/hi}{$-$0.0064}
\defval{posdiff/x40/1024/ced-enc-minus-yocou-loop/govreport-test/mean/lo}{$-$0.0073}
\defval{posdiff/x40/1024/ced-enc-minus-yocou-loop/longdocs/c1-128}{+0.0050}
\defval{posdiff/x40/1024/ced-enc-minus-yocou-loop/longdocs/c1-128/hi}{+0.0056}
\defval{posdiff/x40/1024/ced-enc-minus-yocou-loop/longdocs/c1-128/lo}{+0.0043}
\defval{posdiff/x40/1024/ced-enc-minus-yocou-loop/longdocs/c1024-2048}{$-$0.0107}
\defval{posdiff/x40/1024/ced-enc-minus-yocou-loop/longdocs/c1024-2048/hi}{$-$0.0101}
\defval{posdiff/x40/1024/ced-enc-minus-yocou-loop/longdocs/c1024-2048/lo}{$-$0.0113}
\defval{posdiff/x40/1024/ced-enc-minus-yocou-loop/longdocs/c128-256}{+0.0003}
\defval{posdiff/x40/1024/ced-enc-minus-yocou-loop/longdocs/c128-256/hi}{+0.0009}
\defval{posdiff/x40/1024/ced-enc-minus-yocou-loop/longdocs/c128-256/lo}{$-$0.0003}
\defval{posdiff/x40/1024/ced-enc-minus-yocou-loop/longdocs/c256-512}{$-$0.0041}
\defval{posdiff/x40/1024/ced-enc-minus-yocou-loop/longdocs/c256-512/hi}{$-$0.0035}
\defval{posdiff/x40/1024/ced-enc-minus-yocou-loop/longdocs/c256-512/lo}{$-$0.0046}
\defval{posdiff/x40/1024/ced-enc-minus-yocou-loop/longdocs/c512-1024}{$-$0.0077}
\defval{posdiff/x40/1024/ced-enc-minus-yocou-loop/longdocs/c512-1024/hi}{$-$0.0072}
\defval{posdiff/x40/1024/ced-enc-minus-yocou-loop/longdocs/c512-1024/lo}{$-$0.0083}
\defval{posdiff/x40/1024/ced-enc-minus-yocou-loop/longdocs/mean}{$-$0.0074}
\defval{posdiff/x40/1024/ced-enc-minus-yocou-loop/longdocs/mean/hi}{$-$0.0070}
\defval{posdiff/x40/1024/ced-enc-minus-yocou-loop/longdocs/mean/lo}{$-$0.0079}
\defval{posdiff/x40/1024/ced-enc-minus-yocou-loop/pg19-test/c1-128}{+0.0071}
\defval{posdiff/x40/1024/ced-enc-minus-yocou-loop/pg19-test/c1-128/hi}{+0.0082}
\defval{posdiff/x40/1024/ced-enc-minus-yocou-loop/pg19-test/c1-128/lo}{+0.0059}
\defval{posdiff/x40/1024/ced-enc-minus-yocou-loop/pg19-test/c1024-2048}{$-$0.0086}
\defval{posdiff/x40/1024/ced-enc-minus-yocou-loop/pg19-test/c1024-2048/hi}{$-$0.0077}
\defval{posdiff/x40/1024/ced-enc-minus-yocou-loop/pg19-test/c1024-2048/lo}{$-$0.0095}
\defval{posdiff/x40/1024/ced-enc-minus-yocou-loop/pg19-test/c128-256}{+0.0017}
\defval{posdiff/x40/1024/ced-enc-minus-yocou-loop/pg19-test/c128-256/hi}{+0.0028}
\defval{posdiff/x40/1024/ced-enc-minus-yocou-loop/pg19-test/c128-256/lo}{+0.0005}
\defval{posdiff/x40/1024/ced-enc-minus-yocou-loop/pg19-test/c256-512}{$-$0.0021}
\defval{posdiff/x40/1024/ced-enc-minus-yocou-loop/pg19-test/c256-512/hi}{$-$0.0011}
\defval{posdiff/x40/1024/ced-enc-minus-yocou-loop/pg19-test/c256-512/lo}{$-$0.0031}
\defval{posdiff/x40/1024/ced-enc-minus-yocou-loop/pg19-test/c512-1024}{$-$0.0059}
\defval{posdiff/x40/1024/ced-enc-minus-yocou-loop/pg19-test/c512-1024/hi}{$-$0.0050}
\defval{posdiff/x40/1024/ced-enc-minus-yocou-loop/pg19-test/c512-1024/lo}{$-$0.0068}
\defval{posdiff/x40/1024/ced-enc-minus-yocou-loop/pg19-test/mean}{$-$0.0055}
\defval{posdiff/x40/1024/ced-enc-minus-yocou-loop/pg19-test/mean/hi}{$-$0.0048}
\defval{posdiff/x40/1024/ced-enc-minus-yocou-loop/pg19-test/mean/lo}{$-$0.0063}
\defval{posdiff/x40/1024/ced-enc-minus-yocou-loop/pg19-val/c1-128}{+0.0069}
\defval{posdiff/x40/1024/ced-enc-minus-yocou-loop/pg19-val/c1-128/hi}{+0.0093}
\defval{posdiff/x40/1024/ced-enc-minus-yocou-loop/pg19-val/c1-128/lo}{+0.0046}
\defval{posdiff/x40/1024/ced-enc-minus-yocou-loop/pg19-val/c1024-2048}{$-$0.0082}
\defval{posdiff/x40/1024/ced-enc-minus-yocou-loop/pg19-val/c1024-2048/hi}{$-$0.0071}
\defval{posdiff/x40/1024/ced-enc-minus-yocou-loop/pg19-val/c1024-2048/lo}{$-$0.0094}
\defval{posdiff/x40/1024/ced-enc-minus-yocou-loop/pg19-val/c128-256}{$-$0.0008}
\defval{posdiff/x40/1024/ced-enc-minus-yocou-loop/pg19-val/c128-256/hi}{+0.0016}
\defval{posdiff/x40/1024/ced-enc-minus-yocou-loop/pg19-val/c128-256/lo}{$-$0.0030}
\defval{posdiff/x40/1024/ced-enc-minus-yocou-loop/pg19-val/c256-512}{$-$0.0039}
\defval{posdiff/x40/1024/ced-enc-minus-yocou-loop/pg19-val/c256-512/hi}{$-$0.0021}
\defval{posdiff/x40/1024/ced-enc-minus-yocou-loop/pg19-val/c256-512/lo}{$-$0.0057}
\defval{posdiff/x40/1024/ced-enc-minus-yocou-loop/pg19-val/c512-1024}{$-$0.0061}
\defval{posdiff/x40/1024/ced-enc-minus-yocou-loop/pg19-val/c512-1024/hi}{$-$0.0048}
\defval{posdiff/x40/1024/ced-enc-minus-yocou-loop/pg19-val/c512-1024/lo}{$-$0.0075}
\defval{posdiff/x40/1024/ced-enc-minus-yocou-loop/pg19-val/mean}{$-$0.0058}
\defval{posdiff/x40/1024/ced-enc-minus-yocou-loop/pg19-val/mean/hi}{$-$0.0050}
\defval{posdiff/x40/1024/ced-enc-minus-yocou-loop/pg19-val/mean/lo}{$-$0.0065}
\defval{posdiff/x40/1024/ced-noloop-minus-ced-dec/finepdfs-long/c1-128}{+0.0069}
\defval{posdiff/x40/1024/ced-noloop-minus-ced-dec/finepdfs-long/c1-128/hi}{+0.0083}
\defval{posdiff/x40/1024/ced-noloop-minus-ced-dec/finepdfs-long/c1-128/lo}{+0.0053}
\defval{posdiff/x40/1024/ced-noloop-minus-ced-dec/finepdfs-long/c1024-2048}{+0.0061}
\defval{posdiff/x40/1024/ced-noloop-minus-ced-dec/finepdfs-long/c1024-2048/hi}{+0.0075}
\defval{posdiff/x40/1024/ced-noloop-minus-ced-dec/finepdfs-long/c1024-2048/lo}{+0.0044}
\defval{posdiff/x40/1024/ced-noloop-minus-ced-dec/finepdfs-long/c128-256}{+0.0104}
\defval{posdiff/x40/1024/ced-noloop-minus-ced-dec/finepdfs-long/c128-256/hi}{+0.0118}
\defval{posdiff/x40/1024/ced-noloop-minus-ced-dec/finepdfs-long/c128-256/lo}{+0.0087}
\defval{posdiff/x40/1024/ced-noloop-minus-ced-dec/finepdfs-long/c256-512}{+0.0087}
\defval{posdiff/x40/1024/ced-noloop-minus-ced-dec/finepdfs-long/c256-512/hi}{+0.0100}
\defval{posdiff/x40/1024/ced-noloop-minus-ced-dec/finepdfs-long/c256-512/lo}{+0.0073}
\defval{posdiff/x40/1024/ced-noloop-minus-ced-dec/finepdfs-long/c512-1024}{+0.0072}
\defval{posdiff/x40/1024/ced-noloop-minus-ced-dec/finepdfs-long/c512-1024/hi}{+0.0086}
\defval{posdiff/x40/1024/ced-noloop-minus-ced-dec/finepdfs-long/c512-1024/lo}{+0.0057}
\defval{posdiff/x40/1024/ced-noloop-minus-ced-dec/finepdfs-long/mean}{+0.0070}
\defval{posdiff/x40/1024/ced-noloop-minus-ced-dec/finepdfs-long/mean/hi}{+0.0083}
\defval{posdiff/x40/1024/ced-noloop-minus-ced-dec/finepdfs-long/mean/lo}{+0.0057}
\defval{posdiff/x40/1024/ced-noloop-minus-ced-dec/fineweb-edu-100bt/c1-128}{+0.0072}
\defval{posdiff/x40/1024/ced-noloop-minus-ced-dec/fineweb-edu-100bt/c1-128/hi}{+0.0082}
\defval{posdiff/x40/1024/ced-noloop-minus-ced-dec/fineweb-edu-100bt/c1-128/lo}{+0.0062}
\defval{posdiff/x40/1024/ced-noloop-minus-ced-dec/fineweb-edu-100bt/c1024-2048}{+0.0065}
\defval{posdiff/x40/1024/ced-noloop-minus-ced-dec/fineweb-edu-100bt/c1024-2048/hi}{+0.0076}
\defval{posdiff/x40/1024/ced-noloop-minus-ced-dec/fineweb-edu-100bt/c1024-2048/lo}{+0.0054}
\defval{posdiff/x40/1024/ced-noloop-minus-ced-dec/fineweb-edu-100bt/c128-256}{+0.0071}
\defval{posdiff/x40/1024/ced-noloop-minus-ced-dec/fineweb-edu-100bt/c128-256/hi}{+0.0083}
\defval{posdiff/x40/1024/ced-noloop-minus-ced-dec/fineweb-edu-100bt/c128-256/lo}{+0.0060}
\defval{posdiff/x40/1024/ced-noloop-minus-ced-dec/fineweb-edu-100bt/c256-512}{+0.0068}
\defval{posdiff/x40/1024/ced-noloop-minus-ced-dec/fineweb-edu-100bt/c256-512/hi}{+0.0076}
\defval{posdiff/x40/1024/ced-noloop-minus-ced-dec/fineweb-edu-100bt/c256-512/lo}{+0.0059}
\defval{posdiff/x40/1024/ced-noloop-minus-ced-dec/fineweb-edu-100bt/c512-1024}{+0.0066}
\defval{posdiff/x40/1024/ced-noloop-minus-ced-dec/fineweb-edu-100bt/c512-1024/hi}{+0.0075}
\defval{posdiff/x40/1024/ced-noloop-minus-ced-dec/fineweb-edu-100bt/c512-1024/lo}{+0.0058}
\defval{posdiff/x40/1024/ced-noloop-minus-ced-dec/fineweb-edu-100bt/mean}{+0.0068}
\defval{posdiff/x40/1024/ced-noloop-minus-ced-dec/fineweb-edu-100bt/mean/hi}{+0.0073}
\defval{posdiff/x40/1024/ced-noloop-minus-ced-dec/fineweb-edu-100bt/mean/lo}{+0.0063}
\defval{posdiff/x40/1024/ced-noloop-minus-ced-dec/govreport-test/c1-128}{+0.0052}
\defval{posdiff/x40/1024/ced-noloop-minus-ced-dec/govreport-test/c1-128/hi}{+0.0060}
\defval{posdiff/x40/1024/ced-noloop-minus-ced-dec/govreport-test/c1-128/lo}{+0.0043}
\defval{posdiff/x40/1024/ced-noloop-minus-ced-dec/govreport-test/c1024-2048}{+0.0043}
\defval{posdiff/x40/1024/ced-noloop-minus-ced-dec/govreport-test/c1024-2048/hi}{+0.0049}
\defval{posdiff/x40/1024/ced-noloop-minus-ced-dec/govreport-test/c1024-2048/lo}{+0.0037}
\defval{posdiff/x40/1024/ced-noloop-minus-ced-dec/govreport-test/c128-256}{+0.0043}
\defval{posdiff/x40/1024/ced-noloop-minus-ced-dec/govreport-test/c128-256/hi}{+0.0051}
\defval{posdiff/x40/1024/ced-noloop-minus-ced-dec/govreport-test/c128-256/lo}{+0.0036}
\defval{posdiff/x40/1024/ced-noloop-minus-ced-dec/govreport-test/c256-512}{+0.0042}
\defval{posdiff/x40/1024/ced-noloop-minus-ced-dec/govreport-test/c256-512/hi}{+0.0048}
\defval{posdiff/x40/1024/ced-noloop-minus-ced-dec/govreport-test/c256-512/lo}{+0.0036}
\defval{posdiff/x40/1024/ced-noloop-minus-ced-dec/govreport-test/c512-1024}{+0.0045}
\defval{posdiff/x40/1024/ced-noloop-minus-ced-dec/govreport-test/c512-1024/hi}{+0.0051}
\defval{posdiff/x40/1024/ced-noloop-minus-ced-dec/govreport-test/c512-1024/lo}{+0.0040}
\defval{posdiff/x40/1024/ced-noloop-minus-ced-dec/govreport-test/mean}{+0.0044}
\defval{posdiff/x40/1024/ced-noloop-minus-ced-dec/govreport-test/mean/hi}{+0.0048}
\defval{posdiff/x40/1024/ced-noloop-minus-ced-dec/govreport-test/mean/lo}{+0.0040}
\defval{posdiff/x40/1024/ced-noloop-minus-ced-dec/longdocs/c1-128}{+0.0079}
\defval{posdiff/x40/1024/ced-noloop-minus-ced-dec/longdocs/c1-128/hi}{+0.0086}
\defval{posdiff/x40/1024/ced-noloop-minus-ced-dec/longdocs/c1-128/lo}{+0.0071}
\defval{posdiff/x40/1024/ced-noloop-minus-ced-dec/longdocs/c1024-2048}{+0.0070}
\defval{posdiff/x40/1024/ced-noloop-minus-ced-dec/longdocs/c1024-2048/hi}{+0.0076}
\defval{posdiff/x40/1024/ced-noloop-minus-ced-dec/longdocs/c1024-2048/lo}{+0.0064}
\defval{posdiff/x40/1024/ced-noloop-minus-ced-dec/longdocs/c128-256}{+0.0089}
\defval{posdiff/x40/1024/ced-noloop-minus-ced-dec/longdocs/c128-256/hi}{+0.0096}
\defval{posdiff/x40/1024/ced-noloop-minus-ced-dec/longdocs/c128-256/lo}{+0.0082}
\defval{posdiff/x40/1024/ced-noloop-minus-ced-dec/longdocs/c256-512}{+0.0084}
\defval{posdiff/x40/1024/ced-noloop-minus-ced-dec/longdocs/c256-512/hi}{+0.0090}
\defval{posdiff/x40/1024/ced-noloop-minus-ced-dec/longdocs/c256-512/lo}{+0.0078}
\defval{posdiff/x40/1024/ced-noloop-minus-ced-dec/longdocs/c512-1024}{+0.0079}
\defval{posdiff/x40/1024/ced-noloop-minus-ced-dec/longdocs/c512-1024/hi}{+0.0085}
\defval{posdiff/x40/1024/ced-noloop-minus-ced-dec/longdocs/c512-1024/lo}{+0.0072}
\defval{posdiff/x40/1024/ced-noloop-minus-ced-dec/longdocs/mean}{+0.0076}
\defval{posdiff/x40/1024/ced-noloop-minus-ced-dec/longdocs/mean/hi}{+0.0081}
\defval{posdiff/x40/1024/ced-noloop-minus-ced-dec/longdocs/mean/lo}{+0.0070}
\defval{posdiff/x40/1024/ced-noloop-minus-ced-dec/pg19-test/c1-128}{+0.0116}
\defval{posdiff/x40/1024/ced-noloop-minus-ced-dec/pg19-test/c1-128/hi}{+0.0127}
\defval{posdiff/x40/1024/ced-noloop-minus-ced-dec/pg19-test/c1-128/lo}{+0.0105}
\defval{posdiff/x40/1024/ced-noloop-minus-ced-dec/pg19-test/c1024-2048}{+0.0108}
\defval{posdiff/x40/1024/ced-noloop-minus-ced-dec/pg19-test/c1024-2048/hi}{+0.0118}
\defval{posdiff/x40/1024/ced-noloop-minus-ced-dec/pg19-test/c1024-2048/lo}{+0.0098}
\defval{posdiff/x40/1024/ced-noloop-minus-ced-dec/pg19-test/c128-256}{+0.0120}
\defval{posdiff/x40/1024/ced-noloop-minus-ced-dec/pg19-test/c128-256/hi}{+0.0132}
\defval{posdiff/x40/1024/ced-noloop-minus-ced-dec/pg19-test/c128-256/lo}{+0.0106}
\defval{posdiff/x40/1024/ced-noloop-minus-ced-dec/pg19-test/c256-512}{+0.0123}
\defval{posdiff/x40/1024/ced-noloop-minus-ced-dec/pg19-test/c256-512/hi}{+0.0135}
\defval{posdiff/x40/1024/ced-noloop-minus-ced-dec/pg19-test/c256-512/lo}{+0.0112}
\defval{posdiff/x40/1024/ced-noloop-minus-ced-dec/pg19-test/c512-1024}{+0.0118}
\defval{posdiff/x40/1024/ced-noloop-minus-ced-dec/pg19-test/c512-1024/hi}{+0.0129}
\defval{posdiff/x40/1024/ced-noloop-minus-ced-dec/pg19-test/c512-1024/lo}{+0.0107}
\defval{posdiff/x40/1024/ced-noloop-minus-ced-dec/pg19-test/mean}{+0.0114}
\defval{posdiff/x40/1024/ced-noloop-minus-ced-dec/pg19-test/mean/hi}{+0.0122}
\defval{posdiff/x40/1024/ced-noloop-minus-ced-dec/pg19-test/mean/lo}{+0.0105}
\defval{posdiff/x40/1024/ced-noloop-minus-ced-dec/pg19-val/c1-128}{+0.0089}
\defval{posdiff/x40/1024/ced-noloop-minus-ced-dec/pg19-val/c1-128/hi}{+0.0111}
\defval{posdiff/x40/1024/ced-noloop-minus-ced-dec/pg19-val/c1-128/lo}{+0.0064}
\defval{posdiff/x40/1024/ced-noloop-minus-ced-dec/pg19-val/c1024-2048}{+0.0115}
\defval{posdiff/x40/1024/ced-noloop-minus-ced-dec/pg19-val/c1024-2048/hi}{+0.0125}
\defval{posdiff/x40/1024/ced-noloop-minus-ced-dec/pg19-val/c1024-2048/lo}{+0.0104}
\defval{posdiff/x40/1024/ced-noloop-minus-ced-dec/pg19-val/c128-256}{+0.0097}
\defval{posdiff/x40/1024/ced-noloop-minus-ced-dec/pg19-val/c128-256/hi}{+0.0120}
\defval{posdiff/x40/1024/ced-noloop-minus-ced-dec/pg19-val/c128-256/lo}{+0.0072}
\defval{posdiff/x40/1024/ced-noloop-minus-ced-dec/pg19-val/c256-512}{+0.0103}
\defval{posdiff/x40/1024/ced-noloop-minus-ced-dec/pg19-val/c256-512/hi}{+0.0121}
\defval{posdiff/x40/1024/ced-noloop-minus-ced-dec/pg19-val/c256-512/lo}{+0.0085}
\defval{posdiff/x40/1024/ced-noloop-minus-ced-dec/pg19-val/c512-1024}{+0.0115}
\defval{posdiff/x40/1024/ced-noloop-minus-ced-dec/pg19-val/c512-1024/hi}{+0.0128}
\defval{posdiff/x40/1024/ced-noloop-minus-ced-dec/pg19-val/c512-1024/lo}{+0.0101}
\defval{posdiff/x40/1024/ced-noloop-minus-ced-dec/pg19-val/mean}{+0.0111}
\defval{posdiff/x40/1024/ced-noloop-minus-ced-dec/pg19-val/mean/hi}{+0.0118}
\defval{posdiff/x40/1024/ced-noloop-minus-ced-dec/pg19-val/mean/lo}{+0.0103}
\defval{posdiff/x40/1024/ced-noloop-minus-ced-enc/finepdfs-long/c1-128}{+0.0293}
\defval{posdiff/x40/1024/ced-noloop-minus-ced-enc/finepdfs-long/c1-128/hi}{+0.0310}
\defval{posdiff/x40/1024/ced-noloop-minus-ced-enc/finepdfs-long/c1-128/lo}{+0.0276}
\defval{posdiff/x40/1024/ced-noloop-minus-ced-enc/finepdfs-long/c1024-2048}{+0.0405}
\defval{posdiff/x40/1024/ced-noloop-minus-ced-enc/finepdfs-long/c1024-2048/hi}{+0.0427}
\defval{posdiff/x40/1024/ced-noloop-minus-ced-enc/finepdfs-long/c1024-2048/lo}{+0.0382}
\defval{posdiff/x40/1024/ced-noloop-minus-ced-enc/finepdfs-long/c128-256}{+0.0353}
\defval{posdiff/x40/1024/ced-noloop-minus-ced-enc/finepdfs-long/c128-256/hi}{+0.0373}
\defval{posdiff/x40/1024/ced-noloop-minus-ced-enc/finepdfs-long/c128-256/lo}{+0.0331}
\defval{posdiff/x40/1024/ced-noloop-minus-ced-enc/finepdfs-long/c256-512}{+0.0371}
\defval{posdiff/x40/1024/ced-noloop-minus-ced-enc/finepdfs-long/c256-512/hi}{+0.0388}
\defval{posdiff/x40/1024/ced-noloop-minus-ced-enc/finepdfs-long/c256-512/lo}{+0.0353}
\defval{posdiff/x40/1024/ced-noloop-minus-ced-enc/finepdfs-long/c512-1024}{+0.0388}
\defval{posdiff/x40/1024/ced-noloop-minus-ced-enc/finepdfs-long/c512-1024/hi}{+0.0407}
\defval{posdiff/x40/1024/ced-noloop-minus-ced-enc/finepdfs-long/c512-1024/lo}{+0.0369}
\defval{posdiff/x40/1024/ced-noloop-minus-ced-enc/finepdfs-long/mean}{+0.0386}
\defval{posdiff/x40/1024/ced-noloop-minus-ced-enc/finepdfs-long/mean/hi}{+0.0403}
\defval{posdiff/x40/1024/ced-noloop-minus-ced-enc/finepdfs-long/mean/lo}{+0.0368}
\defval{posdiff/x40/1024/ced-noloop-minus-ced-enc/fineweb-edu-100bt/c1-128}{+0.0230}
\defval{posdiff/x40/1024/ced-noloop-minus-ced-enc/fineweb-edu-100bt/c1-128/hi}{+0.0241}
\defval{posdiff/x40/1024/ced-noloop-minus-ced-enc/fineweb-edu-100bt/c1-128/lo}{+0.0219}
\defval{posdiff/x40/1024/ced-noloop-minus-ced-enc/fineweb-edu-100bt/c1024-2048}{+0.0272}
\defval{posdiff/x40/1024/ced-noloop-minus-ced-enc/fineweb-edu-100bt/c1024-2048/hi}{+0.0287}
\defval{posdiff/x40/1024/ced-noloop-minus-ced-enc/fineweb-edu-100bt/c1024-2048/lo}{+0.0259}
\defval{posdiff/x40/1024/ced-noloop-minus-ced-enc/fineweb-edu-100bt/c128-256}{+0.0246}
\defval{posdiff/x40/1024/ced-noloop-minus-ced-enc/fineweb-edu-100bt/c128-256/hi}{+0.0259}
\defval{posdiff/x40/1024/ced-noloop-minus-ced-enc/fineweb-edu-100bt/c128-256/lo}{+0.0233}
\defval{posdiff/x40/1024/ced-noloop-minus-ced-enc/fineweb-edu-100bt/c256-512}{+0.0245}
\defval{posdiff/x40/1024/ced-noloop-minus-ced-enc/fineweb-edu-100bt/c256-512/hi}{+0.0255}
\defval{posdiff/x40/1024/ced-noloop-minus-ced-enc/fineweb-edu-100bt/c256-512/lo}{+0.0235}
\defval{posdiff/x40/1024/ced-noloop-minus-ced-enc/fineweb-edu-100bt/c512-1024}{+0.0256}
\defval{posdiff/x40/1024/ced-noloop-minus-ced-enc/fineweb-edu-100bt/c512-1024/hi}{+0.0267}
\defval{posdiff/x40/1024/ced-noloop-minus-ced-enc/fineweb-edu-100bt/c512-1024/lo}{+0.0246}
\defval{posdiff/x40/1024/ced-noloop-minus-ced-enc/fineweb-edu-100bt/mean}{+0.0251}
\defval{posdiff/x40/1024/ced-noloop-minus-ced-enc/fineweb-edu-100bt/mean/hi}{+0.0257}
\defval{posdiff/x40/1024/ced-noloop-minus-ced-enc/fineweb-edu-100bt/mean/lo}{+0.0245}
\defval{posdiff/x40/1024/ced-noloop-minus-ced-enc/govreport-test/c1-128}{+0.0176}
\defval{posdiff/x40/1024/ced-noloop-minus-ced-enc/govreport-test/c1-128/hi}{+0.0185}
\defval{posdiff/x40/1024/ced-noloop-minus-ced-enc/govreport-test/c1-128/lo}{+0.0168}
\defval{posdiff/x40/1024/ced-noloop-minus-ced-enc/govreport-test/c1024-2048}{+0.0243}
\defval{posdiff/x40/1024/ced-noloop-minus-ced-enc/govreport-test/c1024-2048/hi}{+0.0249}
\defval{posdiff/x40/1024/ced-noloop-minus-ced-enc/govreport-test/c1024-2048/lo}{+0.0237}
\defval{posdiff/x40/1024/ced-noloop-minus-ced-enc/govreport-test/c128-256}{+0.0202}
\defval{posdiff/x40/1024/ced-noloop-minus-ced-enc/govreport-test/c128-256/hi}{+0.0211}
\defval{posdiff/x40/1024/ced-noloop-minus-ced-enc/govreport-test/c128-256/lo}{+0.0193}
\defval{posdiff/x40/1024/ced-noloop-minus-ced-enc/govreport-test/c256-512}{+0.0221}
\defval{posdiff/x40/1024/ced-noloop-minus-ced-enc/govreport-test/c256-512/hi}{+0.0228}
\defval{posdiff/x40/1024/ced-noloop-minus-ced-enc/govreport-test/c256-512/lo}{+0.0213}
\defval{posdiff/x40/1024/ced-noloop-minus-ced-enc/govreport-test/c512-1024}{+0.0233}
\defval{posdiff/x40/1024/ced-noloop-minus-ced-enc/govreport-test/c512-1024/hi}{+0.0240}
\defval{posdiff/x40/1024/ced-noloop-minus-ced-enc/govreport-test/c512-1024/lo}{+0.0227}
\defval{posdiff/x40/1024/ced-noloop-minus-ced-enc/govreport-test/mean}{+0.0231}
\defval{posdiff/x40/1024/ced-noloop-minus-ced-enc/govreport-test/mean/hi}{+0.0236}
\defval{posdiff/x40/1024/ced-noloop-minus-ced-enc/govreport-test/mean/lo}{+0.0226}
\defval{posdiff/x40/1024/ced-noloop-minus-ced-enc/longdocs/c1-128}{+0.0277}
\defval{posdiff/x40/1024/ced-noloop-minus-ced-enc/longdocs/c1-128/hi}{+0.0284}
\defval{posdiff/x40/1024/ced-noloop-minus-ced-enc/longdocs/c1-128/lo}{+0.0269}
\defval{posdiff/x40/1024/ced-noloop-minus-ced-enc/longdocs/c1024-2048}{+0.0359}
\defval{posdiff/x40/1024/ced-noloop-minus-ced-enc/longdocs/c1024-2048/hi}{+0.0367}
\defval{posdiff/x40/1024/ced-noloop-minus-ced-enc/longdocs/c1024-2048/lo}{+0.0351}
\defval{posdiff/x40/1024/ced-noloop-minus-ced-enc/longdocs/c128-256}{+0.0323}
\defval{posdiff/x40/1024/ced-noloop-minus-ced-enc/longdocs/c128-256/hi}{+0.0332}
\defval{posdiff/x40/1024/ced-noloop-minus-ced-enc/longdocs/c128-256/lo}{+0.0314}
\defval{posdiff/x40/1024/ced-noloop-minus-ced-enc/longdocs/c256-512}{+0.0338}
\defval{posdiff/x40/1024/ced-noloop-minus-ced-enc/longdocs/c256-512/hi}{+0.0345}
\defval{posdiff/x40/1024/ced-noloop-minus-ced-enc/longdocs/c256-512/lo}{+0.0331}
\defval{posdiff/x40/1024/ced-noloop-minus-ced-enc/longdocs/c512-1024}{+0.0349}
\defval{posdiff/x40/1024/ced-noloop-minus-ced-enc/longdocs/c512-1024/hi}{+0.0355}
\defval{posdiff/x40/1024/ced-noloop-minus-ced-enc/longdocs/c512-1024/lo}{+0.0341}
\defval{posdiff/x40/1024/ced-noloop-minus-ced-enc/longdocs/mean}{+0.0346}
\defval{posdiff/x40/1024/ced-noloop-minus-ced-enc/longdocs/mean/hi}{+0.0353}
\defval{posdiff/x40/1024/ced-noloop-minus-ced-enc/longdocs/mean/lo}{+0.0340}
\defval{posdiff/x40/1024/ced-noloop-minus-ced-enc/pg19-test/c1-128}{+0.0361}
\defval{posdiff/x40/1024/ced-noloop-minus-ced-enc/pg19-test/c1-128/hi}{+0.0374}
\defval{posdiff/x40/1024/ced-noloop-minus-ced-enc/pg19-test/c1-128/lo}{+0.0348}
\defval{posdiff/x40/1024/ced-noloop-minus-ced-enc/pg19-test/c1024-2048}{+0.0430}
\defval{posdiff/x40/1024/ced-noloop-minus-ced-enc/pg19-test/c1024-2048/hi}{+0.0441}
\defval{posdiff/x40/1024/ced-noloop-minus-ced-enc/pg19-test/c1024-2048/lo}{+0.0420}
\defval{posdiff/x40/1024/ced-noloop-minus-ced-enc/pg19-test/c128-256}{+0.0416}
\defval{posdiff/x40/1024/ced-noloop-minus-ced-enc/pg19-test/c128-256/hi}{+0.0430}
\defval{posdiff/x40/1024/ced-noloop-minus-ced-enc/pg19-test/c128-256/lo}{+0.0402}
\defval{posdiff/x40/1024/ced-noloop-minus-ced-enc/pg19-test/c256-512}{+0.0424}
\defval{posdiff/x40/1024/ced-noloop-minus-ced-enc/pg19-test/c256-512/hi}{+0.0435}
\defval{posdiff/x40/1024/ced-noloop-minus-ced-enc/pg19-test/c256-512/lo}{+0.0412}
\defval{posdiff/x40/1024/ced-noloop-minus-ced-enc/pg19-test/c512-1024}{+0.0425}
\defval{posdiff/x40/1024/ced-noloop-minus-ced-enc/pg19-test/c512-1024/hi}{+0.0436}
\defval{posdiff/x40/1024/ced-noloop-minus-ced-enc/pg19-test/c512-1024/lo}{+0.0414}
\defval{posdiff/x40/1024/ced-noloop-minus-ced-enc/pg19-test/mean}{+0.0423}
\defval{posdiff/x40/1024/ced-noloop-minus-ced-enc/pg19-test/mean/hi}{+0.0432}
\defval{posdiff/x40/1024/ced-noloop-minus-ced-enc/pg19-test/mean/lo}{+0.0414}
\defval{posdiff/x40/1024/ced-noloop-minus-ced-enc/pg19-val/c1-128}{+0.0315}
\defval{posdiff/x40/1024/ced-noloop-minus-ced-enc/pg19-val/c1-128/hi}{+0.0342}
\defval{posdiff/x40/1024/ced-noloop-minus-ced-enc/pg19-val/c1-128/lo}{+0.0290}
\defval{posdiff/x40/1024/ced-noloop-minus-ced-enc/pg19-val/c1024-2048}{+0.0408}
\defval{posdiff/x40/1024/ced-noloop-minus-ced-enc/pg19-val/c1024-2048/hi}{+0.0420}
\defval{posdiff/x40/1024/ced-noloop-minus-ced-enc/pg19-val/c1024-2048/lo}{+0.0396}
\defval{posdiff/x40/1024/ced-noloop-minus-ced-enc/pg19-val/c128-256}{+0.0386}
\defval{posdiff/x40/1024/ced-noloop-minus-ced-enc/pg19-val/c128-256/hi}{+0.0410}
\defval{posdiff/x40/1024/ced-noloop-minus-ced-enc/pg19-val/c128-256/lo}{+0.0359}
\defval{posdiff/x40/1024/ced-noloop-minus-ced-enc/pg19-val/c256-512}{+0.0396}
\defval{posdiff/x40/1024/ced-noloop-minus-ced-enc/pg19-val/c256-512/hi}{+0.0415}
\defval{posdiff/x40/1024/ced-noloop-minus-ced-enc/pg19-val/c256-512/lo}{+0.0377}
\defval{posdiff/x40/1024/ced-noloop-minus-ced-enc/pg19-val/c512-1024}{+0.0404}
\defval{posdiff/x40/1024/ced-noloop-minus-ced-enc/pg19-val/c512-1024/hi}{+0.0419}
\defval{posdiff/x40/1024/ced-noloop-minus-ced-enc/pg19-val/c512-1024/lo}{+0.0388}
\defval{posdiff/x40/1024/ced-noloop-minus-ced-enc/pg19-val/mean}{+0.0398}
\defval{posdiff/x40/1024/ced-noloop-minus-ced-enc/pg19-val/mean/hi}{+0.0407}
\defval{posdiff/x40/1024/ced-noloop-minus-ced-enc/pg19-val/mean/lo}{+0.0389}
\defval{posdiff/x40/1024/ced-noloop-minus-dense-noloop/finepdfs-long/c1-128}{$-$0.0037}
\defval{posdiff/x40/1024/ced-noloop-minus-dense-noloop/finepdfs-long/c1-128/hi}{$-$0.0015}
\defval{posdiff/x40/1024/ced-noloop-minus-dense-noloop/finepdfs-long/c1-128/lo}{$-$0.0059}
\defval{posdiff/x40/1024/ced-noloop-minus-dense-noloop/finepdfs-long/c1024-2048}{$-$0.0086}
\defval{posdiff/x40/1024/ced-noloop-minus-dense-noloop/finepdfs-long/c1024-2048/hi}{$-$0.0066}
\defval{posdiff/x40/1024/ced-noloop-minus-dense-noloop/finepdfs-long/c1024-2048/lo}{$-$0.0107}
\defval{posdiff/x40/1024/ced-noloop-minus-dense-noloop/finepdfs-long/c128-256}{$-$0.0061}
\defval{posdiff/x40/1024/ced-noloop-minus-dense-noloop/finepdfs-long/c128-256/hi}{$-$0.0039}
\defval{posdiff/x40/1024/ced-noloop-minus-dense-noloop/finepdfs-long/c128-256/lo}{$-$0.0084}
\defval{posdiff/x40/1024/ced-noloop-minus-dense-noloop/finepdfs-long/c256-512}{$-$0.0086}
\defval{posdiff/x40/1024/ced-noloop-minus-dense-noloop/finepdfs-long/c256-512/hi}{$-$0.0067}
\defval{posdiff/x40/1024/ced-noloop-minus-dense-noloop/finepdfs-long/c256-512/lo}{$-$0.0105}
\defval{posdiff/x40/1024/ced-noloop-minus-dense-noloop/finepdfs-long/c512-1024}{$-$0.0084}
\defval{posdiff/x40/1024/ced-noloop-minus-dense-noloop/finepdfs-long/c512-1024/hi}{$-$0.0066}
\defval{posdiff/x40/1024/ced-noloop-minus-dense-noloop/finepdfs-long/c512-1024/lo}{$-$0.0103}
\defval{posdiff/x40/1024/ced-noloop-minus-dense-noloop/finepdfs-long/mean}{$-$0.0081}
\defval{posdiff/x40/1024/ced-noloop-minus-dense-noloop/finepdfs-long/mean/hi}{$-$0.0063}
\defval{posdiff/x40/1024/ced-noloop-minus-dense-noloop/finepdfs-long/mean/lo}{$-$0.0099}
\defval{posdiff/x40/1024/ced-noloop-minus-dense-noloop/fineweb-edu-100bt/c1-128}{$-$0.0052}
\defval{posdiff/x40/1024/ced-noloop-minus-dense-noloop/fineweb-edu-100bt/c1-128/hi}{$-$0.0042}
\defval{posdiff/x40/1024/ced-noloop-minus-dense-noloop/fineweb-edu-100bt/c1-128/lo}{$-$0.0062}
\defval{posdiff/x40/1024/ced-noloop-minus-dense-noloop/fineweb-edu-100bt/c1024-2048}{$-$0.0025}
\defval{posdiff/x40/1024/ced-noloop-minus-dense-noloop/fineweb-edu-100bt/c1024-2048/hi}{$-$0.0015}
\defval{posdiff/x40/1024/ced-noloop-minus-dense-noloop/fineweb-edu-100bt/c1024-2048/lo}{$-$0.0034}
\defval{posdiff/x40/1024/ced-noloop-minus-dense-noloop/fineweb-edu-100bt/c128-256}{$-$0.0038}
\defval{posdiff/x40/1024/ced-noloop-minus-dense-noloop/fineweb-edu-100bt/c128-256/hi}{$-$0.0026}
\defval{posdiff/x40/1024/ced-noloop-minus-dense-noloop/fineweb-edu-100bt/c128-256/lo}{$-$0.0049}
\defval{posdiff/x40/1024/ced-noloop-minus-dense-noloop/fineweb-edu-100bt/c256-512}{$-$0.0021}
\defval{posdiff/x40/1024/ced-noloop-minus-dense-noloop/fineweb-edu-100bt/c256-512/hi}{$-$0.0012}
\defval{posdiff/x40/1024/ced-noloop-minus-dense-noloop/fineweb-edu-100bt/c256-512/lo}{$-$0.0030}
\defval{posdiff/x40/1024/ced-noloop-minus-dense-noloop/fineweb-edu-100bt/c512-1024}{$-$0.0023}
\defval{posdiff/x40/1024/ced-noloop-minus-dense-noloop/fineweb-edu-100bt/c512-1024/hi}{$-$0.0014}
\defval{posdiff/x40/1024/ced-noloop-minus-dense-noloop/fineweb-edu-100bt/c512-1024/lo}{$-$0.0031}
\defval{posdiff/x40/1024/ced-noloop-minus-dense-noloop/fineweb-edu-100bt/mean}{$-$0.0030}
\defval{posdiff/x40/1024/ced-noloop-minus-dense-noloop/fineweb-edu-100bt/mean/hi}{$-$0.0026}
\defval{posdiff/x40/1024/ced-noloop-minus-dense-noloop/fineweb-edu-100bt/mean/lo}{$-$0.0034}
\defval{posdiff/x40/1024/ced-noloop-minus-dense-noloop/govreport-test/c1-128}{$-$0.0042}
\defval{posdiff/x40/1024/ced-noloop-minus-dense-noloop/govreport-test/c1-128/hi}{$-$0.0034}
\defval{posdiff/x40/1024/ced-noloop-minus-dense-noloop/govreport-test/c1-128/lo}{$-$0.0050}
\defval{posdiff/x40/1024/ced-noloop-minus-dense-noloop/govreport-test/c1024-2048}{$-$0.0001}
\defval{posdiff/x40/1024/ced-noloop-minus-dense-noloop/govreport-test/c1024-2048/hi}{+0.0005}
\defval{posdiff/x40/1024/ced-noloop-minus-dense-noloop/govreport-test/c1024-2048/lo}{$-$0.0005}
\defval{posdiff/x40/1024/ced-noloop-minus-dense-noloop/govreport-test/c128-256}{$-$0.0034}
\defval{posdiff/x40/1024/ced-noloop-minus-dense-noloop/govreport-test/c128-256/hi}{$-$0.0026}
\defval{posdiff/x40/1024/ced-noloop-minus-dense-noloop/govreport-test/c128-256/lo}{$-$0.0041}
\defval{posdiff/x40/1024/ced-noloop-minus-dense-noloop/govreport-test/c256-512}{$-$0.0018}
\defval{posdiff/x40/1024/ced-noloop-minus-dense-noloop/govreport-test/c256-512/hi}{$-$0.0012}
\defval{posdiff/x40/1024/ced-noloop-minus-dense-noloop/govreport-test/c256-512/lo}{$-$0.0024}
\defval{posdiff/x40/1024/ced-noloop-minus-dense-noloop/govreport-test/c512-1024}{$-$0.0006}
\defval{posdiff/x40/1024/ced-noloop-minus-dense-noloop/govreport-test/c512-1024/hi}{$-$0.0000}
\defval{posdiff/x40/1024/ced-noloop-minus-dense-noloop/govreport-test/c512-1024/lo}{$-$0.0011}
\defval{posdiff/x40/1024/ced-noloop-minus-dense-noloop/govreport-test/mean}{$-$0.0009}
\defval{posdiff/x40/1024/ced-noloop-minus-dense-noloop/govreport-test/mean/hi}{$-$0.0005}
\defval{posdiff/x40/1024/ced-noloop-minus-dense-noloop/govreport-test/mean/lo}{$-$0.0013}
\defval{posdiff/x40/1024/ced-noloop-minus-dense-noloop/longdocs/c1-128}{$-$0.0039}
\defval{posdiff/x40/1024/ced-noloop-minus-dense-noloop/longdocs/c1-128/hi}{$-$0.0030}
\defval{posdiff/x40/1024/ced-noloop-minus-dense-noloop/longdocs/c1-128/lo}{$-$0.0049}
\defval{posdiff/x40/1024/ced-noloop-minus-dense-noloop/longdocs/c1024-2048}{$-$0.0063}
\defval{posdiff/x40/1024/ced-noloop-minus-dense-noloop/longdocs/c1024-2048/hi}{$-$0.0055}
\defval{posdiff/x40/1024/ced-noloop-minus-dense-noloop/longdocs/c1024-2048/lo}{$-$0.0072}
\defval{posdiff/x40/1024/ced-noloop-minus-dense-noloop/longdocs/c128-256}{$-$0.0060}
\defval{posdiff/x40/1024/ced-noloop-minus-dense-noloop/longdocs/c128-256/hi}{$-$0.0050}
\defval{posdiff/x40/1024/ced-noloop-minus-dense-noloop/longdocs/c128-256/lo}{$-$0.0069}
\defval{posdiff/x40/1024/ced-noloop-minus-dense-noloop/longdocs/c256-512}{$-$0.0064}
\defval{posdiff/x40/1024/ced-noloop-minus-dense-noloop/longdocs/c256-512/hi}{$-$0.0057}
\defval{posdiff/x40/1024/ced-noloop-minus-dense-noloop/longdocs/c256-512/lo}{$-$0.0072}
\defval{posdiff/x40/1024/ced-noloop-minus-dense-noloop/longdocs/c512-1024}{$-$0.0060}
\defval{posdiff/x40/1024/ced-noloop-minus-dense-noloop/longdocs/c512-1024/hi}{$-$0.0052}
\defval{posdiff/x40/1024/ced-noloop-minus-dense-noloop/longdocs/c512-1024/lo}{$-$0.0068}
\defval{posdiff/x40/1024/ced-noloop-minus-dense-noloop/longdocs/mean}{$-$0.0061}
\defval{posdiff/x40/1024/ced-noloop-minus-dense-noloop/longdocs/mean/hi}{$-$0.0053}
\defval{posdiff/x40/1024/ced-noloop-minus-dense-noloop/longdocs/mean/lo}{$-$0.0068}
\defval{posdiff/x40/1024/ced-noloop-minus-dense-noloop/pg19-test/c1-128}{$-$0.0038}
\defval{posdiff/x40/1024/ced-noloop-minus-dense-noloop/pg19-test/c1-128/hi}{$-$0.0023}
\defval{posdiff/x40/1024/ced-noloop-minus-dense-noloop/pg19-test/c1-128/lo}{$-$0.0054}
\defval{posdiff/x40/1024/ced-noloop-minus-dense-noloop/pg19-test/c1024-2048}{$-$0.0103}
\defval{posdiff/x40/1024/ced-noloop-minus-dense-noloop/pg19-test/c1024-2048/hi}{$-$0.0091}
\defval{posdiff/x40/1024/ced-noloop-minus-dense-noloop/pg19-test/c1024-2048/lo}{$-$0.0117}
\defval{posdiff/x40/1024/ced-noloop-minus-dense-noloop/pg19-test/c128-256}{$-$0.0084}
\defval{posdiff/x40/1024/ced-noloop-minus-dense-noloop/pg19-test/c128-256/hi}{$-$0.0069}
\defval{posdiff/x40/1024/ced-noloop-minus-dense-noloop/pg19-test/c128-256/lo}{$-$0.0102}
\defval{posdiff/x40/1024/ced-noloop-minus-dense-noloop/pg19-test/c256-512}{$-$0.0089}
\defval{posdiff/x40/1024/ced-noloop-minus-dense-noloop/pg19-test/c256-512/hi}{$-$0.0076}
\defval{posdiff/x40/1024/ced-noloop-minus-dense-noloop/pg19-test/c256-512/lo}{$-$0.0104}
\defval{posdiff/x40/1024/ced-noloop-minus-dense-noloop/pg19-test/c512-1024}{$-$0.0090}
\defval{posdiff/x40/1024/ced-noloop-minus-dense-noloop/pg19-test/c512-1024/hi}{$-$0.0077}
\defval{posdiff/x40/1024/ced-noloop-minus-dense-noloop/pg19-test/c512-1024/lo}{$-$0.0104}
\defval{posdiff/x40/1024/ced-noloop-minus-dense-noloop/pg19-test/mean}{$-$0.0093}
\defval{posdiff/x40/1024/ced-noloop-minus-dense-noloop/pg19-test/mean/hi}{$-$0.0082}
\defval{posdiff/x40/1024/ced-noloop-minus-dense-noloop/pg19-test/mean/lo}{$-$0.0105}
\defval{posdiff/x40/1024/ced-noloop-minus-dense-noloop/pg19-val/c1-128}{$-$0.0025}
\defval{posdiff/x40/1024/ced-noloop-minus-dense-noloop/pg19-val/c1-128/hi}{$-$0.0001}
\defval{posdiff/x40/1024/ced-noloop-minus-dense-noloop/pg19-val/c1-128/lo}{$-$0.0051}
\defval{posdiff/x40/1024/ced-noloop-minus-dense-noloop/pg19-val/c1024-2048}{$-$0.0086}
\defval{posdiff/x40/1024/ced-noloop-minus-dense-noloop/pg19-val/c1024-2048/hi}{$-$0.0072}
\defval{posdiff/x40/1024/ced-noloop-minus-dense-noloop/pg19-val/c1024-2048/lo}{$-$0.0102}
\defval{posdiff/x40/1024/ced-noloop-minus-dense-noloop/pg19-val/c128-256}{$-$0.0070}
\defval{posdiff/x40/1024/ced-noloop-minus-dense-noloop/pg19-val/c128-256/hi}{$-$0.0046}
\defval{posdiff/x40/1024/ced-noloop-minus-dense-noloop/pg19-val/c128-256/lo}{$-$0.0095}
\defval{posdiff/x40/1024/ced-noloop-minus-dense-noloop/pg19-val/c256-512}{$-$0.0080}
\defval{posdiff/x40/1024/ced-noloop-minus-dense-noloop/pg19-val/c256-512/hi}{$-$0.0058}
\defval{posdiff/x40/1024/ced-noloop-minus-dense-noloop/pg19-val/c256-512/lo}{$-$0.0104}
\defval{posdiff/x40/1024/ced-noloop-minus-dense-noloop/pg19-val/c512-1024}{$-$0.0082}
\defval{posdiff/x40/1024/ced-noloop-minus-dense-noloop/pg19-val/c512-1024/hi}{$-$0.0065}
\defval{posdiff/x40/1024/ced-noloop-minus-dense-noloop/pg19-val/c512-1024/lo}{$-$0.0101}
\defval{posdiff/x40/1024/ced-noloop-minus-dense-noloop/pg19-val/mean}{$-$0.0079}
\defval{posdiff/x40/1024/ced-noloop-minus-dense-noloop/pg19-val/mean/hi}{$-$0.0068}
\defval{posdiff/x40/1024/ced-noloop-minus-dense-noloop/pg19-val/mean/lo}{$-$0.0093}
\defval{posdiff/x40/1024/ced-noloop-minus-yocou-noloop/finepdfs-long/c1-128}{+0.0148}
\defval{posdiff/x40/1024/ced-noloop-minus-yocou-noloop/finepdfs-long/c1-128/hi}{+0.0164}
\defval{posdiff/x40/1024/ced-noloop-minus-yocou-noloop/finepdfs-long/c1-128/lo}{+0.0132}
\defval{posdiff/x40/1024/ced-noloop-minus-yocou-noloop/finepdfs-long/c1024-2048}{$-$0.0002}
\defval{posdiff/x40/1024/ced-noloop-minus-yocou-noloop/finepdfs-long/c1024-2048/hi}{+0.0014}
\defval{posdiff/x40/1024/ced-noloop-minus-yocou-noloop/finepdfs-long/c1024-2048/lo}{$-$0.0020}
\defval{posdiff/x40/1024/ced-noloop-minus-yocou-noloop/finepdfs-long/c128-256}{+0.0096}
\defval{posdiff/x40/1024/ced-noloop-minus-yocou-noloop/finepdfs-long/c128-256/hi}{+0.0111}
\defval{posdiff/x40/1024/ced-noloop-minus-yocou-noloop/finepdfs-long/c128-256/lo}{+0.0080}
\defval{posdiff/x40/1024/ced-noloop-minus-yocou-noloop/finepdfs-long/c256-512}{+0.0049}
\defval{posdiff/x40/1024/ced-noloop-minus-yocou-noloop/finepdfs-long/c256-512/hi}{+0.0062}
\defval{posdiff/x40/1024/ced-noloop-minus-yocou-noloop/finepdfs-long/c256-512/lo}{+0.0035}
\defval{posdiff/x40/1024/ced-noloop-minus-yocou-noloop/finepdfs-long/c512-1024}{+0.0019}
\defval{posdiff/x40/1024/ced-noloop-minus-yocou-noloop/finepdfs-long/c512-1024/hi}{+0.0033}
\defval{posdiff/x40/1024/ced-noloop-minus-yocou-noloop/finepdfs-long/c512-1024/lo}{+0.0002}
\defval{posdiff/x40/1024/ced-noloop-minus-yocou-noloop/finepdfs-long/mean}{+0.0025}
\defval{posdiff/x40/1024/ced-noloop-minus-yocou-noloop/finepdfs-long/mean/hi}{+0.0039}
\defval{posdiff/x40/1024/ced-noloop-minus-yocou-noloop/finepdfs-long/mean/lo}{+0.0009}
\defval{posdiff/x40/1024/ced-noloop-minus-yocou-noloop/fineweb-edu-100bt/c1-128}{$-$0.0009}
\defval{posdiff/x40/1024/ced-noloop-minus-yocou-noloop/fineweb-edu-100bt/c1-128/hi}{+0.0001}
\defval{posdiff/x40/1024/ced-noloop-minus-yocou-noloop/fineweb-edu-100bt/c1-128/lo}{$-$0.0019}
\defval{posdiff/x40/1024/ced-noloop-minus-yocou-noloop/fineweb-edu-100bt/c1024-2048}{$-$0.0041}
\defval{posdiff/x40/1024/ced-noloop-minus-yocou-noloop/fineweb-edu-100bt/c1024-2048/hi}{$-$0.0031}
\defval{posdiff/x40/1024/ced-noloop-minus-yocou-noloop/fineweb-edu-100bt/c1024-2048/lo}{$-$0.0050}
\defval{posdiff/x40/1024/ced-noloop-minus-yocou-noloop/fineweb-edu-100bt/c128-256}{$-$0.0018}
\defval{posdiff/x40/1024/ced-noloop-minus-yocou-noloop/fineweb-edu-100bt/c128-256/hi}{$-$0.0008}
\defval{posdiff/x40/1024/ced-noloop-minus-yocou-noloop/fineweb-edu-100bt/c128-256/lo}{$-$0.0029}
\defval{posdiff/x40/1024/ced-noloop-minus-yocou-noloop/fineweb-edu-100bt/c256-512}{$-$0.0022}
\defval{posdiff/x40/1024/ced-noloop-minus-yocou-noloop/fineweb-edu-100bt/c256-512/hi}{$-$0.0015}
\defval{posdiff/x40/1024/ced-noloop-minus-yocou-noloop/fineweb-edu-100bt/c256-512/lo}{$-$0.0031}
\defval{posdiff/x40/1024/ced-noloop-minus-yocou-noloop/fineweb-edu-100bt/c512-1024}{$-$0.0037}
\defval{posdiff/x40/1024/ced-noloop-minus-yocou-noloop/fineweb-edu-100bt/c512-1024/hi}{$-$0.0028}
\defval{posdiff/x40/1024/ced-noloop-minus-yocou-noloop/fineweb-edu-100bt/c512-1024/lo}{$-$0.0045}
\defval{posdiff/x40/1024/ced-noloop-minus-yocou-noloop/fineweb-edu-100bt/mean}{$-$0.0027}
\defval{posdiff/x40/1024/ced-noloop-minus-yocou-noloop/fineweb-edu-100bt/mean/hi}{$-$0.0023}
\defval{posdiff/x40/1024/ced-noloop-minus-yocou-noloop/fineweb-edu-100bt/mean/lo}{$-$0.0032}
\defval{posdiff/x40/1024/ced-noloop-minus-yocou-noloop/govreport-test/c1-128}{$-$0.0007}
\defval{posdiff/x40/1024/ced-noloop-minus-yocou-noloop/govreport-test/c1-128/hi}{+0.0001}
\defval{posdiff/x40/1024/ced-noloop-minus-yocou-noloop/govreport-test/c1-128/lo}{$-$0.0014}
\defval{posdiff/x40/1024/ced-noloop-minus-yocou-noloop/govreport-test/c1024-2048}{$-$0.0053}
\defval{posdiff/x40/1024/ced-noloop-minus-yocou-noloop/govreport-test/c1024-2048/hi}{$-$0.0048}
\defval{posdiff/x40/1024/ced-noloop-minus-yocou-noloop/govreport-test/c1024-2048/lo}{$-$0.0059}
\defval{posdiff/x40/1024/ced-noloop-minus-yocou-noloop/govreport-test/c128-256}{$-$0.0030}
\defval{posdiff/x40/1024/ced-noloop-minus-yocou-noloop/govreport-test/c128-256/hi}{$-$0.0023}
\defval{posdiff/x40/1024/ced-noloop-minus-yocou-noloop/govreport-test/c128-256/lo}{$-$0.0038}
\defval{posdiff/x40/1024/ced-noloop-minus-yocou-noloop/govreport-test/c256-512}{$-$0.0037}
\defval{posdiff/x40/1024/ced-noloop-minus-yocou-noloop/govreport-test/c256-512/hi}{$-$0.0032}
\defval{posdiff/x40/1024/ced-noloop-minus-yocou-noloop/govreport-test/c256-512/lo}{$-$0.0043}
\defval{posdiff/x40/1024/ced-noloop-minus-yocou-noloop/govreport-test/c512-1024}{$-$0.0045}
\defval{posdiff/x40/1024/ced-noloop-minus-yocou-noloop/govreport-test/c512-1024/hi}{$-$0.0040}
\defval{posdiff/x40/1024/ced-noloop-minus-yocou-noloop/govreport-test/c512-1024/lo}{$-$0.0051}
\defval{posdiff/x40/1024/ced-noloop-minus-yocou-noloop/govreport-test/mean}{$-$0.0045}
\defval{posdiff/x40/1024/ced-noloop-minus-yocou-noloop/govreport-test/mean/hi}{$-$0.0041}
\defval{posdiff/x40/1024/ced-noloop-minus-yocou-noloop/govreport-test/mean/lo}{$-$0.0049}
\defval{posdiff/x40/1024/ced-noloop-minus-yocou-noloop/longdocs/c1-128}{+0.0074}
\defval{posdiff/x40/1024/ced-noloop-minus-yocou-noloop/longdocs/c1-128/hi}{+0.0081}
\defval{posdiff/x40/1024/ced-noloop-minus-yocou-noloop/longdocs/c1-128/lo}{+0.0067}
\defval{posdiff/x40/1024/ced-noloop-minus-yocou-noloop/longdocs/c1024-2048}{$-$0.0037}
\defval{posdiff/x40/1024/ced-noloop-minus-yocou-noloop/longdocs/c1024-2048/hi}{$-$0.0031}
\defval{posdiff/x40/1024/ced-noloop-minus-yocou-noloop/longdocs/c1024-2048/lo}{$-$0.0043}
\defval{posdiff/x40/1024/ced-noloop-minus-yocou-noloop/longdocs/c128-256}{+0.0035}
\defval{posdiff/x40/1024/ced-noloop-minus-yocou-noloop/longdocs/c128-256/hi}{+0.0042}
\defval{posdiff/x40/1024/ced-noloop-minus-yocou-noloop/longdocs/c128-256/lo}{+0.0028}
\defval{posdiff/x40/1024/ced-noloop-minus-yocou-noloop/longdocs/c256-512}{+0.0004}
\defval{posdiff/x40/1024/ced-noloop-minus-yocou-noloop/longdocs/c256-512/hi}{+0.0009}
\defval{posdiff/x40/1024/ced-noloop-minus-yocou-noloop/longdocs/c256-512/lo}{$-$0.0002}
\defval{posdiff/x40/1024/ced-noloop-minus-yocou-noloop/longdocs/c512-1024}{$-$0.0021}
\defval{posdiff/x40/1024/ced-noloop-minus-yocou-noloop/longdocs/c512-1024/hi}{$-$0.0015}
\defval{posdiff/x40/1024/ced-noloop-minus-yocou-noloop/longdocs/c512-1024/lo}{$-$0.0027}
\defval{posdiff/x40/1024/ced-noloop-minus-yocou-noloop/longdocs/mean}{$-$0.0017}
\defval{posdiff/x40/1024/ced-noloop-minus-yocou-noloop/longdocs/mean/hi}{$-$0.0011}
\defval{posdiff/x40/1024/ced-noloop-minus-yocou-noloop/longdocs/mean/lo}{$-$0.0022}
\defval{posdiff/x40/1024/ced-noloop-minus-yocou-noloop/pg19-test/c1-128}{+0.0082}
\defval{posdiff/x40/1024/ced-noloop-minus-yocou-noloop/pg19-test/c1-128/hi}{+0.0095}
\defval{posdiff/x40/1024/ced-noloop-minus-yocou-noloop/pg19-test/c1-128/lo}{+0.0068}
\defval{posdiff/x40/1024/ced-noloop-minus-yocou-noloop/pg19-test/c1024-2048}{$-$0.0055}
\defval{posdiff/x40/1024/ced-noloop-minus-yocou-noloop/pg19-test/c1024-2048/hi}{$-$0.0046}
\defval{posdiff/x40/1024/ced-noloop-minus-yocou-noloop/pg19-test/c1024-2048/lo}{$-$0.0064}
\defval{posdiff/x40/1024/ced-noloop-minus-yocou-noloop/pg19-test/c128-256}{+0.0040}
\defval{posdiff/x40/1024/ced-noloop-minus-yocou-noloop/pg19-test/c128-256/hi}{+0.0052}
\defval{posdiff/x40/1024/ced-noloop-minus-yocou-noloop/pg19-test/c128-256/lo}{+0.0027}
\defval{posdiff/x40/1024/ced-noloop-minus-yocou-noloop/pg19-test/c256-512}{+0.0000}
\defval{posdiff/x40/1024/ced-noloop-minus-yocou-noloop/pg19-test/c256-512/hi}{+0.0011}
\defval{posdiff/x40/1024/ced-noloop-minus-yocou-noloop/pg19-test/c256-512/lo}{$-$0.0009}
\defval{posdiff/x40/1024/ced-noloop-minus-yocou-noloop/pg19-test/c512-1024}{$-$0.0036}
\defval{posdiff/x40/1024/ced-noloop-minus-yocou-noloop/pg19-test/c512-1024/hi}{$-$0.0027}
\defval{posdiff/x40/1024/ced-noloop-minus-yocou-noloop/pg19-test/c512-1024/lo}{$-$0.0045}
\defval{posdiff/x40/1024/ced-noloop-minus-yocou-noloop/pg19-test/mean}{$-$0.0029}
\defval{posdiff/x40/1024/ced-noloop-minus-yocou-noloop/pg19-test/mean/hi}{$-$0.0021}
\defval{posdiff/x40/1024/ced-noloop-minus-yocou-noloop/pg19-test/mean/lo}{$-$0.0036}
\defval{posdiff/x40/1024/ced-noloop-minus-yocou-noloop/pg19-val/c1-128}{+0.0064}
\defval{posdiff/x40/1024/ced-noloop-minus-yocou-noloop/pg19-val/c1-128/hi}{+0.0086}
\defval{posdiff/x40/1024/ced-noloop-minus-yocou-noloop/pg19-val/c1-128/lo}{+0.0040}
\defval{posdiff/x40/1024/ced-noloop-minus-yocou-noloop/pg19-val/c1024-2048}{$-$0.0050}
\defval{posdiff/x40/1024/ced-noloop-minus-yocou-noloop/pg19-val/c1024-2048/hi}{$-$0.0040}
\defval{posdiff/x40/1024/ced-noloop-minus-yocou-noloop/pg19-val/c1024-2048/lo}{$-$0.0060}
\defval{posdiff/x40/1024/ced-noloop-minus-yocou-noloop/pg19-val/c128-256}{+0.0026}
\defval{posdiff/x40/1024/ced-noloop-minus-yocou-noloop/pg19-val/c128-256/hi}{+0.0048}
\defval{posdiff/x40/1024/ced-noloop-minus-yocou-noloop/pg19-val/c128-256/lo}{+0.0003}
\defval{posdiff/x40/1024/ced-noloop-minus-yocou-noloop/pg19-val/c256-512}{$-$0.0004}
\defval{posdiff/x40/1024/ced-noloop-minus-yocou-noloop/pg19-val/c256-512/hi}{+0.0011}
\defval{posdiff/x40/1024/ced-noloop-minus-yocou-noloop/pg19-val/c256-512/lo}{$-$0.0022}
\defval{posdiff/x40/1024/ced-noloop-minus-yocou-noloop/pg19-val/c512-1024}{$-$0.0037}
\defval{posdiff/x40/1024/ced-noloop-minus-yocou-noloop/pg19-val/c512-1024/hi}{$-$0.0024}
\defval{posdiff/x40/1024/ced-noloop-minus-yocou-noloop/pg19-val/c512-1024/lo}{$-$0.0049}
\defval{posdiff/x40/1024/ced-noloop-minus-yocou-noloop/pg19-val/mean}{$-$0.0029}
\defval{posdiff/x40/1024/ced-noloop-minus-yocou-noloop/pg19-val/mean/hi}{$-$0.0022}
\defval{posdiff/x40/1024/ced-noloop-minus-yocou-noloop/pg19-val/mean/lo}{$-$0.0036}
\defval{posdiff/x40/1024/dense-l47-minus-dense-l25/finepdfs-long/c1-128}{+0.0032}
\defval{posdiff/x40/1024/dense-l47-minus-dense-l25/finepdfs-long/c1-128/hi}{+0.0046}
\defval{posdiff/x40/1024/dense-l47-minus-dense-l25/finepdfs-long/c1-128/lo}{+0.0016}
\defval{posdiff/x40/1024/dense-l47-minus-dense-l25/finepdfs-long/c1024-2048}{+0.0122}
\defval{posdiff/x40/1024/dense-l47-minus-dense-l25/finepdfs-long/c1024-2048/hi}{+0.0133}
\defval{posdiff/x40/1024/dense-l47-minus-dense-l25/finepdfs-long/c1024-2048/lo}{+0.0111}
\defval{posdiff/x40/1024/dense-l47-minus-dense-l25/finepdfs-long/c128-256}{+0.0101}
\defval{posdiff/x40/1024/dense-l47-minus-dense-l25/finepdfs-long/c128-256/hi}{+0.0114}
\defval{posdiff/x40/1024/dense-l47-minus-dense-l25/finepdfs-long/c128-256/lo}{+0.0089}
\defval{posdiff/x40/1024/dense-l47-minus-dense-l25/finepdfs-long/c256-512}{+0.0101}
\defval{posdiff/x40/1024/dense-l47-minus-dense-l25/finepdfs-long/c256-512/hi}{+0.0112}
\defval{posdiff/x40/1024/dense-l47-minus-dense-l25/finepdfs-long/c256-512/lo}{+0.0089}
\defval{posdiff/x40/1024/dense-l47-minus-dense-l25/finepdfs-long/c512-1024}{+0.0111}
\defval{posdiff/x40/1024/dense-l47-minus-dense-l25/finepdfs-long/c512-1024/hi}{+0.0121}
\defval{posdiff/x40/1024/dense-l47-minus-dense-l25/finepdfs-long/c512-1024/lo}{+0.0100}
\defval{posdiff/x40/1024/dense-l47-minus-dense-l25/finepdfs-long/mean}{+0.0110}
\defval{posdiff/x40/1024/dense-l47-minus-dense-l25/finepdfs-long/mean/hi}{+0.0118}
\defval{posdiff/x40/1024/dense-l47-minus-dense-l25/finepdfs-long/mean/lo}{+0.0101}
\defval{posdiff/x40/1024/dense-l47-minus-dense-l25/fineweb-edu-100bt/c1-128}{+0.0047}
\defval{posdiff/x40/1024/dense-l47-minus-dense-l25/fineweb-edu-100bt/c1-128/hi}{+0.0058}
\defval{posdiff/x40/1024/dense-l47-minus-dense-l25/fineweb-edu-100bt/c1-128/lo}{+0.0038}
\defval{posdiff/x40/1024/dense-l47-minus-dense-l25/fineweb-edu-100bt/c1024-2048}{+0.0058}
\defval{posdiff/x40/1024/dense-l47-minus-dense-l25/fineweb-edu-100bt/c1024-2048/hi}{+0.0068}
\defval{posdiff/x40/1024/dense-l47-minus-dense-l25/fineweb-edu-100bt/c1024-2048/lo}{+0.0049}
\defval{posdiff/x40/1024/dense-l47-minus-dense-l25/fineweb-edu-100bt/c128-256}{+0.0048}
\defval{posdiff/x40/1024/dense-l47-minus-dense-l25/fineweb-edu-100bt/c128-256/hi}{+0.0059}
\defval{posdiff/x40/1024/dense-l47-minus-dense-l25/fineweb-edu-100bt/c128-256/lo}{+0.0037}
\defval{posdiff/x40/1024/dense-l47-minus-dense-l25/fineweb-edu-100bt/c256-512}{+0.0055}
\defval{posdiff/x40/1024/dense-l47-minus-dense-l25/fineweb-edu-100bt/c256-512/hi}{+0.0063}
\defval{posdiff/x40/1024/dense-l47-minus-dense-l25/fineweb-edu-100bt/c256-512/lo}{+0.0046}
\defval{posdiff/x40/1024/dense-l47-minus-dense-l25/fineweb-edu-100bt/c512-1024}{+0.0055}
\defval{posdiff/x40/1024/dense-l47-minus-dense-l25/fineweb-edu-100bt/c512-1024/hi}{+0.0063}
\defval{posdiff/x40/1024/dense-l47-minus-dense-l25/fineweb-edu-100bt/c512-1024/lo}{+0.0047}
\defval{posdiff/x40/1024/dense-l47-minus-dense-l25/fineweb-edu-100bt/mean}{+0.0053}
\defval{posdiff/x40/1024/dense-l47-minus-dense-l25/fineweb-edu-100bt/mean/hi}{+0.0058}
\defval{posdiff/x40/1024/dense-l47-minus-dense-l25/fineweb-edu-100bt/mean/lo}{+0.0049}
\defval{posdiff/x40/1024/dense-l47-minus-dense-l25/govreport-test/c1-128}{+0.0039}
\defval{posdiff/x40/1024/dense-l47-minus-dense-l25/govreport-test/c1-128/hi}{+0.0047}
\defval{posdiff/x40/1024/dense-l47-minus-dense-l25/govreport-test/c1-128/lo}{+0.0032}
\defval{posdiff/x40/1024/dense-l47-minus-dense-l25/govreport-test/c1024-2048}{+0.0065}
\defval{posdiff/x40/1024/dense-l47-minus-dense-l25/govreport-test/c1024-2048/hi}{+0.0070}
\defval{posdiff/x40/1024/dense-l47-minus-dense-l25/govreport-test/c1024-2048/lo}{+0.0060}
\defval{posdiff/x40/1024/dense-l47-minus-dense-l25/govreport-test/c128-256}{+0.0046}
\defval{posdiff/x40/1024/dense-l47-minus-dense-l25/govreport-test/c128-256/hi}{+0.0052}
\defval{posdiff/x40/1024/dense-l47-minus-dense-l25/govreport-test/c128-256/lo}{+0.0039}
\defval{posdiff/x40/1024/dense-l47-minus-dense-l25/govreport-test/c256-512}{+0.0056}
\defval{posdiff/x40/1024/dense-l47-minus-dense-l25/govreport-test/c256-512/hi}{+0.0062}
\defval{posdiff/x40/1024/dense-l47-minus-dense-l25/govreport-test/c256-512/lo}{+0.0051}
\defval{posdiff/x40/1024/dense-l47-minus-dense-l25/govreport-test/c512-1024}{+0.0061}
\defval{posdiff/x40/1024/dense-l47-minus-dense-l25/govreport-test/c512-1024/hi}{+0.0066}
\defval{posdiff/x40/1024/dense-l47-minus-dense-l25/govreport-test/c512-1024/lo}{+0.0056}
\defval{posdiff/x40/1024/dense-l47-minus-dense-l25/govreport-test/mean}{+0.0060}
\defval{posdiff/x40/1024/dense-l47-minus-dense-l25/govreport-test/mean/hi}{+0.0063}
\defval{posdiff/x40/1024/dense-l47-minus-dense-l25/govreport-test/mean/lo}{+0.0056}
\defval{posdiff/x40/1024/dense-l47-minus-dense-l25/longdocs/c1-128}{+0.0041}
\defval{posdiff/x40/1024/dense-l47-minus-dense-l25/longdocs/c1-128/hi}{+0.0048}
\defval{posdiff/x40/1024/dense-l47-minus-dense-l25/longdocs/c1-128/lo}{+0.0033}
\defval{posdiff/x40/1024/dense-l47-minus-dense-l25/longdocs/c1024-2048}{+0.0091}
\defval{posdiff/x40/1024/dense-l47-minus-dense-l25/longdocs/c1024-2048/hi}{+0.0097}
\defval{posdiff/x40/1024/dense-l47-minus-dense-l25/longdocs/c1024-2048/lo}{+0.0086}
\defval{posdiff/x40/1024/dense-l47-minus-dense-l25/longdocs/c128-256}{+0.0075}
\defval{posdiff/x40/1024/dense-l47-minus-dense-l25/longdocs/c128-256/hi}{+0.0082}
\defval{posdiff/x40/1024/dense-l47-minus-dense-l25/longdocs/c128-256/lo}{+0.0068}
\defval{posdiff/x40/1024/dense-l47-minus-dense-l25/longdocs/c256-512}{+0.0079}
\defval{posdiff/x40/1024/dense-l47-minus-dense-l25/longdocs/c256-512/hi}{+0.0085}
\defval{posdiff/x40/1024/dense-l47-minus-dense-l25/longdocs/c256-512/lo}{+0.0073}
\defval{posdiff/x40/1024/dense-l47-minus-dense-l25/longdocs/c512-1024}{+0.0085}
\defval{posdiff/x40/1024/dense-l47-minus-dense-l25/longdocs/c512-1024/hi}{+0.0090}
\defval{posdiff/x40/1024/dense-l47-minus-dense-l25/longdocs/c512-1024/lo}{+0.0079}
\defval{posdiff/x40/1024/dense-l47-minus-dense-l25/longdocs/mean}{+0.0084}
\defval{posdiff/x40/1024/dense-l47-minus-dense-l25/longdocs/mean/hi}{+0.0088}
\defval{posdiff/x40/1024/dense-l47-minus-dense-l25/longdocs/mean/lo}{+0.0079}
\defval{posdiff/x40/1024/dense-l47-minus-dense-l25/pg19-test/c1-128}{+0.0051}
\defval{posdiff/x40/1024/dense-l47-minus-dense-l25/pg19-test/c1-128/hi}{+0.0064}
\defval{posdiff/x40/1024/dense-l47-minus-dense-l25/pg19-test/c1-128/lo}{+0.0038}
\defval{posdiff/x40/1024/dense-l47-minus-dense-l25/pg19-test/c1024-2048}{+0.0088}
\defval{posdiff/x40/1024/dense-l47-minus-dense-l25/pg19-test/c1024-2048/hi}{+0.0099}
\defval{posdiff/x40/1024/dense-l47-minus-dense-l25/pg19-test/c1024-2048/lo}{+0.0077}
\defval{posdiff/x40/1024/dense-l47-minus-dense-l25/pg19-test/c128-256}{+0.0078}
\defval{posdiff/x40/1024/dense-l47-minus-dense-l25/pg19-test/c128-256/hi}{+0.0092}
\defval{posdiff/x40/1024/dense-l47-minus-dense-l25/pg19-test/c128-256/lo}{+0.0063}
\defval{posdiff/x40/1024/dense-l47-minus-dense-l25/pg19-test/c256-512}{+0.0081}
\defval{posdiff/x40/1024/dense-l47-minus-dense-l25/pg19-test/c256-512/hi}{+0.0093}
\defval{posdiff/x40/1024/dense-l47-minus-dense-l25/pg19-test/c256-512/lo}{+0.0069}
\defval{posdiff/x40/1024/dense-l47-minus-dense-l25/pg19-test/c512-1024}{+0.0084}
\defval{posdiff/x40/1024/dense-l47-minus-dense-l25/pg19-test/c512-1024/hi}{+0.0095}
\defval{posdiff/x40/1024/dense-l47-minus-dense-l25/pg19-test/c512-1024/lo}{+0.0073}
\defval{posdiff/x40/1024/dense-l47-minus-dense-l25/pg19-test/mean}{+0.0083}
\defval{posdiff/x40/1024/dense-l47-minus-dense-l25/pg19-test/mean/hi}{+0.0093}
\defval{posdiff/x40/1024/dense-l47-minus-dense-l25/pg19-test/mean/lo}{+0.0073}
\defval{posdiff/x40/1024/dense-l47-minus-dense-l25/pg19-val/c1-128}{+0.0081}
\defval{posdiff/x40/1024/dense-l47-minus-dense-l25/pg19-val/c1-128/hi}{+0.0105}
\defval{posdiff/x40/1024/dense-l47-minus-dense-l25/pg19-val/c1-128/lo}{+0.0058}
\defval{posdiff/x40/1024/dense-l47-minus-dense-l25/pg19-val/c1024-2048}{+0.0086}
\defval{posdiff/x40/1024/dense-l47-minus-dense-l25/pg19-val/c1024-2048/hi}{+0.0096}
\defval{posdiff/x40/1024/dense-l47-minus-dense-l25/pg19-val/c1024-2048/lo}{+0.0074}
\defval{posdiff/x40/1024/dense-l47-minus-dense-l25/pg19-val/c128-256}{+0.0097}
\defval{posdiff/x40/1024/dense-l47-minus-dense-l25/pg19-val/c128-256/hi}{+0.0122}
\defval{posdiff/x40/1024/dense-l47-minus-dense-l25/pg19-val/c128-256/lo}{+0.0072}
\defval{posdiff/x40/1024/dense-l47-minus-dense-l25/pg19-val/c256-512}{+0.0080}
\defval{posdiff/x40/1024/dense-l47-minus-dense-l25/pg19-val/c256-512/hi}{+0.0098}
\defval{posdiff/x40/1024/dense-l47-minus-dense-l25/pg19-val/c256-512/lo}{+0.0061}
\defval{posdiff/x40/1024/dense-l47-minus-dense-l25/pg19-val/c512-1024}{+0.0099}
\defval{posdiff/x40/1024/dense-l47-minus-dense-l25/pg19-val/c512-1024/hi}{+0.0112}
\defval{posdiff/x40/1024/dense-l47-minus-dense-l25/pg19-val/c512-1024/lo}{+0.0085}
\defval{posdiff/x40/1024/dense-l47-minus-dense-l25/pg19-val/mean}{+0.0089}
\defval{posdiff/x40/1024/dense-l47-minus-dense-l25/pg19-val/mean/hi}{+0.0097}
\defval{posdiff/x40/1024/dense-l47-minus-dense-l25/pg19-val/mean/lo}{+0.0080}
\defval{posdiff/x40/1024/dense-noloop-minus-dense-l25/finepdfs-long/c1-128}{+0.0331}
\defval{posdiff/x40/1024/dense-noloop-minus-dense-l25/finepdfs-long/c1-128/hi}{+0.0351}
\defval{posdiff/x40/1024/dense-noloop-minus-dense-l25/finepdfs-long/c1-128/lo}{+0.0312}
\defval{posdiff/x40/1024/dense-noloop-minus-dense-l25/finepdfs-long/c1024-2048}{+0.0459}
\defval{posdiff/x40/1024/dense-noloop-minus-dense-l25/finepdfs-long/c1024-2048/hi}{+0.0479}
\defval{posdiff/x40/1024/dense-noloop-minus-dense-l25/finepdfs-long/c1024-2048/lo}{+0.0439}
\defval{posdiff/x40/1024/dense-noloop-minus-dense-l25/finepdfs-long/c128-256}{+0.0405}
\defval{posdiff/x40/1024/dense-noloop-minus-dense-l25/finepdfs-long/c128-256/hi}{+0.0425}
\defval{posdiff/x40/1024/dense-noloop-minus-dense-l25/finepdfs-long/c128-256/lo}{+0.0387}
\defval{posdiff/x40/1024/dense-noloop-minus-dense-l25/finepdfs-long/c256-512}{+0.0432}
\defval{posdiff/x40/1024/dense-noloop-minus-dense-l25/finepdfs-long/c256-512/hi}{+0.0451}
\defval{posdiff/x40/1024/dense-noloop-minus-dense-l25/finepdfs-long/c256-512/lo}{+0.0415}
\defval{posdiff/x40/1024/dense-noloop-minus-dense-l25/finepdfs-long/c512-1024}{+0.0442}
\defval{posdiff/x40/1024/dense-noloop-minus-dense-l25/finepdfs-long/c512-1024/hi}{+0.0460}
\defval{posdiff/x40/1024/dense-noloop-minus-dense-l25/finepdfs-long/c512-1024/lo}{+0.0425}
\defval{posdiff/x40/1024/dense-noloop-minus-dense-l25/finepdfs-long/mean}{+0.0440}
\defval{posdiff/x40/1024/dense-noloop-minus-dense-l25/finepdfs-long/mean/hi}{+0.0457}
\defval{posdiff/x40/1024/dense-noloop-minus-dense-l25/finepdfs-long/mean/lo}{+0.0424}
\defval{posdiff/x40/1024/dense-noloop-minus-dense-l25/fineweb-edu-100bt/c1-128}{+0.0260}
\defval{posdiff/x40/1024/dense-noloop-minus-dense-l25/fineweb-edu-100bt/c1-128/hi}{+0.0271}
\defval{posdiff/x40/1024/dense-noloop-minus-dense-l25/fineweb-edu-100bt/c1-128/lo}{+0.0249}
\defval{posdiff/x40/1024/dense-noloop-minus-dense-l25/fineweb-edu-100bt/c1024-2048}{+0.0317}
\defval{posdiff/x40/1024/dense-noloop-minus-dense-l25/fineweb-edu-100bt/c1024-2048/hi}{+0.0330}
\defval{posdiff/x40/1024/dense-noloop-minus-dense-l25/fineweb-edu-100bt/c1024-2048/lo}{+0.0305}
\defval{posdiff/x40/1024/dense-noloop-minus-dense-l25/fineweb-edu-100bt/c128-256}{+0.0287}
\defval{posdiff/x40/1024/dense-noloop-minus-dense-l25/fineweb-edu-100bt/c128-256/hi}{+0.0300}
\defval{posdiff/x40/1024/dense-noloop-minus-dense-l25/fineweb-edu-100bt/c128-256/lo}{+0.0274}
\defval{posdiff/x40/1024/dense-noloop-minus-dense-l25/fineweb-edu-100bt/c256-512}{+0.0282}
\defval{posdiff/x40/1024/dense-noloop-minus-dense-l25/fineweb-edu-100bt/c256-512/hi}{+0.0293}
\defval{posdiff/x40/1024/dense-noloop-minus-dense-l25/fineweb-edu-100bt/c256-512/lo}{+0.0270}
\defval{posdiff/x40/1024/dense-noloop-minus-dense-l25/fineweb-edu-100bt/c512-1024}{+0.0292}
\defval{posdiff/x40/1024/dense-noloop-minus-dense-l25/fineweb-edu-100bt/c512-1024/hi}{+0.0304}
\defval{posdiff/x40/1024/dense-noloop-minus-dense-l25/fineweb-edu-100bt/c512-1024/lo}{+0.0280}
\defval{posdiff/x40/1024/dense-noloop-minus-dense-l25/fineweb-edu-100bt/mean}{+0.0289}
\defval{posdiff/x40/1024/dense-noloop-minus-dense-l25/fineweb-edu-100bt/mean/hi}{+0.0295}
\defval{posdiff/x40/1024/dense-noloop-minus-dense-l25/fineweb-edu-100bt/mean/lo}{+0.0282}
\defval{posdiff/x40/1024/dense-noloop-minus-dense-l25/govreport-test/c1-128}{+0.0216}
\defval{posdiff/x40/1024/dense-noloop-minus-dense-l25/govreport-test/c1-128/hi}{+0.0225}
\defval{posdiff/x40/1024/dense-noloop-minus-dense-l25/govreport-test/c1-128/lo}{+0.0207}
\defval{posdiff/x40/1024/dense-noloop-minus-dense-l25/govreport-test/c1024-2048}{+0.0281}
\defval{posdiff/x40/1024/dense-noloop-minus-dense-l25/govreport-test/c1024-2048/hi}{+0.0288}
\defval{posdiff/x40/1024/dense-noloop-minus-dense-l25/govreport-test/c1024-2048/lo}{+0.0275}
\defval{posdiff/x40/1024/dense-noloop-minus-dense-l25/govreport-test/c128-256}{+0.0244}
\defval{posdiff/x40/1024/dense-noloop-minus-dense-l25/govreport-test/c128-256/hi}{+0.0252}
\defval{posdiff/x40/1024/dense-noloop-minus-dense-l25/govreport-test/c128-256/lo}{+0.0234}
\defval{posdiff/x40/1024/dense-noloop-minus-dense-l25/govreport-test/c256-512}{+0.0259}
\defval{posdiff/x40/1024/dense-noloop-minus-dense-l25/govreport-test/c256-512/hi}{+0.0267}
\defval{posdiff/x40/1024/dense-noloop-minus-dense-l25/govreport-test/c256-512/lo}{+0.0252}
\defval{posdiff/x40/1024/dense-noloop-minus-dense-l25/govreport-test/c512-1024}{+0.0271}
\defval{posdiff/x40/1024/dense-noloop-minus-dense-l25/govreport-test/c512-1024/hi}{+0.0277}
\defval{posdiff/x40/1024/dense-noloop-minus-dense-l25/govreport-test/c512-1024/lo}{+0.0264}
\defval{posdiff/x40/1024/dense-noloop-minus-dense-l25/govreport-test/mean}{+0.0270}
\defval{posdiff/x40/1024/dense-noloop-minus-dense-l25/govreport-test/mean/hi}{+0.0275}
\defval{posdiff/x40/1024/dense-noloop-minus-dense-l25/govreport-test/mean/lo}{+0.0265}
\defval{posdiff/x40/1024/dense-noloop-minus-dense-l25/longdocs/c1-128}{+0.0324}
\defval{posdiff/x40/1024/dense-noloop-minus-dense-l25/longdocs/c1-128/hi}{+0.0332}
\defval{posdiff/x40/1024/dense-noloop-minus-dense-l25/longdocs/c1-128/lo}{+0.0315}
\defval{posdiff/x40/1024/dense-noloop-minus-dense-l25/longdocs/c1024-2048}{+0.0429}
\defval{posdiff/x40/1024/dense-noloop-minus-dense-l25/longdocs/c1024-2048/hi}{+0.0438}
\defval{posdiff/x40/1024/dense-noloop-minus-dense-l25/longdocs/c1024-2048/lo}{+0.0420}
\defval{posdiff/x40/1024/dense-noloop-minus-dense-l25/longdocs/c128-256}{+0.0388}
\defval{posdiff/x40/1024/dense-noloop-minus-dense-l25/longdocs/c128-256/hi}{+0.0397}
\defval{posdiff/x40/1024/dense-noloop-minus-dense-l25/longdocs/c128-256/lo}{+0.0379}
\defval{posdiff/x40/1024/dense-noloop-minus-dense-l25/longdocs/c256-512}{+0.0408}
\defval{posdiff/x40/1024/dense-noloop-minus-dense-l25/longdocs/c256-512/hi}{+0.0415}
\defval{posdiff/x40/1024/dense-noloop-minus-dense-l25/longdocs/c256-512/lo}{+0.0400}
\defval{posdiff/x40/1024/dense-noloop-minus-dense-l25/longdocs/c512-1024}{+0.0416}
\defval{posdiff/x40/1024/dense-noloop-minus-dense-l25/longdocs/c512-1024/hi}{+0.0423}
\defval{posdiff/x40/1024/dense-noloop-minus-dense-l25/longdocs/c512-1024/lo}{+0.0408}
\defval{posdiff/x40/1024/dense-noloop-minus-dense-l25/longdocs/mean}{+0.0414}
\defval{posdiff/x40/1024/dense-noloop-minus-dense-l25/longdocs/mean/hi}{+0.0421}
\defval{posdiff/x40/1024/dense-noloop-minus-dense-l25/longdocs/mean/lo}{+0.0407}
\defval{posdiff/x40/1024/dense-noloop-minus-dense-l25/pg19-test/c1-128}{+0.0424}
\defval{posdiff/x40/1024/dense-noloop-minus-dense-l25/pg19-test/c1-128/hi}{+0.0438}
\defval{posdiff/x40/1024/dense-noloop-minus-dense-l25/pg19-test/c1-128/lo}{+0.0409}
\defval{posdiff/x40/1024/dense-noloop-minus-dense-l25/pg19-test/c1024-2048}{+0.0548}
\defval{posdiff/x40/1024/dense-noloop-minus-dense-l25/pg19-test/c1024-2048/hi}{+0.0562}
\defval{posdiff/x40/1024/dense-noloop-minus-dense-l25/pg19-test/c1024-2048/lo}{+0.0533}
\defval{posdiff/x40/1024/dense-noloop-minus-dense-l25/pg19-test/c128-256}{+0.0515}
\defval{posdiff/x40/1024/dense-noloop-minus-dense-l25/pg19-test/c128-256/hi}{+0.0532}
\defval{posdiff/x40/1024/dense-noloop-minus-dense-l25/pg19-test/c128-256/lo}{+0.0497}
\defval{posdiff/x40/1024/dense-noloop-minus-dense-l25/pg19-test/c256-512}{+0.0532}
\defval{posdiff/x40/1024/dense-noloop-minus-dense-l25/pg19-test/c256-512/hi}{+0.0546}
\defval{posdiff/x40/1024/dense-noloop-minus-dense-l25/pg19-test/c256-512/lo}{+0.0519}
\defval{posdiff/x40/1024/dense-noloop-minus-dense-l25/pg19-test/c512-1024}{+0.0534}
\defval{posdiff/x40/1024/dense-noloop-minus-dense-l25/pg19-test/c512-1024/hi}{+0.0548}
\defval{posdiff/x40/1024/dense-noloop-minus-dense-l25/pg19-test/c512-1024/lo}{+0.0521}
\defval{posdiff/x40/1024/dense-noloop-minus-dense-l25/pg19-test/mean}{+0.0533}
\defval{posdiff/x40/1024/dense-noloop-minus-dense-l25/pg19-test/mean/hi}{+0.0545}
\defval{posdiff/x40/1024/dense-noloop-minus-dense-l25/pg19-test/mean/lo}{+0.0520}
\defval{posdiff/x40/1024/dense-noloop-minus-dense-l25/pg19-val/c1-128}{+0.0387}
\defval{posdiff/x40/1024/dense-noloop-minus-dense-l25/pg19-val/c1-128/hi}{+0.0414}
\defval{posdiff/x40/1024/dense-noloop-minus-dense-l25/pg19-val/c1-128/lo}{+0.0362}
\defval{posdiff/x40/1024/dense-noloop-minus-dense-l25/pg19-val/c1024-2048}{+0.0514}
\defval{posdiff/x40/1024/dense-noloop-minus-dense-l25/pg19-val/c1024-2048/hi}{+0.0528}
\defval{posdiff/x40/1024/dense-noloop-minus-dense-l25/pg19-val/c1024-2048/lo}{+0.0499}
\defval{posdiff/x40/1024/dense-noloop-minus-dense-l25/pg19-val/c128-256}{+0.0480}
\defval{posdiff/x40/1024/dense-noloop-minus-dense-l25/pg19-val/c128-256/hi}{+0.0506}
\defval{posdiff/x40/1024/dense-noloop-minus-dense-l25/pg19-val/c128-256/lo}{+0.0453}
\defval{posdiff/x40/1024/dense-noloop-minus-dense-l25/pg19-val/c256-512}{+0.0499}
\defval{posdiff/x40/1024/dense-noloop-minus-dense-l25/pg19-val/c256-512/hi}{+0.0522}
\defval{posdiff/x40/1024/dense-noloop-minus-dense-l25/pg19-val/c256-512/lo}{+0.0477}
\defval{posdiff/x40/1024/dense-noloop-minus-dense-l25/pg19-val/c512-1024}{+0.0516}
\defval{posdiff/x40/1024/dense-noloop-minus-dense-l25/pg19-val/c512-1024/hi}{+0.0534}
\defval{posdiff/x40/1024/dense-noloop-minus-dense-l25/pg19-val/c512-1024/lo}{+0.0498}
\defval{posdiff/x40/1024/dense-noloop-minus-dense-l25/pg19-val/mean}{+0.0503}
\defval{posdiff/x40/1024/dense-noloop-minus-dense-l25/pg19-val/mean/hi}{+0.0514}
\defval{posdiff/x40/1024/dense-noloop-minus-dense-l25/pg19-val/mean/lo}{+0.0490}
\defval{posdiff/x40/1024/yocou-loop-minus-ced-enc/finepdfs-long/c1-128}{$-$0.0074}
\defval{posdiff/x40/1024/yocou-loop-minus-ced-enc/finepdfs-long/c1-128/hi}{$-$0.0060}
\defval{posdiff/x40/1024/yocou-loop-minus-ced-enc/finepdfs-long/c1-128/lo}{$-$0.0087}
\defval{posdiff/x40/1024/yocou-loop-minus-ced-enc/finepdfs-long/c1024-2048}{+0.0144}
\defval{posdiff/x40/1024/yocou-loop-minus-ced-enc/finepdfs-long/c1024-2048/hi}{+0.0159}
\defval{posdiff/x40/1024/yocou-loop-minus-ced-enc/finepdfs-long/c1024-2048/lo}{+0.0130}
\defval{posdiff/x40/1024/yocou-loop-minus-ced-enc/finepdfs-long/c128-256}{$-$0.0005}
\defval{posdiff/x40/1024/yocou-loop-minus-ced-enc/finepdfs-long/c128-256/hi}{+0.0008}
\defval{posdiff/x40/1024/yocou-loop-minus-ced-enc/finepdfs-long/c128-256/lo}{$-$0.0018}
\defval{posdiff/x40/1024/yocou-loop-minus-ced-enc/finepdfs-long/c256-512}{+0.0058}
\defval{posdiff/x40/1024/yocou-loop-minus-ced-enc/finepdfs-long/c256-512/hi}{+0.0068}
\defval{posdiff/x40/1024/yocou-loop-minus-ced-enc/finepdfs-long/c256-512/lo}{+0.0046}
\defval{posdiff/x40/1024/yocou-loop-minus-ced-enc/finepdfs-long/c512-1024}{+0.0106}
\defval{posdiff/x40/1024/yocou-loop-minus-ced-enc/finepdfs-long/c512-1024/hi}{+0.0120}
\defval{posdiff/x40/1024/yocou-loop-minus-ced-enc/finepdfs-long/c512-1024/lo}{+0.0092}
\defval{posdiff/x40/1024/yocou-loop-minus-ced-enc/finepdfs-long/mean}{+0.0100}
\defval{posdiff/x40/1024/yocou-loop-minus-ced-enc/finepdfs-long/mean/hi}{+0.0111}
\defval{posdiff/x40/1024/yocou-loop-minus-ced-enc/finepdfs-long/mean/lo}{+0.0090}
\defval{posdiff/x40/1024/yocou-loop-minus-ced-enc/fineweb-edu-100bt/c1-128}{+0.0005}
\defval{posdiff/x40/1024/yocou-loop-minus-ced-enc/fineweb-edu-100bt/c1-128/hi}{+0.0015}
\defval{posdiff/x40/1024/yocou-loop-minus-ced-enc/fineweb-edu-100bt/c1-128/lo}{$-$0.0004}
\defval{posdiff/x40/1024/yocou-loop-minus-ced-enc/fineweb-edu-100bt/c1024-2048}{+0.0071}
\defval{posdiff/x40/1024/yocou-loop-minus-ced-enc/fineweb-edu-100bt/c1024-2048/hi}{+0.0081}
\defval{posdiff/x40/1024/yocou-loop-minus-ced-enc/fineweb-edu-100bt/c1024-2048/lo}{+0.0061}
\defval{posdiff/x40/1024/yocou-loop-minus-ced-enc/fineweb-edu-100bt/c128-256}{+0.0020}
\defval{posdiff/x40/1024/yocou-loop-minus-ced-enc/fineweb-edu-100bt/c128-256/hi}{+0.0029}
\defval{posdiff/x40/1024/yocou-loop-minus-ced-enc/fineweb-edu-100bt/c128-256/lo}{+0.0009}
\defval{posdiff/x40/1024/yocou-loop-minus-ced-enc/fineweb-edu-100bt/c256-512}{+0.0029}
\defval{posdiff/x40/1024/yocou-loop-minus-ced-enc/fineweb-edu-100bt/c256-512/hi}{+0.0039}
\defval{posdiff/x40/1024/yocou-loop-minus-ced-enc/fineweb-edu-100bt/c256-512/lo}{+0.0021}
\defval{posdiff/x40/1024/yocou-loop-minus-ced-enc/fineweb-edu-100bt/c512-1024}{+0.0051}
\defval{posdiff/x40/1024/yocou-loop-minus-ced-enc/fineweb-edu-100bt/c512-1024/hi}{+0.0060}
\defval{posdiff/x40/1024/yocou-loop-minus-ced-enc/fineweb-edu-100bt/c512-1024/lo}{+0.0043}
\defval{posdiff/x40/1024/yocou-loop-minus-ced-enc/fineweb-edu-100bt/mean}{+0.0038}
\defval{posdiff/x40/1024/yocou-loop-minus-ced-enc/fineweb-edu-100bt/mean/hi}{+0.0043}
\defval{posdiff/x40/1024/yocou-loop-minus-ced-enc/fineweb-edu-100bt/mean/lo}{+0.0034}
\defval{posdiff/x40/1024/yocou-loop-minus-ced-enc/govreport-test/c1-128}{$-$0.0004}
\defval{posdiff/x40/1024/yocou-loop-minus-ced-enc/govreport-test/c1-128/hi}{+0.0003}
\defval{posdiff/x40/1024/yocou-loop-minus-ced-enc/govreport-test/c1-128/lo}{$-$0.0012}
\defval{posdiff/x40/1024/yocou-loop-minus-ced-enc/govreport-test/c1024-2048}{+0.0091}
\defval{posdiff/x40/1024/yocou-loop-minus-ced-enc/govreport-test/c1024-2048/hi}{+0.0097}
\defval{posdiff/x40/1024/yocou-loop-minus-ced-enc/govreport-test/c1024-2048/lo}{+0.0086}
\defval{posdiff/x40/1024/yocou-loop-minus-ced-enc/govreport-test/c128-256}{+0.0013}
\defval{posdiff/x40/1024/yocou-loop-minus-ced-enc/govreport-test/c128-256/hi}{+0.0021}
\defval{posdiff/x40/1024/yocou-loop-minus-ced-enc/govreport-test/c128-256/lo}{+0.0005}
\defval{posdiff/x40/1024/yocou-loop-minus-ced-enc/govreport-test/c256-512}{+0.0044}
\defval{posdiff/x40/1024/yocou-loop-minus-ced-enc/govreport-test/c256-512/hi}{+0.0049}
\defval{posdiff/x40/1024/yocou-loop-minus-ced-enc/govreport-test/c256-512/lo}{+0.0038}
\defval{posdiff/x40/1024/yocou-loop-minus-ced-enc/govreport-test/c512-1024}{+0.0068}
\defval{posdiff/x40/1024/yocou-loop-minus-ced-enc/govreport-test/c512-1024/hi}{+0.0074}
\defval{posdiff/x40/1024/yocou-loop-minus-ced-enc/govreport-test/c512-1024/lo}{+0.0062}
\defval{posdiff/x40/1024/yocou-loop-minus-ced-enc/govreport-test/mean}{+0.0069}
\defval{posdiff/x40/1024/yocou-loop-minus-ced-enc/govreport-test/mean/hi}{+0.0073}
\defval{posdiff/x40/1024/yocou-loop-minus-ced-enc/govreport-test/mean/lo}{+0.0064}
\defval{posdiff/x40/1024/yocou-loop-minus-ced-enc/longdocs/c1-128}{$-$0.0050}
\defval{posdiff/x40/1024/yocou-loop-minus-ced-enc/longdocs/c1-128/hi}{$-$0.0043}
\defval{posdiff/x40/1024/yocou-loop-minus-ced-enc/longdocs/c1-128/lo}{$-$0.0056}
\defval{posdiff/x40/1024/yocou-loop-minus-ced-enc/longdocs/c1024-2048}{+0.0107}
\defval{posdiff/x40/1024/yocou-loop-minus-ced-enc/longdocs/c1024-2048/hi}{+0.0113}
\defval{posdiff/x40/1024/yocou-loop-minus-ced-enc/longdocs/c1024-2048/lo}{+0.0101}
\defval{posdiff/x40/1024/yocou-loop-minus-ced-enc/longdocs/c128-256}{$-$0.0003}
\defval{posdiff/x40/1024/yocou-loop-minus-ced-enc/longdocs/c128-256/hi}{+0.0003}
\defval{posdiff/x40/1024/yocou-loop-minus-ced-enc/longdocs/c128-256/lo}{$-$0.0009}
\defval{posdiff/x40/1024/yocou-loop-minus-ced-enc/longdocs/c256-512}{+0.0041}
\defval{posdiff/x40/1024/yocou-loop-minus-ced-enc/longdocs/c256-512/hi}{+0.0046}
\defval{posdiff/x40/1024/yocou-loop-minus-ced-enc/longdocs/c256-512/lo}{+0.0035}
\defval{posdiff/x40/1024/yocou-loop-minus-ced-enc/longdocs/c512-1024}{+0.0077}
\defval{posdiff/x40/1024/yocou-loop-minus-ced-enc/longdocs/c512-1024/hi}{+0.0083}
\defval{posdiff/x40/1024/yocou-loop-minus-ced-enc/longdocs/c512-1024/lo}{+0.0072}
\defval{posdiff/x40/1024/yocou-loop-minus-ced-enc/longdocs/mean}{+0.0074}
\defval{posdiff/x40/1024/yocou-loop-minus-ced-enc/longdocs/mean/hi}{+0.0079}
\defval{posdiff/x40/1024/yocou-loop-minus-ced-enc/longdocs/mean/lo}{+0.0070}
\defval{posdiff/x40/1024/yocou-loop-minus-ced-enc/pg19-test/c1-128}{$-$0.0071}
\defval{posdiff/x40/1024/yocou-loop-minus-ced-enc/pg19-test/c1-128/hi}{$-$0.0059}
\defval{posdiff/x40/1024/yocou-loop-minus-ced-enc/pg19-test/c1-128/lo}{$-$0.0082}
\defval{posdiff/x40/1024/yocou-loop-minus-ced-enc/pg19-test/c1024-2048}{+0.0086}
\defval{posdiff/x40/1024/yocou-loop-minus-ced-enc/pg19-test/c1024-2048/hi}{+0.0095}
\defval{posdiff/x40/1024/yocou-loop-minus-ced-enc/pg19-test/c1024-2048/lo}{+0.0077}
\defval{posdiff/x40/1024/yocou-loop-minus-ced-enc/pg19-test/c128-256}{$-$0.0017}
\defval{posdiff/x40/1024/yocou-loop-minus-ced-enc/pg19-test/c128-256/hi}{$-$0.0005}
\defval{posdiff/x40/1024/yocou-loop-minus-ced-enc/pg19-test/c128-256/lo}{$-$0.0028}
\defval{posdiff/x40/1024/yocou-loop-minus-ced-enc/pg19-test/c256-512}{+0.0021}
\defval{posdiff/x40/1024/yocou-loop-minus-ced-enc/pg19-test/c256-512/hi}{+0.0031}
\defval{posdiff/x40/1024/yocou-loop-minus-ced-enc/pg19-test/c256-512/lo}{+0.0011}
\defval{posdiff/x40/1024/yocou-loop-minus-ced-enc/pg19-test/c512-1024}{+0.0059}
\defval{posdiff/x40/1024/yocou-loop-minus-ced-enc/pg19-test/c512-1024/hi}{+0.0068}
\defval{posdiff/x40/1024/yocou-loop-minus-ced-enc/pg19-test/c512-1024/lo}{+0.0050}
\defval{posdiff/x40/1024/yocou-loop-minus-ced-enc/pg19-test/mean}{+0.0055}
\defval{posdiff/x40/1024/yocou-loop-minus-ced-enc/pg19-test/mean/hi}{+0.0063}
\defval{posdiff/x40/1024/yocou-loop-minus-ced-enc/pg19-test/mean/lo}{+0.0048}
\defval{posdiff/x40/1024/yocou-loop-minus-ced-enc/pg19-val/c1-128}{$-$0.0069}
\defval{posdiff/x40/1024/yocou-loop-minus-ced-enc/pg19-val/c1-128/hi}{$-$0.0046}
\defval{posdiff/x40/1024/yocou-loop-minus-ced-enc/pg19-val/c1-128/lo}{$-$0.0093}
\defval{posdiff/x40/1024/yocou-loop-minus-ced-enc/pg19-val/c1024-2048}{+0.0082}
\defval{posdiff/x40/1024/yocou-loop-minus-ced-enc/pg19-val/c1024-2048/hi}{+0.0094}
\defval{posdiff/x40/1024/yocou-loop-minus-ced-enc/pg19-val/c1024-2048/lo}{+0.0071}
\defval{posdiff/x40/1024/yocou-loop-minus-ced-enc/pg19-val/c128-256}{+0.0008}
\defval{posdiff/x40/1024/yocou-loop-minus-ced-enc/pg19-val/c128-256/hi}{+0.0030}
\defval{posdiff/x40/1024/yocou-loop-minus-ced-enc/pg19-val/c128-256/lo}{$-$0.0016}
\defval{posdiff/x40/1024/yocou-loop-minus-ced-enc/pg19-val/c256-512}{+0.0039}
\defval{posdiff/x40/1024/yocou-loop-minus-ced-enc/pg19-val/c256-512/hi}{+0.0057}
\defval{posdiff/x40/1024/yocou-loop-minus-ced-enc/pg19-val/c256-512/lo}{+0.0021}
\defval{posdiff/x40/1024/yocou-loop-minus-ced-enc/pg19-val/c512-1024}{+0.0061}
\defval{posdiff/x40/1024/yocou-loop-minus-ced-enc/pg19-val/c512-1024/hi}{+0.0075}
\defval{posdiff/x40/1024/yocou-loop-minus-ced-enc/pg19-val/c512-1024/lo}{+0.0048}
\defval{posdiff/x40/1024/yocou-loop-minus-ced-enc/pg19-val/mean}{+0.0058}
\defval{posdiff/x40/1024/yocou-loop-minus-ced-enc/pg19-val/mean/hi}{+0.0065}
\defval{posdiff/x40/1024/yocou-loop-minus-ced-enc/pg19-val/mean/lo}{+0.0050}
\defval{posdiff/x40/1024/yocou-loop-minus-dense-l25/finepdfs-long/c1-128}{$-$0.0074}
\defval{posdiff/x40/1024/yocou-loop-minus-dense-l25/finepdfs-long/c1-128/hi}{$-$0.0058}
\defval{posdiff/x40/1024/yocou-loop-minus-dense-l25/finepdfs-long/c1-128/lo}{$-$0.0089}
\defval{posdiff/x40/1024/yocou-loop-minus-dense-l25/finepdfs-long/c1024-2048}{+0.0112}
\defval{posdiff/x40/1024/yocou-loop-minus-dense-l25/finepdfs-long/c1024-2048/hi}{+0.0130}
\defval{posdiff/x40/1024/yocou-loop-minus-dense-l25/finepdfs-long/c1024-2048/lo}{+0.0094}
\defval{posdiff/x40/1024/yocou-loop-minus-dense-l25/finepdfs-long/c128-256}{$-$0.0014}
\defval{posdiff/x40/1024/yocou-loop-minus-dense-l25/finepdfs-long/c128-256/hi}{+0.0000}
\defval{posdiff/x40/1024/yocou-loop-minus-dense-l25/finepdfs-long/c128-256/lo}{$-$0.0029}
\defval{posdiff/x40/1024/yocou-loop-minus-dense-l25/finepdfs-long/c256-512}{+0.0033}
\defval{posdiff/x40/1024/yocou-loop-minus-dense-l25/finepdfs-long/c256-512/hi}{+0.0046}
\defval{posdiff/x40/1024/yocou-loop-minus-dense-l25/finepdfs-long/c256-512/lo}{+0.0020}
\defval{posdiff/x40/1024/yocou-loop-minus-dense-l25/finepdfs-long/c512-1024}{+0.0076}
\defval{posdiff/x40/1024/yocou-loop-minus-dense-l25/finepdfs-long/c512-1024/hi}{+0.0091}
\defval{posdiff/x40/1024/yocou-loop-minus-dense-l25/finepdfs-long/c512-1024/lo}{+0.0059}
\defval{posdiff/x40/1024/yocou-loop-minus-dense-l25/finepdfs-long/mean}{+0.0073}
\defval{posdiff/x40/1024/yocou-loop-minus-dense-l25/finepdfs-long/mean/hi}{+0.0088}
\defval{posdiff/x40/1024/yocou-loop-minus-dense-l25/finepdfs-long/mean/lo}{+0.0058}
\defval{posdiff/x40/1024/yocou-loop-minus-dense-l25/fineweb-edu-100bt/c1-128}{$-$0.0017}
\defval{posdiff/x40/1024/yocou-loop-minus-dense-l25/fineweb-edu-100bt/c1-128/hi}{$-$0.0007}
\defval{posdiff/x40/1024/yocou-loop-minus-dense-l25/fineweb-edu-100bt/c1-128/lo}{$-$0.0026}
\defval{posdiff/x40/1024/yocou-loop-minus-dense-l25/fineweb-edu-100bt/c1024-2048}{+0.0091}
\defval{posdiff/x40/1024/yocou-loop-minus-dense-l25/fineweb-edu-100bt/c1024-2048/hi}{+0.0103}
\defval{posdiff/x40/1024/yocou-loop-minus-dense-l25/fineweb-edu-100bt/c1024-2048/lo}{+0.0080}
\defval{posdiff/x40/1024/yocou-loop-minus-dense-l25/fineweb-edu-100bt/c128-256}{+0.0023}
\defval{posdiff/x40/1024/yocou-loop-minus-dense-l25/fineweb-edu-100bt/c128-256/hi}{+0.0033}
\defval{posdiff/x40/1024/yocou-loop-minus-dense-l25/fineweb-edu-100bt/c128-256/lo}{+0.0012}
\defval{posdiff/x40/1024/yocou-loop-minus-dense-l25/fineweb-edu-100bt/c256-512}{+0.0045}
\defval{posdiff/x40/1024/yocou-loop-minus-dense-l25/fineweb-edu-100bt/c256-512/hi}{+0.0054}
\defval{posdiff/x40/1024/yocou-loop-minus-dense-l25/fineweb-edu-100bt/c256-512/lo}{+0.0036}
\defval{posdiff/x40/1024/yocou-loop-minus-dense-l25/fineweb-edu-100bt/c512-1024}{+0.0064}
\defval{posdiff/x40/1024/yocou-loop-minus-dense-l25/fineweb-edu-100bt/c512-1024/hi}{+0.0073}
\defval{posdiff/x40/1024/yocou-loop-minus-dense-l25/fineweb-edu-100bt/c512-1024/lo}{+0.0054}
\defval{posdiff/x40/1024/yocou-loop-minus-dense-l25/fineweb-edu-100bt/mean}{+0.0046}
\defval{posdiff/x40/1024/yocou-loop-minus-dense-l25/fineweb-edu-100bt/mean/hi}{+0.0050}
\defval{posdiff/x40/1024/yocou-loop-minus-dense-l25/fineweb-edu-100bt/mean/lo}{+0.0041}
\defval{posdiff/x40/1024/yocou-loop-minus-dense-l25/govreport-test/c1-128}{$-$0.0006}
\defval{posdiff/x40/1024/yocou-loop-minus-dense-l25/govreport-test/c1-128/hi}{+0.0001}
\defval{posdiff/x40/1024/yocou-loop-minus-dense-l25/govreport-test/c1-128/lo}{$-$0.0014}
\defval{posdiff/x40/1024/yocou-loop-minus-dense-l25/govreport-test/c1024-2048}{+0.0130}
\defval{posdiff/x40/1024/yocou-loop-minus-dense-l25/govreport-test/c1024-2048/hi}{+0.0137}
\defval{posdiff/x40/1024/yocou-loop-minus-dense-l25/govreport-test/c1024-2048/lo}{+0.0123}
\defval{posdiff/x40/1024/yocou-loop-minus-dense-l25/govreport-test/c128-256}{+0.0022}
\defval{posdiff/x40/1024/yocou-loop-minus-dense-l25/govreport-test/c128-256/hi}{+0.0030}
\defval{posdiff/x40/1024/yocou-loop-minus-dense-l25/govreport-test/c128-256/lo}{+0.0014}
\defval{posdiff/x40/1024/yocou-loop-minus-dense-l25/govreport-test/c256-512}{+0.0064}
\defval{posdiff/x40/1024/yocou-loop-minus-dense-l25/govreport-test/c256-512/hi}{+0.0071}
\defval{posdiff/x40/1024/yocou-loop-minus-dense-l25/govreport-test/c256-512/lo}{+0.0058}
\defval{posdiff/x40/1024/yocou-loop-minus-dense-l25/govreport-test/c512-1024}{+0.0099}
\defval{posdiff/x40/1024/yocou-loop-minus-dense-l25/govreport-test/c512-1024/hi}{+0.0106}
\defval{posdiff/x40/1024/yocou-loop-minus-dense-l25/govreport-test/c512-1024/lo}{+0.0093}
\defval{posdiff/x40/1024/yocou-loop-minus-dense-l25/govreport-test/mean}{+0.0099}
\defval{posdiff/x40/1024/yocou-loop-minus-dense-l25/govreport-test/mean/hi}{+0.0104}
\defval{posdiff/x40/1024/yocou-loop-minus-dense-l25/govreport-test/mean/lo}{+0.0093}
\defval{posdiff/x40/1024/yocou-loop-minus-dense-l25/longdocs/c1-128}{$-$0.0042}
\defval{posdiff/x40/1024/yocou-loop-minus-dense-l25/longdocs/c1-128/hi}{$-$0.0034}
\defval{posdiff/x40/1024/yocou-loop-minus-dense-l25/longdocs/c1-128/lo}{$-$0.0049}
\defval{posdiff/x40/1024/yocou-loop-minus-dense-l25/longdocs/c1024-2048}{+0.0114}
\defval{posdiff/x40/1024/yocou-loop-minus-dense-l25/longdocs/c1024-2048/hi}{+0.0121}
\defval{posdiff/x40/1024/yocou-loop-minus-dense-l25/longdocs/c1024-2048/lo}{+0.0106}
\defval{posdiff/x40/1024/yocou-loop-minus-dense-l25/longdocs/c128-256}{+0.0002}
\defval{posdiff/x40/1024/yocou-loop-minus-dense-l25/longdocs/c128-256/hi}{+0.0010}
\defval{posdiff/x40/1024/yocou-loop-minus-dense-l25/longdocs/c128-256/lo}{$-$0.0006}
\defval{posdiff/x40/1024/yocou-loop-minus-dense-l25/longdocs/c256-512}{+0.0046}
\defval{posdiff/x40/1024/yocou-loop-minus-dense-l25/longdocs/c256-512/hi}{+0.0053}
\defval{posdiff/x40/1024/yocou-loop-minus-dense-l25/longdocs/c256-512/lo}{+0.0039}
\defval{posdiff/x40/1024/yocou-loop-minus-dense-l25/longdocs/c512-1024}{+0.0084}
\defval{posdiff/x40/1024/yocou-loop-minus-dense-l25/longdocs/c512-1024/hi}{+0.0091}
\defval{posdiff/x40/1024/yocou-loop-minus-dense-l25/longdocs/c512-1024/lo}{+0.0077}
\defval{posdiff/x40/1024/yocou-loop-minus-dense-l25/longdocs/mean}{+0.0081}
\defval{posdiff/x40/1024/yocou-loop-minus-dense-l25/longdocs/mean/hi}{+0.0087}
\defval{posdiff/x40/1024/yocou-loop-minus-dense-l25/longdocs/mean/lo}{+0.0075}
\defval{posdiff/x40/1024/yocou-loop-minus-dense-l25/pg19-test/c1-128}{$-$0.0046}
\defval{posdiff/x40/1024/yocou-loop-minus-dense-l25/pg19-test/c1-128/hi}{$-$0.0032}
\defval{posdiff/x40/1024/yocou-loop-minus-dense-l25/pg19-test/c1-128/lo}{$-$0.0060}
\defval{posdiff/x40/1024/yocou-loop-minus-dense-l25/pg19-test/c1024-2048}{+0.0100}
\defval{posdiff/x40/1024/yocou-loop-minus-dense-l25/pg19-test/c1024-2048/hi}{+0.0115}
\defval{posdiff/x40/1024/yocou-loop-minus-dense-l25/pg19-test/c1024-2048/lo}{+0.0085}
\defval{posdiff/x40/1024/yocou-loop-minus-dense-l25/pg19-test/c128-256}{$-$0.0002}
\defval{posdiff/x40/1024/yocou-loop-minus-dense-l25/pg19-test/c128-256/hi}{+0.0013}
\defval{posdiff/x40/1024/yocou-loop-minus-dense-l25/pg19-test/c128-256/lo}{$-$0.0017}
\defval{posdiff/x40/1024/yocou-loop-minus-dense-l25/pg19-test/c256-512}{+0.0041}
\defval{posdiff/x40/1024/yocou-loop-minus-dense-l25/pg19-test/c256-512/hi}{+0.0055}
\defval{posdiff/x40/1024/yocou-loop-minus-dense-l25/pg19-test/c256-512/lo}{+0.0027}
\defval{posdiff/x40/1024/yocou-loop-minus-dense-l25/pg19-test/c512-1024}{+0.0077}
\defval{posdiff/x40/1024/yocou-loop-minus-dense-l25/pg19-test/c512-1024/hi}{+0.0092}
\defval{posdiff/x40/1024/yocou-loop-minus-dense-l25/pg19-test/c512-1024/lo}{+0.0063}
\defval{posdiff/x40/1024/yocou-loop-minus-dense-l25/pg19-test/mean}{+0.0071}
\defval{posdiff/x40/1024/yocou-loop-minus-dense-l25/pg19-test/mean/hi}{+0.0085}
\defval{posdiff/x40/1024/yocou-loop-minus-dense-l25/pg19-test/mean/lo}{+0.0058}
\defval{posdiff/x40/1024/yocou-loop-minus-dense-l25/pg19-val/c1-128}{$-$0.0023}
\defval{posdiff/x40/1024/yocou-loop-minus-dense-l25/pg19-val/c1-128/hi}{+0.0001}
\defval{posdiff/x40/1024/yocou-loop-minus-dense-l25/pg19-val/c1-128/lo}{$-$0.0048}
\defval{posdiff/x40/1024/yocou-loop-minus-dense-l25/pg19-val/c1024-2048}{+0.0103}
\defval{posdiff/x40/1024/yocou-loop-minus-dense-l25/pg19-val/c1024-2048/hi}{+0.0115}
\defval{posdiff/x40/1024/yocou-loop-minus-dense-l25/pg19-val/c1024-2048/lo}{+0.0089}
\defval{posdiff/x40/1024/yocou-loop-minus-dense-l25/pg19-val/c128-256}{+0.0032}
\defval{posdiff/x40/1024/yocou-loop-minus-dense-l25/pg19-val/c128-256/hi}{+0.0056}
\defval{posdiff/x40/1024/yocou-loop-minus-dense-l25/pg19-val/c128-256/lo}{+0.0009}
\defval{posdiff/x40/1024/yocou-loop-minus-dense-l25/pg19-val/c256-512}{+0.0063}
\defval{posdiff/x40/1024/yocou-loop-minus-dense-l25/pg19-val/c256-512/hi}{+0.0080}
\defval{posdiff/x40/1024/yocou-loop-minus-dense-l25/pg19-val/c256-512/lo}{+0.0044}
\defval{posdiff/x40/1024/yocou-loop-minus-dense-l25/pg19-val/c512-1024}{+0.0091}
\defval{posdiff/x40/1024/yocou-loop-minus-dense-l25/pg19-val/c512-1024/hi}{+0.0106}
\defval{posdiff/x40/1024/yocou-loop-minus-dense-l25/pg19-val/c512-1024/lo}{+0.0075}
\defval{posdiff/x40/1024/yocou-loop-minus-dense-l25/pg19-val/mean}{+0.0083}
\defval{posdiff/x40/1024/yocou-loop-minus-dense-l25/pg19-val/mean/hi}{+0.0093}
\defval{posdiff/x40/1024/yocou-loop-minus-dense-l25/pg19-val/mean/lo}{+0.0072}
\defval{posdiff/x40/1024/yocou-noloop-minus-ced-noloop/finepdfs-long/c1-128}{$-$0.0148}
\defval{posdiff/x40/1024/yocou-noloop-minus-ced-noloop/finepdfs-long/c1-128/hi}{$-$0.0132}
\defval{posdiff/x40/1024/yocou-noloop-minus-ced-noloop/finepdfs-long/c1-128/lo}{$-$0.0164}
\defval{posdiff/x40/1024/yocou-noloop-minus-ced-noloop/finepdfs-long/c1024-2048}{+0.0002}
\defval{posdiff/x40/1024/yocou-noloop-minus-ced-noloop/finepdfs-long/c1024-2048/hi}{+0.0020}
\defval{posdiff/x40/1024/yocou-noloop-minus-ced-noloop/finepdfs-long/c1024-2048/lo}{$-$0.0014}
\defval{posdiff/x40/1024/yocou-noloop-minus-ced-noloop/finepdfs-long/c128-256}{$-$0.0096}
\defval{posdiff/x40/1024/yocou-noloop-minus-ced-noloop/finepdfs-long/c128-256/hi}{$-$0.0080}
\defval{posdiff/x40/1024/yocou-noloop-minus-ced-noloop/finepdfs-long/c128-256/lo}{$-$0.0111}
\defval{posdiff/x40/1024/yocou-noloop-minus-ced-noloop/finepdfs-long/c256-512}{$-$0.0049}
\defval{posdiff/x40/1024/yocou-noloop-minus-ced-noloop/finepdfs-long/c256-512/hi}{$-$0.0035}
\defval{posdiff/x40/1024/yocou-noloop-minus-ced-noloop/finepdfs-long/c256-512/lo}{$-$0.0062}
\defval{posdiff/x40/1024/yocou-noloop-minus-ced-noloop/finepdfs-long/c512-1024}{$-$0.0019}
\defval{posdiff/x40/1024/yocou-noloop-minus-ced-noloop/finepdfs-long/c512-1024/hi}{$-$0.0002}
\defval{posdiff/x40/1024/yocou-noloop-minus-ced-noloop/finepdfs-long/c512-1024/lo}{$-$0.0033}
\defval{posdiff/x40/1024/yocou-noloop-minus-ced-noloop/finepdfs-long/mean}{$-$0.0025}
\defval{posdiff/x40/1024/yocou-noloop-minus-ced-noloop/finepdfs-long/mean/hi}{$-$0.0009}
\defval{posdiff/x40/1024/yocou-noloop-minus-ced-noloop/finepdfs-long/mean/lo}{$-$0.0039}
\defval{posdiff/x40/1024/yocou-noloop-minus-ced-noloop/fineweb-edu-100bt/c1-128}{+0.0009}
\defval{posdiff/x40/1024/yocou-noloop-minus-ced-noloop/fineweb-edu-100bt/c1-128/hi}{+0.0019}
\defval{posdiff/x40/1024/yocou-noloop-minus-ced-noloop/fineweb-edu-100bt/c1-128/lo}{$-$0.0001}
\defval{posdiff/x40/1024/yocou-noloop-minus-ced-noloop/fineweb-edu-100bt/c1024-2048}{+0.0041}
\defval{posdiff/x40/1024/yocou-noloop-minus-ced-noloop/fineweb-edu-100bt/c1024-2048/hi}{+0.0050}
\defval{posdiff/x40/1024/yocou-noloop-minus-ced-noloop/fineweb-edu-100bt/c1024-2048/lo}{+0.0031}
\defval{posdiff/x40/1024/yocou-noloop-minus-ced-noloop/fineweb-edu-100bt/c128-256}{+0.0018}
\defval{posdiff/x40/1024/yocou-noloop-minus-ced-noloop/fineweb-edu-100bt/c128-256/hi}{+0.0029}
\defval{posdiff/x40/1024/yocou-noloop-minus-ced-noloop/fineweb-edu-100bt/c128-256/lo}{+0.0008}
\defval{posdiff/x40/1024/yocou-noloop-minus-ced-noloop/fineweb-edu-100bt/c256-512}{+0.0022}
\defval{posdiff/x40/1024/yocou-noloop-minus-ced-noloop/fineweb-edu-100bt/c256-512/hi}{+0.0031}
\defval{posdiff/x40/1024/yocou-noloop-minus-ced-noloop/fineweb-edu-100bt/c256-512/lo}{+0.0015}
\defval{posdiff/x40/1024/yocou-noloop-minus-ced-noloop/fineweb-edu-100bt/c512-1024}{+0.0037}
\defval{posdiff/x40/1024/yocou-noloop-minus-ced-noloop/fineweb-edu-100bt/c512-1024/hi}{+0.0045}
\defval{posdiff/x40/1024/yocou-noloop-minus-ced-noloop/fineweb-edu-100bt/c512-1024/lo}{+0.0028}
\defval{posdiff/x40/1024/yocou-noloop-minus-ced-noloop/fineweb-edu-100bt/mean}{+0.0027}
\defval{posdiff/x40/1024/yocou-noloop-minus-ced-noloop/fineweb-edu-100bt/mean/hi}{+0.0032}
\defval{posdiff/x40/1024/yocou-noloop-minus-ced-noloop/fineweb-edu-100bt/mean/lo}{+0.0023}
\defval{posdiff/x40/1024/yocou-noloop-minus-ced-noloop/govreport-test/c1-128}{+0.0007}
\defval{posdiff/x40/1024/yocou-noloop-minus-ced-noloop/govreport-test/c1-128/hi}{+0.0014}
\defval{posdiff/x40/1024/yocou-noloop-minus-ced-noloop/govreport-test/c1-128/lo}{$-$0.0001}
\defval{posdiff/x40/1024/yocou-noloop-minus-ced-noloop/govreport-test/c1024-2048}{+0.0053}
\defval{posdiff/x40/1024/yocou-noloop-minus-ced-noloop/govreport-test/c1024-2048/hi}{+0.0059}
\defval{posdiff/x40/1024/yocou-noloop-minus-ced-noloop/govreport-test/c1024-2048/lo}{+0.0048}
\defval{posdiff/x40/1024/yocou-noloop-minus-ced-noloop/govreport-test/c128-256}{+0.0030}
\defval{posdiff/x40/1024/yocou-noloop-minus-ced-noloop/govreport-test/c128-256/hi}{+0.0038}
\defval{posdiff/x40/1024/yocou-noloop-minus-ced-noloop/govreport-test/c128-256/lo}{+0.0023}
\defval{posdiff/x40/1024/yocou-noloop-minus-ced-noloop/govreport-test/c256-512}{+0.0037}
\defval{posdiff/x40/1024/yocou-noloop-minus-ced-noloop/govreport-test/c256-512/hi}{+0.0043}
\defval{posdiff/x40/1024/yocou-noloop-minus-ced-noloop/govreport-test/c256-512/lo}{+0.0032}
\defval{posdiff/x40/1024/yocou-noloop-minus-ced-noloop/govreport-test/c512-1024}{+0.0045}
\defval{posdiff/x40/1024/yocou-noloop-minus-ced-noloop/govreport-test/c512-1024/hi}{+0.0051}
\defval{posdiff/x40/1024/yocou-noloop-minus-ced-noloop/govreport-test/c512-1024/lo}{+0.0040}
\defval{posdiff/x40/1024/yocou-noloop-minus-ced-noloop/govreport-test/mean}{+0.0045}
\defval{posdiff/x40/1024/yocou-noloop-minus-ced-noloop/govreport-test/mean/hi}{+0.0049}
\defval{posdiff/x40/1024/yocou-noloop-minus-ced-noloop/govreport-test/mean/lo}{+0.0041}
\defval{posdiff/x40/1024/yocou-noloop-minus-ced-noloop/longdocs/c1-128}{$-$0.0074}
\defval{posdiff/x40/1024/yocou-noloop-minus-ced-noloop/longdocs/c1-128/hi}{$-$0.0067}
\defval{posdiff/x40/1024/yocou-noloop-minus-ced-noloop/longdocs/c1-128/lo}{$-$0.0081}
\defval{posdiff/x40/1024/yocou-noloop-minus-ced-noloop/longdocs/c1024-2048}{+0.0037}
\defval{posdiff/x40/1024/yocou-noloop-minus-ced-noloop/longdocs/c1024-2048/hi}{+0.0043}
\defval{posdiff/x40/1024/yocou-noloop-minus-ced-noloop/longdocs/c1024-2048/lo}{+0.0031}
\defval{posdiff/x40/1024/yocou-noloop-minus-ced-noloop/longdocs/c128-256}{$-$0.0035}
\defval{posdiff/x40/1024/yocou-noloop-minus-ced-noloop/longdocs/c128-256/hi}{$-$0.0028}
\defval{posdiff/x40/1024/yocou-noloop-minus-ced-noloop/longdocs/c128-256/lo}{$-$0.0042}
\defval{posdiff/x40/1024/yocou-noloop-minus-ced-noloop/longdocs/c256-512}{$-$0.0004}
\defval{posdiff/x40/1024/yocou-noloop-minus-ced-noloop/longdocs/c256-512/hi}{+0.0002}
\defval{posdiff/x40/1024/yocou-noloop-minus-ced-noloop/longdocs/c256-512/lo}{$-$0.0009}
\defval{posdiff/x40/1024/yocou-noloop-minus-ced-noloop/longdocs/c512-1024}{+0.0021}
\defval{posdiff/x40/1024/yocou-noloop-minus-ced-noloop/longdocs/c512-1024/hi}{+0.0027}
\defval{posdiff/x40/1024/yocou-noloop-minus-ced-noloop/longdocs/c512-1024/lo}{+0.0015}
\defval{posdiff/x40/1024/yocou-noloop-minus-ced-noloop/longdocs/mean}{+0.0017}
\defval{posdiff/x40/1024/yocou-noloop-minus-ced-noloop/longdocs/mean/hi}{+0.0022}
\defval{posdiff/x40/1024/yocou-noloop-minus-ced-noloop/longdocs/mean/lo}{+0.0011}
\defval{posdiff/x40/1024/yocou-noloop-minus-ced-noloop/pg19-test/c1-128}{$-$0.0082}
\defval{posdiff/x40/1024/yocou-noloop-minus-ced-noloop/pg19-test/c1-128/hi}{$-$0.0068}
\defval{posdiff/x40/1024/yocou-noloop-minus-ced-noloop/pg19-test/c1-128/lo}{$-$0.0095}
\defval{posdiff/x40/1024/yocou-noloop-minus-ced-noloop/pg19-test/c1024-2048}{+0.0055}
\defval{posdiff/x40/1024/yocou-noloop-minus-ced-noloop/pg19-test/c1024-2048/hi}{+0.0064}
\defval{posdiff/x40/1024/yocou-noloop-minus-ced-noloop/pg19-test/c1024-2048/lo}{+0.0046}
\defval{posdiff/x40/1024/yocou-noloop-minus-ced-noloop/pg19-test/c128-256}{$-$0.0040}
\defval{posdiff/x40/1024/yocou-noloop-minus-ced-noloop/pg19-test/c128-256/hi}{$-$0.0027}
\defval{posdiff/x40/1024/yocou-noloop-minus-ced-noloop/pg19-test/c128-256/lo}{$-$0.0052}
\defval{posdiff/x40/1024/yocou-noloop-minus-ced-noloop/pg19-test/c256-512}{$-$0.0000}
\defval{posdiff/x40/1024/yocou-noloop-minus-ced-noloop/pg19-test/c256-512/hi}{+0.0009}
\defval{posdiff/x40/1024/yocou-noloop-minus-ced-noloop/pg19-test/c256-512/lo}{$-$0.0011}
\defval{posdiff/x40/1024/yocou-noloop-minus-ced-noloop/pg19-test/c512-1024}{+0.0036}
\defval{posdiff/x40/1024/yocou-noloop-minus-ced-noloop/pg19-test/c512-1024/hi}{+0.0045}
\defval{posdiff/x40/1024/yocou-noloop-minus-ced-noloop/pg19-test/c512-1024/lo}{+0.0027}
\defval{posdiff/x40/1024/yocou-noloop-minus-ced-noloop/pg19-test/mean}{+0.0029}
\defval{posdiff/x40/1024/yocou-noloop-minus-ced-noloop/pg19-test/mean/hi}{+0.0036}
\defval{posdiff/x40/1024/yocou-noloop-minus-ced-noloop/pg19-test/mean/lo}{+0.0021}
\defval{posdiff/x40/1024/yocou-noloop-minus-ced-noloop/pg19-val/c1-128}{$-$0.0064}
\defval{posdiff/x40/1024/yocou-noloop-minus-ced-noloop/pg19-val/c1-128/hi}{$-$0.0040}
\defval{posdiff/x40/1024/yocou-noloop-minus-ced-noloop/pg19-val/c1-128/lo}{$-$0.0086}
\defval{posdiff/x40/1024/yocou-noloop-minus-ced-noloop/pg19-val/c1024-2048}{+0.0050}
\defval{posdiff/x40/1024/yocou-noloop-minus-ced-noloop/pg19-val/c1024-2048/hi}{+0.0060}
\defval{posdiff/x40/1024/yocou-noloop-minus-ced-noloop/pg19-val/c1024-2048/lo}{+0.0040}
\defval{posdiff/x40/1024/yocou-noloop-minus-ced-noloop/pg19-val/c128-256}{$-$0.0026}
\defval{posdiff/x40/1024/yocou-noloop-minus-ced-noloop/pg19-val/c128-256/hi}{$-$0.0003}
\defval{posdiff/x40/1024/yocou-noloop-minus-ced-noloop/pg19-val/c128-256/lo}{$-$0.0048}
\defval{posdiff/x40/1024/yocou-noloop-minus-ced-noloop/pg19-val/c256-512}{+0.0004}
\defval{posdiff/x40/1024/yocou-noloop-minus-ced-noloop/pg19-val/c256-512/hi}{+0.0022}
\defval{posdiff/x40/1024/yocou-noloop-minus-ced-noloop/pg19-val/c256-512/lo}{$-$0.0011}
\defval{posdiff/x40/1024/yocou-noloop-minus-ced-noloop/pg19-val/c512-1024}{+0.0037}
\defval{posdiff/x40/1024/yocou-noloop-minus-ced-noloop/pg19-val/c512-1024/hi}{+0.0049}
\defval{posdiff/x40/1024/yocou-noloop-minus-ced-noloop/pg19-val/c512-1024/lo}{+0.0024}
\defval{posdiff/x40/1024/yocou-noloop-minus-ced-noloop/pg19-val/mean}{+0.0029}
\defval{posdiff/x40/1024/yocou-noloop-minus-ced-noloop/pg19-val/mean/hi}{+0.0036}
\defval{posdiff/x40/1024/yocou-noloop-minus-ced-noloop/pg19-val/mean/lo}{+0.0022}
\defval{posdiff/x40/1024/yocou-noloop-minus-dense-noloop/finepdfs-long/c1-128}{$-$0.0185}
\defval{posdiff/x40/1024/yocou-noloop-minus-dense-noloop/finepdfs-long/c1-128/hi}{$-$0.0165}
\defval{posdiff/x40/1024/yocou-noloop-minus-dense-noloop/finepdfs-long/c1-128/lo}{$-$0.0207}
\defval{posdiff/x40/1024/yocou-noloop-minus-dense-noloop/finepdfs-long/c1024-2048}{$-$0.0083}
\defval{posdiff/x40/1024/yocou-noloop-minus-dense-noloop/finepdfs-long/c1024-2048/hi}{$-$0.0063}
\defval{posdiff/x40/1024/yocou-noloop-minus-dense-noloop/finepdfs-long/c1024-2048/lo}{$-$0.0105}
\defval{posdiff/x40/1024/yocou-noloop-minus-dense-noloop/finepdfs-long/c128-256}{$-$0.0157}
\defval{posdiff/x40/1024/yocou-noloop-minus-dense-noloop/finepdfs-long/c128-256/hi}{$-$0.0137}
\defval{posdiff/x40/1024/yocou-noloop-minus-dense-noloop/finepdfs-long/c128-256/lo}{$-$0.0178}
\defval{posdiff/x40/1024/yocou-noloop-minus-dense-noloop/finepdfs-long/c256-512}{$-$0.0135}
\defval{posdiff/x40/1024/yocou-noloop-minus-dense-noloop/finepdfs-long/c256-512/hi}{$-$0.0116}
\defval{posdiff/x40/1024/yocou-noloop-minus-dense-noloop/finepdfs-long/c256-512/lo}{$-$0.0153}
\defval{posdiff/x40/1024/yocou-noloop-minus-dense-noloop/finepdfs-long/c512-1024}{$-$0.0103}
\defval{posdiff/x40/1024/yocou-noloop-minus-dense-noloop/finepdfs-long/c512-1024/hi}{$-$0.0084}
\defval{posdiff/x40/1024/yocou-noloop-minus-dense-noloop/finepdfs-long/c512-1024/lo}{$-$0.0122}
\defval{posdiff/x40/1024/yocou-noloop-minus-dense-noloop/finepdfs-long/mean}{$-$0.0106}
\defval{posdiff/x40/1024/yocou-noloop-minus-dense-noloop/finepdfs-long/mean/hi}{$-$0.0088}
\defval{posdiff/x40/1024/yocou-noloop-minus-dense-noloop/finepdfs-long/mean/lo}{$-$0.0124}
\defval{posdiff/x40/1024/yocou-noloop-minus-dense-noloop/fineweb-edu-100bt/c1-128}{$-$0.0043}
\defval{posdiff/x40/1024/yocou-noloop-minus-dense-noloop/fineweb-edu-100bt/c1-128/hi}{$-$0.0034}
\defval{posdiff/x40/1024/yocou-noloop-minus-dense-noloop/fineweb-edu-100bt/c1-128/lo}{$-$0.0053}
\defval{posdiff/x40/1024/yocou-noloop-minus-dense-noloop/fineweb-edu-100bt/c1024-2048}{+0.0016}
\defval{posdiff/x40/1024/yocou-noloop-minus-dense-noloop/fineweb-edu-100bt/c1024-2048/hi}{+0.0027}
\defval{posdiff/x40/1024/yocou-noloop-minus-dense-noloop/fineweb-edu-100bt/c1024-2048/lo}{+0.0006}
\defval{posdiff/x40/1024/yocou-noloop-minus-dense-noloop/fineweb-edu-100bt/c128-256}{$-$0.0020}
\defval{posdiff/x40/1024/yocou-noloop-minus-dense-noloop/fineweb-edu-100bt/c128-256/hi}{$-$0.0009}
\defval{posdiff/x40/1024/yocou-noloop-minus-dense-noloop/fineweb-edu-100bt/c128-256/lo}{$-$0.0030}
\defval{posdiff/x40/1024/yocou-noloop-minus-dense-noloop/fineweb-edu-100bt/c256-512}{+0.0001}
\defval{posdiff/x40/1024/yocou-noloop-minus-dense-noloop/fineweb-edu-100bt/c256-512/hi}{+0.0010}
\defval{posdiff/x40/1024/yocou-noloop-minus-dense-noloop/fineweb-edu-100bt/c256-512/lo}{$-$0.0007}
\defval{posdiff/x40/1024/yocou-noloop-minus-dense-noloop/fineweb-edu-100bt/c512-1024}{+0.0014}
\defval{posdiff/x40/1024/yocou-noloop-minus-dense-noloop/fineweb-edu-100bt/c512-1024/hi}{+0.0023}
\defval{posdiff/x40/1024/yocou-noloop-minus-dense-noloop/fineweb-edu-100bt/c512-1024/lo}{+0.0005}
\defval{posdiff/x40/1024/yocou-noloop-minus-dense-noloop/fineweb-edu-100bt/mean}{$-$0.0003}
\defval{posdiff/x40/1024/yocou-noloop-minus-dense-noloop/fineweb-edu-100bt/mean/hi}{+0.0002}
\defval{posdiff/x40/1024/yocou-noloop-minus-dense-noloop/fineweb-edu-100bt/mean/lo}{$-$0.0008}
\defval{posdiff/x40/1024/yocou-noloop-minus-dense-noloop/govreport-test/c1-128}{$-$0.0035}
\defval{posdiff/x40/1024/yocou-noloop-minus-dense-noloop/govreport-test/c1-128/hi}{$-$0.0028}
\defval{posdiff/x40/1024/yocou-noloop-minus-dense-noloop/govreport-test/c1-128/lo}{$-$0.0044}
\defval{posdiff/x40/1024/yocou-noloop-minus-dense-noloop/govreport-test/c1024-2048}{+0.0053}
\defval{posdiff/x40/1024/yocou-noloop-minus-dense-noloop/govreport-test/c1024-2048/hi}{+0.0059}
\defval{posdiff/x40/1024/yocou-noloop-minus-dense-noloop/govreport-test/c1024-2048/lo}{+0.0047}
\defval{posdiff/x40/1024/yocou-noloop-minus-dense-noloop/govreport-test/c128-256}{$-$0.0003}
\defval{posdiff/x40/1024/yocou-noloop-minus-dense-noloop/govreport-test/c128-256/hi}{+0.0004}
\defval{posdiff/x40/1024/yocou-noloop-minus-dense-noloop/govreport-test/c128-256/lo}{$-$0.0010}
\defval{posdiff/x40/1024/yocou-noloop-minus-dense-noloop/govreport-test/c256-512}{+0.0020}
\defval{posdiff/x40/1024/yocou-noloop-minus-dense-noloop/govreport-test/c256-512/hi}{+0.0025}
\defval{posdiff/x40/1024/yocou-noloop-minus-dense-noloop/govreport-test/c256-512/lo}{+0.0014}
\defval{posdiff/x40/1024/yocou-noloop-minus-dense-noloop/govreport-test/c512-1024}{+0.0040}
\defval{posdiff/x40/1024/yocou-noloop-minus-dense-noloop/govreport-test/c512-1024/hi}{+0.0046}
\defval{posdiff/x40/1024/yocou-noloop-minus-dense-noloop/govreport-test/c512-1024/lo}{+0.0034}
\defval{posdiff/x40/1024/yocou-noloop-minus-dense-noloop/govreport-test/mean}{+0.0036}
\defval{posdiff/x40/1024/yocou-noloop-minus-dense-noloop/govreport-test/mean/hi}{+0.0041}
\defval{posdiff/x40/1024/yocou-noloop-minus-dense-noloop/govreport-test/mean/lo}{+0.0032}
\defval{posdiff/x40/1024/yocou-noloop-minus-dense-noloop/longdocs/c1-128}{$-$0.0113}
\defval{posdiff/x40/1024/yocou-noloop-minus-dense-noloop/longdocs/c1-128/hi}{$-$0.0105}
\defval{posdiff/x40/1024/yocou-noloop-minus-dense-noloop/longdocs/c1-128/lo}{$-$0.0122}
\defval{posdiff/x40/1024/yocou-noloop-minus-dense-noloop/longdocs/c1024-2048}{$-$0.0026}
\defval{posdiff/x40/1024/yocou-noloop-minus-dense-noloop/longdocs/c1024-2048/hi}{$-$0.0018}
\defval{posdiff/x40/1024/yocou-noloop-minus-dense-noloop/longdocs/c1024-2048/lo}{$-$0.0034}
\defval{posdiff/x40/1024/yocou-noloop-minus-dense-noloop/longdocs/c128-256}{$-$0.0094}
\defval{posdiff/x40/1024/yocou-noloop-minus-dense-noloop/longdocs/c128-256/hi}{$-$0.0086}
\defval{posdiff/x40/1024/yocou-noloop-minus-dense-noloop/longdocs/c128-256/lo}{$-$0.0103}
\defval{posdiff/x40/1024/yocou-noloop-minus-dense-noloop/longdocs/c256-512}{$-$0.0068}
\defval{posdiff/x40/1024/yocou-noloop-minus-dense-noloop/longdocs/c256-512/hi}{$-$0.0060}
\defval{posdiff/x40/1024/yocou-noloop-minus-dense-noloop/longdocs/c256-512/lo}{$-$0.0076}
\defval{posdiff/x40/1024/yocou-noloop-minus-dense-noloop/longdocs/c512-1024}{$-$0.0039}
\defval{posdiff/x40/1024/yocou-noloop-minus-dense-noloop/longdocs/c512-1024/hi}{$-$0.0031}
\defval{posdiff/x40/1024/yocou-noloop-minus-dense-noloop/longdocs/c512-1024/lo}{$-$0.0047}
\defval{posdiff/x40/1024/yocou-noloop-minus-dense-noloop/longdocs/mean}{$-$0.0044}
\defval{posdiff/x40/1024/yocou-noloop-minus-dense-noloop/longdocs/mean/hi}{$-$0.0037}
\defval{posdiff/x40/1024/yocou-noloop-minus-dense-noloop/longdocs/mean/lo}{$-$0.0052}
\defval{posdiff/x40/1024/yocou-noloop-minus-dense-noloop/pg19-test/c1-128}{$-$0.0120}
\defval{posdiff/x40/1024/yocou-noloop-minus-dense-noloop/pg19-test/c1-128/hi}{$-$0.0105}
\defval{posdiff/x40/1024/yocou-noloop-minus-dense-noloop/pg19-test/c1-128/lo}{$-$0.0136}
\defval{posdiff/x40/1024/yocou-noloop-minus-dense-noloop/pg19-test/c1024-2048}{$-$0.0048}
\defval{posdiff/x40/1024/yocou-noloop-minus-dense-noloop/pg19-test/c1024-2048/hi}{$-$0.0035}
\defval{posdiff/x40/1024/yocou-noloop-minus-dense-noloop/pg19-test/c1024-2048/lo}{$-$0.0063}
\defval{posdiff/x40/1024/yocou-noloop-minus-dense-noloop/pg19-test/c128-256}{$-$0.0124}
\defval{posdiff/x40/1024/yocou-noloop-minus-dense-noloop/pg19-test/c128-256/hi}{$-$0.0109}
\defval{posdiff/x40/1024/yocou-noloop-minus-dense-noloop/pg19-test/c128-256/lo}{$-$0.0140}
\defval{posdiff/x40/1024/yocou-noloop-minus-dense-noloop/pg19-test/c256-512}{$-$0.0089}
\defval{posdiff/x40/1024/yocou-noloop-minus-dense-noloop/pg19-test/c256-512/hi}{$-$0.0076}
\defval{posdiff/x40/1024/yocou-noloop-minus-dense-noloop/pg19-test/c256-512/lo}{$-$0.0104}
\defval{posdiff/x40/1024/yocou-noloop-minus-dense-noloop/pg19-test/c512-1024}{$-$0.0054}
\defval{posdiff/x40/1024/yocou-noloop-minus-dense-noloop/pg19-test/c512-1024/hi}{$-$0.0041}
\defval{posdiff/x40/1024/yocou-noloop-minus-dense-noloop/pg19-test/c512-1024/lo}{$-$0.0070}
\defval{posdiff/x40/1024/yocou-noloop-minus-dense-noloop/pg19-test/mean}{$-$0.0064}
\defval{posdiff/x40/1024/yocou-noloop-minus-dense-noloop/pg19-test/mean/hi}{$-$0.0052}
\defval{posdiff/x40/1024/yocou-noloop-minus-dense-noloop/pg19-test/mean/lo}{$-$0.0077}
\defval{posdiff/x40/1024/yocou-noloop-minus-dense-noloop/pg19-val/c1-128}{$-$0.0089}
\defval{posdiff/x40/1024/yocou-noloop-minus-dense-noloop/pg19-val/c1-128/hi}{$-$0.0065}
\defval{posdiff/x40/1024/yocou-noloop-minus-dense-noloop/pg19-val/c1-128/lo}{$-$0.0114}
\defval{posdiff/x40/1024/yocou-noloop-minus-dense-noloop/pg19-val/c1024-2048}{$-$0.0036}
\defval{posdiff/x40/1024/yocou-noloop-minus-dense-noloop/pg19-val/c1024-2048/hi}{$-$0.0022}
\defval{posdiff/x40/1024/yocou-noloop-minus-dense-noloop/pg19-val/c1024-2048/lo}{$-$0.0052}
\defval{posdiff/x40/1024/yocou-noloop-minus-dense-noloop/pg19-val/c128-256}{$-$0.0095}
\defval{posdiff/x40/1024/yocou-noloop-minus-dense-noloop/pg19-val/c128-256/hi}{$-$0.0067}
\defval{posdiff/x40/1024/yocou-noloop-minus-dense-noloop/pg19-val/c128-256/lo}{$-$0.0121}
\defval{posdiff/x40/1024/yocou-noloop-minus-dense-noloop/pg19-val/c256-512}{$-$0.0075}
\defval{posdiff/x40/1024/yocou-noloop-minus-dense-noloop/pg19-val/c256-512/hi}{$-$0.0052}
\defval{posdiff/x40/1024/yocou-noloop-minus-dense-noloop/pg19-val/c256-512/lo}{$-$0.0099}
\defval{posdiff/x40/1024/yocou-noloop-minus-dense-noloop/pg19-val/c512-1024}{$-$0.0045}
\defval{posdiff/x40/1024/yocou-noloop-minus-dense-noloop/pg19-val/c512-1024/hi}{$-$0.0031}
\defval{posdiff/x40/1024/yocou-noloop-minus-dense-noloop/pg19-val/c512-1024/lo}{$-$0.0061}
\defval{posdiff/x40/1024/yocou-noloop-minus-dense-noloop/pg19-val/mean}{$-$0.0050}
\defval{posdiff/x40/1024/yocou-noloop-minus-dense-noloop/pg19-val/mean/hi}{$-$0.0038}
\defval{posdiff/x40/1024/yocou-noloop-minus-dense-noloop/pg19-val/mean/lo}{$-$0.0064}
\defval{posdiff/x40/1024/yocou-noloop-minus-yocou-loop/finepdfs-long/c1-128}{+0.0219}
\defval{posdiff/x40/1024/yocou-noloop-minus-yocou-loop/finepdfs-long/c1-128/hi}{+0.0236}
\defval{posdiff/x40/1024/yocou-noloop-minus-yocou-loop/finepdfs-long/c1-128/lo}{+0.0202}
\defval{posdiff/x40/1024/yocou-noloop-minus-yocou-loop/finepdfs-long/c1024-2048}{+0.0264}
\defval{posdiff/x40/1024/yocou-noloop-minus-yocou-loop/finepdfs-long/c1024-2048/hi}{+0.0278}
\defval{posdiff/x40/1024/yocou-noloop-minus-yocou-loop/finepdfs-long/c1024-2048/lo}{+0.0249}
\defval{posdiff/x40/1024/yocou-noloop-minus-yocou-loop/finepdfs-long/c128-256}{+0.0262}
\defval{posdiff/x40/1024/yocou-noloop-minus-yocou-loop/finepdfs-long/c128-256/hi}{+0.0279}
\defval{posdiff/x40/1024/yocou-noloop-minus-yocou-loop/finepdfs-long/c128-256/lo}{+0.0246}
\defval{posdiff/x40/1024/yocou-noloop-minus-yocou-loop/finepdfs-long/c256-512}{+0.0264}
\defval{posdiff/x40/1024/yocou-noloop-minus-yocou-loop/finepdfs-long/c256-512/hi}{+0.0280}
\defval{posdiff/x40/1024/yocou-noloop-minus-yocou-loop/finepdfs-long/c256-512/lo}{+0.0249}
\defval{posdiff/x40/1024/yocou-noloop-minus-yocou-loop/finepdfs-long/c512-1024}{+0.0264}
\defval{posdiff/x40/1024/yocou-noloop-minus-yocou-loop/finepdfs-long/c512-1024/hi}{+0.0277}
\defval{posdiff/x40/1024/yocou-noloop-minus-yocou-loop/finepdfs-long/c512-1024/lo}{+0.0251}
\defval{posdiff/x40/1024/yocou-noloop-minus-yocou-loop/finepdfs-long/mean}{+0.0261}
\defval{posdiff/x40/1024/yocou-noloop-minus-yocou-loop/finepdfs-long/mean/hi}{+0.0273}
\defval{posdiff/x40/1024/yocou-noloop-minus-yocou-loop/finepdfs-long/mean/lo}{+0.0250}
\defval{posdiff/x40/1024/yocou-noloop-minus-yocou-loop/fineweb-edu-100bt/c1-128}{+0.0233}
\defval{posdiff/x40/1024/yocou-noloop-minus-yocou-loop/fineweb-edu-100bt/c1-128/hi}{+0.0245}
\defval{posdiff/x40/1024/yocou-noloop-minus-yocou-loop/fineweb-edu-100bt/c1-128/lo}{+0.0223}
\defval{posdiff/x40/1024/yocou-noloop-minus-yocou-loop/fineweb-edu-100bt/c1024-2048}{+0.0242}
\defval{posdiff/x40/1024/yocou-noloop-minus-yocou-loop/fineweb-edu-100bt/c1024-2048/hi}{+0.0255}
\defval{posdiff/x40/1024/yocou-noloop-minus-yocou-loop/fineweb-edu-100bt/c1024-2048/lo}{+0.0229}
\defval{posdiff/x40/1024/yocou-noloop-minus-yocou-loop/fineweb-edu-100bt/c128-256}{+0.0245}
\defval{posdiff/x40/1024/yocou-noloop-minus-yocou-loop/fineweb-edu-100bt/c128-256/hi}{+0.0258}
\defval{posdiff/x40/1024/yocou-noloop-minus-yocou-loop/fineweb-edu-100bt/c128-256/lo}{+0.0233}
\defval{posdiff/x40/1024/yocou-noloop-minus-yocou-loop/fineweb-edu-100bt/c256-512}{+0.0238}
\defval{posdiff/x40/1024/yocou-noloop-minus-yocou-loop/fineweb-edu-100bt/c256-512/hi}{+0.0248}
\defval{posdiff/x40/1024/yocou-noloop-minus-yocou-loop/fineweb-edu-100bt/c256-512/lo}{+0.0227}
\defval{posdiff/x40/1024/yocou-noloop-minus-yocou-loop/fineweb-edu-100bt/c512-1024}{+0.0242}
\defval{posdiff/x40/1024/yocou-noloop-minus-yocou-loop/fineweb-edu-100bt/c512-1024/hi}{+0.0253}
\defval{posdiff/x40/1024/yocou-noloop-minus-yocou-loop/fineweb-edu-100bt/c512-1024/lo}{+0.0231}
\defval{posdiff/x40/1024/yocou-noloop-minus-yocou-loop/fineweb-edu-100bt/mean}{+0.0240}
\defval{posdiff/x40/1024/yocou-noloop-minus-yocou-loop/fineweb-edu-100bt/mean/hi}{+0.0246}
\defval{posdiff/x40/1024/yocou-noloop-minus-yocou-loop/fineweb-edu-100bt/mean/lo}{+0.0234}
\defval{posdiff/x40/1024/yocou-noloop-minus-yocou-loop/govreport-test/c1-128}{+0.0187}
\defval{posdiff/x40/1024/yocou-noloop-minus-yocou-loop/govreport-test/c1-128/hi}{+0.0195}
\defval{posdiff/x40/1024/yocou-noloop-minus-yocou-loop/govreport-test/c1-128/lo}{+0.0177}
\defval{posdiff/x40/1024/yocou-noloop-minus-yocou-loop/govreport-test/c1024-2048}{+0.0204}
\defval{posdiff/x40/1024/yocou-noloop-minus-yocou-loop/govreport-test/c1024-2048/hi}{+0.0211}
\defval{posdiff/x40/1024/yocou-noloop-minus-yocou-loop/govreport-test/c1024-2048/lo}{+0.0198}
\defval{posdiff/x40/1024/yocou-noloop-minus-yocou-loop/govreport-test/c128-256}{+0.0219}
\defval{posdiff/x40/1024/yocou-noloop-minus-yocou-loop/govreport-test/c128-256/hi}{+0.0228}
\defval{posdiff/x40/1024/yocou-noloop-minus-yocou-loop/govreport-test/c128-256/lo}{+0.0210}
\defval{posdiff/x40/1024/yocou-noloop-minus-yocou-loop/govreport-test/c256-512}{+0.0215}
\defval{posdiff/x40/1024/yocou-noloop-minus-yocou-loop/govreport-test/c256-512/hi}{+0.0220}
\defval{posdiff/x40/1024/yocou-noloop-minus-yocou-loop/govreport-test/c256-512/lo}{+0.0208}
\defval{posdiff/x40/1024/yocou-noloop-minus-yocou-loop/govreport-test/c512-1024}{+0.0211}
\defval{posdiff/x40/1024/yocou-noloop-minus-yocou-loop/govreport-test/c512-1024/hi}{+0.0217}
\defval{posdiff/x40/1024/yocou-noloop-minus-yocou-loop/govreport-test/c512-1024/lo}{+0.0205}
\defval{posdiff/x40/1024/yocou-noloop-minus-yocou-loop/govreport-test/mean}{+0.0207}
\defval{posdiff/x40/1024/yocou-noloop-minus-yocou-loop/govreport-test/mean/hi}{+0.0212}
\defval{posdiff/x40/1024/yocou-noloop-minus-yocou-loop/govreport-test/mean/lo}{+0.0202}
\defval{posdiff/x40/1024/yocou-noloop-minus-yocou-loop/longdocs/c1-128}{+0.0252}
\defval{posdiff/x40/1024/yocou-noloop-minus-yocou-loop/longdocs/c1-128/hi}{+0.0260}
\defval{posdiff/x40/1024/yocou-noloop-minus-yocou-loop/longdocs/c1-128/lo}{+0.0245}
\defval{posdiff/x40/1024/yocou-noloop-minus-yocou-loop/longdocs/c1024-2048}{+0.0289}
\defval{posdiff/x40/1024/yocou-noloop-minus-yocou-loop/longdocs/c1024-2048/hi}{+0.0296}
\defval{posdiff/x40/1024/yocou-noloop-minus-yocou-loop/longdocs/c1024-2048/lo}{+0.0284}
\defval{posdiff/x40/1024/yocou-noloop-minus-yocou-loop/longdocs/c128-256}{+0.0291}
\defval{posdiff/x40/1024/yocou-noloop-minus-yocou-loop/longdocs/c128-256/hi}{+0.0299}
\defval{posdiff/x40/1024/yocou-noloop-minus-yocou-loop/longdocs/c128-256/lo}{+0.0284}
\defval{posdiff/x40/1024/yocou-noloop-minus-yocou-loop/longdocs/c256-512}{+0.0294}
\defval{posdiff/x40/1024/yocou-noloop-minus-yocou-loop/longdocs/c256-512/hi}{+0.0300}
\defval{posdiff/x40/1024/yocou-noloop-minus-yocou-loop/longdocs/c256-512/lo}{+0.0287}
\defval{posdiff/x40/1024/yocou-noloop-minus-yocou-loop/longdocs/c512-1024}{+0.0293}
\defval{posdiff/x40/1024/yocou-noloop-minus-yocou-loop/longdocs/c512-1024/hi}{+0.0299}
\defval{posdiff/x40/1024/yocou-noloop-minus-yocou-loop/longdocs/c512-1024/lo}{+0.0287}
\defval{posdiff/x40/1024/yocou-noloop-minus-yocou-loop/longdocs/mean}{+0.0289}
\defval{posdiff/x40/1024/yocou-noloop-minus-yocou-loop/longdocs/mean/hi}{+0.0294}
\defval{posdiff/x40/1024/yocou-noloop-minus-yocou-loop/longdocs/mean/lo}{+0.0284}
\defval{posdiff/x40/1024/yocou-noloop-minus-yocou-loop/pg19-test/c1-128}{+0.0350}
\defval{posdiff/x40/1024/yocou-noloop-minus-yocou-loop/pg19-test/c1-128/hi}{+0.0363}
\defval{posdiff/x40/1024/yocou-noloop-minus-yocou-loop/pg19-test/c1-128/lo}{+0.0338}
\defval{posdiff/x40/1024/yocou-noloop-minus-yocou-loop/pg19-test/c1024-2048}{+0.0400}
\defval{posdiff/x40/1024/yocou-noloop-minus-yocou-loop/pg19-test/c1024-2048/hi}{+0.0410}
\defval{posdiff/x40/1024/yocou-noloop-minus-yocou-loop/pg19-test/c1024-2048/lo}{+0.0389}
\defval{posdiff/x40/1024/yocou-noloop-minus-yocou-loop/pg19-test/c128-256}{+0.0393}
\defval{posdiff/x40/1024/yocou-noloop-minus-yocou-loop/pg19-test/c128-256/hi}{+0.0407}
\defval{posdiff/x40/1024/yocou-noloop-minus-yocou-loop/pg19-test/c128-256/lo}{+0.0380}
\defval{posdiff/x40/1024/yocou-noloop-minus-yocou-loop/pg19-test/c256-512}{+0.0402}
\defval{posdiff/x40/1024/yocou-noloop-minus-yocou-loop/pg19-test/c256-512/hi}{+0.0414}
\defval{posdiff/x40/1024/yocou-noloop-minus-yocou-loop/pg19-test/c256-512/lo}{+0.0391}
\defval{posdiff/x40/1024/yocou-noloop-minus-yocou-loop/pg19-test/c512-1024}{+0.0402}
\defval{posdiff/x40/1024/yocou-noloop-minus-yocou-loop/pg19-test/c512-1024/hi}{+0.0413}
\defval{posdiff/x40/1024/yocou-noloop-minus-yocou-loop/pg19-test/c512-1024/lo}{+0.0392}
\defval{posdiff/x40/1024/yocou-noloop-minus-yocou-loop/pg19-test/mean}{+0.0397}
\defval{posdiff/x40/1024/yocou-noloop-minus-yocou-loop/pg19-test/mean/hi}{+0.0406}
\defval{posdiff/x40/1024/yocou-noloop-minus-yocou-loop/pg19-test/mean/lo}{+0.0389}
\defval{posdiff/x40/1024/yocou-noloop-minus-yocou-loop/pg19-val/c1-128}{+0.0321}
\defval{posdiff/x40/1024/yocou-noloop-minus-yocou-loop/pg19-val/c1-128/hi}{+0.0347}
\defval{posdiff/x40/1024/yocou-noloop-minus-yocou-loop/pg19-val/c1-128/lo}{+0.0296}
\defval{posdiff/x40/1024/yocou-noloop-minus-yocou-loop/pg19-val/c1024-2048}{+0.0376}
\defval{posdiff/x40/1024/yocou-noloop-minus-yocou-loop/pg19-val/c1024-2048/hi}{+0.0387}
\defval{posdiff/x40/1024/yocou-noloop-minus-yocou-loop/pg19-val/c1024-2048/lo}{+0.0363}
\defval{posdiff/x40/1024/yocou-noloop-minus-yocou-loop/pg19-val/c128-256}{+0.0353}
\defval{posdiff/x40/1024/yocou-noloop-minus-yocou-loop/pg19-val/c128-256/hi}{+0.0377}
\defval{posdiff/x40/1024/yocou-noloop-minus-yocou-loop/pg19-val/c128-256/lo}{+0.0329}
\defval{posdiff/x40/1024/yocou-noloop-minus-yocou-loop/pg19-val/c256-512}{+0.0361}
\defval{posdiff/x40/1024/yocou-noloop-minus-yocou-loop/pg19-val/c256-512/hi}{+0.0382}
\defval{posdiff/x40/1024/yocou-noloop-minus-yocou-loop/pg19-val/c256-512/lo}{+0.0341}
\defval{posdiff/x40/1024/yocou-noloop-minus-yocou-loop/pg19-val/c512-1024}{+0.0379}
\defval{posdiff/x40/1024/yocou-noloop-minus-yocou-loop/pg19-val/c512-1024/hi}{+0.0393}
\defval{posdiff/x40/1024/yocou-noloop-minus-yocou-loop/pg19-val/c512-1024/lo}{+0.0365}
\defval{posdiff/x40/1024/yocou-noloop-minus-yocou-loop/pg19-val/mean}{+0.0370}
\defval{posdiff/x40/1024/yocou-noloop-minus-yocou-loop/pg19-val/mean/hi}{+0.0379}
\defval{posdiff/x40/1024/yocou-noloop-minus-yocou-loop/pg19-val/mean/lo}{+0.0361}
\defval{posfrac/x40/1024/ced-enc-minus-yocou-loop/longdocs/c1024-2048}{30}
\defval{posrel/x40/1024/ced-enc-minus-yocou-loop/longdocs/c1024-2048}{1.1}
\defval{recall/long/1024/ced-dec/bpb}{0.384}
\defval{recall/long/1024/ced-dec/control}{1.443}
\defval{recall/long/1024/ced-enc/bpb}{0.376}
\defval{recall/long/1024/ced-enc/control}{1.407}
\defval{recall/long/1024/ced-noloop/bpb}{0.395}
\defval{recall/long/1024/ced-noloop/control}{1.467}
\defval{recall/long/1024/dense-l47/bpb}{0.388}
\defval{recall/long/1024/dense-l47/control}{1.426}
\defval{recall/long/1024/dense-noloop/bpb}{0.401}
\defval{recall/long/1024/dense-noloop/control}{1.472}
\defval{recall/p140/768/p140-dec-fixed/bpb}{0.533}
\defval{recall/p140/768/p140-dec-fixed/control}{1.580}
\defval{recall/p140/768/p140-enc-fixed/bpb}{0.440}
\defval{recall/p140/768/p140-enc-fixed/control}{1.532}
\defval{recall/p140/768/p140-enc/bpb}{0.481}
\defval{recall/p140/768/p140-enc/control}{1.541}
\defval{recall/p140/768/p140-noloop/bpb}{0.637}
\defval{recall/p140/768/p140-noloop/control}{1.609}
\defval{recall/x40/1024/ced-dec/bpb}{0.477}
\defval{recall/x40/1024/ced-dec/control}{1.528}
\defval{recall/x40/1024/ced-enc-fixed/bpb}{0.388}
\defval{recall/x40/1024/ced-enc-fixed/control}{1.483}
\defval{recall/x40/1024/ced-enc/bpb}{0.388}
\defval{recall/x40/1024/ced-enc/control}{1.482}
\defval{recall/x40/1024/ced-noloop/bpb}{0.471}
\defval{recall/x40/1024/ced-noloop/control}{1.529}
\defval{recall/x40/1024/dense-l25/bpb}{0.413}
\defval{recall/x40/1024/dense-l25/control}{1.483}
\defval{recall/x40/1024/dense-l47/bpb}{0.452}
\defval{recall/x40/1024/dense-l47/control}{1.504}
\defval{recall/x40/1024/dense-noloop/bpb}{0.458}
\defval{recall/x40/1024/dense-noloop/control}{1.554}
\defval{recall/x40/1024/yocou-loop/bpb}{0.521}
\defval{recall/x40/1024/yocou-loop/control}{1.504}
\defval{recall/x40/1024/yocou-noloop/bpb}{0.522}
\defval{recall/x40/1024/yocou-noloop/control}{1.537}
\defval{recall/x40/1536/ced-dec/bpb}{0.367}
\defval{recall/x40/1536/ced-dec/control}{1.413}
\defval{recall/x40/1536/ced-enc/bpb}{0.363}
\defval{recall/x40/1536/ced-enc/control}{1.394}
\defval{recall/x40/1536/ced-noloop/bpb}{0.388}
\defval{recall/x40/1536/ced-noloop/control}{1.432}
\defval{recall/x40/1536/dense-l25/bpb}{0.357}
\defval{recall/x40/1536/dense-l25/control}{1.384}
\defval{recall/x40/1536/dense-noloop/bpb}{0.373}
\defval{recall/x40/1536/dense-noloop/control}{1.444}
\defval{recall/x40/768/ced-dec/bpb}{0.575}
\defval{recall/x40/768/ced-dec/control}{1.593}
\defval{recall/x40/768/ced-enc/bpb}{0.429}
\defval{recall/x40/768/ced-enc/control}{1.552}
\defval{recall/x40/768/ced-noloop/bpb}{0.589}
\defval{recall/x40/768/ced-noloop/control}{1.603}
\defval{recall/x40/768/dense-l25/bpb}{0.446}
\defval{recall/x40/768/dense-l25/control}{1.569}
\defval{recall/x40/768/dense-l47/bpb}{0.512}
\defval{recall/x40/768/dense-l47/control}{1.571}
\defval{recall/x40/768/dense-noloop/bpb}{0.483}
\defval{recall/x40/768/dense-noloop/control}{1.609}
\defval{recall/x40/768/yocou-loop/bpb}{0.551}
\defval{recall/x40/768/yocou-loop/control}{1.574}
\defval{recall/x40/768/yocou-noloop/bpb}{0.631}
\defval{recall/x40/768/yocou-noloop/control}{1.601}
\defval{recalldiff/long/1024/ced-dec-minus-ced-enc}{+0.008}
\defval{recalldiff/long/1024/ced-dec-minus-ced-enc/hi}{+0.009}
\defval{recalldiff/long/1024/ced-dec-minus-ced-enc/lo}{+0.007}
\defval{recalldiff/long/1024/ced-enc-minus-ced-noloop}{$-$0.020}
\defval{recalldiff/long/1024/ced-enc-minus-ced-noloop/hi}{$-$0.018}
\defval{recalldiff/long/1024/ced-enc-minus-ced-noloop/lo}{$-$0.021}
\defval{recalldiff/long/1024/ced-enc-minus-dense-l47}{$-$0.012}
\defval{recalldiff/long/1024/ced-enc-minus-dense-l47/hi}{$-$0.011}
\defval{recalldiff/long/1024/ced-enc-minus-dense-l47/lo}{$-$0.013}
\defval{recalldiff/long/1024/ced-noloop-minus-dense-noloop}{$-$0.005}
\defval{recalldiff/long/1024/ced-noloop-minus-dense-noloop/hi}{$-$0.004}
\defval{recalldiff/long/1024/ced-noloop-minus-dense-noloop/lo}{$-$0.007}
\defval{recalldiff/x40/1024/ced-dec-minus-ced-enc}{+0.089}
\defval{recalldiff/x40/1024/ced-dec-minus-ced-enc/hi}{+0.092}
\defval{recalldiff/x40/1024/ced-dec-minus-ced-enc/lo}{+0.086}
\defval{recalldiff/x40/1024/ced-dec-minus-ced-noloop}{+0.005}
\defval{recalldiff/x40/1024/ced-dec-minus-ced-noloop/hi}{+0.007}
\defval{recalldiff/x40/1024/ced-dec-minus-ced-noloop/lo}{+0.004}
\defval{recalldiff/x40/1024/ced-enc-minus-ced-noloop}{$-$0.084}
\defval{recalldiff/x40/1024/ced-enc-minus-ced-noloop/hi}{$-$0.081}
\defval{recalldiff/x40/1024/ced-enc-minus-ced-noloop/lo}{$-$0.087}
\defval{recalldiff/x40/1024/ced-enc-minus-dense-l25}{$-$0.026}
\defval{recalldiff/x40/1024/ced-enc-minus-dense-l25/hi}{$-$0.024}
\defval{recalldiff/x40/1024/ced-enc-minus-dense-l25/lo}{$-$0.027}
\defval{recalldiff/x40/1024/ced-noloop-minus-dense-noloop}{+0.014}
\defval{recalldiff/x40/1024/ced-noloop-minus-dense-noloop/hi}{+0.016}
\defval{recalldiff/x40/1024/ced-noloop-minus-dense-noloop/lo}{+0.012}
\defval{recalldiff/x40/1024/dense-l25-minus-dense-noloop}{$-$0.044}
\defval{recalldiff/x40/1024/dense-l25-minus-dense-noloop/hi}{$-$0.042}
\defval{recalldiff/x40/1024/dense-l25-minus-dense-noloop/lo}{$-$0.046}
\defval{recalldiff/x40/1024/yocou-loop-minus-dense-l25}{+0.108}
\defval{recalldiff/x40/1024/yocou-loop-minus-dense-l25/hi}{+0.110}
\defval{recalldiff/x40/1024/yocou-loop-minus-dense-l25/lo}{+0.105}
\defval{recalldiff/x40/1536/ced-dec-minus-ced-enc}{+0.004}
\defval{recalldiff/x40/1536/ced-dec-minus-ced-enc/hi}{+0.005}
\defval{recalldiff/x40/1536/ced-dec-minus-ced-enc/lo}{+0.003}
\defval{recalldiff/x40/1536/ced-dec-minus-ced-noloop}{$-$0.021}
\defval{recalldiff/x40/1536/ced-dec-minus-ced-noloop/hi}{$-$0.020}
\defval{recalldiff/x40/1536/ced-dec-minus-ced-noloop/lo}{$-$0.023}
\defval{recalldiff/x40/1536/ced-enc-minus-ced-noloop}{$-$0.025}
\defval{recalldiff/x40/1536/ced-enc-minus-ced-noloop/hi}{$-$0.024}
\defval{recalldiff/x40/1536/ced-enc-minus-ced-noloop/lo}{$-$0.027}
\defval{recalldiff/x40/1536/ced-enc-minus-dense-l25}{+0.006}
\defval{recalldiff/x40/1536/ced-enc-minus-dense-l25/hi}{+0.007}
\defval{recalldiff/x40/1536/ced-enc-minus-dense-l25/lo}{+0.005}
\defval{recalldiff/x40/1536/ced-noloop-minus-dense-noloop}{+0.016}
\defval{recalldiff/x40/1536/ced-noloop-minus-dense-noloop/hi}{+0.017}
\defval{recalldiff/x40/1536/ced-noloop-minus-dense-noloop/lo}{+0.015}
\defval{recalldiff/x40/1536/dense-l25-minus-dense-noloop}{$-$0.015}
\defval{recalldiff/x40/1536/dense-l25-minus-dense-noloop/hi}{$-$0.014}
\defval{recalldiff/x40/1536/dense-l25-minus-dense-noloop/lo}{$-$0.017}
\defval{recalldiff/x40/768/ced-dec-minus-ced-enc}{+0.146}
\defval{recalldiff/x40/768/ced-dec-minus-ced-enc/hi}{+0.150}
\defval{recalldiff/x40/768/ced-dec-minus-ced-enc/lo}{+0.142}
\defval{recalldiff/x40/768/ced-dec-minus-ced-noloop}{$-$0.014}
\defval{recalldiff/x40/768/ced-dec-minus-ced-noloop/hi}{$-$0.012}
\defval{recalldiff/x40/768/ced-dec-minus-ced-noloop/lo}{$-$0.015}
\defval{recalldiff/x40/768/ced-enc-minus-ced-noloop}{$-$0.160}
\defval{recalldiff/x40/768/ced-enc-minus-ced-noloop/hi}{$-$0.156}
\defval{recalldiff/x40/768/ced-enc-minus-ced-noloop/lo}{$-$0.164}
\defval{recalldiff/x40/768/ced-enc-minus-dense-l25}{$-$0.017}
\defval{recalldiff/x40/768/ced-enc-minus-dense-l25/hi}{$-$0.015}
\defval{recalldiff/x40/768/ced-enc-minus-dense-l25/lo}{$-$0.019}
\defval{recalldiff/x40/768/ced-noloop-minus-dense-noloop}{+0.106}
\defval{recalldiff/x40/768/ced-noloop-minus-dense-noloop/hi}{+0.109}
\defval{recalldiff/x40/768/ced-noloop-minus-dense-noloop/lo}{+0.103}
\defval{recalldiff/x40/768/dense-l25-minus-dense-noloop}{$-$0.036}
\defval{recalldiff/x40/768/dense-l25-minus-dense-noloop/hi}{$-$0.035}
\defval{recalldiff/x40/768/dense-l25-minus-dense-noloop/lo}{$-$0.038}
\defval{recalldiff/x40/768/yocou-loop-minus-dense-l25}{+0.105}
\defval{recalldiff/x40/768/yocou-loop-minus-dense-l25/hi}{+0.107}
\defval{recalldiff/x40/768/yocou-loop-minus-dense-l25/lo}{+0.103}
\defval{slope/long/1024/ced-dec-minus-ced-enc/finepdfs-long}{+0.0014}
\defval{slope/long/1024/ced-dec-minus-ced-enc/finepdfs-long/hi}{+0.0021}
\defval{slope/long/1024/ced-dec-minus-ced-enc/finepdfs-long/lo}{+0.0007}
\defval{slope/long/1024/ced-dec-minus-ced-enc/fineweb-edu-100bt}{+0.0011}
\defval{slope/long/1024/ced-dec-minus-ced-enc/fineweb-edu-100bt/hi}{+0.0016}
\defval{slope/long/1024/ced-dec-minus-ced-enc/fineweb-edu-100bt/lo}{+0.0006}
\defval{slope/long/1024/ced-dec-minus-ced-enc/govreport-test}{+0.0013}
\defval{slope/long/1024/ced-dec-minus-ced-enc/govreport-test/hi}{+0.0017}
\defval{slope/long/1024/ced-dec-minus-ced-enc/govreport-test/lo}{+0.0010}
\defval{slope/long/1024/ced-dec-minus-ced-enc/longdocs}{+0.0011}
\defval{slope/long/1024/ced-dec-minus-ced-enc/longdocs/hi}{+0.0014}
\defval{slope/long/1024/ced-dec-minus-ced-enc/longdocs/lo}{+0.0008}
\defval{slope/long/1024/ced-dec-minus-ced-enc/pg19-test}{+0.0007}
\defval{slope/long/1024/ced-dec-minus-ced-enc/pg19-test/hi}{+0.0012}
\defval{slope/long/1024/ced-dec-minus-ced-enc/pg19-test/lo}{+0.0002}
\defval{slope/long/1024/ced-dec-minus-ced-enc/pg19-val}{+0.0002}
\defval{slope/long/1024/ced-dec-minus-ced-enc/pg19-val/hi}{+0.0012}
\defval{slope/long/1024/ced-dec-minus-ced-enc/pg19-val/lo}{$-$0.0007}
\defval{slope/long/1024/ced-dec-minus-dense-l47/finepdfs-long}{+0.0022}
\defval{slope/long/1024/ced-dec-minus-dense-l47/finepdfs-long/hi}{+0.0029}
\defval{slope/long/1024/ced-dec-minus-dense-l47/finepdfs-long/lo}{+0.0016}
\defval{slope/long/1024/ced-dec-minus-dense-l47/fineweb-edu-100bt}{+0.0013}
\defval{slope/long/1024/ced-dec-minus-dense-l47/fineweb-edu-100bt/hi}{+0.0017}
\defval{slope/long/1024/ced-dec-minus-dense-l47/fineweb-edu-100bt/lo}{+0.0008}
\defval{slope/long/1024/ced-dec-minus-dense-l47/govreport-test}{+0.0021}
\defval{slope/long/1024/ced-dec-minus-dense-l47/govreport-test/hi}{+0.0024}
\defval{slope/long/1024/ced-dec-minus-dense-l47/govreport-test/lo}{+0.0018}
\defval{slope/long/1024/ced-dec-minus-dense-l47/longdocs}{+0.0023}
\defval{slope/long/1024/ced-dec-minus-dense-l47/longdocs/hi}{+0.0026}
\defval{slope/long/1024/ced-dec-minus-dense-l47/longdocs/lo}{+0.0021}
\defval{slope/long/1024/ced-dec-minus-dense-l47/pg19-test}{+0.0027}
\defval{slope/long/1024/ced-dec-minus-dense-l47/pg19-test/hi}{+0.0032}
\defval{slope/long/1024/ced-dec-minus-dense-l47/pg19-test/lo}{+0.0022}
\defval{slope/long/1024/ced-dec-minus-dense-l47/pg19-val}{+0.0031}
\defval{slope/long/1024/ced-dec-minus-dense-l47/pg19-val/hi}{+0.0040}
\defval{slope/long/1024/ced-dec-minus-dense-l47/pg19-val/lo}{+0.0021}
\defval{slope/long/1024/ced-enc-minus-dense-l47/finepdfs-long}{+0.0008}
\defval{slope/long/1024/ced-enc-minus-dense-l47/finepdfs-long/hi}{+0.0014}
\defval{slope/long/1024/ced-enc-minus-dense-l47/finepdfs-long/lo}{+0.0003}
\defval{slope/long/1024/ced-enc-minus-dense-l47/fineweb-edu-100bt}{+0.0001}
\defval{slope/long/1024/ced-enc-minus-dense-l47/fineweb-edu-100bt/hi}{+0.0006}
\defval{slope/long/1024/ced-enc-minus-dense-l47/fineweb-edu-100bt/lo}{$-$0.0003}
\defval{slope/long/1024/ced-enc-minus-dense-l47/govreport-test}{+0.0007}
\defval{slope/long/1024/ced-enc-minus-dense-l47/govreport-test/hi}{+0.0010}
\defval{slope/long/1024/ced-enc-minus-dense-l47/govreport-test/lo}{+0.0005}
\defval{slope/long/1024/ced-enc-minus-dense-l47/longdocs}{+0.0012}
\defval{slope/long/1024/ced-enc-minus-dense-l47/longdocs/hi}{+0.0014}
\defval{slope/long/1024/ced-enc-minus-dense-l47/longdocs/lo}{+0.0009}
\defval{slope/long/1024/ced-enc-minus-dense-l47/pg19-test}{+0.0020}
\defval{slope/long/1024/ced-enc-minus-dense-l47/pg19-test/hi}{+0.0025}
\defval{slope/long/1024/ced-enc-minus-dense-l47/pg19-test/lo}{+0.0015}
\defval{slope/long/1024/ced-enc-minus-dense-l47/pg19-val}{+0.0028}
\defval{slope/long/1024/ced-enc-minus-dense-l47/pg19-val/hi}{+0.0037}
\defval{slope/long/1024/ced-enc-minus-dense-l47/pg19-val/lo}{+0.0019}
\defval{slope/long/1024/ced-noloop-minus-ced-dec/finepdfs-long}{+0.0014}
\defval{slope/long/1024/ced-noloop-minus-ced-dec/finepdfs-long/hi}{+0.0019}
\defval{slope/long/1024/ced-noloop-minus-ced-dec/finepdfs-long/lo}{+0.0009}
\defval{slope/long/1024/ced-noloop-minus-ced-dec/fineweb-edu-100bt}{+0.0003}
\defval{slope/long/1024/ced-noloop-minus-ced-dec/fineweb-edu-100bt/hi}{+0.0007}
\defval{slope/long/1024/ced-noloop-minus-ced-dec/fineweb-edu-100bt/lo}{$-$0.0002}
\defval{slope/long/1024/ced-noloop-minus-ced-dec/govreport-test}{+0.0003}
\defval{slope/long/1024/ced-noloop-minus-ced-dec/govreport-test/hi}{+0.0006}
\defval{slope/long/1024/ced-noloop-minus-ced-dec/govreport-test/lo}{+0.0000}
\defval{slope/long/1024/ced-noloop-minus-ced-dec/longdocs}{+0.0011}
\defval{slope/long/1024/ced-noloop-minus-ced-dec/longdocs/hi}{+0.0013}
\defval{slope/long/1024/ced-noloop-minus-ced-dec/longdocs/lo}{+0.0008}
\defval{slope/long/1024/ced-noloop-minus-ced-dec/pg19-test}{+0.0015}
\defval{slope/long/1024/ced-noloop-minus-ced-dec/pg19-test/hi}{+0.0020}
\defval{slope/long/1024/ced-noloop-minus-ced-dec/pg19-test/lo}{+0.0010}
\defval{slope/long/1024/ced-noloop-minus-ced-dec/pg19-val}{+0.0014}
\defval{slope/long/1024/ced-noloop-minus-ced-dec/pg19-val/hi}{+0.0023}
\defval{slope/long/1024/ced-noloop-minus-ced-dec/pg19-val/lo}{+0.0007}
\defval{slope/long/1024/ced-noloop-minus-ced-enc/finepdfs-long}{+0.0028}
\defval{slope/long/1024/ced-noloop-minus-ced-enc/finepdfs-long/hi}{+0.0036}
\defval{slope/long/1024/ced-noloop-minus-ced-enc/finepdfs-long/lo}{+0.0021}
\defval{slope/long/1024/ced-noloop-minus-ced-enc/fineweb-edu-100bt}{+0.0014}
\defval{slope/long/1024/ced-noloop-minus-ced-enc/fineweb-edu-100bt/hi}{+0.0019}
\defval{slope/long/1024/ced-noloop-minus-ced-enc/fineweb-edu-100bt/lo}{+0.0008}
\defval{slope/long/1024/ced-noloop-minus-ced-enc/govreport-test}{+0.0016}
\defval{slope/long/1024/ced-noloop-minus-ced-enc/govreport-test/hi}{+0.0020}
\defval{slope/long/1024/ced-noloop-minus-ced-enc/govreport-test/lo}{+0.0013}
\defval{slope/long/1024/ced-noloop-minus-ced-enc/longdocs}{+0.0022}
\defval{slope/long/1024/ced-noloop-minus-ced-enc/longdocs/hi}{+0.0025}
\defval{slope/long/1024/ced-noloop-minus-ced-enc/longdocs/lo}{+0.0019}
\defval{slope/long/1024/ced-noloop-minus-ced-enc/pg19-test}{+0.0022}
\defval{slope/long/1024/ced-noloop-minus-ced-enc/pg19-test/hi}{+0.0028}
\defval{slope/long/1024/ced-noloop-minus-ced-enc/pg19-test/lo}{+0.0016}
\defval{slope/long/1024/ced-noloop-minus-ced-enc/pg19-val}{+0.0017}
\defval{slope/long/1024/ced-noloop-minus-ced-enc/pg19-val/hi}{+0.0026}
\defval{slope/long/1024/ced-noloop-minus-ced-enc/pg19-val/lo}{+0.0008}
\defval{slope/long/1024/ced-noloop-minus-dense-l47/finepdfs-long}{+0.0036}
\defval{slope/long/1024/ced-noloop-minus-dense-l47/finepdfs-long/hi}{+0.0043}
\defval{slope/long/1024/ced-noloop-minus-dense-l47/finepdfs-long/lo}{+0.0030}
\defval{slope/long/1024/ced-noloop-minus-dense-l47/fineweb-edu-100bt}{+0.0015}
\defval{slope/long/1024/ced-noloop-minus-dense-l47/fineweb-edu-100bt/hi}{+0.0021}
\defval{slope/long/1024/ced-noloop-minus-dense-l47/fineweb-edu-100bt/lo}{+0.0010}
\defval{slope/long/1024/ced-noloop-minus-dense-l47/govreport-test}{+0.0024}
\defval{slope/long/1024/ced-noloop-minus-dense-l47/govreport-test/hi}{+0.0027}
\defval{slope/long/1024/ced-noloop-minus-dense-l47/govreport-test/lo}{+0.0021}
\defval{slope/long/1024/ced-noloop-minus-dense-l47/longdocs}{+0.0034}
\defval{slope/long/1024/ced-noloop-minus-dense-l47/longdocs/hi}{+0.0037}
\defval{slope/long/1024/ced-noloop-minus-dense-l47/longdocs/lo}{+0.0031}
\defval{slope/long/1024/ced-noloop-minus-dense-l47/pg19-test}{+0.0042}
\defval{slope/long/1024/ced-noloop-minus-dense-l47/pg19-test/hi}{+0.0047}
\defval{slope/long/1024/ced-noloop-minus-dense-l47/pg19-test/lo}{+0.0036}
\defval{slope/long/1024/ced-noloop-minus-dense-l47/pg19-val}{+0.0045}
\defval{slope/long/1024/ced-noloop-minus-dense-l47/pg19-val/hi}{+0.0055}
\defval{slope/long/1024/ced-noloop-minus-dense-l47/pg19-val/lo}{+0.0036}
\defval{slope/p140/768/p140-dec-fixed-minus-p140-enc-fixed/finepdfs-long}{+0.0049}
\defval{slope/p140/768/p140-dec-fixed-minus-p140-enc-fixed/finepdfs-long/hi}{+0.0056}
\defval{slope/p140/768/p140-dec-fixed-minus-p140-enc-fixed/finepdfs-long/lo}{+0.0041}
\defval{slope/p140/768/p140-dec-fixed-minus-p140-enc-fixed/fineweb-edu-100bt}{+0.0011}
\defval{slope/p140/768/p140-dec-fixed-minus-p140-enc-fixed/fineweb-edu-100bt/hi}{+0.0016}
\defval{slope/p140/768/p140-dec-fixed-minus-p140-enc-fixed/fineweb-edu-100bt/lo}{+0.0006}
\defval{slope/p140/768/p140-dec-fixed-minus-p140-enc-fixed/govreport-test}{+0.0026}
\defval{slope/p140/768/p140-dec-fixed-minus-p140-enc-fixed/govreport-test/hi}{+0.0029}
\defval{slope/p140/768/p140-dec-fixed-minus-p140-enc-fixed/govreport-test/lo}{+0.0022}
\defval{slope/p140/768/p140-dec-fixed-minus-p140-enc-fixed/longdocs}{+0.0033}
\defval{slope/p140/768/p140-dec-fixed-minus-p140-enc-fixed/longdocs/hi}{+0.0036}
\defval{slope/p140/768/p140-dec-fixed-minus-p140-enc-fixed/longdocs/lo}{+0.0030}
\defval{slope/p140/768/p140-dec-fixed-minus-p140-enc-fixed/pg19-test}{+0.0025}
\defval{slope/p140/768/p140-dec-fixed-minus-p140-enc-fixed/pg19-test/hi}{+0.0030}
\defval{slope/p140/768/p140-dec-fixed-minus-p140-enc-fixed/pg19-test/lo}{+0.0020}
\defval{slope/p140/768/p140-dec-fixed-minus-p140-enc-fixed/pg19-val}{+0.0031}
\defval{slope/p140/768/p140-dec-fixed-minus-p140-enc-fixed/pg19-val/hi}{+0.0040}
\defval{slope/p140/768/p140-dec-fixed-minus-p140-enc-fixed/pg19-val/lo}{+0.0023}
\defval{slope/p140/768/p140-enc-fixed-minus-p140-enc/finepdfs-long}{$-$0.0002}
\defval{slope/p140/768/p140-enc-fixed-minus-p140-enc/finepdfs-long/hi}{+0.0003}
\defval{slope/p140/768/p140-enc-fixed-minus-p140-enc/finepdfs-long/lo}{$-$0.0007}
\defval{slope/p140/768/p140-enc-fixed-minus-p140-enc/fineweb-edu-100bt}{$-$0.0000}
\defval{slope/p140/768/p140-enc-fixed-minus-p140-enc/fineweb-edu-100bt/hi}{+0.0004}
\defval{slope/p140/768/p140-enc-fixed-minus-p140-enc/fineweb-edu-100bt/lo}{$-$0.0004}
\defval{slope/p140/768/p140-enc-fixed-minus-p140-enc/govreport-test}{+0.0001}
\defval{slope/p140/768/p140-enc-fixed-minus-p140-enc/govreport-test/hi}{+0.0003}
\defval{slope/p140/768/p140-enc-fixed-minus-p140-enc/govreport-test/lo}{$-$0.0002}
\defval{slope/p140/768/p140-enc-fixed-minus-p140-enc/longdocs}{$-$0.0003}
\defval{slope/p140/768/p140-enc-fixed-minus-p140-enc/longdocs/hi}{$-$0.0001}
\defval{slope/p140/768/p140-enc-fixed-minus-p140-enc/longdocs/lo}{$-$0.0005}
\defval{slope/p140/768/p140-enc-fixed-minus-p140-enc/pg19-test}{$-$0.0007}
\defval{slope/p140/768/p140-enc-fixed-minus-p140-enc/pg19-test/hi}{$-$0.0003}
\defval{slope/p140/768/p140-enc-fixed-minus-p140-enc/pg19-test/lo}{$-$0.0012}
\defval{slope/p140/768/p140-enc-fixed-minus-p140-enc/pg19-val}{$-$0.0001}
\defval{slope/p140/768/p140-enc-fixed-minus-p140-enc/pg19-val/hi}{+0.0007}
\defval{slope/p140/768/p140-enc-fixed-minus-p140-enc/pg19-val/lo}{$-$0.0008}
\defval{slope/p140/768/p140-noloop-minus-p140-enc/finepdfs-long}{+0.0051}
\defval{slope/p140/768/p140-noloop-minus-p140-enc/finepdfs-long/hi}{+0.0060}
\defval{slope/p140/768/p140-noloop-minus-p140-enc/finepdfs-long/lo}{+0.0042}
\defval{slope/p140/768/p140-noloop-minus-p140-enc/fineweb-edu-100bt}{+0.0016}
\defval{slope/p140/768/p140-noloop-minus-p140-enc/fineweb-edu-100bt/hi}{+0.0021}
\defval{slope/p140/768/p140-noloop-minus-p140-enc/fineweb-edu-100bt/lo}{+0.0010}
\defval{slope/p140/768/p140-noloop-minus-p140-enc/govreport-test}{+0.0026}
\defval{slope/p140/768/p140-noloop-minus-p140-enc/govreport-test/hi}{+0.0030}
\defval{slope/p140/768/p140-noloop-minus-p140-enc/govreport-test/lo}{+0.0022}
\defval{slope/p140/768/p140-noloop-minus-p140-enc/longdocs}{+0.0033}
\defval{slope/p140/768/p140-noloop-minus-p140-enc/longdocs/hi}{+0.0037}
\defval{slope/p140/768/p140-noloop-minus-p140-enc/longdocs/lo}{+0.0029}
\defval{slope/p140/768/p140-noloop-minus-p140-enc/pg19-test}{+0.0023}
\defval{slope/p140/768/p140-noloop-minus-p140-enc/pg19-test/hi}{+0.0029}
\defval{slope/p140/768/p140-noloop-minus-p140-enc/pg19-test/lo}{+0.0017}
\defval{slope/p140/768/p140-noloop-minus-p140-enc/pg19-val}{+0.0031}
\defval{slope/p140/768/p140-noloop-minus-p140-enc/pg19-val/hi}{+0.0040}
\defval{slope/p140/768/p140-noloop-minus-p140-enc/pg19-val/lo}{+0.0023}
\defval{slope/x40/1024/ced-dec-minus-ced-enc/finepdfs-long}{+0.0032}
\defval{slope/x40/1024/ced-dec-minus-ced-enc/finepdfs-long/hi}{+0.0039}
\defval{slope/x40/1024/ced-dec-minus-ced-enc/finepdfs-long/lo}{+0.0026}
\defval{slope/x40/1024/ced-dec-minus-ced-enc/fineweb-edu-100bt}{+0.0011}
\defval{slope/x40/1024/ced-dec-minus-ced-enc/fineweb-edu-100bt/hi}{+0.0016}
\defval{slope/x40/1024/ced-dec-minus-ced-enc/fineweb-edu-100bt/lo}{+0.0006}
\defval{slope/x40/1024/ced-dec-minus-ced-enc/govreport-test}{+0.0013}
\defval{slope/x40/1024/ced-dec-minus-ced-enc/govreport-test/hi}{+0.0017}
\defval{slope/x40/1024/ced-dec-minus-ced-enc/govreport-test/lo}{+0.0010}
\defval{slope/x40/1024/ced-dec-minus-ced-enc/longdocs}{+0.0018}
\defval{slope/x40/1024/ced-dec-minus-ced-enc/longdocs/hi}{+0.0021}
\defval{slope/x40/1024/ced-dec-minus-ced-enc/longdocs/lo}{+0.0015}
\defval{slope/x40/1024/ced-dec-minus-ced-enc/pg19-test}{+0.0008}
\defval{slope/x40/1024/ced-dec-minus-ced-enc/pg19-test/hi}{+0.0013}
\defval{slope/x40/1024/ced-dec-minus-ced-enc/pg19-test/lo}{+0.0004}
\defval{slope/x40/1024/ced-dec-minus-ced-enc/pg19-val}{+0.0001}
\defval{slope/x40/1024/ced-dec-minus-ced-enc/pg19-val/hi}{+0.0010}
\defval{slope/x40/1024/ced-dec-minus-ced-enc/pg19-val/lo}{$-$0.0008}
\defval{slope/x40/1024/ced-dec-minus-dense-l25/finepdfs-long}{+0.0024}
\defval{slope/x40/1024/ced-dec-minus-dense-l25/finepdfs-long/hi}{+0.0031}
\defval{slope/x40/1024/ced-dec-minus-dense-l25/finepdfs-long/lo}{+0.0018}
\defval{slope/x40/1024/ced-dec-minus-dense-l25/fineweb-edu-100bt}{+0.0016}
\defval{slope/x40/1024/ced-dec-minus-dense-l25/fineweb-edu-100bt/hi}{+0.0021}
\defval{slope/x40/1024/ced-dec-minus-dense-l25/fineweb-edu-100bt/lo}{+0.0011}
\defval{slope/x40/1024/ced-dec-minus-dense-l25/govreport-test}{+0.0023}
\defval{slope/x40/1024/ced-dec-minus-dense-l25/govreport-test/hi}{+0.0027}
\defval{slope/x40/1024/ced-dec-minus-dense-l25/govreport-test/lo}{+0.0020}
\defval{slope/x40/1024/ced-dec-minus-dense-l25/longdocs}{+0.0019}
\defval{slope/x40/1024/ced-dec-minus-dense-l25/longdocs/hi}{+0.0021}
\defval{slope/x40/1024/ced-dec-minus-dense-l25/longdocs/lo}{+0.0015}
\defval{slope/x40/1024/ced-dec-minus-dense-l25/pg19-test}{+0.0008}
\defval{slope/x40/1024/ced-dec-minus-dense-l25/pg19-test/hi}{+0.0013}
\defval{slope/x40/1024/ced-dec-minus-dense-l25/pg19-test/lo}{+0.0003}
\defval{slope/x40/1024/ced-dec-minus-dense-l25/pg19-val}{+0.0000}
\defval{slope/x40/1024/ced-dec-minus-dense-l25/pg19-val/hi}{+0.0010}
\defval{slope/x40/1024/ced-dec-minus-dense-l25/pg19-val/lo}{$-$0.0010}
\defval{slope/x40/1024/ced-enc-fixed-minus-ced-enc/finepdfs-long}{$-$0.0004}
\defval{slope/x40/1024/ced-enc-fixed-minus-ced-enc/finepdfs-long/hi}{+0.0001}
\defval{slope/x40/1024/ced-enc-fixed-minus-ced-enc/finepdfs-long/lo}{$-$0.0008}
\defval{slope/x40/1024/ced-enc-fixed-minus-ced-enc/fineweb-edu-100bt}{+0.0001}
\defval{slope/x40/1024/ced-enc-fixed-minus-ced-enc/fineweb-edu-100bt/hi}{+0.0005}
\defval{slope/x40/1024/ced-enc-fixed-minus-ced-enc/fineweb-edu-100bt/lo}{$-$0.0003}
\defval{slope/x40/1024/ced-enc-fixed-minus-ced-enc/govreport-test}{+0.0004}
\defval{slope/x40/1024/ced-enc-fixed-minus-ced-enc/govreport-test/hi}{+0.0006}
\defval{slope/x40/1024/ced-enc-fixed-minus-ced-enc/govreport-test/lo}{+0.0001}
\defval{slope/x40/1024/ced-enc-fixed-minus-ced-enc/longdocs}{+0.0001}
\defval{slope/x40/1024/ced-enc-fixed-minus-ced-enc/longdocs/hi}{+0.0003}
\defval{slope/x40/1024/ced-enc-fixed-minus-ced-enc/longdocs/lo}{$-$0.0001}
\defval{slope/x40/1024/ced-enc-fixed-minus-ced-enc/pg19-test}{+0.0004}
\defval{slope/x40/1024/ced-enc-fixed-minus-ced-enc/pg19-test/hi}{+0.0008}
\defval{slope/x40/1024/ced-enc-fixed-minus-ced-enc/pg19-test/lo}{$-$0.0000}
\defval{slope/x40/1024/ced-enc-fixed-minus-ced-enc/pg19-val}{+0.0003}
\defval{slope/x40/1024/ced-enc-fixed-minus-ced-enc/pg19-val/hi}{+0.0010}
\defval{slope/x40/1024/ced-enc-fixed-minus-ced-enc/pg19-val/lo}{$-$0.0005}
\defval{slope/x40/1024/ced-enc-minus-dense-l25/finepdfs-long}{$-$0.0008}
\defval{slope/x40/1024/ced-enc-minus-dense-l25/finepdfs-long/hi}{$-$0.0002}
\defval{slope/x40/1024/ced-enc-minus-dense-l25/finepdfs-long/lo}{$-$0.0013}
\defval{slope/x40/1024/ced-enc-minus-dense-l25/fineweb-edu-100bt}{+0.0005}
\defval{slope/x40/1024/ced-enc-minus-dense-l25/fineweb-edu-100bt/hi}{+0.0009}
\defval{slope/x40/1024/ced-enc-minus-dense-l25/fineweb-edu-100bt/lo}{+0.0001}
\defval{slope/x40/1024/ced-enc-minus-dense-l25/govreport-test}{+0.0010}
\defval{slope/x40/1024/ced-enc-minus-dense-l25/govreport-test/hi}{+0.0013}
\defval{slope/x40/1024/ced-enc-minus-dense-l25/govreport-test/lo}{+0.0007}
\defval{slope/x40/1024/ced-enc-minus-dense-l25/longdocs}{+0.0001}
\defval{slope/x40/1024/ced-enc-minus-dense-l25/longdocs/hi}{+0.0003}
\defval{slope/x40/1024/ced-enc-minus-dense-l25/longdocs/lo}{$-$0.0002}
\defval{slope/x40/1024/ced-enc-minus-dense-l25/pg19-test}{$-$0.0001}
\defval{slope/x40/1024/ced-enc-minus-dense-l25/pg19-test/hi}{+0.0004}
\defval{slope/x40/1024/ced-enc-minus-dense-l25/pg19-test/lo}{$-$0.0005}
\defval{slope/x40/1024/ced-enc-minus-dense-l25/pg19-val}{$-$0.0001}
\defval{slope/x40/1024/ced-enc-minus-dense-l25/pg19-val/hi}{+0.0008}
\defval{slope/x40/1024/ced-enc-minus-dense-l25/pg19-val/lo}{$-$0.0010}
\defval{slope/x40/1024/ced-enc-minus-yocou-loop/finepdfs-long}{$-$0.0049}
\defval{slope/x40/1024/ced-enc-minus-yocou-loop/finepdfs-long/hi}{$-$0.0044}
\defval{slope/x40/1024/ced-enc-minus-yocou-loop/finepdfs-long/lo}{$-$0.0056}
\defval{slope/x40/1024/ced-enc-minus-yocou-loop/fineweb-edu-100bt}{$-$0.0017}
\defval{slope/x40/1024/ced-enc-minus-yocou-loop/fineweb-edu-100bt/hi}{$-$0.0013}
\defval{slope/x40/1024/ced-enc-minus-yocou-loop/fineweb-edu-100bt/lo}{$-$0.0022}
\defval{slope/x40/1024/ced-enc-minus-yocou-loop/govreport-test}{$-$0.0026}
\defval{slope/x40/1024/ced-enc-minus-yocou-loop/govreport-test/hi}{$-$0.0023}
\defval{slope/x40/1024/ced-enc-minus-yocou-loop/govreport-test/lo}{$-$0.0029}
\defval{slope/x40/1024/ced-enc-minus-yocou-loop/longdocs}{$-$0.0037}
\defval{slope/x40/1024/ced-enc-minus-yocou-loop/longdocs/hi}{$-$0.0034}
\defval{slope/x40/1024/ced-enc-minus-yocou-loop/longdocs/lo}{$-$0.0039}
\defval{slope/x40/1024/ced-enc-minus-yocou-loop/pg19-test}{$-$0.0035}
\defval{slope/x40/1024/ced-enc-minus-yocou-loop/pg19-test/hi}{$-$0.0030}
\defval{slope/x40/1024/ced-enc-minus-yocou-loop/pg19-test/lo}{$-$0.0039}
\defval{slope/x40/1024/ced-enc-minus-yocou-loop/pg19-val}{$-$0.0025}
\defval{slope/x40/1024/ced-enc-minus-yocou-loop/pg19-val/hi}{$-$0.0016}
\defval{slope/x40/1024/ced-enc-minus-yocou-loop/pg19-val/lo}{$-$0.0033}
\defval{slope/x40/1024/ced-noloop-minus-ced-dec/finepdfs-long}{$-$0.0014}
\defval{slope/x40/1024/ced-noloop-minus-ced-dec/finepdfs-long/hi}{$-$0.0009}
\defval{slope/x40/1024/ced-noloop-minus-ced-dec/finepdfs-long/lo}{$-$0.0020}
\defval{slope/x40/1024/ced-noloop-minus-ced-dec/fineweb-edu-100bt}{$-$0.0002}
\defval{slope/x40/1024/ced-noloop-minus-ced-dec/fineweb-edu-100bt/hi}{+0.0003}
\defval{slope/x40/1024/ced-noloop-minus-ced-dec/fineweb-edu-100bt/lo}{$-$0.0007}
\defval{slope/x40/1024/ced-noloop-minus-ced-dec/govreport-test}{+0.0000}
\defval{slope/x40/1024/ced-noloop-minus-ced-dec/govreport-test/hi}{+0.0003}
\defval{slope/x40/1024/ced-noloop-minus-ced-dec/govreport-test/lo}{$-$0.0003}
\defval{slope/x40/1024/ced-noloop-minus-ced-dec/longdocs}{$-$0.0006}
\defval{slope/x40/1024/ced-noloop-minus-ced-dec/longdocs/hi}{$-$0.0003}
\defval{slope/x40/1024/ced-noloop-minus-ced-dec/longdocs/lo}{$-$0.0009}
\defval{slope/x40/1024/ced-noloop-minus-ced-dec/pg19-test}{$-$0.0004}
\defval{slope/x40/1024/ced-noloop-minus-ced-dec/pg19-test/hi}{+0.0001}
\defval{slope/x40/1024/ced-noloop-minus-ced-dec/pg19-test/lo}{$-$0.0008}
\defval{slope/x40/1024/ced-noloop-minus-ced-dec/pg19-val}{+0.0007}
\defval{slope/x40/1024/ced-noloop-minus-ced-dec/pg19-val/hi}{+0.0015}
\defval{slope/x40/1024/ced-noloop-minus-ced-dec/pg19-val/lo}{$-$0.0001}
\defval{slope/x40/1024/ced-noloop-minus-ced-enc/finepdfs-long}{+0.0017}
\defval{slope/x40/1024/ced-noloop-minus-ced-enc/finepdfs-long/hi}{+0.0025}
\defval{slope/x40/1024/ced-noloop-minus-ced-enc/finepdfs-long/lo}{+0.0010}
\defval{slope/x40/1024/ced-noloop-minus-ced-enc/fineweb-edu-100bt}{+0.0009}
\defval{slope/x40/1024/ced-noloop-minus-ced-enc/fineweb-edu-100bt/hi}{+0.0015}
\defval{slope/x40/1024/ced-noloop-minus-ced-enc/fineweb-edu-100bt/lo}{+0.0003}
\defval{slope/x40/1024/ced-noloop-minus-ced-enc/govreport-test}{+0.0013}
\defval{slope/x40/1024/ced-noloop-minus-ced-enc/govreport-test/hi}{+0.0017}
\defval{slope/x40/1024/ced-noloop-minus-ced-enc/govreport-test/lo}{+0.0010}
\defval{slope/x40/1024/ced-noloop-minus-ced-enc/longdocs}{+0.0012}
\defval{slope/x40/1024/ced-noloop-minus-ced-enc/longdocs/hi}{+0.0015}
\defval{slope/x40/1024/ced-noloop-minus-ced-enc/longdocs/lo}{+0.0009}
\defval{slope/x40/1024/ced-noloop-minus-ced-enc/pg19-test}{+0.0004}
\defval{slope/x40/1024/ced-noloop-minus-ced-enc/pg19-test/hi}{+0.0009}
\defval{slope/x40/1024/ced-noloop-minus-ced-enc/pg19-test/lo}{$-$0.0000}
\defval{slope/x40/1024/ced-noloop-minus-ced-enc/pg19-val}{+0.0007}
\defval{slope/x40/1024/ced-noloop-minus-ced-enc/pg19-val/hi}{+0.0015}
\defval{slope/x40/1024/ced-noloop-minus-ced-enc/pg19-val/lo}{$-$0.0000}
\defval{slope/x40/1024/ced-noloop-minus-dense-noloop/finepdfs-long}{$-$0.0007}
\defval{slope/x40/1024/ced-noloop-minus-dense-noloop/finepdfs-long/hi}{$-$0.0000}
\defval{slope/x40/1024/ced-noloop-minus-dense-noloop/finepdfs-long/lo}{$-$0.0014}
\defval{slope/x40/1024/ced-noloop-minus-dense-noloop/fineweb-edu-100bt}{+0.0004}
\defval{slope/x40/1024/ced-noloop-minus-dense-noloop/fineweb-edu-100bt/hi}{+0.0009}
\defval{slope/x40/1024/ced-noloop-minus-dense-noloop/fineweb-edu-100bt/lo}{$-$0.0001}
\defval{slope/x40/1024/ced-noloop-minus-dense-noloop/govreport-test}{+0.0011}
\defval{slope/x40/1024/ced-noloop-minus-dense-noloop/govreport-test/hi}{+0.0014}
\defval{slope/x40/1024/ced-noloop-minus-dense-noloop/govreport-test/lo}{+0.0008}
\defval{slope/x40/1024/ced-noloop-minus-dense-noloop/longdocs}{$-$0.0001}
\defval{slope/x40/1024/ced-noloop-minus-dense-noloop/longdocs/hi}{+0.0002}
\defval{slope/x40/1024/ced-noloop-minus-dense-noloop/longdocs/lo}{$-$0.0003}
\defval{slope/x40/1024/ced-noloop-minus-dense-noloop/pg19-test}{$-$0.0006}
\defval{slope/x40/1024/ced-noloop-minus-dense-noloop/pg19-test/hi}{$-$0.0000}
\defval{slope/x40/1024/ced-noloop-minus-dense-noloop/pg19-test/lo}{$-$0.0011}
\defval{slope/x40/1024/ced-noloop-minus-dense-noloop/pg19-val}{$-$0.0005}
\defval{slope/x40/1024/ced-noloop-minus-dense-noloop/pg19-val/hi}{+0.0004}
\defval{slope/x40/1024/ced-noloop-minus-dense-noloop/pg19-val/lo}{$-$0.0014}
\defval{slope/x40/1024/ced-noloop-minus-yocou-noloop/finepdfs-long}{$-$0.0032}
\defval{slope/x40/1024/ced-noloop-minus-yocou-noloop/finepdfs-long/hi}{$-$0.0027}
\defval{slope/x40/1024/ced-noloop-minus-yocou-noloop/finepdfs-long/lo}{$-$0.0038}
\defval{slope/x40/1024/ced-noloop-minus-yocou-noloop/fineweb-edu-100bt}{$-$0.0008}
\defval{slope/x40/1024/ced-noloop-minus-yocou-noloop/fineweb-edu-100bt/hi}{$-$0.0004}
\defval{slope/x40/1024/ced-noloop-minus-yocou-noloop/fineweb-edu-100bt/lo}{$-$0.0013}
\defval{slope/x40/1024/ced-noloop-minus-yocou-noloop/govreport-test}{$-$0.0008}
\defval{slope/x40/1024/ced-noloop-minus-yocou-noloop/govreport-test/hi}{$-$0.0005}
\defval{slope/x40/1024/ced-noloop-minus-yocou-noloop/govreport-test/lo}{$-$0.0011}
\defval{slope/x40/1024/ced-noloop-minus-yocou-noloop/longdocs}{$-$0.0024}
\defval{slope/x40/1024/ced-noloop-minus-yocou-noloop/longdocs/hi}{$-$0.0021}
\defval{slope/x40/1024/ced-noloop-minus-yocou-noloop/longdocs/lo}{$-$0.0027}
\defval{slope/x40/1024/ced-noloop-minus-yocou-noloop/pg19-test}{$-$0.0032}
\defval{slope/x40/1024/ced-noloop-minus-yocou-noloop/pg19-test/hi}{$-$0.0027}
\defval{slope/x40/1024/ced-noloop-minus-yocou-noloop/pg19-test/lo}{$-$0.0037}
\defval{slope/x40/1024/ced-noloop-minus-yocou-noloop/pg19-val}{$-$0.0026}
\defval{slope/x40/1024/ced-noloop-minus-yocou-noloop/pg19-val/hi}{$-$0.0018}
\defval{slope/x40/1024/ced-noloop-minus-yocou-noloop/pg19-val/lo}{$-$0.0034}
\defval{slope/x40/1024/dense-l47-minus-dense-l25/finepdfs-long}{+0.0007}
\defval{slope/x40/1024/dense-l47-minus-dense-l25/finepdfs-long/hi}{+0.0012}
\defval{slope/x40/1024/dense-l47-minus-dense-l25/finepdfs-long/lo}{+0.0003}
\defval{slope/x40/1024/dense-l47-minus-dense-l25/fineweb-edu-100bt}{+0.0003}
\defval{slope/x40/1024/dense-l47-minus-dense-l25/fineweb-edu-100bt/hi}{+0.0007}
\defval{slope/x40/1024/dense-l47-minus-dense-l25/fineweb-edu-100bt/lo}{$-$0.0001}
\defval{slope/x40/1024/dense-l47-minus-dense-l25/govreport-test}{+0.0006}
\defval{slope/x40/1024/dense-l47-minus-dense-l25/govreport-test/hi}{+0.0009}
\defval{slope/x40/1024/dense-l47-minus-dense-l25/govreport-test/lo}{+0.0004}
\defval{slope/x40/1024/dense-l47-minus-dense-l25/longdocs}{+0.0006}
\defval{slope/x40/1024/dense-l47-minus-dense-l25/longdocs/hi}{+0.0008}
\defval{slope/x40/1024/dense-l47-minus-dense-l25/longdocs/lo}{+0.0003}
\defval{slope/x40/1024/dense-l47-minus-dense-l25/pg19-test}{+0.0003}
\defval{slope/x40/1024/dense-l47-minus-dense-l25/pg19-test/hi}{+0.0008}
\defval{slope/x40/1024/dense-l47-minus-dense-l25/pg19-test/lo}{$-$0.0001}
\defval{slope/x40/1024/dense-l47-minus-dense-l25/pg19-val}{$-$0.0001}
\defval{slope/x40/1024/dense-l47-minus-dense-l25/pg19-val/hi}{+0.0006}
\defval{slope/x40/1024/dense-l47-minus-dense-l25/pg19-val/lo}{$-$0.0010}
\defval{slope/x40/1024/dense-noloop-minus-dense-l25/finepdfs-long}{+0.0017}
\defval{slope/x40/1024/dense-noloop-minus-dense-l25/finepdfs-long/hi}{+0.0024}
\defval{slope/x40/1024/dense-noloop-minus-dense-l25/finepdfs-long/lo}{+0.0011}
\defval{slope/x40/1024/dense-noloop-minus-dense-l25/fineweb-edu-100bt}{+0.0010}
\defval{slope/x40/1024/dense-noloop-minus-dense-l25/fineweb-edu-100bt/hi}{+0.0016}
\defval{slope/x40/1024/dense-noloop-minus-dense-l25/fineweb-edu-100bt/lo}{+0.0005}
\defval{slope/x40/1024/dense-noloop-minus-dense-l25/govreport-test}{+0.0012}
\defval{slope/x40/1024/dense-noloop-minus-dense-l25/govreport-test/hi}{+0.0016}
\defval{slope/x40/1024/dense-noloop-minus-dense-l25/govreport-test/lo}{+0.0009}
\defval{slope/x40/1024/dense-noloop-minus-dense-l25/longdocs}{+0.0013}
\defval{slope/x40/1024/dense-noloop-minus-dense-l25/longdocs/hi}{+0.0016}
\defval{slope/x40/1024/dense-noloop-minus-dense-l25/longdocs/lo}{+0.0010}
\defval{slope/x40/1024/dense-noloop-minus-dense-l25/pg19-test}{+0.0010}
\defval{slope/x40/1024/dense-noloop-minus-dense-l25/pg19-test/hi}{+0.0015}
\defval{slope/x40/1024/dense-noloop-minus-dense-l25/pg19-test/lo}{+0.0005}
\defval{slope/x40/1024/dense-noloop-minus-dense-l25/pg19-val}{+0.0012}
\defval{slope/x40/1024/dense-noloop-minus-dense-l25/pg19-val/hi}{+0.0021}
\defval{slope/x40/1024/dense-noloop-minus-dense-l25/pg19-val/lo}{+0.0003}
\defval{slope/x40/1024/yocou-loop-minus-ced-enc/finepdfs-long}{+0.0049}
\defval{slope/x40/1024/yocou-loop-minus-ced-enc/finepdfs-long/hi}{+0.0056}
\defval{slope/x40/1024/yocou-loop-minus-ced-enc/finepdfs-long/lo}{+0.0044}
\defval{slope/x40/1024/yocou-loop-minus-ced-enc/fineweb-edu-100bt}{+0.0017}
\defval{slope/x40/1024/yocou-loop-minus-ced-enc/fineweb-edu-100bt/hi}{+0.0022}
\defval{slope/x40/1024/yocou-loop-minus-ced-enc/fineweb-edu-100bt/lo}{+0.0013}
\defval{slope/x40/1024/yocou-loop-minus-ced-enc/govreport-test}{+0.0026}
\defval{slope/x40/1024/yocou-loop-minus-ced-enc/govreport-test/hi}{+0.0029}
\defval{slope/x40/1024/yocou-loop-minus-ced-enc/govreport-test/lo}{+0.0023}
\defval{slope/x40/1024/yocou-loop-minus-ced-enc/longdocs}{+0.0037}
\defval{slope/x40/1024/yocou-loop-minus-ced-enc/longdocs/hi}{+0.0039}
\defval{slope/x40/1024/yocou-loop-minus-ced-enc/longdocs/lo}{+0.0034}
\defval{slope/x40/1024/yocou-loop-minus-ced-enc/pg19-test}{+0.0035}
\defval{slope/x40/1024/yocou-loop-minus-ced-enc/pg19-test/hi}{+0.0039}
\defval{slope/x40/1024/yocou-loop-minus-ced-enc/pg19-test/lo}{+0.0030}
\defval{slope/x40/1024/yocou-loop-minus-ced-enc/pg19-val}{+0.0025}
\defval{slope/x40/1024/yocou-loop-minus-ced-enc/pg19-val/hi}{+0.0033}
\defval{slope/x40/1024/yocou-loop-minus-ced-enc/pg19-val/lo}{+0.0016}
\defval{slope/x40/1024/yocou-loop-minus-dense-l25/finepdfs-long}{+0.0042}
\defval{slope/x40/1024/yocou-loop-minus-dense-l25/finepdfs-long/hi}{+0.0049}
\defval{slope/x40/1024/yocou-loop-minus-dense-l25/finepdfs-long/lo}{+0.0035}
\defval{slope/x40/1024/yocou-loop-minus-dense-l25/fineweb-edu-100bt}{+0.0022}
\defval{slope/x40/1024/yocou-loop-minus-dense-l25/fineweb-edu-100bt/hi}{+0.0027}
\defval{slope/x40/1024/yocou-loop-minus-dense-l25/fineweb-edu-100bt/lo}{+0.0018}
\defval{slope/x40/1024/yocou-loop-minus-dense-l25/govreport-test}{+0.0036}
\defval{slope/x40/1024/yocou-loop-minus-dense-l25/govreport-test/hi}{+0.0039}
\defval{slope/x40/1024/yocou-loop-minus-dense-l25/govreport-test/lo}{+0.0033}
\defval{slope/x40/1024/yocou-loop-minus-dense-l25/longdocs}{+0.0037}
\defval{slope/x40/1024/yocou-loop-minus-dense-l25/longdocs/hi}{+0.0040}
\defval{slope/x40/1024/yocou-loop-minus-dense-l25/longdocs/lo}{+0.0034}
\defval{slope/x40/1024/yocou-loop-minus-dense-l25/pg19-test}{+0.0034}
\defval{slope/x40/1024/yocou-loop-minus-dense-l25/pg19-test/hi}{+0.0040}
\defval{slope/x40/1024/yocou-loop-minus-dense-l25/pg19-test/lo}{+0.0029}
\defval{slope/x40/1024/yocou-loop-minus-dense-l25/pg19-val}{+0.0024}
\defval{slope/x40/1024/yocou-loop-minus-dense-l25/pg19-val/hi}{+0.0032}
\defval{slope/x40/1024/yocou-loop-minus-dense-l25/pg19-val/lo}{+0.0016}
\defval{slope/x40/1024/yocou-noloop-minus-ced-noloop/finepdfs-long}{+0.0032}
\defval{slope/x40/1024/yocou-noloop-minus-ced-noloop/finepdfs-long/hi}{+0.0038}
\defval{slope/x40/1024/yocou-noloop-minus-ced-noloop/finepdfs-long/lo}{+0.0027}
\defval{slope/x40/1024/yocou-noloop-minus-ced-noloop/fineweb-edu-100bt}{+0.0008}
\defval{slope/x40/1024/yocou-noloop-minus-ced-noloop/fineweb-edu-100bt/hi}{+0.0013}
\defval{slope/x40/1024/yocou-noloop-minus-ced-noloop/fineweb-edu-100bt/lo}{+0.0004}
\defval{slope/x40/1024/yocou-noloop-minus-ced-noloop/govreport-test}{+0.0008}
\defval{slope/x40/1024/yocou-noloop-minus-ced-noloop/govreport-test/hi}{+0.0011}
\defval{slope/x40/1024/yocou-noloop-minus-ced-noloop/govreport-test/lo}{+0.0005}
\defval{slope/x40/1024/yocou-noloop-minus-ced-noloop/longdocs}{+0.0024}
\defval{slope/x40/1024/yocou-noloop-minus-ced-noloop/longdocs/hi}{+0.0027}
\defval{slope/x40/1024/yocou-noloop-minus-ced-noloop/longdocs/lo}{+0.0021}
\defval{slope/x40/1024/yocou-noloop-minus-ced-noloop/pg19-test}{+0.0032}
\defval{slope/x40/1024/yocou-noloop-minus-ced-noloop/pg19-test/hi}{+0.0037}
\defval{slope/x40/1024/yocou-noloop-minus-ced-noloop/pg19-test/lo}{+0.0027}
\defval{slope/x40/1024/yocou-noloop-minus-ced-noloop/pg19-val}{+0.0026}
\defval{slope/x40/1024/yocou-noloop-minus-ced-noloop/pg19-val/hi}{+0.0034}
\defval{slope/x40/1024/yocou-noloop-minus-ced-noloop/pg19-val/lo}{+0.0018}
\defval{slope/x40/1024/yocou-noloop-minus-dense-noloop/finepdfs-long}{+0.0025}
\defval{slope/x40/1024/yocou-noloop-minus-dense-noloop/finepdfs-long/hi}{+0.0032}
\defval{slope/x40/1024/yocou-noloop-minus-dense-noloop/finepdfs-long/lo}{+0.0019}
\defval{slope/x40/1024/yocou-noloop-minus-dense-noloop/fineweb-edu-100bt}{+0.0012}
\defval{slope/x40/1024/yocou-noloop-minus-dense-noloop/fineweb-edu-100bt/hi}{+0.0017}
\defval{slope/x40/1024/yocou-noloop-minus-dense-noloop/fineweb-edu-100bt/lo}{+0.0007}
\defval{slope/x40/1024/yocou-noloop-minus-dense-noloop/govreport-test}{+0.0019}
\defval{slope/x40/1024/yocou-noloop-minus-dense-noloop/govreport-test/hi}{+0.0022}
\defval{slope/x40/1024/yocou-noloop-minus-dense-noloop/govreport-test/lo}{+0.0016}
\defval{slope/x40/1024/yocou-noloop-minus-dense-noloop/longdocs}{+0.0024}
\defval{slope/x40/1024/yocou-noloop-minus-dense-noloop/longdocs/hi}{+0.0026}
\defval{slope/x40/1024/yocou-noloop-minus-dense-noloop/longdocs/lo}{+0.0021}
\defval{slope/x40/1024/yocou-noloop-minus-dense-noloop/pg19-test}{+0.0026}
\defval{slope/x40/1024/yocou-noloop-minus-dense-noloop/pg19-test/hi}{+0.0031}
\defval{slope/x40/1024/yocou-noloop-minus-dense-noloop/pg19-test/lo}{+0.0021}
\defval{slope/x40/1024/yocou-noloop-minus-dense-noloop/pg19-val}{+0.0021}
\defval{slope/x40/1024/yocou-noloop-minus-dense-noloop/pg19-val/hi}{+0.0030}
\defval{slope/x40/1024/yocou-noloop-minus-dense-noloop/pg19-val/lo}{+0.0012}
\defval{slope/x40/1024/yocou-noloop-minus-yocou-loop/finepdfs-long}{+0.0000}
\defval{slope/x40/1024/yocou-noloop-minus-yocou-loop/finepdfs-long/hi}{+0.0007}
\defval{slope/x40/1024/yocou-noloop-minus-yocou-loop/finepdfs-long/lo}{$-$0.0006}
\defval{slope/x40/1024/yocou-noloop-minus-yocou-loop/fineweb-edu-100bt}{$-$0.0000}
\defval{slope/x40/1024/yocou-noloop-minus-yocou-loop/fineweb-edu-100bt/hi}{+0.0005}
\defval{slope/x40/1024/yocou-noloop-minus-yocou-loop/fineweb-edu-100bt/lo}{$-$0.0006}
\defval{slope/x40/1024/yocou-noloop-minus-yocou-loop/govreport-test}{$-$0.0005}
\defval{slope/x40/1024/yocou-noloop-minus-yocou-loop/govreport-test/hi}{$-$0.0001}
\defval{slope/x40/1024/yocou-noloop-minus-yocou-loop/govreport-test/lo}{$-$0.0008}
\defval{slope/x40/1024/yocou-noloop-minus-yocou-loop/longdocs}{$-$0.0001}
\defval{slope/x40/1024/yocou-noloop-minus-yocou-loop/longdocs/hi}{+0.0002}
\defval{slope/x40/1024/yocou-noloop-minus-yocou-loop/longdocs/lo}{$-$0.0004}
\defval{slope/x40/1024/yocou-noloop-minus-yocou-loop/pg19-test}{+0.0002}
\defval{slope/x40/1024/yocou-noloop-minus-yocou-loop/pg19-test/hi}{+0.0007}
\defval{slope/x40/1024/yocou-noloop-minus-yocou-loop/pg19-test/lo}{$-$0.0003}
\defval{slope/x40/1024/yocou-noloop-minus-yocou-loop/pg19-val}{+0.0009}
\defval{slope/x40/1024/yocou-noloop-minus-yocou-loop/pg19-val/hi}{+0.0017}
\defval{slope/x40/1024/yocou-noloop-minus-yocou-loop/pg19-val/lo}{+0.0001}
\defval{slopeabsm/long/1024/ced-dec-minus-ced-enc/finepdfs-long}{1.4}
\defval{slopeabsm/long/1024/ced-dec-minus-ced-enc/fineweb-edu-100bt}{1.1}
\defval{slopeabsm/long/1024/ced-dec-minus-ced-enc/govreport-test}{1.3}
\defval{slopeabsm/long/1024/ced-dec-minus-ced-enc/longdocs}{1.1}
\defval{slopeabsm/long/1024/ced-dec-minus-ced-enc/pg19-test}{0.7}
\defval{slopeabsm/long/1024/ced-dec-minus-ced-enc/pg19-val}{0.2}
\defval{slopeabsm/long/1024/ced-dec-minus-dense-l47/finepdfs-long}{2.2}
\defval{slopeabsm/long/1024/ced-dec-minus-dense-l47/fineweb-edu-100bt}{1.3}
\defval{slopeabsm/long/1024/ced-dec-minus-dense-l47/govreport-test}{2.1}
\defval{slopeabsm/long/1024/ced-dec-minus-dense-l47/longdocs}{2.3}
\defval{slopeabsm/long/1024/ced-dec-minus-dense-l47/pg19-test}{2.7}
\defval{slopeabsm/long/1024/ced-dec-minus-dense-l47/pg19-val}{3.1}
\defval{slopeabsm/long/1024/ced-enc-minus-dense-l47/finepdfs-long}{0.8}
\defval{slopeabsm/long/1024/ced-enc-minus-dense-l47/fineweb-edu-100bt}{0.1}
\defval{slopeabsm/long/1024/ced-enc-minus-dense-l47/govreport-test}{0.7}
\defval{slopeabsm/long/1024/ced-enc-minus-dense-l47/longdocs}{1.2}
\defval{slopeabsm/long/1024/ced-enc-minus-dense-l47/pg19-test}{2.0}
\defval{slopeabsm/long/1024/ced-enc-minus-dense-l47/pg19-val}{2.8}
\defval{slopeabsm/long/1024/ced-noloop-minus-ced-dec/finepdfs-long}{1.4}
\defval{slopeabsm/long/1024/ced-noloop-minus-ced-dec/fineweb-edu-100bt}{0.3}
\defval{slopeabsm/long/1024/ced-noloop-minus-ced-dec/govreport-test}{0.3}
\defval{slopeabsm/long/1024/ced-noloop-minus-ced-dec/longdocs}{1.1}
\defval{slopeabsm/long/1024/ced-noloop-minus-ced-dec/pg19-test}{1.5}
\defval{slopeabsm/long/1024/ced-noloop-minus-ced-dec/pg19-val}{1.4}
\defval{slopeabsm/long/1024/ced-noloop-minus-ced-enc/finepdfs-long}{2.8}
\defval{slopeabsm/long/1024/ced-noloop-minus-ced-enc/fineweb-edu-100bt}{1.4}
\defval{slopeabsm/long/1024/ced-noloop-minus-ced-enc/govreport-test}{1.6}
\defval{slopeabsm/long/1024/ced-noloop-minus-ced-enc/longdocs}{2.2}
\defval{slopeabsm/long/1024/ced-noloop-minus-ced-enc/pg19-test}{2.2}
\defval{slopeabsm/long/1024/ced-noloop-minus-ced-enc/pg19-val}{1.7}
\defval{slopeabsm/long/1024/ced-noloop-minus-dense-l47/finepdfs-long}{3.6}
\defval{slopeabsm/long/1024/ced-noloop-minus-dense-l47/fineweb-edu-100bt}{1.5}
\defval{slopeabsm/long/1024/ced-noloop-minus-dense-l47/govreport-test}{2.4}
\defval{slopeabsm/long/1024/ced-noloop-minus-dense-l47/longdocs}{3.4}
\defval{slopeabsm/long/1024/ced-noloop-minus-dense-l47/pg19-test}{4.2}
\defval{slopeabsm/long/1024/ced-noloop-minus-dense-l47/pg19-val}{4.5}
\defval{slopeabsm/p140/768/p140-dec-fixed-minus-p140-enc-fixed/finepdfs-long}{4.9}
\defval{slopeabsm/p140/768/p140-dec-fixed-minus-p140-enc-fixed/fineweb-edu-100bt}{1.1}
\defval{slopeabsm/p140/768/p140-dec-fixed-minus-p140-enc-fixed/govreport-test}{2.6}
\defval{slopeabsm/p140/768/p140-dec-fixed-minus-p140-enc-fixed/longdocs}{3.3}
\defval{slopeabsm/p140/768/p140-dec-fixed-minus-p140-enc-fixed/pg19-test}{2.5}
\defval{slopeabsm/p140/768/p140-dec-fixed-minus-p140-enc-fixed/pg19-val}{3.1}
\defval{slopeabsm/p140/768/p140-enc-fixed-minus-p140-enc/finepdfs-long}{0.2}
\defval{slopeabsm/p140/768/p140-enc-fixed-minus-p140-enc/fineweb-edu-100bt}{0.0}
\defval{slopeabsm/p140/768/p140-enc-fixed-minus-p140-enc/govreport-test}{0.1}
\defval{slopeabsm/p140/768/p140-enc-fixed-minus-p140-enc/longdocs}{0.3}
\defval{slopeabsm/p140/768/p140-enc-fixed-minus-p140-enc/pg19-test}{0.7}
\defval{slopeabsm/p140/768/p140-enc-fixed-minus-p140-enc/pg19-val}{0.1}
\defval{slopeabsm/p140/768/p140-noloop-minus-p140-enc/finepdfs-long}{5.1}
\defval{slopeabsm/p140/768/p140-noloop-minus-p140-enc/fineweb-edu-100bt}{1.6}
\defval{slopeabsm/p140/768/p140-noloop-minus-p140-enc/govreport-test}{2.6}
\defval{slopeabsm/p140/768/p140-noloop-minus-p140-enc/longdocs}{3.3}
\defval{slopeabsm/p140/768/p140-noloop-minus-p140-enc/pg19-test}{2.3}
\defval{slopeabsm/p140/768/p140-noloop-minus-p140-enc/pg19-val}{3.1}
\defval{slopeabsm/x40/1024/ced-dec-minus-ced-enc/finepdfs-long}{3.2}
\defval{slopeabsm/x40/1024/ced-dec-minus-ced-enc/fineweb-edu-100bt}{1.1}
\defval{slopeabsm/x40/1024/ced-dec-minus-ced-enc/govreport-test}{1.3}
\defval{slopeabsm/x40/1024/ced-dec-minus-ced-enc/longdocs}{1.8}
\defval{slopeabsm/x40/1024/ced-dec-minus-ced-enc/pg19-test}{0.8}
\defval{slopeabsm/x40/1024/ced-dec-minus-ced-enc/pg19-val}{0.1}
\defval{slopeabsm/x40/1024/ced-dec-minus-dense-l25/finepdfs-long}{2.4}
\defval{slopeabsm/x40/1024/ced-dec-minus-dense-l25/fineweb-edu-100bt}{1.6}
\defval{slopeabsm/x40/1024/ced-dec-minus-dense-l25/govreport-test}{2.3}
\defval{slopeabsm/x40/1024/ced-dec-minus-dense-l25/longdocs}{1.9}
\defval{slopeabsm/x40/1024/ced-dec-minus-dense-l25/pg19-test}{0.8}
\defval{slopeabsm/x40/1024/ced-dec-minus-dense-l25/pg19-val}{0.0}
\defval{slopeabsm/x40/1024/ced-enc-fixed-minus-ced-enc/finepdfs-long}{0.4}
\defval{slopeabsm/x40/1024/ced-enc-fixed-minus-ced-enc/fineweb-edu-100bt}{0.1}
\defval{slopeabsm/x40/1024/ced-enc-fixed-minus-ced-enc/govreport-test}{0.4}
\defval{slopeabsm/x40/1024/ced-enc-fixed-minus-ced-enc/longdocs}{0.1}
\defval{slopeabsm/x40/1024/ced-enc-fixed-minus-ced-enc/pg19-test}{0.4}
\defval{slopeabsm/x40/1024/ced-enc-fixed-minus-ced-enc/pg19-val}{0.3}
\defval{slopeabsm/x40/1024/ced-enc-minus-dense-l25/finepdfs-long}{0.8}
\defval{slopeabsm/x40/1024/ced-enc-minus-dense-l25/fineweb-edu-100bt}{0.5}
\defval{slopeabsm/x40/1024/ced-enc-minus-dense-l25/govreport-test}{1.0}
\defval{slopeabsm/x40/1024/ced-enc-minus-dense-l25/longdocs}{0.1}
\defval{slopeabsm/x40/1024/ced-enc-minus-dense-l25/pg19-test}{0.1}
\defval{slopeabsm/x40/1024/ced-enc-minus-dense-l25/pg19-val}{0.1}
\defval{slopeabsm/x40/1024/ced-enc-minus-yocou-loop/finepdfs-long}{4.9}
\defval{slopeabsm/x40/1024/ced-enc-minus-yocou-loop/fineweb-edu-100bt}{1.7}
\defval{slopeabsm/x40/1024/ced-enc-minus-yocou-loop/govreport-test}{2.6}
\defval{slopeabsm/x40/1024/ced-enc-minus-yocou-loop/longdocs}{3.7}
\defval{slopeabsm/x40/1024/ced-enc-minus-yocou-loop/pg19-test}{3.5}
\defval{slopeabsm/x40/1024/ced-enc-minus-yocou-loop/pg19-val}{2.5}
\defval{slopeabsm/x40/1024/ced-noloop-minus-ced-dec/finepdfs-long}{1.4}
\defval{slopeabsm/x40/1024/ced-noloop-minus-ced-dec/fineweb-edu-100bt}{0.2}
\defval{slopeabsm/x40/1024/ced-noloop-minus-ced-dec/govreport-test}{0.0}
\defval{slopeabsm/x40/1024/ced-noloop-minus-ced-dec/longdocs}{0.6}
\defval{slopeabsm/x40/1024/ced-noloop-minus-ced-dec/pg19-test}{0.4}
\defval{slopeabsm/x40/1024/ced-noloop-minus-ced-dec/pg19-val}{0.7}
\defval{slopeabsm/x40/1024/ced-noloop-minus-ced-enc/finepdfs-long}{1.7}
\defval{slopeabsm/x40/1024/ced-noloop-minus-ced-enc/fineweb-edu-100bt}{0.9}
\defval{slopeabsm/x40/1024/ced-noloop-minus-ced-enc/govreport-test}{1.3}
\defval{slopeabsm/x40/1024/ced-noloop-minus-ced-enc/longdocs}{1.2}
\defval{slopeabsm/x40/1024/ced-noloop-minus-ced-enc/pg19-test}{0.4}
\defval{slopeabsm/x40/1024/ced-noloop-minus-ced-enc/pg19-val}{0.7}
\defval{slopeabsm/x40/1024/ced-noloop-minus-dense-noloop/finepdfs-long}{0.7}
\defval{slopeabsm/x40/1024/ced-noloop-minus-dense-noloop/fineweb-edu-100bt}{0.4}
\defval{slopeabsm/x40/1024/ced-noloop-minus-dense-noloop/govreport-test}{1.1}
\defval{slopeabsm/x40/1024/ced-noloop-minus-dense-noloop/longdocs}{0.1}
\defval{slopeabsm/x40/1024/ced-noloop-minus-dense-noloop/pg19-test}{0.6}
\defval{slopeabsm/x40/1024/ced-noloop-minus-dense-noloop/pg19-val}{0.5}
\defval{slopeabsm/x40/1024/ced-noloop-minus-yocou-noloop/finepdfs-long}{3.2}
\defval{slopeabsm/x40/1024/ced-noloop-minus-yocou-noloop/fineweb-edu-100bt}{0.8}
\defval{slopeabsm/x40/1024/ced-noloop-minus-yocou-noloop/govreport-test}{0.8}
\defval{slopeabsm/x40/1024/ced-noloop-minus-yocou-noloop/longdocs}{2.4}
\defval{slopeabsm/x40/1024/ced-noloop-minus-yocou-noloop/pg19-test}{3.2}
\defval{slopeabsm/x40/1024/ced-noloop-minus-yocou-noloop/pg19-val}{2.6}
\defval{slopeabsm/x40/1024/dense-l47-minus-dense-l25/finepdfs-long}{0.7}
\defval{slopeabsm/x40/1024/dense-l47-minus-dense-l25/fineweb-edu-100bt}{0.3}
\defval{slopeabsm/x40/1024/dense-l47-minus-dense-l25/govreport-test}{0.6}
\defval{slopeabsm/x40/1024/dense-l47-minus-dense-l25/longdocs}{0.6}
\defval{slopeabsm/x40/1024/dense-l47-minus-dense-l25/pg19-test}{0.3}
\defval{slopeabsm/x40/1024/dense-l47-minus-dense-l25/pg19-val}{0.1}
\defval{slopeabsm/x40/1024/dense-noloop-minus-dense-l25/finepdfs-long}{1.7}
\defval{slopeabsm/x40/1024/dense-noloop-minus-dense-l25/fineweb-edu-100bt}{1.0}
\defval{slopeabsm/x40/1024/dense-noloop-minus-dense-l25/govreport-test}{1.2}
\defval{slopeabsm/x40/1024/dense-noloop-minus-dense-l25/longdocs}{1.3}
\defval{slopeabsm/x40/1024/dense-noloop-minus-dense-l25/pg19-test}{1.0}
\defval{slopeabsm/x40/1024/dense-noloop-minus-dense-l25/pg19-val}{1.2}
\defval{slopeabsm/x40/1024/yocou-loop-minus-ced-enc/finepdfs-long}{4.9}
\defval{slopeabsm/x40/1024/yocou-loop-minus-ced-enc/fineweb-edu-100bt}{1.7}
\defval{slopeabsm/x40/1024/yocou-loop-minus-ced-enc/govreport-test}{2.6}
\defval{slopeabsm/x40/1024/yocou-loop-minus-ced-enc/longdocs}{3.7}
\defval{slopeabsm/x40/1024/yocou-loop-minus-ced-enc/pg19-test}{3.5}
\defval{slopeabsm/x40/1024/yocou-loop-minus-ced-enc/pg19-val}{2.5}
\defval{slopeabsm/x40/1024/yocou-loop-minus-dense-l25/finepdfs-long}{4.2}
\defval{slopeabsm/x40/1024/yocou-loop-minus-dense-l25/fineweb-edu-100bt}{2.2}
\defval{slopeabsm/x40/1024/yocou-loop-minus-dense-l25/govreport-test}{3.6}
\defval{slopeabsm/x40/1024/yocou-loop-minus-dense-l25/longdocs}{3.7}
\defval{slopeabsm/x40/1024/yocou-loop-minus-dense-l25/pg19-test}{3.4}
\defval{slopeabsm/x40/1024/yocou-loop-minus-dense-l25/pg19-val}{2.4}
\defval{slopeabsm/x40/1024/yocou-noloop-minus-ced-noloop/finepdfs-long}{3.2}
\defval{slopeabsm/x40/1024/yocou-noloop-minus-ced-noloop/fineweb-edu-100bt}{0.8}
\defval{slopeabsm/x40/1024/yocou-noloop-minus-ced-noloop/govreport-test}{0.8}
\defval{slopeabsm/x40/1024/yocou-noloop-minus-ced-noloop/longdocs}{2.4}
\defval{slopeabsm/x40/1024/yocou-noloop-minus-ced-noloop/pg19-test}{3.2}
\defval{slopeabsm/x40/1024/yocou-noloop-minus-ced-noloop/pg19-val}{2.6}
\defval{slopeabsm/x40/1024/yocou-noloop-minus-dense-noloop/finepdfs-long}{2.5}
\defval{slopeabsm/x40/1024/yocou-noloop-minus-dense-noloop/fineweb-edu-100bt}{1.2}
\defval{slopeabsm/x40/1024/yocou-noloop-minus-dense-noloop/govreport-test}{1.9}
\defval{slopeabsm/x40/1024/yocou-noloop-minus-dense-noloop/longdocs}{2.4}
\defval{slopeabsm/x40/1024/yocou-noloop-minus-dense-noloop/pg19-test}{2.6}
\defval{slopeabsm/x40/1024/yocou-noloop-minus-dense-noloop/pg19-val}{2.1}
\defval{slopeabsm/x40/1024/yocou-noloop-minus-yocou-loop/finepdfs-long}{0.0}
\defval{slopeabsm/x40/1024/yocou-noloop-minus-yocou-loop/fineweb-edu-100bt}{0.0}
\defval{slopeabsm/x40/1024/yocou-noloop-minus-yocou-loop/govreport-test}{0.5}
\defval{slopeabsm/x40/1024/yocou-noloop-minus-yocou-loop/longdocs}{0.1}
\defval{slopeabsm/x40/1024/yocou-noloop-minus-yocou-loop/pg19-test}{0.2}
\defval{slopeabsm/x40/1024/yocou-noloop-minus-yocou-loop/pg19-val}{0.9}
\defval{slopem/long/1024/ced-dec-minus-ced-enc/finepdfs-long}{+1.4}
\defval{slopem/long/1024/ced-dec-minus-ced-enc/finepdfs-long/hi}{+2.1}
\defval{slopem/long/1024/ced-dec-minus-ced-enc/finepdfs-long/lo}{+0.7}
\defval{slopem/long/1024/ced-dec-minus-ced-enc/fineweb-edu-100bt}{+1.1}
\defval{slopem/long/1024/ced-dec-minus-ced-enc/fineweb-edu-100bt/hi}{+1.6}
\defval{slopem/long/1024/ced-dec-minus-ced-enc/fineweb-edu-100bt/lo}{+0.6}
\defval{slopem/long/1024/ced-dec-minus-ced-enc/govreport-test}{+1.3}
\defval{slopem/long/1024/ced-dec-minus-ced-enc/govreport-test/hi}{+1.7}
\defval{slopem/long/1024/ced-dec-minus-ced-enc/govreport-test/lo}{+1.0}
\defval{slopem/long/1024/ced-dec-minus-ced-enc/longdocs}{+1.1}
\defval{slopem/long/1024/ced-dec-minus-ced-enc/longdocs/hi}{+1.4}
\defval{slopem/long/1024/ced-dec-minus-ced-enc/longdocs/lo}{+0.8}
\defval{slopem/long/1024/ced-dec-minus-ced-enc/pg19-test}{+0.7}
\defval{slopem/long/1024/ced-dec-minus-ced-enc/pg19-test/hi}{+1.2}
\defval{slopem/long/1024/ced-dec-minus-ced-enc/pg19-test/lo}{+0.2}
\defval{slopem/long/1024/ced-dec-minus-ced-enc/pg19-val}{+0.2}
\defval{slopem/long/1024/ced-dec-minus-ced-enc/pg19-val/hi}{+1.2}
\defval{slopem/long/1024/ced-dec-minus-ced-enc/pg19-val/lo}{$-$0.7}
\defval{slopem/long/1024/ced-dec-minus-dense-l47/finepdfs-long}{+2.2}
\defval{slopem/long/1024/ced-dec-minus-dense-l47/finepdfs-long/hi}{+2.9}
\defval{slopem/long/1024/ced-dec-minus-dense-l47/finepdfs-long/lo}{+1.6}
\defval{slopem/long/1024/ced-dec-minus-dense-l47/fineweb-edu-100bt}{+1.3}
\defval{slopem/long/1024/ced-dec-minus-dense-l47/fineweb-edu-100bt/hi}{+1.7}
\defval{slopem/long/1024/ced-dec-minus-dense-l47/fineweb-edu-100bt/lo}{+0.8}
\defval{slopem/long/1024/ced-dec-minus-dense-l47/govreport-test}{+2.1}
\defval{slopem/long/1024/ced-dec-minus-dense-l47/govreport-test/hi}{+2.4}
\defval{slopem/long/1024/ced-dec-minus-dense-l47/govreport-test/lo}{+1.8}
\defval{slopem/long/1024/ced-dec-minus-dense-l47/longdocs}{+2.3}
\defval{slopem/long/1024/ced-dec-minus-dense-l47/longdocs/hi}{+2.6}
\defval{slopem/long/1024/ced-dec-minus-dense-l47/longdocs/lo}{+2.1}
\defval{slopem/long/1024/ced-dec-minus-dense-l47/pg19-test}{+2.7}
\defval{slopem/long/1024/ced-dec-minus-dense-l47/pg19-test/hi}{+3.2}
\defval{slopem/long/1024/ced-dec-minus-dense-l47/pg19-test/lo}{+2.2}
\defval{slopem/long/1024/ced-dec-minus-dense-l47/pg19-val}{+3.1}
\defval{slopem/long/1024/ced-dec-minus-dense-l47/pg19-val/hi}{+4.0}
\defval{slopem/long/1024/ced-dec-minus-dense-l47/pg19-val/lo}{+2.1}
\defval{slopem/long/1024/ced-enc-minus-dense-l47/finepdfs-long}{+0.8}
\defval{slopem/long/1024/ced-enc-minus-dense-l47/finepdfs-long/hi}{+1.4}
\defval{slopem/long/1024/ced-enc-minus-dense-l47/finepdfs-long/lo}{+0.3}
\defval{slopem/long/1024/ced-enc-minus-dense-l47/fineweb-edu-100bt}{+0.1}
\defval{slopem/long/1024/ced-enc-minus-dense-l47/fineweb-edu-100bt/hi}{+0.6}
\defval{slopem/long/1024/ced-enc-minus-dense-l47/fineweb-edu-100bt/lo}{$-$0.3}
\defval{slopem/long/1024/ced-enc-minus-dense-l47/govreport-test}{+0.7}
\defval{slopem/long/1024/ced-enc-minus-dense-l47/govreport-test/hi}{+1.0}
\defval{slopem/long/1024/ced-enc-minus-dense-l47/govreport-test/lo}{+0.5}
\defval{slopem/long/1024/ced-enc-minus-dense-l47/longdocs}{+1.2}
\defval{slopem/long/1024/ced-enc-minus-dense-l47/longdocs/hi}{+1.4}
\defval{slopem/long/1024/ced-enc-minus-dense-l47/longdocs/lo}{+0.9}
\defval{slopem/long/1024/ced-enc-minus-dense-l47/pg19-test}{+2.0}
\defval{slopem/long/1024/ced-enc-minus-dense-l47/pg19-test/hi}{+2.5}
\defval{slopem/long/1024/ced-enc-minus-dense-l47/pg19-test/lo}{+1.5}
\defval{slopem/long/1024/ced-enc-minus-dense-l47/pg19-val}{+2.8}
\defval{slopem/long/1024/ced-enc-minus-dense-l47/pg19-val/hi}{+3.7}
\defval{slopem/long/1024/ced-enc-minus-dense-l47/pg19-val/lo}{+1.9}
\defval{slopem/long/1024/ced-noloop-minus-ced-dec/finepdfs-long}{+1.4}
\defval{slopem/long/1024/ced-noloop-minus-ced-dec/finepdfs-long/hi}{+1.9}
\defval{slopem/long/1024/ced-noloop-minus-ced-dec/finepdfs-long/lo}{+0.9}
\defval{slopem/long/1024/ced-noloop-minus-ced-dec/fineweb-edu-100bt}{+0.3}
\defval{slopem/long/1024/ced-noloop-minus-ced-dec/fineweb-edu-100bt/hi}{+0.7}
\defval{slopem/long/1024/ced-noloop-minus-ced-dec/fineweb-edu-100bt/lo}{$-$0.2}
\defval{slopem/long/1024/ced-noloop-minus-ced-dec/govreport-test}{+0.3}
\defval{slopem/long/1024/ced-noloop-minus-ced-dec/govreport-test/hi}{+0.6}
\defval{slopem/long/1024/ced-noloop-minus-ced-dec/govreport-test/lo}{+0.0}
\defval{slopem/long/1024/ced-noloop-minus-ced-dec/longdocs}{+1.1}
\defval{slopem/long/1024/ced-noloop-minus-ced-dec/longdocs/hi}{+1.3}
\defval{slopem/long/1024/ced-noloop-minus-ced-dec/longdocs/lo}{+0.8}
\defval{slopem/long/1024/ced-noloop-minus-ced-dec/pg19-test}{+1.5}
\defval{slopem/long/1024/ced-noloop-minus-ced-dec/pg19-test/hi}{+2.0}
\defval{slopem/long/1024/ced-noloop-minus-ced-dec/pg19-test/lo}{+1.0}
\defval{slopem/long/1024/ced-noloop-minus-ced-dec/pg19-val}{+1.4}
\defval{slopem/long/1024/ced-noloop-minus-ced-dec/pg19-val/hi}{+2.3}
\defval{slopem/long/1024/ced-noloop-minus-ced-dec/pg19-val/lo}{+0.7}
\defval{slopem/long/1024/ced-noloop-minus-ced-enc/finepdfs-long}{+2.8}
\defval{slopem/long/1024/ced-noloop-minus-ced-enc/finepdfs-long/hi}{+3.6}
\defval{slopem/long/1024/ced-noloop-minus-ced-enc/finepdfs-long/lo}{+2.1}
\defval{slopem/long/1024/ced-noloop-minus-ced-enc/fineweb-edu-100bt}{+1.4}
\defval{slopem/long/1024/ced-noloop-minus-ced-enc/fineweb-edu-100bt/hi}{+1.9}
\defval{slopem/long/1024/ced-noloop-minus-ced-enc/fineweb-edu-100bt/lo}{+0.8}
\defval{slopem/long/1024/ced-noloop-minus-ced-enc/govreport-test}{+1.6}
\defval{slopem/long/1024/ced-noloop-minus-ced-enc/govreport-test/hi}{+2.0}
\defval{slopem/long/1024/ced-noloop-minus-ced-enc/govreport-test/lo}{+1.3}
\defval{slopem/long/1024/ced-noloop-minus-ced-enc/longdocs}{+2.2}
\defval{slopem/long/1024/ced-noloop-minus-ced-enc/longdocs/hi}{+2.5}
\defval{slopem/long/1024/ced-noloop-minus-ced-enc/longdocs/lo}{+1.9}
\defval{slopem/long/1024/ced-noloop-minus-ced-enc/pg19-test}{+2.2}
\defval{slopem/long/1024/ced-noloop-minus-ced-enc/pg19-test/hi}{+2.8}
\defval{slopem/long/1024/ced-noloop-minus-ced-enc/pg19-test/lo}{+1.6}
\defval{slopem/long/1024/ced-noloop-minus-ced-enc/pg19-val}{+1.7}
\defval{slopem/long/1024/ced-noloop-minus-ced-enc/pg19-val/hi}{+2.6}
\defval{slopem/long/1024/ced-noloop-minus-ced-enc/pg19-val/lo}{+0.8}
\defval{slopem/long/1024/ced-noloop-minus-dense-l47/finepdfs-long}{+3.6}
\defval{slopem/long/1024/ced-noloop-minus-dense-l47/finepdfs-long/hi}{+4.3}
\defval{slopem/long/1024/ced-noloop-minus-dense-l47/finepdfs-long/lo}{+3.0}
\defval{slopem/long/1024/ced-noloop-minus-dense-l47/fineweb-edu-100bt}{+1.5}
\defval{slopem/long/1024/ced-noloop-minus-dense-l47/fineweb-edu-100bt/hi}{+2.1}
\defval{slopem/long/1024/ced-noloop-minus-dense-l47/fineweb-edu-100bt/lo}{+1.0}
\defval{slopem/long/1024/ced-noloop-minus-dense-l47/govreport-test}{+2.4}
\defval{slopem/long/1024/ced-noloop-minus-dense-l47/govreport-test/hi}{+2.7}
\defval{slopem/long/1024/ced-noloop-minus-dense-l47/govreport-test/lo}{+2.1}
\defval{slopem/long/1024/ced-noloop-minus-dense-l47/longdocs}{+3.4}
\defval{slopem/long/1024/ced-noloop-minus-dense-l47/longdocs/hi}{+3.7}
\defval{slopem/long/1024/ced-noloop-minus-dense-l47/longdocs/lo}{+3.1}
\defval{slopem/long/1024/ced-noloop-minus-dense-l47/pg19-test}{+4.2}
\defval{slopem/long/1024/ced-noloop-minus-dense-l47/pg19-test/hi}{+4.7}
\defval{slopem/long/1024/ced-noloop-minus-dense-l47/pg19-test/lo}{+3.6}
\defval{slopem/long/1024/ced-noloop-minus-dense-l47/pg19-val}{+4.5}
\defval{slopem/long/1024/ced-noloop-minus-dense-l47/pg19-val/hi}{+5.5}
\defval{slopem/long/1024/ced-noloop-minus-dense-l47/pg19-val/lo}{+3.6}
\defval{slopem/p140/768/p140-dec-fixed-minus-p140-enc-fixed/finepdfs-long}{+4.9}
\defval{slopem/p140/768/p140-dec-fixed-minus-p140-enc-fixed/finepdfs-long/hi}{+5.6}
\defval{slopem/p140/768/p140-dec-fixed-minus-p140-enc-fixed/finepdfs-long/lo}{+4.1}
\defval{slopem/p140/768/p140-dec-fixed-minus-p140-enc-fixed/fineweb-edu-100bt}{+1.1}
\defval{slopem/p140/768/p140-dec-fixed-minus-p140-enc-fixed/fineweb-edu-100bt/hi}{+1.6}
\defval{slopem/p140/768/p140-dec-fixed-minus-p140-enc-fixed/fineweb-edu-100bt/lo}{+0.6}
\defval{slopem/p140/768/p140-dec-fixed-minus-p140-enc-fixed/govreport-test}{+2.6}
\defval{slopem/p140/768/p140-dec-fixed-minus-p140-enc-fixed/govreport-test/hi}{+2.9}
\defval{slopem/p140/768/p140-dec-fixed-minus-p140-enc-fixed/govreport-test/lo}{+2.2}
\defval{slopem/p140/768/p140-dec-fixed-minus-p140-enc-fixed/longdocs}{+3.3}
\defval{slopem/p140/768/p140-dec-fixed-minus-p140-enc-fixed/longdocs/hi}{+3.6}
\defval{slopem/p140/768/p140-dec-fixed-minus-p140-enc-fixed/longdocs/lo}{+3.0}
\defval{slopem/p140/768/p140-dec-fixed-minus-p140-enc-fixed/pg19-test}{+2.5}
\defval{slopem/p140/768/p140-dec-fixed-minus-p140-enc-fixed/pg19-test/hi}{+3.0}
\defval{slopem/p140/768/p140-dec-fixed-minus-p140-enc-fixed/pg19-test/lo}{+2.0}
\defval{slopem/p140/768/p140-dec-fixed-minus-p140-enc-fixed/pg19-val}{+3.1}
\defval{slopem/p140/768/p140-dec-fixed-minus-p140-enc-fixed/pg19-val/hi}{+4.0}
\defval{slopem/p140/768/p140-dec-fixed-minus-p140-enc-fixed/pg19-val/lo}{+2.3}
\defval{slopem/p140/768/p140-enc-fixed-minus-p140-enc/finepdfs-long}{$-$0.2}
\defval{slopem/p140/768/p140-enc-fixed-minus-p140-enc/finepdfs-long/hi}{+0.3}
\defval{slopem/p140/768/p140-enc-fixed-minus-p140-enc/finepdfs-long/lo}{$-$0.7}
\defval{slopem/p140/768/p140-enc-fixed-minus-p140-enc/fineweb-edu-100bt}{$-$0.0}
\defval{slopem/p140/768/p140-enc-fixed-minus-p140-enc/fineweb-edu-100bt/hi}{+0.4}
\defval{slopem/p140/768/p140-enc-fixed-minus-p140-enc/fineweb-edu-100bt/lo}{$-$0.4}
\defval{slopem/p140/768/p140-enc-fixed-minus-p140-enc/govreport-test}{+0.1}
\defval{slopem/p140/768/p140-enc-fixed-minus-p140-enc/govreport-test/hi}{+0.3}
\defval{slopem/p140/768/p140-enc-fixed-minus-p140-enc/govreport-test/lo}{$-$0.2}
\defval{slopem/p140/768/p140-enc-fixed-minus-p140-enc/longdocs}{$-$0.3}
\defval{slopem/p140/768/p140-enc-fixed-minus-p140-enc/longdocs/hi}{$-$0.1}
\defval{slopem/p140/768/p140-enc-fixed-minus-p140-enc/longdocs/lo}{$-$0.5}
\defval{slopem/p140/768/p140-enc-fixed-minus-p140-enc/pg19-test}{$-$0.7}
\defval{slopem/p140/768/p140-enc-fixed-minus-p140-enc/pg19-test/hi}{$-$0.3}
\defval{slopem/p140/768/p140-enc-fixed-minus-p140-enc/pg19-test/lo}{$-$1.2}
\defval{slopem/p140/768/p140-enc-fixed-minus-p140-enc/pg19-val}{$-$0.1}
\defval{slopem/p140/768/p140-enc-fixed-minus-p140-enc/pg19-val/hi}{+0.7}
\defval{slopem/p140/768/p140-enc-fixed-minus-p140-enc/pg19-val/lo}{$-$0.8}
\defval{slopem/p140/768/p140-noloop-minus-p140-enc/finepdfs-long}{+5.1}
\defval{slopem/p140/768/p140-noloop-minus-p140-enc/finepdfs-long/hi}{+6.0}
\defval{slopem/p140/768/p140-noloop-minus-p140-enc/finepdfs-long/lo}{+4.2}
\defval{slopem/p140/768/p140-noloop-minus-p140-enc/fineweb-edu-100bt}{+1.6}
\defval{slopem/p140/768/p140-noloop-minus-p140-enc/fineweb-edu-100bt/hi}{+2.1}
\defval{slopem/p140/768/p140-noloop-minus-p140-enc/fineweb-edu-100bt/lo}{+1.0}
\defval{slopem/p140/768/p140-noloop-minus-p140-enc/govreport-test}{+2.6}
\defval{slopem/p140/768/p140-noloop-minus-p140-enc/govreport-test/hi}{+3.0}
\defval{slopem/p140/768/p140-noloop-minus-p140-enc/govreport-test/lo}{+2.2}
\defval{slopem/p140/768/p140-noloop-minus-p140-enc/longdocs}{+3.3}
\defval{slopem/p140/768/p140-noloop-minus-p140-enc/longdocs/hi}{+3.7}
\defval{slopem/p140/768/p140-noloop-minus-p140-enc/longdocs/lo}{+2.9}
\defval{slopem/p140/768/p140-noloop-minus-p140-enc/pg19-test}{+2.3}
\defval{slopem/p140/768/p140-noloop-minus-p140-enc/pg19-test/hi}{+2.9}
\defval{slopem/p140/768/p140-noloop-minus-p140-enc/pg19-test/lo}{+1.7}
\defval{slopem/p140/768/p140-noloop-minus-p140-enc/pg19-val}{+3.1}
\defval{slopem/p140/768/p140-noloop-minus-p140-enc/pg19-val/hi}{+4.0}
\defval{slopem/p140/768/p140-noloop-minus-p140-enc/pg19-val/lo}{+2.3}
\defval{slopem/x40/1024/ced-dec-minus-ced-enc/finepdfs-long}{+3.2}
\defval{slopem/x40/1024/ced-dec-minus-ced-enc/finepdfs-long/hi}{+3.9}
\defval{slopem/x40/1024/ced-dec-minus-ced-enc/finepdfs-long/lo}{+2.6}
\defval{slopem/x40/1024/ced-dec-minus-ced-enc/fineweb-edu-100bt}{+1.1}
\defval{slopem/x40/1024/ced-dec-minus-ced-enc/fineweb-edu-100bt/hi}{+1.6}
\defval{slopem/x40/1024/ced-dec-minus-ced-enc/fineweb-edu-100bt/lo}{+0.6}
\defval{slopem/x40/1024/ced-dec-minus-ced-enc/govreport-test}{+1.3}
\defval{slopem/x40/1024/ced-dec-minus-ced-enc/govreport-test/hi}{+1.7}
\defval{slopem/x40/1024/ced-dec-minus-ced-enc/govreport-test/lo}{+1.0}
\defval{slopem/x40/1024/ced-dec-minus-ced-enc/longdocs}{+1.8}
\defval{slopem/x40/1024/ced-dec-minus-ced-enc/longdocs/hi}{+2.1}
\defval{slopem/x40/1024/ced-dec-minus-ced-enc/longdocs/lo}{+1.5}
\defval{slopem/x40/1024/ced-dec-minus-ced-enc/pg19-test}{+0.8}
\defval{slopem/x40/1024/ced-dec-minus-ced-enc/pg19-test/hi}{+1.3}
\defval{slopem/x40/1024/ced-dec-minus-ced-enc/pg19-test/lo}{+0.4}
\defval{slopem/x40/1024/ced-dec-minus-ced-enc/pg19-val}{+0.1}
\defval{slopem/x40/1024/ced-dec-minus-ced-enc/pg19-val/hi}{+1.0}
\defval{slopem/x40/1024/ced-dec-minus-ced-enc/pg19-val/lo}{$-$0.8}
\defval{slopem/x40/1024/ced-dec-minus-dense-l25/finepdfs-long}{+2.4}
\defval{slopem/x40/1024/ced-dec-minus-dense-l25/finepdfs-long/hi}{+3.1}
\defval{slopem/x40/1024/ced-dec-minus-dense-l25/finepdfs-long/lo}{+1.8}
\defval{slopem/x40/1024/ced-dec-minus-dense-l25/fineweb-edu-100bt}{+1.6}
\defval{slopem/x40/1024/ced-dec-minus-dense-l25/fineweb-edu-100bt/hi}{+2.1}
\defval{slopem/x40/1024/ced-dec-minus-dense-l25/fineweb-edu-100bt/lo}{+1.1}
\defval{slopem/x40/1024/ced-dec-minus-dense-l25/govreport-test}{+2.3}
\defval{slopem/x40/1024/ced-dec-minus-dense-l25/govreport-test/hi}{+2.7}
\defval{slopem/x40/1024/ced-dec-minus-dense-l25/govreport-test/lo}{+2.0}
\defval{slopem/x40/1024/ced-dec-minus-dense-l25/longdocs}{+1.9}
\defval{slopem/x40/1024/ced-dec-minus-dense-l25/longdocs/hi}{+2.1}
\defval{slopem/x40/1024/ced-dec-minus-dense-l25/longdocs/lo}{+1.5}
\defval{slopem/x40/1024/ced-dec-minus-dense-l25/pg19-test}{+0.8}
\defval{slopem/x40/1024/ced-dec-minus-dense-l25/pg19-test/hi}{+1.3}
\defval{slopem/x40/1024/ced-dec-minus-dense-l25/pg19-test/lo}{+0.3}
\defval{slopem/x40/1024/ced-dec-minus-dense-l25/pg19-val}{+0.0}
\defval{slopem/x40/1024/ced-dec-minus-dense-l25/pg19-val/hi}{+1.0}
\defval{slopem/x40/1024/ced-dec-minus-dense-l25/pg19-val/lo}{$-$1.0}
\defval{slopem/x40/1024/ced-enc-fixed-minus-ced-enc/finepdfs-long}{$-$0.4}
\defval{slopem/x40/1024/ced-enc-fixed-minus-ced-enc/finepdfs-long/hi}{+0.1}
\defval{slopem/x40/1024/ced-enc-fixed-minus-ced-enc/finepdfs-long/lo}{$-$0.8}
\defval{slopem/x40/1024/ced-enc-fixed-minus-ced-enc/fineweb-edu-100bt}{+0.1}
\defval{slopem/x40/1024/ced-enc-fixed-minus-ced-enc/fineweb-edu-100bt/hi}{+0.5}
\defval{slopem/x40/1024/ced-enc-fixed-minus-ced-enc/fineweb-edu-100bt/lo}{$-$0.3}
\defval{slopem/x40/1024/ced-enc-fixed-minus-ced-enc/govreport-test}{+0.4}
\defval{slopem/x40/1024/ced-enc-fixed-minus-ced-enc/govreport-test/hi}{+0.6}
\defval{slopem/x40/1024/ced-enc-fixed-minus-ced-enc/govreport-test/lo}{+0.1}
\defval{slopem/x40/1024/ced-enc-fixed-minus-ced-enc/longdocs}{+0.1}
\defval{slopem/x40/1024/ced-enc-fixed-minus-ced-enc/longdocs/hi}{+0.3}
\defval{slopem/x40/1024/ced-enc-fixed-minus-ced-enc/longdocs/lo}{$-$0.1}
\defval{slopem/x40/1024/ced-enc-fixed-minus-ced-enc/pg19-test}{+0.4}
\defval{slopem/x40/1024/ced-enc-fixed-minus-ced-enc/pg19-test/hi}{+0.8}
\defval{slopem/x40/1024/ced-enc-fixed-minus-ced-enc/pg19-test/lo}{$-$0.0}
\defval{slopem/x40/1024/ced-enc-fixed-minus-ced-enc/pg19-val}{+0.3}
\defval{slopem/x40/1024/ced-enc-fixed-minus-ced-enc/pg19-val/hi}{+1.0}
\defval{slopem/x40/1024/ced-enc-fixed-minus-ced-enc/pg19-val/lo}{$-$0.5}
\defval{slopem/x40/1024/ced-enc-minus-dense-l25/finepdfs-long}{$-$0.8}
\defval{slopem/x40/1024/ced-enc-minus-dense-l25/finepdfs-long/hi}{$-$0.2}
\defval{slopem/x40/1024/ced-enc-minus-dense-l25/finepdfs-long/lo}{$-$1.3}
\defval{slopem/x40/1024/ced-enc-minus-dense-l25/fineweb-edu-100bt}{+0.5}
\defval{slopem/x40/1024/ced-enc-minus-dense-l25/fineweb-edu-100bt/hi}{+0.9}
\defval{slopem/x40/1024/ced-enc-minus-dense-l25/fineweb-edu-100bt/lo}{+0.1}
\defval{slopem/x40/1024/ced-enc-minus-dense-l25/govreport-test}{+1.0}
\defval{slopem/x40/1024/ced-enc-minus-dense-l25/govreport-test/hi}{+1.3}
\defval{slopem/x40/1024/ced-enc-minus-dense-l25/govreport-test/lo}{+0.7}
\defval{slopem/x40/1024/ced-enc-minus-dense-l25/longdocs}{+0.1}
\defval{slopem/x40/1024/ced-enc-minus-dense-l25/longdocs/hi}{+0.3}
\defval{slopem/x40/1024/ced-enc-minus-dense-l25/longdocs/lo}{$-$0.2}
\defval{slopem/x40/1024/ced-enc-minus-dense-l25/pg19-test}{$-$0.1}
\defval{slopem/x40/1024/ced-enc-minus-dense-l25/pg19-test/hi}{+0.4}
\defval{slopem/x40/1024/ced-enc-minus-dense-l25/pg19-test/lo}{$-$0.5}
\defval{slopem/x40/1024/ced-enc-minus-dense-l25/pg19-val}{$-$0.1}
\defval{slopem/x40/1024/ced-enc-minus-dense-l25/pg19-val/hi}{+0.8}
\defval{slopem/x40/1024/ced-enc-minus-dense-l25/pg19-val/lo}{$-$1.0}
\defval{slopem/x40/1024/ced-enc-minus-yocou-loop/finepdfs-long}{$-$4.9}
\defval{slopem/x40/1024/ced-enc-minus-yocou-loop/finepdfs-long/hi}{$-$4.4}
\defval{slopem/x40/1024/ced-enc-minus-yocou-loop/finepdfs-long/lo}{$-$5.6}
\defval{slopem/x40/1024/ced-enc-minus-yocou-loop/fineweb-edu-100bt}{$-$1.7}
\defval{slopem/x40/1024/ced-enc-minus-yocou-loop/fineweb-edu-100bt/hi}{$-$1.3}
\defval{slopem/x40/1024/ced-enc-minus-yocou-loop/fineweb-edu-100bt/lo}{$-$2.2}
\defval{slopem/x40/1024/ced-enc-minus-yocou-loop/govreport-test}{$-$2.6}
\defval{slopem/x40/1024/ced-enc-minus-yocou-loop/govreport-test/hi}{$-$2.3}
\defval{slopem/x40/1024/ced-enc-minus-yocou-loop/govreport-test/lo}{$-$2.9}
\defval{slopem/x40/1024/ced-enc-minus-yocou-loop/longdocs}{$-$3.7}
\defval{slopem/x40/1024/ced-enc-minus-yocou-loop/longdocs/hi}{$-$3.4}
\defval{slopem/x40/1024/ced-enc-minus-yocou-loop/longdocs/lo}{$-$3.9}
\defval{slopem/x40/1024/ced-enc-minus-yocou-loop/pg19-test}{$-$3.5}
\defval{slopem/x40/1024/ced-enc-minus-yocou-loop/pg19-test/hi}{$-$3.0}
\defval{slopem/x40/1024/ced-enc-minus-yocou-loop/pg19-test/lo}{$-$3.9}
\defval{slopem/x40/1024/ced-enc-minus-yocou-loop/pg19-val}{$-$2.5}
\defval{slopem/x40/1024/ced-enc-minus-yocou-loop/pg19-val/hi}{$-$1.6}
\defval{slopem/x40/1024/ced-enc-minus-yocou-loop/pg19-val/lo}{$-$3.3}
\defval{slopem/x40/1024/ced-noloop-minus-ced-dec/finepdfs-long}{$-$1.4}
\defval{slopem/x40/1024/ced-noloop-minus-ced-dec/finepdfs-long/hi}{$-$0.9}
\defval{slopem/x40/1024/ced-noloop-minus-ced-dec/finepdfs-long/lo}{$-$2.0}
\defval{slopem/x40/1024/ced-noloop-minus-ced-dec/fineweb-edu-100bt}{$-$0.2}
\defval{slopem/x40/1024/ced-noloop-minus-ced-dec/fineweb-edu-100bt/hi}{+0.3}
\defval{slopem/x40/1024/ced-noloop-minus-ced-dec/fineweb-edu-100bt/lo}{$-$0.7}
\defval{slopem/x40/1024/ced-noloop-minus-ced-dec/govreport-test}{+0.0}
\defval{slopem/x40/1024/ced-noloop-minus-ced-dec/govreport-test/hi}{+0.3}
\defval{slopem/x40/1024/ced-noloop-minus-ced-dec/govreport-test/lo}{$-$0.3}
\defval{slopem/x40/1024/ced-noloop-minus-ced-dec/longdocs}{$-$0.6}
\defval{slopem/x40/1024/ced-noloop-minus-ced-dec/longdocs/hi}{$-$0.3}
\defval{slopem/x40/1024/ced-noloop-minus-ced-dec/longdocs/lo}{$-$0.9}
\defval{slopem/x40/1024/ced-noloop-minus-ced-dec/pg19-test}{$-$0.4}
\defval{slopem/x40/1024/ced-noloop-minus-ced-dec/pg19-test/hi}{+0.1}
\defval{slopem/x40/1024/ced-noloop-minus-ced-dec/pg19-test/lo}{$-$0.8}
\defval{slopem/x40/1024/ced-noloop-minus-ced-dec/pg19-val}{+0.7}
\defval{slopem/x40/1024/ced-noloop-minus-ced-dec/pg19-val/hi}{+1.5}
\defval{slopem/x40/1024/ced-noloop-minus-ced-dec/pg19-val/lo}{$-$0.1}
\defval{slopem/x40/1024/ced-noloop-minus-ced-enc/finepdfs-long}{+1.7}
\defval{slopem/x40/1024/ced-noloop-minus-ced-enc/finepdfs-long/hi}{+2.5}
\defval{slopem/x40/1024/ced-noloop-minus-ced-enc/finepdfs-long/lo}{+1.0}
\defval{slopem/x40/1024/ced-noloop-minus-ced-enc/fineweb-edu-100bt}{+0.9}
\defval{slopem/x40/1024/ced-noloop-minus-ced-enc/fineweb-edu-100bt/hi}{+1.5}
\defval{slopem/x40/1024/ced-noloop-minus-ced-enc/fineweb-edu-100bt/lo}{+0.3}
\defval{slopem/x40/1024/ced-noloop-minus-ced-enc/govreport-test}{+1.3}
\defval{slopem/x40/1024/ced-noloop-minus-ced-enc/govreport-test/hi}{+1.7}
\defval{slopem/x40/1024/ced-noloop-minus-ced-enc/govreport-test/lo}{+1.0}
\defval{slopem/x40/1024/ced-noloop-minus-ced-enc/longdocs}{+1.2}
\defval{slopem/x40/1024/ced-noloop-minus-ced-enc/longdocs/hi}{+1.5}
\defval{slopem/x40/1024/ced-noloop-minus-ced-enc/longdocs/lo}{+0.9}
\defval{slopem/x40/1024/ced-noloop-minus-ced-enc/pg19-test}{+0.4}
\defval{slopem/x40/1024/ced-noloop-minus-ced-enc/pg19-test/hi}{+0.9}
\defval{slopem/x40/1024/ced-noloop-minus-ced-enc/pg19-test/lo}{$-$0.0}
\defval{slopem/x40/1024/ced-noloop-minus-ced-enc/pg19-val}{+0.7}
\defval{slopem/x40/1024/ced-noloop-minus-ced-enc/pg19-val/hi}{+1.5}
\defval{slopem/x40/1024/ced-noloop-minus-ced-enc/pg19-val/lo}{$-$0.0}
\defval{slopem/x40/1024/ced-noloop-minus-dense-noloop/finepdfs-long}{$-$0.7}
\defval{slopem/x40/1024/ced-noloop-minus-dense-noloop/finepdfs-long/hi}{$-$0.0}
\defval{slopem/x40/1024/ced-noloop-minus-dense-noloop/finepdfs-long/lo}{$-$1.4}
\defval{slopem/x40/1024/ced-noloop-minus-dense-noloop/fineweb-edu-100bt}{+0.4}
\defval{slopem/x40/1024/ced-noloop-minus-dense-noloop/fineweb-edu-100bt/hi}{+0.9}
\defval{slopem/x40/1024/ced-noloop-minus-dense-noloop/fineweb-edu-100bt/lo}{$-$0.1}
\defval{slopem/x40/1024/ced-noloop-minus-dense-noloop/govreport-test}{+1.1}
\defval{slopem/x40/1024/ced-noloop-minus-dense-noloop/govreport-test/hi}{+1.4}
\defval{slopem/x40/1024/ced-noloop-minus-dense-noloop/govreport-test/lo}{+0.8}
\defval{slopem/x40/1024/ced-noloop-minus-dense-noloop/longdocs}{$-$0.1}
\defval{slopem/x40/1024/ced-noloop-minus-dense-noloop/longdocs/hi}{+0.2}
\defval{slopem/x40/1024/ced-noloop-minus-dense-noloop/longdocs/lo}{$-$0.3}
\defval{slopem/x40/1024/ced-noloop-minus-dense-noloop/pg19-test}{$-$0.6}
\defval{slopem/x40/1024/ced-noloop-minus-dense-noloop/pg19-test/hi}{$-$0.0}
\defval{slopem/x40/1024/ced-noloop-minus-dense-noloop/pg19-test/lo}{$-$1.1}
\defval{slopem/x40/1024/ced-noloop-minus-dense-noloop/pg19-val}{$-$0.5}
\defval{slopem/x40/1024/ced-noloop-minus-dense-noloop/pg19-val/hi}{+0.4}
\defval{slopem/x40/1024/ced-noloop-minus-dense-noloop/pg19-val/lo}{$-$1.4}
\defval{slopem/x40/1024/ced-noloop-minus-yocou-noloop/finepdfs-long}{$-$3.2}
\defval{slopem/x40/1024/ced-noloop-minus-yocou-noloop/finepdfs-long/hi}{$-$2.7}
\defval{slopem/x40/1024/ced-noloop-minus-yocou-noloop/finepdfs-long/lo}{$-$3.8}
\defval{slopem/x40/1024/ced-noloop-minus-yocou-noloop/fineweb-edu-100bt}{$-$0.8}
\defval{slopem/x40/1024/ced-noloop-minus-yocou-noloop/fineweb-edu-100bt/hi}{$-$0.4}
\defval{slopem/x40/1024/ced-noloop-minus-yocou-noloop/fineweb-edu-100bt/lo}{$-$1.3}
\defval{slopem/x40/1024/ced-noloop-minus-yocou-noloop/govreport-test}{$-$0.8}
\defval{slopem/x40/1024/ced-noloop-minus-yocou-noloop/govreport-test/hi}{$-$0.5}
\defval{slopem/x40/1024/ced-noloop-minus-yocou-noloop/govreport-test/lo}{$-$1.1}
\defval{slopem/x40/1024/ced-noloop-minus-yocou-noloop/longdocs}{$-$2.4}
\defval{slopem/x40/1024/ced-noloop-minus-yocou-noloop/longdocs/hi}{$-$2.1}
\defval{slopem/x40/1024/ced-noloop-minus-yocou-noloop/longdocs/lo}{$-$2.7}
\defval{slopem/x40/1024/ced-noloop-minus-yocou-noloop/pg19-test}{$-$3.2}
\defval{slopem/x40/1024/ced-noloop-minus-yocou-noloop/pg19-test/hi}{$-$2.7}
\defval{slopem/x40/1024/ced-noloop-minus-yocou-noloop/pg19-test/lo}{$-$3.7}
\defval{slopem/x40/1024/ced-noloop-minus-yocou-noloop/pg19-val}{$-$2.6}
\defval{slopem/x40/1024/ced-noloop-minus-yocou-noloop/pg19-val/hi}{$-$1.8}
\defval{slopem/x40/1024/ced-noloop-minus-yocou-noloop/pg19-val/lo}{$-$3.4}
\defval{slopem/x40/1024/dense-l47-minus-dense-l25/finepdfs-long}{+0.7}
\defval{slopem/x40/1024/dense-l47-minus-dense-l25/finepdfs-long/hi}{+1.2}
\defval{slopem/x40/1024/dense-l47-minus-dense-l25/finepdfs-long/lo}{+0.3}
\defval{slopem/x40/1024/dense-l47-minus-dense-l25/fineweb-edu-100bt}{+0.3}
\defval{slopem/x40/1024/dense-l47-minus-dense-l25/fineweb-edu-100bt/hi}{+0.7}
\defval{slopem/x40/1024/dense-l47-minus-dense-l25/fineweb-edu-100bt/lo}{$-$0.1}
\defval{slopem/x40/1024/dense-l47-minus-dense-l25/govreport-test}{+0.6}
\defval{slopem/x40/1024/dense-l47-minus-dense-l25/govreport-test/hi}{+0.9}
\defval{slopem/x40/1024/dense-l47-minus-dense-l25/govreport-test/lo}{+0.4}
\defval{slopem/x40/1024/dense-l47-minus-dense-l25/longdocs}{+0.6}
\defval{slopem/x40/1024/dense-l47-minus-dense-l25/longdocs/hi}{+0.8}
\defval{slopem/x40/1024/dense-l47-minus-dense-l25/longdocs/lo}{+0.3}
\defval{slopem/x40/1024/dense-l47-minus-dense-l25/pg19-test}{+0.3}
\defval{slopem/x40/1024/dense-l47-minus-dense-l25/pg19-test/hi}{+0.8}
\defval{slopem/x40/1024/dense-l47-minus-dense-l25/pg19-test/lo}{$-$0.1}
\defval{slopem/x40/1024/dense-l47-minus-dense-l25/pg19-val}{$-$0.1}
\defval{slopem/x40/1024/dense-l47-minus-dense-l25/pg19-val/hi}{+0.6}
\defval{slopem/x40/1024/dense-l47-minus-dense-l25/pg19-val/lo}{$-$1.0}
\defval{slopem/x40/1024/dense-noloop-minus-dense-l25/finepdfs-long}{+1.7}
\defval{slopem/x40/1024/dense-noloop-minus-dense-l25/finepdfs-long/hi}{+2.4}
\defval{slopem/x40/1024/dense-noloop-minus-dense-l25/finepdfs-long/lo}{+1.1}
\defval{slopem/x40/1024/dense-noloop-minus-dense-l25/fineweb-edu-100bt}{+1.0}
\defval{slopem/x40/1024/dense-noloop-minus-dense-l25/fineweb-edu-100bt/hi}{+1.6}
\defval{slopem/x40/1024/dense-noloop-minus-dense-l25/fineweb-edu-100bt/lo}{+0.5}
\defval{slopem/x40/1024/dense-noloop-minus-dense-l25/govreport-test}{+1.2}
\defval{slopem/x40/1024/dense-noloop-minus-dense-l25/govreport-test/hi}{+1.6}
\defval{slopem/x40/1024/dense-noloop-minus-dense-l25/govreport-test/lo}{+0.9}
\defval{slopem/x40/1024/dense-noloop-minus-dense-l25/longdocs}{+1.3}
\defval{slopem/x40/1024/dense-noloop-minus-dense-l25/longdocs/hi}{+1.6}
\defval{slopem/x40/1024/dense-noloop-minus-dense-l25/longdocs/lo}{+1.0}
\defval{slopem/x40/1024/dense-noloop-minus-dense-l25/pg19-test}{+1.0}
\defval{slopem/x40/1024/dense-noloop-minus-dense-l25/pg19-test/hi}{+1.5}
\defval{slopem/x40/1024/dense-noloop-minus-dense-l25/pg19-test/lo}{+0.5}
\defval{slopem/x40/1024/dense-noloop-minus-dense-l25/pg19-val}{+1.2}
\defval{slopem/x40/1024/dense-noloop-minus-dense-l25/pg19-val/hi}{+2.1}
\defval{slopem/x40/1024/dense-noloop-minus-dense-l25/pg19-val/lo}{+0.3}
\defval{slopem/x40/1024/yocou-loop-minus-ced-enc/finepdfs-long}{+4.9}
\defval{slopem/x40/1024/yocou-loop-minus-ced-enc/finepdfs-long/hi}{+5.6}
\defval{slopem/x40/1024/yocou-loop-minus-ced-enc/finepdfs-long/lo}{+4.4}
\defval{slopem/x40/1024/yocou-loop-minus-ced-enc/fineweb-edu-100bt}{+1.7}
\defval{slopem/x40/1024/yocou-loop-minus-ced-enc/fineweb-edu-100bt/hi}{+2.2}
\defval{slopem/x40/1024/yocou-loop-minus-ced-enc/fineweb-edu-100bt/lo}{+1.3}
\defval{slopem/x40/1024/yocou-loop-minus-ced-enc/govreport-test}{+2.6}
\defval{slopem/x40/1024/yocou-loop-minus-ced-enc/govreport-test/hi}{+2.9}
\defval{slopem/x40/1024/yocou-loop-minus-ced-enc/govreport-test/lo}{+2.3}
\defval{slopem/x40/1024/yocou-loop-minus-ced-enc/longdocs}{+3.7}
\defval{slopem/x40/1024/yocou-loop-minus-ced-enc/longdocs/hi}{+3.9}
\defval{slopem/x40/1024/yocou-loop-minus-ced-enc/longdocs/lo}{+3.4}
\defval{slopem/x40/1024/yocou-loop-minus-ced-enc/pg19-test}{+3.5}
\defval{slopem/x40/1024/yocou-loop-minus-ced-enc/pg19-test/hi}{+3.9}
\defval{slopem/x40/1024/yocou-loop-minus-ced-enc/pg19-test/lo}{+3.0}
\defval{slopem/x40/1024/yocou-loop-minus-ced-enc/pg19-val}{+2.5}
\defval{slopem/x40/1024/yocou-loop-minus-ced-enc/pg19-val/hi}{+3.3}
\defval{slopem/x40/1024/yocou-loop-minus-ced-enc/pg19-val/lo}{+1.6}
\defval{slopem/x40/1024/yocou-loop-minus-dense-l25/finepdfs-long}{+4.2}
\defval{slopem/x40/1024/yocou-loop-minus-dense-l25/finepdfs-long/hi}{+4.9}
\defval{slopem/x40/1024/yocou-loop-minus-dense-l25/finepdfs-long/lo}{+3.5}
\defval{slopem/x40/1024/yocou-loop-minus-dense-l25/fineweb-edu-100bt}{+2.2}
\defval{slopem/x40/1024/yocou-loop-minus-dense-l25/fineweb-edu-100bt/hi}{+2.7}
\defval{slopem/x40/1024/yocou-loop-minus-dense-l25/fineweb-edu-100bt/lo}{+1.8}
\defval{slopem/x40/1024/yocou-loop-minus-dense-l25/govreport-test}{+3.6}
\defval{slopem/x40/1024/yocou-loop-minus-dense-l25/govreport-test/hi}{+3.9}
\defval{slopem/x40/1024/yocou-loop-minus-dense-l25/govreport-test/lo}{+3.3}
\defval{slopem/x40/1024/yocou-loop-minus-dense-l25/longdocs}{+3.7}
\defval{slopem/x40/1024/yocou-loop-minus-dense-l25/longdocs/hi}{+4.0}
\defval{slopem/x40/1024/yocou-loop-minus-dense-l25/longdocs/lo}{+3.4}
\defval{slopem/x40/1024/yocou-loop-minus-dense-l25/pg19-test}{+3.4}
\defval{slopem/x40/1024/yocou-loop-minus-dense-l25/pg19-test/hi}{+4.0}
\defval{slopem/x40/1024/yocou-loop-minus-dense-l25/pg19-test/lo}{+2.9}
\defval{slopem/x40/1024/yocou-loop-minus-dense-l25/pg19-val}{+2.4}
\defval{slopem/x40/1024/yocou-loop-minus-dense-l25/pg19-val/hi}{+3.2}
\defval{slopem/x40/1024/yocou-loop-minus-dense-l25/pg19-val/lo}{+1.6}
\defval{slopem/x40/1024/yocou-noloop-minus-ced-noloop/finepdfs-long}{+3.2}
\defval{slopem/x40/1024/yocou-noloop-minus-ced-noloop/finepdfs-long/hi}{+3.8}
\defval{slopem/x40/1024/yocou-noloop-minus-ced-noloop/finepdfs-long/lo}{+2.7}
\defval{slopem/x40/1024/yocou-noloop-minus-ced-noloop/fineweb-edu-100bt}{+0.8}
\defval{slopem/x40/1024/yocou-noloop-minus-ced-noloop/fineweb-edu-100bt/hi}{+1.3}
\defval{slopem/x40/1024/yocou-noloop-minus-ced-noloop/fineweb-edu-100bt/lo}{+0.4}
\defval{slopem/x40/1024/yocou-noloop-minus-ced-noloop/govreport-test}{+0.8}
\defval{slopem/x40/1024/yocou-noloop-minus-ced-noloop/govreport-test/hi}{+1.1}
\defval{slopem/x40/1024/yocou-noloop-minus-ced-noloop/govreport-test/lo}{+0.5}
\defval{slopem/x40/1024/yocou-noloop-minus-ced-noloop/longdocs}{+2.4}
\defval{slopem/x40/1024/yocou-noloop-minus-ced-noloop/longdocs/hi}{+2.7}
\defval{slopem/x40/1024/yocou-noloop-minus-ced-noloop/longdocs/lo}{+2.1}
\defval{slopem/x40/1024/yocou-noloop-minus-ced-noloop/pg19-test}{+3.2}
\defval{slopem/x40/1024/yocou-noloop-minus-ced-noloop/pg19-test/hi}{+3.7}
\defval{slopem/x40/1024/yocou-noloop-minus-ced-noloop/pg19-test/lo}{+2.7}
\defval{slopem/x40/1024/yocou-noloop-minus-ced-noloop/pg19-val}{+2.6}
\defval{slopem/x40/1024/yocou-noloop-minus-ced-noloop/pg19-val/hi}{+3.4}
\defval{slopem/x40/1024/yocou-noloop-minus-ced-noloop/pg19-val/lo}{+1.8}
\defval{slopem/x40/1024/yocou-noloop-minus-dense-noloop/finepdfs-long}{+2.5}
\defval{slopem/x40/1024/yocou-noloop-minus-dense-noloop/finepdfs-long/hi}{+3.2}
\defval{slopem/x40/1024/yocou-noloop-minus-dense-noloop/finepdfs-long/lo}{+1.9}
\defval{slopem/x40/1024/yocou-noloop-minus-dense-noloop/fineweb-edu-100bt}{+1.2}
\defval{slopem/x40/1024/yocou-noloop-minus-dense-noloop/fineweb-edu-100bt/hi}{+1.7}
\defval{slopem/x40/1024/yocou-noloop-minus-dense-noloop/fineweb-edu-100bt/lo}{+0.7}
\defval{slopem/x40/1024/yocou-noloop-minus-dense-noloop/govreport-test}{+1.9}
\defval{slopem/x40/1024/yocou-noloop-minus-dense-noloop/govreport-test/hi}{+2.2}
\defval{slopem/x40/1024/yocou-noloop-minus-dense-noloop/govreport-test/lo}{+1.6}
\defval{slopem/x40/1024/yocou-noloop-minus-dense-noloop/longdocs}{+2.4}
\defval{slopem/x40/1024/yocou-noloop-minus-dense-noloop/longdocs/hi}{+2.6}
\defval{slopem/x40/1024/yocou-noloop-minus-dense-noloop/longdocs/lo}{+2.1}
\defval{slopem/x40/1024/yocou-noloop-minus-dense-noloop/pg19-test}{+2.6}
\defval{slopem/x40/1024/yocou-noloop-minus-dense-noloop/pg19-test/hi}{+3.1}
\defval{slopem/x40/1024/yocou-noloop-minus-dense-noloop/pg19-test/lo}{+2.1}
\defval{slopem/x40/1024/yocou-noloop-minus-dense-noloop/pg19-val}{+2.1}
\defval{slopem/x40/1024/yocou-noloop-minus-dense-noloop/pg19-val/hi}{+3.0}
\defval{slopem/x40/1024/yocou-noloop-minus-dense-noloop/pg19-val/lo}{+1.2}
\defval{slopem/x40/1024/yocou-noloop-minus-yocou-loop/finepdfs-long}{+0.0}
\defval{slopem/x40/1024/yocou-noloop-minus-yocou-loop/finepdfs-long/hi}{+0.7}
\defval{slopem/x40/1024/yocou-noloop-minus-yocou-loop/finepdfs-long/lo}{$-$0.6}
\defval{slopem/x40/1024/yocou-noloop-minus-yocou-loop/fineweb-edu-100bt}{$-$0.0}
\defval{slopem/x40/1024/yocou-noloop-minus-yocou-loop/fineweb-edu-100bt/hi}{+0.5}
\defval{slopem/x40/1024/yocou-noloop-minus-yocou-loop/fineweb-edu-100bt/lo}{$-$0.6}
\defval{slopem/x40/1024/yocou-noloop-minus-yocou-loop/govreport-test}{$-$0.5}
\defval{slopem/x40/1024/yocou-noloop-minus-yocou-loop/govreport-test/hi}{$-$0.1}
\defval{slopem/x40/1024/yocou-noloop-minus-yocou-loop/govreport-test/lo}{$-$0.8}
\defval{slopem/x40/1024/yocou-noloop-minus-yocou-loop/longdocs}{$-$0.1}
\defval{slopem/x40/1024/yocou-noloop-minus-yocou-loop/longdocs/hi}{+0.2}
\defval{slopem/x40/1024/yocou-noloop-minus-yocou-loop/longdocs/lo}{$-$0.4}
\defval{slopem/x40/1024/yocou-noloop-minus-yocou-loop/pg19-test}{+0.2}
\defval{slopem/x40/1024/yocou-noloop-minus-yocou-loop/pg19-test/hi}{+0.7}
\defval{slopem/x40/1024/yocou-noloop-minus-yocou-loop/pg19-test/lo}{$-$0.3}
\defval{slopem/x40/1024/yocou-noloop-minus-yocou-loop/pg19-val}{+0.9}
\defval{slopem/x40/1024/yocou-noloop-minus-yocou-loop/pg19-val/hi}{+1.7}
\defval{slopem/x40/1024/yocou-noloop-minus-yocou-loop/pg19-val/lo}{+0.1}
\defval{spike/x40/1024/ced-enc-fixed/bpb8000}{0.9259}
\defval{spike/x40/1024/ced-enc-fixed/bpb9000}{0.9464}
\defval{spike/x40/1024/ced-enc-fixed/maxgrad}{19.3}
\defval{spike/x40/1024/ced-enc/bpb8000}{0.9235}
\defval{spike/x40/1024/ced-enc/bpb9000}{0.9192}
\defval{spike/x40/1024/ced-enc/maxgrad}{1.5}
\defval{split/long/1024/ced-dec/tp16_td16}{0.8221}
\defval{split/long/1024/ced-dec/tp16_td2}{0.8472}
\defval{split/long/1024/ced-dec/tp16_td4}{0.8255}
\defval{split/long/1024/ced-dec/tp16_td8}{0.8222}
\defval{split/long/1024/ced-dec/tp1_td16}{0.8226}
\defval{split/long/1024/ced-dec/tp1_td2}{0.8469}
\defval{split/long/1024/ced-dec/tp1_td4}{0.8260}
\defval{split/long/1024/ced-dec/tp1_td8}{0.8227}
\defval{split/long/1024/ced-dec/tp2_td16}{0.8222}
\defval{split/long/1024/ced-dec/tp2_td2}{0.8472}
\defval{split/long/1024/ced-dec/tp2_td4}{0.8256}
\defval{split/long/1024/ced-dec/tp2_td8}{0.8222}
\defval{split/long/1024/ced-dec/tp4_td16}{0.8221}
\defval{split/long/1024/ced-dec/tp4_td2}{0.8472}
\defval{split/long/1024/ced-dec/tp4_td4}{0.8255}
\defval{split/long/1024/ced-dec/tp4_td8}{0.8222}
\defval{split/long/1024/ced-dec/tp8_td16}{0.8221}
\defval{split/long/1024/ced-dec/tp8_td2}{0.8472}
\defval{split/long/1024/ced-dec/tp8_td4}{0.8255}
\defval{split/long/1024/ced-dec/tp8_td8}{0.8222}
\defval{split/long/1024/ced-enc/tp16_td16}{0.8042}
\defval{split/long/1024/ced-enc/tp16_td2}{0.8517}
\defval{split/long/1024/ced-enc/tp16_td4}{0.8118}
\defval{split/long/1024/ced-enc/tp16_td8}{0.8044}
\defval{split/long/1024/ced-enc/tp1_td16}{0.8178}
\defval{split/long/1024/ced-enc/tp1_td2}{0.8633}
\defval{split/long/1024/ced-enc/tp1_td4}{0.8251}
\defval{split/long/1024/ced-enc/tp1_td8}{0.8179}
\defval{split/long/1024/ced-enc/tp2_td16}{0.8075}
\defval{split/long/1024/ced-enc/tp2_td2}{0.8546}
\defval{split/long/1024/ced-enc/tp2_td4}{0.8150}
\defval{split/long/1024/ced-enc/tp2_td8}{0.8077}
\defval{split/long/1024/ced-enc/tp4_td16}{0.8046}
\defval{split/long/1024/ced-enc/tp4_td2}{0.8517}
\defval{split/long/1024/ced-enc/tp4_td4}{0.8121}
\defval{split/long/1024/ced-enc/tp4_td8}{0.8047}
\defval{split/long/1024/ced-enc/tp8_td16}{0.8042}
\defval{split/long/1024/ced-enc/tp8_td2}{0.8517}
\defval{split/long/1024/ced-enc/tp8_td4}{0.8118}
\defval{split/long/1024/ced-enc/tp8_td8}{0.8044}
\defval{split/long/1024/dense-l47/tp16_td16}{0.8090}
\defval{split/long/1024/dense-l47/tp16_td2}{0.9470}
\defval{split/long/1024/dense-l47/tp16_td4}{0.8295}
\defval{split/long/1024/dense-l47/tp16_td8}{0.8096}
\defval{split/long/1024/dense-l47/tp1_td16}{0.8443}
\defval{split/long/1024/dense-l47/tp1_td2}{0.8790}
\defval{split/long/1024/dense-l47/tp1_td4}{0.8533}
\defval{split/long/1024/dense-l47/tp1_td8}{0.8448}
\defval{split/long/1024/dense-l47/tp2_td16}{0.8109}
\defval{split/long/1024/dense-l47/tp2_td2}{0.9563}
\defval{split/long/1024/dense-l47/tp2_td4}{0.8323}
\defval{split/long/1024/dense-l47/tp2_td8}{0.8113}
\defval{split/long/1024/dense-l47/tp4_td16}{0.8092}
\defval{split/long/1024/dense-l47/tp4_td2}{0.9551}
\defval{split/long/1024/dense-l47/tp4_td4}{0.8309}
\defval{split/long/1024/dense-l47/tp4_td8}{0.8098}
\defval{split/long/1024/dense-l47/tp8_td16}{0.8090}
\defval{split/long/1024/dense-l47/tp8_td2}{0.9468}
\defval{split/long/1024/dense-l47/tp8_td4}{0.8295}
\defval{split/long/1024/dense-l47/tp8_td8}{0.8096}
\defval{split/x40/1024/ced-dec/tp16_td16}{0.8574}
\defval{split/x40/1024/ced-dec/tp16_td2}{0.8721}
\defval{split/x40/1024/ced-dec/tp16_td4}{0.8592}
\defval{split/x40/1024/ced-dec/tp16_td8}{0.8573}
\defval{split/x40/1024/ced-dec/tp1_td16}{0.8579}
\defval{split/x40/1024/ced-dec/tp1_td2}{0.8721}
\defval{split/x40/1024/ced-dec/tp1_td4}{0.8595}
\defval{split/x40/1024/ced-dec/tp1_td8}{0.8579}
\defval{split/x40/1024/ced-dec/tp2_td16}{0.8574}
\defval{split/x40/1024/ced-dec/tp2_td2}{0.8721}
\defval{split/x40/1024/ced-dec/tp2_td4}{0.8592}
\defval{split/x40/1024/ced-dec/tp2_td8}{0.8574}
\defval{split/x40/1024/ced-dec/tp4_td16}{0.8574}
\defval{split/x40/1024/ced-dec/tp4_td2}{0.8721}
\defval{split/x40/1024/ced-dec/tp4_td4}{0.8592}
\defval{split/x40/1024/ced-dec/tp4_td8}{0.8573}
\defval{split/x40/1024/ced-dec/tp8_td16}{0.8574}
\defval{split/x40/1024/ced-dec/tp8_td2}{0.8721}
\defval{split/x40/1024/ced-dec/tp8_td4}{0.8592}
\defval{split/x40/1024/ced-dec/tp8_td8}{0.8573}
\defval{split/x40/1024/ced-enc-fixed/tp16_td16}{0.8494}
\defval{split/x40/1024/ced-enc-fixed/tp16_td2}{1.6037}
\defval{split/x40/1024/ced-enc-fixed/tp16_td4}{1.0181}
\defval{split/x40/1024/ced-enc-fixed/tp16_td8}{0.8476}
\defval{split/x40/1024/ced-enc-fixed/tp1_td16}{1.2490}
\defval{split/x40/1024/ced-enc-fixed/tp1_td2}{1.7169}
\defval{split/x40/1024/ced-enc-fixed/tp1_td4}{1.3735}
\defval{split/x40/1024/ced-enc-fixed/tp1_td8}{1.2498}
\defval{split/x40/1024/ced-enc-fixed/tp2_td16}{0.9152}
\defval{split/x40/1024/ced-enc-fixed/tp2_td2}{1.7009}
\defval{split/x40/1024/ced-enc-fixed/tp2_td4}{1.1254}
\defval{split/x40/1024/ced-enc-fixed/tp2_td8}{0.9130}
\defval{split/x40/1024/ced-enc-fixed/tp4_td16}{0.8586}
\defval{split/x40/1024/ced-enc-fixed/tp4_td2}{1.5959}
\defval{split/x40/1024/ced-enc-fixed/tp4_td4}{1.0380}
\defval{split/x40/1024/ced-enc-fixed/tp4_td8}{0.8559}
\defval{split/x40/1024/ced-enc-fixed/tp8_td16}{0.8493}
\defval{split/x40/1024/ced-enc-fixed/tp8_td2}{1.6046}
\defval{split/x40/1024/ced-enc-fixed/tp8_td4}{1.0188}
\defval{split/x40/1024/ced-enc-fixed/tp8_td8}{0.8475}
\defval{split/x40/1024/ced-enc/tp16_td16}{0.8384}
\defval{split/x40/1024/ced-enc/tp16_td2}{0.8835}
\defval{split/x40/1024/ced-enc/tp16_td4}{0.8460}
\defval{split/x40/1024/ced-enc/tp16_td8}{0.8387}
\defval{split/x40/1024/ced-enc/tp1_td16}{0.8519}
\defval{split/x40/1024/ced-enc/tp1_td2}{0.8898}
\defval{split/x40/1024/ced-enc/tp1_td4}{0.8579}
\defval{split/x40/1024/ced-enc/tp1_td8}{0.8521}
\defval{split/x40/1024/ced-enc/tp2_td16}{0.8408}
\defval{split/x40/1024/ced-enc/tp2_td2}{0.8880}
\defval{split/x40/1024/ced-enc/tp2_td4}{0.8486}
\defval{split/x40/1024/ced-enc/tp2_td8}{0.8411}
\defval{split/x40/1024/ced-enc/tp4_td16}{0.8388}
\defval{split/x40/1024/ced-enc/tp4_td2}{0.8840}
\defval{split/x40/1024/ced-enc/tp4_td4}{0.8463}
\defval{split/x40/1024/ced-enc/tp4_td8}{0.8390}
\defval{split/x40/1024/ced-enc/tp8_td16}{0.8384}
\defval{split/x40/1024/ced-enc/tp8_td2}{0.8835}
\defval{split/x40/1024/ced-enc/tp8_td4}{0.8460}
\defval{split/x40/1024/ced-enc/tp8_td8}{0.8387}
\defval{split/x40/1024/dense-l25/tp16_td16}{0.8372}
\defval{split/x40/1024/dense-l25/tp16_td2}{0.9343}
\defval{split/x40/1024/dense-l25/tp16_td4}{0.8534}
\defval{split/x40/1024/dense-l25/tp16_td8}{0.8378}
\defval{split/x40/1024/dense-l25/tp1_td16}{0.8546}
\defval{split/x40/1024/dense-l25/tp1_td2}{0.8876}
\defval{split/x40/1024/dense-l25/tp1_td4}{0.8589}
\defval{split/x40/1024/dense-l25/tp1_td8}{0.8546}
\defval{split/x40/1024/dense-l25/tp2_td16}{0.8393}
\defval{split/x40/1024/dense-l25/tp2_td2}{0.9537}
\defval{split/x40/1024/dense-l25/tp2_td4}{0.8581}
\defval{split/x40/1024/dense-l25/tp2_td8}{0.8398}
\defval{split/x40/1024/dense-l25/tp4_td16}{0.8374}
\defval{split/x40/1024/dense-l25/tp4_td2}{0.9389}
\defval{split/x40/1024/dense-l25/tp4_td4}{0.8543}
\defval{split/x40/1024/dense-l25/tp4_td8}{0.8380}
\defval{split/x40/1024/dense-l25/tp8_td16}{0.8372}
\defval{split/x40/1024/dense-l25/tp8_td2}{0.9343}
\defval{split/x40/1024/dense-l25/tp8_td4}{0.8534}
\defval{split/x40/1024/dense-l25/tp8_td8}{0.8378}
\defval{split/x40/1024/dense-l47/tp16_td16}{0.8427}
\defval{split/x40/1024/dense-l47/tp16_td2}{0.9344}
\defval{split/x40/1024/dense-l47/tp16_td4}{0.8604}
\defval{split/x40/1024/dense-l47/tp16_td8}{0.8435}
\defval{split/x40/1024/dense-l47/tp1_td16}{0.8644}
\defval{split/x40/1024/dense-l47/tp1_td2}{0.8889}
\defval{split/x40/1024/dense-l47/tp1_td4}{0.8693}
\defval{split/x40/1024/dense-l47/tp1_td8}{0.8647}
\defval{split/x40/1024/dense-l47/tp2_td16}{0.8442}
\defval{split/x40/1024/dense-l47/tp2_td2}{0.9514}
\defval{split/x40/1024/dense-l47/tp2_td4}{0.8648}
\defval{split/x40/1024/dense-l47/tp2_td8}{0.8451}
\defval{split/x40/1024/dense-l47/tp4_td16}{0.8429}
\defval{split/x40/1024/dense-l47/tp4_td2}{0.9392}
\defval{split/x40/1024/dense-l47/tp4_td4}{0.8614}
\defval{split/x40/1024/dense-l47/tp4_td8}{0.8437}
\defval{split/x40/1024/dense-l47/tp8_td16}{0.8427}
\defval{split/x40/1024/dense-l47/tp8_td2}{0.9344}
\defval{split/x40/1024/dense-l47/tp8_td4}{0.8604}
\defval{split/x40/1024/dense-l47/tp8_td8}{0.8435}
\defval{split/x40/1024/yocou-loop/tp16_td16}{0.8431}
\defval{split/x40/1024/yocou-loop/tp16_td2}{0.8834}
\defval{split/x40/1024/yocou-loop/tp16_td4}{0.8499}
\defval{split/x40/1024/yocou-loop/tp16_td8}{0.8434}
\defval{split/x40/1024/yocou-loop/tp1_td16}{0.8510}
\defval{split/x40/1024/yocou-loop/tp1_td2}{0.8849}
\defval{split/x40/1024/yocou-loop/tp1_td4}{0.8558}
\defval{split/x40/1024/yocou-loop/tp1_td8}{0.8510}
\defval{split/x40/1024/yocou-loop/tp2_td16}{0.8444}
\defval{split/x40/1024/yocou-loop/tp2_td2}{0.8859}
\defval{split/x40/1024/yocou-loop/tp2_td4}{0.8514}
\defval{split/x40/1024/yocou-loop/tp2_td8}{0.8447}
\defval{split/x40/1024/yocou-loop/tp4_td16}{0.8433}
\defval{split/x40/1024/yocou-loop/tp4_td2}{0.8840}
\defval{split/x40/1024/yocou-loop/tp4_td4}{0.8502}
\defval{split/x40/1024/yocou-loop/tp4_td8}{0.8436}
\defval{split/x40/1024/yocou-loop/tp8_td16}{0.8431}
\defval{split/x40/1024/yocou-loop/tp8_td2}{0.8834}
\defval{split/x40/1024/yocou-loop/tp8_td4}{0.8499}
\defval{split/x40/1024/yocou-loop/tp8_td8}{0.8434}
\defval{split/x40/1536/ced-dec/tp16_td16}{0.8058}
\defval{split/x40/1536/ced-dec/tp16_td2}{0.8270}
\defval{split/x40/1536/ced-dec/tp16_td4}{0.8089}
\defval{split/x40/1536/ced-dec/tp16_td8}{0.8059}
\defval{split/x40/1536/ced-dec/tp1_td16}{0.8073}
\defval{split/x40/1536/ced-dec/tp1_td2}{0.8271}
\defval{split/x40/1536/ced-dec/tp1_td4}{0.8100}
\defval{split/x40/1536/ced-dec/tp1_td8}{0.8073}
\defval{split/x40/1536/ced-dec/tp2_td16}{0.8058}
\defval{split/x40/1536/ced-dec/tp2_td2}{0.8270}
\defval{split/x40/1536/ced-dec/tp2_td4}{0.8090}
\defval{split/x40/1536/ced-dec/tp2_td8}{0.8059}
\defval{split/x40/1536/ced-dec/tp4_td16}{0.8058}
\defval{split/x40/1536/ced-dec/tp4_td2}{0.8270}
\defval{split/x40/1536/ced-dec/tp4_td4}{0.8089}
\defval{split/x40/1536/ced-dec/tp4_td8}{0.8059}
\defval{split/x40/1536/ced-dec/tp8_td16}{0.8058}
\defval{split/x40/1536/ced-dec/tp8_td2}{0.8270}
\defval{split/x40/1536/ced-dec/tp8_td4}{0.8089}
\defval{split/x40/1536/ced-dec/tp8_td8}{0.8059}
\defval{split/x40/1536/ced-enc/tp16_td16}{0.7881}
\defval{split/x40/1536/ced-enc/tp16_td2}{0.8433}
\defval{split/x40/1536/ced-enc/tp16_td4}{0.7972}
\defval{split/x40/1536/ced-enc/tp16_td8}{0.7884}
\defval{split/x40/1536/ced-enc/tp1_td16}{0.8061}
\defval{split/x40/1536/ced-enc/tp1_td2}{0.8481}
\defval{split/x40/1536/ced-enc/tp1_td4}{0.8125}
\defval{split/x40/1536/ced-enc/tp1_td8}{0.8059}
\defval{split/x40/1536/ced-enc/tp2_td16}{0.7911}
\defval{split/x40/1536/ced-enc/tp2_td2}{0.8456}
\defval{split/x40/1536/ced-enc/tp2_td4}{0.7999}
\defval{split/x40/1536/ced-enc/tp2_td8}{0.7914}
\defval{split/x40/1536/ced-enc/tp4_td16}{0.7884}
\defval{split/x40/1536/ced-enc/tp4_td2}{0.8438}
\defval{split/x40/1536/ced-enc/tp4_td4}{0.7975}
\defval{split/x40/1536/ced-enc/tp4_td8}{0.7887}
\defval{split/x40/1536/ced-enc/tp8_td16}{0.7881}
\defval{split/x40/1536/ced-enc/tp8_td2}{0.8434}
\defval{split/x40/1536/ced-enc/tp8_td4}{0.7972}
\defval{split/x40/1536/ced-enc/tp8_td8}{0.7884}
\defval{split/x40/1536/dense-l25/tp16_td16}{0.7871}
\defval{split/x40/1536/dense-l25/tp16_td2}{1.0620}
\defval{split/x40/1536/dense-l25/tp16_td4}{0.8251}
\defval{split/x40/1536/dense-l25/tp16_td8}{0.7882}
\defval{split/x40/1536/dense-l25/tp1_td16}{0.9208}
\defval{split/x40/1536/dense-l25/tp1_td2}{0.8692}
\defval{split/x40/1536/dense-l25/tp1_td4}{0.8804}
\defval{split/x40/1536/dense-l25/tp1_td8}{0.9113}
\defval{split/x40/1536/dense-l25/tp2_td16}{0.7896}
\defval{split/x40/1536/dense-l25/tp2_td2}{1.0743}
\defval{split/x40/1536/dense-l25/tp2_td4}{0.8274}
\defval{split/x40/1536/dense-l25/tp2_td8}{0.7905}
\defval{split/x40/1536/dense-l25/tp4_td16}{0.7874}
\defval{split/x40/1536/dense-l25/tp4_td2}{1.0720}
\defval{split/x40/1536/dense-l25/tp4_td4}{0.8265}
\defval{split/x40/1536/dense-l25/tp4_td8}{0.7885}
\defval{split/x40/1536/dense-l25/tp8_td16}{0.7872}
\defval{split/x40/1536/dense-l25/tp8_td2}{1.0621}
\defval{split/x40/1536/dense-l25/tp8_td4}{0.8252}
\defval{split/x40/1536/dense-l25/tp8_td8}{0.7882}
\defval{split/x40/768/ced-dec/tp16_td16}{0.9040}
\defval{split/x40/768/ced-dec/tp16_td2}{0.9143}
\defval{split/x40/768/ced-dec/tp16_td4}{0.9053}
\defval{split/x40/768/ced-dec/tp16_td8}{0.9039}
\defval{split/x40/768/ced-dec/tp1_td16}{0.9043}
\defval{split/x40/768/ced-dec/tp1_td2}{0.9142}
\defval{split/x40/768/ced-dec/tp1_td4}{0.9055}
\defval{split/x40/768/ced-dec/tp1_td8}{0.9042}
\defval{split/x40/768/ced-dec/tp2_td16}{0.9040}
\defval{split/x40/768/ced-dec/tp2_td2}{0.9142}
\defval{split/x40/768/ced-dec/tp2_td4}{0.9053}
\defval{split/x40/768/ced-dec/tp2_td8}{0.9040}
\defval{split/x40/768/ced-dec/tp4_td16}{0.9040}
\defval{split/x40/768/ced-dec/tp4_td2}{0.9143}
\defval{split/x40/768/ced-dec/tp4_td4}{0.9053}
\defval{split/x40/768/ced-dec/tp4_td8}{0.9039}
\defval{split/x40/768/ced-dec/tp8_td16}{0.9040}
\defval{split/x40/768/ced-dec/tp8_td2}{0.9143}
\defval{split/x40/768/ced-dec/tp8_td4}{0.9053}
\defval{split/x40/768/ced-dec/tp8_td8}{0.9039}
\defval{split/x40/768/ced-enc/tp16_td16}{0.8837}
\defval{split/x40/768/ced-enc/tp16_td2}{0.9216}
\defval{split/x40/768/ced-enc/tp16_td4}{0.8903}
\defval{split/x40/768/ced-enc/tp16_td8}{0.8839}
\defval{split/x40/768/ced-enc/tp1_td16}{0.8945}
\defval{split/x40/768/ced-enc/tp1_td2}{0.9279}
\defval{split/x40/768/ced-enc/tp1_td4}{0.9000}
\defval{split/x40/768/ced-enc/tp1_td8}{0.8946}
\defval{split/x40/768/ced-enc/tp2_td16}{0.8858}
\defval{split/x40/768/ced-enc/tp2_td2}{0.9240}
\defval{split/x40/768/ced-enc/tp2_td4}{0.8924}
\defval{split/x40/768/ced-enc/tp2_td8}{0.8860}
\defval{split/x40/768/ced-enc/tp4_td16}{0.8839}
\defval{split/x40/768/ced-enc/tp4_td2}{0.9218}
\defval{split/x40/768/ced-enc/tp4_td4}{0.8905}
\defval{split/x40/768/ced-enc/tp4_td8}{0.8842}
\defval{split/x40/768/ced-enc/tp8_td16}{0.8837}
\defval{split/x40/768/ced-enc/tp8_td2}{0.9216}
\defval{split/x40/768/ced-enc/tp8_td4}{0.8903}
\defval{split/x40/768/ced-enc/tp8_td8}{0.8839}
\defval{split/x40/768/dense-l25/tp16_td16}{0.8834}
\defval{split/x40/768/dense-l25/tp16_td2}{0.9955}
\defval{split/x40/768/dense-l25/tp16_td4}{0.9024}
\defval{split/x40/768/dense-l25/tp16_td8}{0.8842}
\defval{split/x40/768/dense-l25/tp1_td16}{0.8958}
\defval{split/x40/768/dense-l25/tp1_td2}{0.9254}
\defval{split/x40/768/dense-l25/tp1_td4}{0.8999}
\defval{split/x40/768/dense-l25/tp1_td8}{0.8959}
\defval{split/x40/768/dense-l25/tp2_td16}{0.8852}
\defval{split/x40/768/dense-l25/tp2_td2}{1.0147}
\defval{split/x40/768/dense-l25/tp2_td4}{0.9066}
\defval{split/x40/768/dense-l25/tp2_td8}{0.8860}
\defval{split/x40/768/dense-l25/tp4_td16}{0.8836}
\defval{split/x40/768/dense-l25/tp4_td2}{0.9967}
\defval{split/x40/768/dense-l25/tp4_td4}{0.9027}
\defval{split/x40/768/dense-l25/tp4_td8}{0.8844}
\defval{split/x40/768/dense-l25/tp8_td16}{0.8835}
\defval{split/x40/768/dense-l25/tp8_td2}{0.9956}
\defval{split/x40/768/dense-l25/tp8_td4}{0.9025}
\defval{split/x40/768/dense-l25/tp8_td8}{0.8842}
\defval{split/x40/768/dense-l47/tp16_td16}{0.8899}
\defval{split/x40/768/dense-l47/tp16_td2}{0.9255}
\defval{split/x40/768/dense-l47/tp16_td4}{0.8965}
\defval{split/x40/768/dense-l47/tp16_td8}{0.8902}
\defval{split/x40/768/dense-l47/tp1_td16}{0.9087}
\defval{split/x40/768/dense-l47/tp1_td2}{0.9249}
\defval{split/x40/768/dense-l47/tp1_td4}{0.9080}
\defval{split/x40/768/dense-l47/tp1_td8}{0.9081}
\defval{split/x40/768/dense-l47/tp2_td16}{0.8911}
\defval{split/x40/768/dense-l47/tp2_td2}{0.9272}
\defval{split/x40/768/dense-l47/tp2_td4}{0.8978}
\defval{split/x40/768/dense-l47/tp2_td8}{0.8914}
\defval{split/x40/768/dense-l47/tp4_td16}{0.8901}
\defval{split/x40/768/dense-l47/tp4_td2}{0.9255}
\defval{split/x40/768/dense-l47/tp4_td4}{0.8965}
\defval{split/x40/768/dense-l47/tp4_td8}{0.8903}
\defval{split/x40/768/dense-l47/tp8_td16}{0.8899}
\defval{split/x40/768/dense-l47/tp8_td2}{0.9255}
\defval{split/x40/768/dense-l47/tp8_td4}{0.8965}
\defval{split/x40/768/dense-l47/tp8_td8}{0.8902}
\defval{split/x40/768/yocou-loop/tp16_td16}{0.8879}
\defval{split/x40/768/yocou-loop/tp16_td2}{0.9250}
\defval{split/x40/768/yocou-loop/tp16_td4}{0.8947}
\defval{split/x40/768/yocou-loop/tp16_td8}{0.8882}
\defval{split/x40/768/yocou-loop/tp1_td16}{0.8929}
\defval{split/x40/768/yocou-loop/tp1_td2}{0.9247}
\defval{split/x40/768/yocou-loop/tp1_td4}{0.8985}
\defval{split/x40/768/yocou-loop/tp1_td8}{0.8931}
\defval{split/x40/768/yocou-loop/tp2_td16}{0.8891}
\defval{split/x40/768/yocou-loop/tp2_td2}{0.9283}
\defval{split/x40/768/yocou-loop/tp2_td4}{0.8963}
\defval{split/x40/768/yocou-loop/tp2_td8}{0.8894}
\defval{split/x40/768/yocou-loop/tp4_td16}{0.8881}
\defval{split/x40/768/yocou-loop/tp4_td2}{0.9255}
\defval{split/x40/768/yocou-loop/tp4_td4}{0.8950}
\defval{split/x40/768/yocou-loop/tp4_td8}{0.8884}
\defval{split/x40/768/yocou-loop/tp8_td16}{0.8879}
\defval{split/x40/768/yocou-loop/tp8_td2}{0.9251}
\defval{split/x40/768/yocou-loop/tp8_td4}{0.8947}
\defval{split/x40/768/yocou-loop/tp8_td8}{0.8882}
\defval{tokens/long/1024/ced-dec}{29.6}
\defval{tokens/long/1024/ced-enc}{29.6}
\defval{tokens/long/1024/ced-noloop}{29.6}
\defval{tokens/long/1024/dense-l47}{29.6}
\defval{tokens/long/1024/dense-noloop}{29.6}
\defval{tokens/p140/768/p140-dec-fixed}{22.4}
\defval{tokens/p140/768/p140-enc}{22.4}
\defval{tokens/p140/768/p140-enc-fixed}{22.4}
\defval{tokens/p140/768/p140-noloop}{22.4}
\defval{tokens/x40/1024/ced-dec}{6.0}
\defval{tokens/x40/1024/ced-enc}{6.0}
\defval{tokens/x40/1024/ced-enc-fixed}{6.0}
\defval{tokens/x40/1024/ced-noloop}{6.0}
\defval{tokens/x40/1024/dense-l25}{6.0}
\defval{tokens/x40/1024/dense-l47}{6.0}
\defval{tokens/x40/1024/dense-noloop}{6.0}
\defval{tokens/x40/1024/yocou-loop}{6.0}
\defval{tokens/x40/1024/yocou-noloop}{6.0}
\defval{tokens/x40/1536/ced-dec}{13.6}
\defval{tokens/x40/1536/ced-enc}{13.6}
\defval{tokens/x40/1536/ced-noloop}{13.6}
\defval{tokens/x40/1536/dense-l25}{13.6}
\defval{tokens/x40/1536/dense-noloop}{13.6}
\defval{tokens/x40/768/ced-dec}{3.4}
\defval{tokens/x40/768/ced-enc}{3.4}
\defval{tokens/x40/768/ced-noloop}{3.4}
\defval{tokens/x40/768/dense-l25}{3.4}
\defval{tokens/x40/768/dense-l47}{3.4}
\defval{tokens/x40/768/dense-noloop}{3.4}
\defval{tokens/x40/768/yocou-loop}{3.4}
\defval{tokens/x40/768/yocou-noloop}{3.4}
\defval{tps/long/1024/ced-dec}{0.64}
\defval{tps/long/1024/ced-enc}{0.62}
\defval{tps/long/1024/ced-noloop}{1.38}
\defval{tps/long/1024/dense-l47}{0.78}
\defval{tps/long/1024/dense-noloop}{1.36}
\defval{tps/p140/768/p140-dec-fixed}{1.81}
\defval{tps/p140/768/p140-enc}{1.38}
\defval{tps/p140/768/p140-enc-fixed}{1.76}
\defval{tps/p140/768/p140-noloop}{2.56}
\defval{tps/x40/1024/ced-dec}{0.73}
\defval{tps/x40/1024/ced-enc}{0.72}
\defval{tps/x40/1024/ced-enc-fixed}{0.79}
\defval{tps/x40/1024/ced-noloop}{1.38}
\defval{tps/x40/1024/dense-l25}{0.78}
\defval{tps/x40/1024/dense-l47}{0.78}
\defval{tps/x40/1024/dense-noloop}{1.46}
\defval{tps/x40/1024/yocou-loop}{0.81}
\defval{tps/x40/1024/yocou-noloop}{1.43}
\defval{tps/x40/1536/ced-dec}{0.43}
\defval{tps/x40/1536/ced-enc}{0.42}
\defval{tps/x40/1536/ced-noloop}{0.83}
\defval{tps/x40/1536/dense-l25}{0.45}
\defval{tps/x40/1536/dense-noloop}{0.87}
\defval{tps/x40/768/ced-dec}{0.97}
\defval{tps/x40/768/ced-enc}{0.95}
\defval{tps/x40/768/ced-noloop}{1.77}
\defval{tps/x40/768/dense-l25}{1.04}
\defval{tps/x40/768/dense-l47}{1.04}
\defval{tps/x40/768/dense-noloop}{1.86}
\defval{tps/x40/768/yocou-loop}{1.08}
\defval{tps/x40/768/yocou-noloop}{1.83}
\defval{trunc/x40/1024/ced-enc/L1024}{0.8595}
\defval{trunc/x40/1024/ced-enc/L128}{0.9616}
\defval{trunc/x40/1024/ced-enc/L2048}{0.8465}
\defval{trunc/x40/1024/ced-enc/L256}{0.9133}
\defval{trunc/x40/1024/ced-enc/L512}{0.8807}
\defval{trunc/x40/1024/ced-noloop/L1024}{0.8842}
\defval{trunc/x40/1024/ced-noloop/L128}{0.9831}
\defval{trunc/x40/1024/ced-noloop/L2048}{0.8716}
\defval{trunc/x40/1024/ced-noloop/L256}{0.9362}
\defval{trunc/x40/1024/ced-noloop/L512}{0.9046}
\defval{trunc/x40/1024/dense-l25/L1024}{0.8593}
\defval{trunc/x40/1024/dense-l25/L128}{0.9635}
\defval{trunc/x40/1024/dense-l25/L2048}{0.8457}
\defval{trunc/x40/1024/dense-l25/L256}{0.9143}
\defval{trunc/x40/1024/dense-l25/L512}{0.8809}
\defval{trunc/x40/1024/dense-noloop/L1024}{0.8877}
\defval{trunc/x40/1024/dense-noloop/L128}{0.9888}
\defval{trunc/x40/1024/dense-noloop/L2048}{0.8746}
\defval{trunc/x40/1024/dense-noloop/L256}{0.9412}
\defval{trunc/x40/1024/dense-noloop/L512}{0.9088}
\defval{trunc/x40/1024/yocou-loop/L1024}{0.8625}
\defval{trunc/x40/1024/yocou-loop/L128}{0.9618}
\defval{trunc/x40/1024/yocou-loop/L2048}{0.8503}
\defval{trunc/x40/1024/yocou-loop/L256}{0.9139}
\defval{trunc/x40/1024/yocou-loop/L512}{0.8824}
\defval{trunc/x40/1024/yocou-noloop/L1024}{0.8865}
\defval{trunc/x40/1024/yocou-noloop/L128}{0.9846}
\defval{trunc/x40/1024/yocou-noloop/L2048}{0.8743}
\defval{trunc/x40/1024/yocou-noloop/L256}{0.9378}
\defval{trunc/x40/1024/yocou-noloop/L512}{0.9065}
\defval{truncdrop/x40/1024/ced-enc}{0.115}
\defval{truncdrop/x40/1024/ced-noloop}{0.111}
\defval{truncdrop/x40/1024/dense-l25}{0.118}
\defval{truncdrop/x40/1024/dense-noloop}{0.114}
\defval{truncdrop/x40/1024/yocou-loop}{0.111}
\defval{truncdrop/x40/1024/yocou-noloop}{0.110}
\defval{truncgap/x40/1024/yocou-loop-minus-ced-enc/L1024}{0.0030}
\defval{truncgap/x40/1024/yocou-loop-minus-ced-enc/L128}{0.0001}
\defval{truncgap/x40/1024/yocou-loop-minus-ced-enc/L2048}{0.0038}
\defval{truncgap/x40/1024/yocou-loop-minus-ced-enc/L256}{0.0007}
\defval{truncgap/x40/1024/yocou-loop-minus-ced-enc/L512}{0.0017}
\defval{truncgap/x40/1024/yocou-noloop-minus-ced-noloop/L1024}{0.0023}
\defval{truncgap/x40/1024/yocou-noloop-minus-ced-noloop/L128}{0.0015}
\defval{truncgap/x40/1024/yocou-noloop-minus-ced-noloop/L2048}{0.0027}
\defval{truncgap/x40/1024/yocou-noloop-minus-ced-noloop/L256}{0.0016}
\defval{truncgap/x40/1024/yocou-noloop-minus-ced-noloop/L512}{0.0019}
\defval{tt/long/1024/ced-dec/T1}{0.9092}
\defval{tt/long/1024/ced-dec/T12}{0.8296}
\defval{tt/long/1024/ced-dec/T16}{0.8296}
\defval{tt/long/1024/ced-dec/T2}{0.8543}
\defval{tt/long/1024/ced-dec/T4}{0.8331}
\defval{tt/long/1024/ced-dec/T8}{0.8297}
\defval{tt/long/1024/ced-enc/T1}{0.9552}
\defval{tt/long/1024/ced-enc/T12}{0.8122}
\defval{tt/long/1024/ced-enc/T16}{0.8122}
\defval{tt/long/1024/ced-enc/T2}{0.8622}
\defval{tt/long/1024/ced-enc/T4}{0.8206}
\defval{tt/long/1024/ced-enc/T8}{0.8124}
\defval{tt/long/1024/ced-noloop/T1}{0.8405}
\defval{tt/long/1024/dense-l47/T1}{0.9438}
\defval{tt/long/1024/dense-l47/T12}{0.8176}
\defval{tt/long/1024/dense-l47/T16}{0.8176}
\defval{tt/long/1024/dense-l47/T2}{0.8612}
\defval{tt/long/1024/dense-l47/T4}{0.8245}
\defval{tt/long/1024/dense-l47/T8}{0.8178}
\defval{tt/long/1024/dense-noloop/T1}{0.8406}
\defval{tt/x40/1024/ced-dec/T1}{0.9156}
\defval{tt/x40/1024/ced-dec/T12}{0.8648}
\defval{tt/x40/1024/ced-dec/T16}{0.8648}
\defval{tt/x40/1024/ced-dec/T2}{0.8795}
\defval{tt/x40/1024/ced-dec/T4}{0.8668}
\defval{tt/x40/1024/ced-dec/T8}{0.8648}
\defval{tt/x40/1024/ced-enc-fixed/T1}{2.0066}
\defval{tt/x40/1024/ced-enc-fixed/T12}{0.8476}
\defval{tt/x40/1024/ced-enc-fixed/T16}{0.8477}
\defval{tt/x40/1024/ced-enc-fixed/T2}{1.6951}
\defval{tt/x40/1024/ced-enc-fixed/T4}{1.1182}
\defval{tt/x40/1024/ced-enc-fixed/T8}{0.8484}
\defval{tt/x40/1024/ced-enc/T1}{0.9695}
\defval{tt/x40/1024/ced-enc/T12}{0.8462}
\defval{tt/x40/1024/ced-enc/T16}{0.8462}
\defval{tt/x40/1024/ced-enc/T2}{0.8893}
\defval{tt/x40/1024/ced-enc/T4}{0.8537}
\defval{tt/x40/1024/ced-enc/T8}{0.8465}
\defval{tt/x40/1024/ced-noloop/T1}{0.8716}
\defval{tt/x40/1024/dense-l25/T1}{0.9647}
\defval{tt/x40/1024/dense-l25/T12}{0.8455}
\defval{tt/x40/1024/dense-l25/T16}{0.8455}
\defval{tt/x40/1024/dense-l25/T2}{0.8877}
\defval{tt/x40/1024/dense-l25/T4}{0.8527}
\defval{tt/x40/1024/dense-l25/T8}{0.8457}
\defval{tt/x40/1024/dense-l47/T1}{0.9508}
\defval{tt/x40/1024/dense-l47/T12}{0.8507}
\defval{tt/x40/1024/dense-l47/T16}{0.8507}
\defval{tt/x40/1024/dense-l47/T2}{0.8874}
\defval{tt/x40/1024/dense-l47/T4}{0.8574}
\defval{tt/x40/1024/dense-l47/T8}{0.8511}
\defval{tt/x40/1024/dense-noloop/T1}{0.8746}
\defval{tt/x40/1024/yocou-loop/T1}{0.9563}
\defval{tt/x40/1024/yocou-loop/T12}{0.8501}
\defval{tt/x40/1024/yocou-loop/T16}{0.8501}
\defval{tt/x40/1024/yocou-loop/T2}{0.8880}
\defval{tt/x40/1024/yocou-loop/T4}{0.8567}
\defval{tt/x40/1024/yocou-loop/T8}{0.8503}
\defval{tt/x40/1024/yocou-noloop/T1}{0.8743}
\defval{tt/x40/1536/ced-dec/T1}{0.8795}
\defval{tt/x40/1536/ced-dec/T12}{0.8137}
\defval{tt/x40/1536/ced-dec/T16}{0.8137}
\defval{tt/x40/1536/ced-dec/T2}{0.8345}
\defval{tt/x40/1536/ced-dec/T4}{0.8169}
\defval{tt/x40/1536/ced-dec/T8}{0.8138}
\defval{tt/x40/1536/ced-enc/T1}{0.9359}
\defval{tt/x40/1536/ced-enc/T12}{0.7963}
\defval{tt/x40/1536/ced-enc/T16}{0.7963}
\defval{tt/x40/1536/ced-enc/T2}{0.8447}
\defval{tt/x40/1536/ced-enc/T4}{0.8045}
\defval{tt/x40/1536/ced-enc/T8}{0.7966}
\defval{tt/x40/1536/ced-noloop/T1}{0.8247}
\defval{tt/x40/1536/dense-l25/T1}{0.9306}
\defval{tt/x40/1536/dense-l25/T12}{0.7960}
\defval{tt/x40/1536/dense-l25/T16}{0.7960}
\defval{tt/x40/1536/dense-l25/T2}{0.8425}
\defval{tt/x40/1536/dense-l25/T4}{0.8036}
\defval{tt/x40/1536/dense-l25/T8}{0.7963}
\defval{tt/x40/1536/dense-noloop/T1}{0.8265}
\defval{tt/x40/768/ced-dec/T1}{0.9485}
\defval{tt/x40/768/ced-dec/T12}{0.9108}
\defval{tt/x40/768/ced-dec/T16}{0.9108}
\defval{tt/x40/768/ced-dec/T2}{0.9211}
\defval{tt/x40/768/ced-dec/T4}{0.9122}
\defval{tt/x40/768/ced-dec/T8}{0.9108}
\defval{tt/x40/768/ced-enc/T1}{0.9995}
\defval{tt/x40/768/ced-enc/T12}{0.8919}
\defval{tt/x40/768/ced-enc/T16}{0.8919}
\defval{tt/x40/768/ced-enc/T2}{0.9291}
\defval{tt/x40/768/ced-enc/T4}{0.8985}
\defval{tt/x40/768/ced-enc/T8}{0.8922}
\defval{tt/x40/768/ced-noloop/T1}{0.9142}
\defval{tt/x40/768/dense-l25/T1}{0.9941}
\defval{tt/x40/768/dense-l25/T12}{0.8917}
\defval{tt/x40/768/dense-l25/T16}{0.8917}
\defval{tt/x40/768/dense-l25/T2}{0.9281}
\defval{tt/x40/768/dense-l25/T4}{0.8979}
\defval{tt/x40/768/dense-l25/T8}{0.8919}
\defval{tt/x40/768/dense-l47/T1}{0.9824}
\defval{tt/x40/768/dense-l47/T12}{0.8973}
\defval{tt/x40/768/dense-l47/T16}{0.8973}
\defval{tt/x40/768/dense-l47/T2}{0.9277}
\defval{tt/x40/768/dense-l47/T4}{0.9030}
\defval{tt/x40/768/dense-l47/T8}{0.8976}
\defval{tt/x40/768/dense-noloop/T1}{0.9163}
\defval{tt/x40/768/yocou-loop/T1}{0.9908}
\defval{tt/x40/768/yocou-loop/T12}{0.8950}
\defval{tt/x40/768/yocou-loop/T16}{0.8950}
\defval{tt/x40/768/yocou-loop/T2}{0.9290}
\defval{tt/x40/768/yocou-loop/T4}{0.9013}
\defval{tt/x40/768/yocou-loop/T8}{0.8953}
\defval{tt/x40/768/yocou-noloop/T1}{0.9172}
\defval{wl/agent/dec}{1.31}
\defval{wl/agent/enc}{4.17}
\defval{wl/agent/encround}{6.04}
\defval{wl/agent/hit}{80,000}
\defval{wl/agent/hitrate}{95}
\defval{wl/agent/miss}{4,000}
\defval{wl/agent/out}{300}
\defval{wl/agent/ratio}{3.18}
\defval{wl/agent/share}{45}
\defval{wl/chat/dec}{3.03}
\defval{wl/chat/enc}{4.55}
\defval{wl/chat/encround}{4.82}
\defval{wl/chat/hit}{3,000}
\defval{wl/chat/hitrate}{86}
\defval{wl/chat/miss}{500}
\defval{wl/chat/out}{500}
\defval{wl/chat/ratio}{1.50}
\defval{wl/chat/share}{22}
\defval{wl/longdoc/dec}{1.10}
\defval{wl/longdoc/enc}{7.28}
\defval{wl/longdoc/encround}{7.28}
\defval{wl/longdoc/hit}{0}
\defval{wl/longdoc/hitrate}{0}
\defval{wl/longdoc/miss}{100,000}
\defval{wl/longdoc/out}{1,000}
\defval{wl/longdoc/ratio}{6.61}
\defval{wl/longdoc/share}{97}
\defval{wl/reasoning/dec}{3.76}
\defval{wl/reasoning/enc}{4.09}
\defval{wl/reasoning/encround}{4.09}
\defval{wl/reasoning/hit}{0}
\defval{wl/reasoning/hitrate}{0}
\defval{wl/reasoning/miss}{2,000}
\defval{wl/reasoning/out}{20,000}
\defval{wl/reasoning/ratio}{1.09}
\defval{wl/reasoning/share}{3}

\papertitlerunning{Looping Deeper Without Growing the KV Cache}

\begin{document}

\twocolumn[
\papertitle{Looping Deeper Without Growing the KV Cache: Recurrent Depth for Causal Encoder--Decoders}

\begin{paperauthorlist}
\paperauthor{Jiaqi Leng}{ucsd}
\paperauthor{Minshen Zhang}{ucsd}
\paperauthor{Zhaofeng Sun}{thu}
\paperauthor{Yu-Yang Qian}{nju}
\paperauthor{Peng Zhao}{nju}
\paperauthor{Hao Zhang}{ucsd}
\end{paperauthorlist}

\paperaffiliation{ucsd}{University of California, San Diego}
\paperaffiliation{thu}{Tsinghua University}
\paperaffiliation{nju}{Nanjing University}

\papercorrespondingauthor{Jiaqi Leng}{jileng@ucsd.edu}

\paperkeywords{Looped Transformers, KV Cache, LLM Serving, Causal Encoder-Decoder}

\vskip 0.3in

\begin{abstract}
Looped language models deepen computation without adding parameters, yet caching full-context key--value (KV) states across layers and rounds severely bounds serving batch size. We design a looped causal encoder--decoder (CED) that achieves recurrent depth with zero round-wise KV growth, addressing \textit{where} and \textit{how} to loop an asymmetric stack in which only sparse encoder layers generate global KV. We find that recurrence must be placed in the encoder to iteratively refine global KV representations rather than re-reading static decoder projections, which both drives language-modeling gains and preserves long-context scaling. Furthermore, training the encoder loop under stochastic recurrence depth induces a fixed-point equilibrium where every round losslessly shares the last round's KV cache. Compounding cross-layer and cross-round KV sharing, the resulting \textbf{\ourmethod} matches an unlooped CED's minimal memory footprint and recovers \val{frac/x40/1024/ced-over-dense}--\val{frac/x40/1536/ced-over-dense}\% of a looped Transformer's gain with 1/\val{kvratio/x40/1024/dense-over-ced} of its global KV. To test serving at production scale, we loop the 552B-parameter DeepSeek-V4.1-Flash for 8 rounds on one $4\times$GB200 node: with per-layer and per-round KV the loop cannot fit a single 128K-token request, whereas \ourmethod holds $36.8\times$ less KV per token, fits 38 such requests (as many as the unlooped model), and serves $3.6\times$ higher throughput on 8K-token prompts.
\end{abstract}
]

\begin{NoHyper}\printAffiliationsAndNotice{}\end{NoHyper} 

\section{Introduction}
\label{sec:intro}

\begin{figure*}[t]
\centering
\begin{minipage}[b]{0.58\textwidth}
\centering
\resizebox{\linewidth}{!}{%
\begin{tikzpicture}[x=0.58cm, y=1cm,
  box/.style={draw, minimum width=0.48cm, minimum height=0.42cm, inner sep=0pt, font=\scriptsize},
  S/.style={box, fill=gray!20}, F/.style={box, fill=orange!70, very thick}, R/.style={box, fill=orange!15},
  D/.style={box, fill=blue!20},
  loop/.style={-{Stealth[length=4pt]}, thick, red!70!black},
  lab/.style={font=\scriptsize, anchor=east}]
  \foreach \row/\y/\name in {enc/0/{CED, encoder loop}, dec/-1.35/{CED, decoder loop}} {
    \node[lab] at (-0.55, \y) {\name};
    \foreach \i in {0,1} \node[S] (\row\i) at (\i, \y) {\i};
    \foreach \i in {2,6} \node[F] (\row\i) at (\i, \y) {\i};
    \foreach \i in {3,4,5,7,8,9,10,11} \node[R] (\row\i) at (\i, \y) {\i};
  }
  \node[lab] at (-0.55, -2.7) {Dense loop};
  \foreach \i in {0,...,11} \node[D] (den\i) at (\i, -2.7) {\i};
  \draw[loop] (enc5.north) to[out=65, in=115, looseness=0.75] node[above, font=\scriptsize] {$T$ rounds} (enc2.north);
  \draw[loop] (dec10.north) to[out=65, in=115, looseness=0.75] node[above, font=\scriptsize] {$T$ rounds} (dec7.north);
  \draw[loop] (den5.north) to[out=65, in=115, looseness=0.75] node[above, font=\scriptsize] {$T$ rounds} (den2.north);
  \draw[densely dashed, thick, black!75] (6.5, 0.95) -- (6.5, -1.7);
  \node[font=\scriptsize, anchor=east] at (6.4, 1.0) {$\leftarrow$ prefill};
  \node[font=\scriptsize, anchor=west] at (6.6, 1.0) {skipped by prefill $\rightarrow$};
  \begin{scope}[shift={(-3.6, -3.45)}]
    \node[S, minimum width=0.3cm, minimum height=0.25cm] at (0,0) {}; \node[font=\scriptsize, anchor=west] at (0.25,0) {SWA only};
    \node[F, minimum width=0.3cm, minimum height=0.25cm] at (3.1,0) {}; \node[font=\scriptsize, anchor=west] at (3.35,0) {SWA + global KV source};
    \node[R, minimum width=0.3cm, minimum height=0.25cm] at (8.8,0) {}; \node[font=\scriptsize, anchor=west] at (9.05,0) {SWA + reads global KV};
    \node[D, minimum width=0.3cm, minimum height=0.25cm] at (13.8,0) {}; \node[font=\scriptsize, anchor=west] at (14.05,0) {Full-context KV};
  \end{scope}
\end{tikzpicture}}%
\\[3pt]{\small\textbf{(a) Layer layout and loop placement}}
\end{minipage}\hfill
\begin{minipage}[b]{0.40\textwidth}
\centering
\safegraphic[width=\linewidth]{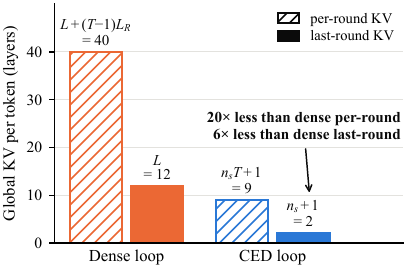}
\\[3pt]{\small\textbf{(b) Resident global KV per token}}
\end{minipage}
\caption{\textbf{Where and how to loop a Causal Encoder--Decoder.}
\emph{(a)}~Our 12-layer models; a loop repeats $L_R{=}4$ layers for $T$ rounds. Prefill runs only the layers left of the dashed line, so an encoder loop puts all $T$ rounds on the prefill path and a decoder loop does not; a dense loop runs every layer in prefill and decode.
\emph{(b)}~Global KV per token in units of one full-context layer ($L{=}12$, $L_R{=}4$, $T{=}\val{cfg/rounds}$, $n_s{=}1$ encoder KV source). Hatched bars keep KV for every round; solid bars keep only the last round's KV, which stochastic-depth training makes lossless (\S\ref{sec:method-train}).}
\label{fig:main}
\end{figure*}

Looped language models scale compute without scaling parameters by repeatedly applying a weight-tied layer block across recurrent rounds~\citep{universal_transformer,saunshi_looped,parcae,huginn,ouro}. Yet in production serving, memory capacity and decode throughput are bottlenecked not by parameter count alone, but by the \emph{resident KV cache} per request~\citep{vllm,gtagla}. To shrink this KV footprint, recent system-co-designed architectures replace homogeneous full-context layers with asymmetric KV routing~\citep{cla,yoco,deepseekv41}. Most notably, the \emph{causal encoder--decoder} (CED) in DeepSeek-V4.1-Flash~\citep{deepseekv41} equips every layer with bounded sliding-window attention (SWA) while restricting full-context (\emph{global}) KV generation to a few encoder layers; all decoder layers share a single global KV projected once from the final encoder output, allowing prompt prefill to bypass the decoder entirely.

Our goal is to combine recurrent looping with a CED to match the modeling quality and long-range scaling of a looped dense Transformer while preserving CED's serving efficiency. In a homogeneous dense Transformer, every layer generates and consumes full-context KV during both prefill and decode. In a CED, by contrast, layer depth directly governs both \emph{global KV generation} and \emph{prefill execution} (Fig.~\ref{fig:main}a), while repeating layers across rounds risks multiplying resident KV states (Fig.~\ref{fig:main}b). Designing a looped CED therefore comes down to two questions: \textbf{(1)~Where in the asymmetric CED stack should recurrence be placed} to maximize modeling capability, and how does that placement affect prefill and decode cost? \textbf{(2)~How can a looped CED be trained and cached} so that multi-round recurrence adds zero resident KV overhead over an unlooped CED without sacrificing quality?

To answer these questions, we compare looping the CED encoder against looping the decoder (Fig.~\ref{fig:main}a), and train the recurrent core under stochastic recurrence depth~\citep{deq,parcae,loopsright2} so that inference caches only the last round's KV rather than all $T$ rounds (Fig.~\ref{fig:main}b). We call the encoder loop trained and cached this way \textbf{\ourmethod}: at any number of rounds, it keeps the resident KV of an unlooped CED. Furthermore, because standard pretraining corpora and benchmarks are dominated by short sequences, we introduce offset-averaged long-document BPB to directly probe long-range routing through CED's global KV bottleneck. Training CED and dense models from scratch across three scales (\S\ref{sec:setup}), paired with a serving cost analysis at DeepSeek-V4.1-Flash scale (\S\ref{sec:cost}), we establish three main findings:
\begin{enumerate}
\item \textbf{Where to loop. Looping the CED encoder matches looped dense quality and long-context scaling across model scales} (\S\ref{sec:res-where}--\S\ref{sec:res-longdoc}). We find that looping the encoder is essential because effective recurrence must iteratively refine the global KV source itself, whereas loops that only re-read static global KV or local sliding windows fail to scale beyond the window. Across all three model sizes and extended training horizons, our encoder-looped CED consistently scales in capability, raising DCLM CORE accuracy by \val{coregain/long/1024/ced-enc} points and matching both the validation BPB and long-context scaling of a looped dense model.
\item \textbf{How to loop. Cross-round and cross-layer KV sharing compound losslessly, adding zero KV overhead and cutting global KV by \val{kvratio/x40/1024/dense-over-ced}--\val{kvratio/x40/1024/denseround-over-ced}$\times$} (\S\ref{sec:res-kv}). Stochastic-depth training induces a contractive fixed point in the CED encoder that makes cross-round KV sharing lossless at inference. Looping thus adds no resident KV over an unlooped CED, retaining only $1/\val{kvratio/x40/1024/dense-over-ced}$ of the global KV of a fixed-point looped dense model and $1/\val{kvratio/x40/1024/denseround-over-ced}$ of a per-round cached one (Fig.~\ref{fig:main}b).
\item \textbf{Serving analysis. Removing per-round KV turns an out-of-memory loop into a servable one at 552B scale} (\S\ref{sec:cost}). Looping the 552B-parameter DeepSeek-V4.1-Flash for 8 rounds with per-layer and per-round KV cannot fit a single 128K-token request on a $4\times$GB200 node; \ourmethod cuts per-token KV by $36.8\times$ and fits 38 such requests like the unlooped model, raising 8K/1K throughput by $3.6\times$ and cutting multi-turn agent TTFT by up to $29.7\times$. The cost of looping is compute, not memory: each round re-runs only the 18-layer recurrent core, and on prompt-heavy workloads the encoder loop pays a prefill premium (\S\ref{sec:cost}).
\end{enumerate}

\section{Preliminaries}
\label{sec:background}

\paragraph{Looped language models.} Following Huginn~\citep{huginn}, modern looped Transformers adopt a ``sandwich'' topology that partitions the network into a non-recurrent prelude $\mathcal{P}$, a weight-tied recurrent core $\mathcal{R}$, and a non-recurrent coda $\mathcal{C}$. For input tokens $x$,
\begin{equation}
  e = \mathcal{P}(x), \qquad h_r = \mathcal{R}(h_{r-1}, e), \qquad y = \mathcal{C}(h_T),
  \label{eq:sandwich}
\end{equation}
where $h_r \in \mathbb{R}^d$ is the per-token state after round $r$ of $T$ (hidden size $d$) and $y$ gives the next-token logits. Concentrating recurrence within a middle core surrounded by dedicated entry and exit blocks consistently outperforms looping the entire layer stack~\citep{saunshi_looped,loopsright1}. To stabilize recurrent dynamics and preserve token-level grounding across rounds, we adopt the gated input injection of Parcae~\citep{parcae}, which blends $\tilde{e} = \mathrm{RMSNorm}(e)$ into the state at each round:
\begin{equation}
  \mathcal{R}(h_{r-1}, e) = f\big(\bar{A} \odot h_{r-1} + \Delta \odot (B\tilde{e})\big),
  \label{eq:injection}
\end{equation}
where $f$ is one core pass, $\odot$ is the element-wise product, and learned $\Delta \in \mathbb{R}^d_{>0}$, $A \in \mathbb{R}^d_{<0}$, $B \in \mathbb{R}^{d \times d}$ give the step size, decay $\bar{A} = \exp(\Delta \odot A)$, and input projection.

\paragraph{Causal encoder--decoder (CED).} Introduced in DeepSeek-V4.1-Flash~\citep{deepseekv41}, a CED partitions an $L$-layer stack into an encoder followed by a decoder. To bound local attention memory, every layer maintains its own sliding-window attention (SWA) cache over the most recent $w$ tokens. Full-context (\emph{global}) KV includes a sparse subset of source layers within the encoder, and a single shared decoder KV entry linearly projected once at the encoder--decoder boundary from the normalized final encoder output. Any layer equipped with global attention jointly attends to its own local SWA cache and its designated global KV source—either the most recent encoder source preceding that layer, or the shared boundary projection—within a single unified softmax. Crucially, because all global KV entries are fully determined by the encoder output, prompt prefill only needs to execute the encoder layers, bypassing the decoder entirely.

\section{Looping a Causal Encoder--Decoder}
\label{sec:method}

Unlike a homogeneous dense Transformer—where every layer generates and consumes full-context KV, making loop depth a purely representational choice—layer depth in a CED is directly coupled to both \emph{global KV routing} and \emph{prefill execution} (\S\ref{sec:background}). Combining recurrence with a CED therefore requires jointly resolving two questions: \textbf{where} in the asymmetric stack to place the recurrent core (\S\ref{sec:method-where}), and \textbf{how} to train and cache recurrent states so that multi-round iteration does not inflate resident KV memory (\S\ref{sec:method-train}).

\subsection{Where to Loop: Architectural Placement and Serving Asymmetry}
\label{sec:method-where}

Because a weight-tied recurrent block cannot straddle the encoder--decoder boundary without breaking the single-projection invariant that enables decoder-free prefill, the CED topology naturally partitions loop placement into two regimes (Figure~\ref{fig:main}a):

\textbf{Encoder loop (Iterative Global KV Refinement).} The recurrent core resides entirely within the encoder and encompasses at least one global KV source layer. Across the $T$ rounds, the core iteratively refines both its local sliding-window representations and the encoder's global KV source. Because the final encoder representation—and by extension the projected decoder global KV—depends on the last round, prompt prefill must run all $T$ encoder rounds.

\textbf{Decoder loop (Static-Context Iterative Readout).} The recurrent core resides entirely within the decoder, downstream of the boundary projection. Here, all global KV entries—both in the encoder and at the boundary—are finalized before the loop begins. Each recurrent round refines its local SWA states while repeatedly querying a \emph{static} global KV cache. Because the decoder is skipped during prefill, this placement leaves prefill compute unchanged relative to an unlooped CED, apart from the last $w$ prompt tokens, which run the decoder to fill its SWA caches.

\paragraph{Compute and execution asymmetry.} 
In a CED, representational role and system execution move in lockstep. During autoregressive decoding, every token traverses the full network, so looping a block of layers for $T$ rounds adds the exact same number of single-layer passes per token in either placement. During prompt prefill, however, a decoder loop executes only the unlooped encoder, whereas an encoder loop must execute all $T$ rounds of its recurrent core before projecting the boundary KV. 

In our 12-layer experimental configuration with a 6-layer encoder, a 4-layer recurrent core, and $T{=}\val{cfg/rounds}$ rounds (Fig.~\ref{fig:main}a), an unlooped CED requires \val{passes/x40/1024/ced-noloop/prefill} layer passes per prompt token (plus the boundary projection), halving the \val{passes/x40/1024/dense-noloop/prefill} passes of an unlooped dense baseline. Looping the decoder preserves this lightweight \val{passes/x40/1024/ced-dec/prefill}-pass prefill, whereas looping the encoder expands prefill to \val{passes/x40/1024/ced-enc/prefill} layer passes per token; during decode, both placements expand execution from \val{passes/x40/1024/ced-noloop/decode} to \val{passes/x40/1024/ced-enc/decode} passes. The two placements also differ in KV footprint. Decoder layers produce no global KV, so a decoder loop keeps the unlooped CED's \val{kvlast/x40/1024/ced-noloop}\,KB of global KV per token (150M models, bf16) even if every round caches its own KV; an encoder loop rewrites its global KV source in every round, so per-round caching grows the global KV to \val{kvround/x40/1024/ced-enc}\,KB per token. Thus, comparing encoder and decoder loops directly tests whether high-quality recurrence requires actively refining the global KV source at the cost of prefill compute, or whether iterating over a fixed global context during decode suffices.

\subsection{How to Loop: Fixed-Point Training and Last-Round KV Caching}
\label{sec:method-train}

Even when a loop improves modeling quality, naively caching distinct keys and values across all $T$ rounds multiplies the resident KV footprint of every looped layer by $T$, negating CED's memory advantage. Eliminating this factor requires bridging a fundamental discrepancy between parallel training and memory-efficient inference.

\paragraph{Per-round training vs.\ last-round inference.}
During training, sequences are processed in parallel across tokens by unrolling the recurrent core: at round $r$, each layer inside the core attends to the round-$r$ keys and values of preceding tokens, which we term \emph{per-round KV}. Caching all $T$ trajectories during serving would require $T$ copies of the core's global and SWA caches: for the 150M encoder loop at $T{=}\val{cfg/rounds}$, the global KV per token grows from \val{kvlast/x40/1024/ced-enc}\,KB to \val{kvround/x40/1024/ced-enc}\,KB (\S\ref{sec:method-where}; Fig.~\ref{fig:main}b, hatched bars). To avoid this overhead, at inference we apply \emph{last-round KV} (or terminal KV sharing~\citep{loopsright2,benice}): once a token completes all $T$ rounds, we retain only its last-round KV states—both the global KV from the core's source layer and the local SWA KV of every looped layer—and discard all intermediate rounds. When a subsequent token executes any round $r$, it attends to its own round-$r$ state for the current step alongside the cached last-round KV for all past tokens.

\paragraph{Fixed-point convergence under stochastic depth.}
Substituting last-round KV for intermediate-round KV at inference avoids distribution shift only if the recurrent core converges toward a contractive fixed point across rounds, so that both the local SWA representations and the global KV source stabilize in late rounds. Unlike a deterministic fixed-depth unroll—which allows the core to learn round-specific feature transformations that diverge across rounds—we sample a stochastic recurrence depth per training sequence from a Poisson distribution with mean $\val{cfg/rounds}$ (clipped to at least $1$, backpropagating gradients through the final $\val{cfg/bwd}$ rounds) while executing a deterministic $\val{cfg/rounds}$ rounds at inference (Table~\ref{tab:hparams}). Exposing the coda and downstream layers to recurrent states across variable depths pressures the core to settle rapidly into a depth-invariant equilibrium~\citep{loopsright2,benice}. When this equilibrium holds (\S\ref{sec:res-kv}), \ourmethod retains the exact same resident KV footprint per token as an unlooped CED (\val{kvlast/x40/1024/ced-enc}\,KB of global KV per token at 150M, against \val{kvround/x40/1024/ced-enc}\,KB with per-round caching).

\section{Experimental Setup}
\label{sec:setup}

\paragraph{Architectures and model scales.} 
We instantiate DeepSeek-V4.1-Flash's asymmetric CED topology across an $L{=}\val{cfg/layers}$-layer stack (Figure~\ref{fig:main}a), split evenly into an encoder (layers 0--5) and a decoder (layers 6--11). Every layer maintains a local SWA cache over a $w{=}\val{cfg/window}$-token window. In the encoder, layers 0--1 use SWA only, while layer 2 serves as the sole encoder global KV source, read by layers 3--5. At the boundary, layer 6 projects the shared decoder global KV from the encoder output, read by layers 7--11. All global-reading layers merge local SWA and global KV in a single softmax (\S\ref{sec:background}). Each token thus stores only two global KV entries—one from the encoder and one from the decoder. 
To isolate CED's cross-layer KV topology, we use standard multi-head attention and dense FFNs without intra-layer KV compression (e.g., GQA or MLA) or MoE modules.
Our dense Transformer baseline uses the exact same dimensions but equips every layer with full-causal KV and no SWA, caching \val{kvratio/x40/1024/dense-over-ced}$\times$ more global KV per token than CED. Fixing this 12-layer layout, we scale hidden dimensions across $d \in \{768, 1024, 1536\}$, yielding approximately 90M, 150M, and 340M non-embedding parameters over a \val{cfg/seq}-token context (adding a loop introduces $\sim$1\% parameters for input injection; full architectural specifications are in Appendix~\ref{app:hparams}).

\paragraph{Compared configurations and ablations.} 
At each scale, we compare an unlooped baseline against a 4-layer recurrent loop ($T{=}\val{cfg/rounds}$ rounds) placed at layers 2--5 for both CED (\emph{encoder loop}) and the dense model, alongside a CED \emph{decoder loop} at layers 7--10. At 90M and 150M, we also train a dense model looped at the middle third (layers 4--7) to match Parcae's default placement~\citep{parcae}. Finally, two 150M CED ablations isolate our core mechanisms: (i) a \emph{fixed-depth} encoder loop trained with a deterministic \val{cfg/rounds}-round unroll to test whether cross-round KV sharing requires stochastic-depth training (\S\ref{sec:res-kv}); and (ii) a \emph{YOCO-U} variant~\citep{yocou} that loops layers 2--5 within an SWA-only encoder (making layer 6 the sole global KV source), testing whether the loop itself must recompute global KV (\S\ref{sec:res-longdoc}).

\paragraph{Training and evaluation.} 
Models are trained from scratch on FineWeb-Edu~\citep{fineweb} using Muon~\citep{moonlight_muon} and AdamW under a unified learning-rate schedule (Appendix~\ref{app:hparams}). Standard runs allocate $\sim$\val{cfg/ratio/x40/1024} tokens per non-embedding parameter (\val{cfg/tokens/x40/768}B, \val{cfg/tokens/x40/1024}B, and \val{cfg/tokens/x40/1536}B tokens for 90M, 150M, and 340M). To evaluate extended scaling, we also train the 150M models for \val{cfg/tokens/long}B tokens, matching the budget of 370M Parcae~\citep{parcae}, our 150M configuration (Parcae counts embeddings). Within each scale and budget, all runs see identical data orders, matching parameter counts and training tokens rather than FLOPs. 

We report upstream bits per byte (BPB) on FineWeb-Edu validation and downstream zero/few-shot accuracy on DCLM CORE and CORE-Extended~\citep{dclm}. Furthermore, because dependencies beyond CED's \val{cfg/window}-token SWA window rely exclusively on global KV—and standard FineWeb-Edu documents are too short to span our \val{cfg/seq}-token context (median \val{data/doc-median} tokens)—we evaluate position-wise BPB across contiguous \val{cfg/seq}-token chunks drawn from three long-document corpora: PG19~\citep{pg19}, FinePDFs ($\ge 4{,}096$ tokens)~\citep{kydlicek2025finepdfs}, and GovReport ($\ge 2{,}048$ tokens)~\citep{govreport} (protocol details in Appendix~\ref{app:longctx}).

\section{Results}
\label{sec:results}

\begin{figure*}[t]
\begin{minipage}[t]{\columnwidth}
\vspace{0pt}
\centering
\safegraphic[width=\columnwidth]{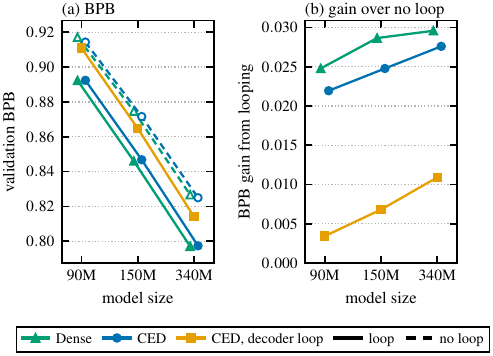}
\caption{\textbf{Validation BPB and recurrent loop gains across model scales.} Model size reflects non-embedding parameters, with each scale trained on \val{cfg/ratio/x40/1024} tokens per block parameter. (a) Upstream validation BPB. (b) BPB reduction of each loop over its unlooped counterpart. Solid: looped (L2--5; decoder loop L7--10); dashed: unlooped. \waitfor{bpb/x40/768/ced-enc,bpb/x40/768/ced-dec,bpb/x40/768/ced-noloop,bpb/x40/768/dense-l25,bpb/x40/768/dense-noloop,bpb/x40/1536/ced-enc,bpb/x40/1536/ced-dec,bpb/x40/1536/ced-noloop,bpb/x40/1536/dense-l25,bpb/x40/1536/dense-noloop}}
\label{fig:width}
\end{minipage}\hfill
\begin{minipage}[t]{\columnwidth}
\vspace{0pt}
\captype{table}
\caption{\textbf{Validation BPB of the 150M models.} Trained for \val{cfg/tokens/x40/1024}B tokens and evaluated at $T{=}\val{cfg/rounds}$ rounds on FineWeb-Edu over the full \val{cfg/seq}-token context (Appendix~\ref{app:eval}); lower is better. ``Gain'' denotes the BPB reduction relative to the corresponding unlooped baseline. Resident global KV is reported per token in bf16 under last-round KV caching.}
\label{tab:main}
\centering
\small
\setlength{\tabcolsep}{2pt}
\begin{tabular}{lcccc}
\toprule
Model & \makecell{Params\\(M)} & \makecell{KV\\(KB)} & BPB & Gain \\
\midrule
Dense, no loop & \sval{params/x40/1024/dense-noloop} & \sval{kvlast/x40/1024/dense-noloop} & \sval{bpb/x40/1024/dense-noloop} & -- \\
Dense, loop L2--5 & \sval{params/x40/1024/dense-l25} & \sval{kvlast/x40/1024/dense-l25} & \sval{bpb/x40/1024/dense-l25} & \sval{gain/x40/1024/dense-l25} \\
Dense, loop L4--7 (Parcae) & \sval{params/x40/1024/dense-l47} & \sval{kvlast/x40/1024/dense-l47} & \sval{bpb/x40/1024/dense-l47} & \sval{gain/x40/1024/dense-l47} \\
\midrule
CED, no loop & \sval{params/x40/1024/ced-noloop} & \sval{kvlast/x40/1024/ced-noloop} & \sval{bpb/x40/1024/ced-noloop} & -- \\
CED, encoder loop L2--5 & \sval{params/x40/1024/ced-enc} & \sval{kvlast/x40/1024/ced-enc} & \textbf{\sval{bpb/x40/1024/ced-enc}} & \textbf{\sval{gain/x40/1024/ced-enc}} \\
CED, decoder loop L7--10 & \sval{params/x40/1024/ced-dec} & \sval{kvlast/x40/1024/ced-dec} & \sval{bpb/x40/1024/ced-dec} & \sval{gain/x40/1024/ced-dec} \\
\midrule
\multicolumn{5}{l}{\emph{Ablations}} \\
CED, encoder loop, fixed rounds & \sval{params/x40/1024/ced-enc-fixed} & \sval{kvlast/x40/1024/ced-enc-fixed} & \sval{bpb/x40/1024/ced-enc-fixed} & \sval{gain/x40/1024/ced-enc-fixed} \\
YOCO-U style, no loop & \sval{params/x40/1024/yocou-noloop} & \sval{kvlast/x40/1024/yocou-noloop} & \sval{bpb/x40/1024/yocou-noloop} & -- \\
YOCO-U style, loop L2--5 & \sval{params/x40/1024/yocou-loop} & \sval{kvlast/x40/1024/yocou-loop} & \sval{bpb/x40/1024/yocou-loop} & \sval{gain/x40/1024/yocou-loop} \\
\bottomrule
\end{tabular}
\end{minipage}
\end{figure*}

We evaluate our models along two complementary axes: overall language modeling quality via validation BPB on FineWeb-Edu across the full \val{cfg/seq}-token training context (Table~\ref{tab:main}, Figure~\ref{fig:width}; \S\ref{sec:res-where}--\S\ref{sec:res-dense}), and true long-range dependency utilization via position-wise BPB on contiguous long documents (Figure~\ref{fig:longctx}; \S\ref{sec:res-longdoc}), followed by cross-round KV sharing (\S\ref{sec:res-kv}) and extended training evaluations (\S\ref{sec:res-long}).

\subsection{Where to Loop: Iterate the Encoder, Not the Decoder}
\label{sec:res-where}

Evaluating validation BPB across placements (Table~\ref{tab:main}) reveals a striking asymmetry: looping the CED encoder yields a substantial quality improvement over the unlooped CED baseline, whereas looping the decoder—despite introducing the exact same parameter overhead, recurrence depth, and per-token decode compute (\val{passes/x40/1024/ced-dec/decode} layer passes per token, \S\ref{sec:method-where})—recovers only \val{frac/x40/768/dec-over-enc}\%, \val{frac/x40/1024/dec-over-enc}\%, and \val{frac/x40/1536/dec-over-enc}\% of the encoder loop's gain at 90M, 150M, and 340M. Crucially, this gap is not an artifact of limited training tokens: even after $5\times$ longer training (\val{cfg/tokens/long}B tokens), the decoder loop captures only \val{frac/long/1024/dec-over-enc}\% of the encoder loop's improvement (Table~\ref{tab:long}).

This disparity stems directly from how each placement interacts with global KV formation (\S\ref{sec:method-where}). Looping the encoder iteratively refines the representations that feed both global caches: layer 2 actively updates its encoder global KV across all $T$ rounds, and the shared decoder KV is projected from the fully refined terminal encoder output $H_E$. Conversely, a decoder loop operates downstream of the boundary projection; its recurrent rounds can only repeatedly query a static global KV representation that was frozen before the loop began. Thus, effective recurrence in a CED requires placing the loop where global representations are constructed rather than where they are merely read.

\subsection{Quality Parity with Looped Dense Transformers at 1/6 Global KV}
\label{sec:res-dense}

When matched for parameter count, loop position (layers 2--5), training tokens, and recurrence depth, our looped CED matches the validation BPB of the looped dense Transformer within 0.001 at all three scales while retaining only one-sixth of the resident global KV (Table~\ref{tab:main}, Figure~\ref{fig:width}). In the unlooped regime, CED slightly outperforms the dense baseline across all three scales; adding the encoder loop amplifies CED capability nearly as strongly as looping benefits the dense stack, capturing \val{frac/x40/768/ced-over-dense}\%, \val{frac/x40/1024/ced-over-dense}\%, and \val{frac/x40/1536/ced-over-dense}\% of the dense loop gain at 90M, 150M, and 340M (the slight difference in relative gain likely reflects that the dense loop iteratively refines four full-KV layers whereas the CED encoder loop refines a single global-KV source alongside local SWA).

We also benchmark against Parcae's default middle-third loop placement (layers 4--7)~\citep{parcae}. In a dense Transformer, looping layers 4--7 slightly underperforms looping layers 2--5 at both 90M and 150M; we therefore use layers 2--5 as our primary dense baseline and include layers 4--7 as a literature reference. Over our extended \val{cfg/tokens/long}B-token training horizon—where the dense reference is looped at layers 4--7—the CED encoder loop finishes ahead by a margin matching this placement offset (Table~\ref{tab:long}), confirming that quality parity between looped CED and looped dense models persists across training scales.

\subsection{Long-Context Evaluation: Recurrence Must Refine Global KV}
\label{sec:res-longdoc}

\begin{figure*}[t]
\centering
\safegraphic[width=\textwidth]{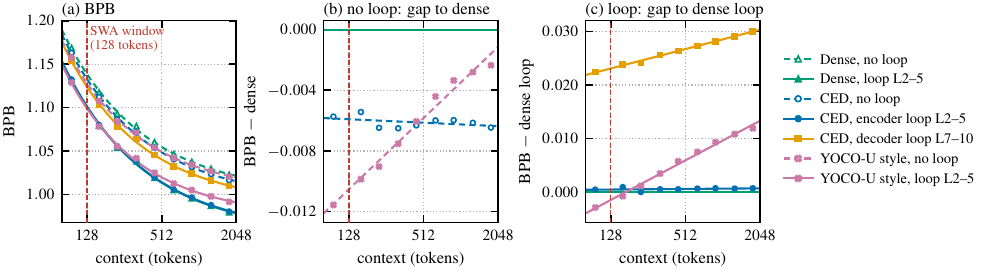}
\caption{\textbf{Position-wise BPB across long documents.} Pooled over PG19 test, FinePDFs, and GovReport (Appendix~\ref{app:longctx}); 150M models trained on \val{cfg/tokens/x40/1024}B tokens. (a) Absolute BPB as a function of available context length; lines fit $a + b\,n^{-c}$. (b) Unlooped models minus the unlooped dense model. (c) Looped models minus the L2--5 dense loop. In (b) and (c), lines are least-squares fits in $\log_2 n$. Markers are measurements; solid curves are looped (layers 2--5; the CED decoder loop, layers 7--10), dashed curves are unlooped. The red dashed line marks the \val{cfg/window}-token SWA window (per-corpus breakdowns in Figure~\ref{fig:longctx-corpora}).}
\label{fig:longctx}
\end{figure*}

Because standard validation corpora are dominated by short documents, aggregate BPB masks how models utilize long-range context. Evaluating position-wise BPB on contiguous long documents (\S\ref{sec:setup}, Figure~\ref{fig:longctx}; values in Table~\ref{tab:longctx}) directly tests whether recurrent computation in a CED can propagate information beyond its local \val{cfg/window}-token SWA window.

\paragraph{Looped CED preserves full-context scaling.} 
Because SWA layers cannot attend beyond \val{cfg/window} tokens, any long-range dependency must route through global KV. Strikingly, the quality gap between the CED encoder loop and the dense loop remains invariant to context length across all three long-document benchmarks (Figure~\ref{fig:longctx}c; Table~\ref{tab:slope}), just as unlooped CED maintains a constant lead over unlooped dense (Figure~\ref{fig:longctx}b). Storing more global KV layers does not by itself improve long-range modeling: the unlooped dense baseline caches $6\times$ more global KV than CED yet performs worst at every context position (Figure~\ref{fig:longctx}a, dashed). Instead, a single iteratively refined encoder KV source suffices—both the dense loop and the CED encoder loop amplify their quality gains over their unlooped counterparts at nearly identical rates as context length scales (and maintain parity over the \val{cfg/tokens/long}B-token horizon, with CED's positional lead over the L4--7 dense loop narrowing only marginally at the longest positions).

\paragraph{Why looping local SWA alone fails at long range (CED vs.\ YOCO-U).} 
Given that looping the lower half of the network is essential (\S\ref{sec:res-where}), a natural question is whether the looped encoder even needs a global KV source, or whether iterating purely over local SWA layers suffices—the architectural choice made by YOCO-U~\citep{yocou}. On aggregate validation BPB (Table~\ref{tab:main}), our YOCO-U-style ablation (which loops layers 2--5 inside an SWA-only encoder, deferring all global KV generation to layer 6) appears highly competitive, recovering \val{frac/x40/1024/yocou-over-enc}\% of the CED encoder loop's average gain. 

Position-wise evaluation on long documents, however, exposes what this short-context average hides (Figure~\ref{fig:longctx}c). Within the \val{cfg/window}-token SWA window, the YOCO-U loop performs slightly better than CED. Beyond the SWA window, however, YOCO-U degrades progressively relative to the dense and CED encoder loops with every doubling of context length across every corpus; by the end of the training context (1,024--2,048 tokens), it forfeits \val{posfrac/x40/1024/ced-enc-minus-yocou-loop/longdocs/c1024-2048}\% of the long-range gain achieved by the CED encoder loop, with the deficit continuing to widen. The CED decoder loop suffers a symmetric failure mode from the opposite side: because its rounds repeatedly query a frozen global KV projection, it falls further behind the dense loop as context grows (Figure~\ref{fig:longctx}c), and still does after $5\times$ longer training. Together, these ablations establish a core design principle: \textit{for recurrent computation to scale across long contexts, the looped block must actively recompute a global KV source.} A copy test isolates this mechanism: each sample's first 1,024 tokens are repeated verbatim, so every copied token must be retrieved through global KV (Figure~\ref{fig:recall} in Appendix~\ref{app:recall}).
At 90M and 150M, unlooped CED recalls worse than unlooped dense (\val{recalldiff/x40/768/ced-noloop-minus-dense-noloop} and \val{recalldiff/x40/1024/ced-noloop-minus-dense-noloop} BPB on the copied half). The encoder loop removes this deficit (\val{recalldiff/x40/768/ced-enc-minus-ced-noloop} and \val{recalldiff/x40/1024/ced-enc-minus-ced-noloop} against unlooped CED; looping the dense model gives \val{recalldiff/x40/768/dense-l25-minus-dense-noloop} and \val{recalldiff/x40/1024/dense-l25-minus-dense-noloop}) and recalls as well as the dense loop, whereas the decoder loop barely changes recall (\val{recalldiff/x40/768/ced-dec-minus-ced-noloop} and \val{recalldiff/x40/1024/ced-dec-minus-ced-noloop}) and the YOCO-U loop recalls worst of the looped models (\val{recalldiff/x40/768/yocou-loop-minus-dense-l25} and \val{recalldiff/x40/1024/yocou-loop-minus-dense-l25} against the dense loop). At 340M the differences shrink, and all five models lie between \val{recall/x40/1536/dense-l25/bpb} and \val{recall/x40/1536/ced-noloop/bpb} BPB.

\subsection{Lossless Cross-Round KV Sharing via Fixed-Point Convergence}
\label{sec:res-kv}

\begin{table*}[t]
\begin{minipage}[t]{\columnwidth}
\caption{\textbf{Extended training (\val{cfg/tokens/long}B tokens) and downstream benchmark accuracy.} 150M models trained to match the token budget of 370M Parcae~\citep{parcae}. BPB$_{\text{train}}$ denotes validation BPB on the training-time evaluation subset (Appendix~\ref{app:eval}). Downstream metrics report zero/few-shot accuracy on DCLM CORE and CORE-Extended (higher is better). Parcae reference rows are cited from~\citet{parcae} (Table~5), which use a different tokenizer and data ordering. Bold: best; underline: second best among our models.
\waitfor{bpb/long/1024/ced-noloop,bpb/long/1024/ced-dec,bpb/long/1024/dense-l47,bpb/long/1024/dense-noloop,core/long/1024/ced-dec,core/long/1024/dense-l47,core/long/1024/dense-noloop}}
\label{tab:long}
\centering
\small
\setlength{\tabcolsep}{2.5pt}
\begin{tabular}{lcccc}
\toprule
Model & BPB$_{\text{train}}$ & BPB & CORE & \makecell{CORE-\\Ext.} \\
\midrule
CED, no loop & \sval{bpbtrain/long/1024/ced-noloop} & \sval{bpb/long/1024/ced-noloop} & \sval{core/long/1024/ced-noloop} & \sval{coreext/long/1024/ced-noloop} \\
CED, encoder loop & \textbf{\sval{bpbtrain/long/1024/ced-enc}} & \textbf{\sval{bpb/long/1024/ced-enc}} & \underline{\sval{core/long/1024/ced-enc}} & \sval{coreext/long/1024/ced-enc} \\
CED, decoder loop & \sval{bpbtrain/long/1024/ced-dec} & \sval{bpb/long/1024/ced-dec} & \sval{core/long/1024/ced-dec} & \underline{\sval{coreext/long/1024/ced-dec}} \\
Dense, no loop & \sval{bpbtrain/long/1024/dense-noloop} & \sval{bpb/long/1024/dense-noloop} & \sval{core/long/1024/dense-noloop} & \sval{coreext/long/1024/dense-noloop} \\
Dense, loop L4--7 & \underline{\sval{bpbtrain/long/1024/dense-l47}} & \underline{\sval{bpb/long/1024/dense-l47}} & \textbf{\sval{core/long/1024/dense-l47}} & \textbf{\sval{coreext/long/1024/dense-l47}} \\
\midrule
Parcae: Transformer & -- & -- & \val{lit/parcae/core/transformer} & \val{lit/parcae/coreext/transformer} \\
Parcae: looped ($T{=}8$) & -- & -- & \val{lit/parcae/core/parcae} & \val{lit/parcae/coreext/parcae} \\
\bottomrule
\end{tabular}
\end{minipage}\hfill
\begin{minipage}[t]{\columnwidth}
\caption{\textbf{Per-round versus last-round KV caching.} Validation BPB under exact fp32 token-by-token decoding at $T{=}\val{cfg/rounds}$ rounds with per-round (Per) or last-round (Last) KV. Scale: non-embedding parameters, training tokens. Last-round KV keeps one terminal copy of the core's global and SWA KV per token, read by all rounds of later tokens.
\waitfor{kvshare/x40/1024/ced-dec/last,kvshare/x40/1024/dense-l25/last,kvshare/x40/1024/ced-enc-fixed/last,kvshare/x40/1536/ced-enc/last,kvshare/x40/1024/yocou-loop/last}}
\label{tab:kvshare}
\centering
\small
\setlength{\tabcolsep}{3.5pt}
\begin{tabular}{@{}lcrrr@{}}
\toprule
Model & Scale & Per & Last & Last$-$Per \\
\midrule
CED, encoder loop & 90M, \val{cfg/tokens/x40/768}B & \sval{kvshare/x40/768/ced-enc/round} & \sval{kvshare/x40/768/ced-enc/last} & \sval{kvshare/x40/768/ced-enc/delta} \\
CED, encoder loop & 150M, \val{cfg/tokens/x40/1024}B & \sval{kvshare/x40/1024/ced-enc/round} & \sval{kvshare/x40/1024/ced-enc/last} & \sval{kvshare/x40/1024/ced-enc/delta} \\
CED, encoder loop & 340M, \val{cfg/tokens/x40/1536}B & \sval{kvshare/x40/1536/ced-enc/round} & \sval{kvshare/x40/1536/ced-enc/last} & \sval{kvshare/x40/1536/ced-enc/delta} \\
CED, encoder loop & 150M, \val{cfg/tokens/long}B & \sval{kvshare/long/1024/ced-enc/round} & \sval{kvshare/long/1024/ced-enc/last} & \sval{kvshare/long/1024/ced-enc/delta} \\
CED, decoder loop & 150M, \val{cfg/tokens/x40/1024}B & \sval{kvshare/x40/1024/ced-dec/round} & \sval{kvshare/x40/1024/ced-dec/last} & \sval{kvshare/x40/1024/ced-dec/delta} \\
Dense, loop L2--5 & 150M, \val{cfg/tokens/x40/1024}B & \sval{kvshare/x40/1024/dense-l25/round} & \sval{kvshare/x40/1024/dense-l25/last} & \sval{kvshare/x40/1024/dense-l25/delta} \\
Dense, loop L4--7 & 150M, \val{cfg/tokens/x40/1024}B & \sval{kvshare/x40/1024/dense-l47/round} & \sval{kvshare/x40/1024/dense-l47/last} & \sval{kvshare/x40/1024/dense-l47/delta} \\
YOCO-U style, loop & 150M, \val{cfg/tokens/x40/1024}B & \sval{kvshare/x40/1024/yocou-loop/round} & \sval{kvshare/x40/1024/yocou-loop/last} & \sval{kvshare/x40/1024/yocou-loop/delta} \\
\midrule
\multicolumn{5}{l}{\emph{Fixed recurrence depth}} \\
CED, encoder loop & 150M, \val{cfg/tokens/x40/1024}B & \sval{kvshare/x40/1024/ced-enc-fixed/round} & \sval{kvshare/x40/1024/ced-enc-fixed/last} & \sval{kvshare/x40/1024/ced-enc-fixed/delta} \\
\bottomrule
\end{tabular}
\end{minipage}
\end{table*}

To verify whether cross-round KV sharing (\S\ref{sec:method-train}) degrades generation quality, Table~\ref{tab:kvshare} evaluates exact token-by-token decoding where every recurrent round queries only the historical tokens' last-round KV cache. Across every model trained with stochastic recurrence depth—whether CED or dense, encoder or decoder loop, 90M to 340M scale, or standard (\val{cfg/tokens/x40/1024}B) versus extended (\val{cfg/tokens/long}B) training—switching from per-round KV to last-round KV alters BPB by less than $0.001$. On all \val{cfg/kvfull/windows} validation samples, the change for the CED encoder loop is at most \val{kvfull/max} BPB at every scale, with 95\% intervals within $\pm$\val{kvfull/maxci} (Appendix~\ref{app:eval}). Consequently, \ourmethod retains only a single copy of its recurrent KV states instead of \val{cfg/rounds} (Figure~\ref{fig:main}b). We further confirm that this invariance holds under a realistic long-document serving protocol—where a long prompt is prefilled in parallel while caching only final-round KV before autoregressively decoding the continuation—yielding zero quality loss across all prompt lengths. In stark contrast, training the exact same CED encoder loop with a deterministic fixed depth ($T{=}\val{cfg/rounds}$) causes last-round KV sharing to fail catastrophically (and degrades baseline per-round stability; preliminary experiments on shallower 6-layer, 45M CED models exhibit the exact same dichotomy).

Appendix~\ref{app:fixedpoint} traces this behavior directly to the recurrent core's dynamical trajectory. Under stochastic-depth training, the CED encoder state contracts geometrically toward a fixed point across late rounds—stabilizing both its local SWA keys/values and its global KV source (\S\ref{sec:method-train})—such that unrolling up to $4\times$ more rounds at test time leaves BPB virtually unchanged. Conversely, deterministic fixed-depth models (trained with truncated backpropagation through the final \val{cfg/bwd} of \val{cfg/rounds} rounds) fail to contract and instead enter a limit cycle of period \val{fp/x40/1024/ced-enc-fixed/fw/period}, rendering last-round KV states incompatible with early rounds~\citep{loopsright2,benice}. While \citet{benice} find that Parcae models from \val{lit/benice/fail-min}M (about our 340M) can require a brief post-training phase to share KV losslessly, stochastic depth alone suffices across all scales studied here, and our serving analysis in \S\ref{sec:cost} assumes that lossless last-round KV caching is maintained at production scale.

\subsection{Extended Training Horizon and Downstream Generalization}
\label{sec:res-long}


Finally, Table~\ref{tab:long} evaluates whether the upstream BPB advantages of looping the CED encoder translate into downstream task accuracy over an extended \val{cfg/tokens/long}B-token training horizon ($\sim$200 tokens per non-embedding parameter). Looping the CED encoder substantially boosts downstream performance over the unlooped CED baseline, improving DCLM CORE by \val{coregain/long/1024/ced-enc} points (95\% paired bootstrap CI [\val{corediff/long/1024/ced-enc-minus-ced-noloop/lo}, \val{corediff/long/1024/ced-enc-minus-ced-noloop/hi}]) and CORE-Extended by \val{coreextgain/long/1024/ced-enc} points. Notably, while absolute scores across studies reflect differences in tokenizer and data ordering, the magnitude of our encoder loop's downstream gain matches that reported by~\citet{parcae} for 370M looped dense models at the identical token budget (\val{coregain/lit/parcae} points on CORE and \val{coreextgain/lit/parcae} on CORE-Extended). Compared against our own looped dense reference—which caches $6\times$ as much global KV—the CED encoder loop leads on validation BPB (\val{bpb/long/1024/ced-enc} vs.\ \val{bpb/long/1024/dense-l47}) and remains statistically indistinguishable on CORE under paired item-bootstrap testing (\val{corediff/long/1024/ced-enc-minus-dense-l47} points, 95\% CI [\val{corediff/long/1024/ced-enc-minus-dense-l47/lo}, \val{corediff/long/1024/ced-enc-minus-dense-l47/hi}]). Moreover, across our standard-budget width sweep against the stronger L2--5 dense loop, the modest downstream gap at 90M and 150M shrinks with model width and becomes statistically indistinguishable from zero at 340M.

Comparing the two CED loop placements on CORE further underscores the role of scale and context length. At standard token budgets, the decoder loop significantly trails the encoder loop across every model size, with its deficit widening to \val{corediff/x40/1536/ced-dec-minus-ced-enc} points (95\% CI [\val{corediff/x40/1536/ced-dec-minus-ced-enc/lo}, \val{corediff/x40/1536/ced-dec-minus-ced-enc/hi}]) at 340M. While extending training to \val{cfg/tokens/long}B tokens allows the 150M decoder loop to narrow this gap on CORE (\val{core/long/1024/ced-dec} vs.\ \val{core/long/1024/ced-enc}), this convergence is confined to short-prompt zero/few-shot items that largely fit within or near the \val{cfg/window}-token local SWA window—leaving the decoder loop's large deficits in validation BPB (\val{bpb/long/1024/ced-dec} vs.\ \val{bpb/long/1024/ced-enc}), long-document scaling (\S\ref{sec:res-longdoc}), and long-range in-context recall (Appendix~\ref{app:recall}) unchanged.

\section{Serving Cost Analysis}
\label{sec:cost}

\begin{figure*}[t]
\centering
\safegraphic[width=\textwidth]{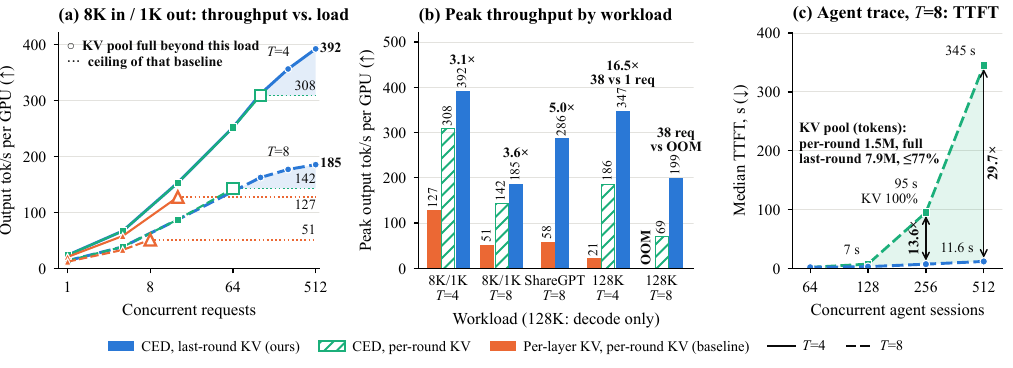}
\caption{\textbf{Serving looped \vfour-Flash on $4\times\text{GB200}$: last-round KV vs.\ per-round KV.}
\emph{(a)}~Throughput vs.\ concurrency on 8K-input / 1K-output requests; open markers show where per-round KV fills the KV pool.
\emph{(b)}~Peak preemption-free throughput; bold ratios are last-round \ced over the baseline with per-layer and per-round KV.
\emph{(c)}~Median time to first token on a multi-turn agent trace.
Setup and further numbers are in Appendix~\ref{app:serving-fig}.}
\label{fig:p1-pareto}
\end{figure*}

Sections~\ref{sec:method}--\ref{sec:results} showed that looping the \ced encoder preserves long-context scaling while stochastic-depth training allows inference to cache only the last round's KV state.
To verify that these dual memory savings translate directly into low-cost production serving at frontier scale, we benchmark a 40-layer, 552B-parameter \ced MoE backbone (\vfour-Flash~\citep{deepseekv41}) in SGLang~\citep{sglang} on NVIDIA GB200 ($4\times\text{GB200}$, TP4) and HGX H200 ($8\times\text{H200}$, TP8) nodes, with dummy or original \vfour-Flash weights, to test serving cost rather than quality (setup in Appendix~\ref{app:serving}).
Our baseline shares the same MoE, sparse attention, and looped layers, but follows standard Transformers by caching global KV at every layer (\emph{per-layer KV}) and follows prior looped models~\citep{parcae} by caching a separate copy at every recurrence round (\emph{per-round KV}, emulated at identical compute by capping the KV pool; Appendix~\ref{app:serving-fig}).
\ced's cross-layer KV sharing shrinks per-token global KV by $11.6\times$, and cross-round KV sharing (last-round KV) eliminates the per-round expansion.
Together, \ourmethod reduces the per-token KV footprint by $\mathbf{36.8\times}$ at $T=8$, expanding the usable KV pool on a single GB200 node to $5.11$\,M tokens regardless of recurrence depth (Table~\ref{tab:cost}).

\begin{table*}[t]
\centering
\caption{\textbf{Production serving summary at $T=8$ on $4\times\text{GB200}$ (TP4) with \vfour-Flash in SGLang.}
Comparing the baseline with \textbf{per-layer and per-round KV}, \textbf{\ced with per-round KV}, and \textbf{\ourmethod (ours; \ced, last-round KV)}; KV pools at the server settings of Figure~\ref{fig:p1-pareto}(a). All three loop \Lr{2}{19} with the same MoE FLOPs per decoded token; the baseline also runs every layer on every prompt token in prefill. Full $T=4$ and $8\times\text{H200}$ results are in Appendix~\ref{app:serving}.}
\label{tab:cost}
\vskip 0.06in
\small
\setlength{\tabcolsep}{4.2pt}
\begin{tabular}{@{}lcccc@{}}
\toprule
\textbf{Workload / Metric ($T{=}8$, GB200)} & \makecell{\textbf{Per-layer KV,}\\\textbf{per-round KV}} & \makecell{\textbf{\ced,}\\\textbf{per-round KV}} & \makecell{\textbf{\ourmethod}\\\textbf{(Ours)}} & \textbf{Advantage (Ours)} \\
\midrule
Global KV per Token (KB) & $60.0$ & $8.5$ & $\mathbf{1.6}$ & $\mathbf{36.8\times}$ \textbf{smaller} ($5.2\times$ vs.\ \ced per-round) \\
Usable KV Pool per Node (M tok) & $0.13$ & $0.98$ & $\mathbf{5.11}$ & $\mathbf{39\times}$ \textbf{larger} ($5.2\times$ vs.\ \ced per-round) \\
\midrule
8K In / 1K Out (tok/s/GPU) & $50.9$ & $142.5$ & $\mathbf{185.1}$ & $\mathbf{3.6\times}$ \textbf{higher} ($1.3\times$ vs.\ \ced per-round) \\
32K HBM Decode (tok/s/GPU) & --- & $289.3$ & $\mathbf{818.5}$ & $\mathbf{2.8\times}$ \textbf{higher decode throughput} \\
128K Long-Context (tok/s/GPU) & \textbf{OOM} ($0$ req) & $7$ reqs max & $\mathbf{199.2}$ ($\mathbf{38}$ reqs) & $\mathbf{38}$ \textbf{reqs vs.\ OOM} ($16.5\times$ at $T{=}4$) \\
Multi-Turn Agent TTFT ($c{=}512$) & --- & $345.1\,\text{s}$ & $\mathbf{11.6\,\text{s}}$ & $\mathbf{29.7\times}$ \textbf{lower latency} \\
\bottomrule
\end{tabular}
\end{table*}

\begin{figure*}[t]
\centering
\safegraphic[width=\textwidth]{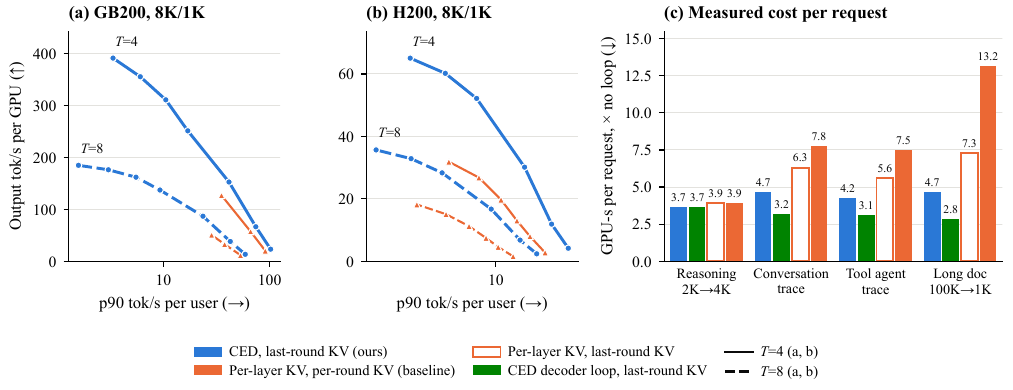}
\caption{\textbf{\ourmethod at interactive speeds and on real workloads.}
\emph{(a, b)}~Output throughput against p90 per-user decode speed on 8K-input / 1K-output requests, preemption-free points only, on \emph{(a)}~one $4\times\text{GB200}$ tray and \emph{(b)}~one $8\times\text{H200}$ node.
\emph{(c)}~Measured GPU-seconds per completed request relative to unlooped \ced on four closed-loop workloads at $T=8$ ($4\times\text{GB200}$; 32, 64, 64 and 8 concurrent requests).
Setup, model weights and further results are in Appendix~\ref{app:serving-more}.}
\label{fig:serving-main}
\end{figure*}

\paragraph{Eliminating the KV capacity wall in single-turn and long-context serving.}
In a large MoE model, high serving throughput requires batching many concurrent requests to amortize expert weight streaming across tokens.
As shown in Figure~\ref{fig:p1-pareto}(a,b) and Table~\ref{tab:cost}, the baseline with per-layer and per-round KV quickly exhausts HBM on standard 8K/1K prompts and cannot fit even a single 128K-context request at $T=8$ (\texttt{OOM}).
Sharing KV across layers (\ced with per-round KV) raises this concurrency ceiling, but per-round KV growth still caps the decode batch size.
Adding cross-round KV sharing (\ourmethod) makes the KV footprint invariant to recurrence depth, raising peak preemption-free throughput on 8K/1K to $\mathbf{3.1\times}$--$\mathbf{3.6\times}$ that of the baseline ($\mathbf{5.0\times}$ on ShareGPT), boosting HBM-saturated decode throughput by up to $\mathbf{2.8\times}$ over per-round \ced (Table~\ref{tab:hbm-decode-app}), and scaling to 38 concurrent 128K requests ($\mathbf{16.5\times}$ higher throughput than the baseline at $T=4$).
Only a small fraction of this gain comes from compute: at low concurrency where no KV pool is full, the baseline is only $\sim\!14\%$ slower (since its prefill runs all layers on every prompt token and every layer keeps its own KV cache and indexer); the rest comes from KV capacity.
At matched per-user interactivity targets, the throughput advantage is much larger than this raw compute gap because the faster model reaches a given per-user speed at higher concurrency (Figure~\ref{fig:serving-main}a,b).
A $2\times 2$ ablation confirms that cross-layer and cross-round sharing each improve throughput and compound multiplicatively (Table~\ref{tab:serving-2x2}, Appendix~\ref{app:serving-more}).
This capacity advantage extends up to 1M-token contexts on $8\times\text{H200}$, where \ourmethod at $T=8$ matches the request capacity of unlooped \ced and decodes $\mathbf{3.5\times}$ faster at 524K context than per-round KV (Table~\ref{tab:serving-longctx}).
\paragraph{Preventing prefix-cache eviction collapse in multi-turn agents.}
A minimal KV footprint is equally critical for multi-turn conversational and agentic workloads, where long prompts consist mostly of shared session history~\citep{mooncake,agentic_coding_wild}.
Because a radix prefix-cache hit in SGLang costs only $\sim\!1\%$ of computing a new prompt token, multi-turn latency hinges on keeping active session histories resident in HBM without eviction.
As shown in Figure~\ref{fig:p1-pareto}(c) and Table~\ref{tab:cost} (bottom), under high concurrency on the Mooncake \texttt{toolagent} trace, per-round KV saturates its cache pool ($100\%$ utilization), triggering continuous evictions that force the server to recompute evicted histories from scratch across all rounds.
Cross-round KV sharing expands the cache pool so it never fills, cutting median time-to-first-token (TTFT) by up to $\mathbf{29.7\times}$ on GB200, where per-round KV stalls for minutes (Figure~\ref{fig:serving-more}c), while similarly preserving higher prefix-cache hit rates on H200 (Appendix~\ref{app:serving-more}).

\paragraph{Compute grows only with the looped layers.}
With the KV capacity bottleneck removed, compute cost grows linearly with recurrence depth, but each extra round re-runs only the recurrent core (Appendix~\ref{app:serving}, Table~\ref{tab:unit-costs-app}).
Because the recurrent core spans less than half of the 40-layer stack, looping $T=8$ rounds increases physical layer passes per decoded token by $4.15\times$ over unlooped \ced, and measured per-step decode latency rises by $\mathbf{3.95\times}$.
Since decode executes every layer under either loop placement, encoder and decoder loops have identical decode cost (Appendix~\ref{app:serving-more}).
Consequently, on decode-dominated workloads such as long chain-of-thought reasoning ($>95\%$ decode time), the encoder loop matches the decoder loop in end-to-end serving cost ($\mathbf{3.7\times}$ that of $T=1$ at $T=8$; Figure~\ref{fig:serving-main}c) while preserving all long-context modeling gains (\S\ref{sec:results}).
Across all four workloads in Figure~\ref{fig:serving-main}(c), \ourmethod consistently uses less GPU time per request than both per-layer KV baselines.
In prefill, the encoder loop still skips the decoder on all but the local window, prefilling uncached prompts and cached long-context suffixes $\mathbf{1.4\times}$--$\mathbf{2.4\times}$ faster than the per-layer KV baseline, which runs every layer over the entire prompt (Table~\ref{tab:serving-prefill}).
Even on uncached prompt-heavy workloads, stochastic-depth training enables asymmetric prefill--decode recurrence ($T_p < T_d$; Appendix~\ref{app:split}, Table~\ref{tab:split}): halving prefill depth from $8$ to $4$ rounds retains over $99\%$ of the $8$-round validation loss improvement while reducing the GB200 prefill cost multiplier from $5.1\times$ to $\mathbf{2.7\times}$ (Table~\ref{tab:unit-costs-app}).

\section{Related Work}
\label{sec:related}

\paragraph{Recurrent-depth models and cross-round KV sharing.} 
Building on early weight-tied and equilibrium networks~\citep{universal_transformer,albert,deq}, modern prelude--core--coda language models and their routed or adapted variants scale reasoning and sample efficiency via recurrent depth~\citep{saunshi_looped,huginn,parcae,ouro,schwethelm,moeut,relaxed_recursive,mor}. To avoid linear KV growth across rounds, prior work either bounds and compresses cross-round caches~\citep{huginn,mor,plt,melt,lla} or reuses only the last round's KV---which collapses fixed-depth models~\citep{lla,schwethelm_cdb,loopsright2} but is lossless when stochastic-depth training induces a contractive fixed point~\citep{parcae,loopsright2,benice}. We extend it to asymmetric SWA and global-KV stacks.

\paragraph{Asymmetric KV topologies and loop placement.} 
Beyond head sharing, latent compression, and local windows~\citep{mqa,gqa,mla_dsv2,gtagla,longformer,streamingllm}, cross-layer KV sharing and hierarchical sparse attention restrict global KV generation to a few or single layers~\citep{cla,yoco,ramba,hsa_lengthgen}, culminating in CED's split between sparse encoder global-KV sources and a single projected decoder KV~\citep{deepseekv41}. Whereas prior recurrent cross-layer designs loop downstream of a frozen prelude KV or only over local SWA layers under fixed depth~\citep{cart,yocou}, we show that preserving long-context quality requires recomputing a global KV source \emph{inside} a stochastically trained encoder loop.

\paragraph{Concurrent work.}
Several concurrent preprints explore nearby points in this design space (Appendix~\ref{app:concurrent}). AsymLoop~\citep{asymloop} loops a CED-style decoder over fixed per-layer encoder KV, but each decoder layer recomputes a full-length local KV in every round, so its cache grows with the number of rounds unless that KV is shared. LPT~\citep{lpt} and \citet{lstmloop} share one round's KV across rounds---the first round's alongside a short local window, or the final round's with a consistency loss---and HLA~\citep{hla} adds a compact latent cache to frozen pretrained Ouro models. In contrast, \ourmethod loops the CED encoder so that each round refines a global KV source, keeps prefill decoder-free apart from the local window, and shares last-round KV through stochastic-depth training alone, without auxiliary losses or full-length per-round caches.

\paragraph{KV-centric serving systems.} 
Modern serving engines maximize throughput through paged KV management, prefix caching, and prefill--decode disaggregation~\citep{vllm,sglang,mooncake,distserve,splitwise}. Because an encoder loop trades prefill compute for a minimal decode KV cache, we evaluate our serving trade-offs (\S\ref{sec:cost}) under empirical prefix-cache hit rates and prefill-to-decode ratios from production traces~\citep{prefixcache_wild,mooncake,agentic_coding_wild}.

\section{Limitations}
\label{sec:limitations}

While our production serving benchmarks evaluate a frontier MoE architecture (\vfour-Flash, with a 552B-parameter backbone), our controlled pretraining quality ablations are conducted on dense models up to 340M parameters within a \val{cfg/seq}-token context. Pretraining a frontier-scale looped \ced MoE end-to-end—potentially incorporating lightweight post-training alignment~\citep{benice} to guarantee fixed-point convergence at extreme scale—and evaluating on broader long-context reasoning benchmarks remain important next steps.

Our serving results measure cost, not quality: the looped \vfour-Flash runs with dummy or original weights that were not trained to loop, so lossless last-round KV at this scale is assumed from our smaller models (\S\ref{sec:res-kv}). Per-round KV is emulated by capping the KV pool, and every run uses a single GB200 tray or H200 node rather than disaggregated or large expert-parallel deployments. Finally, because the encoder loop runs every round over the prompt, it costs more per request than a decoder loop on prompt-heavy workloads (Appendix~\ref{app:serving-more}), and this premium grows with larger decode batches.

\section{Conclusion}
\label{sec:conclusion}

This work introduces \ourmethod and establishes how to combine cross-layer KV consolidation (\ced) with recurrent-depth computation (Looped Transformers) so that looping deeper does not inflate KV memory. \textit{First}, resolving where to loop, we show that recurrence must be placed in the \ced encoder so that each round actively refines a global KV source, matching looped dense Transformers in validation BPB and long-document scaling while avoiding the long-range degradation of decoder or local-window loops. \textit{Second}, resolving how to loop, we demonstrate that training with stochastic recurrence depth drives the recurrent encoder toward a contractive fixed point, enabling lossless cross-round KV sharing with zero additional KV memory over an unlooped model. \textit{Third}, in production serving on $4\times\text{GB200}$ and $8\times\text{H200}$ nodes with \vfour-Flash, these dual savings shrink resident KV memory by $36.8\times$ over loops with per-layer and per-round KV ($5.2\times$ from cross-round sharing alone at $T=8$), delivering $3.1\times$--$16.5\times$ higher throughput across standard and 128K long-context workloads while cutting multi-turn agent TTFT by $13.6\times$--$29.7\times$ against per-round KV.

\section*{Acknowledgements}
We thank NVIDIA for supporting our development on GB200 NVL72 and for providing access to the rack used for
all experiments in this paper.

\clearpage

\bibliography{references}
\bibliographystyle{preprint}

\section*{AI Usage Disclosure}
In preparing this work, the authors used AI-based tools, including Claude Code, in an assistive capacity. Specifically, AI coding assistants were employed to accelerate the implementation of parts of the training, evaluation, and data-processing scripts under the authors' direct specification and verification. AI tools were also used to assist in literature retrieval, citation metadata checking, and language editing and polishing of the manuscript based on the authors' drafts and outlines. All experimental designs, core implementations, data analyses, and scientific conclusions are the original work of the authors. The authors have carefully reviewed and verified all code and text, and take full responsibility for the integrity and content of this paper.

\clearpage
\appendix
\makeatletter
\setlength{\@fptop}{0pt}\setlength{\@fpbot}{0pt plus 1fil}
\setlength{\@dblfptop}{0pt}\setlength{\@dblfpbot}{0pt plus 1fil}
\makeatother
\renewcommand{\topfraction}{0.9}\renewcommand{\bottomfraction}{0.8}\renewcommand{\textfraction}{0.05}
\renewcommand{\dbltopfraction}{0.95}\setcounter{dbltopnumber}{3}\setcounter{topnumber}{3}\setcounter{totalnumber}{6}
\renewcommand{\floatpagefraction}{0.85}\renewcommand{\dblfloatpagefraction}{0.85}
\section{Concurrent work}
\label{app:concurrent}

The works below appeared on arXiv in September and October 2026, while this paper was being finished.
Section~\ref{sec:related} contrasts the closest of them with \ourmethod in brief; here we describe each in more
detail. We have not compared against them experimentally. The fixed-point analyses of
\citet{loopsright2} and BeNice!~\citep{benice}, which we build on throughout, appeared in the same period.

\paragraph{Looped encoder--decoders.} AsymLoop~\citep{asymloop} builds on the CED of DeepSeek-V4.1-Flash and
loops the decoder. A causal encoder runs once and projects a global KV for each decoder layer, which stays fixed
across rounds; each layer of a weight-tied decoder reads it together with a full-length local KV recomputed from
the decoder's state in every round. From 60M to 1B parameters it improves on looping the whole stack, but it is
not compared with an encoder loop. AsymLoop does not discuss caching for decoding, but its decoder has a full-length KV of its own, recomputed every round,
so its cache grows with the number of rounds unless that KV is shared across rounds. Our decoder, like that of
DeepSeek-V4.1-Flash, keeps only sliding-window KV of its own: in our CED, a loop recomputes global KV every round
only if it sits in the encoder and contains a global-KV source. RLT~\citep{rlt} also pairs a parallel causal
encoder, whose output is projected into a global KV that the decoder layers read, with a decoder that keeps
sliding-window KV, but it recurs across tokens rather than across depth: each token's decoder input merges the
final decoder state of the previous token. It is evaluated on algorithmic tasks, and its decoder processes prompt
tokens one after another. Our loops repeat layers within each token and keep prefill parallel across tokens.

\paragraph{Sharing KV across rounds.} LPT~\citep{lpt} pretrains looped models in which only the first round
writes a full-length cache; later rounds read it together with a short window of their own KV. From 150M to 1B
parameters this sharing improves quality over a looped model that caches every round. We share the last round's
KV instead. Both policies cache the looped layers' KV once instead of once per round (LPT adds a short window per
round), and both compose with the reduction across layers that CED provides. In a CED, however, the choice does
not change prefill: the decoder's global KV is projected from the encoder output of the final round, so an
encoder loop runs every round on prompt tokens whichever round's KV the later rounds read. The LSTM-style looped
transformer of \citet{lstmloop} shares one cache across its whole recurrent core: every core block reads the KV
that the last core block writes in the final round, and a consistency loss late in training narrows the gap
between parallel training and recursive decoding. In an ablation, a cache producer placed before the loop gives
higher loss than this late producer at equal training tokens but lower loss at equal training compute; the author
does not conclude that either placement is better in general.

\paragraph{Compressing the caches of pretrained looped models.} HLA~\citep{hla} adds a compact cache to pretrained
Ouro models while keeping their weights frozen: exact KV for a window of recent tokens, and one latent per older
token that every loop reads directly. The cache per token shrinks by an order of magnitude. Like
LLA~\citep{lla}, HLA finds that reusing the last loop's KV in Ouro is unsafe for long generations, consistent with
Ouro's states not reaching a fixed point~\citep{loopsright2}.

\paragraph{Shared global KV with local history.} \citet{sharedglobal_local} studies a non-looped decoder in which a
lower stack produces one global KV that every upper layer reads, through one softmax, together with its own
sliding-window KV, the same split as in a CED decoder. It proves that prefill can rebuild the complete local caches
exactly while each upper layer processes only a suffix of the prompt that shrinks with depth. Our estimate of exact
SWA replay in \S\ref{sec:cost} already shrinks the replayed suffix pass by pass in this way; the proof also lets
queries and feed-forward layers run on a suffix shorter than the one used for keys and values, which would lower
that estimate slightly.

\begin{figure*}[t]
\centering
\safegraphic[width=\textwidth]{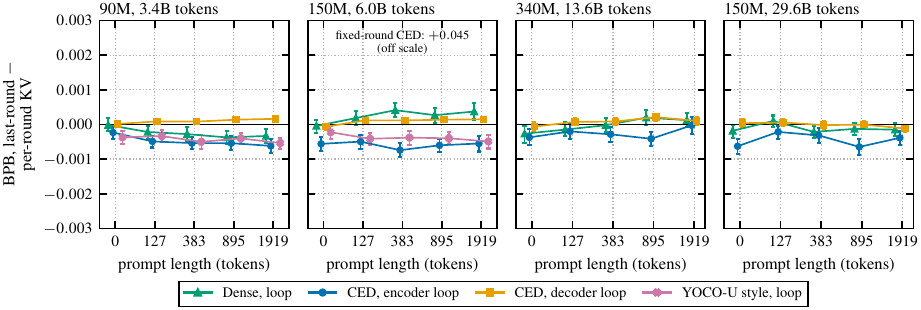}
\caption{\textbf{Last-round KV when serving a prompt.} The prompt is prefilled in one parallel pass, and the next
128 tokens are decoded one at a time; the plot shows their BPB with last-round KV minus with per-round KV, pooled
over PG19 test, FinePDFs and GovReport, with 95\% paired bootstrap intervals. Each panel compares the looped models
of one size and token budget (dense loop: L2--5; L4--7 in the last panel). The fixed-round CED encoder loop
(150M) is off the scale.}
\label{fig:kvprompt}
\end{figure*}

\section{Additional Serving Results}
\label{app:serving-more}

This appendix collects serving measurements that Section~\ref{sec:cost} does not show. They come from the same two
setups: one $4\times\text{GB200}$ tray (TP4/EP4) and one $8\times\text{H200}$ node (TP8/EP8), both running SGLang
\texttt{da64c5cbb} with our plugin. Runs marked \emph{dummy weights} use random weights with standard deviation 0.02,
which route tokens to experts almost uniformly. In an A/B test, dummy weights decoded 15--36\% slower than the real
ones, while the ratios between models moved by 2--4\%. All other runs use the original (real) \vfour-Flash weights, unchanged. In Figure~\ref{fig:serving-main}, panel (a) uses dummy weights and
panels (b) and (c) use the real weights. Every
comparison is at equal rounds $T$, and the names are those of Section~\ref{sec:cost}. Across layers, a model either
shares global KV as \ced does or keeps \emph{per-layer KV} (every layer \Lr{2}{39} caches its own global KV and runs
over every prompt token in prefill). Across rounds, it keeps \emph{last-round KV} or \emph{per-round KV}. The
\emph{baseline} keeps per-layer and per-round KV; \emph{\ourmethod} keeps neither. The \emph{decoder loop} loops
\Lr{21}{38} of \ced instead of the encoder. Per-round KV is always emulated by shrinking the KV pool, so a model and its
per-round variant run identical compute.

\begin{figure*}[t]
\centering
\safegraphic[width=\textwidth]{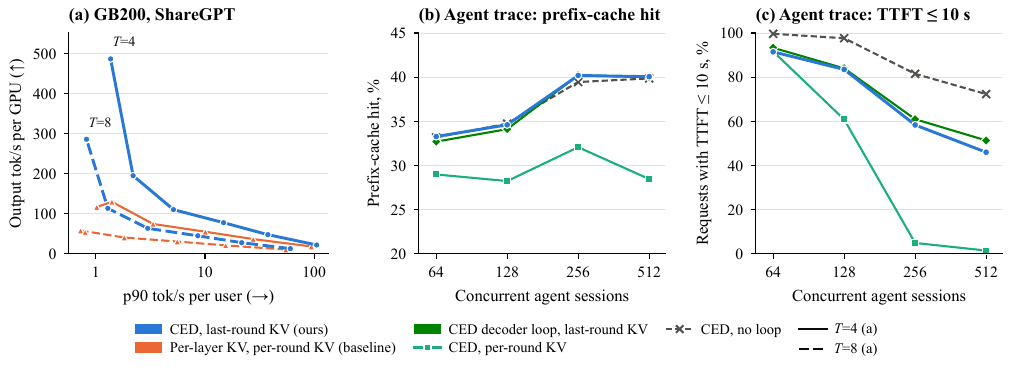}
\caption{\textbf{Further serving measurements.} \emph{(a)}~As Figure~\ref{fig:serving-main}(a), for ShareGPT on
one GB200 tray (dummy weights): output tokens/s per GPU against the p90 per-user decode speed
($1000/\text{p90 TPOT}$), one point per concurrency, preemption-free points only. \emph{(b, c)}~Mooncake tool-agent trace, closed loop, $T=8$, real
weights (the runs of Figure~\ref{fig:p1-pareto}c): \emph{(b)}~prefix-cache hit rate over the second half of each run;
\emph{(c)}~share of requests with TTFT $\le 10$\,s.}
\label{fig:serving-more}
\end{figure*}

\paragraph{The gain holds at interactive speeds.}
The $3.1\times$--$3.6\times$ of Section~\ref{sec:cost} compares peak throughput, which \ourmethod reaches at $c=512$.
Figures~\ref{fig:serving-main}(a, b) and~\ref{fig:serving-more}(a) plot throughput against the p90 per-user speed
instead. Reading the curves by linear interpolation between neighbouring points, at 30 and 50 tokens/s per user \ourmethod serves $1.39\times$--$1.67\times$ the throughput of
the baseline on 8K/1K and $1.42\times$--$1.61\times$ on ShareGPT. No KV pool is full at these speeds, so this part of
the gain comes from cross-layer sharing, not from capacity: the baseline runs every layer on every prompt token
(166 instead of 147 layer passes at $T=8$) and keeps a KV cache and an indexer in all 38 layers. A single request
shows the same: its time to first token (TTFT) on 8K/1K is 0.98\,s instead of 1.47\,s at $T=4$ and 1.68\,s instead of
2.65\,s at $T=8$, and its time per output token is 10\% lower. On H200 with real weights the gap is larger. At equal
concurrency, \ourmethod serves $1.48\times$--$1.64\times$ the throughput of the baseline on 8K/1K ($c=1$--64) and
$1.47\times$--$1.73\times$ on ShareGPT ($c=1$--256); its single-request TTFT is 1.95\,s instead of 3.23\,s ($T=4$)
and 3.56\,s instead of 5.61\,s ($T=8$). On 8K/1K the H200 baseline preempts at $c=112$ ($T=4$) and fills 96\% of its
pool at $c=64$ ($T=8$), while \ourmethod runs $c=256$ at 16\% KV use. On H200 ShareGPT the baseline's pool never
exceeds 28\%, so the $1.72\times$--$1.73\times$ there at $c=256$ comes from cross-layer sharing alone.

\begin{table*}[t]
\caption{\textbf{Cross-layer and cross-round sharing, measured separately} ($4\times\text{GB200}$, dummy weights).
Left: peak preemption-free output tokens/s per GPU on 8K/1K, with the largest preemption-free concurrency in
parentheses. Right: decode tokens/s per GPU at the largest batch of 128K-token requests that fits the KV pool, with
that batch in parentheses. Without a loop, \ced reaches 887.8 and 780.3 and the per-layer-KV model 371.7 and 116.7.
The ratios compare rows at equal $T$.}
\label{tab:serving-2x2}
\vskip 0.06in
\centering
\footnotesize
\setlength{\tabcolsep}{5pt}
\begin{tabular}{@{}llcccc@{}}
\toprule
& & \multicolumn{2}{c}{8K in / 1K out, peak} & \multicolumn{2}{c}{128K decode at the KV limit} \\
\cmidrule(lr){3-4}\cmidrule(lr){5-6}
KV across layers & KV across rounds & $T=4$ & $T=8$ & $T=4$ & $T=8$ \\
\midrule
\ced (\ourmethod, ours) & last round & 391.8 ($c$\,512) & 185.1 ($c$\,512) & 347.4 (38) & 199.2 (38) \\
\ced & per round & 308.4 ($c$\,128) & 142.5 ($c$\,64) & 186.1 (13) & 69.4 (7) \\
per layer & last round & 161.6 ($c$\,32) & 92.4 ($c$\,32) & 51.7 (3) & 29.6 (3) \\
per layer (baseline) & per round & 127.3 ($c$\,16) & 50.9 ($c$\,8) & 21.1 (1) & does not fit (0) \\
\midrule
\multicolumn{2}{@{}l}{Cross-layer sharing, both last-round KV (\ced $\div$ per layer)} & $2.43\times$ & $2.00\times$ & $6.72\times$ & $6.73\times$ \\
\multicolumn{2}{@{}l}{Cross-layer sharing, both per-round KV} & $2.42\times$ & $2.80\times$ & $8.82\times$ & --- \\
\multicolumn{2}{@{}l}{Cross-round sharing in \ced (last round $\div$ per round)} & $1.27\times$ & $1.30\times$ & $1.87\times$ & $2.87\times$ \\
\multicolumn{2}{@{}l}{Cross-round sharing with per-layer KV} & $1.27\times$ & $1.82\times$ & $2.45\times$ & --- \\
\multicolumn{2}{@{}l}{Both (\ourmethod $\div$ baseline)} & $\mathbf{3.08\times}$ & $\mathbf{3.64\times}$ & $\mathbf{16.5\times}$ & 199.2 vs.\ none \\
\bottomrule
\end{tabular}
\end{table*}

\paragraph{Both kinds of sharing contribute, and they multiply.}
Table~\ref{tab:serving-2x2} measures all four combinations of cross-layer sharing (\ced or per-layer KV) and
cross-round sharing (last-round or per-round KV). With last-round KV in both, cross-layer sharing alone gives
$2.0\times$--$2.4\times$ the peak 8K/1K throughput and $6.7\times$ the 128K decode throughput, because \ced fits 38
requests of 128K tokens where per-layer KV fits 3. Cross-round sharing alone gives $1.27\times$--$1.30\times$ in \ced
and $1.27\times$--$1.82\times$ with per-layer KV on 8K/1K, and $1.87\times$--$2.87\times$ in \ced at 128K. The 8K/1K
factors multiply to the headline exactly ($1.27\times 2.43=3.08$, $1.82\times 2.00=3.64$). Cross-round sharing thus
also helps a model with per-layer KV: the benefit of stochastic-depth training is not specific to \ced. At equal batch,
a model and its per-round variant decode within 3\% of each other (one point differs by 6\%), as the emulation intends,
so these gains come from the larger batch that fits. For the same reason, cross-round sharing changes nothing in \ced
at 30 or 50 tokens/s per user (factor 1.00), where its pool is at most a third full. It pays off once the pool fills:
at long context, on agent traces and under heavy load.

\begin{table*}[t]
\caption{\textbf{Prefill cost of running every layer over the prompt.} Top: ratio of per-request prefill time for
uncached prompts (batch 1, real weights); both models keep last-round KV. \emph{Full replay} runs the \ced decoder over
every prompt token instead of the last 128. Bottom: ratio of new prompt tokens prefilled per second behind one cached
prefix ($4\times\text{GB200}$, dummy weights, 4K new tokens per request). Values above 1 mean the \ced variant is
faster.}
\label{tab:serving-prefill}
\vskip 0.06in
\centering
\footnotesize
\setlength{\tabcolsep}{5pt}
\begin{tabular}{@{}llccccc@{}}
\toprule
& & \multicolumn{5}{c}{Prompt length (tokens)} \\
\cmidrule(lr){3-7}
& Ratio & 512 & 2K & 4K & 16K & 64K \\
\midrule
H200 & \ced, no loop: full replay $\div$ 128-token tail & 1.32 & 1.45 & 1.50 & 1.60 & 1.75 \\
H200 & Per-layer KV $\div$ \ced, encoder loop $T=8$ & 1.87 & 1.45 & 1.44 & 1.69 & 2.43 \\
GB200 & Per-layer KV $\div$ \ced, encoder loop $T=8$ & 1.65 & 1.48 & 1.41 & 1.40 & 1.48 \\
\midrule
& & \multicolumn{5}{c}{Context length / prefill chunk (tokens)} \\
\cmidrule(lr){3-7}
& & 64K / 4K & 64K / 8K & 64K / 16K & 128K / 4K & 128K / 8K \\
\midrule
GB200 & \ourmethod $\div$ baseline, $T=4$ & 1.60 & 1.64 & 1.70 & 1.65 & 1.83 \\
\bottomrule
\end{tabular}
\end{table*}

\paragraph{Prefill: skipping the decoder survives looping.}
A \ced prompt runs the encoder and a 128-token decoder tail, and an encoder loop keeps that property
(Table~\ref{tab:serving-prefill}). Even without a loop, running the decoder over the whole prompt costs $1.32\times$ at
512 tokens and $1.75\times$ at 64K tokens on H200, and $1.30\times$ at 8K tokens on GB200. With eight rounds, the
model with per-layer KV takes $1.40\times$--$1.65\times$ (GB200) and $1.44\times$--$2.43\times$ (H200) as long as the
\ced encoder loop to prefill the same prompt. Behind a cached 60K--124K-token prefix, \ourmethod at $T=4$ prefills
$1.60\times$--$1.83\times$ as many new tokens per second as the baseline at the same chunk size, and $1.56\times$ at
$T=8$ (64K context, 4K chunk). The encoder loop's own prefill grows more slowly than its round count
(Table~\ref{tab:unit-costs-app}).

\begin{table*}[t]
\caption{\textbf{Requests that fit the KV pool at long context.} $4\times\text{GB200}$: one tray with the flags of
Figure~\ref{fig:p1-pareto}, dummy weights; requests of $L+256$ tokens; the last column is decode tokens/s per GPU at
the largest 128K batch that fits. The \ced limit of 604 at 8K is computed from the pool (the sweep stopped at 256).
$8\times\text{H200}$: one node, real weights, requests that fit at full length. The H200 per-round pool also charges
each round's sliding-window rings; the GB200 one does not.}
\label{tab:serving-longctx}
\vskip 0.06in
\centering
\footnotesize
\setlength{\tabcolsep}{5pt}
\begin{tabular}{@{}llcccccc@{}}
\toprule
& & \multicolumn{3}{c}{$4\times\text{GB200}$} & \multicolumn{3}{c}{$8\times\text{H200}$} \\
\cmidrule(lr){3-5}\cmidrule(lr){6-8}
Model & KV across rounds & 8K & 128K & tok/s/GPU at 128K & 120K & 524K & 1M \\
\midrule
\ced, no loop & --- & 604 & 38 & 780.3 & 195 & 44 & 23 \\
\ced encoder loop, $T=4$ & last round & 604 & 38 & 347.4 & --- & --- & --- \\
\ced encoder loop, $T=8$ & last round & 604 & 38 & 199.2 & 195 & 44 & 23 \\
\ced decoder loop, $T=8$ & last round & --- & --- & --- & 195 & 44 & 23 \\
\ced encoder loop, $T=4$ & per round & 215 & 13 & 186.1 & --- & --- & --- \\
\ced encoder loop, $T=8$ & per round & 116 & 7 & 69.4 & 23 & 5 & 2 \\
\midrule
Per-layer KV, no loop & --- & 49 & 3 & 116.7 & 16 & 3 & 1 \\
Per-layer KV, $T=4$ & last round & 49 & 3 & 51.7 & --- & --- & --- \\
Per-layer KV, $T=8$ & last round & 49 & 3 & 29.6 & 16 & 3 & 1 \\
Per-layer KV, $T=4$ & per round & 25 & 1 & 21.1 & --- & --- & --- \\
Per-layer KV, $T=8$ & per round & 15 & 0 & --- & --- & --- & --- \\
\bottomrule
\end{tabular}
\end{table*}

\paragraph{Cost per request on four workloads.}
Figure~\ref{fig:serving-main}(c) runs four closed-loop workloads on one GB200 tray with real weights at $T=8$, one run
each:
synthetic reasoning (2K-token prompts, 4K output tokens, 32 concurrent requests), the Mooncake conversation and
tool-agent traces~\citep{mooncake} (64 concurrent requests, starting at minute 10), and synthetic long documents
(100K-token prompts, 1K output tokens, 8 concurrent requests). \ourmethod costs $3.7\times$--$4.7\times$ the GPU time
per request of an unlooped \ced. That is $0.64\times$--$0.93\times$ the cost of per-layer KV with last-round KV, and
$0.36\times$--$0.93\times$ the cost of the baseline. The largest saving is on long documents, where the baseline's pool
holds one 101K-token request at a time. The cost model, evaluated for the same runs, matches the two \ced loops
(measured over predicted 0.78--1.04). It underpredicts per-layer KV with last-round KV by 9--44\% on the three workloads
where that model's KV pool fills (95--100\%), an effect that a model of FLOPs and bytes does not capture. The encoder
loop costs the same as the decoder loop on reasoning ($1.00\times$) and $1.37\times$--$1.64\times$ as much on the three
prompt-heavy workloads, where its rounds run over every prompt token. These runs decode 8--64 requests at a time on one tray. With larger decode batches, as in
expert-parallel serving across many GPUs, each output token costs less and prompt tokens take a larger share of a
request, so the premium grows; running prompts at fewer rounds ($T_p<T_d$, Appendix~\ref{app:split}) lowers it.

\paragraph{Cross-round sharing leaves prefix caching as it is without a loop.}
Figure~\ref{fig:serving-more}(b, c) replays the Mooncake tool-agent trace in closed loop
with 64--512 concurrent sessions, real weights and $T=8$ (the runs of Figure~\ref{fig:p1-pareto}c); each point replays
600--1{,}536 complete requests from the trace. With last-round KV, both the encoder and
the decoder loop hit the prefix cache within 0.8 points as often as the unlooped \ced at every concurrency, so looping
does not change which session histories stay resident. With per-round KV, the 1.5M-token pool is full from $c=128$,
and the hit rate is 4.3, 6.4, 8.1 and 11.6 points lower than with last-round KV at $c=64$, 128, 256 and 512. The
share of requests with TTFT $\le 10$\,s then falls to 4.8\% at $c=256$ (58.3\% with last-round KV) and 1.4\% at
$c=512$ (46.0\%). A synthetic multi-turn agent workload on H200 (80 sessions of 4 turns, 32K--40K-token contexts,
0.2 requests/s) shows the same direction: per-round KV lowers the hit rate of the encoder loop from 72.3\% to 63.8\%
and its completed rate from 0.199 to 0.171 requests/s. Cross-layer sharing matters even more on this workload. In the
same runs, per-layer KV with last-round KV holds 2.0M tokens against \ced's 23.4M, so the session histories (about
3.2M tokens) do not stay resident: its hit rate is 3.4\% and it completes 0.039 requests/s, $5.1\times$ fewer than
\ourmethod.

\paragraph{Encoder and decoder loops decode at the same cost.}
Decode runs every layer under either loop placement. In a pure-decode sweep with real weights on GB200 that fills the KV
pool (8K, 16K and 32K-token contexts, 256 output tokens per request), the encoder loop and the decoder loop at $T=8$
reach 1{,}451 / 1{,}118 / 818 and 1{,}460 / 1{,}112 / 828 output tokens/s per GPU (the 8K values are capped at 768
requests). With last-round KV neither loop adds global KV, so both hold the same pool, and both take 33.1\,ms per decode
step at 8K context and batch 32.

\paragraph{Long and very long contexts.}
With last-round KV, a looped \ced fits as many requests as the unlooped \ced at every $T$
(Table~\ref{tab:serving-longctx}): 38 requests of 128K tokens on one GB200 tray, and 44 of 524K tokens and 23 of 1M
tokens on one H200 node, where the unlooped model with per-layer KV fits 3 and 1. At 524K tokens on H200, last-round KV
decodes 195 tokens/s per node at a batch of 32. Per-round KV fits only 5 such requests and peaks at 55.7 tokens/s at a
batch of 4, $3.5\times$ less. For every model, the long-context decode measurements use KV that is allocated without
being computed, so that they measure decode alone; on H200, decoding over such KV was within 3\% of decoding after a
real prefill.

\section{Evaluation protocol}
\label{app:eval}

\paragraph{Validation BPB.} We tokenize the FineWeb-Edu validation split into one token stream and cut it into
consecutive samples of \val{cfg/seq} tokens, the training context; a sample may span document boundaries. Every
model is scored on all \val{cfg/fwwin} samples, about \val{cfg/fwtok}M tokens. BPB is the total negative
log-likelihood of the predicted tokens in bits, divided by their total number of UTF-8 bytes, so it does not depend
on the tokenizer. Looped models run \val{cfg/rounds} rounds. During training we evaluated only the first
\val{cfg/evalwin/x40} samples (\val{cfg/evalwin/long} for the \val{cfg/tokens/long}B-token runs; BPB$_{\text{train}}$ in
Table~\ref{tab:long}); on the first \val{cfg/evalwin/x40} samples, every model's BPB is within 0.001 of its value on
all samples, and no ranking of models changes.

\paragraph{Last-round KV.} Last-round KV cannot be scored in one parallel pass, because a token's last-round KV
exists only after that token has run all its rounds. We therefore decode every sample token by token with teacher
forcing, in fp32, as \citet{benice} do for shared-KV perplexity. With per-round KV, the parallel pass gives the
same result as token-by-token decoding, which we check in our tests. The deployment protocol of \S\ref{sec:res-kv}
instead prefills a prompt in one parallel pass and decodes only the answer. With last-round KV the prompt keeps only its tokens' final-round KV,
which every round of the answer reads; Figure~\ref{fig:kvprompt} shows that this loses nothing at any prompt length. Table~\ref{tab:kvshare} decodes the
first \val{cfg/evalwin/x40} validation samples. For the CED encoder loop we also decode all
\val{cfg/kvfull/windows} samples of the validation split and bootstrap the paired difference over samples: last-round
minus per-round KV is \val{kvfull/x40/768/ced-enc/delta} [\val{kvfull/x40/768/ced-enc/delta/lo},
\val{kvfull/x40/768/ced-enc/delta/hi}] at 90M, \val{kvfull/x40/1024/ced-enc/delta}
[\val{kvfull/x40/1024/ced-enc/delta/lo}, \val{kvfull/x40/1024/ced-enc/delta/hi}] at 150M,
\val{kvfull/x40/1536/ced-enc/delta} [\val{kvfull/x40/1536/ced-enc/delta/lo}, \val{kvfull/x40/1536/ced-enc/delta/hi}]
at 340M, and \val{kvfull/long/1024/ced-enc/delta} [\val{kvfull/long/1024/ced-enc/delta/lo},
\val{kvfull/long/1024/ced-enc/delta/hi}] after \val{cfg/tokens/long}B tokens.

\paragraph{Downstream intervals.} CORE scores are mean centered accuracy over the suite's tasks. Their 95\%
intervals come from 1,000 bootstrap resamples of items, stratified by task; differences between two models are paired
over items where both have per-item results, and use unpaired resamples otherwise.

\section{Fixed-point analysis}
\label{app:fixedpoint}

\begin{figure}[t]
\centering
\safegraphic[width=\columnwidth]{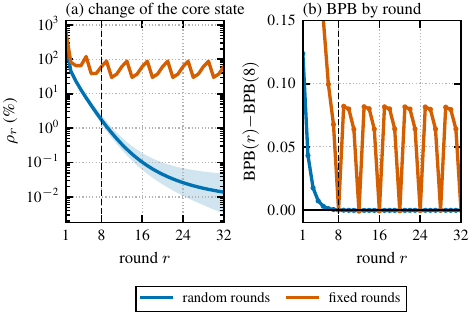}
\caption{\textbf{Fixed-point analysis.} The 150M CED encoder loop (\val{cfg/tokens/x40/1024}B tokens), trained with a
random or a fixed number of rounds and run for up to \val{cfg/fp/rounds} rounds on FineWeb-Edu validation data in fp32. (a) Median relative
change of the core state between rounds, $\rho_r = \lVert h_r - h_{r-1}\rVert / \lVert h_{r-1}\rVert$, over tokens,
with the interquartile band. (b) BPB after $r$ rounds minus BPB after the trained mean of \val{cfg/rounds} rounds
(dashed line).}
\label{fig:fixedpoint}
\end{figure}

\begin{figure*}[t]
\centering
\safegraphic[width=\textwidth]{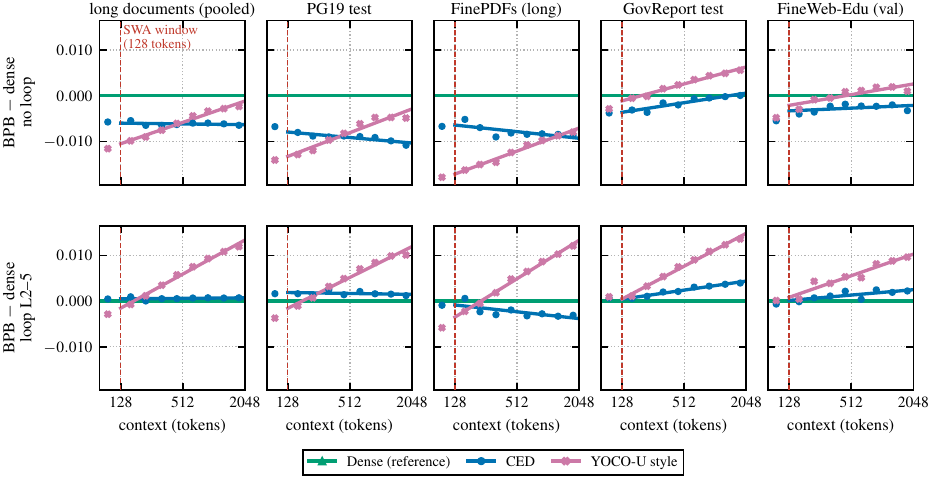}
\caption{\textbf{Gap to dense on each corpus.} BPB minus the dense model with the same loop setting, by tokens of
context, for each corpus (150M
models, \val{cfg/tokens/x40/1024}B tokens). Markers are measurements; lines are least-squares fits in $\log_2$ of
the context beyond the SWA window (red dashed line). Top: no loop. Bottom: loop over layers 2--5. FineWeb-Edu counts
each token's context within its own document and is not averaged over offsets.}
\label{fig:longctx-corpora}
\end{figure*}

We run each loop for up to \val{cfg/fp/rounds} rounds and track $\rho_r$, the relative change of the core state
between rounds, as \citet{benice} do. With random rounds, $\rho_r$ shrinks by a factor of about
\val{fp/x40/1024/ced-enc/fw/contract} per round, from \val{fp/x40/1024/ced-enc/fw/rho/r4}\% at round 4 to
\val{fp/x40/1024/ced-enc/fw/rho/r8}\% at the \val{cfg/rounds} rounds we deploy and
\val{fp/x40/1024/ced-enc/fw/rho/r32}\% at round \val{cfg/fp/rounds}: the core contracts to a fixed point. Between
rounds 7 and 8, layer 2's global KV changes by only \val{fp/x40/1024/ced-enc/fw/kvstep/r8}\%, which is why every
round can read the last round's. From round 8 to \val{cfg/fp/rounds}, BPB stays within
\val{fp/x40/1024/ced-enc/fw/bpbspread/late} of its 8-round value, so running more rounds than trained is safe. The
same holds after \val{cfg/tokens/long}B tokens ($\rho_8 = \val{fp/long/1024/ced-enc/fw/rho/r8}\%$), for the decoder
loop ($\val{fp/x40/1024/ced-dec/fw/rho/r8}\%$) and for the 45M CED ($\val{fp/p140/768/p140-enc/fw/rho/r8}\%$). With
a fixed number of rounds, $\rho_r$ never falls below \val{fp/x40/1024/ced-enc-fixed/fw/rhomin/late}\% after round
8, and layer 2's KV changes by \val{fp/x40/1024/ced-enc-fixed/fw/kvstep/r8}\% per round. The state does not
converge but cycles with period \val{fp/x40/1024/ced-enc-fixed/fw/period}: BPB matches its 8-round value whenever
$r$ is a multiple of \val{fp/x40/1024/ced-enc-fixed/fw/period} and is
\val{fp/x40/1024/ced-enc-fixed/fw/bpbbad/min}--\val{fp/x40/1024/ced-enc-fixed/fw/bpbbad/max} higher otherwise. The
45M fixed-round CED shows the same cycle. Every fixed-round model we have backpropagates through only the last
\val{cfg/bwd} of its \val{cfg/rounds} rounds, which likely causes the period-\val{fp/x40/1024/ced-enc-fixed/fw/period}
cycle; we do not claim it for fixed-depth training in general. These models thus learn an orbit that reaches a
good state at the trained depth, not a fixed point, and the last round's KV is then a poor stand-in for earlier rounds'.

\section{Hyperparameters}
\label{app:hparams}

Every 12-layer run uses the hyperparameters of Table~\ref{tab:hparams}; the three sizes differ only in width, heads,
steps and training tokens.

\begin{table}[tb]
\caption{\textbf{Hyperparameters.} Shared by every 12-layer run, read from the run configurations.}
\label{tab:hparams}
\centering
\footnotesize
\setlength{\tabcolsep}{3pt}
\begin{tabular}{@{}l>{\raggedright\arraybackslash}p{0.5\columnwidth}@{}}
\toprule
Layers & \val{cfg/layers} \\
Model size (non-embedding) & 90M / 150M / 340M \\
Hidden size $d$ & 768 / 1024 / 1536 \\
Heads (query = KV) & \val{cfg/heads/768} / \val{cfg/heads/1024} / \val{cfg/heads/1536}, dimension \val{cfg/headdim} \\
FFN & \val{cfg/ffnmult}$d$, ReLU$^2$ \\
Value embeddings & layers \val{cfg/ve} \\
SWA window & \val{cfg/window} \\
RoPE $\theta$ & \val{cfg/rope} \\
Vocabulary & \val{cfg/vocab} (Qwen3) \\
Sequence length & \val{cfg/seq} \\
Batch & \val{cfg/batch} sequences \\
Optimizer & Muon (matrices), AdamW (others) \\
Learning rate & \val{cfg/lr}, warmup \val{cfg/warmup}\%, linear decay over the last \val{cfg/decay}\% \\
Weight decay & \val{cfg/wd} (AdamW), \val{cfg/muonwd} (Muon) \\
Steps & \val{cfg/steps/x40/768} / \val{cfg/steps/x40/1024} / \val{cfg/steps/x40/1536} \\
Tokens & \val{cfg/tokens/x40/768}B / \val{cfg/tokens/x40/1024}B / \val{cfg/tokens/x40/1536}B; \val{cfg/tokens/long}B for 150M \\
Rounds & $\max(1, \mathrm{Poisson}(\val{cfg/rounds}))$ per sequence; \val{cfg/rounds} at evaluation \\
Backpropagated rounds & last \val{cfg/bwd} \\
Loop-state initialization & \val{cfg/stateinit} (truncated normal); decay gate $\sqrt{\val{cfg/decayinit}}$ \\
\bottomrule
\end{tabular}
\end{table}

\section{Test-time rounds}
\label{app:rounds}

Table~\ref{tab:rounds} evaluates the 150M looped models with $T=1$ to 16 rounds. Running 12 or 16 rounds instead of
the trained mean of \val{cfg/rounds} lowers BPB by at most 0.0004 and never raises it, so these models can run more
rounds than they were trained with (Appendix~\ref{app:fixedpoint}).

\begin{table}[tb]
\caption{\textbf{Test-time rounds.} BPB when evaluating with $T$ rounds; the models were trained with \val{cfg/rounds} rounds on average.
Parallel forward pass with per-round KV on the first \val{cfg/evalwin/x40} validation samples, as in
Table~\ref{tab:kvshare}.}
\label{tab:rounds}
\centering
\footnotesize
\setlength{\tabcolsep}{2pt}
\resizebox{\columnwidth}{!}{\begin{tabular}{@{}lcccccc@{}}
\toprule
Model & $T{=}1$ & 2 & 4 & 8 & 12 & 16 \\
\midrule
CED, encoder loop & \sval{tt/x40/1024/ced-enc/T1} & \sval{tt/x40/1024/ced-enc/T2} & \sval{tt/x40/1024/ced-enc/T4} & \sval{tt/x40/1024/ced-enc/T8} & \sval{tt/x40/1024/ced-enc/T12} & \sval{tt/x40/1024/ced-enc/T16} \\
CED, enc.\ loop, \val{cfg/tokens/long}B & \sval{tt/long/1024/ced-enc/T1} & \sval{tt/long/1024/ced-enc/T2} & \sval{tt/long/1024/ced-enc/T4} & \sval{tt/long/1024/ced-enc/T8} & \sval{tt/long/1024/ced-enc/T12} & \sval{tt/long/1024/ced-enc/T16} \\
CED, decoder loop & \sval{tt/x40/1024/ced-dec/T1} & \sval{tt/x40/1024/ced-dec/T2} & \sval{tt/x40/1024/ced-dec/T4} & \sval{tt/x40/1024/ced-dec/T8} & \sval{tt/x40/1024/ced-dec/T12} & \sval{tt/x40/1024/ced-dec/T16} \\
Dense, loop L2--5 & \sval{tt/x40/1024/dense-l25/T1} & \sval{tt/x40/1024/dense-l25/T2} & \sval{tt/x40/1024/dense-l25/T4} & \sval{tt/x40/1024/dense-l25/T8} & \sval{tt/x40/1024/dense-l25/T12} & \sval{tt/x40/1024/dense-l25/T16} \\
Dense, loop L4--7 & \sval{tt/x40/1024/dense-l47/T1} & \sval{tt/x40/1024/dense-l47/T2} & \sval{tt/x40/1024/dense-l47/T4} & \sval{tt/x40/1024/dense-l47/T8} & \sval{tt/x40/1024/dense-l47/T12} & \sval{tt/x40/1024/dense-l47/T16} \\
\bottomrule
\end{tabular}}
\end{table}

\section{Fewer rounds for prompt tokens}
\label{app:split}

The encoder loop's prefill cost scales with the number of rounds that prompt tokens run. The unified evaluation
splits each validation sequence after its first half. It runs the first half with $T_p$ rounds and caches its
final-round KV, then scores the second half with $T_d$ rounds. Table~\ref{tab:split} shows BPB on the second
half only, so it is not comparable to the other tables. A prefill at $T_p$ rounds runs the same layers as a $T=T_p$
model. On $4\times\text{GB200}$, the measured prefill of $T=4$ and $T=2$ models is $2.72\times$ and $1.54\times$ that of
unlooped \ced, against $5.12\times$ at $T=8$ (Table~\ref{tab:unit-costs-app}).

\begin{table}[tb]
\caption{\textbf{Fewer rounds for prompt tokens.} CED encoder loop after \val{cfg/tokens/long}B tokens: BPB on the second half of each sequence, with
the first half (the prompt) run at $T_p$ rounds and the second half at $T_d$ rounds.}
\label{tab:split}
\centering
\small
\begin{tabular}{lcccc}
\toprule
$T_p$ & $T_d{=}2$ & 4 & 8 & 16 \\
\midrule
1 & \sval{split/long/1024/ced-enc/tp1_td2} & \sval{split/long/1024/ced-enc/tp1_td4} & \sval{split/long/1024/ced-enc/tp1_td8} & \sval{split/long/1024/ced-enc/tp1_td16} \\
2 & \sval{split/long/1024/ced-enc/tp2_td2} & \sval{split/long/1024/ced-enc/tp2_td4} & \sval{split/long/1024/ced-enc/tp2_td8} & \sval{split/long/1024/ced-enc/tp2_td16} \\
4 & \sval{split/long/1024/ced-enc/tp4_td2} & \sval{split/long/1024/ced-enc/tp4_td4} & \sval{split/long/1024/ced-enc/tp4_td8} & \sval{split/long/1024/ced-enc/tp4_td16} \\
8 & \sval{split/long/1024/ced-enc/tp8_td2} & \sval{split/long/1024/ced-enc/tp8_td4} & \sval{split/long/1024/ced-enc/tp8_td8} & \sval{split/long/1024/ced-enc/tp8_td16} \\
16 & \sval{split/long/1024/ced-enc/tp16_td2} & \sval{split/long/1024/ced-enc/tp16_td4} & \sval{split/long/1024/ced-enc/tp16_td8} & \sval{split/long/1024/ced-enc/tp16_td16} \\
\bottomrule
\end{tabular}
\end{table}

\begin{table}[tb]
\caption{\textbf{How fast BPB gaps grow with context.} The least-squares slope of BPB(a)$-$BPB(b) over the four context
buckets beyond the SWA window (128--255, \dots, 1024--2047 tokens) against $\log_2$ of the context, in
$10^{-3}$ BPB per doubling. Positive means that a falls further behind b as the context grows. Long docs pools PG19
test, FinePDFs and GovReport, with a 95\% paired bootstrap interval; FW is the FineWeb-Edu validation set. Hidden
size 1024, \val{cfg/tokens/x40/1024}B tokens.}
\label{tab:slope}
\centering
\small
\setlength{\tabcolsep}{2pt}
\begin{tabular}{lccccc}
\toprule
a $-$ b & Long docs & PG19 & PDFs & Gov. & FW \\
\midrule
\multicolumn{6}{l}{\emph{Gap to dense, loop L2--5}} \\
CED enc.\ loop & \sval{slopem/x40/1024/ced-enc-minus-dense-l25/longdocs} {\scriptsize[\sval{slopem/x40/1024/ced-enc-minus-dense-l25/longdocs/lo}, \sval{slopem/x40/1024/ced-enc-minus-dense-l25/longdocs/hi}]} & \sval{slopem/x40/1024/ced-enc-minus-dense-l25/pg19-test} & \sval{slopem/x40/1024/ced-enc-minus-dense-l25/finepdfs-long} & \sval{slopem/x40/1024/ced-enc-minus-dense-l25/govreport-test} & \sval{slopem/x40/1024/ced-enc-minus-dense-l25/fineweb-edu-100bt} \\
YOCO-U loop & \sval{slopem/x40/1024/yocou-loop-minus-dense-l25/longdocs} {\scriptsize[\sval{slopem/x40/1024/yocou-loop-minus-dense-l25/longdocs/lo}, \sval{slopem/x40/1024/yocou-loop-minus-dense-l25/longdocs/hi}]} & \sval{slopem/x40/1024/yocou-loop-minus-dense-l25/pg19-test} & \sval{slopem/x40/1024/yocou-loop-minus-dense-l25/finepdfs-long} & \sval{slopem/x40/1024/yocou-loop-minus-dense-l25/govreport-test} & \sval{slopem/x40/1024/yocou-loop-minus-dense-l25/fineweb-edu-100bt} \\
\multicolumn{6}{l}{\emph{Gap to dense, no loop}} \\
CED & \sval{slopem/x40/1024/ced-noloop-minus-dense-noloop/longdocs} {\scriptsize[\sval{slopem/x40/1024/ced-noloop-minus-dense-noloop/longdocs/lo}, \sval{slopem/x40/1024/ced-noloop-minus-dense-noloop/longdocs/hi}]} & \sval{slopem/x40/1024/ced-noloop-minus-dense-noloop/pg19-test} & \sval{slopem/x40/1024/ced-noloop-minus-dense-noloop/finepdfs-long} & \sval{slopem/x40/1024/ced-noloop-minus-dense-noloop/govreport-test} & \sval{slopem/x40/1024/ced-noloop-minus-dense-noloop/fineweb-edu-100bt} \\
YOCO-U & \sval{slopem/x40/1024/yocou-noloop-minus-dense-noloop/longdocs} {\scriptsize[\sval{slopem/x40/1024/yocou-noloop-minus-dense-noloop/longdocs/lo}, \sval{slopem/x40/1024/yocou-noloop-minus-dense-noloop/longdocs/hi}]} & \sval{slopem/x40/1024/yocou-noloop-minus-dense-noloop/pg19-test} & \sval{slopem/x40/1024/yocou-noloop-minus-dense-noloop/finepdfs-long} & \sval{slopem/x40/1024/yocou-noloop-minus-dense-noloop/govreport-test} & \sval{slopem/x40/1024/yocou-noloop-minus-dense-noloop/fineweb-edu-100bt} \\
\midrule
\multicolumn{6}{l}{\emph{Loop gain: no loop $-$ loop}} \\
Dense L2--5 & \sval{slopem/x40/1024/dense-noloop-minus-dense-l25/longdocs} {\scriptsize[\sval{slopem/x40/1024/dense-noloop-minus-dense-l25/longdocs/lo}, \sval{slopem/x40/1024/dense-noloop-minus-dense-l25/longdocs/hi}]} & \sval{slopem/x40/1024/dense-noloop-minus-dense-l25/pg19-test} & \sval{slopem/x40/1024/dense-noloop-minus-dense-l25/finepdfs-long} & \sval{slopem/x40/1024/dense-noloop-minus-dense-l25/govreport-test} & \sval{slopem/x40/1024/dense-noloop-minus-dense-l25/fineweb-edu-100bt} \\
CED encoder & \sval{slopem/x40/1024/ced-noloop-minus-ced-enc/longdocs} {\scriptsize[\sval{slopem/x40/1024/ced-noloop-minus-ced-enc/longdocs/lo}, \sval{slopem/x40/1024/ced-noloop-minus-ced-enc/longdocs/hi}]} & \sval{slopem/x40/1024/ced-noloop-minus-ced-enc/pg19-test} & \sval{slopem/x40/1024/ced-noloop-minus-ced-enc/finepdfs-long} & \sval{slopem/x40/1024/ced-noloop-minus-ced-enc/govreport-test} & \sval{slopem/x40/1024/ced-noloop-minus-ced-enc/fineweb-edu-100bt} \\
CED decoder & \sval{slopem/x40/1024/ced-noloop-minus-ced-dec/longdocs} {\scriptsize[\sval{slopem/x40/1024/ced-noloop-minus-ced-dec/longdocs/lo}, \sval{slopem/x40/1024/ced-noloop-minus-ced-dec/longdocs/hi}]} & \sval{slopem/x40/1024/ced-noloop-minus-ced-dec/pg19-test} & \sval{slopem/x40/1024/ced-noloop-minus-ced-dec/finepdfs-long} & \sval{slopem/x40/1024/ced-noloop-minus-ced-dec/govreport-test} & \sval{slopem/x40/1024/ced-noloop-minus-ced-dec/fineweb-edu-100bt} \\
YOCO-U & \sval{slopem/x40/1024/yocou-noloop-minus-yocou-loop/longdocs} {\scriptsize[\sval{slopem/x40/1024/yocou-noloop-minus-yocou-loop/longdocs/lo}, \sval{slopem/x40/1024/yocou-noloop-minus-yocou-loop/longdocs/hi}]} & \sval{slopem/x40/1024/yocou-noloop-minus-yocou-loop/pg19-test} & \sval{slopem/x40/1024/yocou-noloop-minus-yocou-loop/finepdfs-long} & \sval{slopem/x40/1024/yocou-noloop-minus-yocou-loop/govreport-test} & \sval{slopem/x40/1024/yocou-noloop-minus-yocou-loop/fineweb-edu-100bt} \\
\bottomrule
\end{tabular}
\end{table}

\begin{table*}[t]
\caption{\textbf{Long-document BPB by context bucket.} PG19 test, FinePDFs and GovReport (512, 512 and 396 samples
of 2048 tokens, each averaged over six offsets in its document); 150M models. One forward pass per sample with per-round KV and \val{cfg/rounds} rounds for looped
models. Rows above the line trained on \val{cfg/tokens/x40/1024}B tokens.}
\label{tab:longctx}
\centering
\small
\setlength{\tabcolsep}{4pt}
\begin{tabular}{lcccccc}
\toprule
Model & 1--127 & 128--255 & 256--511 & 512--1023 & 1024--2047 & mean \\
\midrule
Dense, no loop & \sval{pos/x40/1024/dense-noloop/longdocs/c1-128} & \sval{pos/x40/1024/dense-noloop/longdocs/c128-256} & \sval{pos/x40/1024/dense-noloop/longdocs/c256-512} & \sval{pos/x40/1024/dense-noloop/longdocs/c512-1024} & \sval{pos/x40/1024/dense-noloop/longdocs/c1024-2048} & \sval{pos/x40/1024/dense-noloop/longdocs/mean} \\
Dense, loop L2--5 & \sval{pos/x40/1024/dense-l25/longdocs/c1-128} & \sval{pos/x40/1024/dense-l25/longdocs/c128-256} & \sval{pos/x40/1024/dense-l25/longdocs/c256-512} & \sval{pos/x40/1024/dense-l25/longdocs/c512-1024} & \sval{pos/x40/1024/dense-l25/longdocs/c1024-2048} & \sval{pos/x40/1024/dense-l25/longdocs/mean} \\
Dense, loop L4--7 & \sval{pos/x40/1024/dense-l47/longdocs/c1-128} & \sval{pos/x40/1024/dense-l47/longdocs/c128-256} & \sval{pos/x40/1024/dense-l47/longdocs/c256-512} & \sval{pos/x40/1024/dense-l47/longdocs/c512-1024} & \sval{pos/x40/1024/dense-l47/longdocs/c1024-2048} & \sval{pos/x40/1024/dense-l47/longdocs/mean} \\
CED, no loop & \sval{pos/x40/1024/ced-noloop/longdocs/c1-128} & \sval{pos/x40/1024/ced-noloop/longdocs/c128-256} & \sval{pos/x40/1024/ced-noloop/longdocs/c256-512} & \sval{pos/x40/1024/ced-noloop/longdocs/c512-1024} & \sval{pos/x40/1024/ced-noloop/longdocs/c1024-2048} & \sval{pos/x40/1024/ced-noloop/longdocs/mean} \\
CED, encoder loop & \sval{pos/x40/1024/ced-enc/longdocs/c1-128} & \sval{pos/x40/1024/ced-enc/longdocs/c128-256} & \sval{pos/x40/1024/ced-enc/longdocs/c256-512} & \sval{pos/x40/1024/ced-enc/longdocs/c512-1024} & \sval{pos/x40/1024/ced-enc/longdocs/c1024-2048} & \sval{pos/x40/1024/ced-enc/longdocs/mean} \\
CED, decoder loop & \sval{pos/x40/1024/ced-dec/longdocs/c1-128} & \sval{pos/x40/1024/ced-dec/longdocs/c128-256} & \sval{pos/x40/1024/ced-dec/longdocs/c256-512} & \sval{pos/x40/1024/ced-dec/longdocs/c512-1024} & \sval{pos/x40/1024/ced-dec/longdocs/c1024-2048} & \sval{pos/x40/1024/ced-dec/longdocs/mean} \\
CED, enc.\ loop, fixed rounds & \sval{pos/x40/1024/ced-enc-fixed/longdocs/c1-128} & \sval{pos/x40/1024/ced-enc-fixed/longdocs/c128-256} & \sval{pos/x40/1024/ced-enc-fixed/longdocs/c256-512} & \sval{pos/x40/1024/ced-enc-fixed/longdocs/c512-1024} & \sval{pos/x40/1024/ced-enc-fixed/longdocs/c1024-2048} & \sval{pos/x40/1024/ced-enc-fixed/longdocs/mean} \\
YOCO-U style, no loop & \sval{pos/x40/1024/yocou-noloop/longdocs/c1-128} & \sval{pos/x40/1024/yocou-noloop/longdocs/c128-256} & \sval{pos/x40/1024/yocou-noloop/longdocs/c256-512} & \sval{pos/x40/1024/yocou-noloop/longdocs/c512-1024} & \sval{pos/x40/1024/yocou-noloop/longdocs/c1024-2048} & \sval{pos/x40/1024/yocou-noloop/longdocs/mean} \\
YOCO-U style, loop L2--5 & \sval{pos/x40/1024/yocou-loop/longdocs/c1-128} & \sval{pos/x40/1024/yocou-loop/longdocs/c128-256} & \sval{pos/x40/1024/yocou-loop/longdocs/c256-512} & \sval{pos/x40/1024/yocou-loop/longdocs/c512-1024} & \sval{pos/x40/1024/yocou-loop/longdocs/c1024-2048} & \sval{pos/x40/1024/yocou-loop/longdocs/mean} \\
\midrule
\multicolumn{7}{l}{\emph{\val{cfg/tokens/long}B training tokens}} \\
CED, no loop & \sval{pos/long/1024/ced-noloop/longdocs/c1-128} & \sval{pos/long/1024/ced-noloop/longdocs/c128-256} & \sval{pos/long/1024/ced-noloop/longdocs/c256-512} & \sval{pos/long/1024/ced-noloop/longdocs/c512-1024} & \sval{pos/long/1024/ced-noloop/longdocs/c1024-2048} & \sval{pos/long/1024/ced-noloop/longdocs/mean} \\
CED, encoder loop & \sval{pos/long/1024/ced-enc/longdocs/c1-128} & \sval{pos/long/1024/ced-enc/longdocs/c128-256} & \sval{pos/long/1024/ced-enc/longdocs/c256-512} & \sval{pos/long/1024/ced-enc/longdocs/c512-1024} & \sval{pos/long/1024/ced-enc/longdocs/c1024-2048} & \sval{pos/long/1024/ced-enc/longdocs/mean} \\
CED, decoder loop & \sval{pos/long/1024/ced-dec/longdocs/c1-128} & \sval{pos/long/1024/ced-dec/longdocs/c128-256} & \sval{pos/long/1024/ced-dec/longdocs/c256-512} & \sval{pos/long/1024/ced-dec/longdocs/c512-1024} & \sval{pos/long/1024/ced-dec/longdocs/c1024-2048} & \sval{pos/long/1024/ced-dec/longdocs/mean} \\
Dense, loop L4--7 & \sval{pos/long/1024/dense-l47/longdocs/c1-128} & \sval{pos/long/1024/dense-l47/longdocs/c128-256} & \sval{pos/long/1024/dense-l47/longdocs/c256-512} & \sval{pos/long/1024/dense-l47/longdocs/c512-1024} & \sval{pos/long/1024/dense-l47/longdocs/c1024-2048} & \sval{pos/long/1024/dense-l47/longdocs/mean} \\
\bottomrule
\end{tabular}
\end{table*}

\begin{figure*}[t]
\centering
\safegraphic[width=\textwidth]{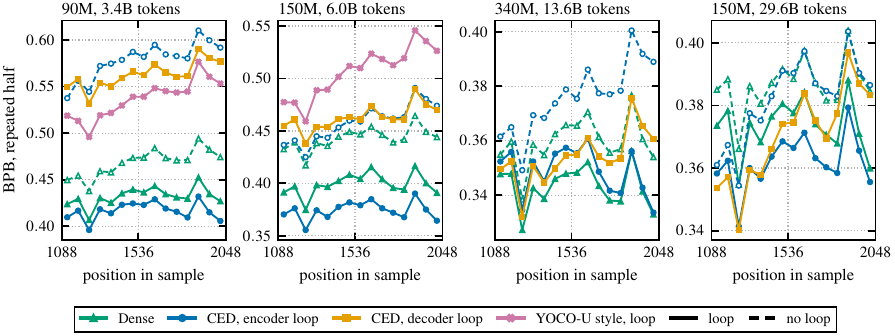}
\caption{\textbf{In-context recall.} BPB on the second half of samples whose tokens 1,024--2,047 repeat tokens
0--1,023 verbatim (PG19 test and arXiv, \val{cfg/recall/windows} samples), per 64-token bin; the first copied bin is
left out. Each panel compares the models of one size and token budget. Solid: looped (L2--5; decoder loop L7--10;
dense L4--7 in the last panel); dashed: unlooped.}
\label{fig:recall}
\end{figure*}

\section{Long-context evaluation}
\label{app:longctx}

\paragraph{Long documents.} PG19 test and FinePDFs contribute 512 samples and GovReport 396, each 2048 contiguous
tokens of one document starting at least 1,024 tokens in; every model scores the same samples. To keep the
difficulty of particular passages out of the position-wise curves, every sample is also scored after moving it
256, 512, 768 and 1,024 tokens earlier in its document and by one random offset of up to 1,024 tokens, and each
context bin averages the six copies, so the text at a given position comes from six places in the document. One forward pass scores each sample,
with \val{cfg/rounds} rounds and per-round KV for looped models. ``Long docs'' pools the three corpora.
Differences between models carry 95\% intervals from a paired bootstrap over samples (\val{cfg/bootreps}
resamples, stratified by corpus). For FineWeb-Edu we take each token's position in its own document as its
context.

\paragraph{BPB by context length.} Table~\ref{tab:longctx} gives the BPB behind Figure~\ref{fig:longctx} for every model, pooled over the three
long-document corpora, by the number of tokens of context. Figure~\ref{fig:longctx-corpora} repeats
Figure~\ref{fig:longctx}b--c for each corpus, and Table~\ref{tab:slope} gives how fast each gap grows with context.

\paragraph{In-context recall.}
\label{app:recall}
To test recall directly, we take the first 1,024 tokens of each PG19 test sample and of 512 arXiv documents from the Proof-Pile-2 test
split~\citep{llemma}, and repeat them verbatim at positions 1,024--2,047. Each copied token lies 1,024 tokens after its source, far beyond the SWA window, so only global KV can retrieve it.
Figure~\ref{fig:recall} shows BPB on the repeated half. Unlooped CED recalls
worse than unlooped dense (\val{recalldiff/x40/768/ced-noloop-minus-dense-noloop} and
\val{recalldiff/x40/1024/ced-noloop-minus-dense-noloop}); the encoder loop removes this deficit
(\val{recalldiff/x40/768/ced-enc-minus-ced-noloop} and \val{recalldiff/x40/1024/ced-enc-minus-ced-noloop} against
unlooped CED), whereas the decoder loop barely changes recall (\val{recalldiff/x40/768/ced-dec-minus-ced-noloop} and
\val{recalldiff/x40/1024/ced-dec-minus-ced-noloop}). The YOCO-U style loop, whose only global KV comes from layer 6,
recalls worst of the looped models (\val{recalldiff/x40/768/yocou-loop-minus-dense-l25} and
\val{recalldiff/x40/1024/yocou-loop-minus-dense-l25} against the dense loop). The differences shrink with scale: at
340M, all five models lie between \val{recall/x40/1536/dense-l25/bpb} and \val{recall/x40/1536/ced-noloop/bpb} BPB.
The CED encoder loop recalls better than the dense loop at 90M and 150M
(\val{recalldiff/x40/768/ced-enc-minus-dense-l25} and \val{recalldiff/x40/1024/ced-enc-minus-dense-l25} BPB), trails
it by \val{recalldiff/x40/1536/ced-enc-minus-dense-l25} [\val{recalldiff/x40/1536/ced-enc-minus-dense-l25/lo},
\val{recalldiff/x40/1536/ced-enc-minus-dense-l25/hi}] at 340M, and recalls better than the L4--7 dense loop of
Table~\ref{tab:long} after \val{cfg/tokens/long}B tokens (\val{recalldiff/long/1024/ced-enc-minus-dense-l47}). Across
the three sizes and both token budgets, it thus recalls within 0.03 BPB of the dense loop while caching one-sixth of its
global KV.

\section{Production Serving Setup and Detailed Results}
\label{app:serving}

\begin{table*}[t]
\centering
\caption{\textbf{Unit prefill and decode cost multipliers relative to unlooped \ced ($T=1$) across $4\times\text{GB200}$ (TP4) and $8\times\text{H200}$ (TP8).}
The analytic columns are a roofline model of FLOPs and bytes at GB200 peak rates, with expert parallelism over 4 GPUs and all of $L_{20}$ run on every prompt token, as in SGLang.
Prefill $a$ is the per-token slope of $t_{\text{pre}}(L) = b + aL + cL^2$, fitted over uncached prompts of 512--65{,}536 tokens at batch 1; for \ced $T=1$ it is $18.8\,\mu$s per prompt token per GB200 tray and $60.8\,\mu$s per H200 node. Decode is the per-step time averaged over 8K and 32K contexts at batches 16--256; for \ced $T=1$ the step at 8K context and batch 32 takes 8.6\,ms. GB200 120K uses distinct 120K-token contexts (per-layer KV, $T=8$: batches 1 and 8 only). Unit costs do not depend on the round policy, so each row holds for last-round and per-round KV alike.}
\label{tab:unit-costs-app}
\vskip 0.06in
\footnotesize
\setlength{\tabcolsep}{2.4pt}
\begin{tabular}{@{}lccccccc@{}}
\toprule
& \multicolumn{3}{c}{\textbf{Analytic Roofline Model}} & \multicolumn{3}{c}{\textbf{Measured $4\times\text{GB200}$ (TP4)}} & \textbf{Measured $8\times\text{H200}$ (TP8)} \\
\cmidrule(lr){2-4} \cmidrule(lr){5-7} \cmidrule(lr){8-8}
\textbf{Model Variant} & Prefill $a$ & Decode $8\text{--}32\text{K}$ & KV Pool & Prefill $a$ & Decode $8\text{--}32\text{K}$ & Decode $120\text{K}$ & Prefill $a$ \\
\midrule
\textbf{\ced, no loop (reference)} & $1.00\times$ & $1.00\times$ & $1.00\times$ & $1.00\times$ & $1.00\times$ & $1.00\times$ & $1.00\times$ \\
\textbf{Stock SGLang (no plugin)} & $1.00\times$ & $1.00\times$ & $1.00\times$ & $1.00\times$ & $1.02\times$ & --- & $0.98\times$ \\
\textbf{\ourmethod, $T=2$} & $1.87\times$ & $1.45\times$ & $1.00\times$ & $1.54\times$ & $1.39\times$ & --- & $1.61\times$ \\
\textbf{\ourmethod, $T=4$} & $3.60\times$ & $2.34\times$ & $1.00\times$ & $2.72\times$ & $2.13\times$ & --- & $2.87\times$ \\
\textbf{\ourmethod, $T=8$} & $7.06\times$ & $4.12\times$ & $\mathbf{1.00\times}$ & $\mathbf{5.12\times}$ & $\mathbf{3.95\times}$ & $\mathbf{3.83\times}$ & $\mathbf{5.42\times}$ \\
\textbf{\ced decoder loop, $T=8$} & $1.00\times^\ddagger$ & $4.13\times$ & $\mathbf{1.00\times}$ & $2.76\times^\ddagger$ & $\mathbf{3.99\times}$ & $\mathbf{3.89\times}$ & $1.80\times^\ddagger$ \\
\midrule
\textbf{Per-layer KV, no loop} & $1.91\times$ & $1.12\times$ & $0.086\times$ & --- & --- & --- & $1.73\times$ \\
\textbf{Per-layer KV, $T=8$} & $8.21\times$ & $4.23\times$ & $0.086\times$ & $\mathbf{6.97\times}$ & $\mathbf{4.24\times}$ & $\mathbf{4.56\times}$ & $\mathbf{6.88\times}$ \\
\bottomrule
\multicolumn{8}{@{}p{0.97\textwidth}@{}}{\scriptsize $^\ddagger$Analytically, the decoder loop runs its extra rounds only on the 128-token tail ($a\times 1.00$, $b\times 7.64$). Measured, its prefill slope is $2.76\times$ (GB200) and $1.80\times$ (H200), and on GB200 its per-request prefill time is $2.6\text{--}2.9\times$ that of \ced $T=1$ at every prompt length from 512 tokens (a single chunk) to 64K. The architecture does not require this extra cost; it comes from our implementation of the decoder-loop prefill.} \\
\end{tabular}
\end{table*}

\begin{table*}[t]
\centering
\caption{\textbf{Pure decode throughput saturation (last-round vs.\ per-round KV) for the \ced encoder loop} on $4\times\text{GB200}$ (real weights; each request has its own context, whose KV is allocated without being computed, and generates 256 tokens). Each cell is the ratio of the two policies' highest decode throughput; in parentheses, the batch at which each was measured. Last-round KV at 8K is capped at 768 requests (pool 80\% full), so the 8K ratios are lower bounds.}
\label{tab:hbm-decode-app}
\vskip 0.06in
\footnotesize
\setlength{\tabcolsep}{5pt}
\begin{tabular}{@{}lccccc@{}}
\toprule
\textbf{Rounds ($T$)} & \makecell{\textbf{Per-round}\\\textbf{KV factor}} & \makecell{\textbf{Pool (last-round}\\\textbf{/ per-round)}} & \textbf{ISL = 8K} & \textbf{ISL = 16K} & \textbf{ISL = 32K} \\
\midrule
\textbf{$T=2$} & $1.60\times$ & $8.11\text{M} \;/\; 5.07\text{M}$ & $\mathbf{\ge 1.12\times}$ ($768$ vs.\ $570$) & $\mathbf{1.17\times}$ ($487$ vs.\ $288$) & $\mathbf{1.34\times}$ ($245$ vs.\ $144$) \\
\textbf{$T=4$} & $2.80\times$ & $8.11\text{M} \;/\; 2.90\text{M}$ & $\mathbf{\ge 1.42\times}$ ($768$ vs.\ $342$) & $\mathbf{1.58\times}$ ($487$ vs.\ $174$) & $\mathbf{1.80\times}$ ($245$ vs.\ $87$) \\
\textbf{$T=8$} & $5.20\times$ & $8.11\text{M} \;/\; 1.56\text{M}$ & $\mathbf{\ge 2.05\times}$ ($768$ vs.\ $184$) & $\mathbf{2.35\times}$ ($487$ vs.\ $93$) & $\mathbf{2.83\times}$ ($245$ vs.\ $47$) \\
\bottomrule
\end{tabular}
\end{table*}

\paragraph{DeepSeek-V4.1-Flash architecture and SGLang plugin.}
DeepSeek-V4.1-Flash~\citep{deepseekv41} is a 40-layer Causal Encoder--Decoder MoE model (hidden dimension $5{,}120$, 384 routed experts per MoE layer with top-6 routing, FP8 attention/dense projections, and MXFP4 expert weights) partitioned into a 21-layer encoder (\Lr{0}{20}) and a 19-layer decoder (\Lr{21}{39}).
Every layer maintains a 128-token sliding-window attention (SWA) ring buffer ($768$ reserved slots per active request in SGLang, totaling $17.94$\,MB/request).
Long-range context is served by only four global KV-source layers ($L_2, L_8, L_{14}, L_{20}$) with compression ratios $(2, 2, 2, 1)$: the three sources inside the looped range \Lr{2}{19} store $3 \times 1/2 = 1.5$ compressed entries per token, and $L_{20}$, whose KV the whole decoder reads, stores one ($2.5$ compressed entries/token total).
In SGLang (\texttt{da64c5cbb}), each compressed entry stores $584$\,B of MLA latent KV (\texttt{fp8\_e4m3} nope + BF16 RoPE + scale) plus $68$\,B of FP4 lightning-indexer keys ($652$\,B/entry), giving $2.5 \times 652 = 1{,}630$\,B/token for \ced vs.\ $29.0 \times 652 = 18{,}908$\,B/token for the per-layer-KV baseline, in which all 38 layers \Lr{2}{39} own distinct global KV caches ($11.60\times$ larger; the analytical FP4-latent cost model uses $288 + 68 = 356$\,B/entry, or $890$\,B vs.\ $10{,}324$\,B/token, with identical ratios).
Under per-round KV, looping \Lr{2}{19} for $T$ rounds multiplies global KV by $1 + \frac{1.5}{2.5}(T-1) = 1 + 0.6(T-1)$ for \ced ($5.20\times$ at $T=8$, or $8{,}476$\,B/tok) and by $1 + \frac{9}{29}(T-1)$ with per-layer KV ($3.17\times$ at $T=8$, or $59{,}984$\,B/tok).

Our weight-tied plugin (\texttt{sglang\_ced\_loop}) wraps SGLang's layer execution loop into a virtual schedule of $(l, r)$ passes while keeping a single copy of physical weights.
For loops whose first layer reuses an earlier source's sparse top-$k$ indices (the decoder loop \Lr{21}{38} reusing $L_{20}$), the plugin snapshots the top-$k$ buffer on round 1 and restores it on rounds $2\dots T$; for $L_{14}$'s Engram table, the embedding lookup and key-value projection execute once on round 1 while the hidden-state gate runs every round.
At $T=1$, the plugin's unit prefill and decode costs are within 2\% of stock SGLang on both $8\times\text{H200}$ (TP8/EP8) and $4\times\text{GB200}$ (TP4/EP4) (Table~\ref{tab:unit-costs-app}), with zero layer-count or NLL discrepancies.
Table~\ref{tab:unit-costs-app} reports the complete unit prefill and decode cost multipliers, and Table~\ref{tab:hbm-decode-app} details the HBM-saturated pure decode sweep across context lengths and rounds.

\paragraph{Setup of Figure~\ref{fig:p1-pareto}.}
\label{app:serving-fig}
All runs use one $4\times\text{GB200}$ tray (TP4/EP4) and one SGLang server.
Within each panel, all models share the server settings, apart from the pool cap that emulates per-round KV. The settings differ between experiments, and so does the KV pool of \ourmethod: $5.11$\,M tokens in panels (a) and (b) and in Table~\ref{tab:cost}, $7.9$\,M in panel (c), and $8.11$\,M in Table~\ref{tab:hbm-decode-app}.
Panels (a) and (b) use the \vfour-Flash architecture with dummy weights, which change the ratios between models by 2--4\% (Appendix~\ref{app:serving-more}); panel (c) uses the real checkpoint weights.
Per-round KV is emulated with identical compute: the KV pool is capped by the per-round byte factor ($5.2\times$ for \ced at $T=8$, $3.17\times$ with per-layer KV; Appendix~\ref{app:serving}).
The baseline with per-layer KV gives every layer \Lr{2}{39} its own global KV and runs every layer on every prompt token in prefill (full replay).
A point is \emph{preemption-free} when no request is retracted; panel (a) draws only such points, and the dotted line carries each baseline's last preemption-free throughput to $c=512$.
\emph{(a)}~8K/1K requests are random token ids with \texttt{ignore\_eos} and $\max(20, 5c)$ prompts per point (one or two repeats).
The baseline (per-layer and per-round KV) fills its pool at $c=16$ ($T=4$) and $c=8$ ($T=8$), \ced with per-round KV at $c=128$ and $c=64$; last-round KV stays preemption-free up to $c=512$.
At $c=512$, last-round KV gives $1.27\times$ ($T=4$) and $1.30\times$ ($T=8$) the peak throughput of per-round KV, and $3.1\times$ and $3.6\times$ that of the baseline.
\emph{(b)}~The 128K groups are pure decode: each request has its own 131K-token context and generates 256 tokens, at the largest batch the KV pool holds ($38$ requests with last-round KV; $13$ and $7$ with per-round KV at $T=4$ and $8$; $1$ for the baseline at $T=4$ and none at $T=8$).
There, last-round KV gives $1.9\times$ ($T=4$) and $2.9\times$ ($T=8$) the throughput of per-round KV.
The ShareGPT group compares \ourmethod with the baseline only.
\emph{(c)}~Mooncake \texttt{toolagent} trace~\citep{mooncake}, closed loop with $c$ concurrent sessions, encoder loop at $T=8$.
With per-round KV the prefix cache holds $1.5$\,M tokens and is full from $c=128$, so evicted session histories are recomputed through all eight rounds; last-round KV holds $7.9$\,M tokens and peaks at $77\%$ use at $c=512$.
Median TTFT drops $13.6\times$ at $c=256$ and $29.7\times$ at $c=512$, while output throughput is within $13\%$ between the two policies.

\end{document}